\documentclass{book}%
\usepackage[T1]{fontenc}%
\usepackage[utf8]{inputenc}%
\usepackage{lmodern}%
\usepackage{textcomp}%
\usepackage{lastpage}%
\usepackage{geometry}%
\usepackage[hidelinks]{hyperref}%
\usepackage{listings}%
\usepackage{makeidx}%
\usepackage{amsmath}%
\usepackage{enumitem}%
\usepackage[breakable]{tcolorbox}%
\usepackage{ragged2e}%
\usepackage{graphicx}%
\title{Applications of Deep Neural Networks with Keras}%
\author{Jeff Heaton}%
\date{Fall 2022.0}%
\makeindex%
\begin{document}%
\normalsize%
\frontmatter%
\maketitle%
\begin{flushleft}%
Publisher: Heaton Research, Inc.%
\par%
Applications of Deep Neural Networks%
\index{neural network}%
\par%
May, 2022%
\par%
Author: {[}Jeffrey Heaton{]}(https://orcid.org/0000{-}0003{-}1496{-}4049%
\par%
ISBN: 9798416344269%
\par%
Edition: 1%
\par%
\par%
\end{flushleft}%
\vspace{2mm}%
\par%
The text and illustrations of Applications of Deep Neural Networks by Jeff Heaton are licensed under CC BY{-}NC{-}SA 4.0. To view a copy of this license, visit%
\index{neural network}%
\href{https://creativecommons.org/licenses/by-nc-sa/4.0}{ CC BY{-}NC{-}SA 4.0}%
.\newline%
All of the book's source code is licensed under the GNU Lesser General Public License as published by the Free Software Foundation; either version 2.1 of the license or (at your option) any later version.%
\href{https://www.gnu.org/licenses/lgpl-3.0.en.html}{ LGPL}%
\par%
\includegraphics[width=1in]{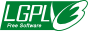}%
\includegraphics[width=1in]{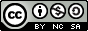}%
\par%
Heaton Research, Encog, the Encog Logo, and the Heaton Research logo are all trademarks of Jeff Heaton in the United States and/or other countries.%
\par%
TRADEMARKS: Heaton Research has attempted throughout this book to distinguish proprietary trademarks from descriptive terms by following the capitalization style used by the manufacturer.%
\par%
The author and publisher have done their best to prepare this book, so the content is based upon the final release of software whenever possible. Portions of the manuscript may be based upon pre{-}release versions supplied by software manufacturer(s). The author and the publisher make no representation or warranties of any kind about the completeness or accuracy of the contents herein and accept no liability of any kind, including but not limited to performance, merchantability, fitness for any particular purpose, or any losses or damages of any kind caused or alleged to be caused directly or indirectly from this book.%
\par%
\textbf{DISCLAIMER}%
\par%
The author, Jeffrey Heaton, makes no warranty or representation, either expressed or implied, concerning the Software or its contents, quality, performance, merchantability, or fitness for a particular purpose. In no event will Jeffrey Heaton, his distributors, or dealers be liable to you or any other party for direct, indirect, special, incidental, consequential, or other damages arising out of the use of or inability to use the Software or its contents even if advised of the possibility of such damage. In the event that the Software includes an online update feature, Heaton Research, Inc. further disclaims any obligation to provide this feature for any specific duration other than the initial posting.%
\index{feature}%
\par%
The exclusion of implied warranties is not permitted by some states. Therefore, the above exclusion may not apply to you. This warranty provides you with specific legal rights; there may be other rights that you may have that vary from state to state. The pricing of the book with the Software by Heaton Research, Inc. reflects the allocation of risk and limitations on liability contained in this agreement of Terms and Conditions.%
\index{SOM}%
\par%
\par%
\tableofcontents%
\newpage%
\chapter{Introduction}%
\label{chap:Introduction}%
Starting in the spring semester of 2016, I began teaching the T81{-}558 Applications of Deep Learning course for Washington University in St. Louis. I never liked Microsoft Powerpoint for technical classes, so I placed my course material, examples, and assignments on GitHub. This material started with code and grew to include enough description that this information evolved into the book you see before you.%
\index{GAN}%
\index{GitHub}%
\index{learning}%
\par%
I license the book's text under the Attribution{-}NonCommercial{-}ShareAlike 4.0 International (CC BY{-}NC{-}SA 4.0) license. Similarly, I offer the book's code under the LGPL license. Though I provide this book both as a relatively inexpensive paperback and Amazon Kindle, you can obtain the book's PDF here:%
\par%
\begin{itemize}[noitemsep]%
\item%
\href{https://arxiv.org/abs/2009.05673}{https://arxiv.org/abs/2009.05673}%
\end{itemize}%
The book's code is available at the following GitHub repository:%
\index{GitHub}%
\par%
\begin{itemize}[noitemsep]%
\item%
\href{https://github.com/jeffheaton/t81_558_deep_learning}{https://github.com/jeffheaton/t81\_558\_deep\_learning}%
\end{itemize}%
If you purchased this book from me, you have my sincere thanks for supporting my ongoing projects. I sell the book as a relatively low{-}cost paperback and Kindle ebook for those who prefer that format or wish to support my projects. I suggest that you look at the above GitHub site, as all of the code for this book is presented there as Jupyter notebooks that are entirely Google CoLab compatible.%
\index{GitHub}%
\par%
This book focuses on the application of deep neural networks. There is some theory; however, I do not focus on recreating neural network fundamentals that tech companies already provide in popular frameworks. The book begins with a quick review of the Python fundamentals needed to learn the subsequent chapters. With Python preliminaries covered, we start with classification and regression neural networks in Keras.%
\index{classification}%
\index{Keras}%
\index{neural network}%
\index{Python}%
\index{regression}%
\index{SOM}%
\par%
In my opinion, PyTorch, Jax, and Keras are the top three deep learning frameworks. When I first created this course, neither PyTorch nor JAX existed. I began the course based on TensorFlow and migrated to Keras the following semester. I believe TensorFlow remains a good choice for a course focusing on the application of deep learning. Some of the third{-}party libraries used for this course use PyTorch; as a result, you will see a blend of both technologies.  StyleGAN and TabGAN both make use of PyTorch.%
\index{GAN}%
\index{Keras}%
\index{learning}%
\index{PyTorch}%
\index{SOM}%
\index{StyleGAN}%
\index{TensorFlow}%
\par%
The technologies that this course is based on change rapidly. I update the Kindle and paperback books according to this schedule. Formal updates to this book typically occur just before each academic year's fall and spring semesters.%
\par%
The source document for this book is Jupyter notebooks. I wrote a Python utility that transforms my course Jupyter notebooks into this book. It is entirely custom, and I may release it as a project someday. However, because this book is based on code and updated twice a year, you may find the occasional typo. I try to minimize errors as much as possible, but please let me know if you see something. I use%
\index{error}%
\index{Python}%
\index{SOM}%
\href{https://www.grammarly.com/}{ Grammarly }%
to find textual issues, but due to the frequently updated nature of this book, I do not run it through a formal editing cycle for each release. I also double{-}check the code with each release to ensure CoLab, Keras, or another third{-}party library did not make a breaking change.%
\index{Keras}%
\par%
The book and course continue to be a work in progress. Many have contributed code, suggestions, fixes, and clarifications to the GitHub repository. Please submit a GitHub issue or a push request with a solution if you find an error.%
\index{error}%
\index{GitHub}%
\par%
\par

\mainmatter%
\chapter{Python Preliminaries}%
\label{chap:PythonPreliminaries}%
\section{Part 1.1: Overview}%
\label{sec:Part1.1Overview}%
Deep learning is a group of exciting new technologies for neural networks.%
\index{learning}%
\index{neural network}%
\cite{lecun2015deep}%
By using a combination of advanced training techniques neural network architectural components, it is now possible to train neural networks of much greater complexity. This book introduces the reader to deep neural networks, regularization units (ReLU), convolution neural networks, and recurrent neural networks. High{-}performance computing (HPC) aspects demonstrate how deep learning can be leveraged both on graphical processing units (GPUs), as well as grids. Deep learning allows a model to learn hierarchies of information in a way that is similar to the function of the human brain. The focus is primarily upon the application of deep learning, with some introduction to the mathematical foundations of deep learning. Readers will make use of the Python programming language to architect a deep learning model for several real{-}world data sets and interpret the results of these networks.%
\index{convolution}%
\index{GPU}%
\index{GPU}%
\index{learning}%
\index{model}%
\index{neural network}%
\index{Python}%
\index{recurrent}%
\index{regularization}%
\index{ReLU}%
\index{ROC}%
\index{ROC}%
\index{SOM}%
\index{training}%
\cite{goodfellow2016deep}%
\par%
\subsection{Origins of Deep Learning}%
\begin{figure}[h]%
\centering%
\includegraphics[width=4in]{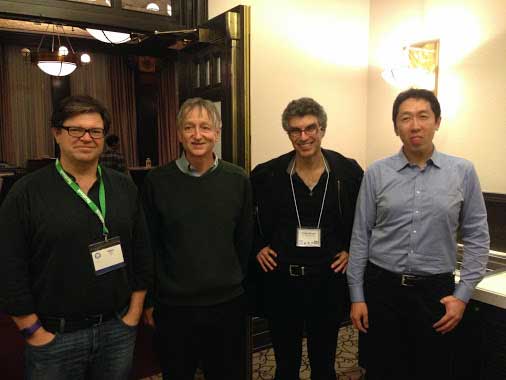}%
\caption{Neural Network Luminaries}%
\label{1.LUM}%
\end{figure}

\label{subsec:OriginsofDeepLearning}%
Neural networks are one of the earliest examples of a machine learning model.  Neural networks were initially introduced in the 1940s and have risen and fallen several times in popularity. The current generation of deep learning begain in 2006 with an improved training algorithm by Geoffrey Hinton.%
\index{algorithm}%
\index{Hinton}%
\index{learning}%
\index{model}%
\index{neural network}%
\index{training}%
\index{training algorithm}%
\cite{hinton2006fast}%
This technique finally allowed neural networks with many layers (deep neural networks) to be efficiently trained. Four researchers have contributed significantly to the development of neural networks.  They have consistently pushed neural network research, both through the ups and downs.  These four luminaries are shown in Figure \ref{1.LUM}.%
\index{layer}%
\index{neural network}%
\par%
\par%

The current luminaries of artificial neural network (ANN) research and ultimately deep learning, in order as appearing in the figure:%
\index{learning}%
\index{neural network}%
\par%
\begin{itemize}[noitemsep]%
\item%
\href{http://yann.lecun.com/}{Yann LeCun}%
, Facebook and New York University {-} Optical character recognition and computer vision using convolutional neural networks (CNN).  The founding father of convolutional nets.%
\index{CNN}%
\index{computer vision}%
\index{convolution}%
\index{convolutional}%
\index{Convolutional Neural Networks}%
\index{neural network}%
\item%
\href{http://www.cs.toronto.edu/~hinton/}{Geoffrey Hinton}%
, Google and University of Toronto. Extensive work on neural networks. Creator of deep learning and early adapter/creator of backpropagation for neural networks.%
\index{backpropagation}%
\index{learning}%
\index{neural network}%
\item%
\href{http://www.iro.umontreal.ca/~bengioy/yoshua_en/index.html}{Yoshua Bengio}%
, University of Montreal and Botler AI. Extensive research into deep learning, neural networks, and machine learning.%
\index{learning}%
\index{neural network}%
\item%
\href{http://www.andrewng.org/}{Andrew Ng}%
, Badiu and Stanford University.  Extensive research into deep learning, neural networks, and application to robotics.%
\index{learning}%
\index{neural network}%
\end{itemize}%
Geoffrey Hinton, Yann LeCun, and Yoshua Bengio won the%
\index{Hinton}%
\index{LeCun}%
\href{https://www.acm.org/media-center/2019/march/turing-award-2018}{ Turing Award }%
for their contributions to deep learning.%
\index{learning}%
\par

\subsection{What is Deep Learning}%
\label{subsec:WhatisDeepLearning}%
The focus of this book is deep learning, which is a prevalent type of machine learning that builds upon the original neural networks popularized in the 1980s. There is very little difference between how a deep neural network is calculated compared with the first neural network.  We've always been able to create and calculate deep neural networks.  A deep neural network is nothing more than a neural network with many layers.  While we've always been able to create/calculate deep neural networks, we've lacked an effective means of training them.  Deep learning provides an efficient means to train deep neural networks.%
\index{calculated}%
\index{layer}%
\index{learning}%
\index{neural network}%
\index{training}%
\par
\begin{figure}[h]%
\centering%
\includegraphics[width=4in]{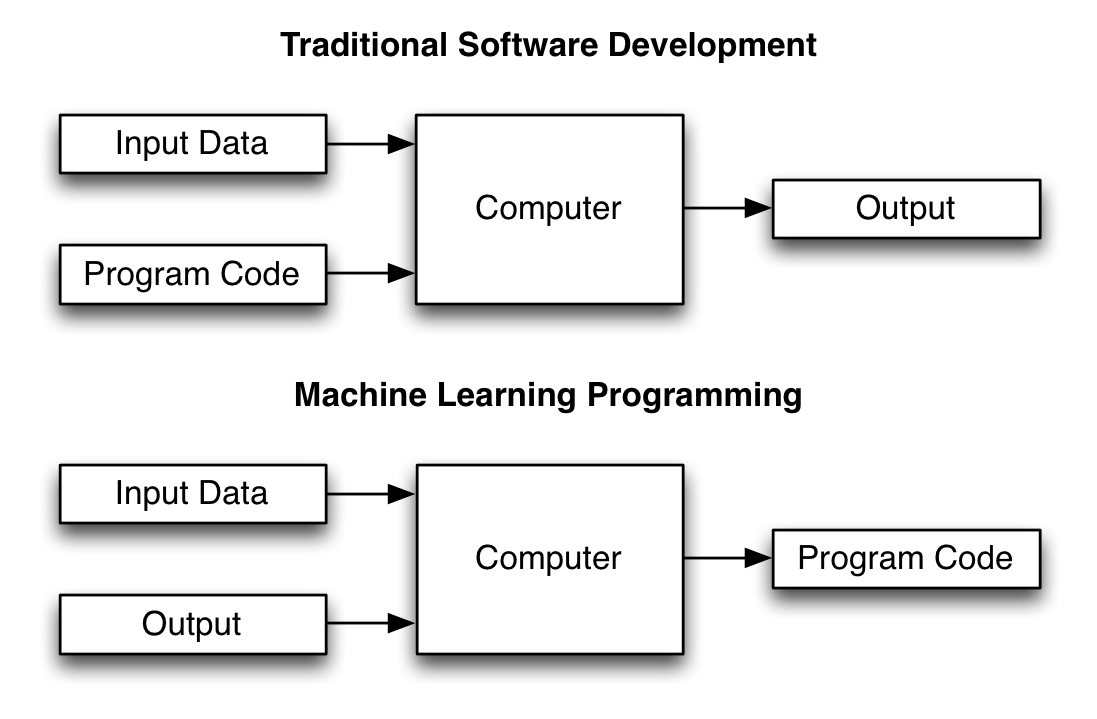}%
\caption{ML vs Traditional Software Development}%
\label{1.ML-DEV}%
\end{figure}

%
If deep learning is a type of machine learning, this begs the question, "What is machine learning?"  Figure \ref{1.ML-DEV}  illustrates how machine learning differs from traditional software development.%
\index{learning}%
\par%

\par
\begin{figure}[h]%
\centering%
\includegraphics[width=4in]{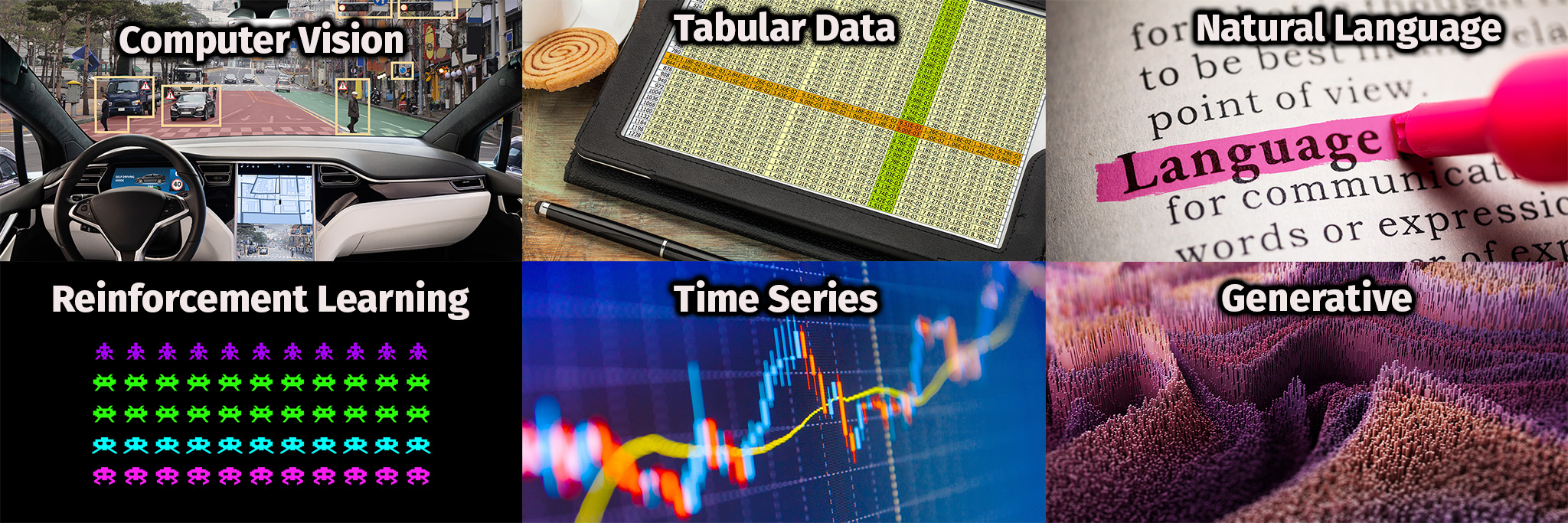}%
\caption{Application of Machine Learning}%
\label{1.ML-DOM}%
\end{figure}

%
\begin{itemize}[noitemsep]%
\item%
\textbf{Traditional Software Development }%
{-} Programmers create programs that specify how to transform input into the desired output.%
\index{input}%
\index{output}%
\item%
\textbf{Machine Learning }%
{-} Programmers create models that can learn to produce the desired output for given input. This learning fills the traditional role of the computer program.%
\index{input}%
\index{learning}%
\index{model}%
\index{output}%
\end{itemize}%
Researchers have applied machine learning to many different areas.  This class explores three specific domains for the application of deep neural networks, as illustrated in Figure \ref{1.ML-DOM}.%
\index{learning}%
\index{neural network}%
\par%

\par%
\begin{itemize}[noitemsep]%
\item%
\textbf{Computer Vision }%
{-} The use of machine learning to detect patterns in visual data. For example, is an image a picture of a cat or a dog.%
\index{learning}%
\item%
\textbf{Tabular Data }%
{-} Several named input values allow the neural network to predict another named value that becomes the output. For example, we are using four measurements of iris flowers to predict the species. This type of data is often called tabular data.%
\index{input}%
\index{iris}%
\index{neural network}%
\index{output}%
\index{predict}%
\index{species}%
\index{tabular data}%
\item%
\textbf{Natural Language Processing (NLP) }%
{-} Deep learning transformers have revolutionized NLP, allowing text sequences to generate more text, images, or classifications.%
\index{classification}%
\index{learning}%
\index{transformer}%
\item%
\textbf{Reinforcement Learning }%
{-} Reinforcement learning trains a neural network to choose ongoing actions so that the algorithm rewards the neural network for optimally completing a task.%
\index{algorithm}%
\index{learning}%
\index{neural network}%
\index{reinforcement learning}%
\item%
\textbf{Time Series }%
{-} The use of machine learning to detect patterns in time. Typical time series applications are financial applications, speech recognition, and even natural language processing (NLP).%
\index{learning}%
\index{ROC}%
\index{ROC}%
\item%
\textbf{Generative Models }%
{-} Neural networks can learn to produce new original synthetic data from input. We will examine StyleGAN, which learns to create new images similar to those it saw during training.%
\index{GAN}%
\index{input}%
\index{neural network}%
\index{StyleGAN}%
\index{training}%
\end{itemize}

\subsection{Regression, Classification and Beyond}%
\label{subsec:Regression,ClassificationandBeyond}%
Machine learning research looks at problems in broad terms of supervised and unsupervised learning. Supervised learning occurs when you know the correct outcome for each item in the training set. On the other hand, unsupervised learning utilizes training sets where no correct outcome is known. Deep learning supports both supervised and unsupervised learning; however, it also adds reinforcement and adversarial learning. Reinforcement learning teaches the neural network to carry out actions based on an environment. Adversarial learning pits two neural networks against each other to learn when the data provides no correct outcomes. Researchers continue to add new deep learning training techniques.%
\index{learning}%
\index{neural network}%
\index{reinforcement learning}%
\index{training}%
\par%
Machine learning practitioners usually divide supervised learning into classification and regression. Classification networks might accept financial data and classify the investment risk as risk or safe. Similarly, a regression neural network outputs a number and might take the same data and return a risk score. Additionally, neural networks can output multiple regression and classification scores simultaneously.%
\index{classification}%
\index{learning}%
\index{neural network}%
\index{output}%
\index{regression}%
\par%
One of the most powerful aspects of neural networks is that the input and output of a neural network can be of many different types, such as:%
\index{input}%
\index{neural network}%
\index{output}%
\par%
\begin{itemize}[noitemsep]%
\item%
An image%
\item%
A series of numbers that could represent text, audio, or another time series%
\item%
A regression number%
\index{regression}%
\item%
A classification class%
\index{classification}%
\end{itemize}

\subsection{Why Deep Learning?}%
\label{subsec:WhyDeepLearning?}%
For tabular data, neural networks often do not perform significantly better that different than other models, such as:%
\index{model}%
\index{neural network}%
\index{tabular data}%
\par%
\begin{itemize}[noitemsep]%
\item%
\href{https://en.wikipedia.org/wiki/Support-vector_machine}{Support Vector Machines}%
\item%
\href{https://en.wikipedia.org/wiki/Random_forest}{Random Forests}%
\item%
\href{https://en.wikipedia.org/wiki/Gradient_boosting}{Gradient Boosted Machines}%
\end{itemize}%
Like these other models, neural networks can perform both%
\index{model}%
\index{neural network}%
\textbf{ classification }%
and%
\textbf{ regression}%
. When applied to relatively low{-}dimensional tabular data tasks, deep neural networks do not necessarily add significant accuracy over other model types. However, most state{-}of{-}the{-}art solutions depend on deep neural networks for images, video, text, and audio data.%
\index{model}%
\index{neural network}%
\index{tabular data}%
\index{video}%
\par

\subsection{Python for Deep Learning}%
\label{subsec:PythonforDeepLearning}%
We will utilize the Python 3.x programming language for this book. Python has some of the widest support for deep learning as a programming language. The two most popular frameworks for deep learning in Python are:%
\index{learning}%
\index{Python}%
\index{SOM}%
\par%
\begin{itemize}[noitemsep]%
\item%
\href{https://www.tensorflow.org/}{TensorFlow/Keras }%
(Google)%
\item%
\href{https://pytorch.org/}{PyTorch }%
(Facebook)%
\end{itemize}%
Overall, this book focused on the application of deep neural networks. This book focuses primarily upon Keras, with some applications in PyTorch. For many tasks, we will utilize Keras directly. We will utilize third{-}party libraries for higher{-}level tasks, such as reinforcement learning, generative adversarial neural networks, and others. These third{-}party libraries may internally make use of either PyTorch or Keras. I chose these libraries based on popularity and application, not whether they used PyTorch or Keras.%
\index{Keras}%
\index{learning}%
\index{neural network}%
\index{PyTorch}%
\index{reinforcement learning}%
\index{SOM}%
\par%
To successfully use this book, you must be able to compile and execute Python code that makes use of TensorFlow for deep learning. There are two options for you to accomplish this:%
\index{learning}%
\index{Python}%
\index{TensorFlow}%
\par%
\begin{itemize}[noitemsep]%
\item%
Install Python, TensorFlow and some IDE (Jupyter, TensorFlow, and others).%
\index{Python}%
\index{SOM}%
\index{TensorFlow}%
\item%
Use%
\href{https://colab.research.google.com/}{ Google CoLab }%
in the cloud, with free GPU access.%
\index{GPU}%
\index{GPU}%
\end{itemize}%
If you look at this notebook on Github, near the top of the document, there are links to videos that describe how to use Google CoLab. There are also videos explaining how to install Python on your local computer. The following sections take you through the process of installing Python on your local computer. This process is essentially the same on Windows, Linux, or Mac. For specific OS instructions, refer to one of the tutorial YouTube videos earlier in this document.%
\index{GitHub}%
\index{link}%
\index{Python}%
\index{ROC}%
\index{ROC}%
\index{video}%
\par%
To install Python on your computer, complete the following instructions:%
\index{Python}%
\par%
\begin{itemize}[noitemsep]%
\item%
\href{https://github.com/jeffheaton/t81_558_deep_learning/blob/master/install/tensorflow-install-jul-2020.ipynb}{Installing Python and TensorFlow {-} Windows/Linux}%
\item%
\href{https://github.com/jeffheaton/t81_558_deep_learning/blob/master/install/tensorflow-install-mac-jan-2021.ipynb}{Installing Python and TensorFlow {-} Mac Intel}%
\item%
\href{https://github.com/jeffheaton/t81_558_deep_learning/blob/master/install/tensorflow-install-mac-metal-jul-2021.ipynb}{Installing Python and TensorFlow {-} Mac M1}%
\end{itemize}

\subsection{Check your Python Installation}%
\label{subsec:CheckyourPythonInstallation}%
Once you've installed Python, you can utilize the following code to check your Python and library versions. If you have a GPU, you can also check to see that Keras recognize it.%
\index{GPU}%
\index{GPU}%
\index{Keras}%
\index{Python}%
\par%
\begin{tcolorbox}[size=title,title=Code,breakable]%
\begin{lstlisting}[language=Python, upquote=true]
# What version of Python do you have?
import sys

import tensorflow.keras
import pandas as pd
import sklearn as sk
import tensorflow as tf

check_gpu = len(tf.config.list_physical_devices('GPU'))>0

print(f"Tensor Flow Version: {tf.__version__}")
print(f"Keras Version: {tensorflow.keras.__version__}")
print()
print(f"Python {sys.version}")
print(f"Pandas {pd.__version__}")
print(f"Scikit-Learn {sk.__version__}")
print("GPU is", "available" if check_gpu \
      else "NOT AVAILABLE")\end{lstlisting}
\tcbsubtitle[before skip=\baselineskip]{Output}%
\begin{lstlisting}[upquote=true]
Tensor Flow Version: 2.8.0
Keras Version: 2.8.0
Python 3.7.13 (default, Mar 16 2022, 17:37:17)
[GCC 7.5.0]
Pandas 1.3.5
Scikit-Learn 1.0.2
GPU is available
\end{lstlisting}
\end{tcolorbox}

\subsection{Module 1 Assignment}%
\label{subsec:Module1Assignment}%
You can find the first assignment here:%
\href{https://github.com/jeffheaton/t81_558_deep_learning/blob/master/assignments/assignment_yourname_class1.ipynb}{ assignment 1}%
\par

\section{Part 1.2: Introduction to Python}%
\label{sec:Part1.2IntroductiontoPython}%
Python is an interpreted, high{-}level, general{-}purpose programming language. Created by Guido van Rossum and first released in 1991, Python's design philosophy emphasizes code readability with its notable use of significant whitespace. Its language constructs and object{-}oriented approach aim to help programmers write clear, logical code for small and large{-}scale projects.  Python has become a common language for machine learning research and is the primary language for TensorFlow.%
\index{learning}%
\index{Python}%
\index{TensorFlow}%
\par%
Python 3.0, released in 2008, was a significant revision of the language that is not entirely backward{-}compatible, and much Python 2 code does not run unmodified on Python 3.  This course makes use of Python 3.  Furthermore, TensorFlow is not compatible with versions of Python earlier than 3. A non{-}profit organization, the Python Software Foundation (PSF), manages and directs resources for Python development. On January 1, 2020, the PSF discontinued the Python 2 language and no longer provides security patches and other improvements. Python interpreters are available for many operating systems.%
\index{GAN}%
\index{Python}%
\index{TensorFlow}%
\par%
The first two modules of this course provide an introduction to some aspects of the Python programming language.  However, entire books focus on Python.  Two modules will not cover every detail of this language.  The reader is encouraged to consult additional sources on the Python language.%
\index{Python}%
\index{SOM}%
\par%
Like most tutorials, we will begin by printing Hello World.%
\par%
\begin{tcolorbox}[size=title,title=Code,breakable]%
\begin{lstlisting}[language=Python, upquote=true]
print("Hello World")\end{lstlisting}
\tcbsubtitle[before skip=\baselineskip]{Output}%
\begin{lstlisting}[upquote=true]
Hello World
\end{lstlisting}
\end{tcolorbox}%
The above code passes a constant string, containing the text "hello world" to a function that is named print.%
\par%
You can also leave comments in your code to explain what you are doing.  Comments can begin anywhere in a line.%
\par%
\begin{tcolorbox}[size=title,title=Code,breakable]%
\begin{lstlisting}[language=Python, upquote=true]
# Single line comment (this has no effect on your program)
print("Hello World") # Say hello\end{lstlisting}
\tcbsubtitle[before skip=\baselineskip]{Output}%
\begin{lstlisting}[upquote=true]
Hello World
\end{lstlisting}
\end{tcolorbox}%
Strings are very versatile and allow your program to process textual information.  Constant string, enclosed in quotes, define literal string values inside your program. Sometimes you may wish to define a larger amount of literal text inside of your program.  This text might consist of multiple lines. The triple quote allows for multiple lines of text.%
\index{ROC}%
\index{ROC}%
\index{SOM}%
\par%
\begin{tcolorbox}[size=title,title=Code,breakable]%
\begin{lstlisting}[language=Python, upquote=true]
print("""Print
Multiple
Lines
""")\end{lstlisting}
\tcbsubtitle[before skip=\baselineskip]{Output}%
\begin{lstlisting}[upquote=true]
Print
Multiple
Lines
\end{lstlisting}
\end{tcolorbox}%
Like many languages Python uses single (') and double (") quotes interchangeably to denote literal string constants. The general convention is that double quotes should enclose actual text, such as words or sentences.  Single quotes should enclose symbolic text, such as error codes.  An example of an error code might be 'HTTP404'.%
\index{error}%
\index{Python}%
\par%
However, there is no difference between single and double quotes in Python, and you may use whichever you like.  The following code makes use of a single quote.%
\index{Python}%
\par%
\begin{tcolorbox}[size=title,title=Code,breakable]%
\begin{lstlisting}[language=Python, upquote=true]
print('Hello World')\end{lstlisting}
\tcbsubtitle[before skip=\baselineskip]{Output}%
\begin{lstlisting}[upquote=true]
Hello World
\end{lstlisting}
\end{tcolorbox}%
In addition to strings, Python allows numbers as literal constants in programs. Python includes support for floating{-}point, integer, complex, and other types of numbers.  This course will not make use of complex numbers.  Unlike strings, quotes do not enclose numbers.%
\index{Python}%
\par%
The presence of a decimal point differentiates floating{-}point and integer numbers.  For example, the value 42 is an integer. Similarly,  42.5 is a floating{-}point number. If you wish to have a floating{-}point number, without a fraction part, you should specify a zero fraction.  The value 42.0 is a floating{-}point number, although it has no fractional part. As an example, the following code prints two numbers.%
\par%
\begin{tcolorbox}[size=title,title=Code,breakable]%
\begin{lstlisting}[language=Python, upquote=true]
print(42)
print(42.5)\end{lstlisting}
\tcbsubtitle[before skip=\baselineskip]{Output}%
\begin{lstlisting}[upquote=true]
42
42.5
\end{lstlisting}
\end{tcolorbox}%
So far, we have only seen how to define literal numeric and string values.  These literal values are constant and do not change as your program runs.  Variables allow your program to hold values that can change as the program runs.  Variables have names that allow you to reference their values. The following code assigns an integer value to a variable named "a" and a string value to a variable named "b."%
\par%
\begin{tcolorbox}[size=title,title=Code,breakable]%
\begin{lstlisting}[language=Python, upquote=true]
a = 10
b = "ten"
print(a)
print(b)\end{lstlisting}
\tcbsubtitle[before skip=\baselineskip]{Output}%
\begin{lstlisting}[upquote=true]
10
ten
\end{lstlisting}
\end{tcolorbox}%
The key feature of variables is that they can change.  The following code demonstrates how to change the values held by variables.%
\index{feature}%
\par%
\begin{tcolorbox}[size=title,title=Code,breakable]%
\begin{lstlisting}[language=Python, upquote=true]
a = 10
print(a)
a = a + 1
print(a)\end{lstlisting}
\tcbsubtitle[before skip=\baselineskip]{Output}%
\begin{lstlisting}[upquote=true]
10
11
\end{lstlisting}
\end{tcolorbox}%
You can mix strings and variables for printing.  This technique is called a formatted or interpolated string.  The variables must be inside of the curly braces.  In Python, this type of string is generally called an f{-}string.  The f{-}string is denoted by placing an "f" just in front of the opening single or double quote that begins the string.  The following code demonstrates the use of an f{-}string to mix several variables with a literal string.%
\index{Python}%
\par%
\begin{tcolorbox}[size=title,title=Code,breakable]%
\begin{lstlisting}[language=Python, upquote=true]
a = 10
print(f'The value of a is {a}')\end{lstlisting}
\tcbsubtitle[before skip=\baselineskip]{Output}%
\begin{lstlisting}[upquote=true]
The value of a is 10
\end{lstlisting}
\end{tcolorbox}%
You can also use f{-}strings with math (called an expression).  Curly braces can enclose any valid Python expression for printing.  The following code demonstrates the use of an expression inside of the curly braces of an f{-}string.%
\index{Python}%
\par%
\begin{tcolorbox}[size=title,title=Code,breakable]%
\begin{lstlisting}[language=Python, upquote=true]
a = 10
print(f'The value of a plus 5 is {a+5}')\end{lstlisting}
\tcbsubtitle[before skip=\baselineskip]{Output}%
\begin{lstlisting}[upquote=true]
The value of a plus 5 is 15
\end{lstlisting}
\end{tcolorbox}%
Python has many ways to print numbers; these are all correct.  However, for this course, we will use f{-}strings. The following code demonstrates some of the varied methods of printing numbers in Python.%
\index{Python}%
\index{SOM}%
\par%
\begin{tcolorbox}[size=title,title=Code,breakable]%
\begin{lstlisting}[language=Python, upquote=true]
a = 5

print(f'a is {a}') # Preferred method for this course.
print('a is {}'.format(a))
print('a is ' + str(a))
print('a is %d' % (a))\end{lstlisting}
\tcbsubtitle[before skip=\baselineskip]{Output}%
\begin{lstlisting}[upquote=true]
a is 5
a is 5
a is 5
a is 5
\end{lstlisting}
\end{tcolorbox}%
You can use if{-}statements to perform logic.  Notice the indents?  These if{-}statements are how Python defines blocks of code to execute together.  A block usually begins after a colon and includes any lines at the same level of indent. Unlike many other programming languages, Python uses whitespace to define blocks of code.  The fact that whitespace is significant to the meaning of program code is a frequent source of annoyance for new programmers of Python.  Tabs and spaces are both used to define the scope in a Python program.  Mixing both spaces and tabs in the same program is not recommended.%
\index{Python}%
\par%
\begin{tcolorbox}[size=title,title=Code,breakable]%
\begin{lstlisting}[language=Python, upquote=true]
a = 5
if a>5:
    print('The variable a is greater than 5.')
else:
    print('The variable a is not greater than 5')\end{lstlisting}
\tcbsubtitle[before skip=\baselineskip]{Output}%
\begin{lstlisting}[upquote=true]
The variable a is not greater than 5
\end{lstlisting}
\end{tcolorbox}%
The following if{-}statement has multiple levels.  It can be easy to indent these levels improperly, so be careful.  This code contains a nested if{-}statement under the first "a==5" if{-}statement.  Only if a is equal to 5 will the nested "b==6" if{-}statement be executed. Also, note that the "elif" command means "else if."%
\par%
\begin{tcolorbox}[size=title,title=Code,breakable]%
\begin{lstlisting}[language=Python, upquote=true]
a = 5
b = 6

if a==5:
    print('The variable a is 5')
    if b==6:
        print('The variable b is also 6')
elif a==6:
    print('The variable a is 6')\end{lstlisting}
\tcbsubtitle[before skip=\baselineskip]{Output}%
\begin{lstlisting}[upquote=true]
The variable a is 5
The variable b is also 6
\end{lstlisting}
\end{tcolorbox}%
It is also important to note that the double equal ("==") operator is used to test the equality of two expressions.  The single equal ("=") operator is only used to assign values to variables in Python.  The greater than (">"), less than ("<"), greater than or equal (">="), less than or equal ("<=") all perform as would generally be accepted.  Testing for inequality is performed with the not equal ("!=") operator.%
\index{Python}%
\par%
It is common in programming languages to loop over a range of numbers.  Python accomplishes this through the use of the%
\index{Python}%
\textbf{ range }%
operation.  Here you can see a%
\textbf{ for }%
loop and a%
\textbf{ range }%
operation that causes the program to loop between 1 and 3.%
\par%
\begin{tcolorbox}[size=title,title=Code,breakable]%
\begin{lstlisting}[language=Python, upquote=true]
for x in range(1, 3):  # If you ever see xrange, you are in Python 2
    print(x)  
# If you ever see print x (no parenthesis), you are in Python 2\end{lstlisting}
\tcbsubtitle[before skip=\baselineskip]{Output}%
\begin{lstlisting}[upquote=true]
1
2
\end{lstlisting}
\end{tcolorbox}%
This code illustrates some incompatibilities between Python 2 and Python 3.  Before Python 3, it was acceptable to leave the parentheses off of a%
\index{Python}%
\index{SOM}%
\textit{ print }%
function call.  This method of invoking the%
\textit{ print }%
command is no longer allowed in Python 3.  Similarly, it used to be a performance improvement to use the%
\index{Python}%
\textit{ xrange }%
command in place of range command at times.  Python 3 incorporated all of the functionality of the%
\index{Python}%
\textit{ xrange }%
Python 2 command into the normal%
\index{Python}%
\textit{ range }%
command.  As a result, the programmer should not use the%
\textit{ xrange }%
command in Python 3.  If you see either of these constructs used in example code, then you are looking at an older Python 2 era example.%
\index{Python}%
\par%
The%
\textit{ range }%
command is used in conjunction with loops to pass over a specific range of numbers.  Cases, where you must loop over specific number ranges, are somewhat uncommon. Generally, programmers use loops on collections of items, rather than hard{-}coding numeric values into your code.  Collections, as well as the operations that loops can perform on them, is covered later in this module.%
\index{SOM}%
\par%
The following is a further example of a looped printing of strings and numbers.%
\par%
\begin{tcolorbox}[size=title,title=Code,breakable]%
\begin{lstlisting}[language=Python, upquote=true]
acc = 0
for x in range(1, 3):
    acc += x
    print(f"Adding {x}, sum so far is {acc}")

print(f"Final sum: {acc}")\end{lstlisting}
\tcbsubtitle[before skip=\baselineskip]{Output}%
\begin{lstlisting}[upquote=true]
Adding 1, sum so far is 1
Adding 2, sum so far is 3
Final sum: 3
\end{lstlisting}
\end{tcolorbox}

\section{Part 1.3: Python Lists, Dictionaries, Sets, and JSON}%
\label{sec:Part1.3PythonLists,Dictionaries,Sets,andJSON}%
Like most modern programming languages, Python includes Lists, Sets, Dictionaries, and other data structures as built{-}in types. The syntax appearance of both of these is similar to JSON. Python and JSON compatibility is discussed later in this module. This course will focus primarily on Lists, Sets, and Dictionaries. It is essential to understand the differences between these three fundamental collection types.%
\index{Python}%
\par%
\begin{itemize}[noitemsep]%
\item%
\textbf{Dictionary }%
{-} A dictionary is a mutable unordered collection that Python indexes with name and value pairs.%
\index{Python}%
\item%
\textbf{List }%
{-} A list is a mutable ordered collection that allows duplicate elements.%
\item%
\textbf{Set }%
{-} A set is a mutable unordered collection with no duplicate elements.%
\item%
\textbf{Tuple }%
{-} A tuple is an immutable ordered collection that allows duplicate elements.%
\end{itemize}%
Most Python collections are mutable, meaning the program can add and remove elements after definition. An immutable collection cannot add or remove items after definition. It is also essential to understand that an ordered collection means that items maintain their order as the program adds them to a collection. This order might not be any specific ordering, such as alphabetic or numeric.%
\index{Python}%
\par%
Lists and tuples are very similar in Python and are often confused. The significant difference is that a list is mutable, but a tuple isn't. So, we include a list when we want to contain similar items and a tuple when we know what information goes into it ahead of time.%
\index{Python}%
\par%
Many programming languages contain a data collection called an array. The array type is noticeably absent in Python. Generally, the programmer will use a list in place of an array in Python. Arrays in most programming languages were fixed{-}length, requiring the program to know the maximum number of elements needed ahead of time. This restriction leads to the infamous array{-}overrun bugs and security issues. The Python list is much more flexible in that the program can dynamically change the size of a list.%
\index{Python}%
\par%
The next sections will look at each collection type in more detail.%
\par%
\subsection{Lists and Tuples}%
\label{subsec:ListsandTuples}%
For a Python program, lists and tuples are very similar. Both lists and tuples hold an ordered collection of items. It is possible to get by as a programmer using only lists and ignoring tuples.%
\index{Python}%
\par%
The primary difference that you will see syntactically is that a list is enclosed by square braces {[}{]}, and a tuple is enclosed by parenthesis (). The following code defines both list and tuple.%
\par%
\begin{tcolorbox}[size=title,title=Code,breakable]%
\begin{lstlisting}[language=Python, upquote=true]
l = ['a', 'b', 'c', 'd']
t = ('a', 'b', 'c', 'd')

print(l)
print(t)\end{lstlisting}
\tcbsubtitle[before skip=\baselineskip]{Output}%
\begin{lstlisting}[upquote=true]
['a', 'b', 'c', 'd']
('a', 'b', 'c', 'd')
\end{lstlisting}
\end{tcolorbox}%
The primary difference you will see programmatically is that a list is mutable, which means the program can change it. A tuple is immutable, which means the program cannot change it. The following code demonstrates that the program can change a list. This code also illustrates that Python indexes lists starting at element 0. Accessing element one modifies the second element in the collection. One advantage of tuples over lists is that tuples are generally slightly faster to iterate over than lists.%
\index{Python}%
\par%
\begin{tcolorbox}[size=title,title=Code,breakable]%
\begin{lstlisting}[language=Python, upquote=true]
l[1] = 'changed'
#t[1] = 'changed' # This would result in an error

print(l)\end{lstlisting}
\tcbsubtitle[before skip=\baselineskip]{Output}%
\begin{lstlisting}[upquote=true]
['a', 'changed', 'c', 'd']
\end{lstlisting}
\end{tcolorbox}%
Like many languages, Python has a for{-}each statement.  This statement allows you to loop over every element in a collection, such as a list or a tuple.%
\index{Python}%
\par%
\begin{tcolorbox}[size=title,title=Code,breakable]%
\begin{lstlisting}[language=Python, upquote=true]
# Iterate over a collection.
for s in l:
    print(s)\end{lstlisting}
\tcbsubtitle[before skip=\baselineskip]{Output}%
\begin{lstlisting}[upquote=true]
a
changed
c
d
\end{lstlisting}
\end{tcolorbox}%
The%
\textbf{ enumerate }%
function is useful for enumerating over a collection and having access to the index of the element that we are currently on.%
\par%
\begin{tcolorbox}[size=title,title=Code,breakable]%
\begin{lstlisting}[language=Python, upquote=true]
# Iterate over a collection, and know where your index.  (Python is zero-based!)
for i,l in enumerate(l):
    print(f"{i}:{l}")\end{lstlisting}
\tcbsubtitle[before skip=\baselineskip]{Output}%
\begin{lstlisting}[upquote=true]
0:a
1:changed
2:c
3:d
\end{lstlisting}
\end{tcolorbox}%
A%
\textbf{ list }%
can have multiple objects added, such as strings.  Duplicate values are allowed.%
\textbf{ Tuples }%
do not allow the program to add additional objects after definition.%
\par%
\begin{tcolorbox}[size=title,title=Code,breakable]%
\begin{lstlisting}[language=Python, upquote=true]
# Manually add items, lists allow duplicates
c = []
c.append('a')
c.append('b')
c.append('c')
c.append('c')
print(c)\end{lstlisting}
\tcbsubtitle[before skip=\baselineskip]{Output}%
\begin{lstlisting}[upquote=true]
['a', 'b', 'c', 'c']
\end{lstlisting}
\end{tcolorbox}%
Ordered collections, such as lists and tuples, allow you to access an element by its index number, as done in the following code. Unordered collections, such as dictionaries and sets, do not allow the program to access them in this way.%
\par%
\begin{tcolorbox}[size=title,title=Code,breakable]%
\begin{lstlisting}[language=Python, upquote=true]
print(c[1])\end{lstlisting}
\tcbsubtitle[before skip=\baselineskip]{Output}%
\begin{lstlisting}[upquote=true]
b
\end{lstlisting}
\end{tcolorbox}%
A%
\textbf{ list }%
can have multiple objects added, such as strings. Duplicate values are allowed. Tuples do not allow the program to add additional objects after definition. The programmer must specify an index for the insert function, an index. These operations are not allowed for tuples because they would result in a change.%
\par%
\begin{tcolorbox}[size=title,title=Code,breakable]%
\begin{lstlisting}[language=Python, upquote=true]
# Insert
c = ['a', 'b', 'c']
c.insert(0, 'a0')
print(c)
# Remove
c.remove('b')
print(c)
# Remove at index
del c[0]
print(c)\end{lstlisting}
\tcbsubtitle[before skip=\baselineskip]{Output}%
\begin{lstlisting}[upquote=true]
['a0', 'a', 'b', 'c']
['a0', 'a', 'c']
['a', 'c']
\end{lstlisting}
\end{tcolorbox}

\subsection{Sets}%
\label{subsec:Sets}%
A Python%
\index{Python}%
\textbf{ set }%
holds an unordered collection of objects, but sets do%
\textit{ not }%
allow duplicates.  If a program adds a duplicate item to a set, only one copy of each item remains in the collection.  Adding a duplicate item to a set does not result in an error.   Any of the following techniques will define a set.%
\index{error}%
\par%
\begin{tcolorbox}[size=title,title=Code,breakable]%
\begin{lstlisting}[language=Python, upquote=true]
s = set()
s = { 'a', 'b', 'c'}
s = set(['a', 'b', 'c'])
print(s)\end{lstlisting}
\tcbsubtitle[before skip=\baselineskip]{Output}%
\begin{lstlisting}[upquote=true]
{'c', 'a', 'b'}
\end{lstlisting}
\end{tcolorbox}%
A%
\textbf{ list }%
is always enclosed in square braces {[}{]}, a%
\textbf{ tuple }%
in parenthesis (), and similarly a%
\textbf{ set }%
is enclosed in curly braces.  Programs can add items to a%
\textbf{ set }%
as they run.  Programs can dynamically add items to a%
\textbf{ set }%
with the%
\textbf{ add }%
function.  It is important to note that the%
\textbf{ append }%
function adds items to lists, whereas the%
\textbf{ add }%
function adds items to a%
\textbf{ set}%
.%
\par%
\begin{tcolorbox}[size=title,title=Code,breakable]%
\begin{lstlisting}[language=Python, upquote=true]
# Manually add items, sets do not allow duplicates
# Sets add, lists append.  I find this annoying.
c = set()
c.add('a')
c.add('b')
c.add('c')
c.add('c')
print(c)\end{lstlisting}
\tcbsubtitle[before skip=\baselineskip]{Output}%
\begin{lstlisting}[upquote=true]
{'c', 'a', 'b'}
\end{lstlisting}
\end{tcolorbox}

\subsection{Maps/Dictionaries/Hash Tables}%
\label{subsec:Maps/Dictionaries/HashTables}%
Many programming languages include the concept of a map, dictionary, or hash table.  These are all very related concepts.  Python provides a dictionary that is essentially a collection of name{-}value pairs.  Programs define dictionaries using curly braces, as seen here.%
\index{Python}%
\par%
\begin{tcolorbox}[size=title,title=Code,breakable]%
\begin{lstlisting}[language=Python, upquote=true]
d = {'name': "Jeff", 'address':"123 Main"}
print(d)
print(d['name'])

if 'name' in d:
    print("Name is defined")

if 'age' in d:
    print("age defined")
else:
    print("age undefined")\end{lstlisting}
\tcbsubtitle[before skip=\baselineskip]{Output}%
\begin{lstlisting}[upquote=true]
{'name': 'Jeff', 'address': '123 Main'}
Jeff
Name is defined
age undefined
\end{lstlisting}
\end{tcolorbox}%
Be careful that you do not attempt to access an undefined key, as this will result in an error.  You can check to see if a key is defined, as demonstrated above.  You can also access the dictionary and provide a default value, as the following code demonstrates.%
\index{error}%
\par%
\begin{tcolorbox}[size=title,title=Code,breakable]%
\begin{lstlisting}[language=Python, upquote=true]
d.get('unknown_key', 'default')\end{lstlisting}
\tcbsubtitle[before skip=\baselineskip]{Output}%
\begin{lstlisting}[upquote=true]
'default'
\end{lstlisting}
\end{tcolorbox}%
You can also access the individual keys and values of a dictionary.%
\par%
\begin{tcolorbox}[size=title,title=Code,breakable]%
\begin{lstlisting}[language=Python, upquote=true]
d = {'name': "Jeff", 'address':"123 Main"}
# All of the keys
print(f"Key: {d.keys()}")

# All of the values
print(f"Values: {d.values()}")\end{lstlisting}
\tcbsubtitle[before skip=\baselineskip]{Output}%
\begin{lstlisting}[upquote=true]
Key: dict_keys(['name', 'address'])
Values: dict_values(['Jeff', '123 Main'])
\end{lstlisting}
\end{tcolorbox}%
Dictionaries and lists can be combined. This syntax is closely related to%
\href{https://en.wikipedia.org/wiki/JSON}{ JSON}%
.  Dictionaries and lists together are a good way to build very complex data structures.  While Python allows quotes (") and apostrophe (') for strings, JSON only allows double{-}quotes ("). We will cover JSON in much greater detail later in this module.%
\index{Python}%
\par%
The following code shows a hybrid usage of dictionaries and lists.%
\par%
\begin{tcolorbox}[size=title,title=Code,breakable]%
\begin{lstlisting}[language=Python, upquote=true]
# Python list & map structures
customers = [
    {"name": "Jeff & Tracy Heaton", "pets": ["Wynton", "Cricket", 
        "Hickory"]},
    {"name": "John Smith", "pets": ["rover"]},
    {"name": "Jane Doe"}
]

print(customers)

for customer in customers:
    print(f"{customer['name']}:{customer.get('pets', 'no pets')}")\end{lstlisting}
\tcbsubtitle[before skip=\baselineskip]{Output}%
\begin{lstlisting}[upquote=true]
[{'name': 'Jeff & Tracy Heaton', 'pets': ['Wynton', 'Cricket',
'Hickory']}, {'name': 'John Smith', 'pets': ['rover']}, {'name': 'Jane
Doe'}]
Jeff & Tracy Heaton:['Wynton', 'Cricket', 'Hickory']
John Smith:['rover']
Jane Doe:no pets
\end{lstlisting}
\end{tcolorbox}%
The variable%
\textbf{ customers }%
is a list that holds three dictionaries that represent customers.  You can think of these dictionaries as records in a table. The fields in these individual records are the keys of the dictionary.  Here the keys%
\textbf{ name }%
and%
\textbf{ pets }%
are fields. However, the field%
\textbf{ pets }%
holds a list of pet names.  There is no limit to how deep you might choose to nest lists and maps.  It is also possible to nest a map inside of a map or a list inside of another list.%
\par

\subsection{More Advanced Lists}%
\label{subsec:MoreAdvancedLists}%
Several advanced features are available for lists that this section introduces. One such function is%
\index{feature}%
\textbf{ zip}%
.  Two lists can be combined into a single list by the%
\textbf{ zip }%
command.  The following code demonstrates the%
\textbf{ zip }%
command.%
\par%
\begin{tcolorbox}[size=title,title=Code,breakable]%
\begin{lstlisting}[language=Python, upquote=true]
a = [1,2,3,4,5]
b = [5,4,3,2,1]

print(zip(a,b))\end{lstlisting}
\tcbsubtitle[before skip=\baselineskip]{Output}%
\begin{lstlisting}[upquote=true]
<zip object at 0x000001802A7A2E08>
\end{lstlisting}
\end{tcolorbox}%
To see the results of the%
\textbf{ zip }%
function, we convert the returned zip object into a list. As you can see, the%
\textbf{ zip }%
function returns a list of tuples.  Each tuple represents a pair of items that the function zipped together.  The order in the two lists was maintained.%
\par%
\begin{tcolorbox}[size=title,title=Code,breakable]%
\begin{lstlisting}[language=Python, upquote=true]
a = [1,2,3,4,5]
b = [5,4,3,2,1]

print(list(zip(a,b)))\end{lstlisting}
\tcbsubtitle[before skip=\baselineskip]{Output}%
\begin{lstlisting}[upquote=true]
[(1, 5), (2, 4), (3, 3), (4, 2), (5, 1)]
\end{lstlisting}
\end{tcolorbox}%
The usual method for using the%
\textbf{ zip }%
command is inside of a for{-}loop.  The following code shows how a for{-}loop can assign a variable to each collection that the program is iterating.%
\par%
\begin{tcolorbox}[size=title,title=Code,breakable]%
\begin{lstlisting}[language=Python, upquote=true]
a = [1,2,3,4,5]
b = [5,4,3,2,1]

for x,y in zip(a,b):
    print(f'{x} - {y}')\end{lstlisting}
\tcbsubtitle[before skip=\baselineskip]{Output}%
\begin{lstlisting}[upquote=true]
1 - 5
2 - 4
3 - 3
4 - 2
5 - 1
\end{lstlisting}
\end{tcolorbox}%
Usually, both collections will be of the same length when passed to the%
\textbf{ zip }%
command.  It is not an error to have collections of different lengths.  As the following code illustrates, the%
\index{error}%
\textbf{ zip }%
command will only process elements up to the length of the smaller collection.%
\index{ROC}%
\index{ROC}%
\par%
\begin{tcolorbox}[size=title,title=Code,breakable]%
\begin{lstlisting}[language=Python, upquote=true]
a = [1,2,3,4,5]
b = [5,4,3]

print(list(zip(a,b)))\end{lstlisting}
\tcbsubtitle[before skip=\baselineskip]{Output}%
\begin{lstlisting}[upquote=true]
[(1, 5), (2, 4), (3, 3)]
\end{lstlisting}
\end{tcolorbox}%
Sometimes you may wish to know the current numeric index when a for{-}loop is iterating through an ordered collection.  Use the%
\index{SOM}%
\textbf{ enumerate }%
command to track the index location for a collection element.  Because the%
\textbf{ enumerate }%
command deals with numeric indexes of the collection, the zip command will assign arbitrary indexes to elements from unordered collections.%
\par%
Consider how you might construct a Python program to change every element greater than 5 to the value of 5.  The following program performs this transformation.  The enumerate command allows the loop to know which element index it is currently on, thus allowing the program to be able to change the value of the current element of the collection.%
\index{Python}%
\par%
\begin{tcolorbox}[size=title,title=Code,breakable]%
\begin{lstlisting}[language=Python, upquote=true]
a = [2, 10, 3, 11, 10, 3, 2, 1]
for i, x in enumerate(a):
    if x>5:
        a[i] = 5
print(a)\end{lstlisting}
\tcbsubtitle[before skip=\baselineskip]{Output}%
\begin{lstlisting}[upquote=true]
[2, 5, 3, 5, 5, 3, 2, 1]
\end{lstlisting}
\end{tcolorbox}%
The comprehension command can dynamically build up a list.  The comprehension below counts from 0 to 9 and adds each value (multiplied by 10) to a list.%
\par%
\begin{tcolorbox}[size=title,title=Code,breakable]%
\begin{lstlisting}[language=Python, upquote=true]
lst = [x*10 for x in range(10)]
print(lst)\end{lstlisting}
\tcbsubtitle[before skip=\baselineskip]{Output}%
\begin{lstlisting}[upquote=true]
[0, 10, 20, 30, 40, 50, 60, 70, 80, 90]
\end{lstlisting}
\end{tcolorbox}%
A dictionary can also be a comprehension.  The general format for this is:%
\par%
\begin{tcolorbox}[size=title,breakable]%
\begin{lstlisting}[upquote=true]
dict_variable = {key:value for (key,value) in dictonary.items()}
\end{lstlisting}
\end{tcolorbox}%
A common use for this is to build up an index to symbolic column names.%
\par%
\begin{tcolorbox}[size=title,title=Code,breakable]%
\begin{lstlisting}[language=Python, upquote=true]
text = ['col-zero','col-one', 'col-two', 'col-three']
lookup = {key:value for (value,key) in enumerate(text)}
print(lookup)\end{lstlisting}
\tcbsubtitle[before skip=\baselineskip]{Output}%
\begin{lstlisting}[upquote=true]
{'col-zero': 0, 'col-one': 1, 'col-two': 2, 'col-three': 3}
\end{lstlisting}
\end{tcolorbox}%
This can be used to easily find the index of a column by name.%
\par%
\begin{tcolorbox}[size=title,title=Code,breakable]%
\begin{lstlisting}[language=Python, upquote=true]
print(f'The index of "col-two" is {lookup["col-two"]}')\end{lstlisting}
\tcbsubtitle[before skip=\baselineskip]{Output}%
\begin{lstlisting}[upquote=true]
The index of "col-two" is 2
\end{lstlisting}
\end{tcolorbox}

\subsection{An Introduction to JSON}%
\label{subsec:AnIntroductiontoJSON}%
Data stored in a CSV file must be flat; it must fit into rows and columns. Most people refer to this type of data as structured or tabular. This data is tabular because the number of columns is the same for every row. Individual rows may be missing a value for a column; however, these rows still have the same columns.%
\index{CSV}%
\par%
This data is convenient for machine learning because most models, such as neural networks, also expect incoming data to be of fixed dimensions. Real{-}world information is not always so tabular. Consider if the rows represent customers. These people might have multiple phone numbers and addresses. How would you describe such data using a fixed number of columns? It would be useful to have a list of these courses in each row that can be variable length for each row or student.%
\index{learning}%
\index{model}%
\index{neural network}%
\par%
JavaScript Object Notation (JSON) is a standard file format that stores data in a hierarchical format similar to eXtensible Markup Language (XML). JSON is nothing more than a hierarchy of lists and dictionaries. Programmers refer to this sort of data as semi{-}structured data or hierarchical data. The following is a sample JSON file.%
\index{Java}%
\index{JavaScript}%
\par%
\begin{tcolorbox}[size=title,breakable]%
\begin{lstlisting}[upquote=true]
{
  "firstName": "John",
  "lastName": "Smith",
  "isAlive": true,
  "age": 27,
  "address": {
    "streetAddress": "21 2nd Street",
    "city": "New York",
    "state": "NY",
    "postalCode": "10021-3100"
  },
  "phoneNumbers": [
    {
      "type": "home",
      "number": "212 555-1234"
    },
    {
      "type": "office",
      "number": "646 555-4567"
    },
    {
      "type": "mobile",
      "number": "123 456-7890"
    }
  ],
  "children": [],
  "spouse": null
}
\end{lstlisting}
\end{tcolorbox}%
The above file may look somewhat like Python code.  You can see curly braces that define dictionaries and square brackets that define lists.  JSON does require there to be a single root element.  A list or dictionary can fulfill this role.  JSON requires double{-}quotes to enclose strings and names.  Single quotes are not allowed in JSON.%
\index{Python}%
\index{SOM}%
\par%
JSON files are always legal JavaScript syntax.  JSON is also generally valid as Python code, as demonstrated by the following Python program.%
\index{Java}%
\index{JavaScript}%
\index{Python}%
\par%
\begin{tcolorbox}[size=title,title=Code,breakable]%
\begin{lstlisting}[language=Python, upquote=true]
jsonHardCoded = {
  "firstName": "John",
  "lastName": "Smith",
  "isAlive": True,
  "age": 27,
  "address": {
    "streetAddress": "21 2nd Street",
    "city": "New York",
    "state": "NY",
    "postalCode": "10021-3100"
  },
  "phoneNumbers": [
    {
      "type": "home",
      "number": "212 555-1234"
    },
    {
      "type": "office",
      "number": "646 555-4567"
    },
    {
      "type": "mobile",
      "number": "123 456-7890"
    }
  ],
  "children": [],
  "spouse": None
}\end{lstlisting}
\end{tcolorbox}%
Generally, it is better to read JSON from files, strings, or the Internet than hard coding, as demonstrated here.  However, for internal data structures, sometimes such hard{-}coding can be useful.%
\index{SOM}%
\par%
Python contains support for JSON.  When a Python program loads a JSON  the root list or dictionary is returned, as demonstrated by the following code.%
\index{Python}%
\par%
\begin{tcolorbox}[size=title,title=Code,breakable]%
\begin{lstlisting}[language=Python, upquote=true]
import json

json_string = '{"first":"Jeff","last":"Heaton"}'
obj = json.loads(json_string)
print(f"First name: {obj['first']}")
print(f"Last name: {obj['last']}")\end{lstlisting}
\tcbsubtitle[before skip=\baselineskip]{Output}%
\begin{lstlisting}[upquote=true]
First name: Jeff
Last name: Heaton
\end{lstlisting}
\end{tcolorbox}%
Python programs can also load JSON from a file or URL.%
\index{Python}%
\par%
\begin{tcolorbox}[size=title,title=Code,breakable]%
\begin{lstlisting}[language=Python, upquote=true]
import requests

r = requests.get("https://raw.githubusercontent.com/jeffheaton/"
                 +"t81_558_deep_learning/master/person.json")
print(r.json())\end{lstlisting}
\tcbsubtitle[before skip=\baselineskip]{Output}%
\begin{lstlisting}[upquote=true]
{'firstName': 'John', 'lastName': 'Smith', 'isAlive': True, 'age': 27,
'address': {'streetAddress': '21 2nd Street', 'city': 'New York',
'state': 'NY', 'postalCode': '10021-3100'}, 'phoneNumbers': [{'type':
'home', 'number': '212 555-1234'}, {'type': 'office', 'number': '646
555-4567'}, {'type': 'mobile', 'number': '123 456-7890'}], 'children':
[], 'spouse': None}
\end{lstlisting}
\end{tcolorbox}%
Python programs can easily generate JSON strings from Python objects of dictionaries and lists.%
\index{Python}%
\par%
\begin{tcolorbox}[size=title,title=Code,breakable]%
\begin{lstlisting}[language=Python, upquote=true]
python_obj = {"first":"Jeff","last":"Heaton"}
print(json.dumps(python_obj))\end{lstlisting}
\tcbsubtitle[before skip=\baselineskip]{Output}%
\begin{lstlisting}[upquote=true]
{"first": "Jeff", "last": "Heaton"}
\end{lstlisting}
\end{tcolorbox}%
A data scientist will generally encounter JSON when they access web services to get their data.  A data scientist might use the techniques presented in this section to convert the semi{-}structured JSON data into tabular data for the program to use with a model such as a neural network.%
\index{data scientist}%
\index{model}%
\index{neural network}%
\index{tabular data}%
\par

\section{Part 1.4: File Handling}%
\label{sec:Part1.4FileHandling}%
Files often contain the data that you use to train your AI programs. Once trained, your models may use real{-}time data to form predictions. These predictions might be made on files too. Regardless of predicting or training, file processing is a vital skill for the AI practitioner.%
\index{model}%
\index{predict}%
\index{ROC}%
\index{ROC}%
\index{training}%
\par%
There are many different types of files that you must process as an AI practitioner. Some of these file types are listed here:%
\index{ROC}%
\index{ROC}%
\index{SOM}%
\par%
\begin{itemize}[noitemsep]%
\item%
\textbf{CSV files }%
(generally have the .csv extension) hold tabular data that resembles spreadsheet data.%
\index{CSV}%
\index{tabular data}%
\item%
\textbf{Image files }%
(generally with the .png or .jpg extension) hold images for computer vision.%
\index{computer vision}%
\item%
\textbf{Text files }%
(often have the .txt extension) hold unstructured text and are essential for natural language processing.%
\index{ROC}%
\index{ROC}%
\item%
\textbf{JSON }%
(often have the .json extension) contain semi{-}structured textual data in a human{-}readable text{-}based format.%
\item%
\textbf{H5 }%
(can have a wide array of extensions) contain semi{-}structured textual data in a human{-}readable text{-}based format. Keras and TensorFlow store neural networks as H5 files.%
\index{Keras}%
\index{neural network}%
\index{TensorFlow}%
\item%
\textbf{Audio Files }%
(often have an extension such as .au or .wav) contain recorded sound.%
\end{itemize}%
Data can come from a variety of sources. In this class, we obtain data from three primary locations:%
\par%
\begin{itemize}[noitemsep]%
\item%
\textbf{Your Hard Drive }%
{-}  This type of data is stored locally, and Python accesses it from a path that looks something like:%
\index{Python}%
\index{SOM}%
\textbf{ c:\textbackslash{}data\textbackslash{}myfile.csv or /Users/jheaton/data/myfile.csv}%
.%
\item%
\textbf{The Internet }%
{-}  This type of data resides in the cloud, and Python accesses it from a URL that looks something like:%
\index{Python}%
\index{SOM}%
\end{itemize}%
https://data.heatonresearch.com/data/t81{-}558/iris.csv.%
\index{CSV}%
\index{iris}%
\par%
\begin{itemize}[noitemsep]%
\item%
\textbf{Google Drive (cloud) }%
{-} If your code in Google CoLab, you use GoogleDrive to save and load some data files. CoLab mounts your GoogleDrive into a path similar to the following:%
\index{SOM}%
\textbf{ /content/drive/My Drive/myfile.csv}%
.%
\end{itemize}%
\subsection{Read a CSV File}%
\label{subsec:ReadaCSVFile}%
Python programs can read CSV files with Pandas. We will see more about Pandas in the next section, but for now, its general format is:%
\index{CSV}%
\index{Python}%
\par%
\begin{tcolorbox}[size=title,title=Code,breakable]%
\begin{lstlisting}[language=Python, upquote=true]
import pandas as pd

df = pd.read_csv("https://data.heatonresearch.com/data/t81-558/iris.csv")\end{lstlisting}
\end{tcolorbox}%
The above command loads%
\href{https://en.wikipedia.org/wiki/Iris_flower_data_set}{ Fisher's Iris data set }%
from the Internet.  It might take a few seconds to load, so it is good to keep the loading code in a separate Jupyter notebook cell so that you do not have to reload it as you test your program.  You can load Internet data, local hard drive, and Google Drive data this way.%
\par%
Now that the data is loaded, you can display the first five rows with this command.%
\par%
\begin{tcolorbox}[size=title,title=Code,breakable]%
\begin{lstlisting}[language=Python, upquote=true]
display(df[0:5])\end{lstlisting}
\tcbsubtitle[before skip=\baselineskip]{Output}%
\begin{tabular}[hbt!]{l|l|l|l|l|l}%
\hline%
&sepal\_l&sepal\_w&petal\_l&petal\_w&species\\%
\hline%
0&5.1&3.5&1.4&0.2&Iris{-}setosa\\%
1&4.9&3.0&1.4&0.2&Iris{-}setosa\\%
2&4.7&3.2&1.3&0.2&Iris{-}setosa\\%
3&4.6&3.1&1.5&0.2&Iris{-}setosa\\%
4&5.0&3.6&1.4&0.2&Iris{-}setosa\\%
\hline%
\end{tabular}%
\vspace{2mm}%
\end{tcolorbox}

\subsection{Read (stream) a Large CSV File}%
\label{subsec:Read(stream)aLargeCSVFile}%
Pandas will read the entire CSV file into memory. Usually, this is fine.  However, at times you may wish to "stream" a huge file.  Streaming allows you to process this file one record at a time.  Because the program does not load all of the data into memory, you can handle huge files.  The following code loads the Iris dataset and calculates averages, one row at a time.  This technique would work for large files.%
\index{CSV}%
\index{dataset}%
\index{iris}%
\index{ROC}%
\index{ROC}%
\par%
\begin{tcolorbox}[size=title,title=Code,breakable]%
\begin{lstlisting}[language=Python, upquote=true]
import csv
import urllib.request
import codecs
import numpy as np

url = "https://data.heatonresearch.com/data/t81-558/iris.csv"
urlstream = urllib.request.urlopen(url)
csvfile = csv.reader(codecs.iterdecode(urlstream, 'utf-8'))
next(csvfile) # Skip header row
sum = np.zeros(4)
count = 0

for line in csvfile:
    # Convert each row to Numpy array
    line2 = np.array(line)[0:4].astype(float)
    
    # If the line is of the right length (skip empty lines), then add
    if len(line2) == 4:
        sum += line2
        count += 1
        
# Calculate the average, and print the average of the 4 iris 
# measurements (features)
print(sum/count)\end{lstlisting}
\tcbsubtitle[before skip=\baselineskip]{Output}%
\begin{lstlisting}[upquote=true]
[5.84333333 3.05733333 3.758      1.19933333]
\end{lstlisting}
\end{tcolorbox}

\subsection{Read a Text File}%
\label{subsec:ReadaTextFile}%
The following code reads the%
\href{https://en.wikipedia.org/wiki/Sonnet_18}{ Sonnet 18 }%
by%
\href{https://en.wikipedia.org/wiki/William_Shakespeare}{ William Shakespeare }%
as a text file.  This code streams the document and reads it line{-}by{-}line.  This code could handle a huge file.%
\par%
\begin{tcolorbox}[size=title,title=Code,breakable]%
\begin{lstlisting}[language=Python, upquote=true]
import urllib.request

url = "https://data.heatonresearch.com/data/t81-558/datasets/sonnet_18.txt"
with urllib.request.urlopen(url) as urlstream:
    for line in codecs.iterdecode(urlstream, 'utf-8'):
        print(line.rstrip())\end{lstlisting}
\tcbsubtitle[before skip=\baselineskip]{Output}%
\begin{lstlisting}[upquote=true]
Sonnet 18 original text
William Shakespeare
Shall I compare thee to a summer's day?
Thou art more lovely and more temperate:
Rough winds do shake the darling buds of May,
And summer's lease hath all too short a date:
Sometime too hot the eye of heaven shines,
And often is his gold complexion dimm'd;
And every fair from fair sometime declines,
By chance or nature's changing course untrimm'd;
But thy eternal summer shall not fade
Nor lose possession of that fair thou owest;
Nor shall Death brag thou wander'st in his shade,
When in eternal lines to time thou growest:
So long as men can breathe or eyes can see,
So long lives this and this gives life to thee.
\end{lstlisting}
\end{tcolorbox}

\subsection{Read an Image}%
\label{subsec:ReadanImage}%
Computer vision is one of the areas that neural networks outshine other models. To support computer vision, the Python programmer needs to understand how to process images.  For this course, we will use the Python PIL package for image processing.  The following code demonstrates how to load an image from a URL and display it.%
\index{computer vision}%
\index{model}%
\index{neural network}%
\index{Python}%
\index{ROC}%
\index{ROC}%
\par%
\begin{tcolorbox}[size=title,title=Code,breakable]%
\begin{lstlisting}[language=Python, upquote=true]
%matplotlib inline
from PIL import Image
import requests
from io import BytesIO

url = "https://data.heatonresearch.com/images/jupyter/brookings.jpeg"

response = requests.get(url)
img = Image.open(BytesIO(response.content))

img\end{lstlisting}
\tcbsubtitle[before skip=\baselineskip]{Output}%
\includegraphics[width=4in]{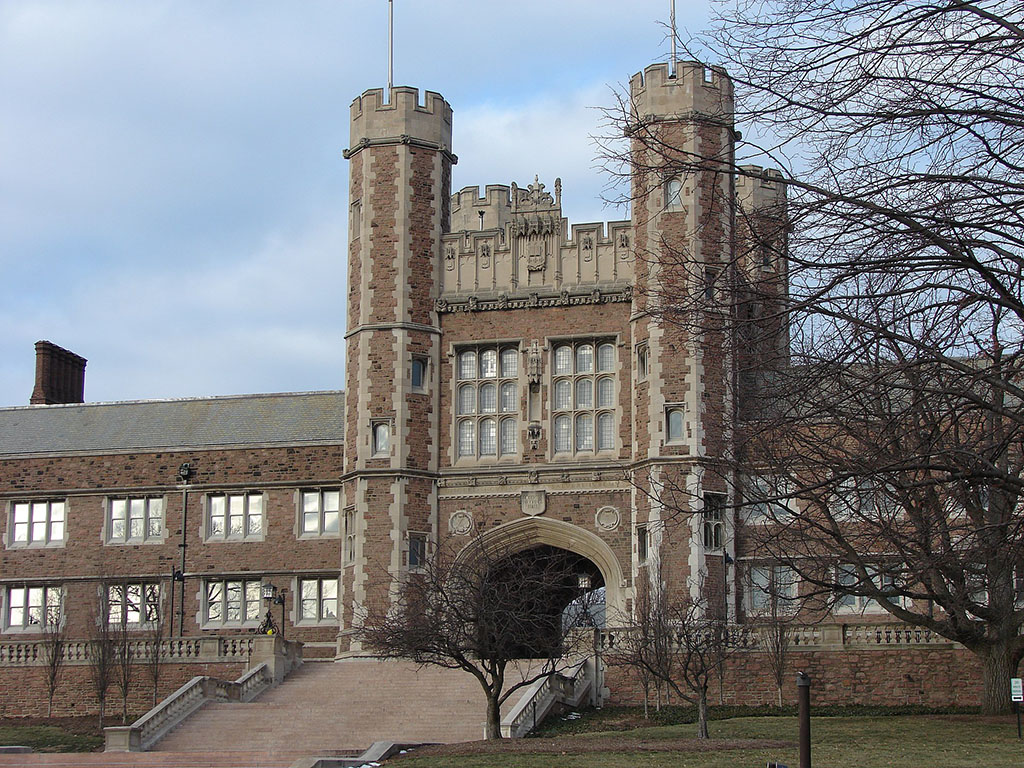}%
\end{tcolorbox}

\section{Part 1.5: Functions, Lambdas, and Map/Reduce}%
\label{sec:Part1.5Functions,Lambdas,andMap/Reduce}%
Functions,%
\textbf{ lambdas}%
, and%
\textbf{ map/reduce }%
can allow you to process your data in advanced ways.  We will introduce these techniques here and expand on them in the next module, which will discuss Pandas.%
\index{ROC}%
\index{ROC}%
\par%
Function parameters can be named or unnamed in Python.  Default values can also be used.  Consider the following function.%
\index{parameter}%
\index{Python}%
\par%
\begin{tcolorbox}[size=title,title=Code,breakable]%
\begin{lstlisting}[language=Python, upquote=true]
def say_hello(speaker, person_to_greet, greeting = "Hello"):
    print(f'{greeting} {person_to_greet}, this is {speaker}.')
    
say_hello('Jeff', "John")
say_hello('Jeff', "John", "Goodbye")
say_hello(speaker='Jeff', person_to_greet="John", greeting = "Goodbye")\end{lstlisting}
\tcbsubtitle[before skip=\baselineskip]{Output}%
\begin{lstlisting}[upquote=true]
Hello John, this is Jeff.
Goodbye John, this is Jeff.
Goodbye John, this is Jeff.
\end{lstlisting}
\end{tcolorbox}%
A function is a way to capture code that is commonly executed.  Consider the following function that can be used to trim white space from a string capitalize the first letter.%
\par%
\begin{tcolorbox}[size=title,title=Code,breakable]%
\begin{lstlisting}[language=Python, upquote=true]
def process_string(str):
    t = str.strip()
    return t[0].upper()+t[1:]\end{lstlisting}
\end{tcolorbox}%
This function can now be called quite easily.%
\par%
\begin{tcolorbox}[size=title,title=Code,breakable]%
\begin{lstlisting}[language=Python, upquote=true]
str = process_string("  hello  ")
print(f'"{str}"')\end{lstlisting}
\tcbsubtitle[before skip=\baselineskip]{Output}%
\begin{lstlisting}[upquote=true]
"Hello"
\end{lstlisting}
\end{tcolorbox}%
Python's%
\index{Python}%
\textbf{ map }%
is a very useful function that is provided in many different programming languages.  The%
\textbf{ map }%
function takes a%
\textbf{ list }%
and applies a function to each member of the%
\textbf{ list }%
and returns a second%
\textbf{ list }%
that is the same size as the first.%
\par%
\begin{tcolorbox}[size=title,title=Code,breakable]%
\begin{lstlisting}[language=Python, upquote=true]
l = ['   apple  ', 'pear ', 'orange', 'pine apple  ']
list(map(process_string, l))\end{lstlisting}
\tcbsubtitle[before skip=\baselineskip]{Output}%
\begin{lstlisting}[upquote=true]
['Apple', 'Pear', 'Orange', 'Pine apple']
\end{lstlisting}
\end{tcolorbox}%
\subsection{Map}%
\label{subsec:Map}%
The%
\textbf{ map }%
function is very similar to the Python%
\index{Python}%
\textbf{ comprehension }%
that we previously explored.  The following%
\textbf{ comprehension }%
accomplishes the same task as the previous call to%
\textbf{ map}%
.%
\par%
\begin{tcolorbox}[size=title,title=Code,breakable]%
\begin{lstlisting}[language=Python, upquote=true]
l = ['   apple  ', 'pear ', 'orange', 'pine apple  ']
l2 = [process_string(x) for x in l]
print(l2)\end{lstlisting}
\tcbsubtitle[before skip=\baselineskip]{Output}%
\begin{lstlisting}[upquote=true]
['Apple', 'Pear', 'Orange', 'Pine apple']
\end{lstlisting}
\end{tcolorbox}%
The choice of using a%
\textbf{ map }%
function or%
\textbf{ comprehension }%
is up to the programmer.  I tend to prefer%
\textbf{ map }%
since it is so common in other programming languages.%
\par

\subsection{Filter}%
\label{subsec:Filter}%
While a%
\textbf{ map function }%
always creates a new%
\textbf{ list }%
of the same size as the original, the%
\textbf{ filter }%
function creates a potentially smaller%
\textbf{ list}%
.%
\par%
\begin{tcolorbox}[size=title,title=Code,breakable]%
\begin{lstlisting}[language=Python, upquote=true]
def greater_than_five(x):
    return x>5

l = [ 1, 10, 20, 3, -2, 0]
l2 = list(filter(greater_than_five, l))
print(l2)\end{lstlisting}
\tcbsubtitle[before skip=\baselineskip]{Output}%
\begin{lstlisting}[upquote=true]
[10, 20]
\end{lstlisting}
\end{tcolorbox}

\subsection{Lambda}%
\label{subsec:Lambda}%
It might seem somewhat tedious to have to create an entire function just to check to see if a value is greater than 5.  A%
\index{SOM}%
\textbf{ lambda }%
saves you this effort.  A lambda is essentially an unnamed function.%
\par%
\begin{tcolorbox}[size=title,title=Code,breakable]%
\begin{lstlisting}[language=Python, upquote=true]
l = [ 1, 10, 20, 3, -2, 0]
l2 = list(filter(lambda x: x>5, l))
print(l2)\end{lstlisting}
\tcbsubtitle[before skip=\baselineskip]{Output}%
\begin{lstlisting}[upquote=true]
[10, 20]
\end{lstlisting}
\end{tcolorbox}

\subsection{Reduce}%
\label{subsec:Reduce}%
Finally, we will make use of%
\textbf{ reduce}%
.  Like%
\textbf{ filter }%
and%
\textbf{ map }%
the%
\textbf{ reduce }%
function also works on a%
\textbf{ list}%
.  However, the result of the%
\textbf{ reduce }%
is a single value.  Consider if you wanted to sum the%
\textbf{ values }%
of a%
\textbf{ list}%
.  The sum is implemented by a%
\textbf{ lambda}%
.%
\par%
\begin{tcolorbox}[size=title,title=Code,breakable]%
\begin{lstlisting}[language=Python, upquote=true]
from functools import reduce

l = [ 1, 10, 20, 3, -2, 0]
result = reduce(lambda x,y: x+y,l)
print(result)\end{lstlisting}
\tcbsubtitle[before skip=\baselineskip]{Output}%
\begin{lstlisting}[upquote=true]
32
\end{lstlisting}
\end{tcolorbox}

\chapter{Python for Machine Learning}%
\label{chap:PythonforMachineLearning}%
\section{Part 2.1: Introduction to Pandas}%
\label{sec:Part2.1IntroductiontoPandas}%
\href{http://pandas.pydata.org/}{Pandas }%
is an open{-}source library providing high{-}performance, easy{-}to{-}use data structures and data analysis tools for the Python programming language.  It is based on the%
\index{Python}%
\href{http://pandas.pydata.org/pandas-docs/stable/generated/pandas.DataFrame.html}{ dataframe }%
concept found in the%
\href{https://www.r-project.org/about.html}{ R programming language}%
.  For this class, Pandas will be the primary means by which we manipulate data to be processed by neural networks.%
\index{neural network}%
\index{ROC}%
\index{ROC}%
\par%
The data frame is a crucial component of Pandas.  We will use it to access the%
\href{https://archive.ics.uci.edu/ml/datasets/Auto+MPG}{ auto{-}mpg dataset}%
.  You can find this dataset on the UCI machine learning repository.  For this class, we will use a version of the Auto MPG dataset, where I added column headers.  You can find my%
\index{dataset}%
\index{learning}%
\href{https://data.heatonresearch.com/data/t81-558/auto-mpg.csv}{ version }%
at%
\href{https://data.heatonresearch.com/}{ https://data.heatonresearch.com/}%
.%
\par%
UCI took this dataset from the StatLib library, which Carnegie Mellon University maintains. The dataset was used in the 1983 American Statistical Association Exposition.  It contains data for 398 cars, including%
\index{dataset}%
\href{https://en.wikipedia.org/wiki/Fuel_economy_in_automobiles}{ mpg}%
,%
\href{https://en.wikipedia.org/wiki/Cylinder_(engine)}{ cylinders}%
,%
\href{https://en.wikipedia.org/wiki/Engine_displacement}{ displacement}%
,%
\href{https://en.wikipedia.org/wiki/Horsepower}{ horsepower }%
, weight, acceleration, model year, origin and the car's name.%
\index{model}%
\par%
The following code loads the MPG dataset into a data frame:%
\index{dataset}%
\par%
\begin{tcolorbox}[size=title,title=Code,breakable]%
\begin{lstlisting}[language=Python, upquote=true]
# Simple dataframe
import os
import pandas as pd

pd.set_option('display.max_columns', 7)
df = pd.read_csv(
    "https://data.heatonresearch.com/data/t81-558/auto-mpg.csv")
display(df[0:5])\end{lstlisting}
\tcbsubtitle[before skip=\baselineskip]{Output}%
\begin{tabular}[hbt!]{l|l|l|l|l|l|l|l}%
\hline%
&mpg&cylinders&displacement&...&year&origin&name\\%
\hline%
0&18.0&8&307.0&...&70&1&chevrolet chevelle malibu\\%
1&15.0&8&350.0&...&70&1&buick skylark 320\\%
2&18.0&8&318.0&...&70&1&plymouth satellite\\%
3&16.0&8&304.0&...&70&1&amc rebel sst\\%
4&17.0&8&302.0&...&70&1&ford torino\\%
\hline%
\end{tabular}%
\vspace{2mm}%
\end{tcolorbox}%
The%
\textbf{ display }%
function provides a cleaner display than merely printing the data frame.  Specifying the maximum rows and columns allows you to achieve greater control over the display.%
\par%
\begin{tcolorbox}[size=title,title=Code,breakable]%
\begin{lstlisting}[language=Python, upquote=true]
pd.set_option('display.max_columns', 7)
pd.set_option('display.max_rows', 5)
display(df)\end{lstlisting}
\tcbsubtitle[before skip=\baselineskip]{Output}%
\begin{tabular}[hbt!]{l|l|l|l|l|l|l|l}%
\hline%
&mpg&cylinders&displacement&...&year&origin&name\\%
\hline%
0&18.0&8&307.0&...&70&1&chevrolet chevelle malibu\\%
1&15.0&8&350.0&...&70&1&buick skylark 320\\%
...&...&...&...&...&...&...&...\\%
396&28.0&4&120.0&...&82&1&ford ranger\\%
397&31.0&4&119.0&...&82&1&chevy s{-}10\\%
\hline%
\end{tabular}%
\vspace{2mm}%
\end{tcolorbox}%
It is possible to generate a second data frame to display statistical information about the first data frame.%
\par%
\begin{tcolorbox}[size=title,title=Code,breakable]%
\begin{lstlisting}[language=Python, upquote=true]
# Strip non-numerics
df = df.select_dtypes(include=['int', 'float'])

headers = list(df.columns.values)
fields = []

for field in headers:
    fields.append({
        'name' : field,
        'mean': df[field].mean(),
        'var': df[field].var(),
        'sdev': df[field].std()
    })

for field in fields:
    print(field)\end{lstlisting}
\tcbsubtitle[before skip=\baselineskip]{Output}%
\begin{lstlisting}[upquote=true]
{'name': 'mpg', 'mean': 23.514572864321607, 'var': 61.089610774274405,
'sdev': 7.815984312565782}
{'name': 'cylinders', 'mean': 5.454773869346734, 'var':
2.893415439920003, 'sdev': 1.7010042445332119}
{'name': 'displacement', 'mean': 193.42587939698493, 'var':
10872.199152247384, 'sdev': 104.26983817119591}
{'name': 'weight', 'mean': 2970.424623115578, 'var':
717140.9905256763, 'sdev': 846.8417741973268}
{'name': 'acceleration', 'mean': 15.568090452261307, 'var':
7.604848233611383, 'sdev': 2.757688929812676}
{'name': 'year', 'mean': 76.01005025125629, 'var': 13.672442818627143,
'sdev': 3.697626646732623}
{'name': 'origin', 'mean': 1.5728643216080402, 'var':
0.6432920268850549, 'sdev': 0.8020548777266148}
\end{lstlisting}
\end{tcolorbox}%
This code outputs a list of dictionaries that hold this statistical information.  This information looks similar to the JSON code seen in Module 1.  If proper JSON is needed, the program should add these records to a list and call the Python JSON library's%
\index{output}%
\index{Python}%
\textbf{ dumps }%
command.%
\par%
The Python program can convert this JSON{-}like information to a data frame for better display.%
\index{Python}%
\par%
\begin{tcolorbox}[size=title,title=Code,breakable]%
\begin{lstlisting}[language=Python, upquote=true]
pd.set_option('display.max_columns', 0)
pd.set_option('display.max_rows', 0)
df2 = pd.DataFrame(fields)
display(df2)\end{lstlisting}
\tcbsubtitle[before skip=\baselineskip]{Output}%
\begin{tabular}[hbt!]{l|l|l|l|l}%
\hline%
&name&mean&var&sdev\\%
\hline%
0&mpg&23.514573&61.089611&7.815984\\%
1&cylinders&5.454774&2.893415&1.701004\\%
2&displacement&193.425879&10872.199152&104.269838\\%
3&weight&2970.424623&717140.990526&846.841774\\%
4&acceleration&15.568090&7.604848&2.757689\\%
5&year&76.010050&13.672443&3.697627\\%
6&origin&1.572864&0.643292&0.802055\\%
\hline%
\end{tabular}%
\vspace{2mm}%
\end{tcolorbox}%
\subsection{Missing Values}%
\label{subsec:MissingValues}%
Missing values are a reality of machine learning.  Ideally, every row of data will have values for all columns.  However, this is rarely the case.  Most of the values are present in the MPG database.  However, there are missing values in the horsepower column.  A common practice is to replace missing values with the median value for that column.  The program calculates the%
\index{learning}%
\href{https://en.wikipedia.org/wiki/Median}{ median}%
.  The following code replaces any NA values in horsepower with the median:%
\par%
\begin{tcolorbox}[size=title,title=Code,breakable]%
\begin{lstlisting}[language=Python, upquote=true]
import os
import pandas as pd

df = pd.read_csv(
    "https://data.heatonresearch.com/data/t81-558/auto-mpg.csv", 
    na_values=['NA', '?'])
print(f"horsepower has na? {pd.isnull(df['horsepower']).values.any()}")
    
print("Filling missing values...")
med = df['horsepower'].median()
df['horsepower'] = df['horsepower'].fillna(med)
# df = df.dropna() # you can also simply drop NA values
                 
print(f"horsepower has na? {pd.isnull(df['horsepower']).values.any()}")\end{lstlisting}
\tcbsubtitle[before skip=\baselineskip]{Output}%
\begin{lstlisting}[upquote=true]
horsepower has na? True
Filling missing values...
horsepower has na? False
\end{lstlisting}
\end{tcolorbox}

\subsection{Dealing with Outliers}%
\label{subsec:DealingwithOutliers}%
Outliers are values that are unusually high or low. We typically consider outliers to be a value that is several standard deviations from the mean. Sometimes outliers are simply errors; this is a result of%
\index{error}%
\index{SOM}%
\index{standard deviation}%
\href{https://en.wikipedia.org/wiki/Observational_error}{ observation error}%
. Outliers can also be truly large or small values that may be difficult to address. The following function can remove such values.%
\par%
\begin{tcolorbox}[size=title,title=Code,breakable]%
\begin{lstlisting}[language=Python, upquote=true]
# Remove all rows where the specified column is +/- sd standard deviations
def remove_outliers(df, name, sd):
    drop_rows = df.index[(np.abs(df[name] - df[name].mean())
                          >= (sd * df[name].std()))]
    df.drop(drop_rows, axis=0, inplace=True)\end{lstlisting}
\end{tcolorbox}%
The code below will drop every row from the Auto MPG dataset where the horsepower is two standard deviations or more above or below the mean.%
\index{dataset}%
\index{standard deviation}%
\par%
\begin{tcolorbox}[size=title,title=Code,breakable]%
\begin{lstlisting}[language=Python, upquote=true]
import pandas as pd
import os
import numpy as np
from sklearn import metrics
from scipy.stats import zscore

df = pd.read_csv(
    "https://data.heatonresearch.com/data/t81-558/auto-mpg.csv",
    na_values=['NA','?'])

# create feature vector
med = df['horsepower'].median()
df['horsepower'] = df['horsepower'].fillna(med)

# Drop the name column
df.drop('name',1,inplace=True)

# Drop outliers in horsepower
print("Length before MPG outliers dropped: {}".format(len(df)))
remove_outliers(df,'mpg',2)
print("Length after MPG outliers dropped: {}".format(len(df)))

pd.set_option('display.max_columns', 0)
pd.set_option('display.max_rows', 5)
display(df)\end{lstlisting}
\tcbsubtitle[before skip=\baselineskip]{Output}%
\begin{tabular}[hbt!]{l|l|l|l|l|l|l|l|l}%
\hline%
&mpg&cylinders&displacement&horsepower&weight&acceleration&year&origin\\%
\hline%
0&18.0&8&307.0&130.0&3504&12.0&70&1\\%
1&15.0&8&350.0&165.0&3693&11.5&70&1\\%
...&...&...&...&...&...&...&...&...\\%
396&28.0&4&120.0&79.0&2625&18.6&82&1\\%
397&31.0&4&119.0&82.0&2720&19.4&82&1\\%
\hline%
\end{tabular}%
\vspace{2mm}%
\begin{lstlisting}[upquote=true]
Length before MPG outliers dropped: 398
Length after MPG outliers dropped: 388
\end{lstlisting}
\end{tcolorbox}

\subsection{Dropping Fields}%
\label{subsec:DroppingFields}%
You must drop fields that are of no value to the neural network.  The following code removes the name column from the MPG dataset.%
\index{dataset}%
\index{neural network}%
\par%
\begin{tcolorbox}[size=title,title=Code,breakable]%
\begin{lstlisting}[language=Python, upquote=true]
import os
import pandas as pd

df = pd.read_csv(
    "https://data.heatonresearch.com/data/t81-558/auto-mpg.csv",
    na_values=['NA','?'])

print(f"Before drop: {list(df.columns)}")
df.drop('name', 1, inplace=True)
print(f"After drop: {list(df.columns)}")\end{lstlisting}
\tcbsubtitle[before skip=\baselineskip]{Output}%
\begin{lstlisting}[upquote=true]
Before drop: ['mpg', 'cylinders', 'displacement', 'horsepower',
'weight', 'acceleration', 'year', 'origin', 'name']
After drop: ['mpg', 'cylinders', 'displacement', 'horsepower',
'weight', 'acceleration', 'year', 'origin']
\end{lstlisting}
\end{tcolorbox}

\subsection{Concatenating Rows and Columns}%
\label{subsec:ConcatenatingRowsandColumns}%
Python can concatenate rows and columns together to form new data frames.  The code below creates a new data frame from the%
\index{Python}%
\textbf{ name }%
and%
\textbf{ horsepower }%
columns from the Auto MPG dataset.  The program does this by concatenating two columns together.%
\index{dataset}%
\par%
\begin{tcolorbox}[size=title,title=Code,breakable]%
\begin{lstlisting}[language=Python, upquote=true]
# Create a new dataframe from name and horsepower

import os
import pandas as pd

df = pd.read_csv(
    "https://data.heatonresearch.com/data/t81-558/auto-mpg.csv",
    na_values=['NA','?'])

col_horsepower = df['horsepower']
col_name = df['name']
result = pd.concat([col_name, col_horsepower], axis=1)

pd.set_option('display.max_columns', 0)
pd.set_option('display.max_rows', 5)
display(result)\end{lstlisting}
\tcbsubtitle[before skip=\baselineskip]{Output}%
\begin{tabular}[hbt!]{l|l|l}%
\hline%
&name&horsepower\\%
\hline%
0&chevrolet chevelle malibu&130.0\\%
1&buick skylark 320&165.0\\%
...&...&...\\%
396&ford ranger&79.0\\%
397&chevy s{-}10&82.0\\%
\hline%
\end{tabular}%
\vspace{2mm}%
\end{tcolorbox}%
The%
\textbf{ concat }%
function can also concatenate rows together.  This code concatenates the first two rows and the last two rows of the Auto MPG dataset.%
\index{dataset}%
\par%
\begin{tcolorbox}[size=title,title=Code,breakable]%
\begin{lstlisting}[language=Python, upquote=true]
# Create a new dataframe from first 2 rows and last 2 rows

import os
import pandas as pd

df = pd.read_csv(
    "https://data.heatonresearch.com/data/t81-558/auto-mpg.csv",
    na_values=['NA','?'])

result = pd.concat([df[0:2],df[-2:]], axis=0)

pd.set_option('display.max_columns', 7)
pd.set_option('display.max_rows', 0)
display(result)\end{lstlisting}
\tcbsubtitle[before skip=\baselineskip]{Output}%
\begin{tabular}[hbt!]{l|l|l|l|l|l|l|l}%
\hline%
&mpg&cylinders&displacement&...&year&origin&name\\%
\hline%
0&18.0&8&307.0&...&70&1&chevrolet chevelle malibu\\%
1&15.0&8&350.0&...&70&1&buick skylark 320\\%
396&28.0&4&120.0&...&82&1&ford ranger\\%
397&31.0&4&119.0&...&82&1&chevy s{-}10\\%
\hline%
\end{tabular}%
\vspace{2mm}%
\end{tcolorbox}

\subsection{Training and Validation}%
\label{subsec:TrainingandValidation}%
We must evaluate a machine learning model based on its ability to predict values that it has never seen before. Because of this, we often divide the training data into a validation and training set. The machine learning model will learn from the training data but ultimately be evaluated based on the validation data.%
\index{learning}%
\index{model}%
\index{predict}%
\index{training}%
\index{validation}%
\par%
\begin{itemize}[noitemsep]%
\item%
\textbf{Training Data }%
{-}%
\textbf{ In Sample Data }%
{-} The data that the neural network used to train.%
\index{neural network}%
\item%
\textbf{Validation Data }%
{-}%
\textbf{ Out of Sample Data }%
{-} The data that the machine learning model is evaluated upon after it is fit to the training data.%
\index{learning}%
\index{model}%
\index{training}%
\end{itemize}%
There are two effective means of dealing with training and validation data:%
\index{training}%
\index{validation}%
\par
\begin{figure}[h]%
\centering%
\includegraphics[width=4in]{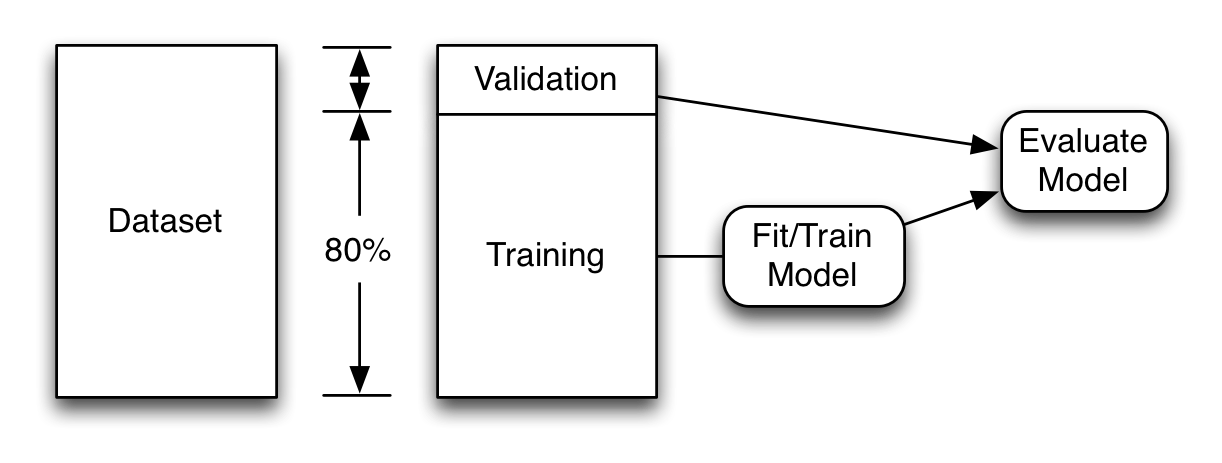}%
\caption{Training and Validation}%
\label{2.TRN-VAL}%
\end{figure}

%
\begin{itemize}[noitemsep]%
\item%
\textbf{Training/Validation Split }%
{-} The program splits the data according to some ratio between a training and validation (hold{-}out) set. Typical rates are 80\% training and 20\% validation.%
\index{SOM}%
\index{training}%
\index{validation}%
\item%
\textbf{K{-}Fold Cross Validation }%
{-} The program splits the data into several folds and models. Because the program creates the same number of models as folds, the program can generate out{-}of{-}sample predictions for the entire dataset.%
\index{dataset}%
\index{model}%
\index{predict}%
\end{itemize}%
The code below splits the MPG data into a training and validation set. The training set uses 80\% of the data, and the validation set uses 20\%. Figure \ref{2.TRN-VAL} shows how we train a model on 80\% of the data and then validated against the remaining 20\%.%
\index{model}%
\index{training}%
\index{validation}%
\par%

\par%
\begin{tcolorbox}[size=title,title=Code,breakable]%
\begin{lstlisting}[language=Python, upquote=true]
import os
import pandas as pd
import numpy as np

df = pd.read_csv(
    "https://data.heatonresearch.com/data/t81-558/auto-mpg.csv",
    na_values=['NA','?'])

# Usually a good idea to shuffle
df = df.reindex(np.random.permutation(df.index)) 

mask = np.random.rand(len(df)) < 0.8
trainDF = pd.DataFrame(df[mask])
validationDF = pd.DataFrame(df[~mask])

print(f"Training DF: {len(trainDF)}")
print(f"Validation DF: {len(validationDF)}")\end{lstlisting}
\tcbsubtitle[before skip=\baselineskip]{Output}%
\begin{lstlisting}[upquote=true]
Training DF: 333
Validation DF: 65
\end{lstlisting}
\end{tcolorbox}

\subsection{Converting a Dataframe to a Matrix}%
\label{subsec:ConvertingaDataframetoaMatrix}%
Neural networks do not directly operate on Python data frames.  A neural network requires a numeric matrix.  The program uses a data frame's%
\index{matrix}%
\index{neural network}%
\index{Python}%
\textbf{ values }%
property to convert the data to a matrix.%
\index{matrix}%
\par%
\begin{tcolorbox}[size=title,title=Code,breakable]%
\begin{lstlisting}[language=Python, upquote=true]
df.values\end{lstlisting}
\tcbsubtitle[before skip=\baselineskip]{Output}%
\begin{lstlisting}[upquote=true]
array([[20.2, 6, 232.0, ..., 79, 1, 'amc concord dl 6'],
       [14.0, 8, 304.0, ..., 74, 1, 'amc matador (sw)'],
       [14.0, 8, 351.0, ..., 71, 1, 'ford galaxie 500'],
       ...,
       [20.2, 6, 200.0, ..., 78, 1, 'ford fairmont (auto)'],
       [26.0, 4, 97.0, ..., 70, 2, 'volkswagen 1131 deluxe sedan'],
       [19.4, 6, 232.0, ..., 78, 1, 'amc concord']], dtype=object)
\end{lstlisting}
\end{tcolorbox}%
You might wish only to convert some of the columns, to leave out the name column, use the following code.%
\index{SOM}%
\par%
\begin{tcolorbox}[size=title,title=Code,breakable]%
\begin{lstlisting}[language=Python, upquote=true]
df[['mpg', 'cylinders', 'displacement', 'horsepower', 'weight',
       'acceleration', 'year', 'origin']].values\end{lstlisting}
\tcbsubtitle[before skip=\baselineskip]{Output}%
\begin{lstlisting}[upquote=true]
array([[ 20.2,   6. , 232. , ...,  18.2,  79. ,   1. ],
       [ 14. ,   8. , 304. , ...,  15.5,  74. ,   1. ],
       [ 14. ,   8. , 351. , ...,  13.5,  71. ,   1. ],
       ...,
       [ 20.2,   6. , 200. , ...,  15.8,  78. ,   1. ],
       [ 26. ,   4. ,  97. , ...,  20.5,  70. ,   2. ],
       [ 19.4,   6. , 232. , ...,  17.2,  78. ,   1. ]])
\end{lstlisting}
\end{tcolorbox}

\subsection{Saving a Dataframe to CSV}%
\label{subsec:SavingaDataframetoCSV}%
Many of the assignments in this course will require that you save a data frame to submit to the instructor.  The following code performs a shuffle and then saves a new copy.%
\par%
\begin{tcolorbox}[size=title,title=Code,breakable]%
\begin{lstlisting}[language=Python, upquote=true]
import os
import pandas as pd
import numpy as np

path = "."

df = pd.read_csv(
    "https://data.heatonresearch.com/data/t81-558/auto-mpg.csv",
    na_values=['NA','?'])

filename_write = os.path.join(path, "auto-mpg-shuffle.csv")
df = df.reindex(np.random.permutation(df.index))
# Specify index = false to not write row numbers
df.to_csv(filename_write, index=False)\end{lstlisting}
\tcbsubtitle[before skip=\baselineskip]{Output}%
\begin{lstlisting}[upquote=true]
Done
\end{lstlisting}
\end{tcolorbox}

\subsection{Saving a Dataframe to Pickle}%
\label{subsec:SavingaDataframetoPickle}%
A variety of software programs can use text files stored as CSV. However, they take longer to generate and can sometimes lose small amounts of precision in the conversion. Generally, you will output to CSV because it is very compatible, even outside of Python. Another format is%
\index{CSV}%
\index{output}%
\index{Python}%
\index{SOM}%
\href{https://docs.python.org/3/library/pickle.html}{ Pickle}%
. The code below stores the Dataframe to Pickle. Pickle stores data in the exact binary representation used by Python. The benefit is that there is no loss of data going to CSV format. The disadvantage is that generally, only Python programs can read Pickle files.%
\index{CSV}%
\index{Python}%
\par%
\begin{tcolorbox}[size=title,title=Code,breakable]%
\begin{lstlisting}[language=Python, upquote=true]
import os
import pandas as pd
import numpy as np
import pickle

path = "."

df = pd.read_csv(
    "https://data.heatonresearch.com/data/t81-558/auto-mpg.csv",
    na_values=['NA','?'])

filename_write = os.path.join(path, "auto-mpg-shuffle.pkl")
df = df.reindex(np.random.permutation(df.index))

with open(filename_write,"wb") as fp:
    pickle.dump(df, fp)\end{lstlisting}
\end{tcolorbox}%
Loading the pickle file back into memory is accomplished by the following lines of code.  Notice that the index numbers are still jumbled from the previous shuffle?  Loading the CSV rebuilt (in the last step) did not preserve these values.%
\index{CSV}%
\par%
\begin{tcolorbox}[size=title,title=Code,breakable]%
\begin{lstlisting}[language=Python, upquote=true]
import os
import pandas as pd
import numpy as np
import pickle

path = "."

df = pd.read_csv(
    "https://data.heatonresearch.com/data/t81-558/auto-mpg.csv",
    na_values=['NA','?'])

filename_read = os.path.join(path, "auto-mpg-shuffle.pkl")

with open(filename_write,"rb") as fp:
    df = pickle.load(fp)

pd.set_option('display.max_columns', 7)
pd.set_option('display.max_rows', 5)
display(df)\end{lstlisting}
\tcbsubtitle[before skip=\baselineskip]{Output}%
\begin{tabular}[hbt!]{l|l|l|l|l|l|l|l}%
\hline%
&mpg&cylinders&displacement&...&year&origin&name\\%
\hline%
387&38.0&6&262.0&...&82&1&oldsmobile cutlass ciera (diesel)\\%
361&25.4&6&168.0&...&81&3&toyota cressida\\%
...&...&...&...&...&...&...&...\\%
358&31.6&4&120.0&...&81&3&mazda 626\\%
237&30.5&4&98.0&...&77&1&chevrolet chevette\\%
\hline%
\end{tabular}%
\vspace{2mm}%
\end{tcolorbox}

\subsection{Module 2 Assignment}%
\label{subsec:Module2Assignment}%
You can find the first assignment here:%
\href{https://github.com/jeffheaton/t81_558_deep_learning/blob/master/assignments/assignment_yourname_class2.ipynb}{ assignment 2}%
\par

\section{Part 2.2: Categorical and Continuous Values}%
\label{sec:Part2.2CategoricalandContinuousValues}%
Neural networks require their input to be a fixed number of columns. This input format is very similar to spreadsheet data; it must be entirely numeric. It is essential to represent the data so that the neural network can train from it. Before we look at specific ways to preprocess data, it is important to consider four basic types of data, as defined by%
\index{input}%
\index{neural network}%
\index{ROC}%
\index{ROC}%
\cite{stevens1946theory}%
. Statisticians commonly refer to as the%
\href{https://en.wikipedia.org/wiki/Level_of_measurement}{ levels of measure}%
:%
\par%
\begin{itemize}[noitemsep]%
\item%
Character Data (strings)%
\begin{itemize}[noitemsep]%
\item%
\textbf{Nominal }%
{-} Individual discrete items, no order. For example, color, zip code, and shape.%
\item%
\textbf{Ordinal }%
{-} Individual distinct items have an implied order. For example, grade level, job title, Starbucks(tm) coffee size (tall, vente, grande)%
\item%
\end{itemize}%
Numeric Data%
\begin{itemize}[noitemsep]%
\item%
\textbf{Interval }%
{-} Numeric values, no defined start.  For example, temperature. You would never say, "yesterday was twice as hot as today."%
\item%
\textbf{Ratio }%
{-} Numeric values, clearly defined start.  For example, speed. You could say, "The first car is going twice as fast as the second."%
\end{itemize}%
\end{itemize}%
\subsection{Encoding Continuous Values}%
\label{subsec:EncodingContinuousValues}%
One common transformation is to normalize the inputs.  It is sometimes valuable to normalize numeric inputs in a standard form so that the program can easily compare these two values.  Consider if a friend told you that he received a 10{-}dollar discount.  Is this a good deal?  Maybe.  But the cost is not normalized.  If your friend purchased a car, the discount is not that good.  If your friend bought lunch, this is an excellent discount!%
\index{input}%
\index{SOM}%
\par%
Percentages are a prevalent form of normalization.  If your friend tells you they got 10\% off, we know that this is a better discount than 5\%.  It does not matter how much the purchase price was.  One widespread machine learning normalization is the Z{-}Score:%
\index{learning}%
\index{Z{-}Score}%
\par%
\vspace{2mm}%
\begin{equation*}
 z = \frac{x - \mu}{\sigma} 
\end{equation*}
\vspace{2mm}%
\par%
To calculate the Z{-}Score, you also need to calculate the mean($\mu$ or $\bar{x}$) and the standard deviation ($\sigma$).  You can calculate the mean with this equation:%
\index{standard deviation}%
\index{Z{-}Score}%
\par%
\vspace{2mm}%
\begin{equation*}
 \mu = \bar{x} = \frac{x_1+x_2+\cdots +x_n}{n} 
\end{equation*}
\vspace{2mm}%
\par%
The standard deviation is calculated as follows:%
\index{calculated}%
\index{standard deviation}%
\par%
\vspace{2mm}%
\begin{equation*}
 \sigma = \sqrt{\frac{1}{N} \sum_{i=1}^N (x_i - \mu)^2} 
\end{equation*}
\vspace{2mm}%
\par%
The following Python code replaces the mpg with a z{-}score.  Cars with average MPG will be near zero, above zero is above average, and below zero is below average.  Z{-}Scores more that 3 above or below are very rare; these are outliers.%
\index{Python}%
\index{Z{-}Score}%
\par%
\begin{tcolorbox}[size=title,title=Code,breakable]%
\begin{lstlisting}[language=Python, upquote=true]
import os
import pandas as pd
from scipy.stats import zscore

df = pd.read_csv(
    "https://data.heatonresearch.com/data/t81-558/auto-mpg.csv",
    na_values=['NA','?'])

pd.set_option('display.max_columns', 7)
pd.set_option('display.max_rows', 5)

df['mpg'] = zscore(df['mpg'])
display(df)\end{lstlisting}
\tcbsubtitle[before skip=\baselineskip]{Output}%
\begin{tabular}[hbt!]{l|l|l|l|l|l|l|l}%
\hline%
&mpg&cylinders&displacement&...&year&origin&name\\%
\hline%
0&{-}0.706439&8&307.0&...&70&1&chevrolet chevelle malibu\\%
1&{-}1.090751&8&350.0&...&70&1&buick skylark 320\\%
...&...&...&...&...&...&...&...\\%
396&0.574601&4&120.0&...&82&1&ford ranger\\%
397&0.958913&4&119.0&...&82&1&chevy s{-}10\\%
\hline%
\end{tabular}%
\vspace{2mm}%
\end{tcolorbox}

\subsection{Encoding Categorical Values as Dummies}%
\label{subsec:EncodingCategoricalValuesasDummies}%
The traditional means of encoding categorical values is to make them dummy variables.  This technique is also called one{-}hot{-}encoding.  Consider the following data set.%
\index{categorical}%
\par%
\begin{tcolorbox}[size=title,title=Code,breakable]%
\begin{lstlisting}[language=Python, upquote=true]
import pandas as pd

df = pd.read_csv(
    "https://data.heatonresearch.com/data/t81-558/jh-simple-dataset.csv",
    na_values=['NA','?'])

pd.set_option('display.max_columns', 7)
pd.set_option('display.max_rows', 5)

display(df)\end{lstlisting}
\tcbsubtitle[before skip=\baselineskip]{Output}%
\begin{tabular}[hbt!]{l|l|l|l|l|l|l|l}%
\hline%
&id&job&area&...&retail\_dense&crime&product\\%
\hline%
0&1&vv&c&...&0.492126&0.071100&b\\%
1&2&kd&c&...&0.342520&0.400809&c\\%
...&...&...&...&...&...&...&...\\%
1998&1999&qp&c&...&0.598425&0.117803&c\\%
1999&2000&pe&c&...&0.539370&0.451973&c\\%
\hline%
\end{tabular}%
\vspace{2mm}%
\end{tcolorbox}%
The%
\textit{ area }%
column is not numeric, so you must encode it with one{-}hot encoding. We display the number of areas and individual values. There are just four values in the%
\textit{ area }%
categorical variable in this case.%
\index{categorical}%
\par%
\begin{tcolorbox}[size=title,title=Code,breakable]%
\begin{lstlisting}[language=Python, upquote=true]
areas = list(df['area'].unique())
print(f'Number of areas: {len(areas)}')
print(f'Areas: {areas}')\end{lstlisting}
\tcbsubtitle[before skip=\baselineskip]{Output}%
\begin{lstlisting}[upquote=true]
Number of areas: 4
Areas: ['c', 'd', 'a', 'b']
\end{lstlisting}
\end{tcolorbox}%
There are four unique values in the%
\textit{ area }%
column.  To encode these dummy variables, we would use four columns, each representing one of the areas.  For each row, one column would have a value of one, the rest zeros.  For this reason, this type of encoding is sometimes called one{-}hot encoding.  The following code shows how you might encode the values "a" through "d."  The value A becomes {[}1,0,0,0{]} and the value B becomes {[}0,1,0,0{]}.%
\index{SOM}%
\par%
\begin{tcolorbox}[size=title,title=Code,breakable]%
\begin{lstlisting}[language=Python, upquote=true]
dummies = pd.get_dummies(['a','b','c','d'],prefix='area')
print(dummies)\end{lstlisting}
\tcbsubtitle[before skip=\baselineskip]{Output}%
\begin{lstlisting}[upquote=true]
area_a  area_b  area_c  area_d
0       1       0       0       0
1       0       1       0       0
2       0       0       1       0
3       0       0       0       1
\end{lstlisting}
\end{tcolorbox}%
We can now encode the actual column.%
\par%
\begin{tcolorbox}[size=title,title=Code,breakable]%
\begin{lstlisting}[language=Python, upquote=true]
dummies = pd.get_dummies(df['area'],prefix='area')
print(dummies[0:10]) # Just show the first 10\end{lstlisting}
\tcbsubtitle[before skip=\baselineskip]{Output}%
\begin{lstlisting}[upquote=true]
area_a  area_b  area_c  area_d
0        0       0       1       0
1        0       0       1       0
..     ...     ...     ...     ...
8        0       0       1       0
9        1       0       0       0
[10 rows x 4 columns]
\end{lstlisting}
\end{tcolorbox}%
For the new dummy/one hot encoded values to be of any use, they must be merged back into the data set.%
\par%
\begin{tcolorbox}[size=title,title=Code,breakable]%
\begin{lstlisting}[language=Python, upquote=true]
df = pd.concat([df,dummies],axis=1)\end{lstlisting}
\end{tcolorbox}%
To encode the%
\textit{ area }%
column, we use the following code. Note that it is necessary to merge these dummies back into the data frame.%
\par%
\begin{tcolorbox}[size=title,title=Code,breakable]%
\begin{lstlisting}[language=Python, upquote=true]
pd.set_option('display.max_columns', 0)
pd.set_option('display.max_rows', 10)

display(df[['id','job','area','income','area_a',
                  'area_b','area_c','area_d']])\end{lstlisting}
\tcbsubtitle[before skip=\baselineskip]{Output}%
\begin{tabular}[hbt!]{l|l|l|l|l|l|l|l|l}%
\hline%
&id&job&area&income&area\_a&area\_b&area\_c&area\_d\\%
\hline%
0&1&vv&c&50876.0&0&0&1&0\\%
1&2&kd&c&60369.0&0&0&1&0\\%
2&3&pe&c&55126.0&0&0&1&0\\%
3&4&11&c&51690.0&0&0&1&0\\%
4&5&kl&d&28347.0&0&0&0&1\\%
...&...&...&...&...&...&...&...&...\\%
1995&1996&vv&c&51017.0&0&0&1&0\\%
1996&1997&kl&d&26576.0&0&0&0&1\\%
1997&1998&kl&d&28595.0&0&0&0&1\\%
1998&1999&qp&c&67949.0&0&0&1&0\\%
1999&2000&pe&c&61467.0&0&0&1&0\\%
\hline%
\end{tabular}%
\vspace{2mm}%
\end{tcolorbox}%
Usually, you will remove the original column%
\textit{ area }%
because the goal is to get the data frame to be entirely numeric for the neural network.%
\index{neural network}%
\par%
\begin{tcolorbox}[size=title,title=Code,breakable]%
\begin{lstlisting}[language=Python, upquote=true]
pd.set_option('display.max_columns', 0)
pd.set_option('display.max_rows', 5)

df.drop('area', axis=1, inplace=True)
display(df[['id','job','income','area_a',
                  'area_b','area_c','area_d']])\end{lstlisting}
\tcbsubtitle[before skip=\baselineskip]{Output}%
\begin{tabular}[hbt!]{l|l|l|l|l|l|l|l}%
\hline%
&id&job&income&area\_a&area\_b&area\_c&area\_d\\%
\hline%
0&1&vv&50876.0&0&0&1&0\\%
1&2&kd&60369.0&0&0&1&0\\%
...&...&...&...&...&...&...&...\\%
1998&1999&qp&67949.0&0&0&1&0\\%
1999&2000&pe&61467.0&0&0&1&0\\%
\hline%
\end{tabular}%
\vspace{2mm}%
\end{tcolorbox}

\subsection{Removing the First Level}%
\label{subsec:RemovingtheFirstLevel}%
The%
\textbf{ pd.concat }%
function also includes a parameter named%
\index{parameter}%
\textit{ drop\_first}%
, which specifies whether to get k{-}1 dummies out of k categorical levels by removing the first level. Why would you want to remove the first level, in this case,%
\index{categorical}%
\textit{ area\_a}%
? This technique provides a more efficient encoding by using the ordinarily unused encoding of {[}0,0,0{]}. We encode the%
\textit{ area }%
to just three columns and map the categorical value of%
\index{categorical}%
\textit{ a }%
to {[}0,0,0{]}. The following code demonstrates this technique.%
\par%
\begin{tcolorbox}[size=title,title=Code,breakable]%
\begin{lstlisting}[language=Python, upquote=true]
import pandas as pd 

dummies = pd.get_dummies(['a','b','c','d'],prefix='area', drop_first=True)
print(dummies)\end{lstlisting}
\tcbsubtitle[before skip=\baselineskip]{Output}%
\begin{lstlisting}[upquote=true]
area_b  area_c  area_d
0       0       0       0
1       1       0       0
2       0       1       0
3       0       0       1
\end{lstlisting}
\end{tcolorbox}%
As you can see from the above data, the%
\textit{ area\_a }%
column is missing, as it%
\textbf{ get\_dummies }%
replaced it by the encoding of {[}0,0,0{]}. The following code shows how to apply this technique to a dataframe.%
\par%
\begin{tcolorbox}[size=title,title=Code,breakable]%
\begin{lstlisting}[language=Python, upquote=true]
import pandas as pd 

# Read the dataset
df = pd.read_csv(
    "https://data.heatonresearch.com/data/t81-558/jh-simple-dataset.csv",
    na_values=['NA','?'])

# encode the area column as dummy variables
dummies = pd.get_dummies(df['area'], drop_first=True, prefix='area')
df = pd.concat([df,dummies],axis=1)
df.drop('area', axis=1, inplace=True)

# display the encoded dataframe
pd.set_option('display.max_columns', 0)
pd.set_option('display.max_rows', 10)

display(df[['id','job','income',
                  'area_b','area_c','area_d']])\end{lstlisting}
\tcbsubtitle[before skip=\baselineskip]{Output}%
\begin{tabular}[hbt!]{l|l|l|l|l|l|l}%
\hline%
&id&job&income&area\_b&area\_c&area\_d\\%
\hline%
0&1&vv&50876.0&0&1&0\\%
1&2&kd&60369.0&0&1&0\\%
2&3&pe&55126.0&0&1&0\\%
3&4&11&51690.0&0&1&0\\%
4&5&kl&28347.0&0&0&1\\%
...&...&...&...&...&...&...\\%
1995&1996&vv&51017.0&0&1&0\\%
1996&1997&kl&26576.0&0&0&1\\%
1997&1998&kl&28595.0&0&0&1\\%
1998&1999&qp&67949.0&0&1&0\\%
1999&2000&pe&61467.0&0&1&0\\%
\hline%
\end{tabular}%
\vspace{2mm}%
\end{tcolorbox}

\subsection{Target Encoding for Categoricals}%
\label{subsec:TargetEncodingforCategoricals}%
Target encoding is a popular technique for Kaggle competitions. Target encoding can sometimes increase the predictive power of a machine learning model. However, it also dramatically increases the risk of overfitting. Because of this risk, you must take care of using this method.%
\index{Kaggle}%
\index{learning}%
\index{model}%
\index{overfitting}%
\index{predict}%
\index{SOM}%
\index{target encoding}%
\par%
Generally, target encoding can only be used on a categorical feature when the output of the machine learning model is numeric (regression).%
\index{categorical}%
\index{feature}%
\index{learning}%
\index{model}%
\index{output}%
\index{regression}%
\index{target encoding}%
\par%
The concept of target encoding is straightforward. For each category, we calculate the average target value for that category. Then to encode, we substitute the percent corresponding to the category that the categorical value has. Unlike dummy variables, where you have a column for each category with target encoding, the program only needs a single column. In this way, target coding is more efficient than dummy variables.%
\index{categorical}%
\index{target encoding}%
\par%
\begin{tcolorbox}[size=title,title=Code,breakable]%
\begin{lstlisting}[language=Python, upquote=true]
# Create a small sample dataset
import pandas as pd
import numpy as np

np.random.seed(43)
df = pd.DataFrame({
    'cont_9': np.random.rand(10)*100,
    'cat_0': ['dog'] * 5 + ['cat'] * 5,
    'cat_1': ['wolf'] * 9 + ['tiger'] * 1,
    'y': [1, 0, 1, 1, 1, 1, 0, 0, 0, 0]
})

pd.set_option('display.max_columns', 0)
pd.set_option('display.max_rows', 0)
display(df)\end{lstlisting}
\tcbsubtitle[before skip=\baselineskip]{Output}%
\begin{tabular}[hbt!]{l|l|l|l|l}%
\hline%
&cont\_9&cat\_0&cat\_1&y\\%
\hline%
0&11.505457&dog&wolf&1\\%
1&60.906654&dog&wolf&0\\%
2&13.339096&dog&wolf&1\\%
3&24.058962&dog&wolf&1\\%
4&32.713906&dog&wolf&1\\%
5&85.913749&cat&wolf&1\\%
6&66.609021&cat&wolf&0\\%
7&54.116221&cat&wolf&0\\%
8&2.901382&cat&wolf&0\\%
9&73.374830&cat&tiger&0\\%
\hline%
\end{tabular}%
\vspace{2mm}%
\end{tcolorbox}%
We want to change them to a number rather than creating dummy variables for "dog" and "cat," we would like to change them to a number. We could use 0 for a cat and 1 for a dog. However, we can encode more information than just that. The simple 0 or 1 would also only work for one animal. Consider what the mean target value is for cat and dog.%
\par%
\begin{tcolorbox}[size=title,title=Code,breakable]%
\begin{lstlisting}[language=Python, upquote=true]
means0 = df.groupby('cat_0')['y'].mean().to_dict()
means0\end{lstlisting}
\tcbsubtitle[before skip=\baselineskip]{Output}%
\begin{lstlisting}[upquote=true]
{'cat': 0.2, 'dog': 0.8}
\end{lstlisting}
\end{tcolorbox}%
The danger is that we are now using the target value ($y$) for training. This technique will potentially lead to overfitting. The possibility of overfitting is even greater if a small number of a particular category. To prevent this from happening, we use a weighting factor. The stronger the weight, the more categories with fewer values will tend towards the overall average of $y$. You can perform this calculation as follows.%
\index{overfitting}%
\index{training}%
\par%
\begin{tcolorbox}[size=title,title=Code,breakable]%
\begin{lstlisting}[language=Python, upquote=true]
df['y'].mean()\end{lstlisting}
\tcbsubtitle[before skip=\baselineskip]{Output}%
\begin{lstlisting}[upquote=true]
0.5
\end{lstlisting}
\end{tcolorbox}%
You can implement target encoding as follows.  For more information on Target Encoding, refer to the article%
\index{target encoding}%
\href{https://maxhalford.github.io/blog/target-encoding/}{ "Target Encoding Done the Right Way"}%
, that I based this code upon.%
\par%
\begin{tcolorbox}[size=title,title=Code,breakable]%
\begin{lstlisting}[language=Python, upquote=true]
def calc_smooth_mean(df1, df2, cat_name, target, weight):
    # Compute the global mean
    mean = df[target].mean()

    # Compute the number of values and the mean of each group
    agg = df.groupby(cat_name)[target].agg(['count', 'mean'])
    counts = agg['count']
    means = agg['mean']

    # Compute the "smoothed" means
    smooth = (counts * means + weight * mean) / (counts + weight)

    # Replace each value by the according smoothed mean
    if df2 is None:
        return df1[cat_name].map(smooth)
    else:
        return df1[cat_name].map(smooth),df2[cat_name].map(smooth.to_dict())\end{lstlisting}
\end{tcolorbox}%
The following code encodes these two categories.%
\par%
\begin{tcolorbox}[size=title,title=Code,breakable]%
\begin{lstlisting}[language=Python, upquote=true]
WEIGHT = 5
df['cat_0_enc'] = calc_smooth_mean(df1=df, df2=None, 
    cat_name='cat_0', target='y', weight=WEIGHT)
df['cat_1_enc'] = calc_smooth_mean(df1=df, df2=None, 
    cat_name='cat_1', target='y', weight=WEIGHT)

pd.set_option('display.max_columns', 0)
pd.set_option('display.max_rows', 0)

display(df)\end{lstlisting}
\tcbsubtitle[before skip=\baselineskip]{Output}%
\begin{tabular}[hbt!]{l|l|l|l|l|l|l}%
\hline%
&cont\_9&cat\_0&cat\_1&y&cat\_0\_enc&cat\_1\_enc\\%
\hline%
0&11.505457&dog&wolf&1&0.65&0.535714\\%
1&60.906654&dog&wolf&0&0.65&0.535714\\%
2&13.339096&dog&wolf&1&0.65&0.535714\\%
3&24.058962&dog&wolf&1&0.65&0.535714\\%
4&32.713906&dog&wolf&1&0.65&0.535714\\%
5&85.913749&cat&wolf&1&0.35&0.535714\\%
6&66.609021&cat&wolf&0&0.35&0.535714\\%
7&54.116221&cat&wolf&0&0.35&0.535714\\%
8&2.901382&cat&wolf&0&0.35&0.535714\\%
9&73.374830&cat&tiger&0&0.35&0.416667\\%
\hline%
\end{tabular}%
\vspace{2mm}%
\end{tcolorbox}

\subsection{Encoding Categorical Values as Ordinal}%
\label{subsec:EncodingCategoricalValuesasOrdinal}%
Typically categoricals will be encoded as dummy variables. However, there might be other techniques to convert categoricals to numeric. Any time there is an order to the categoricals, a number should be used. Consider if you had a categorical that described the current education level of an individual.%
\index{categorical}%
\par%
\begin{itemize}[noitemsep]%
\item%
Kindergarten (0)%
\item%
First Grade (1)%
\item%
Second Grade (2)%
\item%
Third Grade (3)%
\item%
Fourth Grade (4)%
\item%
Fifth Grade (5)%
\item%
Sixth Grade (6)%
\item%
Seventh Grade (7)%
\item%
Eighth Grade (8)%
\item%
High School Freshman (9)%
\item%
High School Sophomore (10)%
\item%
High School Junior (11)%
\item%
High School Senior (12)%
\item%
College Freshman (13)%
\item%
College Sophomore (14)%
\item%
College Junior (15)%
\item%
College Senior (16)%
\item%
Graduate Student (17)%
\item%
PhD Candidate (18)%
\item%
Doctorate (19)%
\item%
Post Doctorate (20)%
\end{itemize}%
The above list has 21 levels and would take 21 dummy variables to encode. However, simply encoding this to dummies would lose the order information. Perhaps the most straightforward approach would be to simply number them and assign the category a single number equal to the value in the parenthesis above. However, we might be able to do even better. A graduate student is likely more than a year so you might increase one value.%
\par

\subsection{High Cardinality Categorical}%
\label{subsec:HighCardinalityCategorical}%
If there were many, perhaps thousands or tens of thousands, then one{-}hot encoding is no longer a good choice. We call these cases high cardinality categorical. We generally encode such values with an embedding layer, which we will discuss later when introducing natural language processing (NLP).%
\index{categorical}%
\index{layer}%
\index{ROC}%
\index{ROC}%
\par

\section{Part 2.3: Grouping, Sorting, and Shuffling}%
\label{sec:Part2.3Grouping,Sorting,andShuffling}%
We will take a look at a few ways to affect an entire Pandas data frame. These techniques will allow us to group, sort, and shuffle data sets. These are all essential operations for both data preprocessing and evaluation.%
\index{ROC}%
\index{ROC}%
\par%
\subsection{Shuffling a Dataset}%
\label{subsec:ShufflingaDataset}%
There may be information lurking in the order of the rows of your dataset. Unless you are dealing with time{-}series data, the order of the rows should not be significant. Consider if your training set included employees in a company. Perhaps this dataset is ordered by the number of years the employees were with the company. It is okay to have an individual column that specifies years of service. However, having the data in this order might be problematic.%
\index{dataset}%
\index{time{-}series}%
\index{training}%
\par%
Consider if you were to split the data into training and validation. You could end up with your validation set having only the newer employees and the training set longer{-}term employees. Separating the data into a k{-}fold cross validation could have similar problems. Because of these issues, it is important to shuffle the data set.%
\index{k{-}fold}%
\index{training}%
\index{validation}%
\par%
Often shuffling and reindexing are both performed together. Shuffling randomizes the order of the data set. However, it does not change the Pandas row numbers. The following code demonstrates a reshuffle. Notice that the program has not reset the row indexes' first column. Generally, this will not cause any issues and allows tracing back to the original order of the data. However, I usually prefer to reset this index. I reason that I typically do not care about the initial position, and there are a few instances where this unordered index can cause issues.%
\index{random}%
\par%
\begin{tcolorbox}[size=title,title=Code,breakable]%
\begin{lstlisting}[language=Python, upquote=true]
import os
import pandas as pd
import numpy as np

df = pd.read_csv(
    "https://data.heatonresearch.com/data/t81-558/auto-mpg.csv", 
    na_values=['NA', '?'])

#np.random.seed(42) # Uncomment this line to get the same shuffle each time
df = df.reindex(np.random.permutation(df.index))

pd.set_option('display.max_columns', 7)
pd.set_option('display.max_rows', 5)
display(df)\end{lstlisting}
\tcbsubtitle[before skip=\baselineskip]{Output}%
\begin{tabular}[hbt!]{l|l|l|l|l|l|l|l}%
\hline%
&mpg&cylinders&displacement&...&year&origin&name\\%
\hline%
117&29.0&4&68.0&...&73&2&fiat 128\\%
245&36.1&4&98.0&...&78&1&ford fiesta\\%
...&...&...&...&...&...&...&...\\%
88&14.0&8&302.0&...&73&1&ford gran torino\\%
26&10.0&8&307.0&...&70&1&chevy c20\\%
\hline%
\end{tabular}%
\vspace{2mm}%
\end{tcolorbox}%
The following code demonstrates a reindex.  Notice how the reindex orders the row indexes.%
\par%
\begin{tcolorbox}[size=title,title=Code,breakable]%
\begin{lstlisting}[language=Python, upquote=true]
pd.set_option('display.max_columns', 7)
pd.set_option('display.max_rows', 5)

df.reset_index(inplace=True, drop=True)
display(df)\end{lstlisting}
\tcbsubtitle[before skip=\baselineskip]{Output}%
\begin{tabular}[hbt!]{l|l|l|l|l|l|l|l}%
\hline%
&mpg&cylinders&displacement&...&year&origin&name\\%
\hline%
0&29.0&4&68.0&...&73&2&fiat 128\\%
1&36.1&4&98.0&...&78&1&ford fiesta\\%
...&...&...&...&...&...&...&...\\%
396&14.0&8&302.0&...&73&1&ford gran torino\\%
397&10.0&8&307.0&...&70&1&chevy c20\\%
\hline%
\end{tabular}%
\vspace{2mm}%
\end{tcolorbox}

\subsection{Sorting a Data Set}%
\label{subsec:SortingaDataSet}%
While it is always good to shuffle a data set before training, during training and preprocessing, you may also wish to sort the data set. Sorting the data set allows you to order the rows in either ascending or descending order for one or more columns. The following code sorts the MPG dataset by name and displays the first car.%
\index{dataset}%
\index{ROC}%
\index{ROC}%
\index{training}%
\par%
\begin{tcolorbox}[size=title,title=Code,breakable]%
\begin{lstlisting}[language=Python, upquote=true]
import os
import pandas as pd

df = pd.read_csv(
    "https://data.heatonresearch.com/data/t81-558/auto-mpg.csv", 
    na_values=['NA', '?'])

df = df.sort_values(by='name', ascending=True)
print(f"The first car is: {df['name'].iloc[0]}")
      
pd.set_option('display.max_columns', 7)
pd.set_option('display.max_rows', 5)
display(df)\end{lstlisting}
\tcbsubtitle[before skip=\baselineskip]{Output}%
\begin{tabular}[hbt!]{l|l|l|l|l|l|l|l}%
\hline%
&mpg&cylinders&displacement&...&year&origin&name\\%
\hline%
96&13.0&8&360.0&...&73&1&amc ambassador brougham\\%
9&15.0&8&390.0&...&70&1&amc ambassador dpl\\%
...&...&...&...&...&...&...&...\\%
325&44.3&4&90.0&...&80&2&vw rabbit c (diesel)\\%
293&31.9&4&89.0&...&79&2&vw rabbit custom\\%
\hline%
\end{tabular}%
\vspace{2mm}%
\begin{lstlisting}[upquote=true]
The first car is: amc ambassador brougham
\end{lstlisting}
\end{tcolorbox}

\subsection{Grouping a Data Set}%
\label{subsec:GroupingaDataSet}%
Grouping is a typical operation on data sets.  Structured Query Language (SQL) calls this operation a "GROUP BY."  Programmers use grouping to summarize data.  Because of this, the summarization row count will usually shrink, and you cannot undo the grouping.  Because of this loss of information, it is essential to keep your original data before the grouping.%
\index{summarization}%
\par%
We use the Auto MPG dataset to demonstrate grouping.%
\index{dataset}%
\par%
\begin{tcolorbox}[size=title,title=Code,breakable]%
\begin{lstlisting}[language=Python, upquote=true]
import os
import pandas as pd

df = pd.read_csv(
    "https://data.heatonresearch.com/data/t81-558/auto-mpg.csv", 
    na_values=['NA', '?'])

pd.set_option('display.max_columns', 7)
pd.set_option('display.max_rows', 5)
display(df)\end{lstlisting}
\tcbsubtitle[before skip=\baselineskip]{Output}%
\begin{tabular}[hbt!]{l|l|l|l|l|l|l|l}%
\hline%
&mpg&cylinders&displacement&...&year&origin&name\\%
\hline%
0&18.0&8&307.0&...&70&1&chevrolet chevelle malibu\\%
1&15.0&8&350.0&...&70&1&buick skylark 320\\%
...&...&...&...&...&...&...&...\\%
396&28.0&4&120.0&...&82&1&ford ranger\\%
397&31.0&4&119.0&...&82&1&chevy s{-}10\\%
\hline%
\end{tabular}%
\vspace{2mm}%
\end{tcolorbox}%
You can use the above data set with the group to perform summaries.  For example, the following code will group cylinders by the average (mean).  This code will provide the grouping.  In addition to%
\textbf{ mean}%
, you can use other aggregating functions, such as%
\textbf{ sum }%
or%
\textbf{ count}%
.%
\par%
\begin{tcolorbox}[size=title,title=Code,breakable]%
\begin{lstlisting}[language=Python, upquote=true]
g = df.groupby('cylinders')['mpg'].mean()
g\end{lstlisting}
\tcbsubtitle[before skip=\baselineskip]{Output}%
\begin{lstlisting}[upquote=true]
cylinders
3    20.550000
4    29.286765
5    27.366667
6    19.985714
8    14.963107
Name: mpg, dtype: float64
\end{lstlisting}
\end{tcolorbox}%
It might be useful to have these%
\textbf{ mean }%
values as a dictionary.%
\par%
\begin{tcolorbox}[size=title,title=Code,breakable]%
\begin{lstlisting}[language=Python, upquote=true]
d = g.to_dict()
d\end{lstlisting}
\tcbsubtitle[before skip=\baselineskip]{Output}%
\begin{lstlisting}[upquote=true]
{3: 20.55,
 4: 29.28676470588236,
 5: 27.366666666666664,
 6: 19.985714285714284,
 8: 14.963106796116508}
\end{lstlisting}
\end{tcolorbox}%
A dictionary allows you to access an individual element quickly.  For example, you could quickly look up the mean for six{-}cylinder cars.  You will see that target encoding, introduced later in this module, uses this technique.%
\index{target encoding}%
\par%
\begin{tcolorbox}[size=title,title=Code,breakable]%
\begin{lstlisting}[language=Python, upquote=true]
d[6]\end{lstlisting}
\tcbsubtitle[before skip=\baselineskip]{Output}%
\begin{lstlisting}[upquote=true]
19.985714285714284
\end{lstlisting}
\end{tcolorbox}%
The code below shows how to count the number of rows that match each cylinder count.%
\par%
\begin{tcolorbox}[size=title,title=Code,breakable]%
\begin{lstlisting}[language=Python, upquote=true]
df.groupby('cylinders')['mpg'].count().to_dict()\end{lstlisting}
\tcbsubtitle[before skip=\baselineskip]{Output}%
\begin{lstlisting}[upquote=true]
{3: 4, 4: 204, 5: 3, 6: 84, 8: 103}
\end{lstlisting}
\end{tcolorbox}

\section{Part 2.4: Apply and Map}%
\label{sec:Part2.4ApplyandMap}%
If you've ever worked with Big Data or functional programming languages before, you've likely heard of map/reduce. Map and reduce are two functions that apply a task you create to a data frame. Pandas supports functional programming techniques that allow you to use functions across en entire data frame. In addition to functions that you write, Pandas also provides several standard functions for use with data frames.%
\par%
\subsection{Using Map with Dataframes}%
\label{subsec:UsingMapwithDataframes}%
The map function allows you to transform a column by mapping certain values in that column to other values. Consider the Auto MPG data set that contains a field%
\textbf{ origin\_name }%
that holds a value between one and three that indicates the geographic origin of each car. We can see how to use the map function to transform this numeric origin into the textual name of each origin.%
\par%
We will begin by loading the Auto MPG data set.%
\par%
\begin{tcolorbox}[size=title,title=Code,breakable]%
\begin{lstlisting}[language=Python, upquote=true]
import os
import pandas as pd
import numpy as np

df = pd.read_csv(
    "https://data.heatonresearch.com/data/t81-558/auto-mpg.csv", 
    na_values=['NA', '?'])

pd.set_option('display.max_columns', 7)
pd.set_option('display.max_rows', 5)

display(df)\end{lstlisting}
\tcbsubtitle[before skip=\baselineskip]{Output}%
\begin{tabular}[hbt!]{l|l|l|l|l|l|l|l}%
\hline%
&mpg&cylinders&displacement&...&year&origin&name\\%
\hline%
0&18.0&8&307.0&...&70&1&chevrolet chevelle malibu\\%
1&15.0&8&350.0&...&70&1&buick skylark 320\\%
...&...&...&...&...&...&...&...\\%
396&28.0&4&120.0&...&82&1&ford ranger\\%
397&31.0&4&119.0&...&82&1&chevy s{-}10\\%
\hline%
\end{tabular}%
\vspace{2mm}%
\end{tcolorbox}%
The%
\textbf{ map }%
method in Pandas operates on a single column.  You provide%
\textbf{ map }%
with a dictionary of values to transform the target column.  The map keys specify what values in the target column should be turned into values specified by those keys.  The following code shows how the map function can transform the numeric values of 1, 2, and 3 into the string values of North America, Europe, and Asia.%
\par%
\begin{tcolorbox}[size=title,title=Code,breakable]%
\begin{lstlisting}[language=Python, upquote=true]
# Apply the map
df['origin_name'] = df['origin'].map(
    {1: 'North America', 2: 'Europe', 3: 'Asia'})

# Shuffle the data, so that we hopefully see
# more regions.
df = df.reindex(np.random.permutation(df.index)) 

# Display
pd.set_option('display.max_columns', 7)
pd.set_option('display.max_rows', 10)
display(df)\end{lstlisting}
\tcbsubtitle[before skip=\baselineskip]{Output}%
\begin{tabular}[hbt!]{l|l|l|l|l|l|l|l}%
\hline%
&mpg&cylinders&displacement&...&origin&name&origin\_name\\%
\hline%
45&18.0&6&258.0&...&1&amc hornet sportabout (sw)&North America\\%
290&15.5&8&351.0&...&1&ford country squire (sw)&North America\\%
313&28.0&4&151.0&...&1&chevrolet citation&North America\\%
82&23.0&4&120.0&...&3&toyouta corona mark ii (sw)&Asia\\%
33&19.0&6&232.0&...&1&amc gremlin&North America\\%
...&...&...&...&...&...&...&...\\%
329&44.6&4&91.0&...&3&honda civic 1500 gl&Asia\\%
326&43.4&4&90.0&...&2&vw dasher (diesel)&Europe\\%
34&16.0&6&225.0&...&1&plymouth satellite custom&North America\\%
118&24.0&4&116.0&...&2&opel manta&Europe\\%
15&22.0&6&198.0&...&1&plymouth duster&North America\\%
\hline%
\end{tabular}%
\vspace{2mm}%
\end{tcolorbox}

\subsection{Using Apply with Dataframes}%
\label{subsec:UsingApplywithDataframes}%
The%
\textbf{ apply }%
function of the data frame can run a function over the entire data frame. You can use either a traditional named function or a lambda function. Python will execute the provided function against each of the rows or columns in the data frame. The%
\index{Python}%
\textbf{ axis }%
parameter specifies that the function is run across rows or columns. For axis = 1, rows are used. The following code calculates a series called%
\index{axis}%
\index{parameter}%
\textbf{ efficiency }%
that is the%
\textbf{ displacement }%
divided by%
\textbf{ horsepower}%
.%
\par%
\begin{tcolorbox}[size=title,title=Code,breakable]%
\begin{lstlisting}[language=Python, upquote=true]
efficiency = df.apply(lambda x: x['displacement']/x['horsepower'], axis=1)
display(efficiency[0:10])\end{lstlisting}
\tcbsubtitle[before skip=\baselineskip]{Output}%
\begin{lstlisting}[upquote=true]
45     2.345455
290    2.471831
313    1.677778
82     1.237113
33     2.320000
249    2.363636
27     1.514286
7      2.046512
302    1.500000
179    1.234694
dtype: float64
\end{lstlisting}
\end{tcolorbox}%
You can now insert this series into the data frame, either as a new column or to replace an existing column.  The following code inserts this new series into the data frame.%
\par%
\begin{tcolorbox}[size=title,title=Code,breakable]%
\begin{lstlisting}[language=Python, upquote=true]
df['efficiency'] = efficiency\end{lstlisting}
\end{tcolorbox}

\subsection{Feature Engineering with Apply and Map}%
\label{subsec:FeatureEngineeringwithApplyandMap}%
In this section, we will see how to calculate a complex feature using map, apply, and grouping. The data set is the following CSV:%
\index{CSV}%
\index{feature}%
\par%
\begin{itemize}[noitemsep]%
\item%
https://www.irs.gov/pub/irs{-}soi/16zpallagi.csv%
\index{CSV}%
\end{itemize}%
This URL contains US Government public data for "SOI Tax Stats {-} Individual Income Tax Statistics."  The entry point to the website is here:%
\par%
\begin{itemize}[noitemsep]%
\item%
https://www.irs.gov/statistics/soi{-}tax{-}stats{-}individual{-}income{-}tax{-}statistics{-}2016{-}zip{-}code{-}data{-}soi%
\end{itemize}%
Documentation describing this data is at the above link.%
\index{link}%
\par%
For this feature, we will attempt to estimate the adjusted gross income (AGI) for each of the zip codes. The data file contains many columns; however, you will only use the following:%
\index{feature}%
\par%
\begin{itemize}[noitemsep]%
\item%
\textbf{STATE }%
{-} The state (e.g., MO)%
\item%
\textbf{zipcode }%
{-} The zipcode (e.g. 63017)%
\item%
\textbf{agi\_stub }%
{-} Six different brackets of annual income (1 through 6)%
\item%
\textbf{N1 }%
{-} The number of tax returns for each of the agi\_stubs%
\end{itemize}%
Note, that the file will have six rows for each zip code for each of the agi\_stub brackets. You can skip zip codes with 0 or 99999.%
\par%
We will create an output CSV with these columns; however, only one row per zip code. Calculate a weighted average of the income brackets. For example, the following six rows are present for 63017:%
\index{CSV}%
\index{output}%
\par%
\begin{tabular}[hbt!]{l|l|l|l|l}%
\hline%
&zipcode &agi\_stub & N1 &\\%
&{-}{-}&{-}{-}&{-}{-} &\\%
&63017     &1 & 4710 &\\%
&63017     &2 & 2780 &\\%
&63017     &3 & 2130 &\\%
&63017     &4 & 2010 &\\%
&63017     &5 & 5240 &\\%
&63017     &6 & 3510 &\\%
\hline%
\end{tabular}%
\vspace{2mm}%
\par%
We must combine these six rows into one.  For privacy reasons, AGI's are broken out into 6 buckets.  We need to combine the buckets and estimate the actual AGI of a zipcode. To do this, consider the values for N1:%
\par%
\begin{itemize}[noitemsep]%
\item%
1 = 1 to 25,000%
\item%
2 = 25,000 to 50,000%
\item%
3 = 50,000 to 75,000%
\item%
4 = 75,000 to 100,000%
\item%
5 = 100,000 to 200,000%
\item%
6 = 200,000 or more%
\end{itemize}%
The median of each of these ranges is approximately:%
\par%
\begin{itemize}[noitemsep]%
\item%
1 = 12,500%
\item%
2 = 37,500%
\item%
3 = 62,500%
\item%
4 = 87,500%
\item%
5 = 112,500%
\item%
6 = 212,500%
\end{itemize}%
Using this, you can estimate 63017's average AGI as:%
\par%
\begin{tcolorbox}[size=title,breakable]%
\begin{lstlisting}[upquote=true]
>>> totalCount = 4710 + 2780 + 2130 + 2010 + 5240 + 3510
>>> totalAGI = 4710 * 12500 + 2780 * 37500 + 2130 * 62500 
    + 2010 * 87500 + 5240 * 112500 + 3510 * 212500
>>> print(totalAGI / totalCount)

88689.89205103042
\end{lstlisting}
\end{tcolorbox}%
We begin by reading the government data.%
\par%
\begin{tcolorbox}[size=title,title=Code,breakable]%
\begin{lstlisting}[language=Python, upquote=true]
import pandas as pd

df=pd.read_csv('https://www.irs.gov/pub/irs-soi/16zpallagi.csv')\end{lstlisting}
\end{tcolorbox}%
First, we trim all zip codes that are either 0 or 99999.  We also select the three fields that we need.%
\par%
\begin{tcolorbox}[size=title,title=Code,breakable]%
\begin{lstlisting}[language=Python, upquote=true]
df=df.loc[(df['zipcode']!=0) & (df['zipcode']!=99999),
          ['STATE','zipcode','agi_stub','N1']]

pd.set_option('display.max_columns', 0)
pd.set_option('display.max_rows', 10)

display(df)\end{lstlisting}
\tcbsubtitle[before skip=\baselineskip]{Output}%
\begin{tabular}[hbt!]{l|l|l|l|l}%
\hline%
&STATE&zipcode&agi\_stub&N1\\%
\hline%
6&AL&35004&1&1510\\%
7&AL&35004&2&1410\\%
8&AL&35004&3&950\\%
9&AL&35004&4&650\\%
10&AL&35004&5&630\\%
...&...&...&...&...\\%
179785&WY&83414&2&40\\%
179786&WY&83414&3&40\\%
179787&WY&83414&4&0\\%
179788&WY&83414&5&40\\%
179789&WY&83414&6&30\\%
\hline%
\end{tabular}%
\vspace{2mm}%
\end{tcolorbox}%
We replace all of the%
\textbf{ agi\_stub }%
values with the correct median values with the%
\textbf{ map }%
function.%
\par%
\begin{tcolorbox}[size=title,title=Code,breakable]%
\begin{lstlisting}[language=Python, upquote=true]
medians = {1:12500,2:37500,3:62500,4:87500,5:112500,6:212500}
df['agi_stub']=df.agi_stub.map(medians)

pd.set_option('display.max_columns', 0)
pd.set_option('display.max_rows', 10)
display(df)\end{lstlisting}
\tcbsubtitle[before skip=\baselineskip]{Output}%
\begin{tabular}[hbt!]{l|l|l|l|l}%
\hline%
&STATE&zipcode&agi\_stub&N1\\%
\hline%
6&AL&35004&12500&1510\\%
7&AL&35004&37500&1410\\%
8&AL&35004&62500&950\\%
9&AL&35004&87500&650\\%
10&AL&35004&112500&630\\%
...&...&...&...&...\\%
179785&WY&83414&37500&40\\%
179786&WY&83414&62500&40\\%
179787&WY&83414&87500&0\\%
179788&WY&83414&112500&40\\%
179789&WY&83414&212500&30\\%
\hline%
\end{tabular}%
\vspace{2mm}%
\end{tcolorbox}%
Next, we group the data frame by zip code.%
\par%
\begin{tcolorbox}[size=title,title=Code,breakable]%
\begin{lstlisting}[language=Python, upquote=true]
groups = df.groupby(by='zipcode')\end{lstlisting}
\end{tcolorbox}%
The program applies a lambda across the groups and calculates the AGI estimate.%
\par%
\begin{tcolorbox}[size=title,title=Code,breakable]%
\begin{lstlisting}[language=Python, upquote=true]
df = pd.DataFrame(groups.apply( 
    lambda x:sum(x['N1']*x['agi_stub'])/sum(x['N1']))) \
    .reset_index()

pd.set_option('display.max_columns', 0)
pd.set_option('display.max_rows', 10)

display(df)\end{lstlisting}
\tcbsubtitle[before skip=\baselineskip]{Output}%
\begin{tabular}[hbt!]{l|l|l}%
\hline%
&zipcode&0\\%
\hline%
0&1001&52895.322940\\%
1&1002&64528.451001\\%
2&1003&15441.176471\\%
3&1005&54694.092827\\%
4&1007&63654.353562\\%
...&...&...\\%
29867&99921&48042.168675\\%
29868&99922&32954.545455\\%
29869&99925&45639.534884\\%
29870&99926&41136.363636\\%
29871&99929&45911.214953\\%
\hline%
\end{tabular}%
\vspace{2mm}%
\end{tcolorbox}%
We can now rename the new%
\textbf{ agi\_estimate }%
column.%
\par%
\begin{tcolorbox}[size=title,title=Code,breakable]%
\begin{lstlisting}[language=Python, upquote=true]
df.columns = ['zipcode','agi_estimate']

pd.set_option('display.max_columns', 0)
pd.set_option('display.max_rows', 10)

display(df)\end{lstlisting}
\tcbsubtitle[before skip=\baselineskip]{Output}%
\begin{tabular}[hbt!]{l|l|l}%
\hline%
&zipcode&agi\_estimate\\%
\hline%
0&1001&52895.322940\\%
1&1002&64528.451001\\%
2&1003&15441.176471\\%
3&1005&54694.092827\\%
4&1007&63654.353562\\%
...&...&...\\%
29867&99921&48042.168675\\%
29868&99922&32954.545455\\%
29869&99925&45639.534884\\%
29870&99926&41136.363636\\%
29871&99929&45911.214953\\%
\hline%
\end{tabular}%
\vspace{2mm}%
\end{tcolorbox}%
Finally, we check to see that our zip code of 63017 got the correct value.%
\par%
\begin{tcolorbox}[size=title,title=Code,breakable]%
\begin{lstlisting}[language=Python, upquote=true]
df[ df['zipcode']==63017 ]\end{lstlisting}
\tcbsubtitle[before skip=\baselineskip]{Output}%
\begin{tabular}[hbt!]{l|l|l}%
\hline%
&zipcode&agi\_estimate\\%
\hline%
19909&63017&88689.892051\\%
\hline%
\end{tabular}%
\vspace{2mm}%
\end{tcolorbox}

\section{Part 2.5: Feature Engineering}%
\label{sec:Part2.5FeatureEngineering}%
Feature engineering is an essential part of machine learning.  For now, we will manually engineer features.  However, later in this course, we will see some techniques for automatic feature engineering.%
\index{feature}%
\index{learning}%
\index{SOM}%
\par%
\subsection{Calculated Fields}%
\label{subsec:CalculatedFields}%
It is possible to add new fields to the data frame that your program calculates from the other fields.  We can create a new column that gives the weight in kilograms.  The equation to calculate a metric weight, given weight in pounds, is:%
\par%
\vspace{2mm}%
\begin{equation*}
 m_{(kg)} = m_{(lb)} \times 0.45359237 
\end{equation*}
\vspace{2mm}%
\par%
The following Python code performs this transformation:%
\index{Python}%
\par%
\begin{tcolorbox}[size=title,title=Code,breakable]%
\begin{lstlisting}[language=Python, upquote=true]
import os
import pandas as pd

df = pd.read_csv(
    "https://data.heatonresearch.com/data/t81-558/auto-mpg.csv", 
    na_values=['NA', '?'])

df.insert(1, 'weight_kg', (df['weight'] * 0.45359237).astype(int))
pd.set_option('display.max_columns', 6)
pd.set_option('display.max_rows', 5)
df\end{lstlisting}
\tcbsubtitle[before skip=\baselineskip]{Output}%
\begin{tabular}[hbt!]{l|l|l|l|l|l|l|l}%
\hline%
&mpg&weight\_kg&cylinders&...&year&origin&name\\%
\hline%
0&18.0&1589&8&...&70&1&chevrolet chevelle malibu\\%
1&15.0&1675&8&...&70&1&buick skylark 320\\%
...&...&...&...&...&...&...&...\\%
396&28.0&1190&4&...&82&1&ford ranger\\%
397&31.0&1233&4&...&82&1&chevy s{-}10\\%
\hline%
\end{tabular}%
\vspace{2mm}%
\end{tcolorbox}

\subsection{Google API Keys}%
\label{subsec:GoogleAPIKeys}%
Sometimes you will use external APIs to obtain data.  The following examples show how to use the Google API keys to encode addresses for use with neural networks.  To use these, you will need your own Google API key.  The key I have below is not a real key; you need to put your own there.  Google will ask for a credit card, but there will be no actual cost unless you use a massive number of lookups.  YOU ARE NOT required to get a Google API key for this class; this only shows you how.  If you want to get a Google API key, visit this site and obtain one for%
\index{neural network}%
\index{SOM}%
\textbf{ geocode}%
.%
\par%
You can obtain your key from this link:%
\index{link}%
\href{https://developers.google.com/maps/documentation/embed/get-api-key}{ Google API Keys}%
.%
\par%
\begin{tcolorbox}[size=title,title=Code,breakable]%
\begin{lstlisting}[language=Python, upquote=true]
if 'GOOGLE_API_KEY' in os.environ:
    # If the API key is defined in an environmental variable,
    # the use the env variable.
    GOOGLE_KEY = os.environ['GOOGLE_API_KEY']    
else:
    # If you have a Google API key of your own, you can also just
    # put it here:
    GOOGLE_KEY = 'REPLACE WITH YOUR GOOGLE API KEY'\end{lstlisting}
\end{tcolorbox}

\subsection{Other Examples: Dealing with Addresses}%
\label{subsec:OtherExamplesDealingwithAddresses}%
Addresses can be difficult to encode into a neural network.  There are many different approaches, and you must consider how you can transform the address into something more meaningful.  Map coordinates can be a good approach.%
\index{neural network}%
\index{SOM}%
\href{https://en.wikipedia.org/wiki/Geographic_coordinate_system}{ latitude and longitude }%
can be a useful encoding.  Thanks to the power of the Internet, it is relatively easy to transform an address into its latitude and longitude values.  The following code determines the coordinates of%
\href{https://wustl.edu/}{ Washington University}%
:%
\par
\begin{figure}[h]%
\centering%
\includegraphics[width=4in]{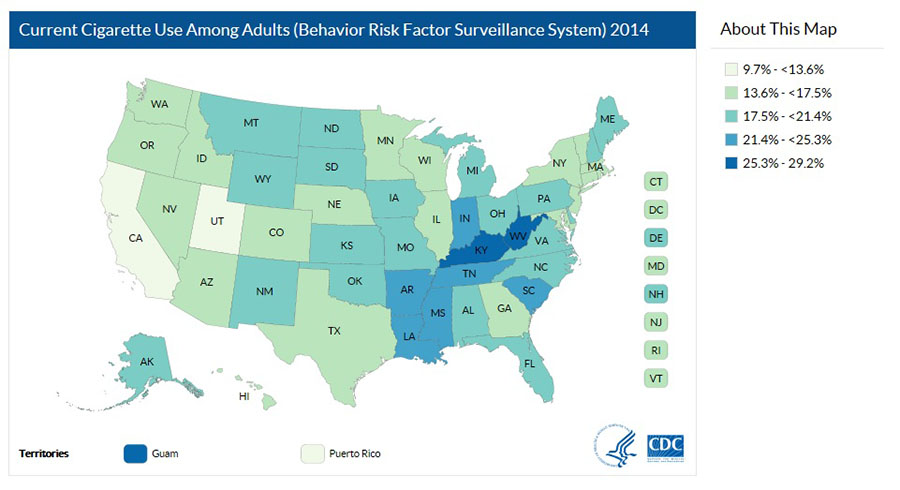}%
\caption{Smokers by State}%
\label{2.SMK}%
\end{figure}

%
\begin{tcolorbox}[size=title,title=Code,breakable]%
\begin{lstlisting}[language=Python, upquote=true]
import requests

address = "1 Brookings Dr, St. Louis, MO 63130"

response = requests.get(
    'https://maps.googleapis.com/maps/api/geocode/json?key={}&address={}' \
    .format(GOOGLE_KEY,address))

resp_json_payload = response.json()

if 'error_message' in resp_json_payload:
    print(resp_json_payload['error_message'])
else:
    print(resp_json_payload['results'][0]['geometry']['location'])\end{lstlisting}
\tcbsubtitle[before skip=\baselineskip]{Output}%
\begin{lstlisting}[upquote=true]
{'lat': 38.6481653, 'lng': -90.3049506}
\end{lstlisting}
\end{tcolorbox}%
They might not be overly helpful if you feed latitude and longitude into the neural network as two features.  These two values would allow your neural network to cluster locations on a map.  Sometimes cluster locations on a map can be useful.  Figure \ref{2.SMK} shows the percentage of the population that smokes in the USA by state.%
\index{feature}%
\index{neural network}%
\index{SOM}%
\par%

\par%
The above map shows that certain behaviors, like smoking, can be clustered by the global region.%
\index{clustered}%
\par%
However, often you will want to transform the coordinates into distances.  It is reasonably easy to estimate the distance between any two points on Earth by using the%
\href{https://en.wikipedia.org/wiki/Great-circle_distance}{ great circle distance }%
between any two points on a sphere:%
\par%
The following code implements this formula:%
\par%
\vspace{2mm}%
\begin{equation*}
 \Delta\sigma=\arccos\bigl(\sin\phi_1\cdot\sin\phi_2+\cos\phi_1\cdot\cos\phi_2\cdot\cos(\Delta\lambda)\bigr) 
\end{equation*}
\vspace{2mm}%
\index{delta}%
\par%
\vspace{2mm}%
\begin{equation*}
 d = r , \Delta\sigma 
\end{equation*}
\vspace{2mm}%
\index{delta}%
\par%
\begin{tcolorbox}[size=title,title=Code,breakable]%
\begin{lstlisting}[language=Python, upquote=true]
from math import sin, cos, sqrt, atan2, radians

URL='https://maps.googleapis.com' + \
    '/maps/api/geocode/json?key={}&address={}'

# Distance function
def distance_lat_lng(lat1,lng1,lat2,lng2):
    # approximate radius of earth in km
    R = 6373.0

    # degrees to radians (lat/lon are in degrees)
    lat1 = radians(lat1)
    lng1 = radians(lng1)
    lat2 = radians(lat2)
    lng2 = radians(lng2)

    dlng = lng2 - lng1
    dlat = lat2 - lat1

    a = sin(dlat / 2)**2 + cos(lat1) * cos(lat2) * sin(dlng / 2)**2
    c = 2 * atan2(sqrt(a), sqrt(1 - a))

    return R * c

# Find lat lon for address
def lookup_lat_lng(address):
    response = requests.get( \
        URL.format(GOOGLE_KEY,address))
    json = response.json()
    if len(json['results']) == 0:
        raise ValueError("Google API error on: {}".format(address))
    map = json['results'][0]['geometry']['location']
    return map['lat'],map['lng']


# Distance between two locations

import requests

address1 = "1 Brookings Dr, St. Louis, MO 63130" 
address2 = "3301 College Ave, Fort Lauderdale, FL 33314"

lat1, lng1 = lookup_lat_lng(address1)
lat2, lng2 = lookup_lat_lng(address2)

print("Distance, St. Louis, MO to Ft. Lauderdale, FL: {} km".format(
        distance_lat_lng(lat1,lng1,lat2,lng2)))\end{lstlisting}
\tcbsubtitle[before skip=\baselineskip]{Output}%
\begin{lstlisting}[upquote=true]
Distance, St. Louis, MO to Ft. Lauderdale, FL: 1685.3019808607426 km
\end{lstlisting}
\end{tcolorbox}%
Distances can be a useful means to encode addresses.  It would help if you considered what distance might be helpful for your dataset.  Consider:%
\index{dataset}%
\par%
\begin{itemize}[noitemsep]%
\item%
Distance to a major metropolitan area%
\item%
Distance to a competitor%
\item%
Distance to a distribution center%
\item%
Distance to a retail outlet%
\end{itemize}%
The following code calculates the distance between 10 universities and Washington University in St. Louis:%
\par%
\begin{tcolorbox}[size=title,title=Code,breakable]%
\begin{lstlisting}[language=Python, upquote=true]
# Encoding other universities by their distance to Washington University

schools = [
    ["Princeton University, Princeton, NJ 08544", 'Princeton'],
    ["Massachusetts Hall, Cambridge, MA 02138", 'Harvard'],
    ["5801 S Ellis Ave, Chicago, IL 60637", 'University of Chicago'],
    ["Yale, New Haven, CT 06520", 'Yale'],
    ["116th St & Broadway, New York, NY 10027", 'Columbia University'],
    ["450 Serra Mall, Stanford, CA 94305", 'Stanford'],
    ["77 Massachusetts Ave, Cambridge, MA 02139", 'MIT'],
    ["Duke University, Durham, NC 27708", 'Duke University'],
    ["University of Pennsylvania, Philadelphia, PA 19104", 
         'University of Pennsylvania'],
    ["Johns Hopkins University, Baltimore, MD 21218", 'Johns Hopkins']
]

lat1, lng1 = lookup_lat_lng("1 Brookings Dr, St. Louis, MO 63130")

for address, name in schools:
    lat2,lng2 = lookup_lat_lng(address)
    dist = distance_lat_lng(lat1,lng1,lat2,lng2)
    print("School '{}', distance to wustl is: {}".format(name,dist))\end{lstlisting}
\tcbsubtitle[before skip=\baselineskip]{Output}%
\begin{lstlisting}[upquote=true]
School 'Princeton', distance to wustl is: 1354.4830895052746
School 'Harvard', distance to wustl is: 1670.6297027161022
School 'University of Chicago', distance to wustl is:
418.0815972177934
School 'Yale', distance to wustl is: 1508.217831712127
School 'Columbia University', distance to wustl is: 1418.2264083295695
School 'Stanford', distance to wustl is: 2780.6829398114114
School 'MIT', distance to wustl is: 1672.4444489665696
School 'Duke University', distance to wustl is: 1046.7970984423719
School 'University of Pennsylvania', distance to wustl is:
1307.19541200423
School 'Johns Hopkins', distance to wustl is: 1184.3831076555425
\end{lstlisting}
\end{tcolorbox}

\chapter{Introduction to TensorFlow}%
\label{chap:IntroductiontoTensorFlow}%
\section{Part 3.1: Deep Learning and Neural Network Introduction}%
\label{sec:Part3.1DeepLearningandNeuralNetworkIntroduction}%
Neural networks were one of the first machine learning models. Their popularity has fallen twice and is now on its third rise. Deep learning implies the use of neural networks. The "deep" in deep learning refers to a neural network with many hidden layers. Because neural networks have been around for so long, they have quite a bit of baggage. Researchers have created many different training algorithms, activation/transfer functions, and structures. This course is only concerned with the latest, most current state{-}of{-}the{-}art techniques for deep neural networks. I will not spend much time discussing the history of neural networks.%
\index{algorithm}%
\index{hidden layer}%
\index{layer}%
\index{learning}%
\index{model}%
\index{neural network}%
\index{training}%
\index{training algorithm}%
\par%
Neural networks accept input and produce output. The input to a neural network is called the feature vector. The size of this vector is always a fixed length. Changing the size of the feature vector usually means recreating the entire neural network. Though the feature vector is called a "vector," this is not always the case. A vector implies a 1D array. Later we will learn about convolutional neural networks (CNNs), which can allow the input size to change without retraining the neural network. Historically the input to a neural network was always 1D. However, with modern neural networks, you might see input data, such as:%
\index{CNN}%
\index{convolution}%
\index{convolutional}%
\index{Convolutional Neural Networks}%
\index{feature}%
\index{input}%
\index{neural network}%
\index{output}%
\index{training}%
\index{vector}%
\par%
\begin{itemize}[noitemsep]%
\item%
\textbf{1D vector }%
{-} Classic input to a neural network, similar to rows in a spreadsheet. Common in predictive modeling.%
\index{input}%
\index{model}%
\index{neural network}%
\index{predict}%
\item%
\textbf{2D Matrix }%
{-} Grayscale image input to a CNN.%
\index{CNN}%
\index{input}%
\item%
\textbf{3D Matrix }%
{-} Color image input to a CNN.%
\index{CNN}%
\index{input}%
\item%
\textbf{nD Matrix }%
{-} Higher{-}order input to a CNN.%
\index{CNN}%
\index{input}%
\end{itemize}%
Before CNNs, programs either encoded images to an intermediate form or sent the image input to a neural network by merely squashing the image matrix into a long array by placing the image's rows side{-}by{-}side. CNNs are different as the matrix passes through the neural network layers.%
\index{CNN}%
\index{input}%
\index{layer}%
\index{matrix}%
\index{neural network}%
\par%
Initially, this book will focus on 1D input to neural networks. However, later modules will focus more heavily on higher dimension input.%
\index{input}%
\index{neural network}%
\par%
The term dimension can be confusing in neural networks. In the sense of a 1D input vector, dimension refers to how many elements are in that 1D array. For example, a neural network with ten input neurons has ten dimensions. However, now that we have CNNs, the input has dimensions. The input to the neural network will%
\index{CNN}%
\index{input}%
\index{input neuron}%
\index{input vector}%
\index{neural network}%
\index{neuron}%
\index{vector}%
\textit{ usually }%
have 1, 2, or 3 dimensions. Four or more dimensions are unusual. You might have a 2D input to a neural network with 64x64 pixels. This configuration would result in 4,096 input neurons. This network is either 2D or 4,096D, depending on which dimensions you reference.%
\index{input}%
\index{input neuron}%
\index{neural network}%
\index{neuron}%
\par%
\subsection{Classification or Regression}%
\label{subsec:ClassificationorRegression}%
Like many models, neural networks can function in classification or regression:%
\index{classification}%
\index{model}%
\index{neural network}%
\index{regression}%
\par
\begin{figure}[h]%
\centering%
\includegraphics[width=4in]{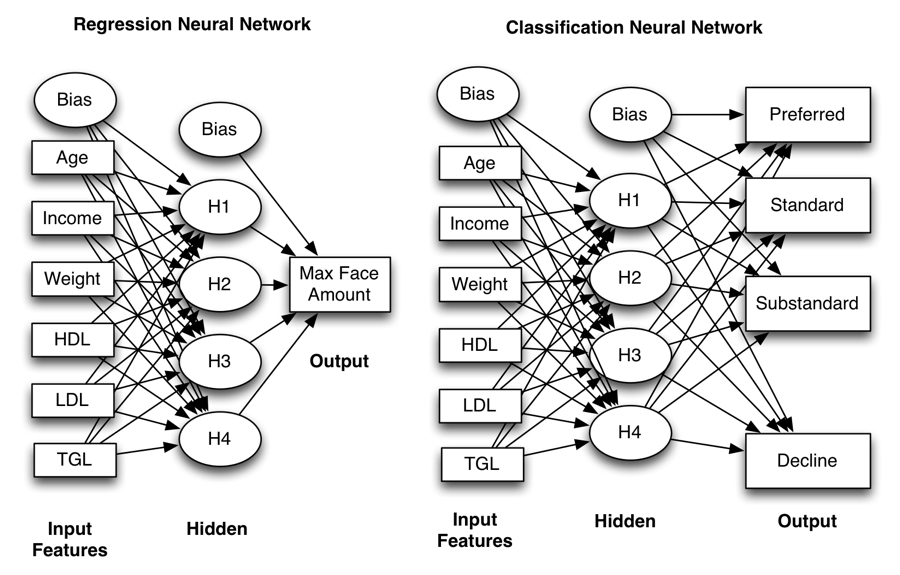}%
\caption{Neural Network Classification and Regression}%
\label{3.CLS-REG}%
\end{figure}

%
\begin{itemize}[noitemsep]%
\item%
\textbf{Regression }%
{-} You expect a number as your neural network's prediction.%
\index{neural network}%
\index{predict}%
\item%
\textbf{Classification }%
{-} You expect a class/category as your neural network's prediction.%
\index{neural network}%
\index{predict}%
\end{itemize}%
A classification and regression neural network is shown by Figure \ref{3.CLS-REG}.%
\index{classification}%
\index{neural network}%
\index{regression}%
\par%

\par%
Notice that the output of the regression neural network is numeric, and the classification output is a class. Regression, or two{-}class classification, networks always have a single output. Classification neural networks have an output neuron for each category.%
\index{classification}%
\index{neural network}%
\index{neuron}%
\index{output}%
\index{output neuron}%
\index{regression}%
\par

\subsection{Neurons and Layers}%
\label{subsec:NeuronsandLayers}%
Most neural network structures use some type of neuron. Many different neural networks exist, and programmers introduce experimental neural network structures. Consequently, it is not possible to cover every neural network architecture. However, there are some commonalities among neural network implementations. A neural network algorithm would typically be composed of individual, interconnected units, even though these units may or may not be called neurons. The name for a neural network processing unit varies among the literature sources. It could be called a node, neuron, or unit.%
\index{algorithm}%
\index{architecture}%
\index{network architecture}%
\index{neural network}%
\index{neuron}%
\index{ROC}%
\index{ROC}%
\index{SOM}%
\par
\begin{figure}[h]%
\centering%
\includegraphics[width=4in]{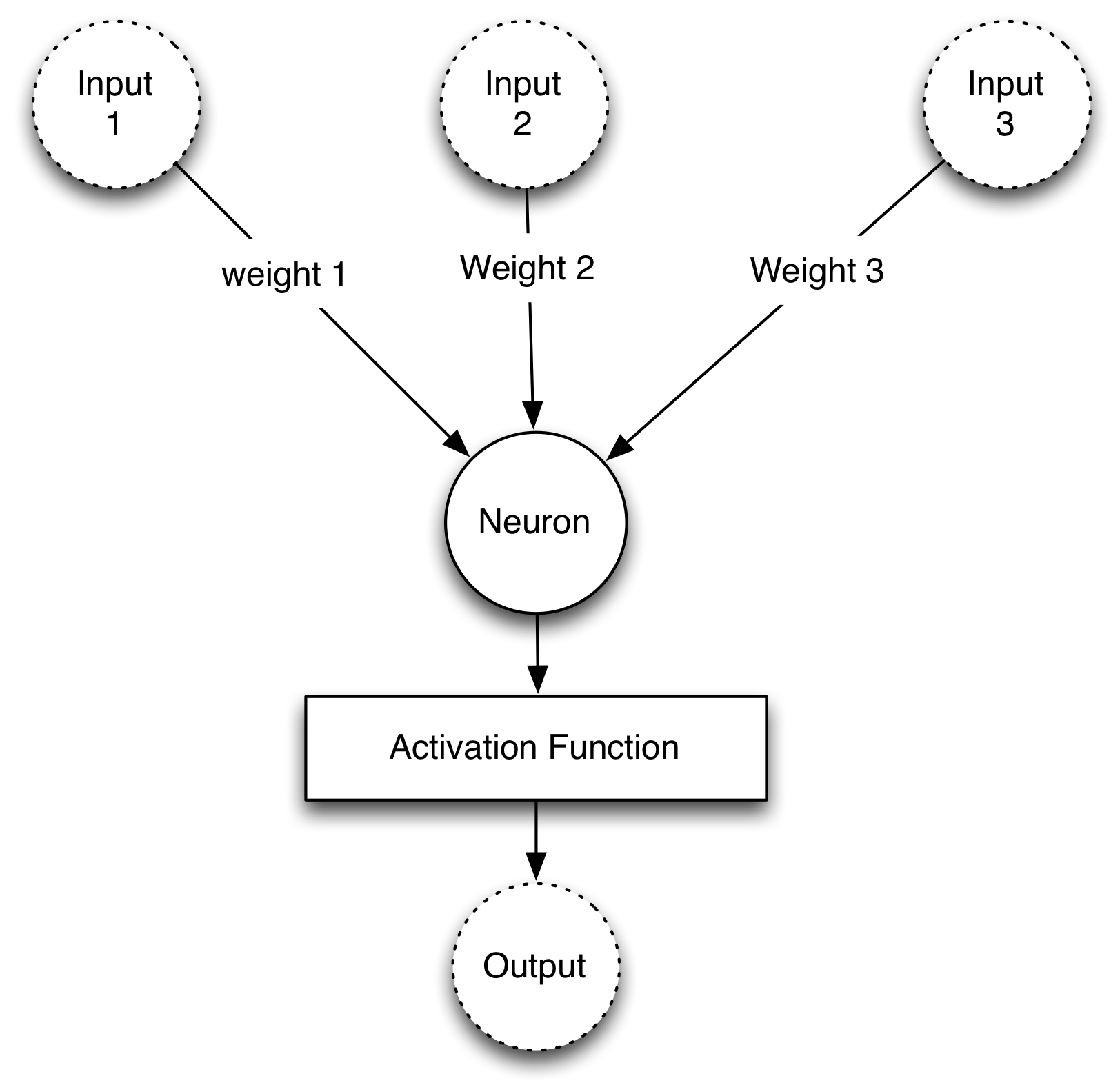}%
\caption{An Artificial Neuron}%
\label{3.ANN}%
\end{figure}

%
A diagram shows the abstract structure of a single artificial neuron in Figure \ref{3.ANN}.%
\index{neuron}%
\par%

\par%
The artificial neuron receives input from one or more sources that may be other neurons or data fed into the network from a computer program. This input is usually floating{-}point or binary. Often binary input is encoded to floating{-}point by representing true or false as 1 or 0. Sometimes the program also depicts the binary information using a bipolar system with true as one and false as {-}1.%
\index{input}%
\index{neuron}%
\index{SOM}%
\par%
An artificial neuron multiplies each of these inputs by a weight. Then it adds these multiplications and passes this sum to an activation function. Some neural networks do not use an activation function. The following equation summarizes the calculated output of a neuron:%
\index{activation function}%
\index{calculated}%
\index{input}%
\index{neural network}%
\index{neuron}%
\index{output}%
\index{SOM}%
\par%
\vspace{2mm}%
\begin{equation*}
 f(x,w) = \phi(\sum_i(\theta_i \cdot x_i)) 
\end{equation*}
\vspace{2mm}%
\par%
In the above equation, the variables $x$ and $\theta$ represent the input and weights of the neuron. The variable $i$ corresponds to the number of weights and inputs. You must always have the same number of weights as inputs. The neural network multiplies each weight by its respective input and feeds the products of these multiplications into an activation function, denoted by the Greek letter $\phi$ (phi). This process results in a single output from the neuron.%
\index{activation function}%
\index{input}%
\index{neural network}%
\index{neuron}%
\index{output}%
\index{ROC}%
\index{ROC}%
\par%
The above neuron has two inputs plus the bias as a third. This neuron might accept the following input feature vector:%
\index{bias}%
\index{feature}%
\index{input}%
\index{neuron}%
\index{vector}%
\par%
\vspace{2mm}%
\begin{equation*}
 [1,2] 
\end{equation*}
\vspace{2mm}%
\par%
Because a bias neuron is present, the program should append the value of one as follows:%
\index{bias}%
\index{bias neuron}%
\index{neuron}%
\par%
\vspace{2mm}%
\begin{equation*}
 [1,2,1] 
\end{equation*}
\vspace{2mm}%
\par%
The weights for a 3{-}input layer (2 real inputs + bias) will always have additional weight for the bias. A weight vector might be:%
\index{bias}%
\index{input}%
\index{input layer}%
\index{layer}%
\index{vector}%
\par%
\vspace{2mm}%
\begin{equation*}
 [ 0.1, 0.2, 0.3] 
\end{equation*}
\vspace{2mm}%
\par%
To calculate the summation, perform the following:%
\par%
$$ 0.1%
\textit{1 + 0.2}%
2 + 0.3*1 = 0.8 $$%
\par%
The program passes a value of 0.8 to the $\phi$ (phi) function, representing the activation function.%
\index{activation function}%
\par
\begin{figure}[h]%
\centering%
\includegraphics[width=3in]{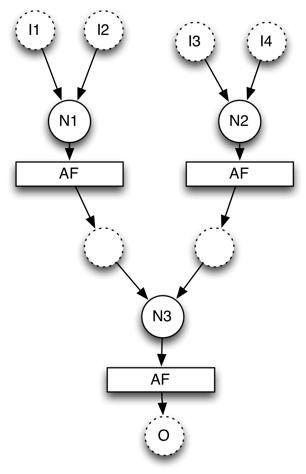}%
\caption{Three Neuron Neural Network}%
\label{3.ANN-3}%
\end{figure}

%
The above figure shows the structure with just one building block. You can chain together many artificial neurons to build an artificial neural network (ANN). Think of the artificial neurons as building blocks for which the input and output circles are the connectors. Figure \ref{3.ANN-3} shows an artificial neural network composed of three neurons:%
\index{input}%
\index{neural network}%
\index{neuron}%
\index{output}%
\par%

\par%
The above diagram shows three interconnected neurons. This representation is essentially this figure, minus a few inputs, repeated three times and then connected. It also has a total of four inputs and a single output. The output of neurons%
\index{input}%
\index{neuron}%
\index{output}%
\textbf{ N1 }%
and%
\textbf{ N2 }%
feed%
\textbf{ N3 }%
to produce the output%
\index{output}%
\textbf{ O}%
.  To calculate the output for this network, we perform the previous equation three times. The first two times calculate%
\index{output}%
\textbf{ N1 }%
and%
\textbf{ N2}%
, and the third calculation uses the output of%
\index{output}%
\textbf{ N1 }%
and%
\textbf{ N2 }%
to calculate%
\textbf{ N3}%
.%
\par
\begin{figure}[h]%
\centering%
\includegraphics[width=3in]{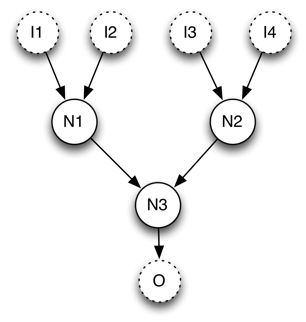}%
\caption{Three Neuron Neural Network}%
\label{3.SANN-3}%
\end{figure}

%
Neural network diagrams do not typically show the detail seen in the previous figure. We can omit the activation functions and intermediate outputs to simplify the chart, resulting in Figure \ref{3.SANN-3}.%
\index{activation function}%
\index{neural network}%
\index{output}%
\par%

\par%
Looking at the previous figure, you can see two additional components of neural networks. First, consider the graph represents the inputs and outputs as abstract dotted line circles. The input and output could be parts of a more extensive neural network. However, the input and output are often a particular type of neuron that accepts data from the computer program using the neural network. The output neurons return a result to the program. This type of neuron is called an input neuron. We will discuss these neurons in the next section. This figure shows the neurons arranged in layers. The input neurons are the first layer, the%
\index{input}%
\index{input neuron}%
\index{layer}%
\index{neural network}%
\index{neuron}%
\index{output}%
\index{output neuron}%
\textbf{ N1 }%
and%
\textbf{ N2 }%
neurons create the second layer, the third layer contains%
\index{layer}%
\index{neuron}%
\textbf{ N3}%
, and the fourth layer has%
\index{layer}%
\textbf{ O}%
.  Most neural networks arrange neurons into layers.%
\index{layer}%
\index{neural network}%
\index{neuron}%
\par
\begin{figure}[h]%
\centering%
\includegraphics[width=4in]{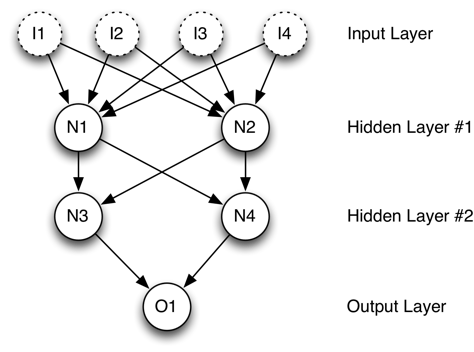}%
\caption{Fully Connected Neural Network Diagram}%
\label{3.F-ANN}%
\end{figure}

%
The neurons that form a layer share several characteristics. First, every neuron in a layer has the same activation function. However, the activation functions employed by each layer may be different. Each of the layers fully connects to the next layer. In other words, every neuron in one layer has a connection to neurons in the previous layer. The former figure is not fully connected. Several layers are missing connections. For example,%
\index{activation function}%
\index{connection}%
\index{layer}%
\index{neuron}%
\textbf{ I1 }%
and%
\textbf{ N2 }%
do not connect. The next neural network in Figure \ref{3.F-ANN} is fully connected and has an additional layer.%
\index{layer}%
\index{neural network}%
\par%

\par%
In this figure, you see a fully connected, multilayered neural network. Networks such as this one will always have an input and output layer. The hidden layer structure determines the name of the network architecture. The network in this figure is a two{-}hidden{-}layer network. Most networks will have between zero and two hidden layers. Without implementing deep learning strategies, networks with more than two hidden layers are rare.%
\index{architecture}%
\index{hidden layer}%
\index{input}%
\index{layer}%
\index{learning}%
\index{network architecture}%
\index{neural network}%
\index{output}%
\index{output layer}%
\par%
You might also notice that the arrows always point downward or forward from the input to the output. Later in this course, we will see recurrent neural networks that form inverted loops among the neurons. This type of neural network is called a feedforward neural network.%
\index{feedforward}%
\index{input}%
\index{neural network}%
\index{neuron}%
\index{output}%
\index{recurrent}%
\par

\subsection{Types of Neurons}%
\label{subsec:TypesofNeurons}%
In the last section, we briefly introduced the idea that different types of neurons exist. Not every neural network will use every kind of neuron. It is also possible for a single neuron to fill the role of several different neuron types. Now we will explain all the neuron types described in the course.%
\index{neural network}%
\index{neuron}%
\par%
There are usually four types of neurons in a neural network:%
\index{neural network}%
\index{neuron}%
\par%
\begin{itemize}[noitemsep]%
\item%
\textbf{Input Neurons }%
{-} We map each input neuron to one element in the feature vector.%
\index{feature}%
\index{input}%
\index{input neuron}%
\index{neuron}%
\index{vector}%
\item%
\textbf{Hidden Neurons }%
{-} Hidden neurons allow the neural network to be abstract and process the input into the output.%
\index{hidden neuron}%
\index{input}%
\index{neural network}%
\index{neuron}%
\index{output}%
\index{ROC}%
\index{ROC}%
\item%
\textbf{Output Neurons }%
{-} Each output neuron calculates one part of the output.%
\index{neuron}%
\index{output}%
\index{output neuron}%
\item%
\textbf{Bias Neurons }%
{-} Work similar to the y{-}intercept of a linear equation.%
\index{linear}%
\end{itemize}%
We place each neuron into a layer:%
\index{layer}%
\index{neuron}%
\par%
\begin{itemize}[noitemsep]%
\item%
\textbf{Input Layer }%
{-} The input layer accepts feature vectors from the dataset. Input layers usually have a bias neuron.%
\index{bias}%
\index{bias neuron}%
\index{dataset}%
\index{feature}%
\index{input}%
\index{input layer}%
\index{layer}%
\index{neuron}%
\index{vector}%
\item%
\textbf{Output Layer }%
{-} The output from the neural network. The output layer does not have a bias neuron.%
\index{bias}%
\index{bias neuron}%
\index{layer}%
\index{neural network}%
\index{neuron}%
\index{output}%
\index{output layer}%
\item%
\textbf{Hidden Layers }%
{-} Layers between the input and output layers. Each hidden layer will usually have a bias neuron.%
\index{bias}%
\index{bias neuron}%
\index{hidden layer}%
\index{input}%
\index{layer}%
\index{neuron}%
\index{output}%
\index{output layer}%
\end{itemize}

\subsection{Input and Output Neurons}%
\label{subsec:InputandOutputNeurons}%
Nearly every neural network has input and output neurons. The input neurons accept data from the program for the network. The output neuron provides processed data from the network back to the program. The program will group these input and output neurons into separate layers called the input and output layers. The program normally represents the input to a neural network as an array or vector. The number of elements contained in the vector must equal the number of input neurons. For example, a neural network with three input neurons might accept the following input vector:%
\index{input}%
\index{input neuron}%
\index{input vector}%
\index{layer}%
\index{neural network}%
\index{neuron}%
\index{output}%
\index{output layer}%
\index{output neuron}%
\index{ROC}%
\index{ROC}%
\index{vector}%
\par%
\vspace{2mm}%
\begin{equation*}
 [0.5, 0.75, 0.2] 
\end{equation*}
\vspace{2mm}%
\par%
Neural networks typically accept floating{-}point vectors as their input. To be consistent, we will represent the output of a single output neuron network as a single{-}element vector. Likewise, neural networks will output a vector with a length equal to the number of output neurons. The output will often be a single value from a single output neuron.%
\index{input}%
\index{neural network}%
\index{neuron}%
\index{output}%
\index{output neuron}%
\index{vector}%
\par

\subsection{Hidden Neurons}%
\label{subsec:HiddenNeurons}%
Hidden neurons have two essential characteristics. First, hidden neurons only receive input from other neurons, such as input or other hidden neurons. Second, hidden neurons only output to other neurons, such as output or other hidden neurons. Hidden neurons help the neural network understand the input and form the output. Programmers often group hidden neurons into fully connected hidden layers. However, these hidden layers do not directly process the incoming data or the eventual output.%
\index{hidden layer}%
\index{hidden neuron}%
\index{input}%
\index{layer}%
\index{neural network}%
\index{neuron}%
\index{output}%
\index{ROC}%
\index{ROC}%
\par%
A common question for programmers concerns the number of hidden neurons in a network. Since the answer to this question is complex, more than one section of the course will include a relevant discussion of the number of hidden neurons. Before deep learning, researchers generally suggested that anything more than a single hidden layer is excessive.%
\index{hidden layer}%
\index{hidden neuron}%
\index{layer}%
\index{learning}%
\index{neuron}%
\cite{hornik1989multilayer}%
Researchers have proven that a single{-}hidden{-}layer neural network can function as a universal approximator. In other words, this network should be able to learn to produce (or approximate) any output from any input as long as it has enough hidden neurons in a single layer.%
\index{hidden neuron}%
\index{input}%
\index{layer}%
\index{neural network}%
\index{neuron}%
\index{output}%
\par%
Training refers to the process that determines good weight values. Before the advent of deep learning, researchers feared additional layers would lengthen training time or encourage overfitting. Both concerns are true; however, increased hardware speeds and clever techniques can mitigate these concerns. Before researchers introduced deep learning techniques, we did not have an efficient way to train a deep network, which is a neural network with many hidden layers. Although a single{-}hidden{-}layer neural network can theoretically learn anything, deep learning facilitates a more complex representation of patterns in the data.%
\index{hidden layer}%
\index{layer}%
\index{learning}%
\index{neural network}%
\index{overfitting}%
\index{ROC}%
\index{ROC}%
\index{training}%
\par

\subsection{Bias Neurons}%
\begin{figure}[h]%
\centering%
\includegraphics[width=4in]{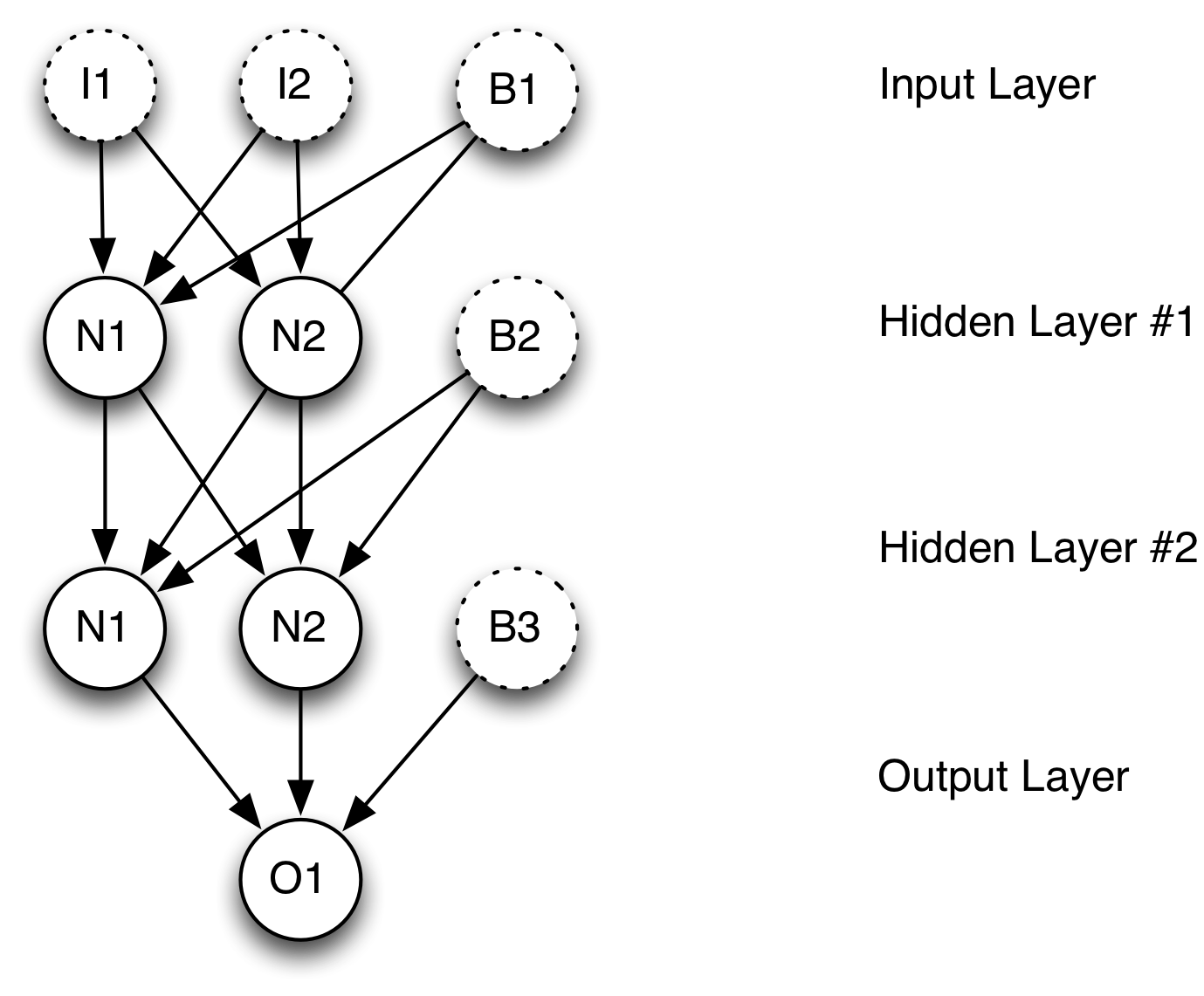}%
\caption{Neural Network with Bias Neurons}%
\label{3.BIAS}%
\end{figure}

\label{subsec:BiasNeurons}%
Programmers add bias neurons to neural networks to help them learn patterns. Bias neurons function like an input neuron that always produces a value of 1. Because the bias neurons have a constant output of 1, they are not connected to the previous layer. The value of 1, called the bias activation, can be set to values other than 1. However, 1 is the most common bias activation. Not all neural networks have bias neurons. Figure \ref{3.BIAS} shows a single{-}hidden{-}layer neural network with bias neurons:%
\index{bias}%
\index{bias neuron}%
\index{input}%
\index{input neuron}%
\index{layer}%
\index{neural network}%
\index{neuron}%
\index{output}%
\par%

\par%
The above network contains three bias neurons. Except for the output layer, every level includes a single bias neuron. Bias neurons allow the program to shift the output of an activation function. We will see precisely how this shifting occurs later in the module when discussing activation functions.%
\index{activation function}%
\index{bias}%
\index{bias neuron}%
\index{layer}%
\index{neuron}%
\index{output}%
\index{output layer}%
\par

\subsection{Other Neuron Types}%
\label{subsec:OtherNeuronTypes}%
The individual units that comprise a neural network are not always called neurons. Researchers will sometimes refer to these neurons as nodes, units, or summations. You will almost always construct neural networks of weighted connections between these units.%
\index{connection}%
\index{neural network}%
\index{neuron}%
\index{SOM}%
\par

\subsection{Why are Bias Neurons Needed?}%
\label{subsec:WhyareBiasNeuronsNeeded?}%
The activation functions from the previous section specify the output of a single neuron. Together, the weight and bias of a neuron shape the output of the activation to produce the desired output. To see how this process occurs, consider the following equation. It represents a single{-}input sigmoid activation neural network.%
\index{activation function}%
\index{bias}%
\index{input}%
\index{neural network}%
\index{neuron}%
\index{output}%
\index{ROC}%
\index{ROC}%
\index{sigmoid}%
\par%
\vspace{2mm}%
\begin{equation*}
 f(x,w,b) = \frac{1}{1 + e^{-(wx+b)}} 
\end{equation*}
\vspace{2mm}%
\par%
The $x$ variable represents the single input to the neural network. The $w$ and $b$ variables specify the weight and bias of the neural network. The above equation combines the weighted sum of the inputs and the sigmoid activation function. For this section, we will consider the sigmoid function because it demonstrates a bias neuron's effect.%
\index{activation function}%
\index{bias}%
\index{bias neuron}%
\index{input}%
\index{neural network}%
\index{neuron}%
\index{sigmoid}%
\par
\begin{figure}[h]%
\centering%
\includegraphics[width=4in]{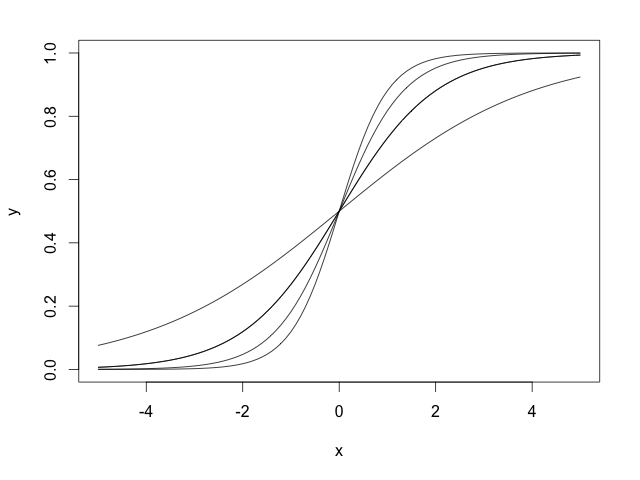}%
\caption{Neuron Weight Shifting}%
\label{3.A-WEIGHT}%
\end{figure}

%
The weights of the neuron allow you to adjust the slope or shape of the activation function. Figure \ref{3.A-WEIGHT} shows the effect on the output of the sigmoid activation function if the weight is varied:%
\index{activation function}%
\index{neuron}%
\index{output}%
\index{sigmoid}%
\index{slope}%
\par%

\par%
The above diagram shows several sigmoid curves using the following parameters:%
\index{parameter}%
\index{sigmoid}%
\par%
\vspace{2mm}%
\begin{equation*}
 f(x,0.5,0.0) 
\end{equation*}
\begin{equation*}
 f(x,1.0,0.0) 
\end{equation*}
\begin{equation*}
 f(x,1.5,0.0) 
\end{equation*}
\begin{equation*}
 f(x,2.0,0.0) 
\end{equation*}
\vspace{2mm}%
\par%
We did not use bias to produce the curves, which is evident in the third parameter of 0 in each case. Using four weight values yields four different sigmoid curves in the above figure. No matter the weight, we always get the same value of 0.5 when%
\index{bias}%
\index{parameter}%
\index{sigmoid}%
\textit{ x }%
is 0 because all curves hit the same point when x is 0. We might need the neural network to produce other values when the input is near 0.5.%
\index{input}%
\index{neural network}%
\par
\begin{figure}[h]%
\centering%
\includegraphics[width=4in]{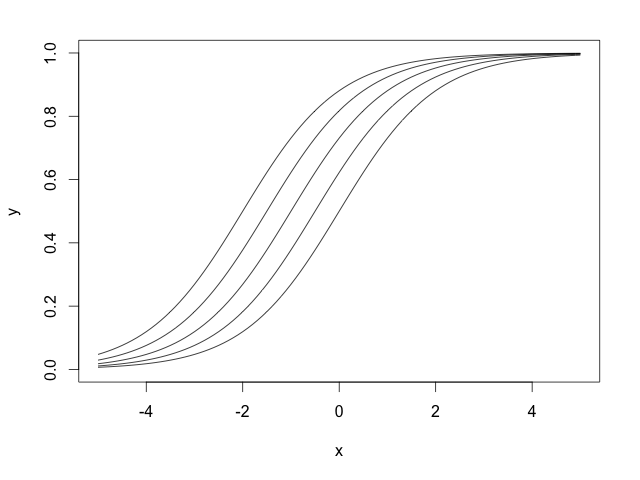}%
\caption{Neuron Bias Shifting}%
\label{3.A-BIAS}%
\end{figure}

%
Bias does shift the sigmoid curve, which allows values other than 0.5 when%
\index{bias}%
\index{sigmoid}%
\textit{ x }%
is near 0. Figure \ref{3.A-BIAS} shows the effect of using a weight of 1.0 with several different biases:%
\index{bias}%
\par%

\par%
The above diagram shows several sigmoid curves with the following parameters:%
\index{parameter}%
\index{sigmoid}%
\par%
\vspace{2mm}%
\begin{equation*}
 f(x,1.0,1.0) 
\end{equation*}
\begin{equation*}
 f(x,1.0,0.5) 
\end{equation*}
\begin{equation*}
 f(x,1.0,1.5) 
\end{equation*}
\begin{equation*}
 f(x,1.0,2.0) 
\end{equation*}
\vspace{2mm}%
\par%
We used a weight of 1.0 for these curves in all cases. When we utilized several different biases, sigmoid curves shifted to the left or right. Because all the curves merge at the top right or bottom left, it is not a complete shift.%
\index{bias}%
\index{sigmoid}%
\par%
When we put bias and weights together, they produced a curve that created the necessary output. The above curves are the output from only one neuron. In a complete network, the output from many different neurons will combine to produce intricate output patterns.%
\index{bias}%
\index{neuron}%
\index{output}%
\par

\subsection{Modern Activation Functions}%
\label{subsec:ModernActivationFunctions}%
Activation functions, also known as transfer functions, are used to calculate the output of each layer of a neural network. Historically neural networks have used a hyperbolic tangent, sigmoid/logistic, or linear activation function. However, modern deep neural networks primarily make use of the following activation functions:%
\index{activation function}%
\index{hyperbolic tangent}%
\index{layer}%
\index{linear}%
\index{neural network}%
\index{output}%
\index{sigmoid}%
\par%
\begin{itemize}[noitemsep]%
\item%
\textbf{Rectified Linear Unit (ReLU) }%
{-} Used for the output of hidden layers.%
\index{hidden layer}%
\index{layer}%
\index{output}%
\cite{glorot2011deep}%
\item%
\textbf{Softmax }%
{-} Used for the output of classification neural networks.%
\index{classification}%
\index{neural network}%
\index{output}%
\item%
\textbf{Linear }%
{-} Used for the output of regression neural networks (or 2{-}class classification).%
\index{classification}%
\index{neural network}%
\index{output}%
\index{regression}%
\end{itemize}

\subsection{Linear Activation Function}%
\label{subsec:LinearActivationFunction}%
The most basic activation function is the linear function because it does not change the neuron output. The following equation 1.2 shows how the program typically implements a linear activation function:%
\index{activation function}%
\index{linear}%
\index{neuron}%
\index{output}%
\par%
\vspace{2mm}%
\begin{equation*}
 \phi(x) = x 
\end{equation*}
\vspace{2mm}%
\par
\begin{figure}[h]%
\centering%
\includegraphics[width=4in]{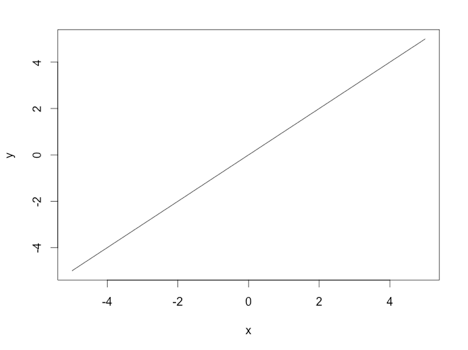}%
\caption{Linear Activation Function}%
\label{3.LIN}%
\end{figure}

%
As you can observe, this activation function simply returns the value that the neuron inputs passed to it.  Figure \ref{3.LIN} shows the graph for a linear activation function:%
\index{activation function}%
\index{input}%
\index{linear}%
\index{neuron}%
\par%

\par%
Regression neural networks, which learn to provide numeric values, will usually use a linear activation function on their output layer. Classification neural networks, which determine an appropriate class for their input, will often utilize a softmax activation function for their output layer.%
\index{activation function}%
\index{classification}%
\index{input}%
\index{layer}%
\index{linear}%
\index{neural network}%
\index{output}%
\index{output layer}%
\index{regression}%
\index{softmax}%
\par

\subsection{Rectified Linear Units (ReLU)}%
\label{subsec:RectifiedLinearUnits(ReLU)}%
Since its introduction, researchers have rapidly adopted the rectified linear unit (ReLU).%
\index{linear}%
\index{ReLU}%
\cite{nair2010rectified}%
Before the ReLU activation function, the programmers generally regarded the hyperbolic tangent as the activation function of choice. Most current research now recommends the ReLU due to superior training results. As a result, most neural networks should utilize the ReLU on hidden layers and either softmax or linear on the output layer. The following equation shows the straightforward ReLU function:%
\index{activation function}%
\index{hidden layer}%
\index{hyperbolic tangent}%
\index{layer}%
\index{linear}%
\index{neural network}%
\index{output}%
\index{output layer}%
\index{ReLU}%
\index{softmax}%
\index{training}%
\par%
\vspace{2mm}%
\begin{equation*}
 \phi(x) = \max(0, x) 
\end{equation*}
\vspace{2mm}%
\par
\begin{figure}[h]%
\centering%
\includegraphics[width=4in]{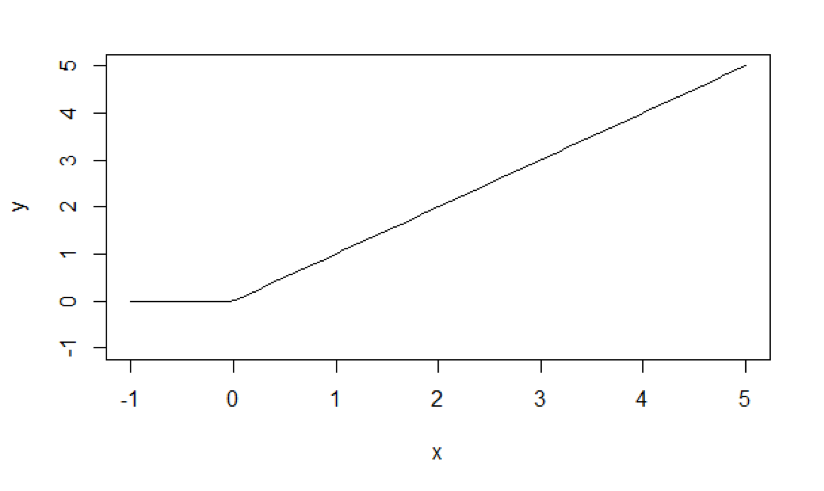}%
\caption{Rectified Linear Units (ReLU)}%
\label{3.RELU}%
\end{figure}

%
Figure \ref{3.RELU} shows the graph of the ReLU activation function:%
\index{activation function}%
\index{ReLU}%
\par%

\par%
Most current research states that the hidden layers of your neural network should use the ReLU activation.%
\index{hidden layer}%
\index{layer}%
\index{neural network}%
\index{ReLU}%
\par

\subsection{Softmax Activation Function}%
\label{subsec:SoftmaxActivationFunction}%
The final activation function that we will examine is the softmax activation function. Along with the linear activation function, you can usually find the softmax function in the output layer of a neural network. Classification neural networks typically employ the softmax function. The neuron with the highest value claims the input as a member of its class. Because it is a preferable method, the softmax activation function forces the neural network's output to represent the probability that the input falls into each of the classes. The neuron's outputs are numeric values without the softmax, with the highest indicating the winning class.%
\index{activation function}%
\index{classification}%
\index{input}%
\index{layer}%
\index{linear}%
\index{neural network}%
\index{neuron}%
\index{output}%
\index{output layer}%
\index{probability}%
\index{softmax}%
\par%
To see how the program uses the softmax activation function, we will look at a typical neural network classification problem. The iris data set contains four measurements for 150 different iris flowers. Each of these flowers belongs to one of three species of iris. When you provide the measurements of a flower, the softmax function allows the neural network to give you the probability that these measurements belong to each of the three species. For example, the neural network might tell you that there is an 80\% chance that the iris is setosa, a 15\% probability that it is virginica, and only a 5\% probability of versicolor. Because these are probabilities, they must add up to 100\%. There could not be an 80\% probability of setosa, a 75\% probability of virginica, and a 20\% probability of versicolor{-}{-}{-}this type of result would be nonsensical.%
\index{activation function}%
\index{classification}%
\index{iris}%
\index{neural network}%
\index{probability}%
\index{softmax}%
\index{species}%
\par%
To classify input data into one of three iris species, you will need one output neuron for each species. The output neurons do not inherently specify the probability of each of the three species. Therefore, it is desirable to provide probabilities that sum to 100\%. The neural network will tell you the likelihood of a flower being each of the three species. To get the probability, use the softmax function in the following equation:%
\index{input}%
\index{iris}%
\index{neural network}%
\index{neuron}%
\index{output}%
\index{output neuron}%
\index{probability}%
\index{softmax}%
\index{species}%
\par%
\vspace{2mm}%
\begin{equation*}
 \phi_i(x) = \frac{exp(x_i)}{\sum_{j}^{ }exp(x_j)} 
\end{equation*}
\vspace{2mm}%
\par%
In the above equation, $i$ represents the index of the output neuron ($\phi$) that the program is calculating, and $j$ represents the indexes of all neurons in the group/level. The variable $x$ designates the array of output neurons. It's important to note that the program calculates the softmax activation differently than the other activation functions in this module. When softmax is the activation function, the output of a single neuron is dependent on the other output neurons.%
\index{activation function}%
\index{neuron}%
\index{output}%
\index{output neuron}%
\index{softmax}%
\par%
To see the softmax function in operation, refer to this%
\index{softmax}%
\href{http://www.heatonresearch.com/aifh/vol3/softmax.html}{ Softmax example website}%
.%
\par%
Consider a trained neural network that classifies data into three categories: the three iris species. In this case, you would use one output neuron for each of the target classes. Consider if the neural network were to output the following:%
\index{iris}%
\index{neural network}%
\index{neuron}%
\index{output}%
\index{output neuron}%
\index{species}%
\par%
\begin{itemize}[noitemsep]%
\item%
\textbf{Neuron 1}%
: setosa: 0.9%
\item%
\textbf{Neuron 2}%
: versicolour: 0.2%
\item%
\textbf{Neuron 3}%
: virginica: 0.4%
\end{itemize}%
The above output shows that the neural network considers the data to represent a setosa iris. However, these numbers are not probabilities. The 0.9 value does not represent a 90\% likelihood of the data representing a setosa. These values sum to 1.5. For the program to treat them as probabilities, they must sum to 1.0. The output vector for this neural network is the following:%
\index{iris}%
\index{neural network}%
\index{output}%
\index{vector}%
\par%
\vspace{2mm}%
\begin{equation*}
 [0.9,0.2,0.4] 
\end{equation*}
\vspace{2mm}%
\par%
If you provide this vector to the softmax function it will return the following vector:%
\index{softmax}%
\index{vector}%
\par%
\vspace{2mm}%
\begin{equation*}
 [0.47548495534876745 , 0.2361188410001125 , 0.28839620365112] 
\end{equation*}
\vspace{2mm}%
\par%
The above three values do sum to 1.0 and can be treated as probabilities.  The likelihood of the data representing a setosa iris is 48\% because the first value in the vector rounds to 0.48 (48\%).  You can calculate this value in the following manner:%
\index{iris}%
\index{vector}%
\par%
\vspace{2mm}%
\begin{equation*}
 sum=\exp(0.9)+\exp(0.2)+\exp(0.4)=5.17283056695839 
\end{equation*}
\begin{equation*}
 j_0= \exp(0.9)/sum = 0.47548495534876745 
\end{equation*}
\begin{equation*}
 j_1= \exp(0.2)/sum = 0.2361188410001125 
\end{equation*}
\begin{equation*}
 j_2= \exp(0.4)/sum = 0.28839620365112 
\end{equation*}
\vspace{2mm}%
\par

\subsection{Step Activation Function}%
\label{subsec:StepActivationFunction}%
The step or threshold activation function is another simple activation function. Neural networks were initially called perceptrons. McCulloch  Pitts (1943) introduced the original perceptron and used a step activation function like the following equation:%
\index{activation function}%
\index{neural network}%
\cite{mcculloch1943logical}%
The step activation is 1 if x>=0.5, and 0 otherwise.%
\par
\begin{figure}[h]%
\centering%
\includegraphics[width=4in]{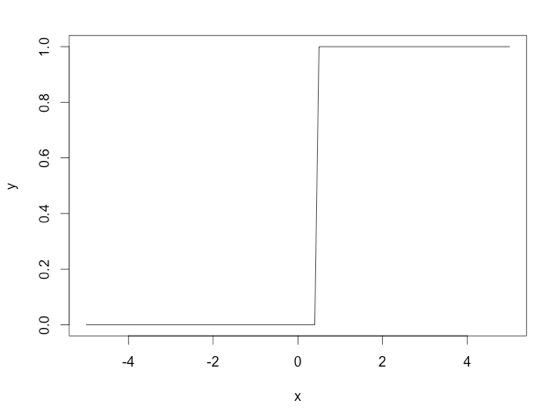}%
\caption{Step Activation Function}%
\label{3.STEP}%
\end{figure}

%
This equation outputs a value of 1.0 for incoming values of 0.5 or higher and 0 for all other values. Step functions, also known as threshold functions, only return 1 (true) for values above the specified threshold, as seen in Figure \ref{3.STEP}.%
\index{output}%
\par%

\par

\subsection{Sigmoid Activation Function}%
\label{subsec:SigmoidActivationFunction}%
The sigmoid or logistic activation function is a common choice for feedforward neural networks that need to output only positive numbers. Despite its widespread use, the hyperbolic tangent or the rectified linear unit (ReLU) activation function is usually a more suitable choice. We introduce the ReLU activation function later in this module. The following equation shows the sigmoid activation function:%
\index{activation function}%
\index{feedforward}%
\index{hyperbolic tangent}%
\index{linear}%
\index{neural network}%
\index{output}%
\index{ReLU}%
\index{sigmoid}%
\par%
\vspace{2mm}%
\begin{equation*}
 \phi(x) = \frac{1}{1 + e^{-x}} 
\end{equation*}
\vspace{2mm}%
\par
\begin{figure}[h]%
\centering%
\includegraphics[width=4in]{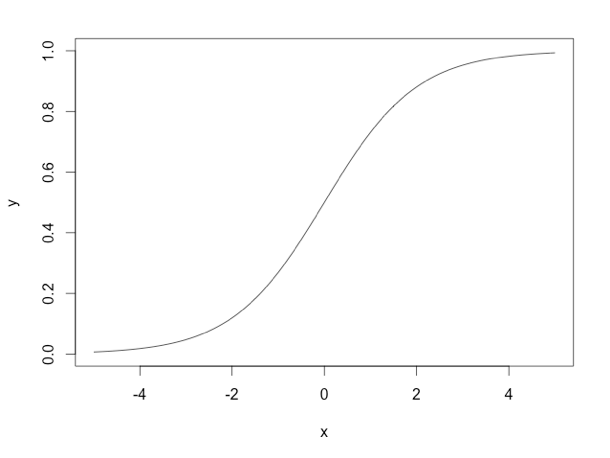}%
\caption{Sigmoid Activation Function}%
\label{3.SIGMOID}%
\end{figure}

%
Use the sigmoid function to ensure that values stay within a relatively small range, as seen in Figure \ref{3.SIGMOID}:%
\index{sigmoid}%
\par%

\par%
As you can see from the above graph, we can force values to a range. Here, the function compressed values above or below 0 to the approximate range between 0 and 1.%
\par

\subsection{Hyperbolic Tangent Activation Function}%
\label{subsec:HyperbolicTangentActivationFunction}%
The hyperbolic tangent function is also a prevalent activation function for neural networks that must output values between {-}1 and 1. This activation function is simply the hyperbolic tangent (tanh) function, as shown in the following equation:%
\index{activation function}%
\index{hyperbolic tangent}%
\index{neural network}%
\index{output}%
\par%
\vspace{2mm}%
\begin{equation*}
 \phi(x) = \tanh(x) 
\end{equation*}
\vspace{2mm}%
\par
\begin{figure}[h]%
\centering%
\includegraphics[width=4in]{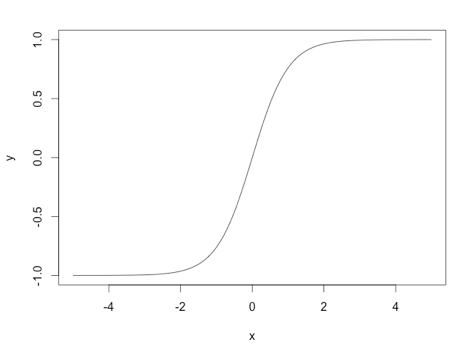}%
\caption{Hyperbolic Tangent Activation Function}%
\label{3.HTAN}%
\end{figure}

%
The graph of the hyperbolic tangent function has a similar shape to the sigmoid activation function, as seen in Figure \ref{3.HTAN}.%
\index{activation function}%
\index{hyperbolic tangent}%
\index{sigmoid}%
\par%

\par%
The hyperbolic tangent function has several advantages over the sigmoid activation function.%
\index{activation function}%
\index{hyperbolic tangent}%
\index{sigmoid}%
\par

\subsection{Why ReLU?}%
\begin{figure}[h]%
\centering%
\includegraphics[width=4in]{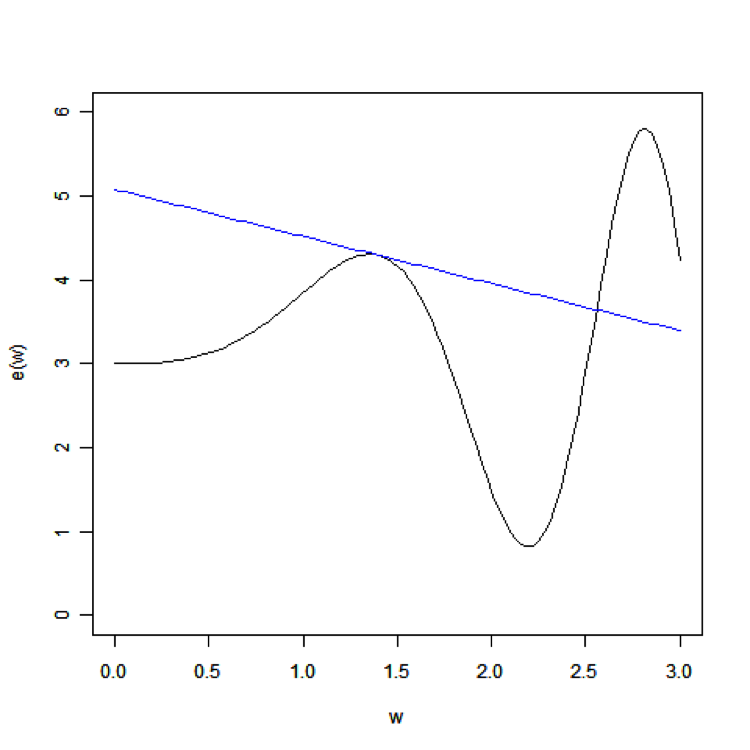}%
\caption{Derivative}%
\label{3.DERV}%
\end{figure}

\label{subsec:WhyReLU?}%
Why is the ReLU activation function so popular? One of the critical improvements to neural networks makes deep learning work.%
\index{activation function}%
\index{learning}%
\index{neural network}%
\index{ReLU}%
\cite{nair2010rectified}%
Before deep learning, the sigmoid activation function was prevalent.  We covered the sigmoid activation function earlier in this module. Frameworks like Keras often train neural networks with gradient descent. For the neural network to use gradient descent, it is necessary to take the derivative of the activation function. The program must derive partial derivatives of each of the weights for the error function. Figure \ref{3.DERV} shows a derivative, the instantaneous rate of change.%
\index{activation function}%
\index{derivative}%
\index{error}%
\index{error function}%
\index{gradient}%
\index{gradient descent}%
\index{Keras}%
\index{learning}%
\index{neural network}%
\index{partial derivative}%
\index{sigmoid}%
\par%

\par%
The derivative of the sigmoid function is given here:%
\index{derivative}%
\index{sigmoid}%
\par%
\vspace{2mm}%
\begin{equation*}
 \phi'(x)=\phi(x)(1-\phi(x)) 
\end{equation*}
\vspace{2mm}%
\par%
Textbooks often give this derivative in other forms. We use the above form for computational efficiency. To see how we determined this derivative,%
\index{derivative}%
\href{http://www.heatonresearch.com/aifh/vol3/deriv_sigmoid.html}{ refer to the following article}%
.%
\par
\begin{figure}[h]%
\centering%
\includegraphics[width=4in]{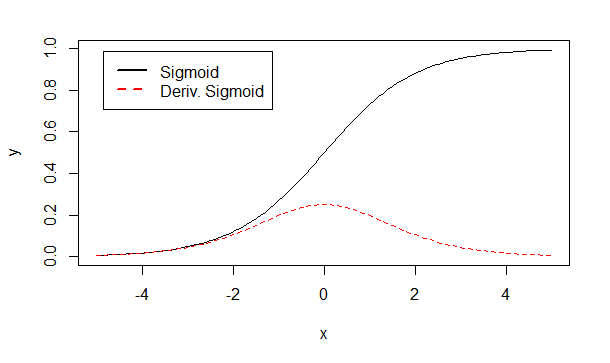}%
\caption{Sigmoid Derivative}%
\label{3.SDERV}%
\end{figure}

%
We present the graph of the sigmoid derivative in Figure \ref{3.SDERV}.%
\index{derivative}%
\index{sigmoid}%
\par%

\par%
The derivative quickly saturates to zero as $x$ moves from zero.  This is not a problem for the derivative of the ReLU, which is given here:%
\index{derivative}%
\index{ReLU}%
\par%
\vspace{2mm}%
\begin{equation*}
 \phi'(x) = \begin{cases} 1 & x > 0 \\ 0 & x \leq 0 \end{cases} 
\end{equation*}
\vspace{2mm}%
\par

\subsection{Module 3 Assignment}%
\label{subsec:Module3Assignment}%
You can find the first assignment here:%
\href{https://github.com/jeffheaton/t81_558_deep_learning/blob/master/assignments/assignment_yourname_class3.ipynb}{ assignment 3}%
\par

\section{Part 3.2: Introduction to Tensorflow and Keras}%
\label{sec:Part3.2IntroductiontoTensorflowandKeras}%
TensorFlow%
\index{TensorFlow}%
\cite{GoogleTensorFlow}%
is an open{-}source software library for machine learning in various kinds of perceptual and language understanding tasks. It is currently used for research and production by different teams in many commercial Google products, such as speech recognition, Gmail, Google Photos, and search, many of which had previously used its predecessor DistBelief. TensorFlow was originally developed by the Google Brain team for Google's research and production purposes and later released under the Apache 2.0 open source license on November 9, 2015.%
\index{learning}%
\index{TensorFlow}%
\par%
\begin{itemize}[noitemsep]%
\item%
\href{https://www.tensorflow.org/}{TensorFlow Homepage}%
\item%
\href{https://github.com/tensorflow/tensorflow}{TensorFlow GitHib}%
\item%
\href{https://groups.google.com/forum/#!forum/tensorflow}{TensorFlow Google Groups Support}%
\item%
\href{https://groups.google.com/a/tensorflow.org/forum/#!forum/discuss}{TensorFlow Google Groups Developer Discussion}%
\item%
\href{https://www.tensorflow.org/resources/faq}{TensorFlow FAQ}%
\end{itemize}%
\subsection{Why TensorFlow}%
\label{subsec:WhyTensorFlow}%
\begin{itemize}[noitemsep]%
\item%
Supported by Google%
\item%
Works well on Windows, Linux, and Mac%
\item%
Excellent GPU support%
\index{GPU}%
\index{GPU}%
\item%
Python is an easy to learn programming language%
\index{Python}%
\item%
Python is extremely popular in the data science community%
\index{Python}%
\end{itemize}

\subsection{Deep Learning Tools}%
\label{subsec:DeepLearningTools}%
TensorFlow is not the only game in town. The biggest competitor to TensorFlow/Keras is PyTorch. Listed below are some of the deep learning toolkits actively being supported:%
\index{Keras}%
\index{learning}%
\index{PyTorch}%
\index{SOM}%
\index{TensorFlow}%
\par%
\begin{itemize}[noitemsep]%
\item%
\textbf{TensorFlow }%
{-} Google's deep learning API.  The focus of this class, along with Keras.%
\index{Keras}%
\index{learning}%
\item%
\textbf{Keras }%
{-} Acts as a higher{-}level to Tensorflow.%
\index{TensorFlow}%
\item%
\textbf{PyTorch }%
{-} PyTorch is an open{-}source machine learning library based on the Torch library, used for computer vision and natural language applications processing. Facebook's AI Research lab primarily develops PyTorch.%
\index{computer vision}%
\index{learning}%
\index{PyTorch}%
\index{ROC}%
\index{ROC}%
\end{itemize}%
Other deep learning tools:%
\index{learning}%
\par%
\begin{itemize}[noitemsep]%
\item%
\textbf{Deeplearning4J }%
{-} Java{-}based. Supports all major platforms. GPU support in Java!%
\index{GPU}%
\index{GPU}%
\index{Java}%
\item%
\textbf{H2O }%
{-} Java{-}based.%
\index{Java}%
\end{itemize}%
In my opinion, the two primary Python libraries for deep learning are PyTorch and Keras. Generally, PyTorch requires more lines of code to perform the deep learning applications presented in this course. This trait of PyTorch gives Keras an easier learning curve than PyTorch. However, if you are creating entirely new neural network structures in a research setting, PyTorch can make for easier access to some of the low{-}level internals of deep learning.%
\index{Keras}%
\index{learning}%
\index{neural network}%
\index{Python}%
\index{PyTorch}%
\index{SOM}%
\par

\subsection{Using TensorFlow Directly}%
\label{subsec:UsingTensorFlowDirectly}%
Most of the time in the course, we will communicate with TensorFlow using Keras%
\index{Keras}%
\index{TensorFlow}%
\cite{franccois2017deep}%
, which allows you to specify the number of hidden layers and create the neural network. TensorFlow is a low{-}level mathematics API, similar to%
\index{hidden layer}%
\index{layer}%
\index{neural network}%
\index{TensorFlow}%
\href{http://www.numpy.org/}{ Numpy}%
. However, unlike Numpy, TensorFlow is built for deep learning. TensorFlow compiles these compute graphs into highly efficient C++/%
\index{learning}%
\index{NumPy}%
\index{TensorFlow}%
\href{https://en.wikipedia.org/wiki/CUDA}{CUDA }%
code.%
\par

\subsection{TensorFlow Linear Algebra Examples}%
\label{subsec:TensorFlowLinearAlgebraExamples}%
TensorFlow is a library for linear algebra. Keras is a higher{-}level abstraction for neural networks that you build upon TensorFlow. In this section, I will demonstrate some basic linear algebra that directly employs TensorFlow and does not use Keras. First, we will see how to multiply a row and column matrix.%
\index{linear algebra}%
\index{Keras}%
\index{linear}%
\index{matrix}%
\index{neural network}%
\index{SOM}%
\index{TensorFlow}%
\par%
\begin{tcolorbox}[size=title,title=Code,breakable]%
\begin{lstlisting}[language=Python, upquote=true]
import tensorflow as tf

# Create a Constant op that produces a 1x2 matrix.  The op is
# added as a node to the default graph.
#
# The value returned by the constructor represents the output
# of the Constant op.
matrix1 = tf.constant([[3., 3.]])

# Create another Constant that produces a 2x1 matrix.
matrix2 = tf.constant([[2.],[2.]])

# Create a Matmul op that takes 'matrix1' and 'matrix2' as inputs.
# The returned value, 'product', represents the result of the matrix
# multiplication.
product = tf.matmul(matrix1, matrix2)

print(product)
print(float(product))\end{lstlisting}
\tcbsubtitle[before skip=\baselineskip]{Output}%
\begin{lstlisting}[upquote=true]
tf.Tensor([[12.]], shape=(1, 1), dtype=float32)
12.0
\end{lstlisting}
\end{tcolorbox}%
This example multiplied two TensorFlow constant tensors.  Next, we will see how to subtract a constant from a variable.%
\index{TensorFlow}%
\par%
\begin{tcolorbox}[size=title,title=Code,breakable]%
\begin{lstlisting}[language=Python, upquote=true]
import tensorflow as tf

x = tf.Variable([1.0, 2.0])
a = tf.constant([3.0, 3.0])

# Add an op to subtract 'a' from 'x'.  Run it and print the result
sub = tf.subtract(x, a)
print(sub)
print(sub.numpy())
# ==> [-2. -1.]\end{lstlisting}
\tcbsubtitle[before skip=\baselineskip]{Output}%
\begin{lstlisting}[upquote=true]
tf.Tensor([-2. -1.], shape=(2,), dtype=float32)
[-2. -1.]
\end{lstlisting}
\end{tcolorbox}%
Of course, variables are only useful if their values can be changed.  The program can accomplish this change in value by calling the assign function.%
\par%
\begin{tcolorbox}[size=title,title=Code,breakable]%
\begin{lstlisting}[language=Python, upquote=true]
x.assign([4.0, 6.0])\end{lstlisting}
\tcbsubtitle[before skip=\baselineskip]{Output}%
\begin{lstlisting}[upquote=true]
<tf.Variable 'UnreadVariable' shape=(2,) dtype=float32,
numpy=array([4., 6.], dtype=float32)>
\end{lstlisting}
\end{tcolorbox}%
The program can now perform the subtraction with this new value.%
\par%
\begin{tcolorbox}[size=title,title=Code,breakable]%
\begin{lstlisting}[language=Python, upquote=true]
sub = tf.subtract(x, a)
print(sub)
print(sub.numpy())\end{lstlisting}
\tcbsubtitle[before skip=\baselineskip]{Output}%
\begin{lstlisting}[upquote=true]
tf.Tensor([1. 3.], shape=(2,), dtype=float32)
[1. 3.]
\end{lstlisting}
\end{tcolorbox}%
In the next section, we will see a TensorFlow example that has nothing to do with neural networks.%
\index{neural network}%
\index{TensorFlow}%
\par

\subsection{TensorFlow Mandelbrot Set Example}%
\label{subsec:TensorFlowMandelbrotSetExample}%
Next, we examine another example where we use TensorFlow directly. To demonstrate that TensorFlow is mathematical and does not only provide neural networks, we will also first use it for a non{-}machine learning rendering task. The code presented here can render a%
\index{learning}%
\index{neural network}%
\index{TensorFlow}%
\href{https://en.wikipedia.org/wiki/Mandelbrot_set}{ Mandelbrot set}%
. Note, I based this code on a Mandelbrot%
\index{Mandelbrot}%
\href{https://chromium.googlesource.com/external/github.com/tensorflow/tensorflow/+/r0.10/tensorflow/g3doc/tutorials/mandelbrot/index.md}{ example }%
that I originally found with TensorFlow 1.0. I've updated the code slightly to comply with current versions of TensorFlow.%
\index{TensorFlow}%
\par%
\begin{tcolorbox}[size=title,title=Code,breakable]%
\begin{lstlisting}[language=Python, upquote=true]
# Import libraries for simulation
import tensorflow as tf
import numpy as np

# Imports for visualization
import PIL.Image
from io import BytesIO
from IPython.display import Image, display

def DisplayFractal(a, fmt='jpeg'):
  """Display an array of iteration counts as a
     colorful picture of a fractal."""
  a_cyclic = (6.28*a/20.0).reshape(list(a.shape)+[1])
  img = np.concatenate([10+20*np.cos(a_cyclic),
                        30+50*np.sin(a_cyclic),
                        155-80*np.cos(a_cyclic)], 2)
  img[a==a.max()] = 0
  a = img
  a = np.uint8(np.clip(a, 0, 255))
  f = BytesIO()
  PIL.Image.fromarray(a).save(f, fmt)
  display(Image(data=f.getvalue()))

# Use NumPy to create a 2D array of complex numbers

Y, X = np.mgrid[-1.3:1.3:0.005, -2:1:0.005]
Z = X+1j*Y

xs = tf.constant(Z.astype(np.complex64))
zs = tf.Variable(xs)
ns = tf.Variable(tf.zeros_like(xs, tf.float32))



# Operation to update the zs and the iteration count.
#
# Note: We keep computing zs after they diverge! This
#       is very wasteful! There are better, if a little
#       less simple, ways to do this.
#
for i in range(200):
    # Compute the new values of z: z^2 + x
    zs_ = zs*zs + xs

    # Have we diverged with this new value?
    not_diverged = tf.abs(zs_) < 4

    zs.assign(zs_),
    ns.assign_add(tf.cast(not_diverged, tf.float32))
    
DisplayFractal(ns.numpy())\end{lstlisting}
\tcbsubtitle[before skip=\baselineskip]{Output}%
\includegraphics[width=4in]{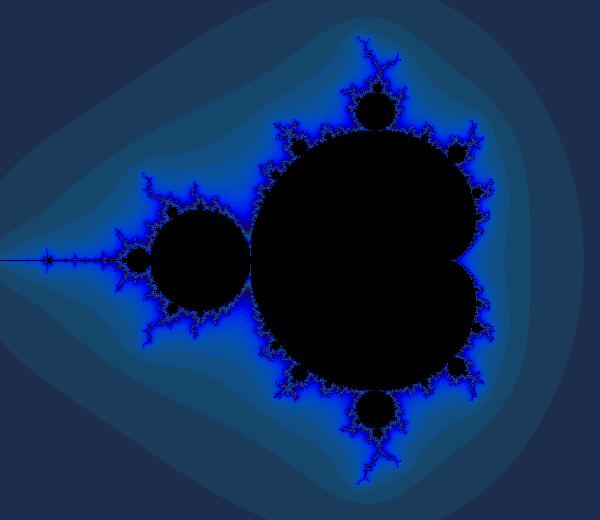}%
\end{tcolorbox}%
Mandlebrot rendering programs are both simple and infinitely complex at the same time. This view shows the entire Mandlebrot universe simultaneously, as a view completely zoomed out. However, if you zoom in on any non{-}black portion of the plot, you will find infinite hidden complexity.%
\par

\subsection{Introduction to Keras}%
\label{subsec:IntroductiontoKeras}%
\href{https://keras.io/}{Keras }%
is a layer on top of Tensorflow that makes it much easier to create neural networks. Rather than define the graphs, as you see above, you set the individual layers of the network with a much more high{-}level API. Unless you are researching entirely new structures of deep neural networks, it is unlikely that you need to program TensorFlow directly.%
\index{layer}%
\index{neural network}%
\index{TensorFlow}%
\par%
\textbf{For this class, we will usually use TensorFlow through Keras, rather than direct TensorFlow}%
\par

\subsection{Simple TensorFlow Regression: MPG}%
\label{subsec:SimpleTensorFlowRegressionMPG}%
This example shows how to encode the MPG dataset for regression. This dataset is slightly more complicated than Iris because:%
\index{dataset}%
\index{iris}%
\index{regression}%
\par%
\begin{itemize}[noitemsep]%
\item%
Input has both numeric and categorical%
\index{categorical}%
\index{input}%
\item%
Input has missing values%
\index{input}%
\end{itemize}%
This example uses functions defined above in this notepad, the "%
\href{https://github.com/jeffheaton/t81_558_deep_learning/blob/master/jeffs_helpful.ipynb}{helpful functions}%
". These functions allow you to build the feature vector for a neural network. Consider the following:%
\index{feature}%
\index{neural network}%
\index{vector}%
\par%
To encode categorical values that are part of the feature vector, use the functions from above if the categorical value is the target (as was the case with Iris, use the same technique as Iris). The iris technique allows you to decode back to Iris text strings from the predictions.%
\index{categorical}%
\index{feature}%
\index{iris}%
\index{predict}%
\index{vector}%
\par%
\begin{tcolorbox}[size=title,title=Code,breakable]%
\begin{lstlisting}[language=Python, upquote=true]
from tensorflow.keras.models import Sequential
from tensorflow.keras.layers import Dense, Activation
import pandas as pd
import io
import os
import requests
import numpy as np
from sklearn import metrics

df = pd.read_csv(
    "https://data.heatonresearch.com/data/t81-558/auto-mpg.csv", 
    na_values=['NA', '?'])

cars = df['name']

# Handle missing value
df['horsepower'] = df['horsepower'].fillna(df['horsepower'].median())

# Pandas to Numpy
x = df[['cylinders', 'displacement', 'horsepower', 'weight',
       'acceleration', 'year', 'origin']].values
y = df['mpg'].values # regression

# Build the neural network
model = Sequential()
model.add(Dense(25, input_dim=x.shape[1], activation='relu')) # Hidden 1
model.add(Dense(10, activation='relu')) # Hidden 2
model.add(Dense(1)) # Output
model.compile(loss='mean_squared_error', optimizer='adam')
model.fit(x,y,verbose=2,epochs=100)\end{lstlisting}
\tcbsubtitle[before skip=\baselineskip]{Output}%
\begin{lstlisting}[upquote=true]
...
13/13 - 0s - loss: 139.3435
Epoch 100/100
13/13 - 0s - loss: 135.2217
\end{lstlisting}
\end{tcolorbox}

\subsection{Introduction to Neural Network Hyperparameters}%
\label{subsec:IntroductiontoNeuralNetworkHyperparameters}%
If you look at the above code, you will see that the neural network contains four layers. The first layer is the input layer because it contains the%
\index{input}%
\index{input layer}%
\index{layer}%
\index{neural network}%
\textbf{ input\_dim }%
parameter that the programmer sets to be the number of inputs the dataset has. The network needs one input neuron for every column in the data set (including dummy variables).%
\index{dataset}%
\index{input}%
\index{input neuron}%
\index{neuron}%
\index{parameter}%
\par%
There are also several hidden layers, with 25 and 10 neurons each. You might be wondering how the programmer chose these numbers. Selecting a hidden neuron structure is one of the most common questions about neural networks. Unfortunately, there is no right answer. These are hyperparameters. They are settings that can affect neural network performance, yet there are no clearly defined means of setting them.%
\index{hidden layer}%
\index{hidden neuron}%
\index{hyperparameter}%
\index{layer}%
\index{neural network}%
\index{neuron}%
\index{parameter}%
\par%
In general, more hidden neurons mean more capability to fit complex problems. However, too many neurons can lead to overfitting and lengthy training times. Too few can lead to underfitting the problem and will sacrifice accuracy. Also, how many layers you have is another hyperparameter. In general, more layers allow the neural network to perform more of its feature engineering and data preprocessing. But this also comes at the expense of training times and the risk of overfitting. In general, you will see that neuron counts start larger near the input layer and tend to shrink towards the output layer in a triangular fashion.%
\index{feature}%
\index{hidden neuron}%
\index{hyperparameter}%
\index{input}%
\index{input layer}%
\index{layer}%
\index{neural network}%
\index{neuron}%
\index{output}%
\index{output layer}%
\index{overfitting}%
\index{parameter}%
\index{ROC}%
\index{ROC}%
\index{training}%
\par%
Some techniques use machine learning to optimize these values. These will be discussed in%
\index{learning}%
\index{SOM}%
\href{t81_558_class_08_3_keras_hyperparameters.ipynb}{ Module 8.3}%
.%
\par

\subsection{Controlling the Amount of Output}%
\label{subsec:ControllingtheAmountofOutput}%
The program produces one line of output for each training epoch. You can eliminate this output by setting the verbose setting of the fit command:%
\index{output}%
\index{training}%
\par%
\begin{itemize}[noitemsep]%
\item%
\textbf{verbose=0 }%
{-} No progress output (use with Jupyter if you do not want output).%
\index{output}%
\item%
\textbf{verbose=1 }%
{-} Display progress bar, does not work well with Jupyter.%
\item%
\textbf{verbose=2 }%
{-} Summary progress output (use with Jupyter if you want to know the loss at each epoch).%
\index{output}%
\end{itemize}

\subsection{Regression Prediction}%
\label{subsec:RegressionPrediction}%
Next, we will perform actual predictions. The program assigns these predictions to the%
\index{predict}%
\textbf{ pred }%
variable. These are all MPG predictions from the neural network. Notice that this is a 2D array? You can always see the dimensions of what Keras returns by printing out%
\index{Keras}%
\index{neural network}%
\index{predict}%
\textbf{ pred.shape}%
. Neural networks can return multiple values, so the result is always an array. Here the neural network only returns one value per prediction (there are 398 cars, so 398 predictions). However, a 2D range is needed because the neural network has the potential of returning more than one value.%
\index{neural network}%
\index{predict}%
\par%
\begin{tcolorbox}[size=title,title=Code,breakable]%
\begin{lstlisting}[language=Python, upquote=true]
pred = model.predict(x)
print(f"Shape: {pred.shape}")
print(pred[0:10])\end{lstlisting}
\tcbsubtitle[before skip=\baselineskip]{Output}%
\begin{lstlisting}[upquote=true]
Shape: (398, 1)
[[22.539425]
 [27.995203]
 [25.851433]
 [25.711117]
 [23.701847]
 [31.893755]
 [35.556503]
 [34.45243 ]
 [36.27014 ]
 [31.358776]]
\end{lstlisting}
\end{tcolorbox}%
We would like to see how good these predictions are.  We know the correct MPG for each car so we can measure how close the neural network was.%
\index{neural network}%
\index{predict}%
\par%
\begin{tcolorbox}[size=title,title=Code,breakable]%
\begin{lstlisting}[language=Python, upquote=true]
# Measure RMSE error.  RMSE is common for regression.
score = np.sqrt(metrics.mean_squared_error(pred,y))
print(f"Final score (RMSE): {score}")\end{lstlisting}
\tcbsubtitle[before skip=\baselineskip]{Output}%
\begin{lstlisting}[upquote=true]
Final score (RMSE): 11.552907365195134
\end{lstlisting}
\end{tcolorbox}%
The number printed above is the average number of predictions above or below the expected output. We can also print out the first ten cars with predictions and actual MPG.%
\index{output}%
\index{predict}%
\par%
\begin{tcolorbox}[size=title,title=Code,breakable]%
\begin{lstlisting}[language=Python, upquote=true]
# Sample predictions
for i in range(10):
    print(f"{i+1}. Car name: {cars[i]}, MPG: {y[i]}, " 
          + f"predicted MPG: {pred[i]}")\end{lstlisting}
\tcbsubtitle[before skip=\baselineskip]{Output}%
\begin{lstlisting}[upquote=true]
1. Car name: chevrolet chevelle malibu, MPG: 18.0, predicted MPG:
[22.539425]
2. Car name: buick skylark 320, MPG: 15.0, predicted MPG: [27.995203]
3. Car name: plymouth satellite, MPG: 18.0, predicted MPG: [25.851433]
4. Car name: amc rebel sst, MPG: 16.0, predicted MPG: [25.711117]
5. Car name: ford torino, MPG: 17.0, predicted MPG: [23.701847]
6. Car name: ford galaxie 500, MPG: 15.0, predicted MPG: [31.893755]
7. Car name: chevrolet impala, MPG: 14.0, predicted MPG: [35.556503]
8. Car name: plymouth fury iii, MPG: 14.0, predicted MPG: [34.45243]
9. Car name: pontiac catalina, MPG: 14.0, predicted MPG: [36.27014]
10. Car name: amc ambassador dpl, MPG: 15.0, predicted MPG:
[31.358776]
\end{lstlisting}
\end{tcolorbox}

\subsection{Simple TensorFlow Classification: Iris}%
\begin{figure}[h]%
\centering%
\includegraphics[width=4in]{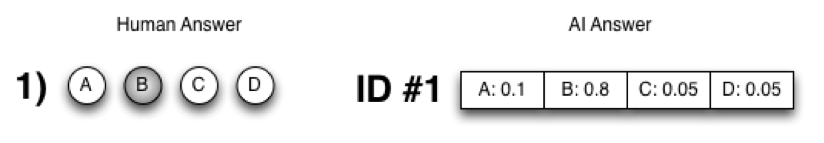}%
\caption{Classification Neural Network Output}%
\label{3.EXAM}%
\end{figure}

\label{subsec:SimpleTensorFlowClassificationIris}%
Classification is how a neural network attempts to classify the input into one or more classes.  The simplest way of evaluating a classification network is to track the percentage of training set items classified incorrectly.  We typically score human results in this manner.  For example, you might have taken multiple{-}choice exams in school in which you had to shade in a bubble for choices A, B, C, or D.  If you chose the wrong letter on a 10{-}question exam, you would earn a 90\%.  In the same way, we can grade computers; however, most classification algorithms do not merely choose A, B, C, or D.  Computers typically report a classification as their percent confidence in each class.  Figure \ref{3.EXAM} shows how a computer and a human might respond to question number 1 on an exam.%
\index{algorithm}%
\index{classification}%
\index{input}%
\index{neural network}%
\index{training}%
\par%

\par%
As you can see, the human test taker marked the first question as "B." However, the computer test taker had an 80\% (0.8) confidence in "B" and was also somewhat sure with 10\% (0.1) on "A." The computer then distributed the remaining points to the other two.  In the simplest sense, the machine would get 80\% of the score for this question if the correct answer were "B." The computer would get only 5\% (0.05) of the points if the correct answer were "D."%
\index{SOM}%
\par%
We just saw a straightforward example of how to perform the Iris classification using TensorFlow.  The iris.csv file is used rather than using the built{-}in data that many Google examples require.%
\index{classification}%
\index{CSV}%
\index{iris}%
\index{TensorFlow}%
\par%
\textbf{Make sure that you always run previous code blocks.  If you run the code block below, without the code block above, you will get errors}%
\par%
\begin{tcolorbox}[size=title,title=Code,breakable]%
\begin{lstlisting}[language=Python, upquote=true]
import pandas as pd
import io
import requests
import numpy as np
from sklearn import metrics
from tensorflow.keras.models import Sequential
from tensorflow.keras.layers import Dense, Activation
from tensorflow.keras.callbacks import EarlyStopping

df = pd.read_csv(
    "https://data.heatonresearch.com/data/t81-558/iris.csv", 
    na_values=['NA', '?'])

# Convert to numpy - Classification
x = df[['sepal_l', 'sepal_w', 'petal_l', 'petal_w']].values
dummies = pd.get_dummies(df['species']) # Classification
species = dummies.columns
y = dummies.values


# Build neural network
model = Sequential()
model.add(Dense(50, input_dim=x.shape[1], activation='relu')) # Hidden 1
model.add(Dense(25, activation='relu')) # Hidden 2
model.add(Dense(y.shape[1],activation='softmax')) # Output

model.compile(loss='categorical_crossentropy', optimizer='adam')
model.fit(x,y,verbose=2,epochs=100)\end{lstlisting}
\tcbsubtitle[before skip=\baselineskip]{Output}%
\begin{lstlisting}[upquote=true]
...
5/5 - 0s - loss: 0.0851
Epoch 100/100
5/5 - 0s - loss: 0.0880
\end{lstlisting}
\end{tcolorbox}%
\begin{tcolorbox}[size=title,title=Code,breakable]%
\begin{lstlisting}[language=Python, upquote=true]
# Print out number of species found:

print(species)\end{lstlisting}
\tcbsubtitle[before skip=\baselineskip]{Output}%
\begin{lstlisting}[upquote=true]
Index(['Iris-setosa', 'Iris-versicolor', 'Iris-virginica'],
dtype='object')
\end{lstlisting}
\end{tcolorbox}%
Now that you have a neural network trained, we would like to be able to use it. The following code makes use of our neural network. Exactly like before, we will generate predictions. Notice that three values come back for each of the 150 iris flowers. There were three types of iris (Iris{-}setosa, Iris{-}versicolor, and Iris{-}virginica).%
\index{iris}%
\index{neural network}%
\index{predict}%
\par%
\begin{tcolorbox}[size=title,title=Code,breakable]%
\begin{lstlisting}[language=Python, upquote=true]
pred = model.predict(x)
print(f"Shape: {pred.shape}")
print(pred[0:10])\end{lstlisting}
\tcbsubtitle[before skip=\baselineskip]{Output}%
\begin{lstlisting}[upquote=true]
Shape: (150, 3)
[[9.9768412e-01 2.3087766e-03 7.1474560e-06]
 [9.9349666e-01 6.4763017e-03 2.6995105e-05]
 [9.9618298e-01 3.7991456e-03 1.7790366e-05]
 [9.9207532e-01 7.8882594e-03 3.6453897e-05]
 [9.9791318e-01 2.0800228e-03 6.7602941e-06]
 [9.9684995e-01 3.1442614e-03 5.8112000e-06]
 [9.9547136e-01 4.5086881e-03 1.9946103e-05]
 [9.9625921e-01 3.7288493e-03 1.2040506e-05]
 [9.9011189e-01 9.8296851e-03 5.8434536e-05]
 [9.9447203e-01 5.5067884e-03 2.1272421e-05]]
\end{lstlisting}
\end{tcolorbox}%
If you would like to turn of scientific notation, the following line can be used:%
\par%
\begin{tcolorbox}[size=title,title=Code,breakable]%
\begin{lstlisting}[language=Python, upquote=true]
np.set_printoptions(suppress=True)\end{lstlisting}
\end{tcolorbox}%
Now we see these values rounded up.%
\par%
\begin{tcolorbox}[size=title,title=Code,breakable]%
\begin{lstlisting}[language=Python, upquote=true]
print(y[0:10])\end{lstlisting}
\tcbsubtitle[before skip=\baselineskip]{Output}%
\begin{lstlisting}[upquote=true]
[[1 0 0]
 [1 0 0]
 [1 0 0]
 [1 0 0]
 [1 0 0]
 [1 0 0]
 [1 0 0]
 [1 0 0]
 [1 0 0]
 [1 0 0]]
\end{lstlisting}
\end{tcolorbox}%
Usually, the program considers the column with the highest prediction to be the prediction of the neural network.  It is easy to convert the predictions to the expected iris species.  The argmax function finds the index of the maximum prediction for each row.%
\index{iris}%
\index{neural network}%
\index{predict}%
\index{species}%
\par%
\begin{tcolorbox}[size=title,title=Code,breakable]%
\begin{lstlisting}[language=Python, upquote=true]
predict_classes = np.argmax(pred,axis=1)
expected_classes = np.argmax(y,axis=1)
print(f"Predictions: {predict_classes}")
print(f"Expected: {expected_classes}")\end{lstlisting}
\tcbsubtitle[before skip=\baselineskip]{Output}%
\begin{lstlisting}[upquote=true]
Predictions: [0 0 0 0 0 0 0 0 0 0 0 0 0 0 0 0 0 0 0 0 0 0 0 0 0 0 0 0
0 0 0 0 0 0 0 0 0
 0 0 0 0 0 0 0 0 0 0 0 0 0 1 1 1 1 1 1 1 1 1 1 1 1 1 1 1 1 1 1 1 1 2 1
2 1
 1 1 1 1 1 1 1 1 1 2 2 1 1 1 1 1 1 1 1 1 1 1 1 1 1 1 2 2 2 2 2 2 2 2 2
2 2
 2 2 2 2 2 2 2 2 2 2 2 2 2 2 2 2 2 2 2 2 2 2 2 2 2 2 2 2 2 2 2 2 2 2 2
2 2
 2 2]
Expected: [0 0 0 0 0 0 0 0 0 0 0 0 0 0 0 0 0 0 0 0 0 0 0 0 0 0 0 0 0 0
0 0 0 0 0 0 0
 0 0 0 0 0 0 0 0 0 0 0 0 0 1 1 1 1 1 1 1 1 1 1 1 1 1 1 1 1 1 1 1 1 1 1
1 1
 1 1 1 1 1 1 1 1 1 1 1 1 1 1 1 1 1 1 1 1 1 1 1 1 1 1 2 2 2 2 2 2 2 2 2
2 2
 2 2 2 2 2 2 2 2 2 2 2 2 2 2 2 2 2 2 2 2 2 2 2 2 2 2 2 2 2 2 2 2 2 2 2
2 2
 2 2]
\end{lstlisting}
\end{tcolorbox}%
Of course, it is straightforward to turn these indexes back into iris species. We use the species list that we created earlier.%
\index{iris}%
\index{species}%
\par%
\begin{tcolorbox}[size=title,title=Code,breakable]%
\begin{lstlisting}[language=Python, upquote=true]
print(species[predict_classes[1:10]])\end{lstlisting}
\tcbsubtitle[before skip=\baselineskip]{Output}%
\begin{lstlisting}[upquote=true]
Index(['Iris-setosa', 'Iris-setosa', 'Iris-setosa', 'Iris-setosa',
       'Iris-setosa', 'Iris-setosa', 'Iris-setosa', 'Iris-setosa',
       'Iris-setosa'],
      dtype='object')
\end{lstlisting}
\end{tcolorbox}%
Accuracy might be a more easily understood error metric.  It is essentially a test score.  For all of the iris predictions, what percent were correct?  The downside is it does not consider how confident the neural network was in each prediction.%
\index{error}%
\index{iris}%
\index{neural network}%
\index{predict}%
\par%
\begin{tcolorbox}[size=title,title=Code,breakable]%
\begin{lstlisting}[language=Python, upquote=true]
from sklearn.metrics import accuracy_score

correct = accuracy_score(expected_classes,predict_classes)
print(f"Accuracy: {correct}")\end{lstlisting}
\tcbsubtitle[before skip=\baselineskip]{Output}%
\begin{lstlisting}[upquote=true]
Accuracy: 0.9733333333333334
\end{lstlisting}
\end{tcolorbox}%
The code below performs two ad hoc predictions.  The first prediction is a single iris flower, and the second predicts two iris flowers.  Notice that the%
\index{iris}%
\index{predict}%
\textbf{ argmax }%
in the second prediction requires%
\index{predict}%
\textbf{ axis=1}%
?  Since we have a 2D array now, we must specify which axis to take the%
\index{axis}%
\textbf{ argmax }%
over.  The value%
\textbf{ axis=1 }%
specifies we want the max column index for each row.%
\par%
\begin{tcolorbox}[size=title,title=Code,breakable]%
\begin{lstlisting}[language=Python, upquote=true]
sample_flower = np.array( [[5.0,3.0,4.0,2.0]], dtype=float)
pred = model.predict(sample_flower)
print(pred)
pred = np.argmax(pred)
print(f"Predict that {sample_flower} is: {species[pred]}")\end{lstlisting}
\tcbsubtitle[before skip=\baselineskip]{Output}%
\begin{lstlisting}[upquote=true]
[[0.00065001 0.17222181 0.8271282 ]]
Predict that [[5. 3. 4. 2.]] is: Iris-virginica
\end{lstlisting}
\end{tcolorbox}%
You can also predict two sample flowers.%
\index{predict}%
\par%
\begin{tcolorbox}[size=title,title=Code,breakable]%
\begin{lstlisting}[language=Python, upquote=true]
sample_flower = np.array( [[5.0,3.0,4.0,2.0],[5.2,3.5,1.5,0.8]],\
        dtype=float)
pred = model.predict(sample_flower)
print(pred)
pred = np.argmax(pred,axis=1)
print(f"Predict that these two flowers {sample_flower} ")
print(f"are: {species[pred]}")\end{lstlisting}
\tcbsubtitle[before skip=\baselineskip]{Output}%
\begin{lstlisting}[upquote=true]
[[0.00065001 0.17222157 0.8271284 ]
 [0.9887937  0.01117751 0.00002886]]
Predict that these two flowers [[5.  3.  4.  2. ]
 [5.2 3.5 1.5 0.8]]
are: Index(['Iris-virginica', 'Iris-setosa'], dtype='object')
\end{lstlisting}
\end{tcolorbox}

\section{Part 3.3: Saving and Loading a Keras Neural Network}%
\label{sec:Part3.3SavingandLoadingaKerasNeuralNetwork}%
Complex neural networks will take a long time to fit/train.  It is helpful to be able to save these neural networks so that you can reload them later.  A reloaded neural network will not require retraining.  Keras provides three formats for neural network saving.%
\index{Keras}%
\index{neural network}%
\index{training}%
\par%
\begin{itemize}[noitemsep]%
\item%
\textbf{JSON }%
{-} Stores the neural network structure (no weights) in the%
\index{neural network}%
\href{https://en.wikipedia.org/wiki/JSON}{ JSON file format}%
.%
\item%
\textbf{HDF5 }%
{-} Stores the complete neural network (with weights) in the%
\index{neural network}%
\href{https://en.wikipedia.org/wiki/Hierarchical_Data_Format}{ HDF5 file format}%
. Do not confuse HDF5 with%
\href{https://en.wikipedia.org/wiki/Apache_Hadoop}{ HDFS}%
.  They are different.  We do not use HDFS in this class.%
\end{itemize}%
Usually, you will want to save in HDF5.%
\par%
\begin{tcolorbox}[size=title,title=Code,breakable]%
\begin{lstlisting}[language=Python, upquote=true]
from tensorflow.keras.models import Sequential
from tensorflow.keras.layers import Dense, Activation
import pandas as pd
import io
import os
import requests
import numpy as np
from sklearn import metrics

save_path = "."

df = pd.read_csv(
    "https://data.heatonresearch.com/data/t81-558/auto-mpg.csv", 
    na_values=['NA', '?'])

cars = df['name']

# Handle missing value
df['horsepower'] = df['horsepower'].fillna(df['horsepower'].median())

# Pandas to Numpy
x = df[['cylinders', 'displacement', 'horsepower', 'weight',
       'acceleration', 'year', 'origin']].values
y = df['mpg'].values # regression

# Build the neural network
model = Sequential()
model.add(Dense(25, input_dim=x.shape[1], activation='relu')) # Hidden 1
model.add(Dense(10, activation='relu')) # Hidden 2
model.add(Dense(1)) # Output
model.compile(loss='mean_squared_error', optimizer='adam')
model.fit(x,y,verbose=2,epochs=100)

# Predict
pred = model.predict(x)

# Measure RMSE error.  RMSE is common for regression.
score = np.sqrt(metrics.mean_squared_error(pred,y))
print(f"Before save score (RMSE): {score}")

# save neural network structure to JSON (no weights)
model_json = model.to_json()
with open(os.path.join(save_path,"network.json"), "w") as json_file:
    json_file.write(model_json)

# save entire network to HDF5 (save everything, suggested)
model.save(os.path.join(save_path,"network.h5"))\end{lstlisting}
\tcbsubtitle[before skip=\baselineskip]{Output}%
\begin{lstlisting}[upquote=true]
...
13/13 - 0s - loss: 50.2118 - 25ms/epoch - 2ms/step
Epoch 100/100
13/13 - 0s - loss: 49.8828 - 25ms/epoch - 2ms/step
Before save score (RMSE): 7.044431690300903
\end{lstlisting}
\end{tcolorbox}%
The code below sets up a neural network and reads the data (for predictions), but it does not clear the model directory or fit the neural network. The code loads the weights from the previous fit. Now we reload the network and perform another prediction. The RMSE should match the previous one exactly if we saved and reloaded the neural network correctly.%
\index{model}%
\index{MSE}%
\index{neural network}%
\index{predict}%
\index{RMSE}%
\index{RMSE}%
\par%
\begin{tcolorbox}[size=title,title=Code,breakable]%
\begin{lstlisting}[language=Python, upquote=true]
from tensorflow.keras.models import load_model
model2 = load_model(os.path.join(save_path,"network.h5"))
pred = model2.predict(x)
# Measure RMSE error.  RMSE is common for regression.
score = np.sqrt(metrics.mean_squared_error(pred,y))
print(f"After load score (RMSE): {score}")\end{lstlisting}
\tcbsubtitle[before skip=\baselineskip]{Output}%
\begin{lstlisting}[upquote=true]
After load score (RMSE): 7.044431690300903
\end{lstlisting}
\end{tcolorbox}

\section{Part 3.4: Early Stopping in Keras to Prevent Overfitting}%
\begin{figure}[h]%
\centering%
\includegraphics[width=4in]{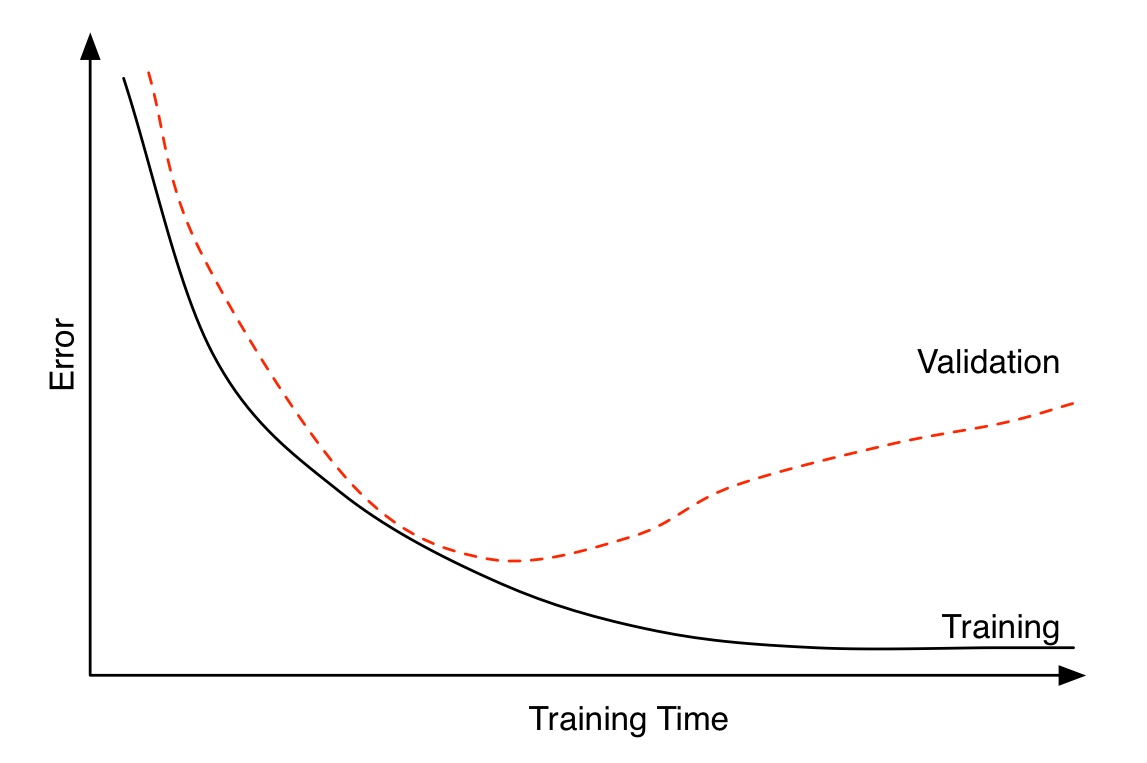}%
\caption{Training vs. Validation Error for Overfitting}%
\label{3.OVER}%
\end{figure}

\label{sec:Part3.4EarlyStoppinginKerastoPreventOverfitting}%
It can be difficult to determine how many epochs to cycle through to train a neural network. Overfitting will occur if you train the neural network for too many epochs, and the neural network will not perform well on new data, despite attaining a good accuracy on the training set. Overfitting occurs when a neural network is trained to the point that it begins to memorize rather than generalize, as demonstrated in Figure \ref{3.OVER}.%
\index{neural network}%
\index{overfitting}%
\index{training}%
\par%

\par%
It is important to segment the original dataset into several datasets:%
\index{dataset}%
\par%
\begin{itemize}[noitemsep]%
\item%
\textbf{Training Set}%
\item%
\textbf{Validation Set}%
\item%
\textbf{Holdout Set}%
\end{itemize}%
You can construct these sets in several different ways. The following programs demonstrate some of these.%
\index{SOM}%
\par
\begin{figure}[h]%
\centering%
\includegraphics[width=4in]{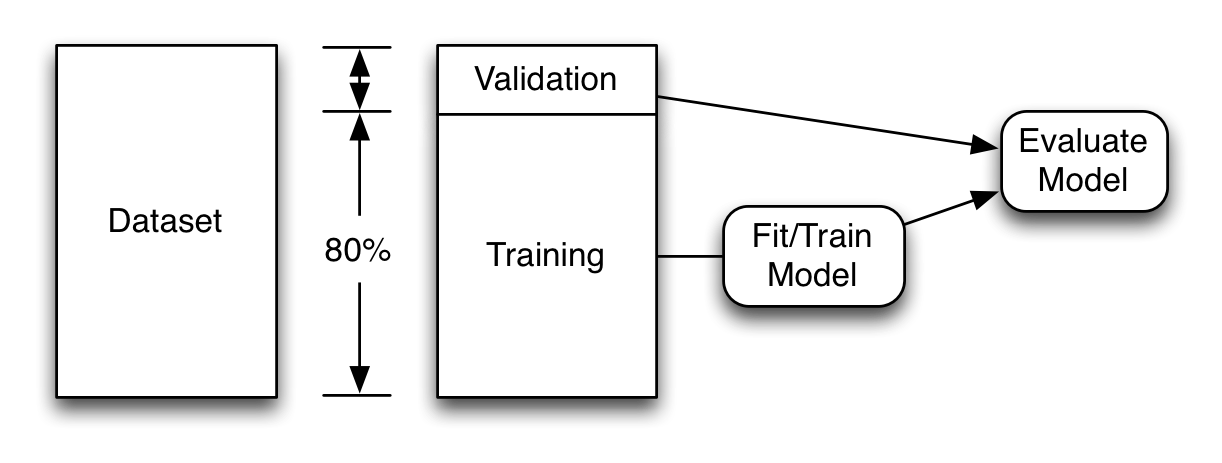}%
\caption{Training with a Validation Set}%
\label{3.VAL}%
\end{figure}

%
The first method is a training and validation set. We use the training data to train the neural network until the validation set no longer improves. This attempts to stop at a near{-}optimal training point. This method will only give accurate "out of sample" predictions for the validation set; this is usually 20\% of the data. The predictions for the training data will be overly optimistic, as these were the data that we used to train the neural network. Figure \ref{3.VAL} demonstrates how we divide the dataset.%
\index{dataset}%
\index{neural network}%
\index{predict}%
\index{training}%
\index{validation}%
\par%

\par%
\subsection{Early Stopping with Classification}%
\label{subsec:EarlyStoppingwithClassification}%
We will now see an example of classification training with early stopping. We will train the neural network until the error no longer improves on the validation set.%
\index{classification}%
\index{early stopping}%
\index{error}%
\index{neural network}%
\index{training}%
\index{validation}%
\par%
\begin{tcolorbox}[size=title,title=Code,breakable]%
\begin{lstlisting}[language=Python, upquote=true]
import pandas as pd
import io
import requests
import numpy as np
from sklearn import metrics
from sklearn.model_selection import train_test_split
from tensorflow.keras.models import Sequential
from tensorflow.keras.layers import Dense, Activation
from tensorflow.keras.callbacks import EarlyStopping

df = pd.read_csv(
    "https://data.heatonresearch.com/data/t81-558/iris.csv", 
    na_values=['NA', '?'])

# Convert to numpy - Classification
x = df[['sepal_l', 'sepal_w', 'petal_l', 'petal_w']].values
dummies = pd.get_dummies(df['species']) # Classification
species = dummies.columns
y = dummies.values

# Split into validation and training sets
x_train, x_test, y_train, y_test = train_test_split(    
    x, y, test_size=0.25, random_state=42)

# Build neural network
model = Sequential()
model.add(Dense(50, input_dim=x.shape[1], activation='relu')) # Hidden 1
model.add(Dense(25, activation='relu')) # Hidden 2
model.add(Dense(y.shape[1],activation='softmax')) # Output
model.compile(loss='categorical_crossentropy', optimizer='adam')

monitor = EarlyStopping(monitor='val_loss', min_delta=1e-3, patience=5, 
        verbose=1, mode='auto', restore_best_weights=True)
model.fit(x_train,y_train,validation_data=(x_test,y_test),
        callbacks=[monitor],verbose=2,epochs=1000)\end{lstlisting}
\tcbsubtitle[before skip=\baselineskip]{Output}%
\begin{lstlisting}[upquote=true]
Train on 112 samples, validate on 38 samples
...
112/112 - 0s - loss: 0.1017 - val_loss: 0.0926
Epoch 107/1000
Restoring model weights from the end of the best epoch.
112/112 - 0s - loss: 0.1001 - val_loss: 0.0869
Epoch 00107: early stopping
\end{lstlisting}
\end{tcolorbox}%
There are a number of parameters that are specified to the%
\index{parameter}%
\textbf{ EarlyStopping }%
object.%
\par%
\begin{itemize}[noitemsep]%
\item%
\textbf{min\_delta }%
This value should be kept small. It simply means the minimum change in error to be registered as an improvement.  Setting it even smaller will not likely have a great deal of impact.%
\index{error}%
\item%
\textbf{patience }%
How long should the training wait for the validation error to improve?%
\index{error}%
\index{training}%
\index{validation}%
\item%
\textbf{verbose }%
How much progress information do you want?%
\item%
\textbf{mode }%
In general, always set this to "auto".  This allows you to specify if the error should be minimized or maximized.  Consider accuracy, where higher numbers are desired vs log{-}loss/RMSE where lower numbers are desired.%
\index{error}%
\index{log{-}loss}%
\index{MSE}%
\index{RMSE}%
\index{RMSE}%
\item%
\textbf{restore\_best\_weights }%
This should always be set to true.  This restores the weights to the values they were at when the validation set is the highest.  Unless you are manually tracking the weights yourself (we do not use this technique in this course), you should have Keras perform this step for you.%
\index{Keras}%
\index{validation}%
\end{itemize}%
As you can see from above, the entire number of requested epochs were not used.  The neural network training stopped once the validation set no longer improved.%
\index{neural network}%
\index{training}%
\index{validation}%
\par%
\begin{tcolorbox}[size=title,title=Code,breakable]%
\begin{lstlisting}[language=Python, upquote=true]
from sklearn.metrics import accuracy_score

pred = model.predict(x_test)
predict_classes = np.argmax(pred,axis=1)
expected_classes = np.argmax(y_test,axis=1)
correct = accuracy_score(expected_classes,predict_classes)
print(f"Accuracy: {correct}")\end{lstlisting}
\tcbsubtitle[before skip=\baselineskip]{Output}%
\begin{lstlisting}[upquote=true]
Accuracy: 1.0
\end{lstlisting}
\end{tcolorbox}

\subsection{Early Stopping with Regression}%
\label{subsec:EarlyStoppingwithRegression}%
The following code demonstrates how we can apply early stopping to a regression problem.  The technique is similar to the early stopping for classification code that we just saw.%
\index{classification}%
\index{early stopping}%
\index{regression}%
\par%
\begin{tcolorbox}[size=title,title=Code,breakable]%
\begin{lstlisting}[language=Python, upquote=true]
from tensorflow.keras.models import Sequential
from tensorflow.keras.layers import Dense, Activation
import pandas as pd
import io
import os
import requests
import numpy as np
from sklearn import metrics

df = pd.read_csv(
    "https://data.heatonresearch.com/data/t81-558/auto-mpg.csv", 
    na_values=['NA', '?'])

cars = df['name']

# Handle missing value
df['horsepower'] = df['horsepower'].fillna(df['horsepower'].median())

# Pandas to Numpy
x = df[['cylinders', 'displacement', 'horsepower', 'weight',
       'acceleration', 'year', 'origin']].values
y = df['mpg'].values # regression

# Split into validation and training sets
x_train, x_test, y_train, y_test = train_test_split(    
    x, y, test_size=0.25, random_state=42)

# Build the neural network
model = Sequential()
model.add(Dense(25, input_dim=x.shape[1], activation='relu')) # Hidden 1
model.add(Dense(10, activation='relu')) # Hidden 2
model.add(Dense(1)) # Output
model.compile(loss='mean_squared_error', optimizer='adam')

monitor = EarlyStopping(monitor='val_loss', min_delta=1e-3, 
        patience=5, verbose=1, mode='auto',
        restore_best_weights=True)
model.fit(x_train,y_train,validation_data=(x_test,y_test),
        callbacks=[monitor], verbose=2,epochs=1000)\end{lstlisting}
\tcbsubtitle[before skip=\baselineskip]{Output}%
\begin{lstlisting}[upquote=true]
Train on 298 samples, validate on 100 samples
...
298/298 - 0s - loss: 34.0591 - val_loss: 29.3044
Epoch 317/1000
Restoring model weights from the end of the best epoch.
298/298 - 0s - loss: 32.9764 - val_loss: 29.1071
Epoch 00317: early stopping
\end{lstlisting}
\end{tcolorbox}%
Finally, we evaluate the error.%
\index{error}%
\par%
\begin{tcolorbox}[size=title,title=Code,breakable]%
\begin{lstlisting}[language=Python, upquote=true]
# Measure RMSE error.  RMSE is common for regression.
pred = model.predict(x_test)
score = np.sqrt(metrics.mean_squared_error(pred,y_test))
print(f"Final score (RMSE): {score}")\end{lstlisting}
\tcbsubtitle[before skip=\baselineskip]{Output}%
\begin{lstlisting}[upquote=true]
Final score (RMSE): 5.291219300799398
\end{lstlisting}
\end{tcolorbox}

\section{Part 3.5: Extracting Weights and Manual Network Calculation}%
\label{sec:Part3.5ExtractingWeightsandManualNetworkCalculation}%
\subsection{Weight Initialization}%
\label{subsec:WeightInitialization}%
The weights of a neural network determine the output for the neural network. The training process can adjust these weights, so the neural network produces useful output. Most neural network training algorithms begin by initializing the weights to a random state. Training then progresses through iterations that continuously improve the weights to produce better output.%
\index{algorithm}%
\index{continuous}%
\index{iteration}%
\index{neural network}%
\index{output}%
\index{random}%
\index{ROC}%
\index{ROC}%
\index{training}%
\index{training algorithm}%
\par%
The random weights of a neural network impact how well that neural network can be trained. If a neural network fails to train, you can remedy the problem by simply restarting with a new set of random weights. However, this solution can be frustrating when you are experimenting with the architecture of a neural network and trying different combinations of hidden layers and neurons. If you add a new layer, and the network's performance improves, you must ask yourself if this improvement resulted from the new layer or from a new set of weights. Because of this uncertainty, we look for two key attributes in a weight initialization algorithm:%
\index{algorithm}%
\index{architecture}%
\index{hidden layer}%
\index{layer}%
\index{neural network}%
\index{neuron}%
\index{random}%
\par%
\begin{itemize}[noitemsep]%
\item%
How consistently does this algorithm provide good weights?%
\index{algorithm}%
\item%
How much of an advantage do the weights of the algorithm provide?%
\index{algorithm}%
\end{itemize}%
One of the most common yet least practical approaches to weight initialization is to set the weights to random values within a specific range. Numbers between {-}1 and +1 or {-}5 and +5 are often the choice. If you want to ensure that you get the same set of random weights each time, you should use a seed. The seed specifies a set of predefined random weights to use. For example, a seed of 1000 might produce random weights of 0.5, 0.75, and 0.2. These values are still random; you cannot predict them, yet you will always get these values when you choose a seed of 1000. \newline%
Not all seeds are created equal. One problem with random weight initialization is that the random weights created by some seeds are much more difficult to train than others. The weights can be so bad that training is impossible. If you cannot train a neural network with a particular weight set, you should generate a new set of weights using a different seed.%
\index{neural network}%
\index{predict}%
\index{random}%
\index{SOM}%
\index{training}%
\par%
Because weight initialization is a problem, considerable research has been around it. By default, Keras uses the Xavier weight initialization algorithm, introduced in 2006 by Glorot  Bengio%
\index{algorithm}%
\index{Keras}%
\index{Xavier}%
\cite{glorot2010understanding}%
, produces good weights with reasonable consistency. This relatively simple algorithm uses normally distributed random numbers.%
\index{algorithm}%
\index{random}%
\par%
To use the Xavier weight initialization, it is necessary to understand that normally distributed random numbers are not the typical random numbers between 0 and 1 that most programming languages generate. Normally distributed random numbers are centered on a mean ($\mu$, mu) that is typically 0. If 0 is the center (mean), then you will get an equal number of random numbers above and below 0. The next question is how far these random numbers will venture from 0. In theory, you could end up with both positive and negative numbers close to the maximum positive and negative ranges supported by your computer. However, the reality is that you will more likely see random numbers that are between 0 and three standard deviations from the center.%
\index{random}%
\index{standard deviation}%
\index{Xavier}%
\par%
The standard deviation ($\sigma$, sigma) parameter specifies the size of this standard deviation. For example, if you specified a standard deviation of 10, you would mainly see random numbers between {-}30 and +30, and the numbers nearer to 0 have a much higher probability of being selected.%
\index{parameter}%
\index{probability}%
\index{random}%
\index{standard deviation}%
\par%
The above figure illustrates that the center, which in this case is 0, will be generated with a 0.4 (40\%) probability. Additionally, the probability decreases very quickly beyond {-}2 or +2 standard deviations. By defining the center and how large the standard deviations are, you can control the range of random numbers that you will receive.%
\index{probability}%
\index{random}%
\index{standard deviation}%
\par%
The Xavier weight initialization sets all weights to normally distributed random numbers. These weights are always centered at 0; however, their standard deviation varies depending on how many connections are present for the current layer of weights. Specifically, Equation 4.2 can determine the standard deviation:%
\index{connection}%
\index{layer}%
\index{random}%
\index{standard deviation}%
\index{Xavier}%
\par%
\vspace{2mm}%
\begin{equation*}
 Var(W) = \frac{2}{n_{in}+n_{out}} 
\end{equation*}
\vspace{2mm}%
\par
\begin{figure}[h]%
\centering%
\includegraphics[width=3in]{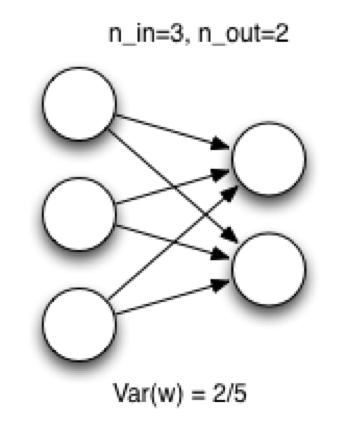}%
\caption{Xavier Weight Initialization}%
\label{3.XAVIER}%
\end{figure}

%
The above equation shows how to obtain the variance for all weights. The square root of the variance is the standard deviation. Most random number generators accept a standard deviation rather than a variance. As a result, you usually need to take the square root of the above equation. Figure \ref{3.XAVIER} shows how this algorithm might initialize one layer.%
\index{algorithm}%
\index{layer}%
\index{random}%
\index{standard deviation}%
\index{Xavier}%
\par%

\par%
We complete this process for each layer in the neural network.%
\index{layer}%
\index{neural network}%
\index{ROC}%
\index{ROC}%
\par

\subsection{Manual Neural Network Calculation}%
\label{subsec:ManualNeuralNetworkCalculation}%
This section will build a neural network and analyze it down the individual weights. We will train a simple neural network that learns the XOR function. It is not hard to hand{-}code the neurons to provide an%
\index{neural network}%
\index{neuron}%
\index{XOR}%
\href{https://en.wikipedia.org/wiki/Exclusive_or}{ XOR function}%
; however, we will allow Keras for simplicity to train this network for us. The neural network is small, with two inputs, two hidden neurons, and a single output. We will use 100K epochs on the ADAM optimizer. This approach is overkill, but it gets the result, and our focus here is not on tuning.%
\index{ADAM}%
\index{hidden neuron}%
\index{input}%
\index{Keras}%
\index{neural network}%
\index{neuron}%
\index{output}%
\par%
\begin{tcolorbox}[size=title,title=Code,breakable]%
\begin{lstlisting}[language=Python, upquote=true]
from tensorflow.keras.models import Sequential
from tensorflow.keras.layers import Dense, Activation
import numpy as np

# Create a dataset for the XOR function
x = np.array([
    [0,0],
    [1,0],
    [0,1],
    [1,1]
])

y = np.array([
    0,
    1,
    1,
    0
])

# Build the network
# sgd = optimizers.SGD(lr=0.01, decay=1e-6, momentum=0.9, nesterov=True)

done = False
cycle = 1

while not done:
    print("Cycle #{}".format(cycle))
    cycle+=1
    model = Sequential()
    model.add(Dense(2, input_dim=2, activation='relu')) 
    model.add(Dense(1)) 
    model.compile(loss='mean_squared_error', optimizer='adam')
    model.fit(x,y,verbose=0,epochs=10000)

    # Predict
    pred = model.predict(x)
    
    # Check if successful.  It takes several runs with this 
    # small of a network
    done = pred[0]<0.01 and pred[3]<0.01 and pred[1] > 0.9 \
        and pred[2] > 0.9 
    print(pred)\end{lstlisting}
\tcbsubtitle[before skip=\baselineskip]{Output}%
\begin{lstlisting}[upquote=true]
Cycle #1
[[0.49999997]
 [0.49999997]
 [0.49999997]
 [0.49999997]]
Cycle #2
[[0.33333334]
 [1.        ]
 [0.33333334]
 [0.33333334]]
Cycle #3
[[0.33333334]
 [1.        ]
 [0.33333334]
 [0.33333334]]
Cycle #4
[[0.]
 [1.]
 [1.]
 [0.]]
\end{lstlisting}
\end{tcolorbox}%
\begin{tcolorbox}[size=title,title=Code,breakable]%
\begin{lstlisting}[language=Python, upquote=true]
pred[3]\end{lstlisting}
\tcbsubtitle[before skip=\baselineskip]{Output}%
\begin{lstlisting}[upquote=true]
array([0.], dtype=float32)
\end{lstlisting}
\end{tcolorbox}%
The output above should have two numbers near 0.0 for the first and fourth spots (input {[}0,0{]} and {[}1,1{]}).  The middle two numbers should be near 1.0 (input {[}1,0{]} and {[}0,1{]}).  These numbers are in scientific notation.  Due to random starting weights, it is sometimes necessary to run the above through several cycles to get a good result.%
\index{input}%
\index{output}%
\index{random}%
\index{SOM}%
\par%
Now that we've trained the neural network, we can dump the weights.%
\index{neural network}%
\par%
\begin{tcolorbox}[size=title,title=Code,breakable]%
\begin{lstlisting}[language=Python, upquote=true]
# Dump weights
for layerNum, layer in enumerate(model.layers):
    weights = layer.get_weights()[0]
    biases = layer.get_weights()[1]
    
    for toNeuronNum, bias in enumerate(biases):
        print(f'{layerNum}B -> L{layerNum+1}N{toNeuronNum}: {bias}')
    
    for fromNeuronNum, wgt in enumerate(weights):
        for toNeuronNum, wgt2 in enumerate(wgt):
            print(f'L{layerNum}N{fromNeuronNum} \
                  -> L{layerNum+1}N{toNeuronNum} = {wgt2}')\end{lstlisting}
\tcbsubtitle[before skip=\baselineskip]{Output}%
\begin{lstlisting}[upquote=true]
0B -> L1N0: 1.3025760914331386e-08
0B -> L1N1: -1.4192625741316078e-08
L0N0                   -> L1N0 = 0.659289538860321
L0N0                   -> L1N1 = -0.9533336758613586
L0N1                   -> L1N0 = -0.659289538860321
L0N1                   -> L1N1 = 0.9533336758613586
1B -> L2N0: -1.9757269598130733e-08
L1N0                   -> L2N0 = 1.5167843103408813
L1N1                   -> L2N0 = 1.0489506721496582
\end{lstlisting}
\end{tcolorbox}%
If you rerun this, you probably get different weights.  There are many ways to solve the XOR function.%
\index{XOR}%
\par%
In the next section, we copy/paste the weights from above and recreate the calculations done by the neural network.  Because weights can change with each training, the weights used for the below code came from this:%
\index{neural network}%
\index{training}%
\par%
\begin{tcolorbox}[size=title,breakable]%
\begin{lstlisting}[upquote=true]
0B -> L1N0: -1.2913415431976318
0B -> L1N1: -3.021530048386012e-08
L0N0 -> L1N0 = 1.2913416624069214
L0N0 -> L1N1 = 1.1912699937820435
L0N1 -> L1N0 = 1.2913411855697632
L0N1 -> L1N1 = 1.1912697553634644
1B -> L2N0: 7.626241297587034e-36
L1N0 -> L2N0 = -1.548777461051941
L1N1 -> L2N0 = 0.8394404649734497
\end{lstlisting}
\end{tcolorbox}%
\begin{tcolorbox}[size=title,title=Code,breakable]%
\begin{lstlisting}[language=Python, upquote=true]
input0 = 0
input1 = 1

hidden0Sum = (input0*1.3)+(input1*1.3)+(-1.3)
hidden1Sum = (input0*1.2)+(input1*1.2)+(0)

print(hidden0Sum) # 0
print(hidden1Sum) # 1.2

hidden0 = max(0,hidden0Sum)
hidden1 = max(0,hidden1Sum)

print(hidden0) # 0
print(hidden1) # 1.2

outputSum = (hidden0*-1.6)+(hidden1*0.8)+(0)
print(outputSum) # 0.96

output = max(0,outputSum)

print(output) # 0.96\end{lstlisting}
\tcbsubtitle[before skip=\baselineskip]{Output}%
\begin{lstlisting}[upquote=true]
0.0
1.2
0
1.2
0.96
0.96
\end{lstlisting}
\end{tcolorbox}

\chapter{Training for Tabular Data}%
\label{chap:TrainingforTabularData}%
\section{Part 4.1: Encoding a Feature Vector for Keras Deep Learning}%
\label{sec:Part4.1EncodingaFeatureVectorforKerasDeepLearning}%
Neural networks can accept many types of data. We will begin with tabular data, where there are well{-}defined rows and columns. This data is what you would typically see in Microsoft Excel. Neural networks require numeric input. This numeric form is called a feature vector. Each input neurons receive one feature (or column) from this vector. Each row of training data typically becomes one vector. This section will see how to encode the following tabular data into a feature vector. You can see an example of tabular data below.%
\index{feature}%
\index{input}%
\index{input neuron}%
\index{neural network}%
\index{neuron}%
\index{tabular data}%
\index{training}%
\index{vector}%
\par%
\begin{tcolorbox}[size=title,title=Code,breakable]%
\begin{lstlisting}[language=Python, upquote=true]
import pandas as pd

pd.set_option('display.max_columns', 7) 
pd.set_option('display.max_rows', 5)

df = pd.read_csv(
    "https://data.heatonresearch.com/data/t81-558/jh-simple-dataset.csv",
    na_values=['NA','?'])

pd.set_option('display.max_columns', 9)
pd.set_option('display.max_rows', 5)

display(df)\end{lstlisting}
\tcbsubtitle[before skip=\baselineskip]{Output}%
\begin{tabular}[hbt!]{l|l|l|l|l|l|l|l|l|l}%
\hline%
&id&job&area&income&...&pop\_dense&retail\_dense&crime&product\\%
\hline%
0&1&vv&c&50876.0&...&0.885827&0.492126&0.071100&b\\%
1&2&kd&c&60369.0&...&0.874016&0.342520&0.400809&c\\%
...&...&...&...&...&...&...&...&...&...\\%
1998&1999&qp&c&67949.0&...&0.909449&0.598425&0.117803&c\\%
1999&2000&pe&c&61467.0&...&0.925197&0.539370&0.451973&c\\%
\hline%
\end{tabular}%
\vspace{2mm}%
\end{tcolorbox}%
You can make the following observations from the above data:%
\par%
\begin{itemize}[noitemsep]%
\item%
The target column is the column that you seek to predict.  There are several candidates here.  However, we will initially use the column "product".  This field specifies what product someone bought.%
\index{predict}%
\index{SOM}%
\item%
There is an ID column.  You should exclude his column because it contains no information useful for prediction.%
\index{predict}%
\item%
Many of these fields are numeric and might not require further processing.%
\index{ROC}%
\index{ROC}%
\item%
The income column does have some missing values.%
\index{SOM}%
\item%
There are categorical values: job, area, and product.%
\index{categorical}%
\end{itemize}%
To begin with, we will convert the job code into dummy variables.%
\par%
\begin{tcolorbox}[size=title,title=Code,breakable]%
\begin{lstlisting}[language=Python, upquote=true]
pd.set_option('display.max_columns', 7) 
pd.set_option('display.max_rows', 5)

dummies = pd.get_dummies(df['job'],prefix="job")
print(dummies.shape)

pd.set_option('display.max_columns', 9)
pd.set_option('display.max_rows', 10)

display(dummies)\end{lstlisting}
\tcbsubtitle[before skip=\baselineskip]{Output}%
\begin{tabular}[hbt!]{l|l|l|l|l|l|l|l|l|l}%
\hline%
&job\_11&job\_al&job\_am&job\_ax&...&job\_rn&job\_sa&job\_vv&job\_zz\\%
\hline%
0&0&0&0&0&...&0&0&1&0\\%
1&0&0&0&0&...&0&0&0&0\\%
2&0&0&0&0&...&0&0&0&0\\%
3&1&0&0&0&...&0&0&0&0\\%
4&0&0&0&0&...&0&0&0&0\\%
...&...&...&...&...&...&...&...&...&...\\%
1995&0&0&0&0&...&0&0&1&0\\%
1996&0&0&0&0&...&0&0&0&0\\%
1997&0&0&0&0&...&0&0&0&0\\%
1998&0&0&0&0&...&0&0&0&0\\%
1999&0&0&0&0&...&0&0&0&0\\%
\hline%
\end{tabular}%
\vspace{2mm}%
\begin{lstlisting}[upquote=true]
(2000, 33)
\end{lstlisting}
\end{tcolorbox}%
Because there are 33 different job codes, there are 33 dummy variables.  We also specified a prefix because the job codes (such as "ax") are not that meaningful by themselves.  Something such as "job\_ax" also tells us the origin of this field.%
\index{MSE}%
\index{SOM}%
\par%
Next, we must merge these dummies back into the main data frame.  We also drop the original "job" field, as the dummies now represent it.%
\par%
\begin{tcolorbox}[size=title,title=Code,breakable]%
\begin{lstlisting}[language=Python, upquote=true]
pd.set_option('display.max_columns', 7) 
pd.set_option('display.max_rows', 5)

df = pd.concat([df,dummies],axis=1)
df.drop('job', axis=1, inplace=True)

pd.set_option('display.max_columns', 9)
pd.set_option('display.max_rows', 10)

display(df)\end{lstlisting}
\tcbsubtitle[before skip=\baselineskip]{Output}%
\begin{tabular}[hbt!]{l|l|l|l|l|l|l|l|l|l}%
\hline%
&id&area&income&aspect&...&job\_rn&job\_sa&job\_vv&job\_zz\\%
\hline%
0&1&c&50876.0&13.100000&...&0&0&1&0\\%
1&2&c&60369.0&18.625000&...&0&0&0&0\\%
2&3&c&55126.0&34.766667&...&0&0&0&0\\%
3&4&c&51690.0&15.808333&...&0&0&0&0\\%
4&5&d&28347.0&40.941667&...&0&0&0&0\\%
...&...&...&...&...&...&...&...&...&...\\%
1995&1996&c&51017.0&38.233333&...&0&0&1&0\\%
1996&1997&d&26576.0&33.358333&...&0&0&0&0\\%
1997&1998&d&28595.0&39.425000&...&0&0&0&0\\%
1998&1999&c&67949.0&5.733333&...&0&0&0&0\\%
1999&2000&c&61467.0&16.891667&...&0&0&0&0\\%
\hline%
\end{tabular}%
\vspace{2mm}%
\end{tcolorbox}%
We also introduce dummy variables for the area column.%
\par%
\begin{tcolorbox}[size=title,title=Code,breakable]%
\begin{lstlisting}[language=Python, upquote=true]
pd.set_option('display.max_columns', 7) 
pd.set_option('display.max_rows', 5)

df = pd.concat([df,pd.get_dummies(df['area'],prefix="area")],axis=1)
df.drop('area', axis=1, inplace=True)

pd.set_option('display.max_columns', 9)
pd.set_option('display.max_rows', 10)
display(df)\end{lstlisting}
\tcbsubtitle[before skip=\baselineskip]{Output}%
\begin{tabular}[hbt!]{l|l|l|l|l|l|l|l|l|l}%
\hline%
&id&income&aspect&subscriptions&...&area\_a&area\_b&area\_c&area\_d\\%
\hline%
0&1&50876.0&13.100000&1&...&0&0&1&0\\%
1&2&60369.0&18.625000&2&...&0&0&1&0\\%
2&3&55126.0&34.766667&1&...&0&0&1&0\\%
3&4&51690.0&15.808333&1&...&0&0&1&0\\%
4&5&28347.0&40.941667&3&...&0&0&0&1\\%
...&...&...&...&...&...&...&...&...&...\\%
1995&1996&51017.0&38.233333&1&...&0&0&1&0\\%
1996&1997&26576.0&33.358333&2&...&0&0&0&1\\%
1997&1998&28595.0&39.425000&3&...&0&0&0&1\\%
1998&1999&67949.0&5.733333&0&...&0&0&1&0\\%
1999&2000&61467.0&16.891667&0&...&0&0&1&0\\%
\hline%
\end{tabular}%
\vspace{2mm}%
\end{tcolorbox}%
The last remaining transformation is to fill in missing income values.%
\par%
\begin{tcolorbox}[size=title,title=Code,breakable]%
\begin{lstlisting}[language=Python, upquote=true]
med = df['income'].median()
df['income'] = df['income'].fillna(med)\end{lstlisting}
\end{tcolorbox}%
There are more advanced ways of filling in missing values, but they require more analysis. The idea would be to see if another field might hint at what the income was. For example, it might be beneficial to calculate a median income for each area or job category. This technique is something to keep in mind for the class Kaggle competition.%
\index{Kaggle}%
\index{SOM}%
\par%
At this point, the Pandas dataframe is ready to be converted to Numpy for neural network training. We need to know a list of the columns that will make up%
\index{neural network}%
\index{NumPy}%
\index{training}%
\textit{ x }%
(the predictors or inputs) and%
\index{input}%
\index{predict}%
\textit{ y }%
(the target).%
\par%
The complete list of columns is:%
\par%
\begin{tcolorbox}[size=title,title=Code,breakable]%
\begin{lstlisting}[language=Python, upquote=true]
print(list(df.columns))\end{lstlisting}
\tcbsubtitle[before skip=\baselineskip]{Output}%
\begin{lstlisting}[upquote=true]
['id', 'income', 'aspect', 'subscriptions', 'dist_healthy',
'save_rate', 'dist_unhealthy', 'age', 'pop_dense', 'retail_dense',
'crime', 'product', 'job_11', 'job_al', 'job_am', 'job_ax', 'job_bf',
'job_by', 'job_cv', 'job_de', 'job_dz', 'job_e2', 'job_f8', 'job_gj',
'job_gv', 'job_kd', 'job_ke', 'job_kl', 'job_kp', 'job_ks', 'job_kw',
'job_mm', 'job_nb', 'job_nn', 'job_ob', 'job_pe', 'job_po', 'job_pq',
'job_pz', 'job_qp', 'job_qw', 'job_rn', 'job_sa', 'job_vv', 'job_zz',
'area_a', 'area_b', 'area_c', 'area_d']
\end{lstlisting}
\end{tcolorbox}%
This data includes both the target and predictors.  We need a list with the target removed.  We also remove%
\index{predict}%
\textbf{ id }%
because it is not useful for prediction.%
\index{predict}%
\par%
\begin{tcolorbox}[size=title,title=Code,breakable]%
\begin{lstlisting}[language=Python, upquote=true]
x_columns = df.columns.drop('product').drop('id')
print(list(x_columns))\end{lstlisting}
\tcbsubtitle[before skip=\baselineskip]{Output}%
\begin{lstlisting}[upquote=true]
['income', 'aspect', 'subscriptions', 'dist_healthy', 'save_rate',
'dist_unhealthy', 'age', 'pop_dense', 'retail_dense', 'crime',
'job_11', 'job_al', 'job_am', 'job_ax', 'job_bf', 'job_by', 'job_cv',
'job_de', 'job_dz', 'job_e2', 'job_f8', 'job_gj', 'job_gv', 'job_kd',
'job_ke', 'job_kl', 'job_kp', 'job_ks', 'job_kw', 'job_mm', 'job_nb',
'job_nn', 'job_ob', 'job_pe', 'job_po', 'job_pq', 'job_pz', 'job_qp',
'job_qw', 'job_rn', 'job_sa', 'job_vv', 'job_zz', 'area_a', 'area_b',
'area_c', 'area_d']
\end{lstlisting}
\end{tcolorbox}%
\subsection{Generate X and Y for a Classification Neural Network}%
\label{subsec:GenerateXandYforaClassificationNeuralNetwork}%
We can now generate%
\textit{ x }%
and%
\textit{ y}%
.  Note that this is how we generate y for a classification problem.  Regression would not use dummies and would encode the numeric value of the target.%
\index{classification}%
\index{regression}%
\par%
\begin{tcolorbox}[size=title,title=Code,breakable]%
\begin{lstlisting}[language=Python, upquote=true]
# Convert to numpy - Classification
x_columns = df.columns.drop('product').drop('id')
x = df[x_columns].values
dummies = pd.get_dummies(df['product']) # Classification
products = dummies.columns
y = dummies.values\end{lstlisting}
\end{tcolorbox}%
We can display the%
\textit{ x }%
and%
\textit{ y }%
matrices.%
\par%
\begin{tcolorbox}[size=title,title=Code,breakable]%
\begin{lstlisting}[language=Python, upquote=true]
print(x)
print(y)\end{lstlisting}
\tcbsubtitle[before skip=\baselineskip]{Output}%
\begin{lstlisting}[upquote=true]
[[5.08760000e+04 1.31000000e+01 1.00000000e+00 ... 0.00000000e+00
  1.00000000e+00 0.00000000e+00]
 [6.03690000e+04 1.86250000e+01 2.00000000e+00 ... 0.00000000e+00
  1.00000000e+00 0.00000000e+00]
 [5.51260000e+04 3.47666667e+01 1.00000000e+00 ... 0.00000000e+00
  1.00000000e+00 0.00000000e+00]
 ...
 [2.85950000e+04 3.94250000e+01 3.00000000e+00 ... 0.00000000e+00
  0.00000000e+00 1.00000000e+00]
 [6.79490000e+04 5.73333333e+00 0.00000000e+00 ... 0.00000000e+00
  1.00000000e+00 0.00000000e+00]
 [6.14670000e+04 1.68916667e+01 0.00000000e+00 ... 0.00000000e+00
  1.00000000e+00 0.00000000e+00]]
[[0 1 0 ... 0 0 0]
 [0 0 1 ... 0 0 0]
 [0 1 0 ... 0 0 0]
 ...
 [0 0 0 ... 0 1 0]
 [0 0 1 ... 0 0 0]
 [0 0 1 ... 0 0 0]]
\end{lstlisting}
\end{tcolorbox}%
The x and y values are now ready for a neural network.  Make sure that you construct the neural network for a classification problem.  Specifically,%
\index{classification}%
\index{neural network}%
\par%
\begin{itemize}[noitemsep]%
\item%
Classification neural networks have an output neuron count equal to the number of classes.%
\index{classification}%
\index{neural network}%
\index{neuron}%
\index{output}%
\index{output neuron}%
\item%
Classification neural networks should use%
\index{classification}%
\index{neural network}%
\textbf{ categorical\_crossentropy }%
and a%
\textbf{ softmax }%
activation function on the output layer.%
\index{activation function}%
\index{layer}%
\index{output}%
\index{output layer}%
\end{itemize}

\subsection{Generate X and Y for a Regression Neural Network}%
\label{subsec:GenerateXandYforaRegressionNeuralNetwork}%
The program generates the%
\textit{ x }%
values the say way for a regression neural network.  However,%
\index{neural network}%
\index{regression}%
\textit{ y }%
does not use dummies.  Make sure to replace%
\textbf{ income }%
with your actual target.%
\par%
\begin{tcolorbox}[size=title,title=Code,breakable]%
\begin{lstlisting}[language=Python, upquote=true]
y = df['income'].values\end{lstlisting}
\end{tcolorbox}

\subsection{Module 4 Assignment}%
\label{subsec:Module4Assignment}%
You can find the first assignment here:%
\href{https://github.com/jeffheaton/t81_558_deep_learning/blob/master/assignments/assignment_yourname_class4.ipynb}{ assignment 4}%
\par

\section{Part 4.2: Multiclass Classification with ROC and AUC}%
\label{sec:Part4.2MulticlassClassificationwithROCandAUC}%
The output of modern neural networks can be of many different forms. However, classically, neural network output has typically been one of the following:%
\index{neural network}%
\index{output}%
\par%
\begin{itemize}[noitemsep]%
\item%
\textbf{Binary Classification }%
{-} Classification between two possibilities (positive and negative). Common in medical testing, does the person has the disease (positive) or not (negative).%
\index{classification}%
\item%
\textbf{Classification }%
{-} Classification between more than 2.  The iris dataset (3{-}way classification).%
\index{classification}%
\index{dataset}%
\index{iris}%
\item%
\textbf{Regression }%
{-} Numeric prediction.  How many MPG does a car get? (covered in next video)%
\index{predict}%
\index{video}%
\end{itemize}%
We will look at some visualizations for all three in this section.%
\index{SOM}%
\index{visualization}%
\par%
It is important to evaluate the false positives and negatives in the results produced by a neural network. We will now look at assessing error for both classification and regression neural networks.%
\index{classification}%
\index{error}%
\index{false positive}%
\index{neural network}%
\index{regression}%
\par%
\subsection{Binary Classification and ROC Charts}%
\label{subsec:BinaryClassificationandROCCharts}%
Binary classification occurs when a neural network must choose between two options: true/false, yes/no, correct/incorrect, or buy/sell. To see how to use binary classification, we will consider a classification system for a credit card company. This system will either "issue a credit card" or "decline a credit card." This classification system must decide how to respond to a new potential customer.%
\index{classification}%
\index{neural network}%
\par%
When you have only two classes that you can consider, the objective function's score is the number of false{-}positive predictions versus the number of false negatives. False negatives and false positives are both types of errors, and it is essential to understand the difference. For the previous example, issuing a credit card would be positive. A false positive occurs when a model decides to issue a credit card to someone who will not make payments as agreed. A false negative happens when a model denies a credit card to someone who would have made payments as agreed.%
\index{error}%
\index{false negative}%
\index{false positive}%
\index{model}%
\index{predict}%
\index{SOM}%
\par%
Because only two options exist, we can choose the mistake that is the more serious type of error, a false positive or a false negative. For most banks issuing credit cards, a false positive is worse than a false negative. Declining a potentially good credit card holder is better than accepting a credit card holder who would cause the bank to undertake expensive collection activities.%
\index{error}%
\index{false negative}%
\index{false positive}%
\par%
Consider the following program that uses the%
\href{https://data.heatonresearch.com/data/t81-558/wcbreast_wdbc.csv}{ wcbreast\_wdbc dataset }%
to classify if a breast tumor is cancerous (malignant) or not (benign).%
\par%
\begin{tcolorbox}[size=title,title=Code,breakable]%
\begin{lstlisting}[language=Python, upquote=true]
import pandas as pd

df = pd.read_csv(
    "https://data.heatonresearch.com/data/t81-558/wcbreast_wdbc.csv",
    na_values=['NA','?'])

pd.set_option('display.max_columns', 5)
pd.set_option('display.max_rows', 5)

display(df)\end{lstlisting}
\tcbsubtitle[before skip=\baselineskip]{Output}%
\begin{tabular}[hbt!]{l|l|l|l|l|l}%
\hline%
&id&diagnosis&...&worst\_symmetry&worst\_fractal\_dimension\\%
\hline%
0&842302&M&...&0.4601&0.11890\\%
1&842517&M&...&0.2750&0.08902\\%
...&...&...&...&...&...\\%
567&927241&M&...&0.4087&0.12400\\%
568&92751&B&...&0.2871&0.07039\\%
\hline%
\end{tabular}%
\vspace{2mm}%
\end{tcolorbox}%
ROC curves can be a bit confusing. However, they are prevalent in analytics. It is essential to know how to read them. Even their name is confusing. Do not worry about their name; the receiver operating characteristic curve (ROC) comes from electrical engineering (EE).%
\index{ROC}%
\index{ROC}%
\par%
Binary classification is common in medical testing. Often you want to diagnose if someone has a disease. This diagnosis can lead to two types of errors, known as false positives and false negatives:%
\index{classification}%
\index{error}%
\index{false negative}%
\index{false positive}%
\index{SOM}%
\par
\begin{figure}[h]%
\centering%
\includegraphics[width=4in]{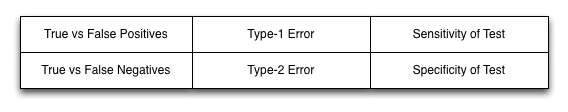}%
\caption{Type of Error}%
\label{4.ETYP}%
\end{figure}

%
\begin{itemize}[noitemsep]%
\item%
\textbf{False Positive }%
{-} Your test (neural network) indicated that the patient had the disease; however, the patient did not.%
\index{neural network}%
\item%
\textbf{False Negative }%
{-} Your test (neural network) indicated that the patient did not have the disease; however, the patient did have the disease.%
\index{neural network}%
\item%
\textbf{True Positive }%
{-} Your test (neural network) correctly identified that the patient had the disease.%
\index{neural network}%
\item%
\textbf{True Negative }%
{-} Your test (neural network) correctly identified that the patient did not have the disease.%
\index{neural network}%
\end{itemize}%
Figure \ref{4.ETYP} shows you these types of errors.%
\index{error}%
\par%

\par%
Neural networks classify in terms of the probability of it being positive. However, at what possibility do you give a positive result? Is the cutoff 50\%? 90\%? Where you set, this cutoff is called the threshold. Anything above the cutoff is positive; anything below is negative. Setting this cutoff allows the model to be more sensitive or specific:%
\index{model}%
\index{neural network}%
\index{probability}%
\par%
More info on Sensitivity vs. Specificity:%
\index{sensitivity}%
\index{specificity}%
\href{https://www.youtube.com/watch?v=Z5TtopYX1Gc}{ Khan Academy}%
\par%
\begin{tcolorbox}[size=title,title=Code,breakable]%
\begin{lstlisting}[language=Python, upquote=true]
%matplotlib inline
import matplotlib.pyplot as plt
import numpy as np
import scipy.stats as stats
import math

mu1 = -2
mu2 = 2
variance = 1
sigma = math.sqrt(variance)
x1 = np.linspace(mu1 - 5*sigma, mu1 + 4*sigma, 100)
x2 = np.linspace(mu2 - 5*sigma, mu2 + 4*sigma, 100)
plt.plot(x1, stats.norm.pdf(x1, mu1, sigma)/1,color="green", 
         linestyle='dashed')
plt.plot(x2, stats.norm.pdf(x2, mu2, sigma)/1,color="red")
plt.axvline(x=-2,color="black")
plt.axvline(x=0,color="black")
plt.axvline(x=+2,color="black")
plt.text(-2.7,0.55,"Sensitive")
plt.text(-0.7,0.55,"Balanced")
plt.text(1.7,0.55,"Specific")
plt.ylim([0,0.53])
plt.xlim([-5,5])
plt.legend(['Negative','Positive'])
plt.yticks([])
plt.show()\end{lstlisting}
\tcbsubtitle[before skip=\baselineskip]{Output}%
\includegraphics[width=3in]{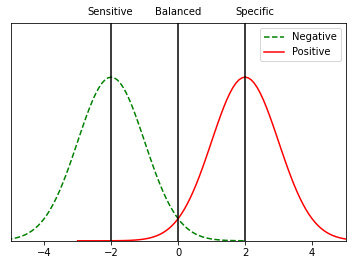}%
\end{tcolorbox}%
We will now train a neural network for the Wisconsin breast cancer dataset. We begin by preprocessing the data. Because we have all numeric data, we compute a z{-}score for each column.%
\index{dataset}%
\index{neural network}%
\index{ROC}%
\index{ROC}%
\index{Z{-}Score}%
\par%
\begin{tcolorbox}[size=title,title=Code,breakable]%
\begin{lstlisting}[language=Python, upquote=true]
from scipy.stats import zscore

x_columns = df.columns.drop('diagnosis').drop('id')
for col in x_columns:
    df[col] = zscore(df[col])

# Convert to numpy - Regression
x = df[x_columns].values
y = df['diagnosis'].map({'M':1,"B":0}).values # Binary classification, 
                                              # M is 1 and B is 0\end{lstlisting}
\end{tcolorbox}%
We can now define two functions. The first function plots a confusion matrix. The second function plots a ROC chart.%
\index{confusion matrix}%
\index{matrix}%
\index{ROC}%
\index{ROC}%
\par%
\begin{tcolorbox}[size=title,title=Code,breakable]%
\begin{lstlisting}[language=Python, upquote=true]
%matplotlib inline
import matplotlib.pyplot as plt
from sklearn.metrics import roc_curve, auc

# Plot a confusion matrix.
# cm is the confusion matrix, names are the names of the classes.
def plot_confusion_matrix(cm, names, title='Confusion matrix', 
                            cmap=plt.cm.Blues):
    plt.imshow(cm, interpolation='nearest', cmap=cmap)
    plt.title(title)
    plt.colorbar()
    tick_marks = np.arange(len(names))
    plt.xticks(tick_marks, names, rotation=45)
    plt.yticks(tick_marks, names)
    plt.tight_layout()
    plt.ylabel('True label')
    plt.xlabel('Predicted label')
    

# Plot an ROC. pred - the predictions, y - the expected output.
def plot_roc(pred,y):
    fpr, tpr, _ = roc_curve(y, pred)
    roc_auc = auc(fpr, tpr)

    plt.figure()
    plt.plot(fpr, tpr, label='ROC curve (area = %0.2f)' % roc_auc)
    plt.plot([0, 1], [0, 1], 'k--')
    plt.xlim([0.0, 1.0])
    plt.ylim([0.0, 1.05])
    plt.xlabel('False Positive Rate')
    plt.ylabel('True Positive Rate')
    plt.title('Receiver Operating Characteristic (ROC)')
    plt.legend(loc="lower right")
    plt.show()\end{lstlisting}
\end{tcolorbox}

\subsection{ROC Chart Example}%
\label{subsec:ROCChartExample}%
The following code demonstrates how to implement a ROC chart in Python.%
\index{Python}%
\index{ROC}%
\index{ROC}%
\par%
\begin{tcolorbox}[size=title,title=Code,breakable]%
\begin{lstlisting}[language=Python, upquote=true]
# Classification neural network
import numpy as np
import tensorflow.keras
from tensorflow.keras.models import Sequential
from tensorflow.keras.layers import Dense, Activation
from tensorflow.keras.callbacks import EarlyStopping
from sklearn.model_selection import train_test_split

# Split into train/test
x_train, x_test, y_train, y_test = train_test_split(    
    x, y, test_size=0.25, random_state=42)

model = Sequential()
model.add(Dense(100, input_dim=x.shape[1], activation='relu',
                kernel_initializer='random_normal'))
model.add(Dense(50,activation='relu',kernel_initializer='random_normal'))
model.add(Dense(25,activation='relu',kernel_initializer='random_normal'))
model.add(Dense(1,activation='sigmoid',kernel_initializer='random_normal'))
model.compile(loss='binary_crossentropy', 
              optimizer=tensorflow.keras.optimizers.Adam(),
              metrics =['accuracy'])
monitor = EarlyStopping(monitor='val_loss', min_delta=1e-3, 
    patience=5, verbose=1, mode='auto', restore_best_weights=True)

model.fit(x_train,y_train,validation_data=(x_test,y_test),
          callbacks=[monitor],verbose=2,epochs=1000)\end{lstlisting}
\tcbsubtitle[before skip=\baselineskip]{Output}%
\begin{lstlisting}[upquote=true]
...
14/14 - 0s - loss: 0.0458 - accuracy: 0.9836 - val_loss: 0.0486 -
val_accuracy: 0.9860 - 119ms/epoch - 8ms/step
Epoch 13/1000
Restoring model weights from the end of the best epoch: 8.
14/14 - 0s - loss: 0.0417 - accuracy: 0.9883 - val_loss: 0.0477 -
val_accuracy: 0.9860 - 124ms/epoch - 9ms/step
Epoch 13: early stopping
\end{lstlisting}
\end{tcolorbox}%
\begin{tcolorbox}[size=title,title=Code,breakable]%
\begin{lstlisting}[language=Python, upquote=true]
pred = model.predict(x_test)
plot_roc(pred,y_test)\end{lstlisting}
\tcbsubtitle[before skip=\baselineskip]{Output}%
\includegraphics[width=3in]{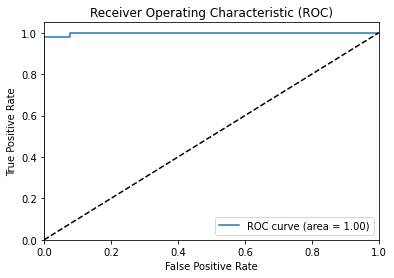}%
\end{tcolorbox}

\subsection{Multiclass Classification Error Metrics}%
\label{subsec:MulticlassClassificationErrorMetrics}%
If you want to predict more than one outcome, you will need more than one output neuron. Because a single neuron can predict two results, a neural network with two output neurons is somewhat rare. If there are three or more outcomes, there will be three or more output neurons. The following sections will examine several metrics for evaluating classification error. We will assess the following classification neural network.%
\index{classification}%
\index{error}%
\index{neural network}%
\index{neuron}%
\index{output}%
\index{output neuron}%
\index{predict}%
\index{SOM}%
\par%
\begin{tcolorbox}[size=title,title=Code,breakable]%
\begin{lstlisting}[language=Python, upquote=true]
import pandas as pd
from scipy.stats import zscore

# Read the data set
df = pd.read_csv(
    "https://data.heatonresearch.com/data/t81-558/jh-simple-dataset.csv",
    na_values=['NA','?'])

# Generate dummies for job
df = pd.concat([df,pd.get_dummies(df['job'],prefix="job")],axis=1)
df.drop('job', axis=1, inplace=True)

# Generate dummies for area
df = pd.concat([df,pd.get_dummies(df['area'],prefix="area")],axis=1)
df.drop('area', axis=1, inplace=True)

# Missing values for income
med = df['income'].median()
df['income'] = df['income'].fillna(med)

# Standardize ranges
df['income'] = zscore(df['income'])
df['aspect'] = zscore(df['aspect'])
df['save_rate'] = zscore(df['save_rate'])
df['age'] = zscore(df['age'])
df['subscriptions'] = zscore(df['subscriptions'])

# Convert to numpy - Classification
x_columns = df.columns.drop('product').drop('id')
x = df[x_columns].values
dummies = pd.get_dummies(df['product']) # Classification
products = dummies.columns
y = dummies.values\end{lstlisting}
\end{tcolorbox}%
\begin{tcolorbox}[size=title,title=Code,breakable]%
\begin{lstlisting}[language=Python, upquote=true]
# Classification neural network
import numpy as np
import tensorflow.keras
from tensorflow.keras.models import Sequential
from tensorflow.keras.layers import Dense, Activation
from tensorflow.keras.callbacks import EarlyStopping
from sklearn.model_selection import train_test_split

# Split into train/test
x_train, x_test, y_train, y_test = train_test_split(    
    x, y, test_size=0.25, random_state=42)

model = Sequential()
model.add(Dense(100, input_dim=x.shape[1], activation='relu',
                kernel_initializer='random_normal'))
model.add(Dense(50,activation='relu',kernel_initializer='random_normal'))
model.add(Dense(25,activation='relu',kernel_initializer='random_normal'))
model.add(Dense(y.shape[1],activation='softmax',
                kernel_initializer='random_normal'))
model.compile(loss='categorical_crossentropy', 
              optimizer=tensorflow.keras.optimizers.Adam(),
              metrics =['accuracy'])
monitor = EarlyStopping(monitor='val_loss', min_delta=1e-3, patience=5, 
                        verbose=1, mode='auto', restore_best_weights=True)
model.fit(x_train,y_train,validation_data=(x_test,y_test),
          callbacks=[monitor],verbose=2,epochs=1000)\end{lstlisting}
\tcbsubtitle[before skip=\baselineskip]{Output}%
\begin{lstlisting}[upquote=true]
...
47/47 - 0s - loss: 0.6624 - accuracy: 0.7147 - val_loss: 0.7527 -
val_accuracy: 0.6800 - 328ms/epoch - 7ms/step
Epoch 21/1000
Restoring model weights from the end of the best epoch: 16.
47/47 - 1s - loss: 0.6558 - accuracy: 0.7160 - val_loss: 0.7653 -
val_accuracy: 0.6720 - 527ms/epoch - 11ms/step
Epoch 21: early stopping
\end{lstlisting}
\end{tcolorbox}

\subsection{Calculate Classification Accuracy}%
\label{subsec:CalculateClassificationAccuracy}%
Accuracy is the number of rows where the neural network correctly predicted the target class.  Accuracy is only used for classification, not regression.%
\index{classification}%
\index{neural network}%
\index{predict}%
\index{regression}%
\par%
\vspace{2mm}%
\begin{equation*}
 accuracy = \frac{c}{N} 
\end{equation*}
\vspace{2mm}%
\par%
Where $c$ is the number correct and $N$ is the size of the evaluated set (training or validation). Higher accuracy numbers are desired.%
\index{training}%
\index{validation}%
\par%
As we just saw, by default, Keras will return the percent probability for each class. We can change these prediction probabilities into the actual iris predicted with%
\index{iris}%
\index{Keras}%
\index{predict}%
\index{probability}%
\textbf{ argmax}%
.%
\par%
\begin{tcolorbox}[size=title,title=Code,breakable]%
\begin{lstlisting}[language=Python, upquote=true]
pred = model.predict(x_test)
pred = np.argmax(pred,axis=1) 
# raw probabilities to chosen class (highest probability)\end{lstlisting}
\end{tcolorbox}%
Now that we have the actual iris flower predicted, we can calculate the percent accuracy (how many were correctly classified).%
\index{iris}%
\index{predict}%
\par%
\begin{tcolorbox}[size=title,title=Code,breakable]%
\begin{lstlisting}[language=Python, upquote=true]
from sklearn import metrics

y_compare = np.argmax(y_test,axis=1) 
score = metrics.accuracy_score(y_compare, pred)
print("Accuracy score: {}".format(score))\end{lstlisting}
\tcbsubtitle[before skip=\baselineskip]{Output}%
\begin{lstlisting}[upquote=true]
Accuracy score: 0.7
\end{lstlisting}
\end{tcolorbox}

\subsection{Calculate Classification Log Loss}%
\label{subsec:CalculateClassificationLogLoss}%
Accuracy is like a final exam with no partial credit.  However, neural networks can predict a probability of each of the target classes.  Neural networks will give high probabilities to predictions that are more likely.  Log loss is an error metric that penalizes confidence in wrong answers. Lower log loss values are desired.%
\index{error}%
\index{neural network}%
\index{predict}%
\index{probability}%
\par%
The following code shows the output of predict\_proba:%
\index{output}%
\index{predict}%
\par%
\begin{tcolorbox}[size=title,title=Code,breakable]%
\begin{lstlisting}[language=Python, upquote=true]
from IPython.display import display

# Don't display numpy in scientific notation
np.set_printoptions(precision=4)
np.set_printoptions(suppress=True)

# Generate predictions
pred = model.predict(x_test)

print("Numpy array of predictions")
display(pred[0:5])

print("As percent probability")
print(pred[0]*100)

score = metrics.log_loss(y_test, pred)
print("Log loss score: {}".format(score))

# raw probabilities to chosen class (highest probability)
pred = np.argmax(pred,axis=1)\end{lstlisting}
\tcbsubtitle[before skip=\baselineskip]{Output}%
\begin{lstlisting}[upquote=true]
Numpy array of predictions
array([[0.    , 0.1201, 0.7286, 0.1494, 0.0018, 0.    , 0.    ],
       [0.    , 0.6962, 0.3016, 0.0001, 0.0022, 0.    , 0.    ],
       [0.    , 0.7234, 0.2708, 0.0003, 0.0053, 0.0001, 0.    ],
       [0.    , 0.3836, 0.6039, 0.0086, 0.0039, 0.    , 0.    ],
       [0.    , 0.0609, 0.6303, 0.3079, 0.001 , 0.    , 0.    ]],
      dtype=float32)As percent probability
[ 0.0001 12.0143 72.8578 14.9446  0.1823  0.0009  0.0001]
Log loss score: 0.7423401429280638
\end{lstlisting}
\end{tcolorbox}%
\href{https://www.kaggle.com/wiki/LogarithmicLoss}{Log loss }%
is calculated as follows:%
\index{calculated}%
\par%
\vspace{2mm}%
\begin{equation*}
 \mbox{log loss} = -\frac{1}{N}\sum_{i=1}^N {( {y}_i\log(\hat{y}_i) + (1 - {y}_i)\log(1 - \hat{y}_i))} 
\end{equation*}
\vspace{2mm}%
\par%
You should use this equation only as an objective function for classifications that have two outcomes. The variable y{-}hat is the neural network's prediction, and the variable y is the known correct answer.  In this case, y will always be 0 or 1.  The training data have no probabilities. The neural network classifies it either into one class (1) or the other (0).%
\index{classification}%
\index{neural network}%
\index{predict}%
\index{training}%
\par%
The variable N represents the number of elements in the training set the number of questions in the test.  We divide by N because this process is customary for an average.  We also begin the equation with a negative because the log function is always negative over the domain 0 to 1.  This negation allows a positive score for the training to minimize.%
\index{ROC}%
\index{ROC}%
\index{training}%
\par%
You will notice two terms are separated by the addition (+).  Each contains a log function.  Because y will be either 0 or 1, then one of these two terms will cancel out to 0.  If y is 0, then the first term will reduce to 0.  If y is 1, then the second term will be 0.%
\par%
If your prediction for the first class of a two{-}class prediction is y{-}hat, then your prediction for the second class is 1 minus y{-}hat.  Essentially, if your prediction for class A is 70\% (0.7), then your prediction for class B is 30\% (0.3).  Your score will increase by the log of your prediction for the correct class.  If the neural network had predicted 1.0 for class A, and the correct answer was A, your score would increase by log (1), which is 0. For log loss, we seek a low score, so a correct answer results in 0.  Some of these log values for a neural network's probability estimate for the correct class:%
\index{neural network}%
\index{predict}%
\index{probability}%
\index{SOM}%
\par%
\begin{itemize}[noitemsep]%
\item%
{-}log(1.0) = 0%
\item%
{-}log(0.95) = 0.02%
\item%
{-}log(0.9) = 0.05%
\item%
{-}log(0.8) = 0.1%
\item%
{-}log(0.5) = 0.3%
\item%
{-}log(0.1) = 1%
\item%
{-}log(0.01) = 2%
\item%
{-}log(1.0e{-}12) = 12%
\item%
{-}log(0.0) = negative infinity%
\end{itemize}%
As you can see, giving a low confidence to the correct answer affects the score the most.  Because log (0) is negative infinity, we typically impose a minimum value.  Of course, the above log values are for a single training set element.  We will average the log values for the entire training set.%
\index{training}%
\par%
The log function is useful to penalizing wrong answers.  The following code demonstrates the utility of the log function:%
\par%
\begin{tcolorbox}[size=title,title=Code,breakable]%
\begin{lstlisting}[language=Python, upquote=true]
%matplotlib inline
from matplotlib.pyplot import figure, show
from numpy import arange, sin, pi

#t = arange(1e-5, 5.0, 0.00001)
#t = arange(1.0, 5.0, 0.00001) # computer scientists
t = arange(0.0, 1.0, 0.00001)  # data     scientists

fig = figure(1,figsize=(12, 10))

ax1 = fig.add_subplot(211)
ax1.plot(t, np.log(t))
ax1.grid(True)
ax1.set_ylim((-8, 1.5))
ax1.set_xlim((-0.1, 2))
ax1.set_xlabel('x')
ax1.set_ylabel('y')
ax1.set_title('log(x)')

show()\end{lstlisting}
\tcbsubtitle[before skip=\baselineskip]{Output}%
\includegraphics[width=4in]{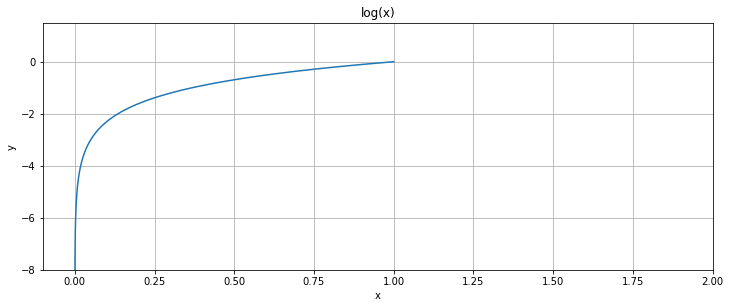}%
\end{tcolorbox}

\subsection{Confusion Matrix}%
\label{subsec:ConfusionMatrix}%
A confusion matrix shows which predicted classes are often confused for the other classes. The vertical axis (y) represents the true labels and the horizontal axis (x) represents the predicted labels. When the true label and predicted label are the same, the highest values occur down the diagonal extending from the upper left to the lower right. The other values, outside the diagonal, represent incorrect predictions. For example, in the confusion matrix below, the value in row 2, column 1 shows how often the predicted value A occurred when it should have been B.%
\index{axis}%
\index{confusion matrix}%
\index{matrix}%
\index{predict}%
\par%
\begin{tcolorbox}[size=title,title=Code,breakable]%
\begin{lstlisting}[language=Python, upquote=true]
import numpy as np
from sklearn import svm, datasets
from sklearn.model_selection import train_test_split
from sklearn.metrics import confusion_matrix

# Compute confusion matrix
cm = confusion_matrix(y_compare, pred)
np.set_printoptions(precision=2)

# Normalize the confusion matrix by row (i.e by the number of samples
# in each class)
cm_normalized = cm.astype('float') / cm.sum(axis=1)[:, np.newaxis]
print('Normalized confusion matrix')
print(cm_normalized)
plt.figure()
plot_confusion_matrix(cm_normalized, products, 
        title='Normalized confusion matrix')

plt.show()\end{lstlisting}
\tcbsubtitle[before skip=\baselineskip]{Output}%
\includegraphics[width=3in]{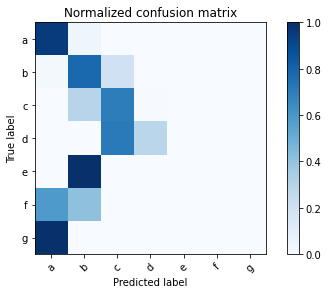}%
\begin{lstlisting}[upquote=true]
Normalized confusion matrix
[[0.95 0.05 0.   0.   0.   0.   0.  ]
 [0.02 0.78 0.2  0.   0.   0.   0.  ]
 [0.   0.29 0.7  0.01 0.   0.   0.  ]
 [0.   0.   0.71 0.29 0.   0.   0.  ]
 [0.   1.   0.   0.   0.   0.   0.  ]
 [0.59 0.41 0.   0.   0.   0.   0.  ]
 [1.   0.   0.   0.   0.   0.   0.  ]]
\end{lstlisting}
\end{tcolorbox}

\section{Part 4.3: Keras Regression for Deep Neural Networks with RMSE}%
\label{sec:Part4.3KerasRegressionforDeepNeuralNetworkswithRMSE}%
We evaluate regression results differently than classification.  Consider the following code that trains a neural network for regression on the data set%
\index{classification}%
\index{neural network}%
\index{regression}%
\textbf{ jh{-}simple{-}dataset.csv}%
.  We begin by preparing the data set.%
\par%
\begin{tcolorbox}[size=title,title=Code,breakable]%
\begin{lstlisting}[language=Python, upquote=true]
import pandas as pd
from scipy.stats import zscore
from sklearn.model_selection import train_test_split
import matplotlib.pyplot as plt

# Read the data set
df = pd.read_csv(
    "https://data.heatonresearch.com/data/t81-558/jh-simple-dataset.csv",
    na_values=['NA','?'])

# Generate dummies for job
df = pd.concat([df,pd.get_dummies(df['job'],prefix="job")],axis=1)
df.drop('job', axis=1, inplace=True)

# Generate dummies for area
df = pd.concat([df,pd.get_dummies(df['area'],prefix="area")],axis=1)
df.drop('area', axis=1, inplace=True)

# Generate dummies for product
df = pd.concat([df,pd.get_dummies(df['product'],prefix="product")],axis=1)
df.drop('product', axis=1, inplace=True)

# Missing values for income
med = df['income'].median()
df['income'] = df['income'].fillna(med)

# Standardize ranges
df['income'] = zscore(df['income'])
df['aspect'] = zscore(df['aspect'])
df['save_rate'] = zscore(df['save_rate'])
df['subscriptions'] = zscore(df['subscriptions'])

# Convert to numpy - Classification
x_columns = df.columns.drop('age').drop('id')
x = df[x_columns].values
y = df['age'].values

# Create train/test
x_train, x_test, y_train, y_test = train_test_split(    
    x, y, test_size=0.25, random_state=42)\end{lstlisting}
\end{tcolorbox}%
Next, we create a neural network to fit the data we just loaded.%
\index{neural network}%
\par%
\begin{tcolorbox}[size=title,title=Code,breakable]%
\begin{lstlisting}[language=Python, upquote=true]
from tensorflow.keras.models import Sequential
from tensorflow.keras.layers import Dense, Activation
from tensorflow.keras.callbacks import EarlyStopping

# Build the neural network
model = Sequential()
model.add(Dense(25, input_dim=x.shape[1], activation='relu')) # Hidden 1
model.add(Dense(10, activation='relu')) # Hidden 2
model.add(Dense(1)) # Output
model.compile(loss='mean_squared_error', optimizer='adam')
monitor = EarlyStopping(monitor='val_loss', min_delta=1e-3, 
                        patience=5, verbose=1, mode='auto', 
                        restore_best_weights=True)
model.fit(x_train,y_train,validation_data=(x_test,y_test),
          callbacks=[monitor],verbose=2,epochs=1000)\end{lstlisting}
\tcbsubtitle[before skip=\baselineskip]{Output}%
\begin{lstlisting}[upquote=true]
Train on 1500 samples, validate on 500 samples
...
1500/1500 - 0s - loss: 0.4081 - val_loss: 0.5540
Epoch 124/1000
Restoring model weights from the end of the best epoch.
1500/1500 - 0s - loss: 0.4353 - val_loss: 0.5538
Epoch 00124: early stopping
\end{lstlisting}
\end{tcolorbox}%
\subsection{Mean Square Error}%
\label{subsec:MeanSquareError}%
The mean square error (MSE) is the sum of the squared differences between the prediction ($\hat{y}$) and the expected ($y$).  MSE values are not of a particular unit. If an MSE value has decreased for a model, that is good. However, beyond this, there is not much more you can determine. We seek to achieve low MSE values. The following equation demonstrates how to calculate MSE.%
\index{error}%
\index{model}%
\index{MSE}%
\index{predict}%
\par%
\vspace{2mm}%
\begin{equation*}
 \mbox{MSE} = \frac{1}{n} \sum_{i=1}^n \left(\hat{y}_i - y_i\right)^2 
\end{equation*}
\vspace{2mm}%
\index{MSE}%
\par%
The following code calculates the MSE on the predictions from the neural network.%
\index{MSE}%
\index{neural network}%
\index{predict}%
\par%
\begin{tcolorbox}[size=title,title=Code,breakable]%
\begin{lstlisting}[language=Python, upquote=true]
from sklearn import metrics

# Predict
pred = model.predict(x_test)

# Measure MSE error.  
score = metrics.mean_squared_error(pred,y_test)
print("Final score (MSE): {}".format(score))\end{lstlisting}
\tcbsubtitle[before skip=\baselineskip]{Output}%
\begin{lstlisting}[upquote=true]
Final score (MSE): 0.5463447829677607
\end{lstlisting}
\end{tcolorbox}

\subsection{Root Mean Square Error}%
\label{subsec:RootMeanSquareError}%
The root mean square (RMSE) is essentially the square root of the MSE. Because of this, the RMSE error is in the same units as the training data outcome. We desire Low RMSE values. The following equation calculates RMSE.%
\index{error}%
\index{MSE}%
\index{RMSE}%
\index{RMSE}%
\index{training}%
\par%
\vspace{2mm}%
\begin{equation*}
 \mbox{RMSE} = \sqrt{\frac{1}{n} \sum_{i=1}^n \left(\hat{y}_i - y_i\right)^2} 
\end{equation*}
\vspace{2mm}%
\index{MSE}%
\index{RMSE}%
\index{RMSE}%
\par%
\begin{tcolorbox}[size=title,title=Code,breakable]%
\begin{lstlisting}[language=Python, upquote=true]
import numpy as np

# Measure RMSE error.  RMSE is common for regression.
score = np.sqrt(metrics.mean_squared_error(pred,y_test))
print("Final score (RMSE): {}".format(score))\end{lstlisting}
\tcbsubtitle[before skip=\baselineskip]{Output}%
\begin{lstlisting}[upquote=true]
Final score (RMSE): 0.7391513938076291
\end{lstlisting}
\end{tcolorbox}

\subsection{Lift Chart}%
\label{subsec:LiftChart}%
We often visualize the results of regression with a lift chart. To generate a lift chart, perform the following activities:%
\index{regression}%
\par%
\begin{itemize}[noitemsep]%
\item%
Sort the data by expected output and plot these values.%
\index{output}%
\item%
For every point on the x{-}axis, plot that same data point's predicted value in another color.%
\index{axis}%
\index{predict}%
\item%
The x{-}axis is just 0 to 100\% of the dataset. The expected always starts low and ends high.%
\index{axis}%
\index{dataset}%
\item%
The y{-}axis is ranged according to the values predicted.%
\index{axis}%
\index{predict}%
\end{itemize}%
You can interpret the lift chart as follows:%
\par%
\begin{itemize}[noitemsep]%
\item%
The expected and predict lines should be close. Notice where one is above the other.%
\index{predict}%
\item%
The below chart is the most accurate for lower ages.%
\end{itemize}%
\begin{tcolorbox}[size=title,title=Code,breakable]%
\begin{lstlisting}[language=Python, upquote=true]
# Regression chart.
def chart_regression(pred, y, sort=True):
    t = pd.DataFrame({'pred': pred, 'y': y.flatten()})
    if sort:
        t.sort_values(by=['y'], inplace=True)
    plt.plot(t['y'].tolist(), label='expected')
    plt.plot(t['pred'].tolist(), label='prediction')
    plt.ylabel('output')
    plt.legend()
    plt.show()
    
# Plot the chart
chart_regression(pred.flatten(),y_test)\end{lstlisting}
\tcbsubtitle[before skip=\baselineskip]{Output}%
\includegraphics[width=3in]{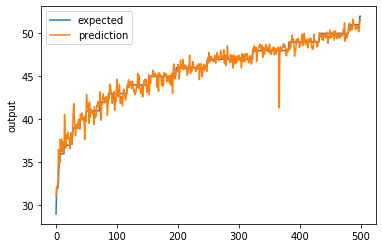}%
\end{tcolorbox}

\section{Part 4.4: Training Neural Networks}%
\label{sec:Part4.4TrainingNeuralNetworks}%
Backpropagation%
\index{backpropagation}%
\cite{rumelhart1986learning}%
is one of the most common methods for training a neural network. Rumelhart, Hinton,  Williams introduced backpropagation, and it remains popular today. Programmers frequently train deep neural networks with backpropagation because it scales really well when run on graphical processing units (GPUs). To understand this algorithm for neural networks, we must examine how to train it as well as how it processes a pattern.%
\index{algorithm}%
\index{backpropagation}%
\index{GPU}%
\index{GPU}%
\index{Hinton}%
\index{neural network}%
\index{ROC}%
\index{ROC}%
\index{training}%
\par%
Researchers have extended classic backpropagation and modified to give rise to many different training algorithms. This section will discuss the most commonly used training algorithms for neural networks. We begin with classic backpropagation and end the chapter with stochastic gradient descent (SGD).%
\index{algorithm}%
\index{backpropagation}%
\index{gradient}%
\index{gradient descent}%
\index{neural network}%
\index{SGD}%
\index{stochastic}%
\index{stochastic gradient descent}%
\index{training}%
\index{training algorithm}%
\par%
Backpropagation is the primary means of determining a neural network's weights during training. Backpropagation works by calculating a weight change amount ($v_t$) for every weight($\theta$, theta) in the neural network. This value is subtracted from every weight by the following equation:%
\index{backpropagation}%
\index{neural network}%
\index{training}%
\par%
\vspace{2mm}%
\begin{equation*}
 \theta_t = \theta_{t-1} - v_t 
\end{equation*}
\vspace{2mm}%
\par%
We repeat this process for every iteration($t$). The training algorithm determines how we calculate the weight change. Classic backpropagation calculates a gradient ($\nabla$, nabla) for every weight in the neural network for the neural network's error function ($J$). We scale the gradient by a learning rate ($\eta$, eta).%
\index{algorithm}%
\index{backpropagation}%
\index{error}%
\index{error function}%
\index{gradient}%
\index{iteration}%
\index{learning}%
\index{learning rate}%
\index{neural network}%
\index{ROC}%
\index{ROC}%
\index{training}%
\index{training algorithm}%
\par%
\vspace{2mm}%
\begin{equation*}
 v_t = \eta \nabla_{\theta_{t-1}} J(\theta_{t-1}) 
\end{equation*}
\vspace{2mm}%
\par%
The learning rate is an important concept for backpropagation training. Setting the learning rate can be complex:%
\index{backpropagation}%
\index{learning}%
\index{learning rate}%
\index{Propagation Training}%
\index{training}%
\par%
\begin{itemize}[noitemsep]%
\item%
Too low a learning rate will usually converge to a reasonable solution; however, the process will be prolonged.%
\index{learning}%
\index{learning rate}%
\index{ROC}%
\index{ROC}%
\item%
Too high of a learning rate will either fail outright or converge to a higher error than a better learning rate.%
\index{error}%
\index{learning}%
\index{learning rate}%
\end{itemize}%
Common values for learning rate are: 0.1, 0.01, 0.001, etc.%
\index{learning}%
\index{learning rate}%
\par%
Backpropagation is a gradient descent type, and many texts will use these two terms interchangeably. Gradient descent refers to calculating a gradient on each weight in the neural network for each training element. Because the neural network will not output the expected value for a training element, the gradient of each weight will indicate how to modify each weight to achieve the expected output. If the neural network did output exactly what was expected, the gradient for each weight would be 0, indicating that no change to the weight is necessary.%
\index{backpropagation}%
\index{gradient}%
\index{gradient descent}%
\index{neural network}%
\index{output}%
\index{training}%
\par%
The gradient is the derivative of the error function at the weight's current value. The error function measures the distance of the neural network's output from the expected output. We can use gradient descent, a process in which each weight's gradient value can reach even lower values of the error function.%
\index{derivative}%
\index{error}%
\index{error function}%
\index{gradient}%
\index{gradient descent}%
\index{neural network}%
\index{output}%
\index{ROC}%
\index{ROC}%
\par%
The gradient is the partial derivative of each weight in the neural network concerning the error function. Each weight has a gradient that is the slope of the error function. Weight is a connection between two neurons. Calculating the gradient of the error function allows the training method to determine whether it should increase or decrease the weight. In turn, this determination will decrease the error of the neural network. The error is the difference between the expected output and actual output of the neural network. Many different training methods called propagation{-}training algorithms utilize gradients. In all of them, the sign of the gradient tells the neural network the following information:%
\index{algorithm}%
\index{connection}%
\index{derivative}%
\index{error}%
\index{error function}%
\index{gradient}%
\index{neural network}%
\index{neuron}%
\index{output}%
\index{partial derivative}%
\index{slope}%
\index{training}%
\index{training algorithm}%
\par%
\begin{itemize}[noitemsep]%
\item%
Zero gradient {-} The weight does not contribute to the neural network's error.%
\index{error}%
\index{gradient}%
\index{neural network}%
\item%
Negative gradient {-} The algorithm should increase the weight to lower error.%
\index{algorithm}%
\index{error}%
\index{gradient}%
\item%
Positive gradient {-} The algorithm should decrease the weight to lower error.%
\index{algorithm}%
\index{error}%
\index{gradient}%
\end{itemize}%
Because many algorithms depend on gradient calculation, we will begin with an analysis of this process. First of all, let's examine the gradient. Essentially, training is a search for the set of weights that will cause the neural network to have the lowest error for a training set. If we had infinite computation resources, we would try every possible combination of weights to determine the one that provided the lowest error during the training.%
\index{algorithm}%
\index{error}%
\index{gradient}%
\index{gradient calculation}%
\index{neural network}%
\index{ROC}%
\index{ROC}%
\index{training}%
\par%
Because we do not have unlimited computing resources, we have to use some shortcuts to prevent the need to examine every possible weight combination. These training methods utilize clever techniques to avoid performing a brute{-}force search of all weight values. This type of exhaustive search would be impossible because even small networks have an infinite number of weight combinations.%
\index{SOM}%
\index{training}%
\par
\begin{figure}[h]%
\centering%
\includegraphics[width=4in]{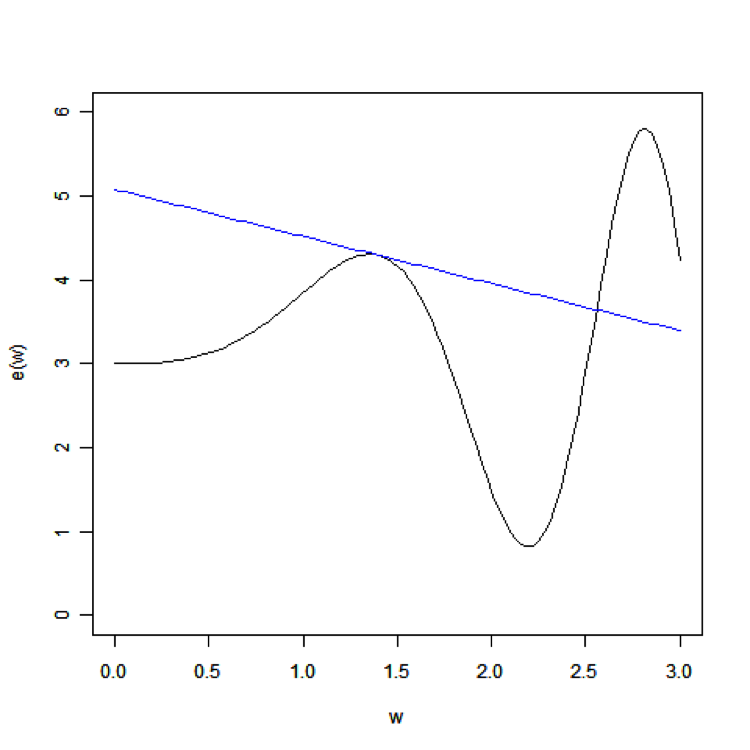}%
\caption{Derivative}%
\label{4.DRV}%
\end{figure}

%
Consider a chart that shows the error of a neural network for each possible weight. Figure \ref{4.DRV} is a graph that demonstrates the error for a single weight:%
\index{error}%
\index{neural network}%
\par%

\par%
Looking at this chart, you can easily see that the optimal weight is where the line has the lowest y{-}value. The problem is that we see only the error for the current value of the weight; we do not see the entire graph because that process would require an exhaustive search. However, we can determine the slope of the error curve at a particular weight. In the above chart, we see the slope of the error curve at 1.5. The straight line barely touches the error curve at 1.5 gives the slope. In this case, the slope, or gradient, is {-}0.5622. The negative slope indicates that an increase in the weight will lower the error.\newline%
The gradient is the instantaneous slope of the error function at the specified weight. The derivative of the error curve at that point gives the gradient. This line tells us the steepness of the error function at the given weight.%
\index{derivative}%
\index{error}%
\index{error function}%
\index{gradient}%
\index{ROC}%
\index{ROC}%
\index{slope}%
\linebreak%
Derivatives are one of the most fundamental concepts in calculus. For this book, you need to understand that a derivative provides the slope of a function at a specific point. A training technique and this slope can give you the information to adjust the weight for a lower error. Using our working definition of the gradient, we will show how to calculate it.%
\index{calculus}%
\index{derivative}%
\index{error}%
\index{gradient}%
\index{slope}%
\index{training}%
\par%
\subsection{Momentum Backpropagation}%
\label{subsec:MomentumBackpropagation}%
Momentum adds another term to the calculation of $v_t$:%
\index{momentum}%
\par%
\vspace{2mm}%
\begin{equation*}
 v_t = \eta \nabla_{\theta_{t-1}} J(\theta_{t-1}) + \lambda v_{t-1} 
\end{equation*}
\vspace{2mm}%
\par%
Like the learning rate, momentum adds another training parameter that scales the effect of momentum. Momentum backpropagation has two training parameters: learning rate ($\eta$, eta) and momentum ($\lambda$, lambda). Momentum adds the scaled value of the previous weight change amount ($v_{t-1}$) to the current weight change amount($v_t$).%
\index{backpropagation}%
\index{learning}%
\index{learning rate}%
\index{momentum}%
\index{parameter}%
\index{training}%
\par
\begin{figure}[h]%
\centering%
\includegraphics[width=3in]{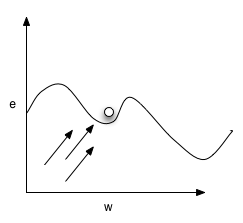}%
\caption{Momentum}%
\label{4.MTM}%
\end{figure}

%
This technique has the effect of adding additional force behind the direction a weight is moving. Figure \ref{4.MTM} shows how this might allow the weight to escape local minima.%
\par%

\par%
A typical value for momentum is 0.9.%
\index{momentum}%
\par

\subsection{Batch and Online Backpropagation}%
\label{subsec:BatchandOnlineBackpropagation}%
How often should the weights of a neural network be updated?  We can calculate gradients for a training set element.  These gradients can also be summed together into batches, and the weights updated once per batch.%
\index{gradient}%
\index{neural network}%
\index{training}%
\par%
\begin{itemize}[noitemsep]%
\item%
\textbf{Online Training }%
{-} Update the weights based on gradients calculated from a single training set element.%
\index{calculated}%
\index{gradient}%
\index{training}%
\item%
\textbf{Batch Training }%
{-} Update the weights based on the sum of the gradients over all training set elements.%
\index{gradient}%
\index{training}%
\item%
\textbf{Batch Size }%
{-} Update the weights based on the sum of some batch size of training set elements.%
\index{SOM}%
\index{training}%
\item%
\textbf{Mini{-}Batch Training }%
{-} The same as batch size, but with minimal batch size.  Mini{-}batches are very popular, often in the 32{-}64 element range.%
\index{mini{-}batch}%
\end{itemize}%
Because the batch size is smaller than the full training set size, it may take several batches to make it completely through the training set.%
\index{training}%
\par%
\begin{itemize}[noitemsep]%
\item%
\textbf{Step/Iteration }%
{-} The number of processed batches.%
\index{ROC}%
\index{ROC}%
\item%
\textbf{Epoch }%
{-} The number of times the algorithm processed the complete training set.%
\index{algorithm}%
\index{ROC}%
\index{ROC}%
\index{training}%
\end{itemize}

\subsection{Stochastic Gradient Descent}%
\label{subsec:StochasticGradientDescent}%
Stochastic gradient descent (SGD) is currently one of the most popular neural network training algorithms.  It works very similarly to Batch/Mini{-}Batch training, except that the batches are made up of a random set of training elements.%
\index{algorithm}%
\index{gradient}%
\index{gradient descent}%
\index{mini{-}batch}%
\index{neural network}%
\index{random}%
\index{SGD}%
\index{stochastic}%
\index{stochastic gradient descent}%
\index{training}%
\index{training algorithm}%
\par
\begin{figure}[h]%
\centering%
\includegraphics[width=4in]{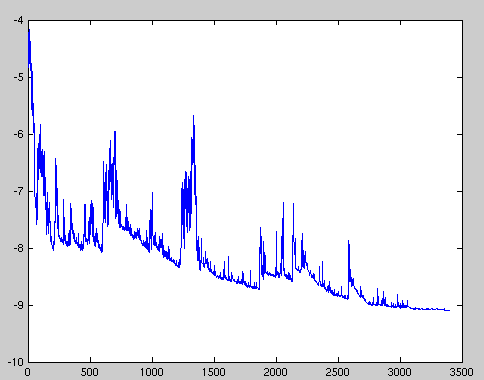}%
\caption{SGD Error}%
\label{4.SGD}%
\end{figure}

%
This technique leads to a very irregular convergence in error during training, as shown in Figure \ref{4.SGD}.%
\index{error}%
\index{SGD}%
\index{training}%
\par%

\href{https://en.wikipedia.org/wiki/Stochastic_gradient_descent}{Image from Wikipedia}%
\par%
Because the neural network is trained on a random sample of the complete training set each time, the error does not make a smooth transition downward.  However, the error usually does go down.%
\index{error}%
\index{neural network}%
\index{random}%
\index{training}%
\par%
Advantages to SGD include:%
\index{SGD}%
\par%
\begin{itemize}[noitemsep]%
\item%
Computationally efficient.  Each training step can be relatively fast, even with a huge training set.%
\index{training}%
\item%
Decreases overfitting by focusing on only a portion of the training set each step.%
\index{overfitting}%
\index{training}%
\end{itemize}

\subsection{Other Techniques}%
\label{subsec:OtherTechniques}%
One problem with simple backpropagation training algorithms is that they are susceptible to learning rate and momentum. This technique is difficult because:%
\index{algorithm}%
\index{backpropagation}%
\index{learning}%
\index{learning rate}%
\index{momentum}%
\index{Propagation Training}%
\index{training}%
\index{training algorithm}%
\par%
\begin{itemize}[noitemsep]%
\item%
Learning rate must be adjusted to a small enough level to train an accurate neural network.%
\index{learning}%
\index{learning rate}%
\index{neural network}%
\item%
Momentum must be large enough to overcome local minima yet small enough not to destabilize the training.%
\index{momentum}%
\index{training}%
\item%
A single learning rate/momentum is often not good enough for the entire training process. It is often helpful to automatically decrease the learning rate as the training progresses.%
\index{learning}%
\index{learning rate}%
\index{momentum}%
\index{ROC}%
\index{ROC}%
\index{training}%
\item%
All weights share a single learning rate/momentum.%
\index{learning}%
\index{learning rate}%
\index{momentum}%
\end{itemize}%
Other training techniques:%
\index{training}%
\par%
\begin{itemize}[noitemsep]%
\item%
\textbf{Resilient Propagation }%
{-} Use only the magnitude of the gradient and allow each neuron to learn at its rate. There is no need for learning rate/momentum; however, it only works in full batch mode.%
\index{gradient}%
\index{learning}%
\index{learning rate}%
\index{momentum}%
\index{neuron}%
\item%
\textbf{Nesterov accelerated gradient }%
{-} Helps mitigate the risk of choosing a bad mini{-}batch.%
\index{mini{-}batch}%
\item%
\textbf{Adagrad }%
{-} Allows an automatically decaying per{-}weight learning rate and momentum concept.%
\index{learning}%
\index{learning rate}%
\index{momentum}%
\item%
\textbf{Adadelta }%
{-} Extension of Adagrad that seeks to reduce its aggressive, monotonically decreasing learning rate.%
\index{learning}%
\index{learning rate}%
\item%
\textbf{Non{-}Gradient Methods }%
{-} Non{-}gradient methods can%
\index{gradient}%
\textit{ sometimes }%
be useful, though rarely outperform gradient{-}based backpropagation methods.  These include:%
\index{backpropagation}%
\index{gradient}%
\href{https://en.wikipedia.org/wiki/Simulated_annealing}{ simulated annealing}%
,%
\href{https://en.wikipedia.org/wiki/Genetic_algorithm}{ genetic algorithms}%
,%
\href{https://en.wikipedia.org/wiki/Particle_swarm_optimization}{ particle swarm optimization}%
,%
\href{https://en.wikipedia.org/wiki/Nelder%E2%80%93Mead_method}{ Nelder Mead}%
, and%
\href{https://en.wikipedia.org/wiki/Category:Optimization_algorithms_and_methods}{ many more}%
.%
\end{itemize}

\subsection{ADAM Update}%
\label{subsec:ADAMUpdate}%
ADAM is the first training algorithm you should try.  It is very effective.  Kingma and Ba (2014) introduced the Adam update rule that derives its name from the adaptive moment estimates.%
\index{ADAM}%
\index{algorithm}%
\index{training}%
\index{training algorithm}%
\cite{kingma2014adam}%
Adam estimates the first (mean) and second (variance) moments to determine the weight corrections.  Adam begins with an exponentially decaying average of past gradients (m):%
\index{ADAM}%
\index{gradient}%
\par%
\vspace{2mm}%
\begin{equation*}
 m_t = \beta_1 m_{t-1} + (1-\beta_1) g_t 
\end{equation*}
\vspace{2mm}%
\par%
This average accomplishes a similar goal as classic momentum update; however, its value is calculated automatically based on the current gradient ($g_t$).  The update rule then calculates the second moment ($v_t$):%
\index{calculated}%
\index{gradient}%
\index{momentum}%
\par%
\vspace{2mm}%
\begin{equation*}
 v_t = \beta_2 v_{t-1} + (1-\beta_2) g_t^2 
\end{equation*}
\vspace{2mm}%
\par%
The values $m_t$ and $v_t$ are estimates of the gradients' first moment (the mean) and the second moment (the uncentered variance).  However, they will be strongly biased towards zero in the initial training cycles.  The first moment's bias is corrected as follows.%
\index{bias}%
\index{gradient}%
\index{training}%
\par%
\vspace{2mm}%
\begin{equation*}
 \hat{m}_t = \frac{m_t}{1-\beta^t_1} 
\end{equation*}
\vspace{2mm}%
\par%
Similarly, the second moment is also corrected:%
\par%
\vspace{2mm}%
\begin{equation*}
 \hat{v}_t = \frac{v_t}{1-\beta_2^t} 
\end{equation*}
\vspace{2mm}%
\par%
These bias{-}corrected first and second moment estimates are applied to the ultimate Adam update rule, as follows:%
\index{ADAM}%
\index{bias}%
\par%
\vspace{2mm}%
\begin{equation*}
 \theta_t = \theta_{t-1} - \frac{\alpha \cdot \hat{m}_t}{\sqrt{\hat{v}_t}+\eta} \hat{m}_t 
\end{equation*}
\vspace{2mm}%
\par%
Adam is very tolerant to initial learning rate (\textbackslash{}alpha) and other training parameters. Kingma and Ba (2014)  propose default values of 0.9 for $\beta_1$, 0.999 for $\beta_2$, and 10{-}8 for $\eta$.%
\index{ADAM}%
\index{learning}%
\index{learning rate}%
\index{parameter}%
\index{training}%
\par

\subsection{Methods Compared}%
\label{subsec:MethodsCompared}%
The following image shows how each of these algorithms train. It is animated, so it is not displayed in the printed book, but can be accessed from here:%
\index{algorithm}%
\href{https://bit.ly/3kykkbn}{ https://bit.ly/3kykkbn}%
.%
\par%
Image credits:%
\href{https://scholar.google.com/citations?user=dOad5HoAAAAJ&amp;hl=en}{ Alec Radford}%
\par

\subsection{Specifying the Update Rule in Keras}%
\label{subsec:SpecifyingtheUpdateRuleinKeras}%
TensorFlow allows the update rule to be set to one of:%
\index{TensorFlow}%
\par%
\begin{itemize}[noitemsep]%
\item%
Adagrad%
\item%
\textbf{Adam}%
\item%
Ftrl%
\item%
Momentum%
\index{momentum}%
\item%
RMSProp%
\item%
\textbf{SGD}%
\end{itemize}%
\begin{tcolorbox}[size=title,title=Code,breakable]%
\begin{lstlisting}[language=Python, upquote=true]
%matplotlib inline

from tensorflow.keras.models import Sequential
from tensorflow.keras.layers import Dense, Activation
from tensorflow.keras.callbacks import EarlyStopping
from scipy.stats import zscore
from sklearn.model_selection import train_test_split
import pandas as pd
import matplotlib.pyplot as plt


# Regression chart.
def chart_regression(pred, y, sort=True):
    t = pd.DataFrame({'pred': pred, 'y': y.flatten()})
    if sort:
        t.sort_values(by=['y'], inplace=True)
    plt.plot(t['y'].tolist(), label='expected')
    plt.plot(t['pred'].tolist(), label='prediction')
    plt.ylabel('output')
    plt.legend()
    plt.show()

# Read the data set
df = pd.read_csv(
    "https://data.heatonresearch.com/data/t81-558/jh-simple-dataset.csv",
    na_values=['NA','?'])

# Generate dummies for job
df = pd.concat([df,pd.get_dummies(df['job'],prefix="job")],axis=1)
df.drop('job', axis=1, inplace=True)

# Generate dummies for area
df = pd.concat([df,pd.get_dummies(df['area'],prefix="area")],axis=1)
df.drop('area', axis=1, inplace=True)

# Generate dummies for product
df = pd.concat([df,pd.get_dummies(df['product'],prefix="product")],axis=1)
df.drop('product', axis=1, inplace=True)

# Missing values for income
med = df['income'].median()
df['income'] = df['income'].fillna(med)

# Standardize ranges
df['income'] = zscore(df['income'])
df['aspect'] = zscore(df['aspect'])
df['save_rate'] = zscore(df['save_rate'])
df['subscriptions'] = zscore(df['subscriptions'])

# Convert to numpy - Classification
x_columns = df.columns.drop('age').drop('id')
x = df[x_columns].values
y = df['age'].values

# Create train/test
x_train, x_test, y_train, y_test = train_test_split(    
    x, y, test_size=0.25, random_state=42)

# Build the neural network
model = Sequential()
model.add(Dense(25, input_dim=x.shape[1], activation='relu')) # Hidden 1
model.add(Dense(10, activation='relu')) # Hidden 2
model.add(Dense(1)) # Output
model.compile(loss='mean_squared_error', optimizer='adam') # Modify here
monitor = EarlyStopping(monitor='val_loss', min_delta=1e-3, patience=5, 
                        verbose=1, mode='auto', restore_best_weights=True)
model.fit(x_train,y_train,validation_data=(x_test,y_test),
          callbacks=[monitor],verbose=0,epochs=1000)

# Plot the chart
pred = model.predict(x_test)
chart_regression(pred.flatten(),y_test)\end{lstlisting}
\tcbsubtitle[before skip=\baselineskip]{Output}%
\includegraphics[width=3in]{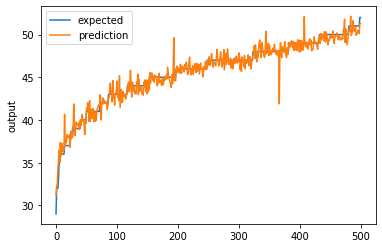}%
\begin{lstlisting}[upquote=true]
Restoring model weights from the end of the best epoch.
Epoch 00105: early stopping
\end{lstlisting}
\end{tcolorbox}

\section{Part 4.5: Error Calculation from Scratch}%
\label{sec:Part4.5ErrorCalculationfromScratch}%
We will now look at how to calculate RMSE and logloss by hand.  RMSE is typically used for regression. We begin by calculating RMSE with libraries.%
\index{MSE}%
\index{regression}%
\index{RMSE}%
\index{RMSE}%
\par%
\begin{tcolorbox}[size=title,title=Code,breakable]%
\begin{lstlisting}[language=Python, upquote=true]
from sklearn import metrics
import numpy as np

predicted = [1.1,1.9,3.4,4.2,4.3]
expected = [1,2,3,4,5]

score_mse = metrics.mean_squared_error(predicted,expected)
score_rmse = np.sqrt(score_mse)
print("Score (MSE): {}".format(score_mse))
print("Score (RMSE): {}".format(score_rmse))\end{lstlisting}
\tcbsubtitle[before skip=\baselineskip]{Output}%
\begin{lstlisting}[upquote=true]
Score (MSE): 0.14200000000000007
Score (RMSE): 0.37682887362833556
\end{lstlisting}
\end{tcolorbox}%
We can also calculate without libraries.%
\par%
\begin{tcolorbox}[size=title,title=Code,breakable]%
\begin{lstlisting}[language=Python, upquote=true]
score_mse = ((predicted[0]-expected[0])**2 + (predicted[1]-expected[1])**2 
+ (predicted[2]-expected[2])**2 + (predicted[3]-expected[3])**2
+ (predicted[4]-expected[4])**2)/len(predicted)
score_rmse = np.sqrt(score_mse)
    
print("Score (MSE): {}".format(score_mse))
print("Score (RMSE): {}".format(score_rmse))\end{lstlisting}
\tcbsubtitle[before skip=\baselineskip]{Output}%
\begin{lstlisting}[upquote=true]
Score (MSE): 0.14200000000000007
Score (RMSE): 0.37682887362833556
\end{lstlisting}
\end{tcolorbox}%
\subsection{Classification}%
\label{subsec:Classification}%
We will now look at how to calculate a logloss by hand. For this, we look at a binary prediction. The predicted is some number between 0{-}1 that indicates the probability true (1). The expected is always 0 or 1. Therefore, a prediction of 1.0 is completely correct if the expected is 1 and completely wrong if the expected is 0.%
\index{predict}%
\index{probability}%
\index{SOM}%
\par%
\begin{tcolorbox}[size=title,title=Code,breakable]%
\begin{lstlisting}[language=Python, upquote=true]
from sklearn import metrics

expected = [1,1,0,0,0]
predicted = [0.9,0.99,0.1,0.05,0.06]

print(metrics.log_loss(expected,predicted))\end{lstlisting}
\tcbsubtitle[before skip=\baselineskip]{Output}%
\begin{lstlisting}[upquote=true]
0.06678801305495843
\end{lstlisting}
\end{tcolorbox}%
Now we attempt to calculate the same logloss manually.%
\par%
\begin{tcolorbox}[size=title,title=Code,breakable]%
\begin{lstlisting}[language=Python, upquote=true]
import numpy as np

score_logloss = (np.log(1.0-np.abs(expected[0]-predicted[0]))+\
np.log(1.0-np.abs(expected[1]-predicted[1]))+\
np.log(1.0-np.abs(expected[2]-predicted[2]))+\
np.log(1.0-np.abs(expected[3]-predicted[3]))+\
np.log(1.0-np.abs(expected[4]-predicted[4])))\
*(-1/len(predicted))

print(f'Score Logloss {score_logloss}')\end{lstlisting}
\tcbsubtitle[before skip=\baselineskip]{Output}%
\begin{lstlisting}[upquote=true]
Score Logloss 0.06678801305495843
\end{lstlisting}
\end{tcolorbox}

\chapter{Regularization and Dropout}%
\label{chap:RegularizationandDropout}%
\section{Part 5.1: Introduction to Regularization: Ridge and Lasso}%
\label{sec:Part5.1IntroductiontoRegularizationRidgeandLasso}%
Regularization is a technique that reduces overfitting, which occurs when neural networks attempt to memorize training data rather than learn from it. Humans are capable of overfitting as well. Before examining how a machine accidentally overfits, we will first explore how humans can suffer from it.%
\index{neural network}%
\index{overfitting}%
\index{regularization}%
\index{training}%
\par%
Human programmers often take certification exams to show their competence in a given programming language. To help prepare for these exams, the test makers often make practice exams available. Consider a programmer who enters a loop of taking the practice exam, studying more, and then retaking the practice exam. The programmer has memorized much of the practice exam at some point rather than learning the techniques necessary to figure out the individual questions. The programmer has now overfitted for the practice exam. When this programmer takes the real exam, his actual score will likely be lower than what he earned on the practice exam.%
\index{learning}%
\index{SOM}%
\par%
Although a neural network received a high score on its training data, this result does not mean that the same neural network will score high on data that was not inside the training set. A computer can overfit as well. Regularization is one of the techniques that can prevent overfitting. Several different regularization techniques exist. Most work by analyzing and potentially modifying the weights of a neural network as it trains.%
\index{neural network}%
\index{overfitting}%
\index{regularization}%
\index{training}%
\par%
\subsection{L1 and L2 Regularization}%
\label{subsec:L1andL2Regularization}%
L1 and L2 regularization are two standard regularization techniques that can reduce the effects of overfitting. These algorithms can either work with an objective function or as part of the backpropagation algorithm. The regularization algorithm is attached to the training algorithm by adding an objective in both cases.%
\index{algorithm}%
\index{backpropagation}%
\index{L1}%
\index{L2}%
\index{overfitting}%
\index{regularization}%
\index{training}%
\index{training algorithm}%
\par%
These algorithms work by adding a weight penalty to the neural network training. This penalty encourages the neural network to keep the weights to small values. Both L1 and L2 calculate this penalty differently. You can add this penalty calculation to the calculated gradients for gradient{-}descent{-}based algorithms, such as backpropagation. The penalty is negatively combined with the objective score for objective{-}function{-}based training, such as simulated annealing.%
\index{algorithm}%
\index{annealing}%
\index{backpropagation}%
\index{calculated}%
\index{gradient}%
\index{L1}%
\index{L2}%
\index{neural network}%
\index{simulated annealing}%
\index{training}%
\par%
We will look at linear regression to see how L1 and L2 regularization work. The following code sets up the auto{-}mpg data for this purpose.%
\index{L1}%
\index{L2}%
\index{linear}%
\index{regression}%
\index{regularization}%
\par%
\begin{tcolorbox}[size=title,title=Code,breakable]%
\begin{lstlisting}[language=Python, upquote=true]
from sklearn.linear_model import LassoCV
import pandas as pd
import os
import numpy as np
from sklearn import metrics
from scipy.stats import zscore
from sklearn.model_selection import train_test_split 

df = pd.read_csv(
    "https://data.heatonresearch.com/data/t81-558/auto-mpg.csv", 
    na_values=['NA', '?'])

# Handle missing value
df['horsepower'] = df['horsepower'].fillna(df['horsepower'].median())

# Pandas to Numpy
names = ['cylinders', 'displacement', 'horsepower', 'weight',
       'acceleration', 'year', 'origin']
x = df[names].values
y = df['mpg'].values # regression

# Split into train/test
x_train, x_test, y_train, y_test = train_test_split(    
    x, y, test_size=0.25, random_state=45)\end{lstlisting}
\end{tcolorbox}%
We will use the data just loaded for several examples. The first examples in this part use several forms of linear regression. For linear regression, it is helpful to examine the model's coefficients. The following function is utilized to display these coefficients.%
\index{linear}%
\index{model}%
\index{regression}%
\par%
\begin{tcolorbox}[size=title,title=Code,breakable]%
\begin{lstlisting}[language=Python, upquote=true]
# Simple function to evaluate the coefficients of a regression
%matplotlib inline    
from IPython.display import display, HTML    

def report_coef(names,coef,intercept):
    r = pd.DataFrame( { 'coef': coef, 'positive': coef>=0  }, index = names )
    r = r.sort_values(by=['coef'])
    display(r)
    print(f"Intercept: {intercept}")
    r['coef'].plot(kind='barh', color=r['positive'].map(
        {True: 'b', False: 'r'}))\end{lstlisting}
\end{tcolorbox}

\subsection{Linear Regression}%
\label{subsec:LinearRegression}%
Before jumping into L1/L2 regularization, we begin with linear regression.  Researchers first introduced the L1/L2 form of regularization for%
\index{L1}%
\index{L2}%
\index{linear}%
\index{regression}%
\index{regularization}%
\href{https://en.wikipedia.org/wiki/Linear_regression}{ linear regression}%
.  We can also make use of L1/L2 for neural networks.  To fully understand L1/L2 we will begin with how we can use them with linear regression.%
\index{L1}%
\index{L2}%
\index{linear}%
\index{neural network}%
\index{regression}%
\par%
The following code uses linear regression to fit the auto{-}mpg data set.  The RMSE reported will not be as good as a neural network.%
\index{linear}%
\index{MSE}%
\index{neural network}%
\index{regression}%
\index{RMSE}%
\index{RMSE}%
\par%
\begin{tcolorbox}[size=title,title=Code,breakable]%
\begin{lstlisting}[language=Python, upquote=true]
import sklearn

# Create linear regression
regressor = sklearn.linear_model.LinearRegression()

# Fit/train linear regression
regressor.fit(x_train,y_train)
# Predict
pred = regressor.predict(x_test)

# Measure RMSE error.  RMSE is common for regression.
score = np.sqrt(metrics.mean_squared_error(pred,y_test))
print(f"Final score (RMSE): {score}")

report_coef(
  names,
  regressor.coef_,
  regressor.intercept_)\end{lstlisting}
\tcbsubtitle[before skip=\baselineskip]{Output}%
\begin{tabular}[hbt!]{l|l|l}%
\hline%
&coef&positive\\%
\hline%
cylinders&{-}0.427721&False\\%
weight&{-}0.007255&False\\%
horsepower&{-}0.005491&False\\%
displacement&0.020166&True\\%
acceleration&0.138575&True\\%
year&0.783047&True\\%
origin&1.003762&True\\%
\hline%
\end{tabular}%
\vspace{2mm}%
\includegraphics[width=4in]{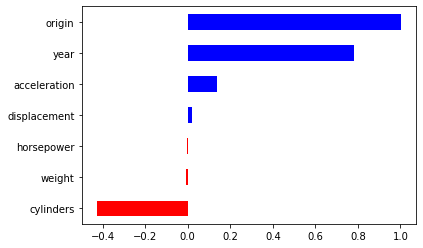}%
\begin{lstlisting}[upquote=true]
Final score (RMSE): 3.0019345985860784
Intercept: -19.101231042200112
\end{lstlisting}
\end{tcolorbox}

\subsection{L1 (Lasso) Regularization}%
\label{subsec:L1(Lasso)Regularization}%
L1 regularization, also called LASSO (Least Absolute Shrinkage and Selection Operator) should be used to create sparsity in the neural network. In other words, the L1 algorithm will push many weight connections to near 0. When the weight is near 0, the program drops it from the network. Dropping weighted connections will create a sparse neural network.%
\index{algorithm}%
\index{connection}%
\index{L1}%
\index{neural network}%
\index{regularization}%
\index{sparse}%
\par%
Feature selection is a useful byproduct of sparse neural networks. Features are the values that the training set provides to the input neurons. Once all the weights of an input neuron reach 0, the neural network training determines that the feature is unnecessary. If your data set has many unnecessary input features, L1 regularization can help the neural network detect and ignore unnecessary features.%
\index{feature}%
\index{input}%
\index{input neuron}%
\index{L1}%
\index{neural network}%
\index{neuron}%
\index{regularization}%
\index{sparse}%
\index{training}%
\par%
L1 is implemented by adding the following error to the objective to minimize:%
\index{error}%
\index{L1}%
\par%
\vspace{2mm}%
\begin{equation*}
 E_1 = \alpha \sum_w{ |w| } 
\end{equation*}
\vspace{2mm}%
\par%
You should use L1 regularization to create sparsity in the neural network. In other words, the L1 algorithm will push many weight connections to near 0. When the weight is near 0, the program drops it from the network. Dropping weighted connections will create a sparse neural network.%
\index{algorithm}%
\index{connection}%
\index{L1}%
\index{neural network}%
\index{regularization}%
\index{sparse}%
\par%
The following code demonstrates lasso regression. Notice the effect of the coefficients compared to the previous section that used linear regression.%
\index{linear}%
\index{regression}%
\par%
\begin{tcolorbox}[size=title,title=Code,breakable]%
\begin{lstlisting}[language=Python, upquote=true]
import sklearn
from sklearn.linear_model import Lasso

# Create linear regression
regressor = Lasso(random_state=0,alpha=0.1)

# Fit/train LASSO
regressor.fit(x_train,y_train)
# Predict
pred = regressor.predict(x_test)

# Measure RMSE error.  RMSE is common for regression.
score = np.sqrt(metrics.mean_squared_error(pred,y_test))
print(f"Final score (RMSE): {score}")

report_coef(
  names,
  regressor.coef_,
  regressor.intercept_)\end{lstlisting}
\tcbsubtitle[before skip=\baselineskip]{Output}%
\begin{tabular}[hbt!]{l|l|l}%
\hline%
&coef&positive\\%
\hline%
cylinders&{-}0.012995&False\\%
weight&{-}0.007328&False\\%
horsepower&{-}0.002715&False\\%
displacement&0.011601&True\\%
acceleration&0.114391&True\\%
origin&0.708222&True\\%
year&0.777480&True\\%
\hline%
\end{tabular}%
\vspace{2mm}%
\includegraphics[width=4in]{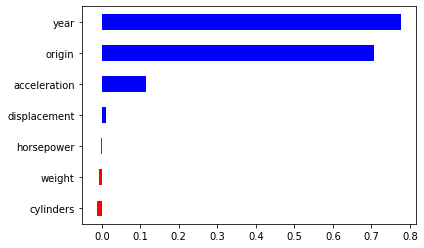}%
\begin{lstlisting}[upquote=true]
Final score (RMSE): 3.0604021904033303
Intercept: -18.506677982383252
\end{lstlisting}
\end{tcolorbox}

\subsection{L2 (Ridge) Regularization}%
\label{subsec:L2(Ridge)Regularization}%
You should use Tikhonov/Ridge/L2 regularization when you are less concerned about creating a space network and are more concerned about low weight values.  The lower weight values will typically lead to less overfitting.%
\index{L2}%
\index{overfitting}%
\index{regularization}%
\par%
\vspace{2mm}%
\begin{equation*}
 E_2 = \alpha \sum_w{ w^2 } 
\end{equation*}
\vspace{2mm}%
\par%
Like the L1 algorithm, the $\alpha$ value determines how important the L2 objective is compared to the neural network's error.  Typical L2 values are below 0.1 (10\%).  The main calculation performed by L2 is the summing of the squares of all of the weights.  The algorithm will not sum bias values.%
\index{algorithm}%
\index{bias}%
\index{error}%
\index{L1}%
\index{L2}%
\index{neural network}%
\par%
You should use L2 regularization when you are less concerned about creating a space network and are more concerned about low weight values.  The lower weight values will typically lead to less overfitting.  Generally, L2 regularization will produce better overall performance than L1.  However, L1 might be useful in situations with many inputs, and you can prune some of the weaker inputs.%
\index{input}%
\index{L1}%
\index{L2}%
\index{overfitting}%
\index{regularization}%
\index{SOM}%
\par%
The following code uses L2 with linear regression (Ridge regression):%
\index{L2}%
\index{linear}%
\index{regression}%
\par%
\begin{tcolorbox}[size=title,title=Code,breakable]%
\begin{lstlisting}[language=Python, upquote=true]
import sklearn
from sklearn.linear_model import Ridge

# Create linear regression
regressor = Ridge(alpha=1)

# Fit/train Ridge
regressor.fit(x_train,y_train)
# Predict
pred = regressor.predict(x_test)

# Measure RMSE error.  RMSE is common for regression.
score = np.sqrt(metrics.mean_squared_error(pred,y_test))
print("Final score (RMSE): {score}")

report_coef(
  names,
  regressor.coef_,
  regressor.intercept_)\end{lstlisting}
\tcbsubtitle[before skip=\baselineskip]{Output}%
\begin{tabular}[hbt!]{l|l|l}%
\hline%
&coef&positive\\%
\hline%
cylinders&{-}0.421393&False\\%
weight&{-}0.007257&False\\%
horsepower&{-}0.005385&False\\%
displacement&0.020006&True\\%
acceleration&0.138470&True\\%
year&0.782889&True\\%
origin&0.994621&True\\%
\hline%
\end{tabular}%
\vspace{2mm}%
\includegraphics[width=4in]{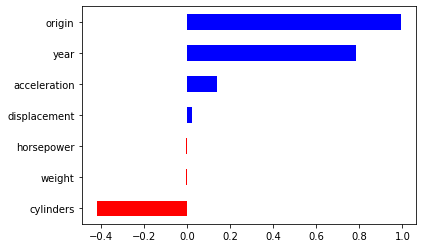}%
\begin{lstlisting}[upquote=true]
Final score (RMSE): {score}
Intercept: -19.07980074425469
\end{lstlisting}
\end{tcolorbox}

\subsection{ElasticNet Regularization}%
\label{subsec:ElasticNetRegularization}%
The ElasticNet regression combines both L1 and L2.  Both penalties are applied.  The amount of L1 and L2 are governed by the parameters alpha and beta.%
\index{L1}%
\index{L2}%
\index{parameter}%
\index{regression}%
\par%
\vspace{2mm}%
\begin{equation*}
 a * {\rm L}1 + b * {\rm L}2 
\end{equation*}
\vspace{2mm}%
\par%
\begin{tcolorbox}[size=title,title=Code,breakable]%
\begin{lstlisting}[language=Python, upquote=true]
import sklearn
from sklearn.linear_model import ElasticNet

# Create linear regression
regressor = ElasticNet(alpha=0.1, l1_ratio=0.1)

# Fit/train LASSO
regressor.fit(x_train,y_train)
# Predict
pred = regressor.predict(x_test)

# Measure RMSE error.  RMSE is common for regression.
score = np.sqrt(metrics.mean_squared_error(pred,y_test))
print(f"Final score (RMSE): {score}")

report_coef(
  names,
  regressor.coef_,
  regressor.intercept_)\end{lstlisting}
\tcbsubtitle[before skip=\baselineskip]{Output}%
\begin{tabular}[hbt!]{l|l|l}%
\hline%
&coef&positive\\%
\hline%
cylinders&{-}0.274010&False\\%
weight&{-}0.007303&False\\%
horsepower&{-}0.003231&False\\%
displacement&0.016194&True\\%
acceleration&0.132348&True\\%
year&0.777482&True\\%
origin&0.782781&True\\%
\hline%
\end{tabular}%
\vspace{2mm}%
\includegraphics[width=4in]{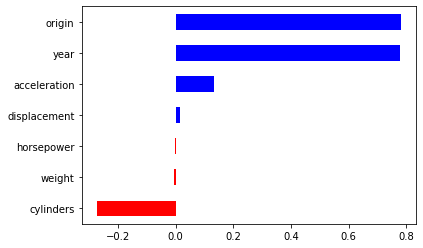}%
\begin{lstlisting}[upquote=true]
Final score (RMSE): 3.0450899960775013
Intercept: -18.389355690429767
\end{lstlisting}
\end{tcolorbox}

\section{Part 5.2: Using K{-}Fold Cross{-}validation with Keras}%
\label{sec:Part5.2UsingK{-}FoldCross{-}validationwithKeras}%
You can use cross{-}validation for a variety of purposes in predictive modeling:%
\index{model}%
\index{predict}%
\index{validation}%
\par
\begin{figure}[h]%
\centering%
\includegraphics[width=4in]{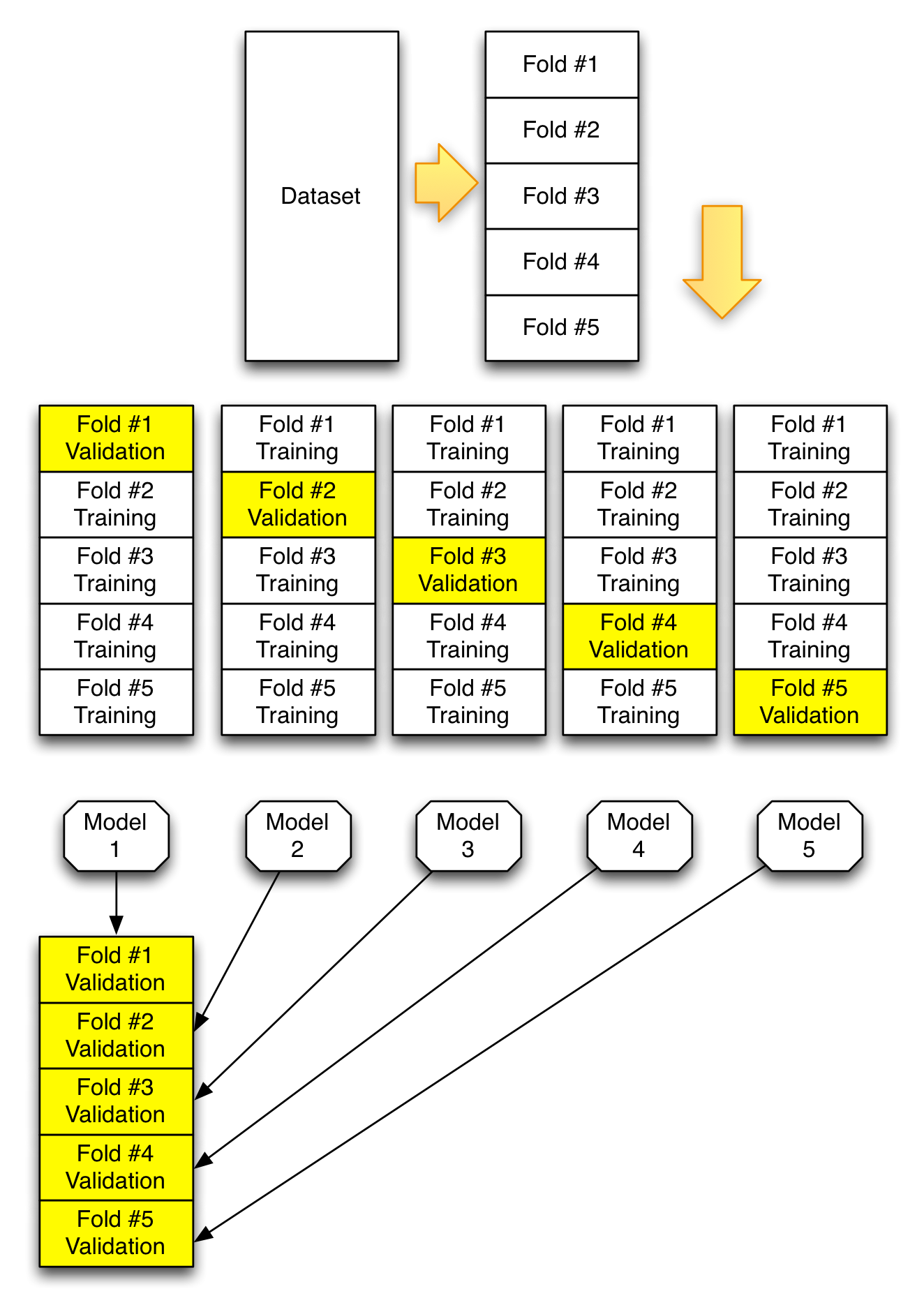}%
\caption{K{-}Fold Crossvalidation}%
\label{5.CROSS}%
\end{figure}

%
\begin{itemize}[noitemsep]%
\item%
Generating out{-}of{-}sample predictions from a neural network%
\index{neural network}%
\index{predict}%
\item%
Estimate a good number of epochs to train a neural network for (early stopping)%
\index{early stopping}%
\index{neural network}%
\item%
Evaluate the effectiveness of certain hyperparameters, such as activation functions, neuron counts, and layer counts%
\index{activation function}%
\index{hyperparameter}%
\index{layer}%
\index{neuron}%
\index{parameter}%
\end{itemize}%
Cross{-}validation uses several folds and multiple models to provide each data segment a chance to serve as both the validation and training set. Figure \ref{5.CROSS} shows cross{-}validation.%
\index{model}%
\index{training}%
\index{validation}%
\par%

\par%
It is important to note that each fold will have one model (neural network). To generate predictions for new data (not present in the training set), predictions from the fold models can be handled in several ways:%
\index{model}%
\index{neural network}%
\index{predict}%
\index{training}%
\par%
\begin{itemize}[noitemsep]%
\item%
Choose the model with the highest validation score as the final model.%
\index{model}%
\index{validation}%
\item%
Preset new data to the five models (one for each fold) and average the result (this is an%
\index{model}%
\href{https://en.wikipedia.org/wiki/Ensemble_learning}{ ensemble}%
).%
\item%
Retrain a new model (using the same settings as the cross{-}validation) on the entire dataset. Train for as many epochs and with the same hidden layer structure.%
\index{dataset}%
\index{hidden layer}%
\index{layer}%
\index{model}%
\index{validation}%
\end{itemize}%
Generally, I prefer the last approach and will retrain a model on the entire data set once I have selected hyper{-}parameters. Of course, I will always set aside a final holdout set for model validation that I do not use in any aspect of the training process.%
\index{model}%
\index{parameter}%
\index{ROC}%
\index{ROC}%
\index{training}%
\index{validation}%
\par%
\subsection{Regression vs Classification K{-}Fold Cross{-}Validation}%
\label{subsec:RegressionvsClassificationK{-}FoldCross{-}Validation}%
Regression and classification are handled somewhat differently concerning cross{-}validation. Regression is the simpler case where you can break up the data set into K folds with little regard for where each item lands. For regression, the data items should fall into the folds as randomly as possible. It is also important to remember that not every fold will necessarily have the same number of data items. It is not always possible for the data set to be evenly divided into K folds. For regression cross{-}validation, we will use the Scikit{-}Learn class%
\index{classification}%
\index{random}%
\index{regression}%
\index{SOM}%
\index{validation}%
\textbf{ KFold}%
.%
\par%
Cross{-}validation for classification could also use the%
\index{classification}%
\index{validation}%
\textbf{ KFold }%
object; however, this technique would not ensure that the class balance remains the same in each fold as in the original. The balance of classes that a model was trained on must remain the same (or similar) to the training set. Drift in this distribution is one of the most important things to monitor after a trained model has been placed into actual use. Because of this, we want to make sure that the cross{-}validation itself does not introduce an unintended shift. This technique is called stratified sampling and is accomplished by using the Scikit{-}Learn object%
\index{model}%
\index{training}%
\index{validation}%
\textbf{ StratifiedKFold }%
in place of%
\textbf{ KFold }%
whenever you use classification. In summary, you should use the following two objects in Scikit{-}Learn:%
\index{classification}%
\par%
\begin{itemize}[noitemsep]%
\item%
\textbf{KFold }%
When dealing with a regression problem.%
\index{regression}%
\item%
\textbf{StratifiedKFold }%
When dealing with a classification problem.%
\index{classification}%
\end{itemize}%
The following two sections demonstrate cross{-}validation with classification and regression.%
\index{classification}%
\index{regression}%
\index{validation}%
\par

\subsection{Out{-}of{-}Sample Regression Predictions with K{-}Fold Cross{-}Validation}%
\label{subsec:Out{-}of{-}SampleRegressionPredictionswithK{-}FoldCross{-}Validation}%
The following code trains the simple dataset using a 5{-}fold cross{-}validation. The expected performance of a neural network of the type trained here would be the score for the generated out{-}of{-}sample predictions. We begin by preparing a feature vector using the%
\index{dataset}%
\index{feature}%
\index{neural network}%
\index{predict}%
\index{validation}%
\index{vector}%
\textbf{ jh{-}simple{-}dataset }%
to predict age. This model is set up as a regression problem.%
\index{model}%
\index{predict}%
\index{regression}%
\par%
\begin{tcolorbox}[size=title,title=Code,breakable]%
\begin{lstlisting}[language=Python, upquote=true]
import pandas as pd
from scipy.stats import zscore
from sklearn.model_selection import train_test_split

# Read the data set
df = pd.read_csv(
    "https://data.heatonresearch.com/data/t81-558/jh-simple-dataset.csv",
    na_values=['NA','?'])

# Generate dummies for job
df = pd.concat([df,pd.get_dummies(df['job'],prefix="job")],axis=1)
df.drop('job', axis=1, inplace=True)

# Generate dummies for area
df = pd.concat([df,pd.get_dummies(df['area'],prefix="area")],axis=1)
df.drop('area', axis=1, inplace=True)

# Generate dummies for product
df = pd.concat([df,pd.get_dummies(df['product'],prefix="product")],axis=1)
df.drop('product', axis=1, inplace=True)

# Missing values for income
med = df['income'].median()
df['income'] = df['income'].fillna(med)

# Standardize ranges
df['income'] = zscore(df['income'])
df['aspect'] = zscore(df['aspect'])
df['save_rate'] = zscore(df['save_rate'])
df['subscriptions'] = zscore(df['subscriptions'])

# Convert to numpy - Classification
x_columns = df.columns.drop('age').drop('id')
x = df[x_columns].values
y = df['age'].values\end{lstlisting}
\end{tcolorbox}%
Now that the feature vector is created a 5{-}fold cross{-}validation can be performed to generate out{-}of{-}sample predictions.  We will assume 500 epochs and not use early stopping.  Later we will see how we can estimate a more optimal epoch count.%
\index{early stopping}%
\index{feature}%
\index{predict}%
\index{validation}%
\index{vector}%
\par%
\begin{tcolorbox}[size=title,title=Code,breakable]%
\begin{lstlisting}[language=Python, upquote=true]
EPOCHS=500

import pandas as pd
import os
import numpy as np
from sklearn import metrics
from scipy.stats import zscore
from sklearn.model_selection import KFold
from tensorflow.keras.models import Sequential
from tensorflow.keras.layers import Dense, Activation

# Cross-Validate
kf = KFold(5, shuffle=True, random_state=42) # Use for KFold classification
oos_y = []
oos_pred = []

fold = 0
for train, test in kf.split(x):
    fold+=1
    print(f"Fold #{fold}")
        
    x_train = x[train]
    y_train = y[train]
    x_test = x[test]
    y_test = y[test]
    
    model = Sequential()
    model.add(Dense(20, input_dim=x.shape[1], activation='relu'))
    model.add(Dense(10, activation='relu'))
    model.add(Dense(1))
    model.compile(loss='mean_squared_error', optimizer='adam')
    
    model.fit(x_train,y_train,validation_data=(x_test,y_test),verbose=0,
              epochs=EPOCHS)
    
    pred = model.predict(x_test)
    
    oos_y.append(y_test)
    oos_pred.append(pred)    

    # Measure this fold's RMSE
    score = np.sqrt(metrics.mean_squared_error(pred,y_test))
    print(f"Fold score (RMSE): {score}")

# Build the oos prediction list and calculate the error.
oos_y = np.concatenate(oos_y)
oos_pred = np.concatenate(oos_pred)
score = np.sqrt(metrics.mean_squared_error(oos_pred,oos_y))
print(f"Final, out of sample score (RMSE): {score}")    
    
# Write the cross-validated prediction
oos_y = pd.DataFrame(oos_y)
oos_pred = pd.DataFrame(oos_pred)
oosDF = pd.concat( [df, oos_y, oos_pred],axis=1 )
#oosDF.to_csv(filename_write,index=False)\end{lstlisting}
\tcbsubtitle[before skip=\baselineskip]{Output}%
\begin{lstlisting}[upquote=true]
Fold #1
Fold score (RMSE): 0.6814299426511208
Fold #2
Fold score (RMSE): 0.45486513719487165
Fold #3
Fold score (RMSE): 0.571615041876392
Fold #4
Fold score (RMSE): 0.46416356081116916
Fold #5
Fold score (RMSE): 1.0426518491685475
Final, out of sample score (RMSE): 0.678316077597408
\end{lstlisting}
\end{tcolorbox}%
As you can see, the above code also reports the average number of epochs needed.  A common technique is to then train on the entire dataset for the average number of epochs required.%
\index{dataset}%
\par

\subsection{Classification with Stratified K{-}Fold Cross{-}Validation}%
\label{subsec:ClassificationwithStratifiedK{-}FoldCross{-}Validation}%
The following code trains and fits the%
\textbf{ jh}%
{-}simple{-}dataset dataset with cross{-}validation to generate out{-}of{-}sample.  It also writes the out{-}of{-}sample (predictions on the test set) results.%
\index{dataset}%
\index{predict}%
\index{validation}%
\par%
It is good to perform stratified k{-}fold cross{-}validation with classification data.  This technique ensures that the percentages of each class remain the same across all folds.  Use the%
\index{classification}%
\index{k{-}fold}%
\index{validation}%
\textbf{ StratifiedKFold }%
object instead of the%
\textbf{ KFold }%
object used in the regression.%
\index{regression}%
\par%
\begin{tcolorbox}[size=title,title=Code,breakable]%
\begin{lstlisting}[language=Python, upquote=true]
import pandas as pd
from scipy.stats import zscore

# Read the data set
df = pd.read_csv(
    "https://data.heatonresearch.com/data/t81-558/jh-simple-dataset.csv",
    na_values=['NA','?'])

# Generate dummies for job
df = pd.concat([df,pd.get_dummies(df['job'],prefix="job")],axis=1)
df.drop('job', axis=1, inplace=True)

# Generate dummies for area
df = pd.concat([df,pd.get_dummies(df['area'],prefix="area")],axis=1)
df.drop('area', axis=1, inplace=True)

# Missing values for income
med = df['income'].median()
df['income'] = df['income'].fillna(med)

# Standardize ranges
df['income'] = zscore(df['income'])
df['aspect'] = zscore(df['aspect'])
df['save_rate'] = zscore(df['save_rate'])
df['age'] = zscore(df['age'])
df['subscriptions'] = zscore(df['subscriptions'])

# Convert to numpy - Classification
x_columns = df.columns.drop('product').drop('id')
x = df[x_columns].values
dummies = pd.get_dummies(df['product']) # Classification
products = dummies.columns
y = dummies.values\end{lstlisting}
\end{tcolorbox}%
We will assume 500 epochs and not use early stopping.  Later we will see how we can estimate a more optimal epoch count.%
\index{early stopping}%
\par%
\begin{tcolorbox}[size=title,title=Code,breakable]%
\begin{lstlisting}[language=Python, upquote=true]
import pandas as pd
import os
import numpy as np
from sklearn import metrics
from sklearn.model_selection import StratifiedKFold
from tensorflow.keras.models import Sequential
from tensorflow.keras.layers import Dense, Activation

# np.argmax(pred,axis=1)
# Cross-validate
# Use for StratifiedKFold classification
kf = StratifiedKFold(5, shuffle=True, random_state=42) 
    
oos_y = []
oos_pred = []
fold = 0

# Must specify y StratifiedKFold for
for train, test in kf.split(x,df['product']):  
    fold+=1
    print(f"Fold #{fold}")
        
    x_train = x[train]
    y_train = y[train]
    x_test = x[test]
    y_test = y[test]
    
    model = Sequential()
    # Hidden 1
    model.add(Dense(50, input_dim=x.shape[1], activation='relu')) 
    model.add(Dense(25, activation='relu')) # Hidden 2
    model.add(Dense(y.shape[1],activation='softmax')) # Output
    model.compile(loss='categorical_crossentropy', optimizer='adam')

    model.fit(x_train,y_train,validation_data=(x_test,y_test),
              verbose=0, epochs=EPOCHS)
    
    pred = model.predict(x_test)
    
    oos_y.append(y_test)
    # raw probabilities to chosen class (highest probability)
    pred = np.argmax(pred,axis=1) 
    oos_pred.append(pred)  

    # Measure this fold's accuracy
    y_compare = np.argmax(y_test,axis=1) # For accuracy calculation
    score = metrics.accuracy_score(y_compare, pred)
    print(f"Fold score (accuracy): {score}")

# Build the oos prediction list and calculate the error.
oos_y = np.concatenate(oos_y)
oos_pred = np.concatenate(oos_pred)
oos_y_compare = np.argmax(oos_y,axis=1) # For accuracy calculation

score = metrics.accuracy_score(oos_y_compare, oos_pred)
print(f"Final score (accuracy): {score}")    
    
# Write the cross-validated prediction
oos_y = pd.DataFrame(oos_y)
oos_pred = pd.DataFrame(oos_pred)
oosDF = pd.concat( [df, oos_y, oos_pred],axis=1 )
#oosDF.to_csv(filename_write,index=False)\end{lstlisting}
\tcbsubtitle[before skip=\baselineskip]{Output}%
\begin{lstlisting}[upquote=true]
Fold #1
Fold score (accuracy): 0.6325
Fold #2
Fold score (accuracy): 0.6725
Fold #3
Fold score (accuracy): 0.6975
Fold #4
Fold score (accuracy): 0.6575
Fold #5
Fold score (accuracy): 0.675
Final score (accuracy): 0.667
\end{lstlisting}
\end{tcolorbox}

\subsection{Training with both a Cross{-}Validation and a Holdout Set}%
\label{subsec:TrainingwithbothaCross{-}ValidationandaHoldoutSet}%
If you have a considerable amount of data, it is always valuable to set aside a holdout set before you cross{-}validate. This holdout set will be the final evaluation before using your model for its real{-}world use. Figure \ref{5.} HOLDOUT shows this division.%
\index{model}%
\par%

\begin{figure}[h]%
\centering%
\includegraphics[width=4in]{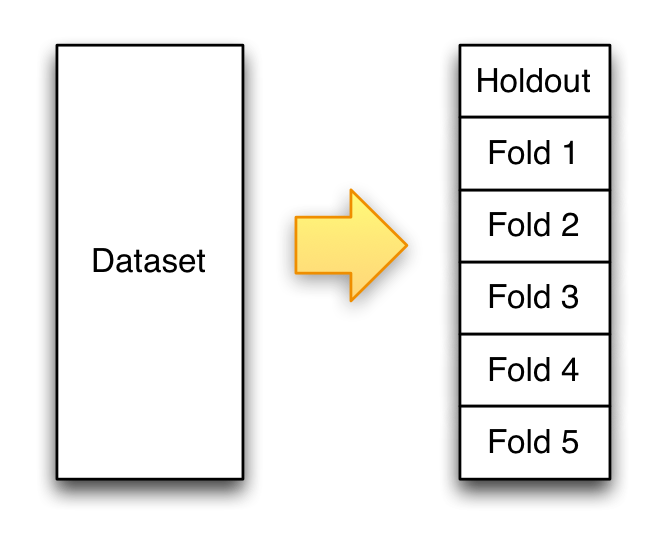}%
\caption{Cross{-}Validation and a Holdout Set}%
\label{5. HOLDOUT}%
\end{figure}

\par%
The following program uses a holdout set and then still cross{-}validates.%
\par%
\begin{tcolorbox}[size=title,title=Code,breakable]%
\begin{lstlisting}[language=Python, upquote=true]
import pandas as pd
from scipy.stats import zscore
from sklearn.model_selection import train_test_split

# Read the data set
df = pd.read_csv(
    "https://data.heatonresearch.com/data/t81-558/jh-simple-dataset.csv",
    na_values=['NA','?'])

# Generate dummies for job
df = pd.concat([df,pd.get_dummies(df['job'],prefix="job")],axis=1)
df.drop('job', axis=1, inplace=True)

# Generate dummies for area
df = pd.concat([df,pd.get_dummies(df['area'],prefix="area")],axis=1)
df.drop('area', axis=1, inplace=True)

# Generate dummies for product
df = pd.concat([df,pd.get_dummies(df['product'],prefix="product")],axis=1)
df.drop('product', axis=1, inplace=True)

# Missing values for income
med = df['income'].median()
df['income'] = df['income'].fillna(med)

# Standardize ranges
df['income'] = zscore(df['income'])
df['aspect'] = zscore(df['aspect'])
df['save_rate'] = zscore(df['save_rate'])
df['subscriptions'] = zscore(df['subscriptions'])

# Convert to numpy - Classification
x_columns = df.columns.drop('age').drop('id')
x = df[x_columns].values
y = df['age'].values\end{lstlisting}
\end{tcolorbox}%
Now that the data has been preprocessed, we are ready to build the neural network.%
\index{neural network}%
\index{ROC}%
\index{ROC}%
\par%
\begin{tcolorbox}[size=title,title=Code,breakable]%
\begin{lstlisting}[language=Python, upquote=true]
from sklearn.model_selection import train_test_split
import pandas as pd
import os
import numpy as np
from sklearn import metrics
from scipy.stats import zscore
from sklearn.model_selection import KFold

# Keep a 10% holdout
x_main, x_holdout, y_main, y_holdout = train_test_split(    
    x, y, test_size=0.10) 


# Cross-validate
kf = KFold(5)
    
oos_y = []
oos_pred = []
fold = 0
for train, test in kf.split(x_main):        
    fold+=1
    print(f"Fold #{fold}")
        
    x_train = x_main[train]
    y_train = y_main[train]
    x_test = x_main[test]
    y_test = y_main[test]
    
    model = Sequential()
    model.add(Dense(20, input_dim=x.shape[1], activation='relu'))
    model.add(Dense(5, activation='relu'))
    model.add(Dense(1))
    model.compile(loss='mean_squared_error', optimizer='adam')
    
    model.fit(x_train,y_train,validation_data=(x_test,y_test),
              verbose=0,epochs=EPOCHS)
    
    pred = model.predict(x_test)
    
    oos_y.append(y_test)
    oos_pred.append(pred) 

    # Measure accuracy
    score = np.sqrt(metrics.mean_squared_error(pred,y_test))
    print(f"Fold score (RMSE): {score}")


# Build the oos prediction list and calculate the error.
oos_y = np.concatenate(oos_y)
oos_pred = np.concatenate(oos_pred)
score = np.sqrt(metrics.mean_squared_error(oos_pred,oos_y))
print()
print(f"Cross-validated score (RMSE): {score}")    
    
# Write the cross-validated prediction (from the last neural network)
holdout_pred = model.predict(x_holdout)

score = np.sqrt(metrics.mean_squared_error(holdout_pred,y_holdout))
print(f"Holdout score (RMSE): {score}")\end{lstlisting}
\tcbsubtitle[before skip=\baselineskip]{Output}%
\begin{lstlisting}[upquote=true]
Fold #1
Fold score (RMSE): 0.544195299216696
Fold #2
Fold score (RMSE): 0.48070599342910353
Fold #3
Fold score (RMSE): 0.7034584765928998
Fold #4
Fold score (RMSE): 0.5397141785190473
Fold #5
Fold score (RMSE): 24.126205213080077
Cross-validated score (RMSE): 10.801732731207947
Holdout score (RMSE): 24.097657947297677
\end{lstlisting}
\end{tcolorbox}

\section{Part 5.3: L1 and L2 Regularization to Decrease Overfitting}%
\label{sec:Part5.3L1andL2RegularizationtoDecreaseOverfitting}%
L1 and L2 regularization are two common regularization techniques that can reduce the effects of overfitting%
\index{L1}%
\index{L2}%
\index{overfitting}%
\index{regularization}%
\cite{ng2004feature}%
. These algorithms can either work with an objective function or as a part of the backpropagation algorithm. In both cases, the regularization algorithm is attached to the training algorithm by adding an objective.%
\index{algorithm}%
\index{backpropagation}%
\index{regularization}%
\index{training}%
\index{training algorithm}%
\par%
These algorithms work by adding a weight penalty to the neural network training. This penalty encourages the neural network to keep the weights to small values. Both L1 and L2 calculate this penalty differently. You can add this penalty calculation to the calculated gradients for gradient{-}descent{-}based algorithms, such as backpropagation. The penalty is negatively combined with the objective score for objective{-}function{-}based training, such as simulated annealing.%
\index{algorithm}%
\index{annealing}%
\index{backpropagation}%
\index{calculated}%
\index{gradient}%
\index{L1}%
\index{L2}%
\index{neural network}%
\index{simulated annealing}%
\index{training}%
\par
\begin{figure}[h]%
\centering%
\includegraphics[width=4in]{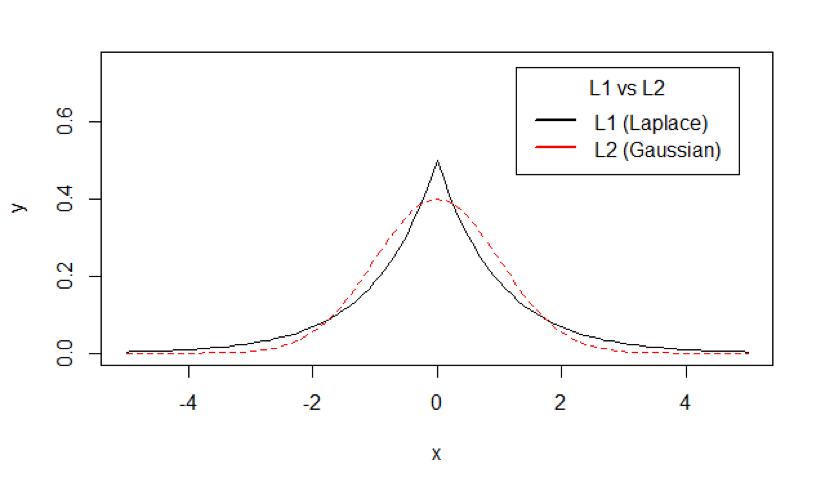}%
\caption{L1 vs L2}%
\label{5.L1L2}%
\end{figure}

%
Both L1 and L2 work differently in that they penalize the size of the weight. L2 will force the weights into a pattern similar to a Gaussian distribution; the L1 will force the weights into a pattern similar to a Laplace distribution, as demonstrated in Figure \ref{5.L1L2}.%
\index{Gaussian}%
\index{L1}%
\index{L2}%
\par%

\par%
As you can see, L1 algorithm is more tolerant of weights further from 0, whereas the L2 algorithm is less tolerant. We will highlight other important differences between L1 and L2 in the following sections. You also need to note that both L1 and L2 count their penalties based only on weights; they do not count penalties on bias values. Keras allows%
\index{algorithm}%
\index{bias}%
\index{Keras}%
\index{L1}%
\index{L2}%
\href{http://tensorlayer.readthedocs.io/en/stable/modules/cost.html}{ l1/l2 to be directly added to your network}%
.%
\par%
\begin{tcolorbox}[size=title,title=Code,breakable]%
\begin{lstlisting}[language=Python, upquote=true]
import pandas as pd
from scipy.stats import zscore

# Read the data set
df = pd.read_csv(
    "https://data.heatonresearch.com/data/t81-558/jh-simple-dataset.csv",
    na_values=['NA','?'])

# Generate dummies for job
df = pd.concat([df,pd.get_dummies(df['job'],prefix="job")],axis=1)
df.drop('job', axis=1, inplace=True)

# Generate dummies for area
df = pd.concat([df,pd.get_dummies(df['area'],prefix="area")],axis=1)
df.drop('area', axis=1, inplace=True)

# Missing values for income
med = df['income'].median()
df['income'] = df['income'].fillna(med)

# Standardize ranges
df['income'] = zscore(df['income'])
df['aspect'] = zscore(df['aspect'])
df['save_rate'] = zscore(df['save_rate'])
df['age'] = zscore(df['age'])
df['subscriptions'] = zscore(df['subscriptions'])

# Convert to numpy - Classification
x_columns = df.columns.drop('product').drop('id')
x = df[x_columns].values
dummies = pd.get_dummies(df['product']) # Classification
products = dummies.columns
y = dummies.values\end{lstlisting}
\end{tcolorbox}%
We now create a Keras network with L1 regression.%
\index{Keras}%
\index{L1}%
\index{regression}%
\par%
\begin{tcolorbox}[size=title,title=Code,breakable]%
\begin{lstlisting}[language=Python, upquote=true]
import pandas as pd
import os
import numpy as np
from sklearn import metrics
from sklearn.model_selection import KFold
from tensorflow.keras.models import Sequential
from tensorflow.keras.layers import Dense, Activation
from tensorflow.keras import regularizers

# Cross-validate
kf = KFold(5, shuffle=True, random_state=42)
    
oos_y = []
oos_pred = []
fold = 0

for train, test in kf.split(x):
    fold+=1
    print(f"Fold #{fold}")
        
    x_train = x[train]
    y_train = y[train]
    x_test = x[test]
    y_test = y[test]
    
    #kernel_regularizer=regularizers.l2(0.01),
    
    model = Sequential()
    # Hidden 1
    model.add(Dense(50, input_dim=x.shape[1], 
            activation='relu',
             activity_regularizer=regularizers.l1(1e-4))) 
    # Hidden 2
    model.add(Dense(25, activation='relu', 
                    activity_regularizer=regularizers.l1(1e-4))) 
     # Output
    model.add(Dense(y.shape[1],activation='softmax'))
    model.compile(loss='categorical_crossentropy', optimizer='adam')

    model.fit(x_train,y_train,validation_data=(x_test,y_test),
              verbose=0,epochs=500)
    
    pred = model.predict(x_test)
    
    oos_y.append(y_test)
    # raw probabilities to chosen class (highest probability)
    pred = np.argmax(pred,axis=1) 
    oos_pred.append(pred)        

    # Measure this fold's accuracy
    y_compare = np.argmax(y_test,axis=1) # For accuracy calculation
    score = metrics.accuracy_score(y_compare, pred)
    print(f"Fold score (accuracy): {score}")


# Build the oos prediction list and calculate the error.
oos_y = np.concatenate(oos_y)
oos_pred = np.concatenate(oos_pred)
oos_y_compare = np.argmax(oos_y,axis=1) # For accuracy calculation

score = metrics.accuracy_score(oos_y_compare, oos_pred)
print(f"Final score (accuracy): {score}")    
    
# Write the cross-validated prediction
oos_y = pd.DataFrame(oos_y)
oos_pred = pd.DataFrame(oos_pred)
oosDF = pd.concat( [df, oos_y, oos_pred],axis=1 )
#oosDF.to_csv(filename_write,index=False)\end{lstlisting}
\tcbsubtitle[before skip=\baselineskip]{Output}%
\begin{lstlisting}[upquote=true]
Fold #1
Fold score (accuracy): 0.64
Fold #2
Fold score (accuracy): 0.6775
Fold #3
Fold score (accuracy): 0.6825
Fold #4
Fold score (accuracy): 0.6675
Fold #5
Fold score (accuracy): 0.645
Final score (accuracy): 0.6625
\end{lstlisting}
\end{tcolorbox}

\section{Part 5.4: Drop Out for Keras to Decrease Overfitting}%
\label{sec:Part5.4DropOutforKerastoDecreaseOverfitting}%
Hinton, Srivastava, Krizhevsky, Sutskever,  Salakhutdinov (2012) introduced the dropout regularization algorithm.%
\index{algorithm}%
\index{dropout}%
\index{Hinton}%
\index{regularization}%
\cite{srivastava2014dropout}%
Although dropout works differently than L1 and L2, it accomplishes the same goal{-}{-}{-}the prevention of overfitting. However, the algorithm does the task by actually removing neurons and connections{-}{-}{-}at least temporarily. Unlike L1 and L2, no weight penalty is added. Dropout does not directly seek to train small weights.\newline%
Dropout works by causing hidden neurons of the neural network to be unavailable during part of the training. Dropping part of the neural network causes the remaining portion to be trained to still achieve a good score even without the dropped neurons. This technique decreases co{-}adaptation between neurons, which results in less overfitting.%
\index{algorithm}%
\index{connection}%
\index{dropout}%
\index{hidden neuron}%
\index{L1}%
\index{L2}%
\index{neural network}%
\index{neuron}%
\index{overfitting}%
\index{training}%
\par%
Most neural network frameworks implement dropout as a separate layer. Dropout layers function like a regular, densely connected neural network layer. The only difference is that the dropout layers will periodically drop some of their neurons during training. You can use dropout layers on regular feedforward neural networks.%
\index{dropout}%
\index{dropout layer}%
\index{feedforward}%
\index{layer}%
\index{neural network}%
\index{neuron}%
\index{SOM}%
\index{training}%
\par
\begin{figure}[h]%
\centering%
\includegraphics[width=4in]{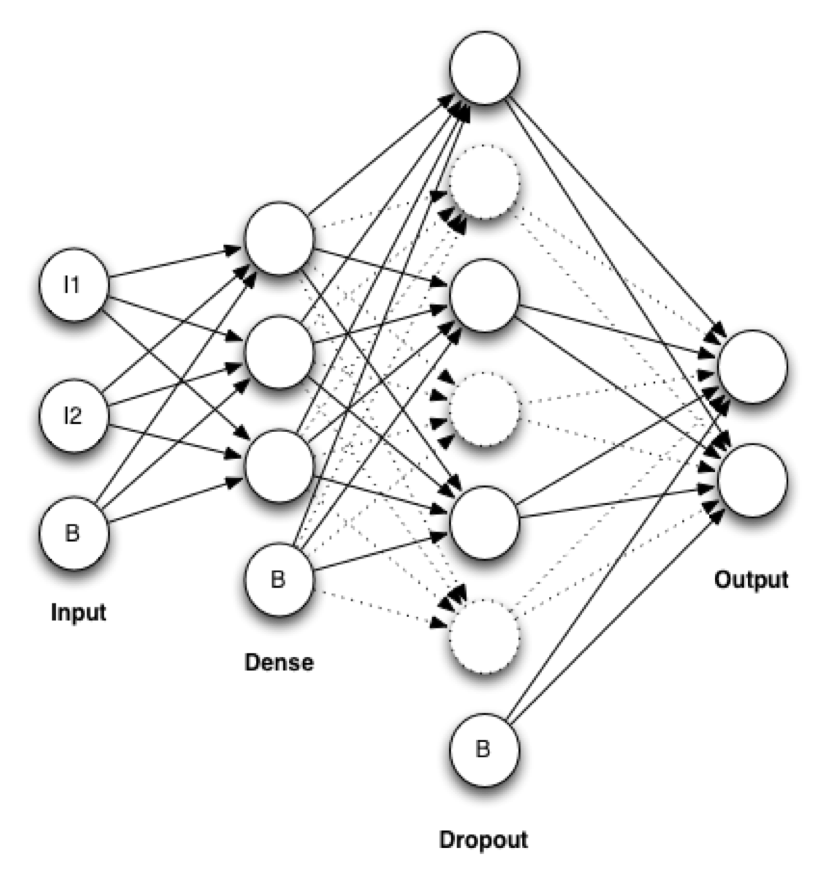}%
\caption{Dropout Regularization}%
\label{5.DROPOUT}%
\end{figure}

%
The program implements a dropout layer as a dense layer that can eliminate some of its neurons. Contrary to popular belief about the dropout layer, the program does not permanently remove these discarded neurons. A dropout layer does not lose any of its neurons during the training process, and it will still have the same number of neurons after training. In this way, the program only temporarily masks the neurons rather than dropping them. \newline%
Figure \ref{5.DROPOUT} shows how a dropout layer might be situated with other layers.%
\index{dense layer}%
\index{dropout}%
\index{dropout layer}%
\index{layer}%
\index{neuron}%
\index{ROC}%
\index{ROC}%
\index{SOM}%
\index{training}%
\par%

\par%
The discarded neurons and their connections are shown as dashed lines. The input layer has two input neurons as well as a bias neuron. The second layer is a dense layer with three neurons and a bias neuron. The third layer is a dropout layer with six regular neurons even though the program has dropped 50\% of them. While the program drops these neurons, it neither calculates nor trains them. However, the final neural network will use all of these neurons for the output. As previously mentioned, the program only temporarily discards the neurons.%
\index{bias}%
\index{bias neuron}%
\index{connection}%
\index{dense layer}%
\index{dropout}%
\index{dropout layer}%
\index{input}%
\index{input layer}%
\index{input neuron}%
\index{layer}%
\index{neural network}%
\index{neuron}%
\index{output}%
\par%
The program chooses different sets of neurons from the dropout layer during subsequent training iterations. Although we chose a probability of 50\% for dropout, the computer will not necessarily drop three neurons. It is as if we flipped a coin for each of the dropout candidate neurons to choose if that neuron was dropped out. You must know that the program should never drop the bias neuron. Only the regular neurons on a dropout layer are candidates.\newline%
The implementation of the training algorithm influences the process of discarding neurons. The dropout set frequently changes once per training iteration or batch. The program can also provide intervals where all neurons are present. Some neural network frameworks give additional hyper{-}parameters to allow you to specify exactly the rate of this interval.%
\index{algorithm}%
\index{bias}%
\index{bias neuron}%
\index{dropout}%
\index{dropout layer}%
\index{iteration}%
\index{layer}%
\index{neural network}%
\index{neuron}%
\index{parameter}%
\index{probability}%
\index{ROC}%
\index{ROC}%
\index{SOM}%
\index{training}%
\index{training algorithm}%
\par%
Why dropout is capable of decreasing overfitting is a common question. The answer is that dropout can reduce the chance of codependency developing between two neurons. Two neurons that develop codependency will not be able to operate effectively when one is dropped out. As a result, the neural network can no longer rely on the presence of every neuron, and it trains accordingly. This characteristic decreases its ability to memorize the information presented, thereby forcing generalization.%
\index{dropout}%
\index{neural network}%
\index{neuron}%
\index{overfitting}%
\par%
Dropout also decreases overfitting by forcing a bootstrapping process upon the neural network. Bootstrapping is a prevalent ensemble technique. Ensembling is a technique of machine learning that combines multiple models to produce a better result than those achieved by individual models. The ensemble is a term that originates from the musical ensembles in which the final music product that the audience hears is the combination of many instruments.%
\index{bootstrapping}%
\index{dropout}%
\index{ensemble}%
\index{learning}%
\index{model}%
\index{neural network}%
\index{overfitting}%
\index{ROC}%
\index{ROC}%
\par%
Bootstrapping is one of the most simple ensemble techniques. The bootstrapping programmer simply trains several neural networks to perform precisely the same task. However, each neural network will perform differently because of some training techniques and the random numbers used in the neural network weight initialization. The difference in weights causes the performance variance. The output from this ensemble of neural networks becomes the average output of the members taken together. This process decreases overfitting through the consensus of differently trained neural networks.%
\index{bootstrapping}%
\index{ensemble}%
\index{neural network}%
\index{output}%
\index{overfitting}%
\index{random}%
\index{ROC}%
\index{ROC}%
\index{SOM}%
\index{training}%
\par%
Dropout works somewhat like bootstrapping. You might think of each neural network that results from a different set of neurons being dropped out as an individual member in an ensemble. As training progresses, the program creates more neural networks in this way. However, dropout does not require the same amount of processing as bootstrapping. The new neural networks created are temporary; they exist only for a training iteration. The final result is also a single neural network rather than an ensemble of neural networks to be averaged together.%
\index{bootstrapping}%
\index{dropout}%
\index{ensemble}%
\index{iteration}%
\index{neural network}%
\index{neuron}%
\index{ROC}%
\index{ROC}%
\index{SOM}%
\index{training}%
\par%
The following animation shows how dropout works:%
\index{dropout}%
\href{https://yusugomori.com/projects/deep-learning/dropout-relu}{ animation link}%
\par%
\begin{tcolorbox}[size=title,title=Code,breakable]%
\begin{lstlisting}[language=Python, upquote=true]
import pandas as pd
from scipy.stats import zscore

# Read the data set
df = pd.read_csv(
    "https://data.heatonresearch.com/data/t81-558/jh-simple-dataset.csv",
    na_values=['NA','?'])

# Generate dummies for job
df = pd.concat([df,pd.get_dummies(df['job'],prefix="job")],axis=1)
df.drop('job', axis=1, inplace=True)

# Generate dummies for area
df = pd.concat([df,pd.get_dummies(df['area'],prefix="area")],axis=1)
df.drop('area', axis=1, inplace=True)

# Missing values for income
med = df['income'].median()
df['income'] = df['income'].fillna(med)

# Standardize ranges
df['income'] = zscore(df['income'])
df['aspect'] = zscore(df['aspect'])
df['save_rate'] = zscore(df['save_rate'])
df['age'] = zscore(df['age'])
df['subscriptions'] = zscore(df['subscriptions'])

# Convert to numpy - Classification
x_columns = df.columns.drop('product').drop('id')
x = df[x_columns].values
dummies = pd.get_dummies(df['product']) # Classification
products = dummies.columns
y = dummies.values\end{lstlisting}
\end{tcolorbox}%
Now we will see how to apply dropout to classification.%
\index{classification}%
\index{dropout}%
\par%
\begin{tcolorbox}[size=title,title=Code,breakable]%
\begin{lstlisting}[language=Python, upquote=true]
########################################
# Keras with dropout for Classification
########################################

import pandas as pd
import os
import numpy as np
from sklearn import metrics
from sklearn.model_selection import KFold
from tensorflow.keras.models import Sequential
from tensorflow.keras.layers import Dense, Activation, Dropout
from tensorflow.keras import regularizers

# Cross-validate
kf = KFold(5, shuffle=True, random_state=42)
    
oos_y = []
oos_pred = []
fold = 0

for train, test in kf.split(x):
    fold+=1
    print(f"Fold #{fold}")
        
    x_train = x[train]
    y_train = y[train]
    x_test = x[test]
    y_test = y[test]
    
    #kernel_regularizer=regularizers.l2(0.01),
    
    model = Sequential()
    model.add(Dense(50, input_dim=x.shape[1], activation='relu')) # Hidden 1
    model.add(Dropout(0.5))
    model.add(Dense(25, activation='relu', \
                activity_regularizer=regularizers.l1(1e-4))) # Hidden 2
    # Usually do not add dropout after final hidden layer
    #model.add(Dropout(0.5)) 
    model.add(Dense(y.shape[1],activation='softmax')) # Output
    model.compile(loss='categorical_crossentropy', optimizer='adam')

    model.fit(x_train,y_train,validation_data=(x_test,y_test),\
              verbose=0,epochs=500)
    
    pred = model.predict(x_test)
    
    oos_y.append(y_test)
    # raw probabilities to chosen class (highest probability)
    pred = np.argmax(pred,axis=1) 
    oos_pred.append(pred)        

    # Measure this fold's accuracy
    y_compare = np.argmax(y_test,axis=1) # For accuracy calculation
    score = metrics.accuracy_score(y_compare, pred)
    print(f"Fold score (accuracy): {score}")


# Build the oos prediction list and calculate the error.
oos_y = np.concatenate(oos_y)
oos_pred = np.concatenate(oos_pred)
oos_y_compare = np.argmax(oos_y,axis=1) # For accuracy calculation

score = metrics.accuracy_score(oos_y_compare, oos_pred)
print(f"Final score (accuracy): {score}")    
    
# Write the cross-validated prediction
oos_y = pd.DataFrame(oos_y)
oos_pred = pd.DataFrame(oos_pred)
oosDF = pd.concat( [df, oos_y, oos_pred],axis=1 )
#oosDF.to_csv(filename_write,index=False)\end{lstlisting}
\tcbsubtitle[before skip=\baselineskip]{Output}%
\begin{lstlisting}[upquote=true]
Fold #1
Fold score (accuracy): 0.68
Fold #2
Fold score (accuracy): 0.695
Fold #3
Fold score (accuracy): 0.7425
Fold #4
Fold score (accuracy): 0.71
Fold #5
Fold score (accuracy): 0.6625
Final score (accuracy): 0.698
\end{lstlisting}
\end{tcolorbox}

\section{Part 5.5: Benchmarking Regularization Techniques}%
\label{sec:Part5.5BenchmarkingRegularizationTechniques}%
Quite a few hyperparameters have been introduced so far.  Tweaking each of these values can have an effect on the score obtained by your neural networks.  Some of the hyperparameters seen so far include:%
\index{hyperparameter}%
\index{neural network}%
\index{parameter}%
\index{SOM}%
\par%
\begin{itemize}[noitemsep]%
\item%
Number of layers in the neural network%
\index{layer}%
\index{neural network}%
\item%
How many neurons in each layer%
\index{layer}%
\index{neuron}%
\item%
What activation functions to use on each layer%
\index{activation function}%
\index{layer}%
\item%
Dropout percent on each layer%
\index{dropout}%
\index{layer}%
\item%
L1 and L2 values on each layer%
\index{L1}%
\index{L2}%
\index{layer}%
\end{itemize}%
To try out each of these hyperparameters you will need to run train neural networks with multiple settings for each hyperparameter.  However, you may have noticed that neural networks often produce somewhat different results when trained multiple times.  This is because the neural networks start with random weights.  Because of this it is necessary to fit and evaluate a neural network times to ensure that one set of hyperparameters are actually better than another.  Bootstrapping can be an effective means of benchmarking (comparing) two sets of hyperparameters.%
\index{bootstrapping}%
\index{hyperparameter}%
\index{neural network}%
\index{parameter}%
\index{random}%
\index{SOM}%
\par%
Bootstrapping is similar to cross{-}validation.  Both go through a number of cycles/folds providing validation and training sets.  However, bootstrapping can have an unlimited number of cycles.  Bootstrapping chooses a new train and validation split each cycle, with replacement.  The fact that each cycle is chosen with replacement means that, unlike cross validation, there will often be repeated rows selected between cycles.  If you run the bootstrap for enough cycles, there will be duplicate cycles.%
\index{bootstrapping}%
\index{training}%
\index{validation}%
\par%
In this part we will use bootstrapping for hyperparameter benchmarking.  We will train a neural network for a specified number of splits (denoted by the SPLITS constant).  For these examples we use 100.  We will compare the average score at the end of the 100.  By the end of the cycles the mean score will have converged somewhat.  This ending score will be a much better basis of comparison than a single cross{-}validation.  Additionally, the average number of epochs will be tracked to give an idea of a possible optimal value.  Because the early stopping validation set is also used to evaluate the the neural network as well, it might be slightly inflated.  This is because we are both stopping and evaluating on the same sample.  However, we are using the scores only as relative measures to determine the superiority of one set of hyperparameters to another, so this slight inflation should not present too much of a problem.%
\index{bootstrapping}%
\index{early stopping}%
\index{hyperparameter}%
\index{neural network}%
\index{parameter}%
\index{SOM}%
\index{validation}%
\par%
Because we are benchmarking, we will display the amount of time taken for each cycle.  The following function can be used to nicely format a time span.%
\par%
\begin{tcolorbox}[size=title,title=Code,breakable]%
\begin{lstlisting}[language=Python, upquote=true]
# Nicely formatted time string
def hms_string(sec_elapsed):
    h = int(sec_elapsed / (60 * 60))
    m = int((sec_elapsed % (60 * 60)) / 60)
    s = sec_elapsed % 60
    return "{}:{:>02}:{:>05.2f}".format(h, m, s)\end{lstlisting}
\end{tcolorbox}%
\subsection{Bootstrapping for Regression}%
\label{subsec:BootstrappingforRegression}%
Regression bootstrapping uses the%
\index{bootstrapping}%
\index{regression}%
\textbf{ ShuffleSplit }%
object to perform the splits.  This technique is similar to%
\textbf{ KFold }%
for cross{-}validation; no balancing occurs.  We will attempt to predict the age column for the%
\index{predict}%
\index{validation}%
\textbf{ jh{-}simple{-}dataset}%
; the following code loads this data.%
\par%
\begin{tcolorbox}[size=title,title=Code,breakable]%
\begin{lstlisting}[language=Python, upquote=true]
import pandas as pd
from scipy.stats import zscore
from sklearn.model_selection import train_test_split

# Read the data set
df = pd.read_csv(
    "https://data.heatonresearch.com/data/t81-558/jh-simple-dataset.csv",
    na_values=['NA','?'])

# Generate dummies for job
df = pd.concat([df,pd.get_dummies(df['job'],prefix="job")],axis=1)
df.drop('job', axis=1, inplace=True)

# Generate dummies for area
df = pd.concat([df,pd.get_dummies(df['area'],prefix="area")],axis=1)
df.drop('area', axis=1, inplace=True)

# Generate dummies for product
df = pd.concat([df,pd.get_dummies(df['product'],prefix="product")],axis=1)
df.drop('product', axis=1, inplace=True)

# Missing values for income
med = df['income'].median()
df['income'] = df['income'].fillna(med)

# Standardize ranges
df['income'] = zscore(df['income'])
df['aspect'] = zscore(df['aspect'])
df['save_rate'] = zscore(df['save_rate'])
df['subscriptions'] = zscore(df['subscriptions'])

# Convert to numpy - Classification
x_columns = df.columns.drop('age').drop('id')
x = df[x_columns].values
y = df['age'].values\end{lstlisting}
\end{tcolorbox}%
The following code performs the bootstrap.  The architecture of the neural network can be adjusted to compare many different configurations.%
\index{architecture}%
\index{neural network}%
\par%
\begin{tcolorbox}[size=title,title=Code,breakable]%
\begin{lstlisting}[language=Python, upquote=true]
import pandas as pd
import os
import numpy as np
import time
import statistics
from sklearn import metrics
from sklearn.model_selection import StratifiedKFold
from tensorflow.keras.models import Sequential
from tensorflow.keras.layers import Dense, Activation
from tensorflow.keras import regularizers
from tensorflow.keras.callbacks import EarlyStopping
from sklearn.model_selection import ShuffleSplit

SPLITS = 50

# Bootstrap
boot = ShuffleSplit(n_splits=SPLITS, test_size=0.1, random_state=42)

# Track progress
mean_benchmark = []
epochs_needed = []
num = 0

# Loop through samples
for train, test in boot.split(x):
    start_time = time.time()
    num+=1

    # Split train and test
    x_train = x[train]
    y_train = y[train]
    x_test = x[test]
    y_test = y[test]

    # Construct neural network
    model = Sequential()
    model.add(Dense(20, input_dim=x_train.shape[1], activation='relu'))
    model.add(Dense(10, activation='relu'))
    model.add(Dense(1))
    model.compile(loss='mean_squared_error', optimizer='adam')
    
    monitor = EarlyStopping(monitor='val_loss', min_delta=1e-3, 
        patience=5, verbose=0, mode='auto', restore_best_weights=True)

    # Train on the bootstrap sample
    model.fit(x_train,y_train,validation_data=(x_test,y_test),
              callbacks=[monitor],verbose=0,epochs=1000)
    epochs = monitor.stopped_epoch
    epochs_needed.append(epochs)
    
    # Predict on the out of boot (validation)
    pred = model.predict(x_test)
  
    # Measure this bootstrap's log loss
    score = np.sqrt(metrics.mean_squared_error(pred,y_test))
    mean_benchmark.append(score)
    m1 = statistics.mean(mean_benchmark)
    m2 = statistics.mean(epochs_needed)
    mdev = statistics.pstdev(mean_benchmark)
    
    # Record this iteration
    time_took = time.time() - start_time
    print(f"#{num}: score={score:.6f}, mean score={m1:.6f},"
          f" stdev={mdev:.6f}", 
          f" epochs={epochs}, mean epochs={int(m2)}", 
          f" time={hms_string(time_took)}")\end{lstlisting}
\tcbsubtitle[before skip=\baselineskip]{Output}%
\begin{lstlisting}[upquote=true]
#1: score=0.630750, mean score=0.630750, stdev=0.000000  epochs=147,
mean epochs=147  time=0:00:12.56
#2: score=1.020895, mean score=0.825823, stdev=0.195072  epochs=101,
mean epochs=124  time=0:00:08.70
#3: score=0.803801, mean score=0.818482, stdev=0.159614  epochs=155,
mean epochs=134  time=0:00:20.85
#4: score=0.540871, mean score=0.749079, stdev=0.183188  epochs=122,
mean epochs=131  time=0:00:10.64
#5: score=0.802589, mean score=0.759781, stdev=0.165240  epochs=116,
mean epochs=128  time=0:00:10.84
#6: score=0.862807, mean score=0.776952, stdev=0.155653  epochs=108,
mean epochs=124  time=0:00:10.65
#7: score=0.550373, mean score=0.744584, stdev=0.164478  epochs=131,
mean epochs=125  time=0:00:10.85
#8: score=0.659148, mean score=0.733904, stdev=0.156428  epochs=118,

...

mean epochs=116  time=0:00:09.33
#49: score=0.911419, mean score=0.747607, stdev=0.185098  epochs=124,
mean epochs=116  time=0:00:10.66
#50: score=0.599252, mean score=0.744639, stdev=0.184411  epochs=132,
mean epochs=116  time=0:00:20.91
\end{lstlisting}
\end{tcolorbox}%
The bootstrapping process for classification is similar, and I present it in the next section.%
\index{bootstrapping}%
\index{classification}%
\index{ROC}%
\index{ROC}%
\par

\subsection{Bootstrapping for Classification}%
\label{subsec:BootstrappingforClassification}%
Regression bootstrapping uses the%
\index{bootstrapping}%
\index{regression}%
\textbf{ StratifiedShuffleSplit }%
class to perform the splits.  This class is similar to%
\textbf{ StratifiedKFold }%
for cross{-}validation, as the classes are balanced so that the sampling does not affect proportions.  To demonstrate this technique, we will attempt to predict the product column for the%
\index{predict}%
\index{validation}%
\textbf{ jh{-}simple{-}dataset}%
; the following code loads this data.%
\par%
\begin{tcolorbox}[size=title,title=Code,breakable]%
\begin{lstlisting}[language=Python, upquote=true]
import pandas as pd
from scipy.stats import zscore

# Read the data set
df = pd.read_csv(
    "https://data.heatonresearch.com/data/t81-558/jh-simple-dataset.csv",
    na_values=['NA','?'])

# Generate dummies for job
df = pd.concat([df,pd.get_dummies(df['job'],prefix="job")],axis=1)
df.drop('job', axis=1, inplace=True)

# Generate dummies for area
df = pd.concat([df,pd.get_dummies(df['area'],prefix="area")],axis=1)
df.drop('area', axis=1, inplace=True)

# Missing values for income
med = df['income'].median()
df['income'] = df['income'].fillna(med)

# Standardize ranges
df['income'] = zscore(df['income'])
df['aspect'] = zscore(df['aspect'])
df['save_rate'] = zscore(df['save_rate'])
df['age'] = zscore(df['age'])
df['subscriptions'] = zscore(df['subscriptions'])

# Convert to numpy - Classification
x_columns = df.columns.drop('product').drop('id')
x = df[x_columns].values
dummies = pd.get_dummies(df['product']) # Classification
products = dummies.columns
y = dummies.values\end{lstlisting}
\end{tcolorbox}%
We now run this data through a number of splits specified by the SPLITS variable. We track the average error through each of these splits.%
\index{error}%
\par%
\begin{tcolorbox}[size=title,title=Code,breakable]%
\begin{lstlisting}[language=Python, upquote=true]
import pandas as pd
import os
import numpy as np
import time
import statistics
from sklearn import metrics
from sklearn.model_selection import StratifiedKFold
from tensorflow.keras.models import Sequential
from tensorflow.keras.layers import Dense, Activation
from tensorflow.keras import regularizers
from tensorflow.keras.callbacks import EarlyStopping
from sklearn.model_selection import StratifiedShuffleSplit

SPLITS = 50

# Bootstrap
boot = StratifiedShuffleSplit(n_splits=SPLITS, test_size=0.1, 
                                random_state=42)

# Track progress
mean_benchmark = []
epochs_needed = []
num = 0

# Loop through samples
for train, test in boot.split(x,df['product']):
    start_time = time.time()
    num+=1

    # Split train and test
    x_train = x[train]
    y_train = y[train]
    x_test = x[test]
    y_test = y[test]

    # Construct neural network
    model = Sequential()
    model.add(Dense(50, input_dim=x.shape[1], activation='relu')) # Hidden 1
    model.add(Dense(25, activation='relu')) # Hidden 2
    model.add(Dense(y.shape[1],activation='softmax')) # Output
    model.compile(loss='categorical_crossentropy', optimizer='adam')
    monitor = EarlyStopping(monitor='val_loss', min_delta=1e-3, 
        patience=25, verbose=0, mode='auto', restore_best_weights=True)

    # Train on the bootstrap sample
    model.fit(x_train,y_train,validation_data=(x_test,y_test),
              callbacks=[monitor],verbose=0,epochs=1000)
    epochs = monitor.stopped_epoch
    epochs_needed.append(epochs)
    
    # Predict on the out of boot (validation)
    pred = model.predict(x_test)
  
    # Measure this bootstrap's log loss
    y_compare = np.argmax(y_test,axis=1) # For log loss calculation
    score = metrics.log_loss(y_compare, pred)
    mean_benchmark.append(score)
    m1 = statistics.mean(mean_benchmark)
    m2 = statistics.mean(epochs_needed)
    mdev = statistics.pstdev(mean_benchmark)
    
    # Record this iteration
    time_took = time.time() - start_time
    print(f"#{num}: score={score:.6f}, mean score={m1:.6f}," +\
          f"stdev={mdev:.6f}, epochs={epochs}, mean epochs={int(m2)}," +\
          f" time={hms_string(time_took)}")\end{lstlisting}
\tcbsubtitle[before skip=\baselineskip]{Output}%
\begin{lstlisting}[upquote=true]
#1: score=0.666342, mean score=0.666342,stdev=0.000000, epochs=66,
mean epochs=66, time=0:00:06.31
#2: score=0.645598, mean score=0.655970,stdev=0.010372, epochs=59,
mean epochs=62, time=0:00:10.63
#3: score=0.676924, mean score=0.662955,stdev=0.013011, epochs=66,
mean epochs=63, time=0:00:10.64
#4: score=0.672602, mean score=0.665366,stdev=0.012017, epochs=84,
mean epochs=68, time=0:00:08.20
#5: score=0.667274, mean score=0.665748,stdev=0.010776, epochs=73,
mean epochs=69, time=0:00:10.65
#6: score=0.706372, mean score=0.672518,stdev=0.018055, epochs=50,
mean epochs=66, time=0:00:04.81
#7: score=0.687937, mean score=0.674721,stdev=0.017565, epochs=71,
mean epochs=67, time=0:00:06.89
#8: score=0.734794, mean score=0.682230,stdev=0.025781, epochs=43,

...

mean epochs=66, time=0:00:04.14
#49: score=0.665493, mean score=0.673305,stdev=0.049060, epochs=60,
mean epochs=66, time=0:00:10.65
#50: score=0.692625, mean score=0.673691,stdev=0.048642, epochs=55,
mean epochs=65, time=0:00:05.22
\end{lstlisting}
\end{tcolorbox}

\subsection{Benchmarking}%
\label{subsec:Benchmarking}%
Now that we've seen how to bootstrap with both classification and regression, we can start to try to optimize the hyperparameters for the%
\index{classification}%
\index{hyperparameter}%
\index{parameter}%
\index{regression}%
\textbf{ jh{-}simple{-}dataset }%
data.  For this example, we will encode for classification of the product column.  Evaluation will be in log loss.%
\index{classification}%
\par%
\begin{tcolorbox}[size=title,title=Code,breakable]%
\begin{lstlisting}[language=Python, upquote=true]
import pandas as pd
from scipy.stats import zscore

# Read the data set
df = pd.read_csv(
    "https://data.heatonresearch.com/data/t81-558/jh-simple-dataset.csv",
    na_values=['NA','?'])

# Generate dummies for job
df = pd.concat([df,pd.get_dummies(df['job'],prefix="job")],axis=1)
df.drop('job', axis=1, inplace=True)

# Generate dummies for area
df = pd.concat([df,pd.get_dummies(df['area'],prefix="area")],
               axis=1)
df.drop('area', axis=1, inplace=True)

# Missing values for income
med = df['income'].median()
df['income'] = df['income'].fillna(med)

# Standardize ranges
df['income'] = zscore(df['income'])
df['aspect'] = zscore(df['aspect'])
df['save_rate'] = zscore(df['save_rate'])
df['age'] = zscore(df['age'])
df['subscriptions'] = zscore(df['subscriptions'])

# Convert to numpy - Classification
x_columns = df.columns.drop('product').drop('id')
x = df[x_columns].values
dummies = pd.get_dummies(df['product']) # Classification
products = dummies.columns
y = dummies.values\end{lstlisting}
\end{tcolorbox}%
I performed some optimization, and the code has the best settings that I could determine. Later in this book, we will see how we can use an automatic process to optimize the hyperparameters.%
\index{hyperparameter}%
\index{optimization}%
\index{parameter}%
\index{ROC}%
\index{ROC}%
\index{SOM}%
\par%
\begin{tcolorbox}[size=title,title=Code,breakable]%
\begin{lstlisting}[language=Python, upquote=true]
import pandas as pd
import os
import numpy as np
import time
import tensorflow.keras.initializers
import statistics
from sklearn import metrics
from sklearn.model_selection import StratifiedKFold
from tensorflow.keras.models import Sequential
from tensorflow.keras.layers import Dense, Activation, Dropout
from tensorflow.keras import regularizers
from tensorflow.keras.callbacks import EarlyStopping
from sklearn.model_selection import StratifiedShuffleSplit
from tensorflow.keras.layers import LeakyReLU,PReLU

SPLITS = 100

# Bootstrap
boot = StratifiedShuffleSplit(n_splits=SPLITS, test_size=0.1)

# Track progress
mean_benchmark = []
epochs_needed = []
num = 0

# Loop through samples
for train, test in boot.split(x,df['product']):
    start_time = time.time()
    num+=1

    # Split train and test
    x_train = x[train]
    y_train = y[train]
    x_test = x[test]
    y_test = y[test]

    # Construct neural network
    model = Sequential()
    model.add(Dense(100, input_dim=x.shape[1], activation=PReLU(), \
        kernel_regularizer=regularizers.l2(1e-4))) # Hidden 1
    model.add(Dropout(0.5))
    model.add(Dense(100, activation=PReLU(), \
        activity_regularizer=regularizers.l2(1e-4))) # Hidden 2
    model.add(Dropout(0.5))
    model.add(Dense(100, activation=PReLU(), \
        activity_regularizer=regularizers.l2(1e-4)
    )) # Hidden 3
#    model.add(Dropout(0.5)) - Usually better performance 
# without dropout on final layer
    model.add(Dense(y.shape[1],activation='softmax')) # Output
    model.compile(loss='categorical_crossentropy', optimizer='adam')
    monitor = EarlyStopping(monitor='val_loss', min_delta=1e-3, 
        patience=100, verbose=0, mode='auto', restore_best_weights=True)

    # Train on the bootstrap sample
    model.fit(x_train,y_train,validation_data=(x_test,y_test), \
              callbacks=[monitor],verbose=0,epochs=1000)
    epochs = monitor.stopped_epoch
    epochs_needed.append(epochs)
    
    # Predict on the out of boot (validation)
    pred = model.predict(x_test)
  
    # Measure this bootstrap's log loss
    y_compare = np.argmax(y_test,axis=1) # For log loss calculation
    score = metrics.log_loss(y_compare, pred)
    mean_benchmark.append(score)
    m1 = statistics.mean(mean_benchmark)
    m2 = statistics.mean(epochs_needed)
    mdev = statistics.pstdev(mean_benchmark)
    
    # Record this iteration
    time_took = time.time() - start_time
    print(f"#{num}: score={score:.6f}, mean score={m1:.6f},"
          f"stdev={mdev:.6f}, epochs={epochs},"
          f"mean epochs={int(m2)}, time={hms_string(time_took)}")\end{lstlisting}
\tcbsubtitle[before skip=\baselineskip]{Output}%
\begin{lstlisting}[upquote=true]
#1: score=0.642887, mean score=0.642887,stdev=0.000000,
epochs=325,mean epochs=325, time=0:00:42.10
#2: score=0.555518, mean score=0.599202,stdev=0.043684,
epochs=208,mean epochs=266, time=0:00:41.74
#3: score=0.605537, mean score=0.601314,stdev=0.035793,
epochs=187,mean epochs=240, time=0:00:24.22
#4: score=0.609415, mean score=0.603339,stdev=0.031195,
epochs=250,mean epochs=242, time=0:00:41.72
#5: score=0.619657, mean score=0.606603,stdev=0.028655,
epochs=201,mean epochs=234, time=0:00:26.10
#6: score=0.638641, mean score=0.611943,stdev=0.028755,
epochs=172,mean epochs=223, time=0:00:41.73
#7: score=0.671137, mean score=0.620399,stdev=0.033731,
epochs=203,mean epochs=220, time=0:00:26.58
#8: score=0.635294, mean score=0.622261,stdev=0.031935,

...

epochs=173,mean epochs=196, time=0:00:22.70
#99: score=0.697473, mean score=0.649279,stdev=0.042577,
epochs=172,mean epochs=196, time=0:00:41.79
#100: score=0.678298, mean score=0.649569,stdev=0.042462,
epochs=169,mean epochs=196, time=0:00:21.90
\end{lstlisting}
\end{tcolorbox}

\chapter{Convolutional Neural Networks (CNN) for Computer Vision}%
\label{chap:ConvolutionalNeuralNetworks(CNN)forComputerVision}%
\section{Part 6.1: Image Processing in Python}%
\label{sec:Part6.1ImageProcessinginPython}%
Computer vision requires processing images. These images might come from a video stream, a camera, or files on a storage drive. We begin this chapter by looking at how to process images with Python. To use images in Python, we will make use of the Pillow package. The following program uses Pillow to load and display an image.%
\index{computer vision}%
\index{Python}%
\index{ROC}%
\index{ROC}%
\index{video}%
\par%
\begin{tcolorbox}[size=title,title=Code,breakable]%
\begin{lstlisting}[language=Python, upquote=true]
from PIL import Image, ImageFile
from matplotlib.pyplot import imshow
import requests
from io import BytesIO
import numpy as np

%matplotlib inline

url = "https://data.heatonresearch.com/images/jupyter/brookings.jpeg"

response = requests.get(url,headers={'User-Agent': 'Mozilla/5.0'})
img = Image.open(BytesIO(response.content))
img.load()

print(np.asarray(img))

img\end{lstlisting}
\tcbsubtitle[before skip=\baselineskip]{Output}%
\includegraphics[width=4in]{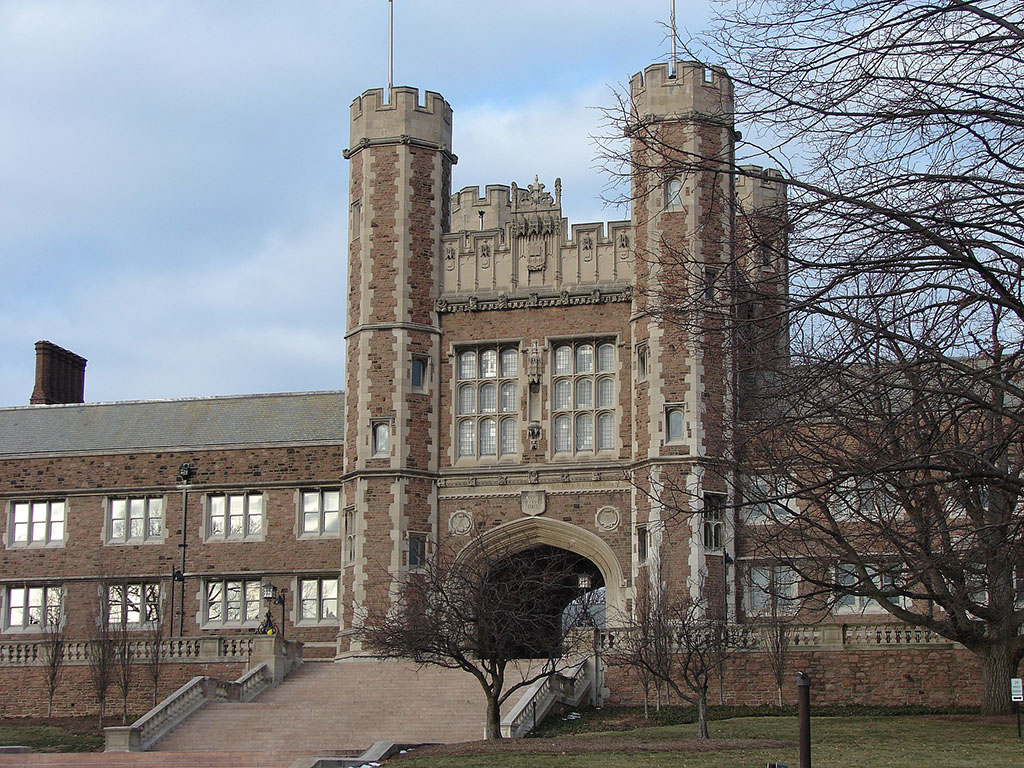}%
\begin{lstlisting}[upquote=true]
[[[199 213 240]
  [200 214 240]
  [200 214 240]
  ...
  [ 86  34  96]
  [ 48   4  57]
  [ 57  21  65]]
 [[199 213 239]
  [200 214 240]
  [200 214 240]
  ...
  [215 215 251]
  [252 242 255]
  [237 218 250]]
 [[200 214 240]

...

  [131  98  91]
  ...
  [ 86  82  57]
  [ 89  85  60]
  [ 89  85  60]]]
\end{lstlisting}
\end{tcolorbox}%
\subsection{Creating Images from Pixels in Python}%
\label{subsec:CreatingImagesfromPixelsinPython}%
You can use Pillow to create an image from a 3D NumPy cube{-}shaped array.  The rows and columns specify the pixels.  The third dimension (size 3) defines red, green, and blue color values.  The following code demonstrates creating a simple image from a NumPy array.%
\index{NumPy}%
\par%
\begin{tcolorbox}[size=title,title=Code,breakable]%
\begin{lstlisting}[language=Python, upquote=true]
from PIL import Image
import numpy as np

w, h = 64, 64
data = np.zeros((h, w, 3), dtype=np.uint8)

# Yellow
for row in range(32):
    for col in range(32):
        data[row,col] = [255,255,0]
        
# Red
for row in range(32):
    for col in range(32):
        data[row+32,col] = [255,0,0]
        
# Green
for row in range(32):
    for col in range(32):
        data[row+32,col+32] = [0,255,0]        
        
# Blue
for row in range(32):
    for col in range(32):
        data[row,col+32] = [0,0,255]                
        

img = Image.fromarray(data, 'RGB')
img\end{lstlisting}
\tcbsubtitle[before skip=\baselineskip]{Output}%
\includegraphics[width=1in]{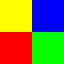}%
\end{tcolorbox}

\subsection{Transform Images in Python (at the pixel level)}%
\label{subsec:TransformImagesinPython(atthepixellevel)}%
We can combine the last two programs and modify images.  Here we take the mean color of each pixel and form a grayscale image.%
\par%
\begin{tcolorbox}[size=title,title=Code,breakable]%
\begin{lstlisting}[language=Python, upquote=true]
from PIL import Image, ImageFile
from matplotlib.pyplot import imshow
import requests
from io import BytesIO

%matplotlib inline

url = "https://data.heatonresearch.com/images/jupyter/brookings.jpeg"
response = requests.get(url,headers={'User-Agent': 'Mozilla/5.0'})

img = Image.open(BytesIO(response.content))
img.load()

img_array = np.asarray(img)
rows = img_array.shape[0]
cols = img_array.shape[1]

print("Rows: {}, Cols: {}".format(rows,cols))

# Create new image
img2_array = np.zeros((rows, cols, 3), dtype=np.uint8)
for row in range(rows):
    for col in range(cols):
        t = np.mean(img_array[row,col])
        img2_array[row,col] = [t,t,t]

img2 = Image.fromarray(img2_array, 'RGB')
img2\end{lstlisting}
\tcbsubtitle[before skip=\baselineskip]{Output}%
\includegraphics[width=4in]{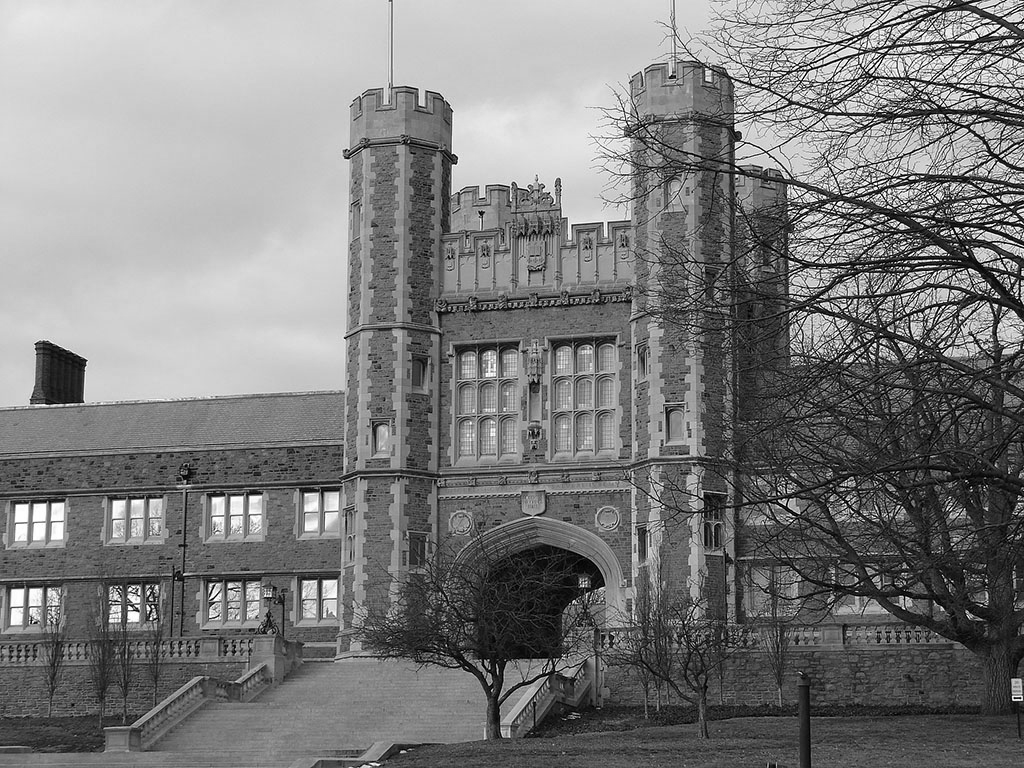}%
\begin{lstlisting}[upquote=true]
Rows: 768, Cols: 1024
\end{lstlisting}
\end{tcolorbox}

\subsection{Standardize Images}%
\label{subsec:StandardizeImages}%
When processing several images together, it is sometimes essential to standardize them.  The following code reads a sequence of images and causes them to all be of the same size and perfectly square.  If the input images are not square, cropping will occur.%
\index{input}%
\index{ROC}%
\index{ROC}%
\index{SOM}%
\par%
\begin{tcolorbox}[size=title,title=Code,breakable]%
\begin{lstlisting}[language=Python, upquote=true]
%matplotlib inline
from PIL import Image, ImageFile
from matplotlib.pyplot import imshow
import requests
import numpy as np
from io import BytesIO
from IPython.display import display, HTML

images = [
  "https://data.heatonresearch.com/images/jupyter/brookings.jpeg",
  "https://data.heatonresearch.com/images/jupyter/SeigleHall.jpeg",
  "https://data.heatonresearch.com/images/jupyter/WUSTLKnight.jpeg"   
]

def crop_square(image):        
    width, height = image.size
    
    # Crop the image, centered
    new_width = min(width,height)
    new_height = new_width
    left = (width - new_width)/2
    top = (height - new_height)/2
    right = (width + new_width)/2
    bottom = (height + new_height)/2
    return image.crop((left, top, right, bottom))
    
x = [] 
    
for url in images:
    ImageFile.LOAD_TRUNCATED_IMAGES = False
    response = requests.get(url,headers={'User-Agent': 'Mozilla/5.0'})
    img = Image.open(BytesIO(response.content))
    img.load()
    img = crop_square(img)
    img = img.resize((128,128), Image.ANTIALIAS)
    print(url)
    display(img)
    img_array = np.asarray(img)
    img_array = img_array.flatten()
    img_array = img_array.astype(np.float32)
    img_array = (img_array-128)/128
    x.append(img_array)
    

x = np.array(x)

print(x.shape)\end{lstlisting}
\tcbsubtitle[before skip=\baselineskip]{Output}%
\includegraphics[width=1in]{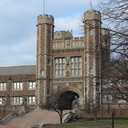}%
\begin{lstlisting}[upquote=true]
https://data.heatonresearch.com/images/jupyter/brookings.jpeg
\end{lstlisting}
\includegraphics[width=1in]{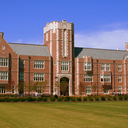}%
\begin{lstlisting}[upquote=true]
https://data.heatonresearch.com/images/jupyter/SeigleHall.jpeg
\end{lstlisting}
\includegraphics[width=1in]{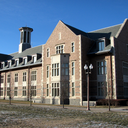}%
\begin{lstlisting}[upquote=true]
https://data.heatonresearch.com/images/jupyter/WUSTLKnight.jpeg
\end{lstlisting}
\begin{lstlisting}[upquote=true]
(3, 49152)
\end{lstlisting}
\end{tcolorbox}

\subsection{Adding Noise to an Image}%
\label{subsec:AddingNoisetoanImage}%
Sometimes it is beneficial to add noise to images. We might use noise to augment images to generate more training data or modify images to test the recognition capabilities of neural networks. It is essential to see how to add noise to an image. There are many ways to add such noise. The following code adds random black squares to the image to produce noise.%
\index{neural network}%
\index{random}%
\index{SOM}%
\index{training}%
\par%
\begin{tcolorbox}[size=title,title=Code,breakable]%
\begin{lstlisting}[language=Python, upquote=true]
from PIL import Image, ImageFile
from matplotlib.pyplot import imshow
import requests
from io import BytesIO

%matplotlib inline


def add_noise(a):
    a2 = a.copy()
    rows = a2.shape[0]
    cols = a2.shape[1]
    s = int(min(rows,cols)/20) # size of spot is 1/20 of smallest dimension
    
    for i in range(100):
        x = np.random.randint(cols-s)
        y = np.random.randint(rows-s)
        a2[y:(y+s),x:(x+s)] = 0
        
    return a2

url = "https://data.heatonresearch.com/images/jupyter/brookings.jpeg"

response = requests.get(url,headers={'User-Agent': 'Mozilla/5.0'})
img = Image.open(BytesIO(response.content))
img.load()

img_array = np.asarray(img)
rows = img_array.shape[0]
cols = img_array.shape[1]

print("Rows: {}, Cols: {}".format(rows,cols))

# Create new image
img2_array = img_array.astype(np.uint8)
print(img2_array.shape)
img2_array = add_noise(img2_array)
img2 = Image.fromarray(img2_array, 'RGB')
img2\end{lstlisting}
\tcbsubtitle[before skip=\baselineskip]{Output}%
\includegraphics[width=4in]{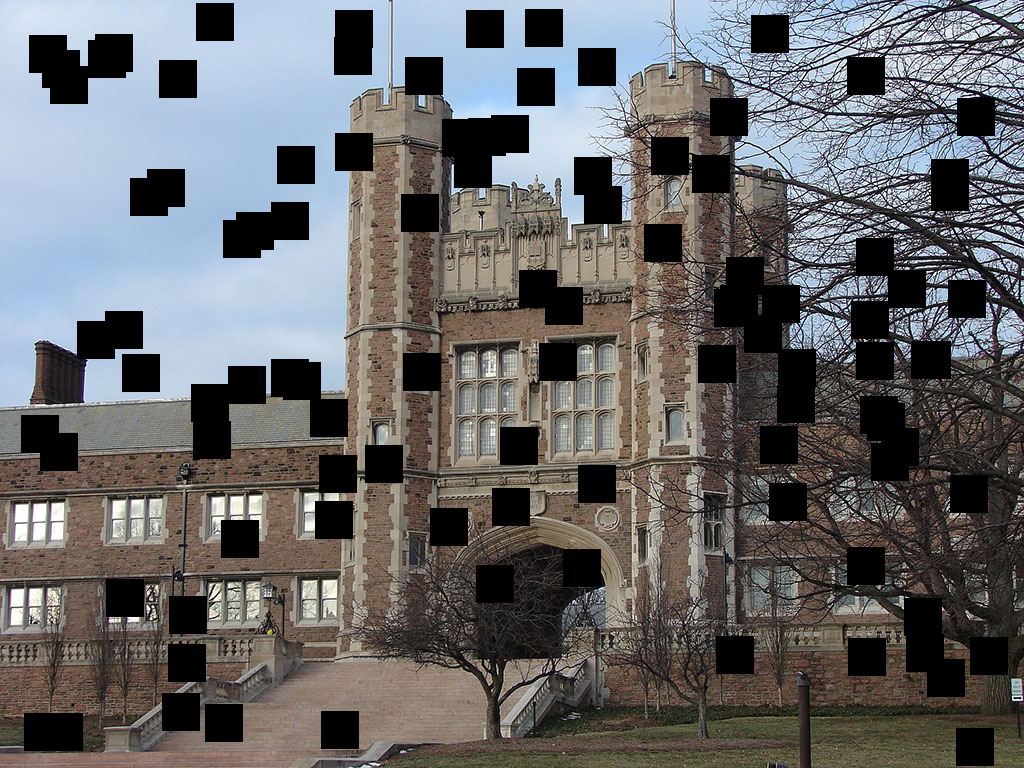}%
\begin{lstlisting}[upquote=true]
Rows: 768, Cols: 1024
(768, 1024, 3)
\end{lstlisting}
\end{tcolorbox}

\subsection{Preprocessing Many Images}%
\label{subsec:PreprocessingManyImages}%
To download images, we define several paths. We will download sample images of paperclips from the URL specified by%
\textbf{ DOWNLOAD\_SOURCE}%
. Once downloaded, we will unzip and perform the preprocessing on these paper clips. I mean for this code as a starting point for other image preprocessing.%
\index{ROC}%
\index{ROC}%
\par%
\begin{tcolorbox}[size=title,title=Code,breakable]%
\begin{lstlisting}[language=Python, upquote=true]
import os

URL = "https://github.com/jeffheaton/data-mirror/releases/"
#DOWNLOAD_SOURCE = URL+"download/v1/iris-image.zip"
DOWNLOAD_SOURCE = URL+"download/v1/paperclips.zip"
DOWNLOAD_NAME = DOWNLOAD_SOURCE[DOWNLOAD_SOURCE.rfind('/')+1:]

if COLAB:
  PATH = "/content"
  EXTRACT_TARGET = os.path.join(PATH,"clips")
  SOURCE = os.path.join(PATH, "/content/clips/paperclips")
  TARGET = os.path.join(PATH,"/content/clips-processed")
else:
  # I used this locally on my machine, you may need different
  PATH = "/Users/jeff/temp"
  EXTRACT_TARGET = os.path.join(PATH,"clips")
  SOURCE = os.path.join(PATH, "clips/paperclips")
  TARGET = os.path.join(PATH,"clips-processed")\end{lstlisting}
\end{tcolorbox}%
Next, we download the images. This part depends on the origin of your images. The following code downloads images from a URL, where a ZIP file contains the images. The code unzips the ZIP file.%
\par%
\begin{tcolorbox}[size=title,title=Code,breakable]%
\begin{lstlisting}[language=Python, upquote=true]
!wget -O {os.path.join(PATH,DOWNLOAD_NAME)} {DOWNLOAD_SOURCE}
!mkdir -p {SOURCE}
!mkdir -p {TARGET}
!mkdir -p {EXTRACT_TARGET}
!unzip -o -j -d {SOURCE} {os.path.join(PATH, DOWNLOAD_NAME)} >/dev/null\end{lstlisting}
\end{tcolorbox}%
The following code contains functions that we use to preprocess the images. The%
\index{ROC}%
\index{ROC}%
\textbf{ crop\_square }%
function converts images to a square by cropping extra data. The%
\textbf{ scale }%
function increases or decreases the size of an image. The%
\textbf{ standardize }%
function ensures an image is full color; a mix of color and grayscale images can be problematic.%
\par%
\begin{tcolorbox}[size=title,title=Code,breakable]%
\begin{lstlisting}[language=Python, upquote=true]
import imageio
import glob
from tqdm import tqdm
from PIL import Image
import os
        
def scale(img, scale_width, scale_height):
    # Scale the image
    img = img.resize((
        scale_width, 
        scale_height), 
        Image.ANTIALIAS)
            
    return img

def standardize(image):
    rgbimg = Image.new("RGB", image.size)
    rgbimg.paste(image)
    return rgbimg

def fail_below(image, check_width, check_height):
    width, height = image.size
    assert width == check_width
    assert height == check_height\end{lstlisting}
\end{tcolorbox}%
Next, we loop through each image. The images are loaded, and you can apply any desired transformations. Ultimately, the script saves the images as JPG.%
\par%
\begin{tcolorbox}[size=title,title=Code,breakable]%
\begin{lstlisting}[language=Python, upquote=true]
files = glob.glob(os.path.join(SOURCE,"*.jpg"))

for file in tqdm(files):
    try:
        target = ""
        name = os.path.basename(file)
        filename, _ = os.path.splitext(name)
        img = Image.open(file)
        img = standardize(img)
        img = crop_square(img)
        img = scale(img, 128, 128)
        #fail_below(img, 128, 128)

        target = os.path.join(TARGET,filename+".jpg")
        img.save(target, quality=25)
    except KeyboardInterrupt:
        print("Keyboard interrupt")
        break
    except AssertionError:
        print("Assertion")
        break
    except:
        print("Unexpected exception while processing image source: " \
              f"{file}, target: {target}" , exc_info=True)\end{lstlisting}
\end{tcolorbox}%
Now we can zip the preprocessed files and store them somewhere.%
\index{ROC}%
\index{ROC}%
\index{SOM}%
\par

\subsection{Module 6 Assignment}%
\label{subsec:Module6Assignment}%
You can find the first assignment here:%
\href{https://github.com/jeffheaton/t81_558_deep_learning/blob/master/assignments/assignment_yourname_class6.ipynb}{ assignment 6}%
\par

\section{Part 6.2: Keras Neural Networks for Digits and Fashion MNIST}%
\label{sec:Part6.2KerasNeuralNetworksforDigitsandFashionMNIST}%
This module will focus on computer vision. There are some important differences and similarities with previous neural networks.%
\index{computer vision}%
\index{neural network}%
\index{SOM}%
\par%
\begin{itemize}[noitemsep]%
\item%
We will usually use classification, though regression is still an option.%
\index{classification}%
\index{regression}%
\item%
The input to the neural network is now 3D (height, width, color)%
\index{input}%
\index{neural network}%
\item%
Data are not transformed; no z{-}scores or dummy variables.%
\index{Z{-}Score}%
\item%
Processing time is much longer.%
\index{ROC}%
\index{ROC}%
\item%
We now have different layer times: dense layers (just like before), convolution layers, and max{-}pooling layers.%
\index{convolution}%
\index{dense layer}%
\index{layer}%
\item%
Data will no longer arrive as CSV files. TensorFlow provides some utilities for going directly from the image to the input for a neural network.%
\index{CSV}%
\index{input}%
\index{neural network}%
\index{SOM}%
\index{TensorFlow}%
\end{itemize}%
\subsection{Common Computer Vision Data Sets}%
\label{subsec:CommonComputerVisionDataSets}%
There are many data sets for computer vision. Two of the most popular classic datasets are the MNIST digits data set and the CIFAR image data sets. We will not use either of these datasets in this course, but it is important to be familiar with them since neural network texts often refer to them.%
\index{computer vision}%
\index{dataset}%
\index{digit}%
\index{MNIST}%
\index{neural network}%
\par
\begin{figure}[h]%
\centering%
\includegraphics[width=4in]{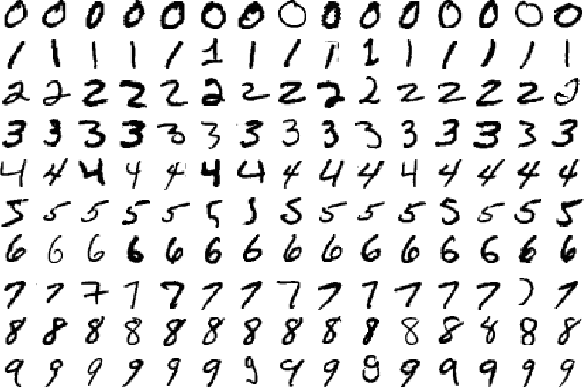}%
\caption{MNIST Data Set}%
\label{6.MNIST}%
\end{figure}

%
The%
\href{http://yann.lecun.com/exdb/mnist/}{ MNIST Digits Data Set }%
is very popular in the neural network research community. You can see a sample of it in Figure \ref{6.MNIST}.%
\index{MNIST}%
\index{neural network}%
\par%

\par
\begin{figure}[h]%
\centering%
\includegraphics[width=4in]{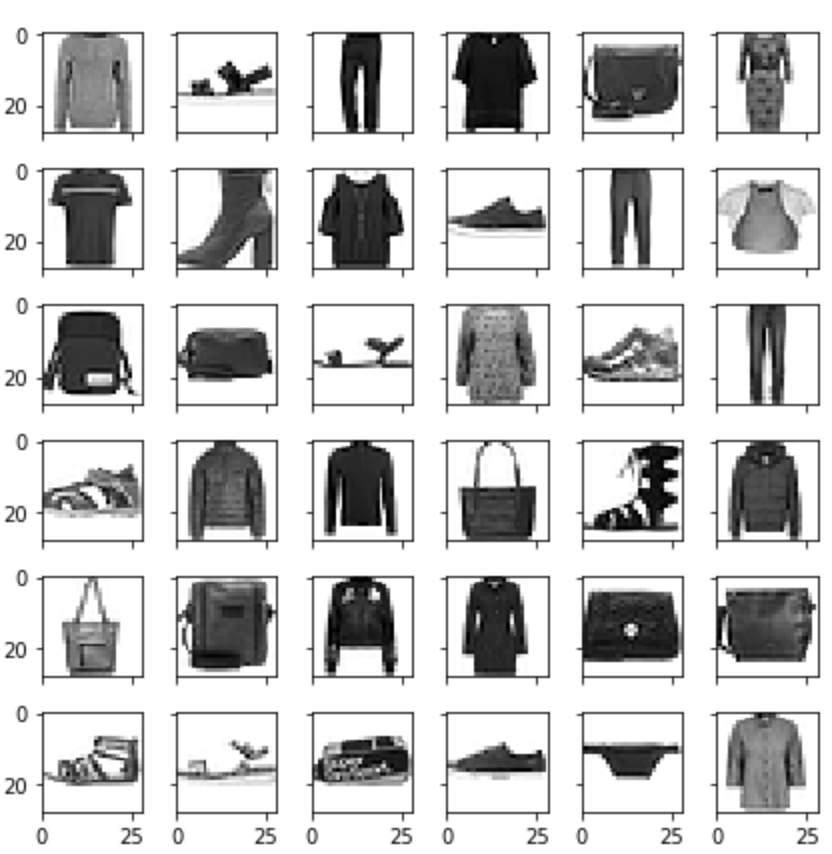}%
\caption{MNIST Fashon Data Set}%
\label{6.MNIST-FASHION}%
\end{figure}

%
\href{https://www.kaggle.com/zalando-research/fashionmnist}{Fashion{-}MNIST }%
is a dataset of%
\index{dataset}%
\href{https://jobs.zalando.com/tech/}{ Zalando }%
's article images{-}{-}{-}consisting of a training set of 60,000 examples and a test set of 10,000 examples. Each example is a 28x28 grayscale image associated with a label from 10 classes. Fashion{-}MNIST is a direct drop{-}in replacement for the original%
\index{MNIST}%
\index{training}%
\href{http://yann.lecun.com/exdb/mnist/}{ MNIST dataset }%
for benchmarking machine learning algorithms. It shares the same image size and structure of training and testing splits. You can see this data in Figure \ref{6.MNIST-FASHION}.%
\index{algorithm}%
\index{learning}%
\index{MNIST}%
\index{training}%
\par%

\par%
The%
\href{https://www.cs.toronto.edu/~kriz/cifar.html}{ CIFAR{-}10 and CIFAR{-}100 }%
datasets are also frequently used by the neural network research community.%
\index{dataset}%
\index{neural network}%
\par%

\begin{figure}[h]%
\centering%
\includegraphics[width=4in]{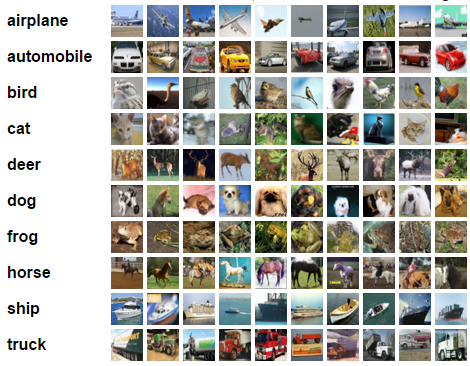}%
\caption{CIFAR Data Set}%
\label{6.CIFAR}%
\end{figure}

\par%
The CIFAR{-}10 data set contains low{-}rez images that are divided into 10 classes. The CIFAR{-}100 data set contains 100 classes in a hierarchy.%
\par

\subsection{Convolutional Neural Networks (CNNs)}%
\label{subsec:ConvolutionalNeuralNetworks(CNNs)}%
The convolutional neural network (CNN) is a neural network technology that has profoundly impacted the area of computer vision (CV). Fukushima  (1980)%
\index{CNN}%
\index{computer vision}%
\index{convolution}%
\index{convolutional}%
\index{neural network}%
\cite{fukushima1980neocognitron}%
introduced the original concept of a convolutional neural network, and   LeCun, Bottou, Bengio  Haffner (1998)%
\index{convolution}%
\index{convolutional}%
\index{LeCun}%
\index{neural network}%
\cite{lecun1995convolutional}%
greatly improved this work. From this research, Yan LeCun introduced the famous LeNet{-}5 neural network architecture. This chapter follows the LeNet{-}5 style of convolutional neural network.%
\index{architecture}%
\index{convolution}%
\index{convolutional}%
\index{LeCun}%
\index{LeNET{-}5}%
\index{network architecture}%
\index{neural network}%
\linebreak%
Although computer vision primarily uses CNNs, this technology has some applications outside of the field. You need to realize that if you want to utilize CNNs on non{-}visual data, you must find a way to encode your data to mimic the properties of visual data.%
\index{CNN}%
\index{computer vision}%
\index{SOM}%
\par%
The order of the input array elements is crucial to the training. In contrast, most neural networks that are not CNNs treat their input data as a long vector of values, and the order in which you arrange the incoming features in this vector is irrelevant. You cannot change the order for these types of neural networks after you have trained the network.%
\index{CNN}%
\index{feature}%
\index{input}%
\index{neural network}%
\index{training}%
\index{vector}%
\par%
The CNN network arranges the inputs into a grid. This arrangement worked well with images because the pixels in closer proximity to each other are important to each other. The order of pixels in an image is significant. The human body is a relevant example of this type of order. For the design of the face, we are accustomed to eyes being near to each other.%
\index{CNN}%
\index{input}%
\par
\begin{figure}[h]%
\centering%
\includegraphics[width=4in]{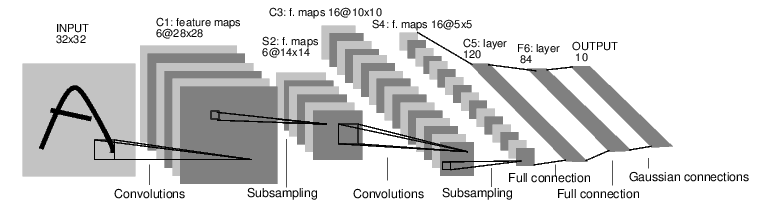}%
\caption{A LeNET{-}5 Network (LeCun, 1998)}%
\label{6.LENET}%
\end{figure}

%
This advance in CNNs is due to years of research on biological eyes. In other words, CNNs utilize overlapping fields of input to simulate features of biological eyes. Until this breakthrough, AI had been unable to reproduce the capabilities of biological vision.\newline%
Scale, rotation, and noise have presented challenges for AI computer vision research. You can observe the complexity of biological eyes in the example that follows. A friend raises a sheet of paper with a large number written on it. As your friend moves nearer to you, the number is still identifiable. In the same way, you can still identify the number when your friend rotates the paper. Lastly, your friend creates noise by drawing lines on the page, but you can still identify the number. As you can see, these examples demonstrate the high function of the biological eye and allow you to understand better the research breakthrough of CNNs. That is, this neural network can process scale, rotation, and noise in the field of computer vision. You can see this network structure in Figure \ref{6.LENET}.%
\index{CNN}%
\index{computer vision}%
\index{feature}%
\index{input}%
\index{neural network}%
\index{ROC}%
\index{ROC}%
\par%

\par%
So far, we have only seen one layer type (dense layers). By the end of this book we will have seen:%
\index{dense layer}%
\index{layer}%
\par%
\begin{itemize}[noitemsep]%
\item%
\textbf{Dense Layers }%
{-} Fully connected layers.%
\index{layer}%
\item%
\textbf{Convolution Layers }%
{-} Used to scan across images.%
\item%
\textbf{Max Pooling Layers }%
{-} Used to downsample images.%
\index{downsample}%
\item%
\textbf{Dropout Layers }%
{-} Used to add regularization.%
\index{regularization}%
\item%
\textbf{LSTM and Transformer Layers }%
{-} Used for time series data.%
\end{itemize}

\subsection{Convolution Layers}%
\label{subsec:ConvolutionLayers}%
The first layer that we will examine is the convolutional layer. We will begin by looking at the hyper{-}parameters that you must specify for a convolutional layer in most neural network frameworks that support the CNN:%
\index{CNN}%
\index{convolution}%
\index{convolutional}%
\index{layer}%
\index{neural network}%
\index{parameter}%
\par%
\begin{itemize}[noitemsep]%
\item%
Number of filters%
\item%
Filter Size%
\item%
Stride%
\item%
Padding%
\item%
Activation Function/Non{-}Linearity%
\index{activation function}%
\index{linear}%
\end{itemize}%
The primary purpose of a convolutional layer is to detect features such as edges, lines, blobs of color, and other visual elements. The filters can detect these features. The more filters we give to a convolutional layer, the more features it can see.%
\index{convolution}%
\index{convolutional}%
\index{feature}%
\index{layer}%
\par%
A filter is a square{-}shaped object that scans over the image. A grid can represent the individual pixels of a grid. You can think of the convolutional layer as a smaller grid that sweeps left to right over each image row. There is also a hyperparameter that specifies both the width and height of the square{-}shaped filter. The following figure shows this configuration in which you see the six convolutional filters sweeping over the image grid:%
\index{convolution}%
\index{convolutional}%
\index{hyperparameter}%
\index{layer}%
\index{parameter}%
\par%
A convolutional layer has weights between it and the previous layer or image grid. Each pixel on each convolutional layer is a weight. Therefore, the number of weights between a convolutional layer and its predecessor layer or image field is the following:%
\index{convolution}%
\index{convolutional}%
\index{layer}%
\par%
\begin{tcolorbox}[size=title,breakable]%
\begin{lstlisting}[upquote=true]
[FilterSize] * [FilterSize] * [# of Filters]
\end{lstlisting}
\end{tcolorbox}%
For example, if the filter size were 5 (5x5) for 10 filters, there would be 250 weights.%
\par
\begin{figure}[h]%
\centering%
\includegraphics[width=3in]{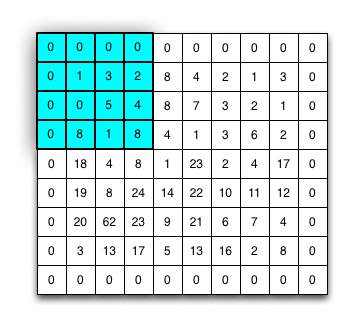}%
\caption{Convolutional Neural Network}%
\label{6.CNN}%
\end{figure}

%
You need to understand how the convolutional filters sweep across the previous layer's output or image grid. Figure \ref{6.CNN} illustrates the sweep:%
\index{CNN}%
\index{convolution}%
\index{convolutional}%
\index{layer}%
\index{output}%
\par%

\par%
The above figure shows a convolutional filter with 4 and a padding size of 1. The padding size is responsible for the border of zeros in the area that the filter sweeps. Even though the image is 8x7, the extra padding provides a virtual image size of 9x8 for the filter to sweep across. The stride specifies the number of positions the convolutional filters will stop. The convolutional filters move to the right, advancing by the number of cells specified in the stride. Once you reach the far right, the convolutional filter moves back to the far left; then, it moves down by the stride amount and\newline%
continues to the right again.%
\index{convolution}%
\index{convolutional}%
\par%
Some constraints exist concerning the size of the stride. The stride cannot be 0. The convolutional filter would never move if you set the stride. Furthermore, neither the stride nor the convolutional filter size can be larger than the previous grid. There are additional constraints on the stride (%
\index{convolution}%
\index{convolutional}%
\index{SOM}%
\textit{s}%
), padding (%
\textit{p}%
), and the filter width (%
\textit{f}%
) for an image of width (%
\textit{w}%
). Specifically, the convolutional filter must be able to start at the far left or top border, move a certain number of strides, and land on the far right or bottom border. The following equation shows the number of steps a convolutional operator\newline%
must take to cross the image:%
\index{convolution}%
\index{convolutional}%
\par%
\vspace{2mm}%
\begin{equation*}
 steps = \frac{w - f + 2p}{s}+1 
\end{equation*}
\vspace{2mm}%
\par%
The number of steps must be an integer. In other words, it cannot have decimal places. The purpose of the padding (%
\textit{p}%
) is to be adjusted to make this equation become an integer value.%
\par

\subsection{Max Pooling Layers}%
\label{subsec:MaxPoolingLayers}%
Max{-}pool layers downsample a 3D box to a new one with smaller dimensions. Typically, you can always place a max{-}pool layer immediately following the convolutional layer. The LENET shows the max{-}pool layer immediately after layers C1 and C3. These max{-}pool layers progressively decrease the size of the dimensions of the 3D boxes passing through them. This technique can avoid overfitting (Krizhevsky, Sutskever  Hinton, 2012).%
\index{convolution}%
\index{convolutional}%
\index{downsample}%
\index{Hinton}%
\index{layer}%
\index{overfitting}%
\par%
A pooling layer has the following hyper{-}parameters:%
\index{layer}%
\index{parameter}%
\par%
\begin{itemize}[noitemsep]%
\item%
Spatial Extent (%
\textit{f}%
)%
\item%
Stride (%
\textit{s}%
)%
\end{itemize}%
Unlike convolutional layers, max{-}pool layers do not use padding. Additionally, max{-}pool layers have no weights, so training does not affect them. These layers downsample their 3D box input. The 3D box output by a max{-}pool layer will have a width equal to this equation:%
\index{convolution}%
\index{convolutional}%
\index{downsample}%
\index{input}%
\index{layer}%
\index{output}%
\index{training}%
\par%
\vspace{2mm}%
\begin{equation*}
 w_2 = \frac{w_1 - f}{s} + 1 
\end{equation*}
\vspace{2mm}%
\par%
The height of the 3D box produced by the max{-}pool layer is calculated similarly with this equation:%
\index{calculated}%
\index{layer}%
\par%
\vspace{2mm}%
\begin{equation*}
 h_2 = \frac{h_1 - f}{s} + 1 
\end{equation*}
\vspace{2mm}%
\par%
The depth of the 3D box produced by the max{-}pool layer is equal to the depth the 3D box received as input. The most common setting for the hyper{-}parameters of a max{-}pool layer is f=2 and s=2. The spatial extent (f) specifies that boxes of 2x2 will be scaled down to single pixels. Of these four pixels, the pixel with the maximum value will represent the 2x2 pixel in the new grid. Because squares of size 4 are replaced with size 1, 75\% of the pixel information is lost. The following figure shows this transformation as a 6x6 grid becomes a 3x3:%
\index{input}%
\index{layer}%
\index{parameter}%
\par%

\begin{figure}[h]%
\centering%
\includegraphics[width=3in]{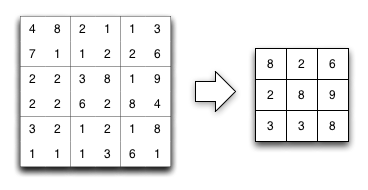}%
\caption{Max Pooling Layer}%
\label{6.MAXPOOL}%
\end{figure}

\par%
Of course, the above diagram shows each pixel as a single number. A grayscale image would have this characteristic. We usually take the average of the three numbers for an RGB image to determine which pixel has the maximum value.%
\par

\subsection{Regression Convolutional Neural Networks}%
\label{subsec:RegressionConvolutionalNeuralNetworks}%
We will now look at two examples, one for regression and another for classification. For supervised computer vision, your dataset will need some labels. For classification, this label usually specifies what the image is a picture of. For regression, this "label" is some numeric quantity the image should produce, such as a count. We will look at two different means of providing this label.%
\index{classification}%
\index{computer vision}%
\index{dataset}%
\index{regression}%
\index{SOM}%
\par%
The first example will show how to handle regression with convolution neural networks. We will provide an image and expect the neural network to count items in that image. We will use a%
\index{convolution}%
\index{neural network}%
\index{regression}%
\href{https://www.kaggle.com/jeffheaton/count-the-paperclips}{ dataset }%
that I created that contains a random number of paperclips. The following code will download this dataset for you.%
\index{dataset}%
\index{random}%
\par%
\begin{tcolorbox}[size=title,title=Code,breakable]%
\begin{lstlisting}[language=Python, upquote=true]
import os

URL = "https://github.com/jeffheaton/data-mirror/releases/"
DOWNLOAD_SOURCE = URL+"download/v1/paperclips.zip"
DOWNLOAD_NAME = DOWNLOAD_SOURCE[DOWNLOAD_SOURCE.rfind('/')+1:]

if COLAB:
  PATH = "/content"
else:
  # I used this locally on my machine, you may need different
  PATH = "/Users/jeff/temp"

EXTRACT_TARGET = os.path.join(PATH,"clips")
SOURCE = os.path.join(EXTRACT_TARGET, "paperclips")\end{lstlisting}
\end{tcolorbox}%
Next, we download the images. This part depends on the origin of your images. The following code downloads images from a URL, where a ZIP file contains the images. The code unzips the ZIP file.%
\par%
\begin{tcolorbox}[size=title,title=Code,breakable]%
\begin{lstlisting}[language=Python, upquote=true]
!wget -O {os.path.join(PATH,DOWNLOAD_NAME)} {DOWNLOAD_SOURCE}
!mkdir -p {SOURCE}
!mkdir -p {TARGET}
!mkdir -p {EXTRACT_TARGET}
!unzip -o -j -d {SOURCE} {os.path.join(PATH, DOWNLOAD_NAME)} >/dev/null\end{lstlisting}
\end{tcolorbox}%
The labels are contained in a CSV file named%
\index{CSV}%
\textbf{ train.csv}%
for regression. This file has just two labels,%
\index{regression}%
\textbf{ id }%
and%
\textbf{ clip\_count}%
. The ID specifies the filename; for example, row id 1 corresponds to the file%
\textbf{ clips{-}1.jpg}%
. The following code loads the labels for the training set and creates a new column, named%
\index{training}%
\textbf{ filename}%
, that contains the filename of each image, based on the%
\textbf{ id }%
column.%
\par%
\begin{tcolorbox}[size=title,title=Code,breakable]%
\begin{lstlisting}[language=Python, upquote=true]
import pandas as pd

df = pd.read_csv(
    os.path.join(SOURCE,"train.csv"), 
    na_values=['NA', '?'])

df['filename']="clips-"+df["id"].astype(str)+".jpg"\end{lstlisting}
\end{tcolorbox}%
This results in the following dataframe.%
\par%
\begin{tcolorbox}[size=title,title=Code,breakable]%
\begin{lstlisting}[language=Python, upquote=true]
df\end{lstlisting}
\tcbsubtitle[before skip=\baselineskip]{Output}%
\begin{tabular}[hbt!]{l|l|l|l}%
\hline%
&id&clip\_count&filename\\%
\hline%
0&30001&11&clips{-}30001.jpg\\%
1&30002&2&clips{-}30002.jpg\\%
2&30003&26&clips{-}30003.jpg\\%
3&30004&41&clips{-}30004.jpg\\%
4&30005&49&clips{-}30005.jpg\\%
...&...&...&...\\%
19995&49996&35&clips{-}49996.jpg\\%
19996&49997&54&clips{-}49997.jpg\\%
19997&49998&72&clips{-}49998.jpg\\%
19998&49999&24&clips{-}49999.jpg\\%
19999&50000&35&clips{-}50000.jpg\\%
\hline%
\end{tabular}%
\vspace{2mm}%
\end{tcolorbox}%
Separate into a training and validation (for early stopping)%
\index{early stopping}%
\index{training}%
\index{validation}%
\par%
\begin{tcolorbox}[size=title,title=Code,breakable]%
\begin{lstlisting}[language=Python, upquote=true]
TRAIN_PCT = 0.9
TRAIN_CUT = int(len(df) * TRAIN_PCT)

df_train = df[0:TRAIN_CUT]
df_validate = df[TRAIN_CUT:]

print(f"Training size: {len(df_train)}")
print(f"Validate size: {len(df_validate)}")\end{lstlisting}
\tcbsubtitle[before skip=\baselineskip]{Output}%
\begin{lstlisting}[upquote=true]
Training size: 18000
Validate size: 2000
\end{lstlisting}
\end{tcolorbox}%
We are now ready to create two ImageDataGenerator objects. We currently use a generator, which creates additional training data by manipulating the source material. This technique can produce considerably stronger neural networks. The generator below flips the images both vertically and horizontally. Keras will train the neuron network both on the original images and the flipped images. This augmentation increases the size of the training data considerably. Module 6.4 goes deeper into the transformations you can perform. You can also specify a target size to resize the images automatically.%
\index{Keras}%
\index{neural network}%
\index{neuron}%
\index{training}%
\par%
The function%
\textbf{ flow\_from\_dataframe }%
loads the labels from a Pandas dataframe connected to our%
\textbf{ train.csv }%
file. When we demonstrate classification, we will use the%
\index{classification}%
\textbf{ flow\_from\_directory}%
; which loads the labels from the directory structure rather than a CSV.%
\index{CSV}%
\par%
\begin{tcolorbox}[size=title,title=Code,breakable]%
\begin{lstlisting}[language=Python, upquote=true]
import tensorflow as tf
import keras_preprocessing
from keras_preprocessing import image
from keras_preprocessing.image import ImageDataGenerator

training_datagen = ImageDataGenerator(
  rescale = 1./255,
  horizontal_flip=True,
  vertical_flip=True,
  fill_mode='nearest')

train_generator = training_datagen.flow_from_dataframe(
        dataframe=df_train,
        directory=SOURCE,
        x_col="filename",
        y_col="clip_count",
        target_size=(256, 256),
        batch_size=32,
        class_mode='other')

validation_datagen = ImageDataGenerator(rescale = 1./255)

val_generator = validation_datagen.flow_from_dataframe(
        dataframe=df_validate,
        directory=SOURCE,
        x_col="filename",
        y_col="clip_count",
        target_size=(256, 256),
        class_mode='other')\end{lstlisting}
\tcbsubtitle[before skip=\baselineskip]{Output}%
\begin{lstlisting}[upquote=true]
Found 18000 validated image filenames.
Found 2000 validated image filenames.
\end{lstlisting}
\end{tcolorbox}%
We can now train the neural network. The code to build and train the neural network is not that different than in the previous modules. We will use the Keras Sequential class to provide layers to the neural network. We now have several new layer types that we did not previously see.%
\index{Keras}%
\index{layer}%
\index{neural network}%
\par%
\begin{itemize}[noitemsep]%
\item%
\textbf{Conv2D }%
{-} The convolution layers.%
\index{convolution}%
\index{layer}%
\item%
\textbf{MaxPooling2D }%
{-} The max{-}pooling layers.%
\index{layer}%
\item%
\textbf{Flatten }%
{-} Flatten the 2D (and higher) tensors to allow a Dense layer to process.%
\index{dense layer}%
\index{layer}%
\index{ROC}%
\index{ROC}%
\item%
\textbf{Dense }%
{-} Dense layers, the same as demonstrated previously. Dense layers often form the final output layers of the neural network.%
\index{dense layer}%
\index{layer}%
\index{neural network}%
\index{output}%
\index{output layer}%
\end{itemize}%
The training code is very similar to previously. This code is for regression, so a final linear activation is used, along with mean\_squared\_error for the loss function. The generator provides both the%
\index{error}%
\index{linear}%
\index{regression}%
\index{training}%
\textit{ x }%
and%
\textit{ y }%
matrixes we previously supplied.%
\index{matrix}%
\par%
\begin{tcolorbox}[size=title,title=Code,breakable]%
\begin{lstlisting}[language=Python, upquote=true]
from tensorflow.keras.callbacks import EarlyStopping
import time

model = tf.keras.models.Sequential([
    # Note the input shape is the desired size of the image 150x150 
    # with 3 bytes color.
    # This is the first convolution
    tf.keras.layers.Conv2D(64, (3,3), activation='relu', 
        input_shape=(256, 256, 3)),
    tf.keras.layers.MaxPooling2D(2, 2),
    # The second convolution
    tf.keras.layers.Conv2D(64, (3,3), activation='relu'),
    tf.keras.layers.MaxPooling2D(2,2),
    tf.keras.layers.Flatten(),
    # 512 neuron hidden layer
    tf.keras.layers.Dense(512, activation='relu'),
    tf.keras.layers.Dense(1, activation='linear')
])


model.summary()
epoch_steps = 250 # needed for 2.2
validation_steps = len(df_validate)
model.compile(loss = 'mean_squared_error', optimizer='adam')
monitor = EarlyStopping(monitor='val_loss', min_delta=1e-3, 
        patience=5, verbose=1, mode='auto',
        restore_best_weights=True)

start_time = time.time()
history = model.fit(train_generator,  
  verbose = 1, 
  validation_data=val_generator, callbacks=[monitor], epochs=25)

elapsed_time = time.time() - start_time
print("Elapsed time: {}".format(hms_string(elapsed_time)))\end{lstlisting}
\tcbsubtitle[before skip=\baselineskip]{Output}%
\begin{lstlisting}[upquote=true]
Model: "sequential"
_________________________________________________________________
 Layer (type)                Output Shape              Param #
=================================================================
 conv2d (Conv2D)             (None, 254, 254, 64)      1792
 max_pooling2d (MaxPooling2D  (None, 127, 127, 64)     0
 )
 conv2d_1 (Conv2D)           (None, 125, 125, 64)      36928
 max_pooling2d_1 (MaxPooling  (None, 62, 62, 64)       0
 2D)
 flatten (Flatten)           (None, 246016)            0
 dense (Dense)               (None, 512)               125960704
 dense_1 (Dense)             (None, 1)                 513
=================================================================
Total params: 125,999,937

...

3.2399 - val_loss: 4.0449
Epoch 25/25
563/563 [==============================] - 53s 94ms/step - loss:
3.2823 - val_loss: 4.4899
Elapsed time: 0:22:22.78
\end{lstlisting}
\end{tcolorbox}%
This code will run very slowly if you do not use a GPU. The above code takes approximately 13 minutes with a GPU.%
\index{GPU}%
\index{GPU}%
\par

\subsection{Score Regression Image Data}%
\label{subsec:ScoreRegressionImageData}%
Scoring/predicting from a generator is a bit different than training. We do not want augmented images, and we do not wish to have the dataset shuffled. For scoring, we want a prediction for each input. We construct the generator as follows:%
\index{dataset}%
\index{input}%
\index{predict}%
\index{training}%
\par%
\begin{itemize}[noitemsep]%
\item%
shuffle=False%
\item%
batch\_size=1%
\item%
class\_mode=None%
\end{itemize}%
We use a%
\textbf{ batch\_size }%
of 1 to guarantee that we do not run out of GPU memory if our prediction set is large. You can increase this value for better performance. The%
\index{GPU}%
\index{GPU}%
\index{predict}%
\textbf{ class\_mode }%
is None because there is no%
\textit{ y}%
, or label. After all, we are predicting.%
\index{predict}%
\par%
\begin{tcolorbox}[size=title,title=Code,breakable]%
\begin{lstlisting}[language=Python, upquote=true]
df_test = pd.read_csv(
    os.path.join(SOURCE,"test.csv"), 
    na_values=['NA', '?'])

df_test['filename']="clips-"+df_test["id"].astype(str)+".jpg"

test_datagen = ImageDataGenerator(rescale = 1./255)

test_generator = validation_datagen.flow_from_dataframe(
        dataframe=df_test,
        directory=SOURCE,
        x_col="filename",
        batch_size=1,
        shuffle=False,
        target_size=(256, 256),
        class_mode=None)\end{lstlisting}
\tcbsubtitle[before skip=\baselineskip]{Output}%
\begin{lstlisting}[upquote=true]
Found 5000 validated image filenames.
\end{lstlisting}
\end{tcolorbox}%
We need to reset the generator to ensure we are always at the beginning.%
\par%
\begin{tcolorbox}[size=title,title=Code,breakable]%
\begin{lstlisting}[language=Python, upquote=true]
test_generator.reset()
pred = model.predict(test_generator,steps=len(df_test))\end{lstlisting}
\end{tcolorbox}%
We can now generate a CSV file to hold the predictions.%
\index{CSV}%
\index{predict}%
\par%
\begin{tcolorbox}[size=title,title=Code,breakable]%
\begin{lstlisting}[language=Python, upquote=true]
df_submit = pd.DataFrame({'id':df_test['id'],'clip_count':pred.flatten()})
df_submit.to_csv(os.path.join(PATH,"submit.csv"),index=False)\end{lstlisting}
\end{tcolorbox}

\subsection{Classification Neural Networks}%
\label{subsec:ClassificationNeuralNetworks}%
Just like earlier in this module, we will load data. However, this time we will use a dataset of images of three different types of the iris flower. This zip file contains three different directories that specify each image's label. The directories are named the same as the labels:%
\index{dataset}%
\index{iris}%
\par%
\begin{itemize}[noitemsep]%
\item%
iris{-}setosa%
\index{iris}%
\item%
iris{-}versicolour%
\index{iris}%
\item%
iris{-}virginica%
\index{iris}%
\end{itemize}%
\begin{tcolorbox}[size=title,title=Code,breakable]%
\begin{lstlisting}[language=Python, upquote=true]
import os

URL = "https://github.com/jeffheaton/data-mirror/releases"
DOWNLOAD_SOURCE = URL+"/download/v1/iris-image.zip"
DOWNLOAD_NAME = DOWNLOAD_SOURCE[DOWNLOAD_SOURCE.rfind('/')+1:]

if COLAB:
  PATH = "/content"
  EXTRACT_TARGET = os.path.join(PATH,"iris")
  SOURCE = EXTRACT_TARGET # In this case its the same, no subfolder
else:
  # I used this locally on my machine, you may need different
  PATH = "/Users/jeff/temp"
  EXTRACT_TARGET = os.path.join(PATH,"iris")
  SOURCE = EXTRACT_TARGET # In this case its the same, no subfolder\end{lstlisting}
\end{tcolorbox}%
Just as before, we unzip the images.%
\par%
\begin{tcolorbox}[size=title,title=Code,breakable]%
\begin{lstlisting}[language=Python, upquote=true]
!wget -O {os.path.join(PATH,DOWNLOAD_NAME)} {DOWNLOAD_SOURCE}
!mkdir -p {SOURCE}
!mkdir -p {TARGET}
!mkdir -p {EXTRACT_TARGET}
!unzip -o -d {EXTRACT_TARGET} {os.path.join(PATH, DOWNLOAD_NAME)} >/dev/null\end{lstlisting}
\end{tcolorbox}%
You can see these folders with the following command.%
\par%
\begin{tcolorbox}[size=title,title=Code,breakable]%
\begin{lstlisting}[language=Python, upquote=true]
!ls /content/iris\end{lstlisting}
\tcbsubtitle[before skip=\baselineskip]{Output}%
\begin{lstlisting}[upquote=true]
iris-setosa  iris-versicolour  iris-virginica
\end{lstlisting}
\end{tcolorbox}%
We set up the generator, similar to before.  This time we use flow\_from\_directory to get the labels from the directory structure.%
\par%
\begin{tcolorbox}[size=title,title=Code,breakable]%
\begin{lstlisting}[language=Python, upquote=true]
import tensorflow as tf
import keras_preprocessing
from keras_preprocessing import image
from keras_preprocessing.image import ImageDataGenerator

training_datagen = ImageDataGenerator(
  rescale = 1./255,
  horizontal_flip=True,
  vertical_flip=True,
  width_shift_range=[-200,200],
  rotation_range=360,

  fill_mode='nearest')

train_generator = training_datagen.flow_from_directory(
    directory=SOURCE, target_size=(256, 256), 
    class_mode='categorical', batch_size=32, shuffle=True)

validation_datagen = ImageDataGenerator(rescale = 1./255)

validation_generator = validation_datagen.flow_from_directory(
    directory=SOURCE, target_size=(256, 256), 
    class_mode='categorical', batch_size=32, shuffle=True)\end{lstlisting}
\tcbsubtitle[before skip=\baselineskip]{Output}%
\begin{lstlisting}[upquote=true]
Found 421 images belonging to 3 classes.
Found 421 images belonging to 3 classes.
\end{lstlisting}
\end{tcolorbox}%
Training the neural network with classification is similar to regression.%
\index{classification}%
\index{neural network}%
\index{regression}%
\index{training}%
\par%
\begin{tcolorbox}[size=title,title=Code,breakable]%
\begin{lstlisting}[language=Python, upquote=true]
from tensorflow.keras.callbacks import EarlyStopping

class_count = len(train_generator.class_indices)

model = tf.keras.models.Sequential([
    # Note the input shape is the desired size of the image 
    # 300x300 with 3 bytes color
    # This is the first convolution
    tf.keras.layers.Conv2D(16, (3,3), activation='relu', 
        input_shape=(256, 256, 3)),
    tf.keras.layers.MaxPooling2D(2, 2),
    # The second convolution
    tf.keras.layers.Conv2D(32, (3,3), activation='relu'),
    tf.keras.layers.Dropout(0.5),
    tf.keras.layers.MaxPooling2D(2,2),
    # The third convolution
    tf.keras.layers.Conv2D(64, (3,3), activation='relu'),
    tf.keras.layers.Dropout(0.5),
    tf.keras.layers.MaxPooling2D(2,2),
    # The fourth convolution
    tf.keras.layers.Conv2D(64, (3,3), activation='relu'),
    tf.keras.layers.MaxPooling2D(2,2),
    # The fifth convolution
    tf.keras.layers.Conv2D(64, (3,3), activation='relu'),
    tf.keras.layers.MaxPooling2D(2,2),
    # Flatten the results to feed into a DNN
    
    tf.keras.layers.Flatten(),
    tf.keras.layers.Dropout(0.5),
    # 512 neuron hidden layer
    tf.keras.layers.Dense(512, activation='relu'),
    # Only 1 output neuron. It will contain a value from 0-1 
    tf.keras.layers.Dense(class_count, activation='softmax')
])

model.summary()

model.compile(loss = 'categorical_crossentropy', optimizer='adam')

model.fit(train_generator, epochs=50, steps_per_epoch=10, 
                    verbose = 1)\end{lstlisting}
\tcbsubtitle[before skip=\baselineskip]{Output}%
\begin{lstlisting}[upquote=true]
Model: "sequential_1"
_________________________________________________________________
 Layer (type)                Output Shape              Param #
=================================================================
 conv2d_2 (Conv2D)           (None, 254, 254, 16)      448
 max_pooling2d_2 (MaxPooling  (None, 127, 127, 16)     0
 2D)
 conv2d_3 (Conv2D)           (None, 125, 125, 32)      4640
 dropout (Dropout)           (None, 125, 125, 32)      0
 max_pooling2d_3 (MaxPooling  (None, 62, 62, 32)       0
 2D)
 conv2d_4 (Conv2D)           (None, 60, 60, 64)        18496
 dropout_1 (Dropout)         (None, 60, 60, 64)        0
 max_pooling2d_4 (MaxPooling  (None, 30, 30, 64)       0
 2D)

...

_________________________________________________________________
...
10/10 [==============================] - 5s 458ms/step - loss: 0.7957
Epoch 50/50
10/10 [==============================] - 5s 501ms/step - loss: 0.8670
\end{lstlisting}
\end{tcolorbox}%
The iris image dataset is not easy to predict; it turns out that a tabular dataset of measurements is more manageable.  However, we can achieve a 63\%.%
\index{dataset}%
\index{iris}%
\index{predict}%
\index{tabular data}%
\par%
\begin{tcolorbox}[size=title,title=Code,breakable]%
\begin{lstlisting}[language=Python, upquote=true]
from sklearn.metrics import accuracy_score
import numpy as np

validation_generator.reset()
pred = model.predict(validation_generator)

predict_classes = np.argmax(pred,axis=1)
expected_classes = validation_generator.classes

correct = accuracy_score(expected_classes,predict_classes)
print(f"Accuracy: {correct}")\end{lstlisting}
\tcbsubtitle[before skip=\baselineskip]{Output}%
\begin{lstlisting}[upquote=true]
Accuracy: 0.6389548693586699
\end{lstlisting}
\end{tcolorbox}

\subsection{Other Resources}%
\label{subsec:OtherResources}%
\begin{itemize}[noitemsep]%
\item%
\href{http://image-net.org/challenges/LSVRC/2014/index}{Imagenet:Large Scale Visual Recognition Challenge 2014}%
\item%
\href{http://cs.stanford.edu/people/karpathy/}{Andrej Karpathy }%
{-} PhD student/instructor at Stanford.%
\item%
\href{http://cs231n.stanford.edu/}{CS231n Convolutional Neural Networks for Visual Recognition }%
{-} Stanford course on computer vision/CNN's.%
\index{CNN}%
\index{computer vision}%
\item%
\href{http://cs231n.github.io/}{CS231n {-} GitHub}%
\item%
\href{http://cs.stanford.edu/people/karpathy/convnetjs/}{ConvNetJS }%
{-} JavaScript library for deep learning.%
\index{Java}%
\index{JavaScript}%
\index{learning}%
\end{itemize}

\section{Part 6.3: Transfer Learning for Computer Vision}%
\label{sec:Part6.3TransferLearningforComputerVision}%
Many advanced prebuilt neural networks are available for computer vision, and Keras provides direct access to many networks. Transfer learning is the technique where you use these prebuilt neural networks. Module 9 takes a deeper look at transfer learning.%
\index{computer vision}%
\index{Keras}%
\index{learning}%
\index{neural network}%
\index{transfer learning}%
\par%
There are several different levels of transfer learning.%
\index{learning}%
\index{transfer learning}%
\par%
\begin{itemize}[noitemsep]%
\item%
Use a prebuilt neural network in its entirety%
\index{neural network}%
\item%
Use a prebuilt neural network's structure%
\index{neural network}%
\item%
Use a prebuilt neural network's weights%
\index{neural network}%
\end{itemize}%
We will begin by using the MobileNet prebuilt neural network in its entirety. MobileNet will be loaded and allowed to classify simple images. We can already classify 1,000 images through this technique without ever having trained the network.%
\index{neural network}%
\par%
\begin{tcolorbox}[size=title,title=Code,breakable]%
\begin{lstlisting}[language=Python, upquote=true]
import pandas as pd
import numpy as np
import os
import tensorflow.keras
import matplotlib.pyplot as plt
from tensorflow.keras.layers import Dense,GlobalAveragePooling2D
from tensorflow.keras.applications import MobileNet
from tensorflow.keras.preprocessing import image
from tensorflow.keras.applications.mobilenet import preprocess_input
from tensorflow.keras.preprocessing.image import ImageDataGenerator
from tensorflow.keras.models import Model
from tensorflow.keras.optimizers import Adam\end{lstlisting}
\end{tcolorbox}%
We begin by downloading weights for a MobileNet trained for the imagenet dataset, which will take some time to download the first time you train the network.%
\index{dataset}%
\index{SOM}%
\par%
\begin{tcolorbox}[size=title,title=Code,breakable]%
\begin{lstlisting}[language=Python, upquote=true]
model = MobileNet(weights='imagenet',include_top=True)\end{lstlisting}
\end{tcolorbox}%
The loaded network is a Keras neural network. However, this is a neural network that a third party engineered on advanced hardware. Merely looking at the structure of an advanced state{-}of{-}the{-}art neural network can be educational.%
\index{Keras}%
\index{neural network}%
\par%
\begin{tcolorbox}[size=title,title=Code,breakable]%
\begin{lstlisting}[language=Python, upquote=true]
model.summary()\end{lstlisting}
\tcbsubtitle[before skip=\baselineskip]{Output}%
\begin{lstlisting}[upquote=true]
Model: "mobilenet_1.00_224"
_________________________________________________________________
 Layer (type)                Output Shape              Param #
=================================================================
 input_1 (InputLayer)        [(None, 224, 224, 3)]     0
 conv1 (Conv2D)              (None, 112, 112, 32)      864
 conv1_bn (BatchNormalizatio  (None, 112, 112, 32)     128
 n)
 conv1_relu (ReLU)           (None, 112, 112, 32)      0
 conv_dw_1 (DepthwiseConv2D)  (None, 112, 112, 32)     288
 conv_dw_1_bn (BatchNormaliz  (None, 112, 112, 32)     128
 ation)
 conv_dw_1_relu (ReLU)       (None, 112, 112, 32)      0
 conv_pw_1 (Conv2D)          (None, 112, 112, 64)      2048
 conv_pw_1_bn (BatchNormaliz  (None, 112, 112, 64)     256

...

=================================================================
Total params: 4,253,864
Trainable params: 4,231,976
Non-trainable params: 21,888
_________________________________________________________________
\end{lstlisting}
\end{tcolorbox}%
Several clues to neural network architecture become evident when examining the above structure.%
\index{architecture}%
\index{network architecture}%
\index{neural network}%
\par%
We will now use the MobileNet to classify several image URLs below.  You can add additional URLs of your own to see how well the MobileNet can classify.%
\par%
\begin{tcolorbox}[size=title,title=Code,breakable]%
\begin{lstlisting}[language=Python, upquote=true]
%matplotlib inline
from PIL import Image, ImageFile
from matplotlib.pyplot import imshow
import requests
import numpy as np
from io import BytesIO
from IPython.display import display, HTML
from tensorflow.keras.applications.mobilenet import decode_predictions

IMAGE_WIDTH = 224
IMAGE_HEIGHT = 224
IMAGE_CHANNELS = 3

ROOT = "https://data.heatonresearch.com/data/t81-558/images/"

def make_square(img):
    cols,rows = img.size
    
    if rows>cols:
        pad = (rows-cols)/2
        img = img.crop((pad,0,cols,cols))
    else:
        pad = (cols-rows)/2
        img = img.crop((0,pad,rows,rows))
    
    return img
        

def classify_image(url):
  x = []
  ImageFile.LOAD_TRUNCATED_IMAGES = False
  response = requests.get(url)
  img = Image.open(BytesIO(response.content))
  img.load()
  img = img.resize((IMAGE_WIDTH,IMAGE_HEIGHT),Image.ANTIALIAS)

  x = image.img_to_array(img)
  x = np.expand_dims(x, axis=0)
  x = preprocess_input(x)
  x = x[:,:,:,:3] # maybe an alpha channel
  pred = model.predict(x)

  display(img)
  print(np.argmax(pred,axis=1))

  lst = decode_predictions(pred, top=5)
  for itm in lst[0]:
      print(itm)\end{lstlisting}
\end{tcolorbox}%
We can now classify an example image.  You can specify the URL of any image you wish to classify.%
\par%
\begin{tcolorbox}[size=title,title=Code,breakable]%
\begin{lstlisting}[language=Python, upquote=true]
classify_image(ROOT+"soccer_ball.jpg")\end{lstlisting}
\tcbsubtitle[before skip=\baselineskip]{Output}%
\includegraphics[width=3in]{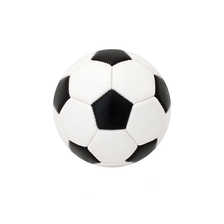}%
\begin{lstlisting}[upquote=true]
[805]
Downloading data from https://storage.googleapis.com/download.tensorfl
ow.org/data/imagenet_class_index.json
40960/35363 [==================================] - 0s 0us/step
49152/35363 [=========================================] - 0s 0us/step
('n04254680', 'soccer_ball', 0.9999938)
('n03530642', 'honeycomb', 3.862412e-06)
('n03255030', 'dumbbell', 4.442458e-07)
('n02782093', 'balloon', 3.7038987e-07)
('n04548280', 'wall_clock', 3.143911e-07)
\end{lstlisting}
\end{tcolorbox}%
\begin{tcolorbox}[size=title,title=Code,breakable]%
\begin{lstlisting}[language=Python, upquote=true]
classify_image(ROOT+"race_truck.jpg")\end{lstlisting}
\tcbsubtitle[before skip=\baselineskip]{Output}%
\includegraphics[width=3in]{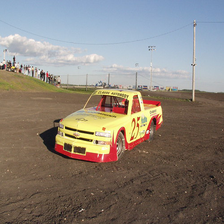}%
\end{tcolorbox}%
Overall, the neural network is doing quite well.%
\index{neural network}%
\par%
For many applications, MobileNet might be entirely acceptable as an image classifier. However, if you need to classify very specialized images, not in the 1,000 image types supported by imagenet, it is necessary to use transfer learning.%
\index{learning}%
\index{transfer learning}%
\par%
\subsection{Using the Structure of ResNet}%
\label{subsec:UsingtheStructureofResNet}%
We will train a neural network to count the number of paper clips in images. We will make use of the structure of the ResNet neural network. There are several significant changes that we will make to ResNet to apply to this task. First, ResNet is a classifier; we wish to perform a regression to count. Secondly, we want to change the image resolution that ResNet uses. We will not use the weights from ResNet; changing this resolution invalidates the current weights. Thus, it will be necessary to retrain the network.%
\index{neural network}%
\index{regression}%
\index{ResNet}%
\par%
\begin{tcolorbox}[size=title,title=Code,breakable]%
\begin{lstlisting}[language=Python, upquote=true]
import os
URL = "https://github.com/jeffheaton/data-mirror/"
DOWNLOAD_SOURCE = URL+"releases/download/v1/paperclips.zip"
DOWNLOAD_NAME = DOWNLOAD_SOURCE[DOWNLOAD_SOURCE.rfind('/')+1:]

if COLAB:
  PATH = "/content"
else:
  # I used this locally on my machine, you may need different
  PATH = "/Users/jeff/temp"

EXTRACT_TARGET = os.path.join(PATH,"clips")
SOURCE = os.path.join(EXTRACT_TARGET, "paperclips")\end{lstlisting}
\tcbsubtitle[before skip=\baselineskip]{Output}%
\begin{lstlisting}[upquote=true]
[751]
('n04037443', 'racer', 0.7131951)
('n03100240', 'convertible', 0.100896776)
('n04285008', 'sports_car', 0.0770768)
('n03930630', 'pickup', 0.02635305)
('n02704792', 'amphibian', 0.011636169)
\end{lstlisting}
\end{tcolorbox}%
Next, we download the images. This part depends on the origin of your images. The following code downloads images from a URL, where a ZIP file contains the images. The code unzips the ZIP file.%
\par%
\begin{tcolorbox}[size=title,title=Code,breakable]%
\begin{lstlisting}[language=Python, upquote=true]
!wget -O {os.path.join(PATH,DOWNLOAD_NAME)} {DOWNLOAD_SOURCE}
!mkdir -p {SOURCE}
!mkdir -p {TARGET}
!mkdir -p {EXTRACT_TARGET}
!unzip -o -j -d {SOURCE} {os.path.join(PATH, DOWNLOAD_NAME)} >/dev/null\end{lstlisting}
\end{tcolorbox}%
The labels are contained in a CSV file named%
\index{CSV}%
\textbf{ train.csv }%
for the regression. This file has just two labels,%
\index{regression}%
\textbf{ id }%
and%
\textbf{ clip\_count}%
. The ID specifies the filename; for example, row id 1 corresponds to the file%
\textbf{ clips{-}1.jpg}%
. The following code loads the labels for the training set and creates a new column, named%
\index{training}%
\textbf{ filename}%
, that contains the filename of each image, based on the%
\textbf{ id }%
column.%
\par%
\begin{tcolorbox}[size=title,title=Code,breakable]%
\begin{lstlisting}[language=Python, upquote=true]
df_train = pd.read_csv(os.path.join(SOURCE, "train.csv"))
df_train['filename'] = "clips-" + df_train.id.astype(str) + ".jpg"\end{lstlisting}
\end{tcolorbox}%
We want to use early stopping. To do this, we need a validation set. We will break the data into 80 percent test data and 20 validation. Do not confuse this validation data with the test set provided by Kaggle. This validation set is unique to your program and is for early stopping.%
\index{early stopping}%
\index{Kaggle}%
\index{validation}%
\par%
\begin{tcolorbox}[size=title,title=Code,breakable]%
\begin{lstlisting}[language=Python, upquote=true]
TRAIN_PCT = 0.9
TRAIN_CUT = int(len(df_train) * TRAIN_PCT)

df_train_cut = df_train[0:TRAIN_CUT]
df_validate_cut = df_train[TRAIN_CUT:]

print(f"Training size: {len(df_train_cut)}")
print(f"Validate size: {len(df_validate_cut)}")\end{lstlisting}
\tcbsubtitle[before skip=\baselineskip]{Output}%
\begin{lstlisting}[upquote=true]
Training size: 18000
Validate size: 2000
\end{lstlisting}
\end{tcolorbox}%
Next, we create the generators that will provide the images to the neural network during training. We normalize the images so that the RGB colors between 0{-}255 become ratios between 0 and 1. We also use the%
\index{neural network}%
\index{training}%
\textbf{ flow\_from\_dataframe }%
generator to connect the Pandas dataframe to the actual image files. We see here a straightforward implementation; you might also wish to use some of the image transformations provided by the data generator.%
\index{SOM}%
\par%
The%
\textbf{ HEIGHT }%
and%
\textbf{ WIDTH }%
constants specify the dimensions to which the image will be scaled (or expanded). It is probably not a good idea to expand the images.%
\par%
\begin{tcolorbox}[size=title,title=Code,breakable]%
\begin{lstlisting}[language=Python, upquote=true]
import tensorflow as tf
import keras_preprocessing
from keras_preprocessing import image
from keras_preprocessing.image import ImageDataGenerator

WIDTH = 256
HEIGHT = 256

training_datagen = ImageDataGenerator(
  rescale = 1./255,
  horizontal_flip=True,
  #vertical_flip=True,
  fill_mode='nearest')

train_generator = training_datagen.flow_from_dataframe(
        dataframe=df_train_cut,
        directory=SOURCE,
        x_col="filename",
        y_col="clip_count",
        target_size=(HEIGHT, WIDTH),
        # Keeping the training batch size small 
        # USUALLY increases performance
        batch_size=32, 
        class_mode='raw')

validation_datagen = ImageDataGenerator(rescale = 1./255)

val_generator = validation_datagen.flow_from_dataframe(
        dataframe=df_validate_cut,
        directory=SOURCE,
        x_col="filename",
        y_col="clip_count",
        target_size=(HEIGHT, WIDTH),
        # Make the validation batch size as large as you 
        # have memory for
        batch_size=256, 
        class_mode='raw')\end{lstlisting}
\tcbsubtitle[before skip=\baselineskip]{Output}%
\begin{lstlisting}[upquote=true]
Found 18000 validated image filenames.
Found 2000 validated image filenames.
\end{lstlisting}
\end{tcolorbox}%
We will now use a ResNet neural network as a basis for our neural network. We will redefine both the input shape and output of the ResNet model, so we will not transfer the weights. Since we redefine the input, the weights are of minimal value. We begin by loading, from Keras, the ResNet50 network. We specify%
\index{input}%
\index{Keras}%
\index{model}%
\index{neural network}%
\index{output}%
\index{ResNet}%
\textbf{ include\_top }%
as False because we will change the input resolution. We also specify%
\index{input}%
\textbf{ weights }%
as false because we must retrain the network after changing the top input layers.%
\index{input}%
\index{input layer}%
\index{layer}%
\par%
\begin{tcolorbox}[size=title,title=Code,breakable]%
\begin{lstlisting}[language=Python, upquote=true]
from tensorflow.keras.applications.resnet50 import ResNet50
from tensorflow.keras.layers import Input

input_tensor = Input(shape=(HEIGHT, WIDTH, 3))

base_model = ResNet50(
    include_top=False, weights=None, input_tensor=input_tensor,
    input_shape=None)\end{lstlisting}
\end{tcolorbox}%
Now we must add a few layers to the end of the neural network so that it becomes a regression model.%
\index{layer}%
\index{model}%
\index{neural network}%
\index{regression}%
\par%
\begin{tcolorbox}[size=title,title=Code,breakable]%
\begin{lstlisting}[language=Python, upquote=true]
from tensorflow.keras.layers import Dense, GlobalAveragePooling2D
from tensorflow.keras.models import Model

x=base_model.output
x=GlobalAveragePooling2D()(x)
x=Dense(1024,activation='relu')(x) 
x=Dense(1024,activation='relu')(x) 
model=Model(inputs=base_model.input,outputs=Dense(1)(x))\end{lstlisting}
\end{tcolorbox}%
We train like before; the only difference is that we do not define the entire neural network here.%
\index{neural network}%
\par%
\begin{tcolorbox}[size=title,title=Code,breakable]%
\begin{lstlisting}[language=Python, upquote=true]
from tensorflow.keras.callbacks import EarlyStopping
from tensorflow.keras.metrics import RootMeanSquaredError

# Important, calculate a valid step size for the validation dataset
STEP_SIZE_VALID=val_generator.n//val_generator.batch_size

model.compile(loss = 'mean_squared_error', optimizer='adam', 
              metrics=[RootMeanSquaredError(name="rmse")])
monitor = EarlyStopping(monitor='val_loss', min_delta=1e-3, 
        patience=50, verbose=1, mode='auto',
        restore_best_weights=True)

history = model.fit(train_generator, epochs=100, steps_per_epoch=250, 
                    validation_data = val_generator, callbacks=[monitor],
                    verbose = 1, validation_steps=STEP_SIZE_VALID)\end{lstlisting}
\tcbsubtitle[before skip=\baselineskip]{Output}%
\begin{lstlisting}[upquote=true]
...
250/250 [==============================] - 61s 243ms/step - loss:
1.9211 - rmse: 1.3860 - val_loss: 17.0489 - val_rmse: 4.1290
Epoch 72/100
250/250 [==============================] - 61s 243ms/step - loss:
2.3726 - rmse: 1.5403 - val_loss: 167.8536 - val_rmse: 12.9558
\end{lstlisting}
\end{tcolorbox}

\section{Part 6.4: Inside Augmentation}%
\label{sec:Part6.4InsideAugmentation}%
The%
\href{https://www.tensorflow.org/api_docs/python/tf/keras/preprocessing/image/ImageDataGenerator}{ ImageDataGenerator }%
class provides many options for image augmentation.  Deciding which augmentations to use can impact the effectiveness of your model. This part will visualize some of these augmentations that you might use to train your neural network. We begin by loading a sample image to augment.%
\index{model}%
\index{neural network}%
\index{SOM}%
\par%
\begin{tcolorbox}[size=title,title=Code,breakable]%
\begin{lstlisting}[language=Python, upquote=true]
import urllib.request
import shutil
from IPython.display import Image

URL =  "https://github.com/jeffheaton/t81_558_deep_learning/" +\
  "blob/master/photos/landscape.jpg?raw=true"
LOCAL_IMG_FILE = "/content/landscape.jpg"

with urllib.request.urlopen(URL) as response, \
  open(LOCAL_IMG_FILE, 'wb') as out_file:
    shutil.copyfileobj(response, out_file)

Image(filename=LOCAL_IMG_FILE)\end{lstlisting}
\tcbsubtitle[before skip=\baselineskip]{Output}%
\includegraphics[width=4in]{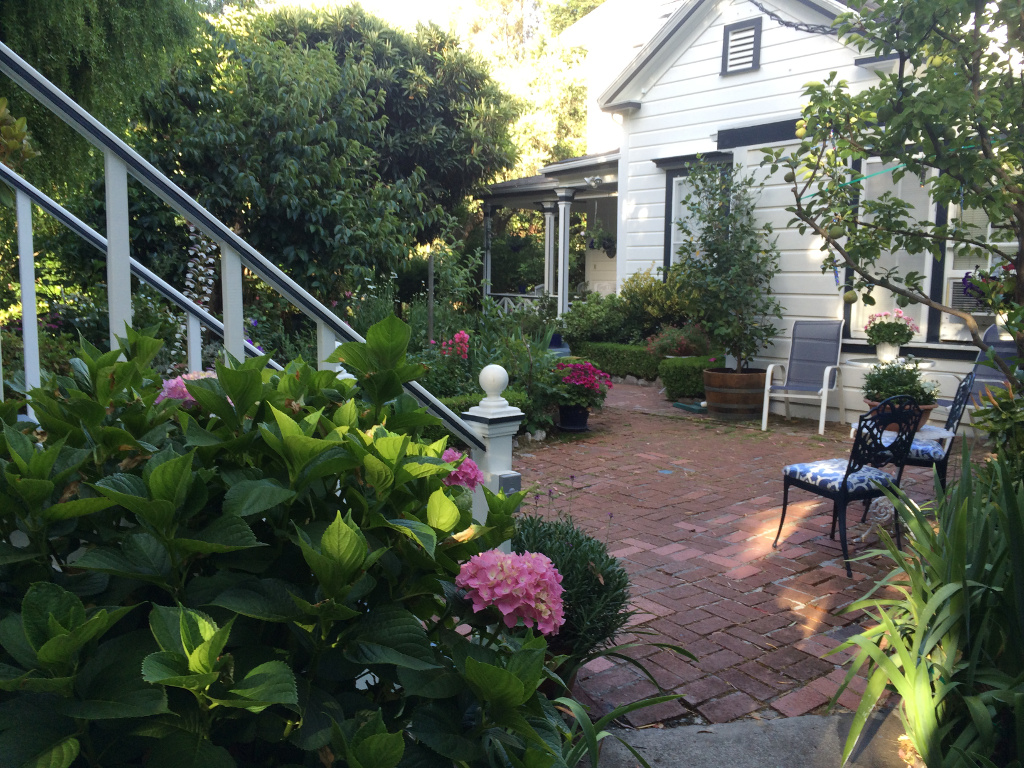}%
\end{tcolorbox}%
Next, we introduce a simple utility function to visualize four images sampled from any generator.%
\index{sampled}%
\par%
\begin{tcolorbox}[size=title,title=Code,breakable]%
\begin{lstlisting}[language=Python, upquote=true]
from numpy import expand_dims
from keras.preprocessing.image import load_img
from keras.preprocessing.image import img_to_array
from keras.preprocessing.image import ImageDataGenerator
from matplotlib import pyplot
import matplotlib.pyplot as plt
import numpy as np
import matplotlib

def visualize_generator(img_file, gen):
	# Load the requested image
  img = load_img(img_file)
  data = img_to_array(img)
  samples = expand_dims(data, 0)

	# Generat augumentations from the generator
  it = gen.flow(samples, batch_size=1)
  images = []
  for i in range(4):
    batch = it.next()
    image = batch[0].astype('uint8')
    images.append(image)

  images = np.array(images)

	# Create a grid of 4 images from the generator
  index, height, width, channels = images.shape
  nrows = index//2
    
  grid = (images.reshape(nrows, 2, height, width, channels)
            .swapaxes(1,2)
            .reshape(height*nrows, width*2, 3))
  
  fig = plt.figure(figsize=(15., 15.))
  plt.axis('off')
  plt.imshow(grid)\end{lstlisting}
\end{tcolorbox}%
We begin by flipping the image. Some images may not make sense to flip, such as this landscape.  However, if you expect "noise" in your data where some images may be flipped, then this augmentation may be useful, even if it violates physical reality.%
\index{SOM}%
\par%
\begin{tcolorbox}[size=title,title=Code,breakable]%
\begin{lstlisting}[language=Python, upquote=true]
visualize_generator(
  LOCAL_IMG_FILE,
  ImageDataGenerator(horizontal_flip=True, vertical_flip=True))\end{lstlisting}
\tcbsubtitle[before skip=\baselineskip]{Output}%
\includegraphics[width=4in]{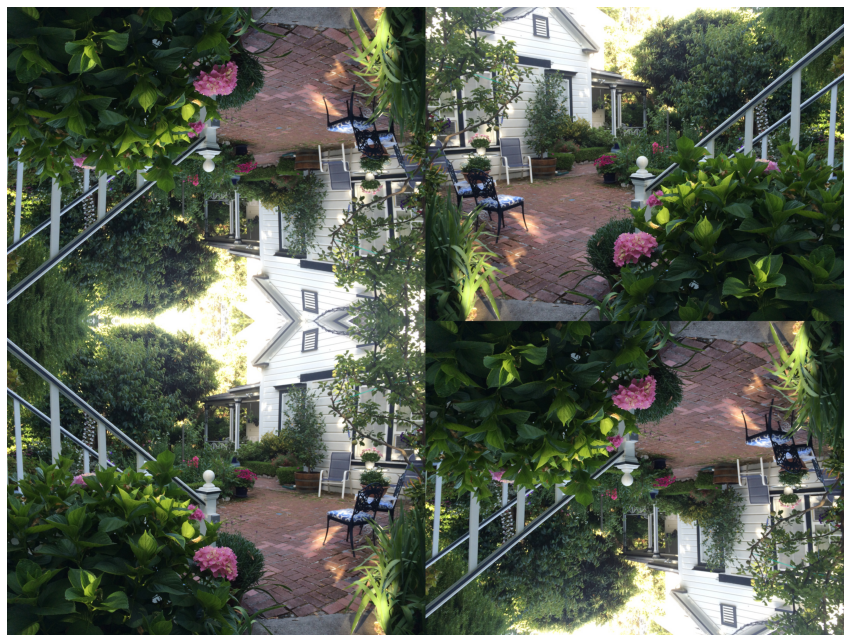}%
\end{tcolorbox}%
Next, we will try moving the image. Notice how part of the image is missing? There are various ways to fill in the missing data, as controlled by%
\textbf{ fill\_mode}%
. In this case, we simply use the nearest pixel to fill. It is also possible to rotate images.%
\par%
\begin{tcolorbox}[size=title,title=Code,breakable]%
\begin{lstlisting}[language=Python, upquote=true]
visualize_generator(
    LOCAL_IMG_FILE,
    ImageDataGenerator(width_shift_range=[-200,200], 
        fill_mode='nearest'))\end{lstlisting}
\tcbsubtitle[before skip=\baselineskip]{Output}%
\includegraphics[width=4in]{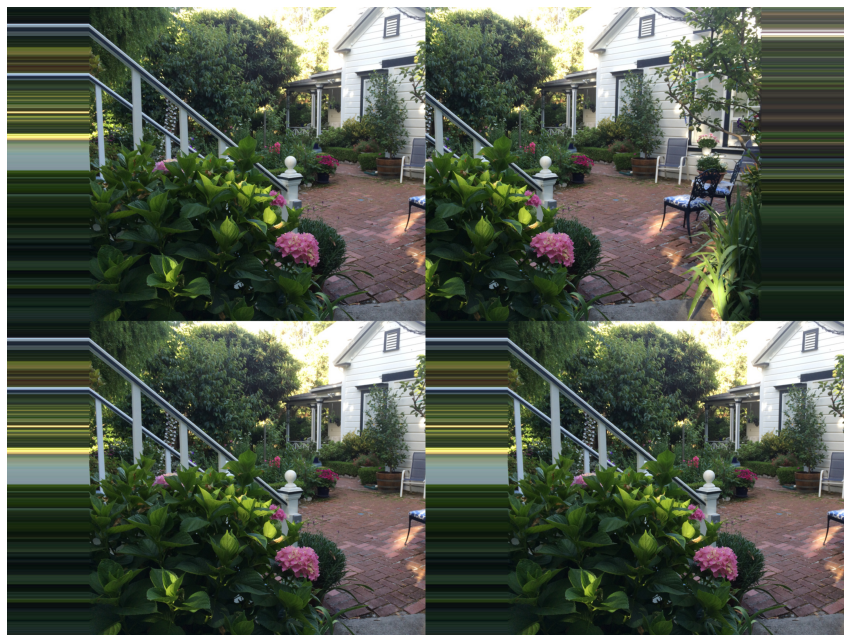}%
\end{tcolorbox}%
We can also adjust brightness.%
\par%
\begin{tcolorbox}[size=title,title=Code,breakable]%
\begin{lstlisting}[language=Python, upquote=true]
visualize_generator(
  LOCAL_IMG_FILE,
  ImageDataGenerator(brightness_range=[0,1]))

# brightness_range=None, shear_range=0.0\end{lstlisting}
\tcbsubtitle[before skip=\baselineskip]{Output}%
\includegraphics[width=4in]{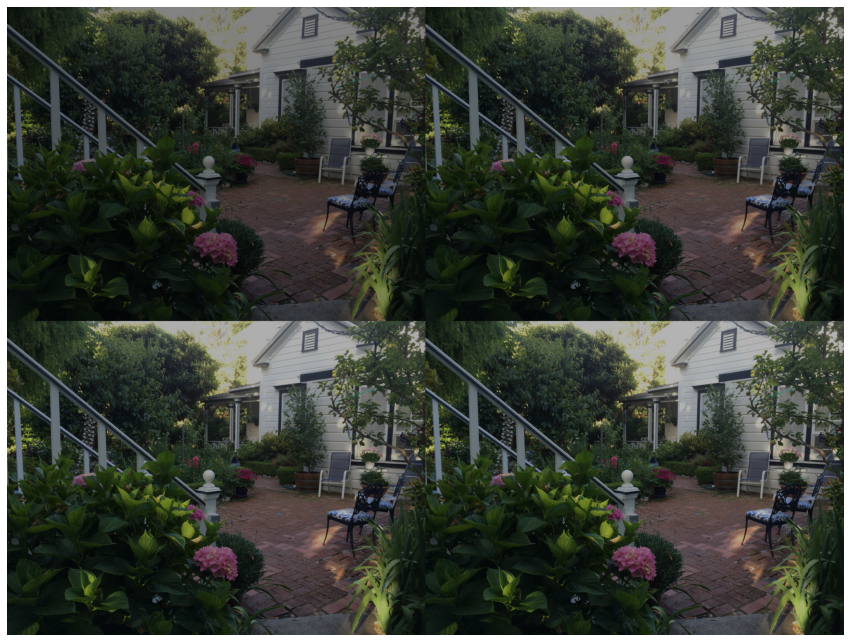}%
\end{tcolorbox}%
Shearing may not be appropriate for all image types, it stretches the image.%
\par%
\begin{tcolorbox}[size=title,title=Code,breakable]%
\begin{lstlisting}[language=Python, upquote=true]
visualize_generator(
  LOCAL_IMG_FILE,
  ImageDataGenerator(shear_range=30))\end{lstlisting}
\tcbsubtitle[before skip=\baselineskip]{Output}%
\includegraphics[width=4in]{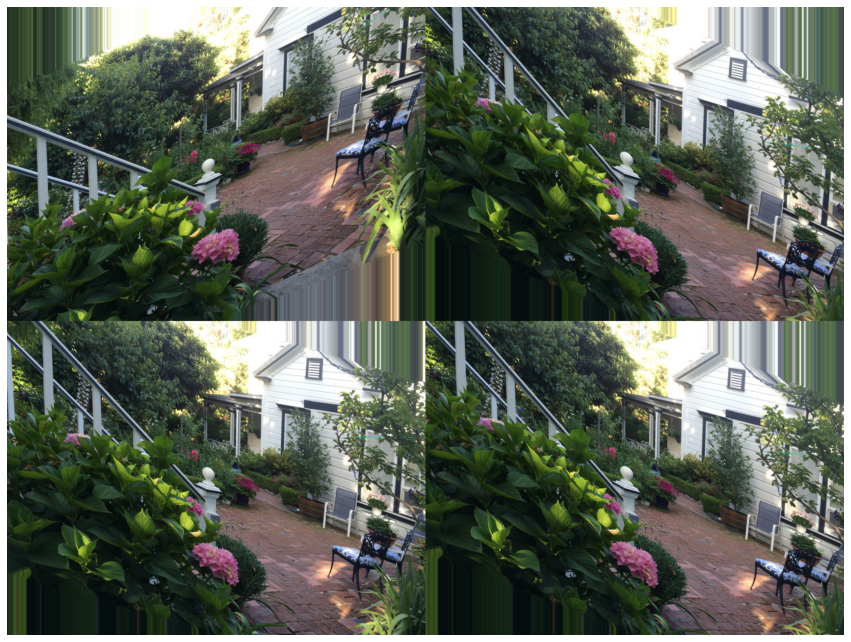}%
\end{tcolorbox}%
It is also possible to rotate images.%
\par%
\begin{tcolorbox}[size=title,title=Code,breakable]%
\begin{lstlisting}[language=Python, upquote=true]
visualize_generator(
  LOCAL_IMG_FILE,
  ImageDataGenerator(rotation_range=30))\end{lstlisting}
\tcbsubtitle[before skip=\baselineskip]{Output}%
\includegraphics[width=4in]{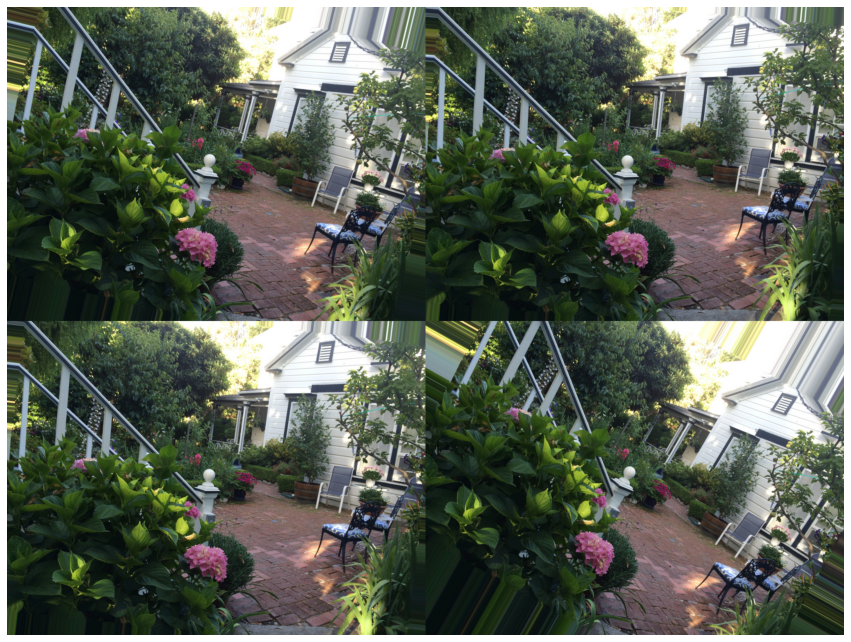}%
\end{tcolorbox}

\section{Part 6.5: Recognizing Multiple Images with YOLO5}%
\label{sec:Part6.5RecognizingMultipleImageswithYOLO5}%
Programmers typically design convolutional neural networks to classify a single item centered in an image. However, as humans, we can recognize many items in our field of view in real{-}time. It is advantageous to recognize multiple items in a single image. One of the most advanced means of doing this is YOLOv5. You Only Look Once (YOLO) was introduced by Joseph Redmon, who supported YOLO up through V3.%
\index{convolution}%
\index{convolutional}%
\index{Convolutional Neural Networks}%
\index{neural network}%
\index{YOLO}%
\index{YOLO}%
\cite{redmon2016you}%
The fact that YOLO must only look once speaks to the efficiency of the algorithm. In this context, to "look" means to perform one scan over the image. It is also possible to run YOLO on live video streams.%
\index{algorithm}%
\index{context}%
\index{video}%
\index{YOLO}%
\index{YOLO}%
\par%
Joseph Redmon left computer vision to pursue other interests. The current version, YOLOv5 is supported by the startup company%
\index{computer vision}%
\index{YOLO}%
\index{YOLO}%
\href{https://ultralytics.com/}{ Ultralytics}%
, who released the open{-}source library that we use in this class.%
\cite{zhu2021tph}%
\par%
Researchers have trained YOLO on a variety of different computer image datasets. The version of YOLO weights used in this course is from the dataset Common Objects in Context (COCO).%
\index{context}%
\index{dataset}%
\index{YOLO}%
\index{YOLO}%
\cite{lin2014microsoft}%
This dataset contains images labeled into 80 different classes. COCO is the source of the file coco.txt used in this module.%
\index{dataset}%
\par%
\subsection{Using YOLO in Python}%
\label{subsec:UsingYOLOinPython}%
To use YOLO in Python, we will use the open{-}source library provided by Ultralytics.%
\index{Python}%
\index{YOLO}%
\index{YOLO}%
\par%
\begin{itemize}[noitemsep]%
\item%
\href{https://github.com/ultralytics/yolov5}{YOLOv5 GitHub}%
\end{itemize}%
The code provided in this notebook works equally well when run either locally or from Google CoLab. It is easier to run YOLOv5 from CoLab, which is recommended for this course.%
\index{YOLO}%
\index{YOLO}%
\par%
We begin by obtaining an image to classify.%
\par%
\begin{tcolorbox}[size=title,title=Code,breakable]%
\begin{lstlisting}[language=Python, upquote=true]
import urllib.request
import shutil
from IPython.display import Image
!mkdir /content/images/

URL = "https://github.com/jeffheaton/t81_558_deep_learning"
URL += "/raw/master/photos/jeff_cook.jpg"
LOCAL_IMG_FILE = "/content/images/jeff_cook.jpg"

with urllib.request.urlopen(URL) as response, \
  open(LOCAL_IMG_FILE, 'wb') as out_file:
    shutil.copyfileobj(response, out_file)

Image(filename=LOCAL_IMG_FILE)\end{lstlisting}
\tcbsubtitle[before skip=\baselineskip]{Output}%
\includegraphics[width=3in]{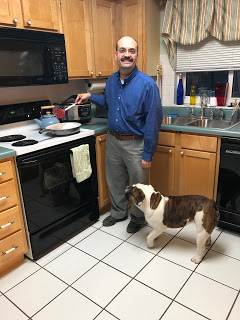}%
\end{tcolorbox}

\subsection{Installing YOLOv5}%
\label{subsec:InstallingYOLOv5}%
YOLO is not available directly through either PIP or CONDA. Additionally, YOLO is not installed in Google CoLab by default. Therefore, whether you wish to use YOLO through CoLab or run it locally, you need to go through several steps to install it. This section describes the process of installing YOLO. The same steps apply to either CoLab or a local install. For CoLab, you must repeat these steps each time the system restarts your virtual environment. You must perform these steps only once for your virtual Python environment for a local install. If you are installing locally, install to the same virtual environment you created for this course. The following commands install YOLO directly from its GitHub repository.%
\index{GitHub}%
\index{Python}%
\index{ROC}%
\index{ROC}%
\index{YOLO}%
\index{YOLO}%
\par%
\begin{tcolorbox}[size=title,title=Code,breakable]%
\begin{lstlisting}[language=Python, upquote=true]
!git clone https://github.com/ultralytics/yolov5 --tag 6.1
!mv /content/6.1 /content/yolov5
%cd /content/yolov5
%pip install -qr requirements.txt

from yolov5 import utils
display = utils.notebook_init()\end{lstlisting}
\tcbsubtitle[before skip=\baselineskip]{Output}%
\begin{lstlisting}[upquote=true]
Setup complete  (12 CPUs, 83.5 GB RAM, 39.9/166.8 GB disk)
\end{lstlisting}
\end{tcolorbox}%
Next, we will run YOLO from the command line and classify the previously downloaded kitchen picture.  You can run this classification on any image you choose.%
\index{classification}%
\index{YOLO}%
\index{YOLO}%
\par%
\begin{tcolorbox}[size=title,title=Code,breakable]%
\begin{lstlisting}[language=Python, upquote=true]
!python detect.py --weights yolov5s.pt --img 640 \
  --conf 0.25 --source /content/images/

URL = '/content/yolov5/runs/detect/exp/jeff_cook.jpg'
display.Image(filename=URL, width=300)\end{lstlisting}
\tcbsubtitle[before skip=\baselineskip]{Output}%
\includegraphics[width=3in]{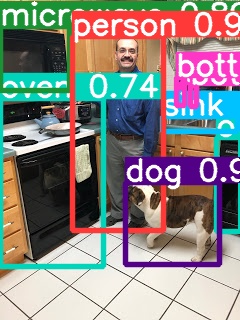}%
\begin{lstlisting}[upquote=true]
Downloading https://ultralytics.com/assets/Arial.ttf to
/root/.config/Ultralytics/Arial.ttf...
detect: weights=['yolov5s.pt'], source=/content/images/,
data=data/coco128.yaml, imgsz=[640, 640], conf_thres=0.25,
iou_thres=0.45, max_det=1000, device=, view_img=False, save_txt=False,
save_conf=False, save_crop=False, nosave=False, classes=None,
agnostic_nms=False, augment=False, visualize=False, update=False,
project=runs/detect, name=exp, exist_ok=False, line_thickness=3,
hide_labels=False, hide_conf=False, half=False, dnn=False
YOLOv5  v6.1-85-g6f4eb95 torch 1.10.0+cu111 CUDA:0 (A100-SXM4-40GB,
40536MiB)
Downloading https://github.com/ultralytics/yolov5/releases/download/v6
.1/yolov5s.pt to yolov5s.pt...
100% 14.1M/14.1M [00:00<00:00, 135MB/s]
Fusing layers...

...

image 1/1 /content/images/jeff_cook.jpg: 640x480 1 person, 1 dog, 3
bottles, 1 microwave, 2 ovens, 1 sink, Done. (0.016s)
Speed: 0.6ms pre-process, 15.9ms inference, 29.3ms NMS per image at
shape (1, 3, 640, 640)
Results saved to runs/detect/exp
\end{lstlisting}
\end{tcolorbox}

\subsection{Running YOLOv5}%
\label{subsec:RunningYOLOv5}%
In addition to the command line execution, we just saw.  The following code adds the downloaded YOLOv5 to Python's environment, allowing%
\index{Python}%
\index{YOLO}%
\index{YOLO}%
\textbf{ yolov5 }%
to be imported like a regular Python library.%
\index{Python}%
\par%
\begin{tcolorbox}[size=title,title=Code,breakable]%
\begin{lstlisting}[language=Python, upquote=true]
import sys
sys.path.append(str("/content/yolov5"))

from yolov5 import utils
display = utils.notebook_init()\end{lstlisting}
\tcbsubtitle[before skip=\baselineskip]{Output}%
\begin{lstlisting}[upquote=true]
Setup complete  (12 CPUs, 83.5 GB RAM, 39.9/166.8 GB disk)
\end{lstlisting}
\end{tcolorbox}%
Next, we obtain an image to classify. For this example, the program loads the image from a URL. YOLOv5 expects that the image is in the format of a Numpy array. We use PIL to obtain this image. We will convert it to the proper format for PyTorch and YOLOv5 later.%
\index{NumPy}%
\index{PyTorch}%
\index{YOLO}%
\index{YOLO}%
\par%
\begin{tcolorbox}[size=title,title=Code,breakable]%
\begin{lstlisting}[language=Python, upquote=true]
from PIL import Image
import requests
from io import BytesIO
import torchvision.transforms.functional as TF

url = "https://raw.githubusercontent.com/jeffheaton/"\
    "t81_558_deep_learning/master/images/cook.jpg"
response = requests.get(url,headers={'User-Agent': 'Mozilla/5.0'})
img = Image.open(BytesIO(response.content))\end{lstlisting}
\end{tcolorbox}%
The following libraries are needed to classify this image.%
\par%
\begin{tcolorbox}[size=title,title=Code,breakable]%
\begin{lstlisting}[language=Python, upquote=true]
import argparse
import os
import sys
from pathlib import Path

import cv2
import torch
import torch.backends.cudnn as cudnn

from models.common import DetectMultiBackend
from utils.datasets import IMG_FORMATS, VID_FORMATS, LoadImages, LoadStreams
from utils.general import (LOGGER, check_file, check_img_size, check_imshow, 
                           check_requirements, colorstr,
                           increment_path, non_max_suppression, 
                           print_args, scale_coords, strip_optimizer, 
                           xyxy2xywh)
from utils.plots import Annotator, colors, save_one_box
from utils.torch_utils import select_device, time_sync\end{lstlisting}
\end{tcolorbox}%
We are now ready to load YOLO with pretrained weights provided by the creators of YOLO.  It is also possible to train YOLO to recognize images of your own.%
\index{YOLO}%
\index{YOLO}%
\par%
\begin{tcolorbox}[size=title,title=Code,breakable]%
\begin{lstlisting}[language=Python, upquote=true]
device = select_device('')
weights = '/content/yolov5/yolov5s.pt'
imgsz = [img.height, img.width]
original_size = imgsz
model = DetectMultiBackend(weights, device=device, dnn=False)
stride, names, pt, jit, onnx, engine = model.stride, model.names, \
    model.pt, model.jit, model.onnx, model.engine
imgsz = check_img_size(imgsz, s=stride)  # check image size
print(f"Original size: {original_size}")
print(f"YOLO input size: {imgsz}")\end{lstlisting}
\tcbsubtitle[before skip=\baselineskip]{Output}%
\begin{lstlisting}[upquote=true]
Original size: [320, 240]
YOLO input size: [320, 256]
\end{lstlisting}
\end{tcolorbox}%
The creators of YOLOv5 built upon PyTorch, which has a particular format for images.  PyTorch images are generally a 4D matrix of the following dimensions:%
\index{matrix}%
\index{PyTorch}%
\index{YOLO}%
\index{YOLO}%
\par%
\begin{itemize}[noitemsep]%
\item%
batch\_size, channels, height, width%
\end{itemize}%
This code converts the previously loaded PIL image into this format.%
\par%
\begin{tcolorbox}[size=title,title=Code,breakable]%
\begin{lstlisting}[language=Python, upquote=true]
import numpy as np
source = '/content/images/'

conf_thres=0.25  # confidence threshold
iou_thres=0.45  # NMS IOU threshold
classes = None
agnostic_nms=False,  # class-agnostic NMS
max_det=1000

model.warmup(imgsz=(1, 3, *imgsz))  # warmup
dt, seen = [0.0, 0.0, 0.0], 0

# https://stackoverflow.com/questions/50657449/
# convert-image-to-proper-dimension-pytorch
img2 = img.resize([imgsz[1],imgsz[0]], Image.ANTIALIAS)
    
img_raw = torch.from_numpy(np.asarray(img2)).to(device)
img_raw = img_raw.float()  # uint8 to fp16/32
img_raw /= 255  # 0 - 255 to 0.0 - 1.0
img_raw = img_raw.unsqueeze_(0)
img_raw = img_raw.permute(0, 3, 1, 2)
print(img_raw.shape)\end{lstlisting}
\tcbsubtitle[before skip=\baselineskip]{Output}%
\begin{lstlisting}[upquote=true]
torch.Size([1, 3, 320, 256])
\end{lstlisting}
\end{tcolorbox}%
With the image converted, we are now ready to present the image to YOLO and obtain predictions.%
\index{predict}%
\index{YOLO}%
\index{YOLO}%
\par%
\begin{tcolorbox}[size=title,title=Code,breakable]%
\begin{lstlisting}[language=Python, upquote=true]
pred = model(img_raw, augment=False, visualize=False)
pred = non_max_suppression(pred, conf_thres, iou_thres, classes, 
  agnostic_nms, max_det=max_det)\end{lstlisting}
\end{tcolorbox}%
We now convert these raw predictions into the bounding boxes, labels, and confidences for each of the images that YOLO recognized.%
\index{predict}%
\index{YOLO}%
\index{YOLO}%
\par%
\begin{tcolorbox}[size=title,title=Code,breakable]%
\begin{lstlisting}[language=Python, upquote=true]
results = []
for i, det in enumerate(pred):  # per image
  gn = torch.tensor(img_raw.shape)[[1, 0, 1, 0]]  

  if len(det):
      # Rescale boxes from img_size to im0 size
      det[:, :4] = scale_coords(original_size, det[:, :4], imgsz).round()

      # Write results
      for *xyxy, conf, cls in reversed(det):
        xywh = (xyxy2xywh(torch.tensor(xyxy).view(1, 4)) / \
                gn).view(-1).tolist()
        # Choose between xyxy and xywh as your desired format.
        results.append([names[int(cls)], float(conf), [*xyxy]])\end{lstlisting}
\end{tcolorbox}%
We can now see the results from the classification. We will display the first 3.%
\index{classification}%
\par%
\begin{tcolorbox}[size=title,title=Code,breakable]%
\begin{lstlisting}[language=Python, upquote=true]
for itm in results[0:3]:
  print(itm)\end{lstlisting}
\tcbsubtitle[before skip=\baselineskip]{Output}%
\begin{lstlisting}[upquote=true]
['bowl', 0.28484195470809937, [tensor(55., device='cuda:0'),
tensor(120., device='cuda:0'), tensor(93., device='cuda:0'),
tensor(134., device='cuda:0')]]
['oven', 0.31531617045402527, [tensor(245., device='cuda:0'),
tensor(128., device='cuda:0'), tensor(256., device='cuda:0'),
tensor(231., device='cuda:0')]]
['bottle', 0.3567507565021515, [tensor(215., device='cuda:0'),
tensor(80., device='cuda:0'), tensor(223., device='cuda:0'),
tensor(101., device='cuda:0')]]
\end{lstlisting}
\end{tcolorbox}%
It is important to note that the%
\textbf{ yolo }%
class instantiated here is a callable object, which can fill the role of both an object and a function. Acting as a function,%
\textit{ yolo }%
returns three arrays named%
\textbf{ boxes}%
,%
\textbf{ scores}%
, and%
\textbf{ classes }%
that are of the same length.  The function returns all sub{-}images found with a score above the minimum threshold.  Additionally, the%
\textbf{ yolo }%
function returns an array named called%
\textbf{ nums}%
. The first element of the%
\textbf{ nums }%
array specifies how many sub{-}images YOLO found to be above the score threshold.%
\index{YOLO}%
\index{YOLO}%
\par%
\begin{itemize}[noitemsep]%
\item%
\textbf{boxes }%
{-} The bounding boxes for each sub{-}image detected in the image sent to YOLO.%
\index{YOLO}%
\index{YOLO}%
\item%
\textbf{scores }%
{-} The confidence for each of the sub{-}images detected.%
\item%
\textbf{classes }%
{-} The string class names for each item.  These are COCO names such as "person" or "dog."%
\item%
\textbf{nums }%
{-} The number of images above the threshold.%
\end{itemize}%
Your program should use these values to perform whatever actions you wish due to the input image.  The following code displays the images detected above the threshold.%
\index{input}%
\par%
To demonstrate the correctness of the results obtained, we draw bounding boxes over the original image.%
\par%
\begin{tcolorbox}[size=title,title=Code,breakable]%
\begin{lstlisting}[language=Python, upquote=true]
from PIL import Image, ImageDraw

img3 = img.copy()
draw = ImageDraw.Draw(img3)

for itm in results:
  b = itm[2]
  print(b)
  draw.rectangle(b)

img3\end{lstlisting}
\tcbsubtitle[before skip=\baselineskip]{Output}%
\includegraphics[width=3in]{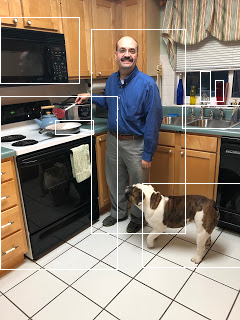}%
\begin{lstlisting}[upquote=true]
[tensor(55., device='cuda:0'), tensor(120., device='cuda:0'),
tensor(93., device='cuda:0'), tensor(134., device='cuda:0')]
[tensor(245., device='cuda:0'), tensor(128., device='cuda:0'),
tensor(256., device='cuda:0'), tensor(231., device='cuda:0')]
[tensor(215., device='cuda:0'), tensor(80., device='cuda:0'),
tensor(223., device='cuda:0'), tensor(101., device='cuda:0')]
[tensor(182., device='cuda:0'), tensor(105., device='cuda:0'),
tensor(256., device='cuda:0'), tensor(128., device='cuda:0')]
[tensor(200., device='cuda:0'), tensor(71., device='cuda:0'),
tensor(210., device='cuda:0'), tensor(101., device='cuda:0')]
[tensor(0., device='cuda:0'), tensor(96., device='cuda:0'),
tensor(117., device='cuda:0'), tensor(269., device='cuda:0')]
[tensor(0., device='cuda:0'), tensor(17., device='cuda:0'),
tensor(79., device='cuda:0'), tensor(83., device='cuda:0')]
[tensor(91., device='cuda:0'), tensor(29., device='cuda:0'),
tensor(185., device='cuda:0'), tensor(233., device='cuda:0')]
[tensor(142., device='cuda:0'), tensor(183., device='cuda:0'),
tensor(253., device='cuda:0'), tensor(267., device='cuda:0')]
\end{lstlisting}
\end{tcolorbox}

\subsection{Module 6 Assignment}%
\label{subsec:Module6Assignment}%
You can find the first assignment here:%
\href{https://github.com/jeffheaton/t81_558_deep_learning/blob/master/assignments/assignment_yourname_class6.ipynb}{ assignment 6}%
\par

\chapter{Generative Adversarial Networks}%
\label{chap:GenerativeAdversarialNetworks}%
\section{Part 7.1: Introduction to GANS for Image and Data Generation}%
\label{sec:Part7.1IntroductiontoGANSforImageandDataGeneration}%
A generative adversarial network (GAN) is a class of machine learning systems invented by Ian Goodfellow in 2014.%
\index{GAN}%
\index{learning}%
\cite{goodfellow2014generative}%
Two neural networks compete with each other in a game. The GAN training algorithm starts with a training set and learns to generate new data with the same distributions as the training set. For example, a GAN trained on photographs can generate new photographs that look at least superficially authentic to human observers, having many realistic characteristics.%
\index{algorithm}%
\index{GAN}%
\index{neural network}%
\index{training}%
\index{training algorithm}%
\par%
This chapter makes use of the PyTorch framework rather than Keras/TensorFlow. While there are versions of%
\index{Keras}%
\index{PyTorch}%
\index{TensorFlow}%
\href{https://github.com/jeffheaton/t81_558_deep_learning/blob/5e2528a08c302c82919001a3c3c8364c29c1b999/t81_558_class_07_3_style_gan.ipynb}{ StyleGAN2{-}ADA that work with TensorFlow 1.0}%
, NVIDIA has switched to PyTorch for StyleGAN. Running this notebook in this notebook in Google CoLab is the most straightforward means of completing this chapter. Because of this, I designed this notebook to run in Google CoLab. It will take some modifications if you wish to run it locally.%
\index{GAN}%
\index{NVIDIA}%
\index{PyTorch}%
\index{SOM}%
\index{StyleGAN}%
\par
\begin{figure}[h]%
\centering%
\includegraphics[width=4in]{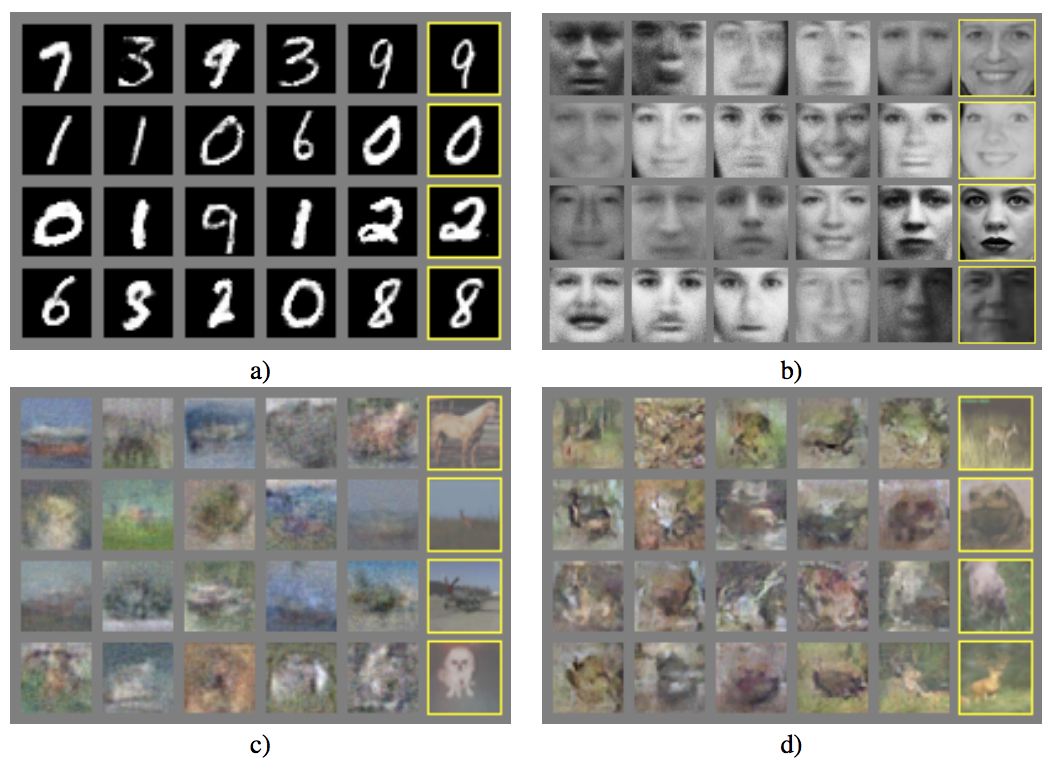}%
\caption{GAN Generated Images}%
\label{7.GANS}%
\end{figure}

%
This original StyleGAN paper used neural networks to automatically generate images for several previously seen datasets: MINST and CIFAR. However, it also included the Toronto Face Dataset (a private dataset used by some researchers). You can see some of these images in Figure \ref{7.GANS}.%
\index{dataset}%
\index{GAN}%
\index{neural network}%
\index{SOM}%
\index{StyleGAN}%
\par%

\par%
Only sub{-}figure D made use of convolutional neural networks. Figures A{-}C make use of fully connected neural networks. As we will see in this module, the researchers significantly increased the role of convolutional neural networks for GANs.%
\index{convolution}%
\index{convolutional}%
\index{Convolutional Neural Networks}%
\index{GAN}%
\index{neural network}%
\par
\begin{figure}[h]%
\centering%
\includegraphics[width=4in]{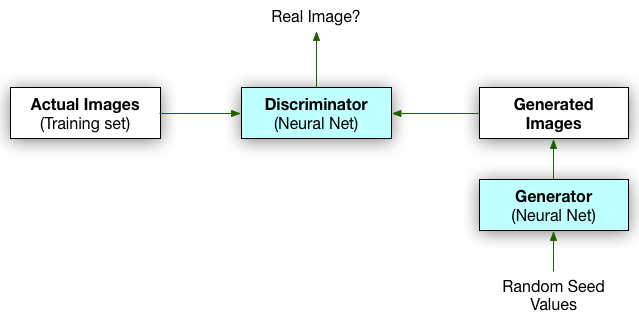}%
\caption{GAN Structure}%
\label{7.GAN-FLOW}%
\end{figure}

%
We call a GAN a generative model because it generates new data. You can see the overall process in Figure \ref{7.GAN-FLOW}.%
\index{GAN}%
\index{model}%
\index{ROC}%
\index{ROC}%
\par%

\par%
\subsection{Face Generation with StyleGAN and Python}%
\begin{figure}[h]%
\centering%
\includegraphics[width=4in]{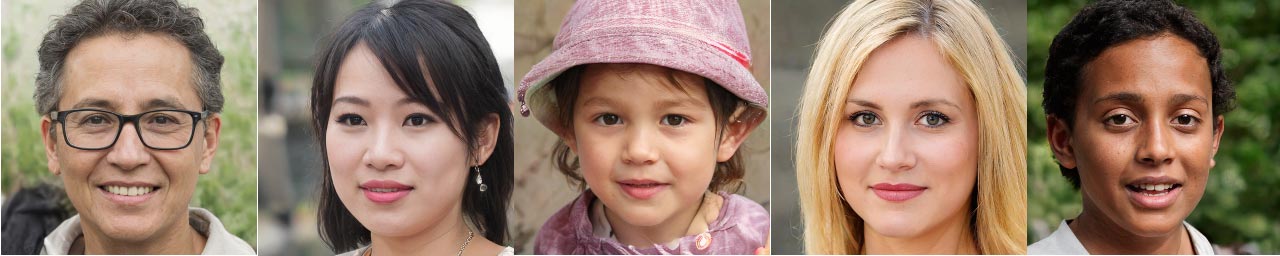}%
\caption{StyleGAN2 Generated Faces}%
\label{7.STY-GAN}%
\end{figure}

\label{subsec:FaceGenerationwithStyleGANandPython}%
GANs have appeared frequently in the media, showcasing their ability to generate highly photorealistic faces. One significant step forward for realistic face generation was the NVIDIA StyleGAN series. NVIDIA introduced the origional StyleGAN in 2018.%
\index{GAN}%
\index{NVIDIA}%
\index{StyleGAN}%
\cite{karras2019style}%
StyleGAN was followed by StyleGAN2 in 2019, which improved the quality of StyleGAN by removing certian artifacts.%
\index{GAN}%
\index{StyleGAN}%
\cite{karras2019analyzing}%
Most recently, in 2020, NVIDIA released StyleGAN2 adaptive discriminator augmentation (ADA), which will be the focus of this module.%
\index{GAN}%
\index{NVIDIA}%
\index{StyleGAN}%
\cite{karras2020training}%
We will see both how to train StyleGAN2 ADA on any arbitray set of images; as well as use pretrained weights provided by NVIDIA. The NVIDIA weights allow us to generate high resolution photorealistic looking faces, such seen in Figure \ref{7.STY-GAN}.%
\index{GAN}%
\index{NVIDIA}%
\index{StyleGAN}%
\par%

\par%
The above images were generated with StyleGAN2, using Google CoLab. Following the instructions in this section, you will be able to create faces like this of your own. StyleGAN2 images are usually 1,024 x 1,024 in resolution.  An example of a full{-}resolution StyleGAN image can be%
\index{GAN}%
\index{StyleGAN}%
\href{https://raw.githubusercontent.com/jeffheaton/t81_558_deep_learning/master/images/stylegan2-hires.jpg}{ found here}%
.%
\par
\begin{figure}[h]%
\centering%
\includegraphics[width=4in]{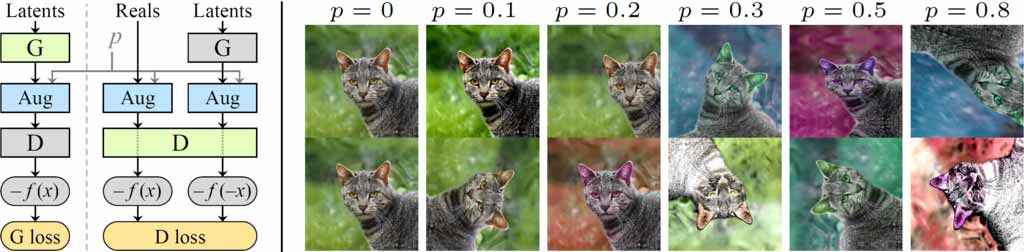}%
\caption{StyleGAN2 ADA Training}%
\label{7.STY-GAN-ADA}%
\end{figure}

%
The primary advancement introduced by the adaptive discriminator augmentation is that the algorithm augments the training images in real{-}time. Image augmentation is a common technique in many convolution neural network applications. Augmentation has the effect of increasing the size of the training set. Where StyleGAN2 previously required over 30K images for an effective to develop an effective neural network; now much fewer are needed. I used 2K images to train the fish generating GAN for this section. Figure \ref{7.STY-GAN-ADA} demonstrates the ADA process.%
\index{algorithm}%
\index{convolution}%
\index{GAN}%
\index{neural network}%
\index{ROC}%
\index{ROC}%
\index{StyleGAN}%
\index{training}%
\par%

\par%
The figure shows the increasing probability of augmentation as $p$ increases. For small image sets, the discriminator will generally memorize the image set unless the training algorithm makes use of augmentation. Once this memorization occurs, the discriminator is no longer providing useful information to the training of the generator.%
\index{algorithm}%
\index{probability}%
\index{training}%
\index{training algorithm}%
\par
\begin{figure}[h]%
\centering%
\includegraphics[width=4in]{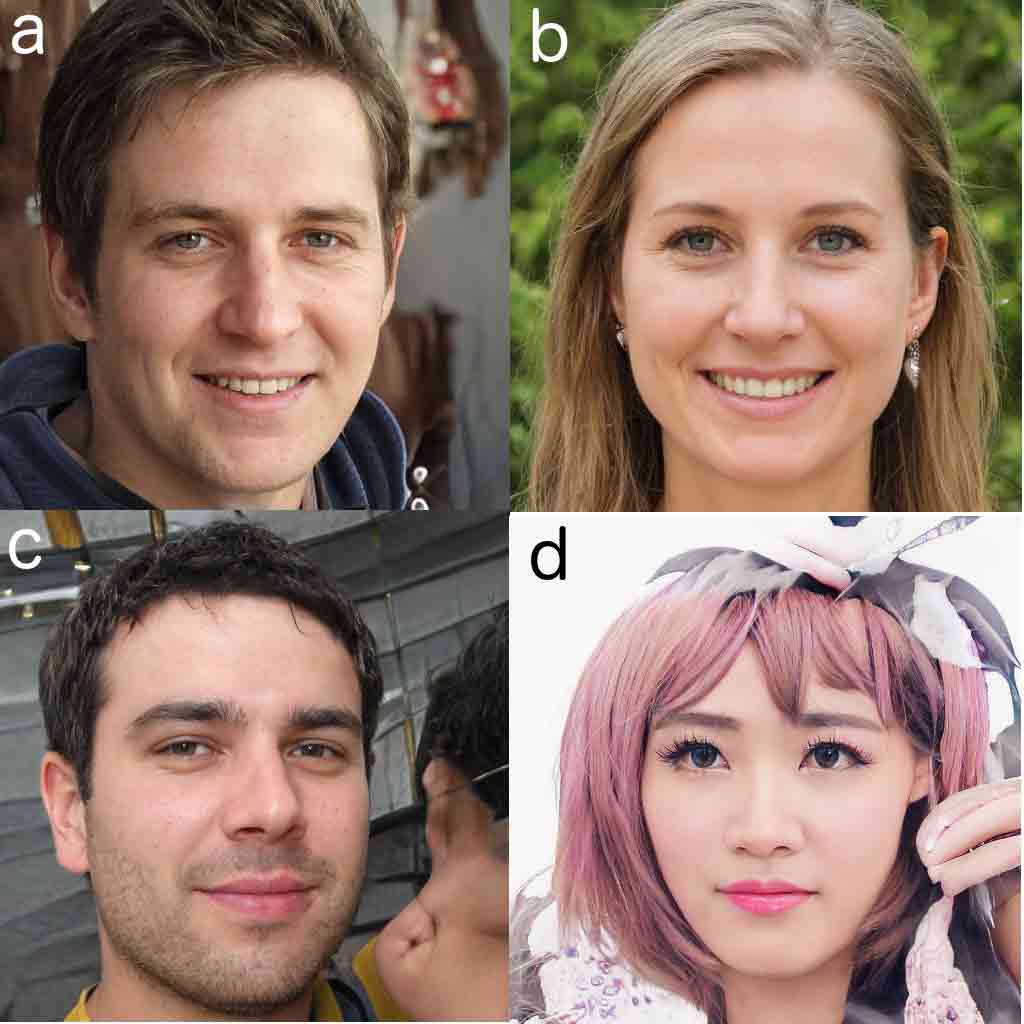}%
\caption{StyleGAN2 Face}%
\label{7.STYLEGAN2}%
\end{figure}

%
While the above images look much more realistic than images generated earlier in this course, they are not perfect. Look at Figure \ref{7.STYLEGAN2}. There are usually several tell{-}tail signs that you are looking at a computer{-}generated image. One of the most obvious is usually the surreal, dream{-}like backgrounds. The background does not look obviously fake at first glance; however, upon closer inspection, you usually can't quite discern what a GAN{-}generated background is. Also, look at the image character's left eye. It is slightly unrealistic looking, especially near the eyelashes.%
\index{GAN}%
\index{StyleGAN}%
\par%
Look at the following GAN face. Can you spot any imperfections?%
\index{GAN}%
\par%

\par%
\begin{itemize}[noitemsep]%
\item%
Image A demonstrates the abstract backgrounds usually associated with a GAN{-}generated image.%
\index{GAN}%
\item%
Image B exhibits issues that earrings often present for GANs. GANs sometimes have problems with symmetry, particularly earrings.%
\index{GAN}%
\index{SOM}%
\item%
Image C contains an abstract background and a highly distorted secondary image.%
\item%
Image D also contains a highly distorted secondary image that might be a hand.%
\end{itemize}%
Several websites allow you to generate GANs of your own without any software.%
\index{GAN}%
\par%
\begin{itemize}[noitemsep]%
\item%
\href{https://www.thispersondoesnotexist.com/}{This Person Does not Exist}%
\item%
\href{http://www.whichfaceisreal.com/}{Which Face is Real}%
\end{itemize}%
The first site generates high{-}resolution images of human faces. The second site presents a quiz to see if you can detect the difference between a real and fake human face image.%
\par%
In this chapter, you will learn to create your own StyleGAN pictures using Python.%
\index{GAN}%
\index{Python}%
\index{StyleGAN}%
\par

\subsection{Generating High Rez GAN Faces with Google CoLab}%
\label{subsec:GeneratingHighRezGANFaceswithGoogleCoLab}%
This notebook demonstrates how to run%
\href{https://github.com/NVlabs/stylegan2-ada}{ NVidia StyleGAN2 ADA }%
inside a Google CoLab notebook.  I suggest you use this to generate GAN faces from a pretrained model.  If you try to train your own, you will run into compute limitations of Google CoLab. Make sure to run this code on a GPU instance.  GPU is assumed.%
\index{GAN}%
\index{GPU}%
\index{GPU}%
\index{model}%
\par%
First, we clone StyleGAN3 from GitHub.%
\index{GAN}%
\index{GitHub}%
\index{StyleGAN}%
\par%
\begin{tcolorbox}[size=title,title=Code,breakable]%
\begin{lstlisting}[language=Python, upquote=true]
!git clone https://github.com/NVlabs/stylegan3.git
!pip install ninja\end{lstlisting}
\end{tcolorbox}%
Verify that StyleGAN has been cloned.%
\index{GAN}%
\index{StyleGAN}%
\par%
\begin{tcolorbox}[size=title,title=Code,breakable]%
\begin{lstlisting}[language=Python, upquote=true]
!ls /content/stylegan3\end{lstlisting}
\tcbsubtitle[before skip=\baselineskip]{Output}%
\begin{lstlisting}[upquote=true]
avg_spectra.py   Dockerfile       gen_video.py  metrics      train.py
calc_metrics.py  docs             gui_utils     README.md
visualizer.py
dataset_tool.py  environment.yml  legacy.py     torch_utils  viz
dnnlib           gen_images.py    LICENSE.txt   training
\end{lstlisting}
\end{tcolorbox}

\subsection{Run StyleGan From Command Line}%
\label{subsec:RunStyleGanFromCommandLine}%
Add the StyleGAN folder to Python so that you can import it. I based this code below on code from NVidia for the original StyleGAN paper. When you use StyleGAN you will generally create a GAN from a seed number. This seed is an integer, such as 6600, that will generate a unique image. The seed generates a latent vector containing 512 floating{-}point values. The GAN code uses the seed to generate these 512 values. The seed value is easier to represent in code than a 512 value vector; however, while a small change to the latent vector results in a slight change to the image, even a small change to the integer seed value will produce a radically different image.%
\index{GAN}%
\index{latent vector}%
\index{NVIDIA}%
\index{Python}%
\index{StyleGAN}%
\index{vector}%
\par%
\begin{tcolorbox}[size=title,title=Code,breakable]%
\begin{lstlisting}[language=Python, upquote=true]
URL = "https://api.ngc.nvidia.com/v2/models/nvidia/research/"\
      "stylegan3/versions/1/files/stylegan3-r-ffhq-1024x1024.pkl"

!python /content/stylegan3/gen_images.py \
    --network={URL} \
  --outdir=/content/results --seeds=6600-6625\end{lstlisting}
\end{tcolorbox}%
We can now display the images created.%
\par%
\begin{tcolorbox}[size=title,title=Code,breakable]%
\begin{lstlisting}[language=Python, upquote=true]
!ls /content/results\end{lstlisting}
\tcbsubtitle[before skip=\baselineskip]{Output}%
\begin{lstlisting}[upquote=true]
seed6600.png  seed6606.png  seed6612.png  seed6618.png  seed6624.png
seed6601.png  seed6607.png  seed6613.png  seed6619.png  seed6625.png
seed6602.png  seed6608.png  seed6614.png  seed6620.png
seed6603.png  seed6609.png  seed6615.png  seed6621.png
seed6604.png  seed6610.png  seed6616.png  seed6622.png
seed6605.png  seed6611.png  seed6617.png  seed6623.png
\end{lstlisting}
\end{tcolorbox}%
Next, copy the images to a folder of your choice on GDrive.%
\par%
\begin{tcolorbox}[size=title,title=Code,breakable]%
\begin{lstlisting}[language=Python, upquote=true]
!cp /content/results/* \
    /content/drive/My\ Drive/projects/stylegan3\end{lstlisting}
\end{tcolorbox}

\subsection{Run StyleGAN From Python Code}%
\label{subsec:RunStyleGANFromPythonCode}%
Add the StyleGAN folder to Python so that you can import it.%
\index{GAN}%
\index{Python}%
\index{StyleGAN}%
\par%
\begin{tcolorbox}[size=title,title=Code,breakable]%
\begin{lstlisting}[language=Python, upquote=true]
import sys
sys.path.insert(0, "/content/stylegan3")
import pickle
import os
import numpy as np
import PIL.Image
from IPython.display import Image
import matplotlib.pyplot as plt
import IPython.display
import torch
import dnnlib
import legacy

def seed2vec(G, seed):
  return np.random.RandomState(seed).randn(1, G.z_dim)

def display_image(image):
  plt.axis('off')
  plt.imshow(image)
  plt.show()

def generate_image(G, z, truncation_psi):
    # Render images for dlatents initialized from random seeds.
    Gs_kwargs = {
        'output_transform': dict(func=tflib.convert_images_to_uint8, 
         nchw_to_nhwc=True),
        'randomize_noise': False
    }
    if truncation_psi is not None:
        Gs_kwargs['truncation_psi'] = truncation_psi

    label = np.zeros([1] + G.input_shapes[1][1:])
    # [minibatch, height, width, channel]
    images = G.run(z, label, **G_kwargs) 
    return images[0]

def get_label(G, device, class_idx):
  label = torch.zeros([1, G.c_dim], device=device)
  if G.c_dim != 0:
      if class_idx is None:
          ctx.fail("Must specify class label with --class when using "\
            "a conditional network")
      label[:, class_idx] = 1
  else:
      if class_idx is not None:
          print ("warn: --class=lbl ignored when running on "\
            "an unconditional network")
  return label

def generate_image(device, G, z, truncation_psi=1.0, noise_mode='const', 
                   class_idx=None):
  z = torch.from_numpy(z).to(device)
  label = get_label(G, device, class_idx)
  img = G(z, label, truncation_psi=truncation_psi, noise_mode=noise_mode)
  img = (img.permute(0, 2, 3, 1) * 127.5 + 128).clamp(0, 255).to(\
      torch.uint8)
  return PIL.Image.fromarray(img[0].cpu().numpy(), 'RGB')\end{lstlisting}
\end{tcolorbox}%
\begin{tcolorbox}[size=title,title=Code,breakable]%
\begin{lstlisting}[language=Python, upquote=true]
#URL = "https://github.com/jeffheaton/pretrained-gan-fish/releases/"\
#  "download/1.0.0/fish-gan-2020-12-09.pkl"
#URL = "https://github.com/jeffheaton/pretrained-merry-gan-mas/releases/"\
#  "download/v1/christmas-gan-2020-12-03.pkl"
URL = "https://api.ngc.nvidia.com/v2/models/nvidia/research/stylegan3/"\
  "versions/1/files/stylegan3-r-ffhq-1024x1024.pkl"

print(f'Loading networks from "{URL}"...')
device = torch.device('cuda')
with dnnlib.util.open_url(URL) as f:
    G = legacy.load_network_pkl(f)['G_ema'].to(device) # type: ignore\end{lstlisting}
\tcbsubtitle[before skip=\baselineskip]{Output}%
\begin{lstlisting}[upquote=true]
Loading networks from "https://api.ngc.nvidia.com/v2/models/nvidia/res
earch/stylegan3/versions/1/files/stylegan3-r-ffhq-1024x1024.pkl"...
\end{lstlisting}
\end{tcolorbox}%
We can now generate images from integer seed codes in Python.%
\index{Python}%
\par%
\begin{tcolorbox}[size=title,title=Code,breakable]%
\begin{lstlisting}[language=Python, upquote=true]
# Choose your own starting and ending seed.
SEED_FROM = 1000
SEED_TO = 1003

# Generate the images for the seeds.
for i in range(SEED_FROM, SEED_TO):
  print(f"Seed {i}")
  z = seed2vec(G, i)
  img = generate_image(device, G, z)
  display_image(img)\end{lstlisting}
\tcbsubtitle[before skip=\baselineskip]{Output}%
\includegraphics[width=3in]{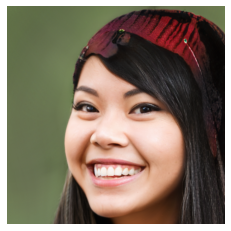}%
\begin{lstlisting}[upquote=true]
Seed 1000
Setting up PyTorch plugin "bias_act_plugin"... Done.
Setting up PyTorch plugin "filtered_lrelu_plugin"... Done.
\end{lstlisting}
\includegraphics[width=3in]{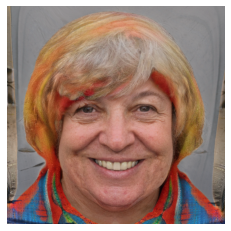}%
\begin{lstlisting}[upquote=true]
Seed 1001
\end{lstlisting}
\includegraphics[width=3in]{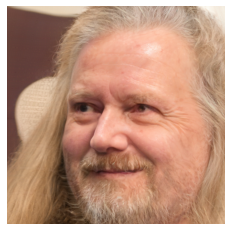}%
\begin{lstlisting}[upquote=true]
Seed 1002
\end{lstlisting}
\end{tcolorbox}

\subsection{Examining the Latent Vector}%
\begin{figure}[h]%
\centering%
\includegraphics[width=4in]{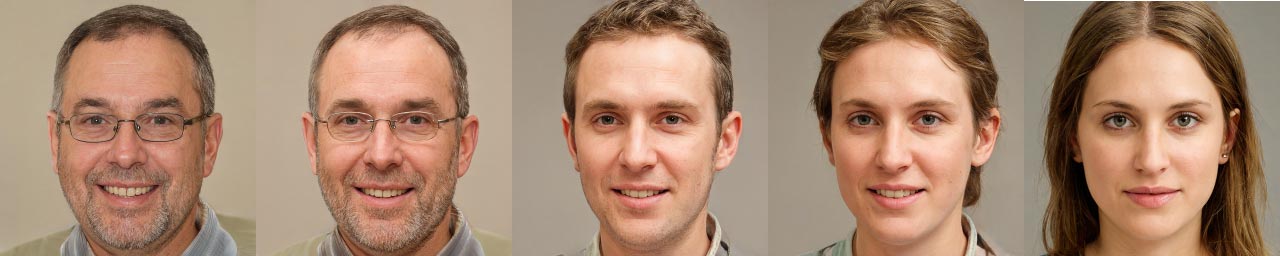}%
\caption{Transforming the Latent Vector}%
\label{7.LVEC}%
\end{figure}

\label{subsec:ExaminingtheLatentVector}%
Figure \ref{7.LVEC} shows the effects of transforming the latent vector between two images. We accomplish this transformation by slowly moving one 512{-}value latent vector to another 512 vector. A high{-}dimension point between two latent vectors will appear similar to both of the two endpoint latent vectors. Images that have similar latent vectors will appear similar to each other.%
\index{latent vector}%
\index{vector}%
\par%

\par%
\begin{tcolorbox}[size=title,title=Code,breakable]%
\begin{lstlisting}[language=Python, upquote=true]
def expand_seed(seeds, vector_size):
  result = []

  for seed in seeds:
    rnd = np.random.RandomState(seed)
    result.append( rnd.randn(1, vector_size) ) 
  return result

#URL = "https://github.com/jeffheaton/pretrained-gan-fish/releases/"\
#  "download/1.0.0/fish-gan-2020-12-09.pkl"
#URL = "https://github.com/jeffheaton/pretrained-merry-gan-mas/releases/"\
#  "download/v1/christmas-gan-2020-12-03.pkl"
#URL = "https://nvlabs-fi-cdn.nvidia.com/stylegan2-ada/pretrained/ffhq.pkl"
URL = "https://api.ngc.nvidia.com/v2/models/nvidia/research/stylegan3/"\
  "versions/1/files/stylegan3-r-ffhq-1024x1024.pkl"

print(f'Loading networks from "{URL}"...')
device = torch.device('cuda')
with dnnlib.util.open_url(URL) as f:
    G = legacy.load_network_pkl(f)['G_ema'].to(device) # type: ignore

vector_size = G.z_dim
# range(8192,8300)
seeds = expand_seed( [8192+1,8192+9], vector_size)
#generate_images(Gs, seeds,truncation_psi=0.5)
print(seeds[0].shape)\end{lstlisting}
\tcbsubtitle[before skip=\baselineskip]{Output}%
\begin{lstlisting}[upquote=true]
Loading networks from "https://api.ngc.nvidia.com/v2/models/nvidia/res
earch/stylegan3/versions/1/files/stylegan3-r-ffhq-1024x1024.pkl"...
(1, 512)
\end{lstlisting}
\end{tcolorbox}%
The following code will move between the provided seeds.  The constant STEPS specify how many frames there should be between each seed.%
\par%
\begin{tcolorbox}[size=title,title=Code,breakable]%
\begin{lstlisting}[language=Python, upquote=true]
# Choose your seeds to morph through and the number of steps to 
# take to get to each.

SEEDS = [6624,6618,6616] # Better for faces
#SEEDS = [1000,1003,1001] # Better for fish
STEPS = 100

# Remove any prior results
!rm /content/results/* 

from tqdm.notebook import tqdm

os.makedirs("./results/", exist_ok=True)

# Generate the images for the video.
idx = 0
for i in range(len(SEEDS)-1):
  v1 = seed2vec(G, SEEDS[i])
  v2 = seed2vec(G, SEEDS[i+1])

  diff = v2 - v1
  step = diff / STEPS
  current = v1.copy()

  for j in tqdm(range(STEPS), desc=f"Seed {SEEDS[i]}"):
    current = current + step
    img = generate_image(device, G, current)
    img.save(f'./results/frame-{idx}.png')
    idx+=1
 
# Link the images into a video.
!ffmpeg -r 30 -i /content/results/frame-%d.png -vcodec mpeg4 -y movie.mp4\end{lstlisting}
\end{tcolorbox}%
You can now download the generated video.%
\index{video}%
\par%
\begin{tcolorbox}[size=title,title=Code,breakable]%
\begin{lstlisting}[language=Python, upquote=true]
from google.colab import files
files.download('movie.mp4')\end{lstlisting}
\tcbsubtitle[before skip=\baselineskip]{Output}%
\begin{lstlisting}[upquote=true]
<IPython.core.display.Javascript
object><IPython.core.display.Javascript object>
\end{lstlisting}
\end{tcolorbox}

\subsection{Module 7 Assignment}%
\label{subsec:Module7Assignment}%
You can find the first assignment here:%
\href{https://github.com/jeffheaton/t81_558_deep_learning/blob/master/assignments/assignment_yourname_class7.ipynb}{ assignment 7}%
\par

\section{Part 7.2: Train StyleGAN3 with your Images}%
\label{sec:Part7.2TrainStyleGAN3withyourImages}%
Training GANs with StyleGAN is resource{-}intensive. The NVIDA StyleGAN researchers used computers with eight high{-}end GPUs for the high{-}resolution face GANs trained by NVIDIA. The GPU used by NVIDIA is an A100, which has more memory and cores than the P100 or V100 offered by even Colab Pro+. In this part, we will use StyleGAN2 to train rather than StyleGAN3. You can use networks trained with StyleGAN2 from StyleGAN3; however, StyleGAN3 usually is more effective at training than StyleGAN2.%
\index{GAN}%
\index{GPU}%
\index{GPU}%
\index{NVIDIA}%
\index{StyleGAN}%
\index{training}%
\par%
Unfortunately, StyleGAN3 is compute{-}intensive and will perform slowly on any GPU that is not the latest Ampere technology. Because Colab does not provide such technology, I am keeping the training guide at the StyleGAN2 level. Switching to StyleGAN3 is relatively easy, as will be pointed out later.%
\index{GAN}%
\index{GPU}%
\index{GPU}%
\index{StyleGAN}%
\index{training}%
\par%
Make sure that you are running this notebook with a GPU runtime. You can train GANs with either Google Colab Free or Pro. I recommend at least the Pro version due to better GPU instances, longer runtimes, and timeouts. Additionally, the capability of Google Colab Pro to run in the background is valuable when training GANs, as you can close your browser or reboot your laptop while training continues.%
\index{GAN}%
\index{GPU}%
\index{GPU}%
\index{training}%
\par%
You will store your training data and trained neural networks to GDRIVE. For GANs, I lay out my GDRIVE like this:%
\index{GAN}%
\index{neural network}%
\index{training}%
\par%
\begin{itemize}[noitemsep]%
\item%
./data/gan/images {-} RAW images I wish to train on.%
\index{GAN}%
\item%
./data/gan/datasets {-} Actual training datasets that I convert from the raw images.%
\index{dataset}%
\index{GAN}%
\index{training}%
\item%
./data/gan/experiments {-} The output from StyleGAN2, my image previews, and saved network snapshots.%
\index{GAN}%
\index{output}%
\index{StyleGAN}%
\end{itemize}%
You will mount the drive at the following location.%
\par%
\begin{tcolorbox}[size=title,breakable]%
\begin{lstlisting}[upquote=true]
/content/drive/MyDrive/data
\end{lstlisting}
\end{tcolorbox}%
\subsection{What Sort of GPU do you Have?}%
\label{subsec:WhatSortofGPUdoyouHave?}%
The type of GPU assigned to you by Colab will significantly affect your training time. Some sample times that I achieved with Colab are given here. I've found that Colab Pro generally starts you with a V100, however, if you run scripts non{-}stop for 24hrs straight for a few days in a row, you will generally be throttled back to a P100.%
\index{GPU}%
\index{GPU}%
\index{SOM}%
\index{training}%
\par%
\begin{itemize}[noitemsep]%
\item%
1024x1024 {-} V100 {-} 566 sec/tick (CoLab Pro)%
\item%
1024x1024 {-} P100 {-} 1819 sec/tick (CoLab Pro)%
\item%
1024x1024 {-} T4 {-} 2188 sec/tick (CoLab Free)%
\end{itemize}%
By comparison, a 1024x1024 GAN trained with StyleGAN3 on a V100 is 3087 sec/tick.%
\index{GAN}%
\index{StyleGAN}%
\par%
If you use Google CoLab Pro, generally, it will not disconnect before 24 hours, even if you (but not your script) are inactive. Free CoLab WILL disconnect a perfectly good running script if you do not interact for a few hours. The following describes how to circumvent this issue.%
\par%
\begin{itemize}[noitemsep]%
\item%
\href{https://stackoverflow.com/questions/57113226/how-to-prevent-google-colab-from-disconnecting}{How to prevent Google Colab from disconnecting?}%
\end{itemize}

\subsection{Set Up New Environment}%
\label{subsec:SetUpNewEnvironment}%
You will likely need to train for >24 hours. Colab will disconnect you. You must be prepared to restart training when this eventually happens. Training is divided into ticks, every so many ticks (50 by default), your neural network is evaluated, and a snapshot is saved. When CoLab shuts down, all training after the last snapshot is lost. It might seem desirable to snapshot after each tick; however, this snapshotting process itself takes nearly an hour. Learning an optimal snapshot size for your resolution and training data is important.%
\index{learning}%
\index{neural network}%
\index{ROC}%
\index{ROC}%
\index{training}%
\par%
We will mount GDRIVE so that you will save your snapshots there. You must also place your training images in GDRIVE.%
\index{training}%
\par%
You must also install NVIDIA StyleGAN2 ADA PyTorch. We also need to downgrade PyTorch to a version that supports StyleGAN.%
\index{GAN}%
\index{NVIDIA}%
\index{PyTorch}%
\index{StyleGAN}%
\par%
\begin{tcolorbox}[size=title,title=Code,breakable]%
\begin{lstlisting}[language=Python, upquote=true]
!pip install torch==1.8.1 torchvision==0.9.1
!git clone https://github.com/NVlabs/stylegan2-ada-pytorch.git
!pip install ninja\end{lstlisting}
\end{tcolorbox}

\subsection{Find Your Files}%
\label{subsec:FindYourFiles}%
The drive is mounted to the following location.%
\par%
\begin{tcolorbox}[size=title,breakable]%
\begin{lstlisting}[upquote=true]
/content/drive/MyDrive/data
\end{lstlisting}
\end{tcolorbox}%
It might be helpful to use an%
\textbf{\texttt{ ls }}%
command to establish the exact path for your images.%
\par%
\begin{tcolorbox}[size=title,title=Code,breakable]%
\begin{lstlisting}[language=Python, upquote=true]
!ls /content/drive/MyDrive/data/gan/images\end{lstlisting}
\end{tcolorbox}

\subsection{Convert Your Images}%
\label{subsec:ConvertYourImages}%
You must convert your images into a data set form that PyTorch can directly utilize. The following command converts your images and writes the resulting data set to another directory.%
\index{PyTorch}%
\par%
\begin{tcolorbox}[size=title,title=Code,breakable]%
\begin{lstlisting}[language=Python, upquote=true]
CMD = "python /content/stylegan2-ada-pytorch/dataset_tool.py "\
  "--source /content/drive/MyDrive/data/gan/images/circuit "\
  "--dest /content/drive/MyDrive/data/gan/dataset/circuit"

!{CMD}\end{lstlisting}
\end{tcolorbox}%
You can use the following command to clear out the newly created dataset.  If something goes wrong and you need to clean up your images and rerun the above command, you should delete your partially completed dataset directory.%
\index{dataset}%
\index{SOM}%
\par%
\begin{tcolorbox}[size=title,title=Code,breakable]%
\begin{lstlisting}[language=Python, upquote=true]
#!rm -R /content/drive/MyDrive/data/gan/dataset/circuit/*\end{lstlisting}
\end{tcolorbox}

\subsection{Clean Up your Images}%
\label{subsec:CleanUpyourImages}%
All images must have the same dimensions and color depth.  This code can identify images that have issues.%
\par%
\begin{tcolorbox}[size=title,title=Code,breakable]%
\begin{lstlisting}[language=Python, upquote=true]
from os import listdir
from os.path import isfile, join
import os
from PIL import Image
from tqdm.notebook import tqdm

IMAGE_PATH = '/content/drive/MyDrive/data/gan/images/fish'
files = [f for f in listdir(IMAGE_PATH) if isfile(join(IMAGE_PATH, f))]

base_size = None
for file in tqdm(files):
  file2 = os.path.join(IMAGE_PATH,file)
  img = Image.open(file2)
  sz = img.size
  if base_size and sz!=base_size:
    print(f"Inconsistant size: {file2}")
  elif img.mode!='RGB':
    print(f"Inconsistant color format: {file2}")
  else:
    base_size = sz\end{lstlisting}
\end{tcolorbox}

\subsection{Perform Initial Training}%
\label{subsec:PerformInitialTraining}%
This code performs the initial training.  Set SNAP low enough to get a snapshot before Colab forces you to quit.%
\index{training}%
\par%
\begin{tcolorbox}[size=title,title=Code,breakable]%
\begin{lstlisting}[language=Python, upquote=true]
import os

# Modify these to suit your needs
EXPERIMENTS = "/content/drive/MyDrive/data/gan/experiments"
DATA = "/content/drive/MyDrive/data/gan/dataset/circuit"
SNAP = 10

# Build the command and run it
cmd = f"/usr/bin/python3 /content/stylegan2-ada-pytorch/train.py "\
  f"--snap {SNAP} --outdir {EXPERIMENTS} --data {DATA}"
!{cmd}\end{lstlisting}
\end{tcolorbox}

\subsection{Resume Training}%
\label{subsec:ResumeTraining}%
You can now resume training after you are interrupted by something in the pervious step.%
\index{SOM}%
\index{training}%
\par%
\begin{tcolorbox}[size=title,title=Code,breakable]%
\begin{lstlisting}[language=Python, upquote=true]
import os

# Modify these to suit your needs
EXPERIMENTS = "/content/drive/MyDrive/data/gan/experiments"
NETWORK = "network-snapshot-000100.pkl"
RESUME = os.path.join(EXPERIMENTS, \
                "00008-circuit-auto1-resumecustom", NETWORK)
DATA = "/content/drive/MyDrive/data/gan/dataset/circuit"
SNAP = 10

# Build the command and run it
cmd = f"/usr/bin/python3 /content/stylegan2-ada-pytorch/train.py "\
  f"--snap {SNAP} --resume {RESUME} --outdir {EXPERIMENTS} --data {DATA}"
!{cmd}\end{lstlisting}
\end{tcolorbox}

\section{Part 7.3: Exploring the StyleGAN Latent Vector}%
\label{sec:Part7.3ExploringtheStyleGANLatentVector}%
StyleGAN seeds, such as 3000, are only random number seeds used to generate much longer 512{-}length latent vectors, which create the GAN image.  If you make a small change to the seed, for example, change 3000 to 3001, StyleGAN will create an entirely different picture.  However, if you make a small change to a few latent vector values, the image will only change slightly.  In this part, we will see how we can fine{-}tune the latent vector to control, to some degree, the resulting GAN image appearance.%
\index{GAN}%
\index{latent vector}%
\index{random}%
\index{SOM}%
\index{StyleGAN}%
\index{vector}%
\par%
\subsection{Installing Needed Software}%
\label{subsec:InstallingNeededSoftware}%
We begin by installing StyleGAN.%
\index{GAN}%
\index{StyleGAN}%
\par%
\begin{tcolorbox}[size=title,title=Code,breakable]%
\begin{lstlisting}[language=Python, upquote=true]
!git clone https://github.com/NVlabs/stylegan3.git
!pip install ninja\end{lstlisting}
\end{tcolorbox}%
We will use the same functions introduced in the previous part to generate GAN seeds and images.%
\index{GAN}%
\par%
\begin{tcolorbox}[size=title,title=Code,breakable]%
\begin{lstlisting}[language=Python, upquote=true]
import sys
sys.path.insert(0, "/content/stylegan3")
import pickle
import os
import numpy as np
import PIL.Image
from IPython.display import Image
import matplotlib.pyplot as plt
import IPython.display
import torch
import dnnlib
import legacy

def seed2vec(G, seed):
  return np.random.RandomState(seed).randn(1, G.z_dim)

def display_image(image):
  plt.axis('off')
  plt.imshow(image)
  plt.show()

def generate_image(G, z, truncation_psi):
    # Render images for dlatents initialized from random seeds.
    Gs_kwargs = {
        'output_transform': dict(func=tflib.convert_images_to_uint8, 
        nchw_to_nhwc=True),
        'randomize_noise': False
    }
    if truncation_psi is not None:
        Gs_kwargs['truncation_psi'] = truncation_psi

    label = np.zeros([1] + G.input_shapes[1][1:])
    # [minibatch, height, width, channel]
    images = G.run(z, label, **G_kwargs) 
    return images[0]

def get_label(G, device, class_idx):
  label = torch.zeros([1, G.c_dim], device=device)
  if G.c_dim != 0:
      if class_idx is None:
          ctx.fail('Must specify class label with --class'\
                   'when using a conditional network')
      label[:, class_idx] = 1
  else:
      if class_idx is not None:
          print ('warn: --class=lbl ignored when running '\
            'on an unconditional network')
  return label

def generate_image(device, G, z, truncation_psi=1.0, 
                   noise_mode='const', class_idx=None):
  z = torch.from_numpy(z).to(device)
  label = get_label(G, device, class_idx)
  img = G(z, label, truncation_psi=truncation_psi, 
          noise_mode=noise_mode)
  img = (img.permute(0, 2, 3, 1) * 127.5 + 128)\
    .clamp(0, 255).to(torch.uint8)
  return PIL.Image.fromarray(img[0].cpu().numpy(), 'RGB')\end{lstlisting}
\end{tcolorbox}%
Next, we load the NVIDIA FFHQ (faces) GAN.  We could use any StyleGAN pretrained GAN network here.%
\index{GAN}%
\index{NVIDIA}%
\index{StyleGAN}%
\par%
\begin{tcolorbox}[size=title,title=Code,breakable]%
\begin{lstlisting}[language=Python, upquote=true]
# HIDE CODE

URL = "https://api.ngc.nvidia.com/v2/models/nvidia/research/"\
  "stylegan3/versions/1/files/stylegan3-r-ffhq-1024x1024.pkl"

print('Loading networks from "%s"...' % URL)
device = torch.device('cuda')
with dnnlib.util.open_url(URL) as fp:
    G = legacy.load_network_pkl(fp)['G_ema']\
      .requires_grad_(False).to(device)\end{lstlisting}
\tcbsubtitle[before skip=\baselineskip]{Output}%
\begin{lstlisting}[upquote=true]
Loading networks from "https://api.ngc.nvidia.com/v2/models/nvidia/res
earch/stylegan3/versions/1/files/stylegan3-r-ffhq-1024x1024.pkl"...
Downloading https://api.ngc.nvidia.com/v2/models/nvidia/research/style
gan3/versions/1/files/stylegan3-r-ffhq-1024x1024.pkl ... done
\end{lstlisting}
\end{tcolorbox}

\subsection{Generate and View GANS from Seeds}%
\label{subsec:GenerateandViewGANSfromSeeds}%
We will begin by generating a few seeds to evaluate potential starting points for our fine{-}tuning. Try out different seeds ranges until you have a seed that looks close to what you wish to fine{-}tune.%
\par%
\begin{tcolorbox}[size=title,title=Code,breakable]%
\begin{lstlisting}[language=Python, upquote=true]
# Choose your own starting and ending seed.
SEED_FROM = 4020
SEED_TO = 4023

# Generate the images for the seeds.
for i in range(SEED_FROM, SEED_TO):
  print(f"Seed {i}")
  z = seed2vec(G, i)
  img = generate_image(device, G, z)
  display_image(img)\end{lstlisting}
\tcbsubtitle[before skip=\baselineskip]{Output}%
\includegraphics[width=3in]{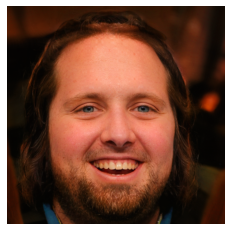}%
\begin{lstlisting}[upquote=true]
Seed 4020
Setting up PyTorch plugin "bias_act_plugin"... Done.
Setting up PyTorch plugin "filtered_lrelu_plugin"... Done.
\end{lstlisting}
\begin{lstlisting}[upquote=true]
...
\end{lstlisting}
\end{tcolorbox}

\subsection{Fine{-}tune an Image}%
\label{subsec:Fine{-}tuneanImage}%
If you find a seed you like, you can fine{-}tune it by directly adjusting the latent vector.  First, choose the seed to fine{-}tune.%
\index{latent vector}%
\index{vector}%
\par%
\begin{tcolorbox}[size=title,title=Code,breakable]%
\begin{lstlisting}[language=Python, upquote=true]
START_SEED = 4022

current = seed2vec(G, START_SEED)\end{lstlisting}
\end{tcolorbox}%
Next, generate and display the current vector. You will return to this point for each iteration of the finetuning.%
\index{iteration}%
\index{vector}%
\par%
\begin{tcolorbox}[size=title,title=Code,breakable]%
\begin{lstlisting}[language=Python, upquote=true]
img = generate_image(device, G, current)

SCALE = 0.5
display_image(img)\end{lstlisting}
\tcbsubtitle[before skip=\baselineskip]{Output}%
\includegraphics[width=3in]{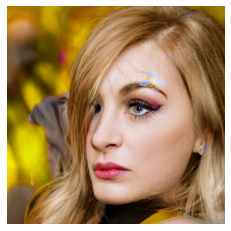}%
\end{tcolorbox}%
Choose an explore size; this is the number of different potential images chosen by moving in 10 different directions.  Run this code once and then again anytime you wish to change the ten directions you are exploring.  You might change the ten directions if you are no longer seeing improvements.%
\par%
\begin{tcolorbox}[size=title,title=Code,breakable]%
\begin{lstlisting}[language=Python, upquote=true]
EXPLORE_SIZE = 25

explore = []
for i in range(EXPLORE_SIZE):
  explore.append( np.random.rand(1, 512) - 0.5 )\end{lstlisting}
\end{tcolorbox}%
Each image displayed from running this code shows a potential direction that we can move in the latent vector.  Choose one image that you like and change MOVE\_DIRECTION to indicate this decision.  Once you rerun the code, the code will give you a new set of potential directions.  Continue this process until you have a latent vector that you like.%
\index{latent vector}%
\index{ROC}%
\index{ROC}%
\index{vector}%
\par%
\begin{tcolorbox}[size=title,title=Code,breakable]%
\begin{lstlisting}[language=Python, upquote=true]
# Choose the direction to move.  Choose -1 for the initial iteration.   
MOVE_DIRECTION = -1
SCALE = 0.5

if MOVE_DIRECTION >=0:
  current = current + explore[MOVE_DIRECTION]

for i, mv in enumerate(explore):
  print(f"Direction {i}")
  z = current + mv
  img = generate_image(device, G, z)
  display_image(img)\end{lstlisting}
\tcbsubtitle[before skip=\baselineskip]{Output}%
\includegraphics[width=3in]{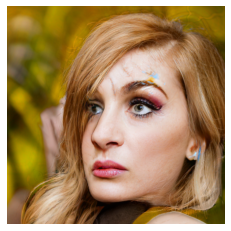}%
\begin{lstlisting}[upquote=true]
Direction 0
\end{lstlisting}
\begin{lstlisting}[upquote=true]
...
\end{lstlisting}
\end{tcolorbox}

\section{Part 7.4: GANS to Enhance Old Photographs Deoldify}%
\label{sec:Part7.4GANStoEnhanceOldPhotographsDeoldify}%
For the last two parts of this module, we will examine two applications of GANs. The first application is named%
\index{GAN}%
\href{https://deoldify.ai/}{ deoldify}%
, which uses a PyTorche{-}based GAN to transform old photographs into more modern{-}looking images. The complete%
\index{GAN}%
\index{PyTorch}%
\href{https://github.com/jantic/DeOldify}{ source code }%
to Deoldify is provided, along with several examples%
\href{https://colab.research.google.com/github/jantic/DeOldify/blob/master/ImageColorizerColab.ipynb}{ notebooks }%
upon which I based this part.%
\par%
\subsection{Install Needed Software}%
\label{subsec:InstallNeededSoftware}%
We begin by cloning the deoldify repository.%
\par%
\begin{tcolorbox}[size=title,title=Code,breakable]%
\begin{lstlisting}[language=Python, upquote=true]
!git clone https://github.com/jantic/DeOldify.git DeOldify
%cd DeOldify\end{lstlisting}
\end{tcolorbox}%
Install any additional Python packages needed.%
\index{Python}%
\par%
\begin{tcolorbox}[size=title,title=Code,breakable]%
\begin{lstlisting}[language=Python, upquote=true]
!pip install -r colab_requirements.txt\end{lstlisting}
\end{tcolorbox}%
Install the pretrained weights for deoldify.%
\par%
\begin{tcolorbox}[size=title,title=Code,breakable]%
\begin{lstlisting}[language=Python, upquote=true]
!mkdir './models/'
CMD = "wget https://data.deepai.org/deoldify/ColorizeArtistic_gen.pth"\
  " -O ./models/ColorizeArtistic_gen.pth"
!{CMD}\end{lstlisting}
\end{tcolorbox}%
The authors of deoldify suggest that you might wish to include a watermark to let others know that AI{-}enhanced this picture. The following code downloads this standard watermark. The authors describe the watermark as follows:%
\par%
"This places a watermark icon of a palette at the bottom left corner of the image. The authors intend this practice to be a standard way to convey to others viewing the image that AI colorizes it. We want to help promote this as a standard, especially as the technology continues to improve and the distinction between real and fake becomes harder to discern. This palette watermark practice was initiated and led by the MyHeritage in the MyHeritage In Color feature (which uses a newer version of DeOldify than what you're using here)."%
\index{feature}%
\par%
\begin{tcolorbox}[size=title,title=Code,breakable]%
\begin{lstlisting}[language=Python, upquote=true]
CMD = "wget https://media.githubusercontent.com/media/jantic/"\
  "DeOldify/master/resource_images/watermark.png "\
  "-O /content/DeOldify/resource_images/watermark.png"
!{CMD}\end{lstlisting}
\end{tcolorbox}

\subsection{Initialize Torch Device}%
\label{subsec:InitializeTorchDevice}%
First, we must initialize a Torch device.  If we have a GPU available, we will detect it here.  I assume that you will run this code from Google CoLab, with a GPU.  It is possible to run this code from a local GPU; however, some modification will be necessary.%
\index{GPU}%
\index{GPU}%
\index{SOM}%
\par%
\begin{tcolorbox}[size=title,title=Code,breakable]%
\begin{lstlisting}[language=Python, upquote=true]
import sys

#NOTE:  This must be the first call in order to work properly!
from deoldify import device
from deoldify.device_id import DeviceId
#choices:  CPU, GPU0...GPU7
device.set(device=DeviceId.GPU0)

import torch

if not torch.cuda.is_available():
  print('GPU not available.')
else:
  print('Using GPU.')\end{lstlisting}
\tcbsubtitle[before skip=\baselineskip]{Output}%
\begin{lstlisting}[upquote=true]
Using GPU.
\end{lstlisting}
\end{tcolorbox}%
We can now call the model. I will enhance an image from my childhood, probably taken in the late 1970s. The picture shows three miniature schnauzers. My childhood dog (Scooby) is on the left, followed by his mom and sister. Overall, a stunning improvement. However, the red in the fire engine riding toy is lost, and the red color of the picnic table where the three dogs were sitting.%
\index{model}%
\par%
\begin{tcolorbox}[size=title,title=Code,breakable]%
\begin{lstlisting}[language=Python, upquote=true]
import fastai
from deoldify.visualize import *
import warnings
from urllib.parse import urlparse
import os

warnings.filterwarnings("ignore", category=UserWarning, 
          message=".*?Your .*? set is empty.*?")

URL = 'https://raw.githubusercontent.com/jeffheaton/'\
  't81_558_deep_learning/master/photos/scooby_family.jpg' 

!wget {URL}

a = urlparse(URL)
before_file = os.path.basename(a.path)

RENDER_FACTOR = 35  
WATERMARK = False

colorizer = get_image_colorizer(artistic=True)

after_image = colorizer.get_transformed_image(
    before_file, render_factor=RENDER_FACTOR, 
    watermarked=WATERMARK)
#print("Starting image:")\end{lstlisting}
\end{tcolorbox}%
You can see the starting image here.%
\par%
\begin{tcolorbox}[size=title,title=Code,breakable]%
\begin{lstlisting}[language=Python, upquote=true]
from IPython import display
display.Image(URL)\end{lstlisting}
\tcbsubtitle[before skip=\baselineskip]{Output}%
\includegraphics[width=3in]{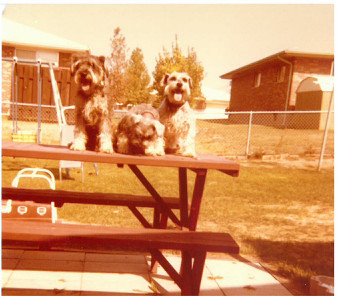}%
\end{tcolorbox}%
You can see the deoldify version here. Please note that these two images will look similar in a black and white book. To see it in color, visit this%
\href{https://github.com/jeffheaton/t81_558_deep_learning/blob/master/t81_558_class_07_4_deoldify.ipynb}{ link}%
.%
\par%
\begin{tcolorbox}[size=title,title=Code,breakable]%
\begin{lstlisting}[language=Python, upquote=true]
after_image\end{lstlisting}
\tcbsubtitle[before skip=\baselineskip]{Output}%
\includegraphics[width=3in]{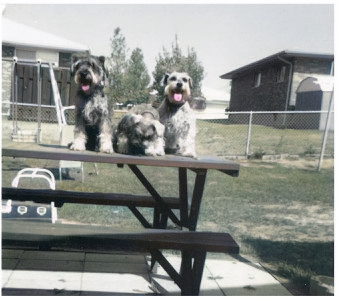}%
\end{tcolorbox}

\section{Part 7.5: GANs for Tabular Synthetic Data Generation}%
\label{sec:Part7.5GANsforTabularSyntheticDataGeneration}%
Typically GANs are used to generate images. However, we can also generate tabular data from a GAN. In this part, we will use the Python tabgan utility to create fake data from tabular data. Specifically, we will use the Auto MPG dataset to train a GAN to generate fake cars.%
\index{dataset}%
\index{GAN}%
\index{Python}%
\index{tabular data}%
\href{https://arxiv.org/pdf/2010.00638.pdf}{ Cite:ashrapov2020tabular}%
\par%
\subsection{Installing Tabgan}%
\label{subsec:InstallingTabgan}%
Pytorch is the foundation of the tabgan neural network utility. The following code installs the needed software to run tabgan in Google Colab.%
\index{GAN}%
\index{neural network}%
\index{PyTorch}%
\par%
\begin{tcolorbox}[size=title,title=Code,breakable]%
\begin{lstlisting}[language=Python, upquote=true]
CMD = "wget https://raw.githubusercontent.com/Diyago/"\
  "GAN-for-tabular-data/master/requirements.txt"

!{CMD}
!pip install -r requirements.txt
!pip install tabgan\end{lstlisting}
\end{tcolorbox}%
Note, after installing; you may see this message:%
\par%
\begin{itemize}[noitemsep]%
\item%
You must restart the runtime in order to use newly installed versions.%
\end{itemize}%
If so, click the "restart runtime" button just under the message. Then rerun this notebook, and you should not receive further issues.%
\par

\subsection{Loading the Auto MPG Data and Training a Neural Network}%
\label{subsec:LoadingtheAutoMPGDataandTrainingaNeuralNetwork}%
We will begin by generating fake data for the Auto MPG dataset we have previously seen. The tabgan library can generate categorical (textual) and continuous (numeric) data. However, it cannot generate unstructured data, such as the name of the automobile. Car names, such as "AMC Rebel SST" cannot be replicated by the GAN, because every row has a different car name; it is a textual but non{-}categorical value.%
\index{categorical}%
\index{continuous}%
\index{dataset}%
\index{GAN}%
\par%
The following code is similar to what we have seen before. We load the AutoMPG dataset. The tabgan library requires Pandas dataframe to train. Because of this, we keep both the Pandas and Numpy values.%
\index{dataset}%
\index{GAN}%
\index{NumPy}%
\par%
\begin{tcolorbox}[size=title,title=Code,breakable]%
\begin{lstlisting}[language=Python, upquote=true]
from tensorflow.keras.models import Sequential
from tensorflow.keras.layers import Dense, Activation
from tensorflow.keras.callbacks import EarlyStopping
from sklearn.model_selection import train_test_split
import pandas as pd
import io
import os
import requests
import numpy as np
from sklearn import metrics

df = pd.read_csv(
    "https://data.heatonresearch.com/data/t81-558/auto-mpg.csv", 
    na_values=['NA', '?'])

COLS_USED = ['cylinders', 'displacement', 'horsepower', 'weight', 
          'acceleration', 'year', 'origin','mpg']
COLS_TRAIN = ['cylinders', 'displacement', 'horsepower', 'weight', 
          'acceleration', 'year', 'origin']

df = df[COLS_USED]

# Handle missing value
df['horsepower'] = df['horsepower'].fillna(df['horsepower'].median())


# Split into training and test sets
df_x_train, df_x_test, df_y_train, df_y_test = train_test_split(
    df.drop("mpg", axis=1),
    df["mpg"],
    test_size=0.20,
    #shuffle=False,
    random_state=42,
)

# Create dataframe versions for tabular GAN
df_x_test, df_y_test = df_x_test.reset_index(drop=True), \
  df_y_test.reset_index(drop=True)
df_y_train = pd.DataFrame(df_y_train)
df_y_test = pd.DataFrame(df_y_test)

# Pandas to Numpy
x_train = df_x_train.values
x_test = df_x_test.values
y_train = df_y_train.values
y_test = df_y_test.values

# Build the neural network
model = Sequential()
# Hidden 1
model.add(Dense(50, input_dim=x_train.shape[1], activation='relu')) 
model.add(Dense(25, activation='relu')) # Hidden 2
model.add(Dense(12, activation='relu')) # Hidden 2
model.add(Dense(1)) # Output
model.compile(loss='mean_squared_error', optimizer='adam')

monitor = EarlyStopping(monitor='val_loss', min_delta=1e-3, 
        patience=5, verbose=1, mode='auto',
        restore_best_weights=True)
model.fit(x_train,y_train,validation_data=(x_test,y_test),
        callbacks=[monitor], verbose=2,epochs=1000)\end{lstlisting}
\end{tcolorbox}%
We now evaluate the trained neural network to see the RMSE. We will use this trained neural network to compare the accuracy between the original data and the GAN{-}generated data. We will later see that you can use such comparisons for anomaly detection. We can use this technique can be used for security systems. If a neural network trained on original data does not perform well on new data, then the new data may be suspect or fake.%
\index{GAN}%
\index{MSE}%
\index{neural network}%
\index{RMSE}%
\index{RMSE}%
\par%
\begin{tcolorbox}[size=title,title=Code,breakable]%
\begin{lstlisting}[language=Python, upquote=true]
pred = model.predict(x_test)
score = np.sqrt(metrics.mean_squared_error(pred,y_test))
print("Final score (RMSE): {}".format(score))\end{lstlisting}
\tcbsubtitle[before skip=\baselineskip]{Output}%
\begin{lstlisting}[upquote=true]
Final score (RMSE): 4.33633936452545
\end{lstlisting}
\end{tcolorbox}

\subsection{Training a GAN for Auto MPG}%
\label{subsec:TrainingaGANforAutoMPG}%
Next, we will train the GAN to generate fake data from the original MPG data. There are quite a few options that you can fine{-}tune for the GAN. The example presented here uses most of the default values. These are the usual hyperparameters that must be tuned for any model and require some experimentation for optimal results. To learn more about tabgab refer to its paper or this%
\index{GAN}%
\index{hyperparameter}%
\index{model}%
\index{parameter}%
\index{SOM}%
\href{https://towardsdatascience.com/review-of-gans-for-tabular-data-a30a2199342}{ Medium article}%
, written by the creator of tabgan.%
\index{GAN}%
\par%
\begin{tcolorbox}[size=title,title=Code,breakable]%
\begin{lstlisting}[language=Python, upquote=true]
from tabgan.sampler import GANGenerator
import pandas as pd
import numpy as np
from sklearn.model_selection import train_test_split

gen_x, gen_y = GANGenerator(gen_x_times=1.1, cat_cols=None,
           bot_filter_quantile=0.001, top_filter_quantile=0.999, \
              is_post_process=True,
           adversarial_model_params={
               "metrics": "rmse", "max_depth": 2, "max_bin": 100, 
               "learning_rate": 0.02, "random_state": \
                42, "n_estimators": 500,
           }, pregeneration_frac=2, only_generated_data=False,\
           gan_params = {"batch_size": 500, "patience": 25, \
          "epochs" : 500,}).generate_data_pipe(df_x_train, df_y_train,\
          df_x_test, deep_copy=True, only_adversarial=False, \
          use_adversarial=True)\end{lstlisting}
\tcbsubtitle[before skip=\baselineskip]{Output}%
\begin{lstlisting}[upquote=true]
Fitting CTGAN transformers for each column:   0%|          | 0/8
[00:00<?, ?it/s]Training CTGAN, epochs::   0%|          | 0/500
[00:00<?, ?it/s]
\end{lstlisting}
\end{tcolorbox}%
Note: if you receive an error running the above code, you likely need to restart the runtime. You should have a "restart runtime" button in the output from the second cell. Once you restart the runtime, rerun all of the cells. This step is necessary as tabgan requires specific versions of some packages.%
\index{error}%
\index{GAN}%
\index{output}%
\index{SOM}%
\par

\subsection{Evaluating the GAN Results}%
\label{subsec:EvaluatingtheGANResults}%
If we display the results, we can see that the GAN{-}generated data looks similar to the original. Some values, typically whole numbers in the original data, have fractional values in the synthetic data.%
\index{GAN}%
\index{SOM}%
\par%
\begin{tcolorbox}[size=title,title=Code,breakable]%
\begin{lstlisting}[language=Python, upquote=true]
gen_x\end{lstlisting}
\tcbsubtitle[before skip=\baselineskip]{Output}%
\begin{tabular}[hbt!]{l|l|l|l|l|l|l|l}%
\hline%
&cylinders&displacement&horsepower&weight&acceleration&year&origin\\%
\hline%
0&5&296.949632&106.872450&2133&18.323035&73&2\\%
1&5&247.744505&97.532052&2233&19.490136&75&2\\%
2&4&259.648421&108.111921&2424&19.898952&79&3\\%
3&5&319.208637&93.764364&2054&19.420225&78&3\\%
4&4&386.237667&129.837418&1951&20.989091&82&2\\%
...&...&...&...&...&...&...&...\\%
542&8&304.000000&150.000000&3672&11.500000&72&1\\%
543&8&304.000000&150.000000&3433&12.000000&70&1\\%
544&4&98.000000&80.000000&2164&15.000000&72&1\\%
545&4&97.500000&80.000000&2126&17.000000&72&1\\%
546&5&138.526374&68.958515&2497&13.495784&71&1\\%
\hline%
\end{tabular}%
\vspace{2mm}%
\end{tcolorbox}%
Finally, we present the synthetic data to the previously trained neural network to see how accurately we can predict the synthetic targets.  As we can see, you lose some RMSE accuracy by going to synthetic data.%
\index{MSE}%
\index{neural network}%
\index{predict}%
\index{RMSE}%
\index{RMSE}%
\index{SOM}%
\par%
\begin{tcolorbox}[size=title,title=Code,breakable]%
\begin{lstlisting}[language=Python, upquote=true]
# Predict
pred = model.predict(gen_x.values)
score = np.sqrt(metrics.mean_squared_error(pred,gen_y.values))
print("Final score (RMSE): {}".format(score))\end{lstlisting}
\tcbsubtitle[before skip=\baselineskip]{Output}%
\begin{lstlisting}[upquote=true]
Final score (RMSE): 9.083745225633098
\end{lstlisting}
\end{tcolorbox}

\chapter{Kaggle Data Sets}%
\label{chap:KaggleDataSets}%
\section{Part 8.1: Introduction to Kaggle}%
\label{sec:Part8.1IntroductiontoKaggle}%
\href{http://www.kaggle.com}{Kaggle }%
runs competitions where data scientists compete to provide the best model to fit the data. A simple project to get started with Kaggle is the%
\index{data scientist}%
\index{Kaggle}%
\index{model}%
\href{https://www.kaggle.com/c/titanic-gettingStarted}{ Titanic data set}%
. Most Kaggle competitions end on a specific date. Website organizers have scheduled the Titanic competition to end on December 31, 20xx (with the year usually rolling forward). However, they have already extended the deadline several times, and an extension beyond 2014 is also possible. Second, the Titanic data set is considered a tutorial data set. There is no prize, and your score in the competition does not count towards becoming a Kaggle Master.%
\index{GAN}%
\index{Kaggle}%
\par%
\subsection{Kaggle Ranks}%
\label{subsec:KaggleRanks}%
You achieve Kaggle ranks by earning gold, silver, and bronze medals.%
\index{Kaggle}%
\par%
\begin{itemize}[noitemsep]%
\item%
\href{https://www.kaggle.com/rankings}{Kaggle Top Users}%
\item%
\href{https://www.kaggle.com/stasg7}{Current Top Kaggle User's Profile Page}%
\item%
\href{https://www.kaggle.com/jeffheaton}{Jeff Heaton's (your instructor) Kaggle Profile}%
\item%
\href{https://www.kaggle.com/progression}{Current Kaggle Ranking System}%
\end{itemize}

\subsection{Typical Kaggle Competition}%
\label{subsec:TypicalKaggleCompetition}%
A typical Kaggle competition will have several components.  Consider the Titanic tutorial:%
\index{Kaggle}%
\par%
\begin{itemize}[noitemsep]%
\item%
\href{https://www.kaggle.com/c/titanic}{Competition Summary Page}%
\item%
\href{https://www.kaggle.com/c/titanic/data}{Data Page}%
\item%
\href{https://www.kaggle.com/c/titanic/details/evaluation}{Evaluation Description Page}%
\item%
\href{https://www.kaggle.com/c/titanic/leaderboard}{Leaderboard}%
\end{itemize}

\subsection{How Kaggle Competition Scoring}%
\begin{figure}[h]%
\centering%
\includegraphics[width=4in]{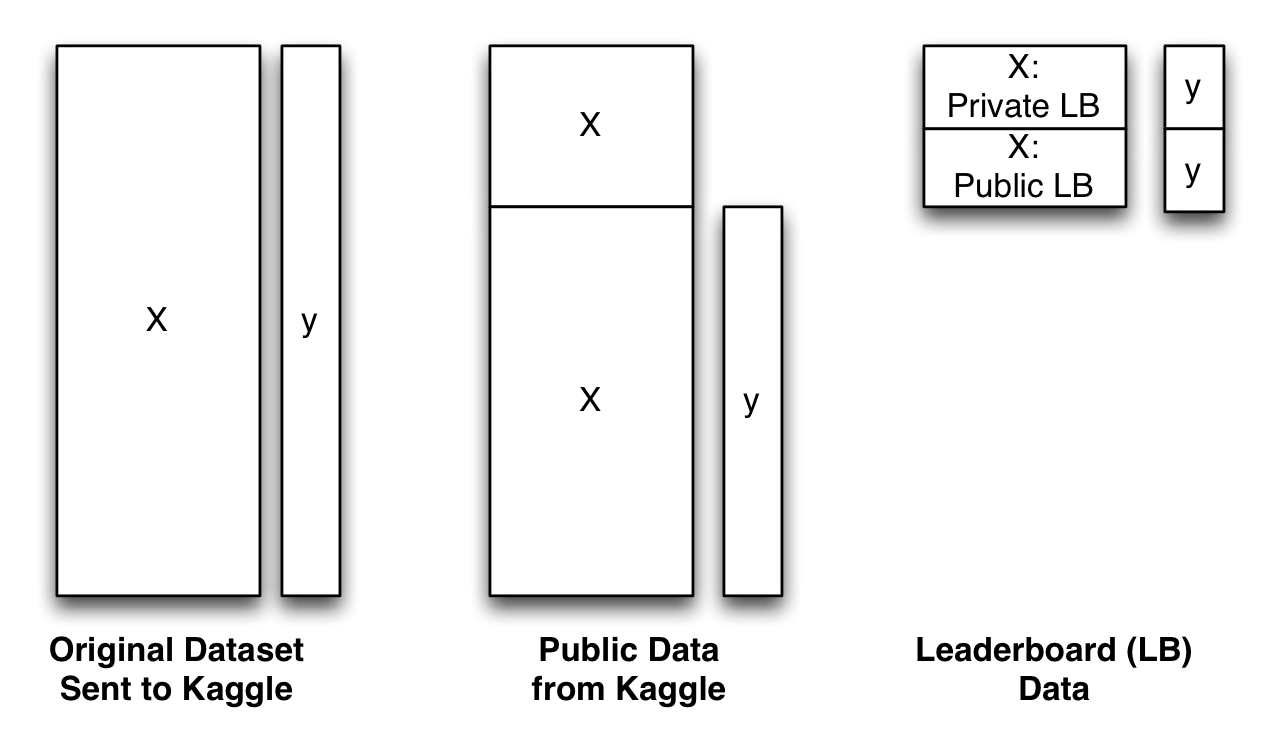}%
\caption{How Kaggle Competition Scoring}%
\label{8.SCORE}%
\end{figure}

\label{subsec:HowKaggleCompetitionScoring}%
Kaggle is provided with a data set by the competition sponsor, as seen in Figure \ref{8.SCORE}. Kaggle divides this data set as follows:%
\index{Kaggle}%
\par%
\begin{itemize}[noitemsep]%
\item%
\textbf{Complete Data Set }%
{-} This is the complete data set.%
\begin{itemize}[noitemsep]%
\item%
\textbf{Training Data Set }%
{-} This dataset provides both the inputs and the outcomes for the training portion of the data set.%
\index{dataset}%
\index{input}%
\index{training}%
\item%
\textbf{Test Data Set }%
{-} This dataset provides the complete test data; however, it does not give the outcomes. Your submission file should contain the predicted results for this data set.%
\index{dataset}%
\index{predict}%
\begin{itemize}[noitemsep]%
\item%
\textbf{Public Leaderboard }%
{-} Kaggle does not tell you what part of the test data set contributes to the public leaderboard. Your public score is calculated based on this part of the data set.%
\index{calculated}%
\index{Kaggle}%
\item%
\textbf{Private Leaderboard }%
{-} Likewise, Kaggle does not tell you what part of the test data set contributes to the public leaderboard. Your final score/rank is calculated based on this part. You do not see your private leaderboard score until the end.%
\index{calculated}%
\index{Kaggle}%
\end{itemize}%
\end{itemize}%
\end{itemize}%

\par

\subsection{Preparing a Kaggle Submission}%
\label{subsec:PreparingaKaggleSubmission}%
You do not submit the code to your solution to Kaggle. For competitions, you are scored entirely on the accuracy of your submission file. A Kaggle submission file is always a CSV file that contains the%
\index{CSV}%
\index{Kaggle}%
\textbf{ Id }%
of the row you are predicting and the answer. For the titanic competition, a submission file looks something like this:%
\index{predict}%
\index{SOM}%
\par%
\begin{tcolorbox}[size=title,breakable]%
\begin{lstlisting}[upquote=true]
PassengerId,Survived
892,0
893,1
894,1
895,0
896,0
897,1
...
\end{lstlisting}
\end{tcolorbox}%
The above file states the prediction for each of the various passengers. You should only predict on ID's that are in the test file. Likewise, you should render a prediction for every row in the test file. Some competitions will have different formats for their answers. For example, a multi{-}classification will usually have a column for each class and your predictions for each class.%
\index{classification}%
\index{predict}%
\index{SOM}%
\par

\subsection{Select Kaggle Competitions}%
\label{subsec:SelectKaggleCompetitions}%
There have been many exciting competitions on Kaggle; these are some of my favorites. Some select predictive modeling competitions which use tabular data include:%
\index{Kaggle}%
\index{model}%
\index{predict}%
\index{SOM}%
\index{tabular data}%
\par%
\begin{itemize}[noitemsep]%
\item%
\href{https://www.kaggle.com/c/otto-group-product-classification-challenge}{Otto Group Product Classification Challenge}%
\item%
\href{https://www.kaggle.com/c/galaxy-zoo-the-galaxy-challenge}{Galaxy Zoo {-} The Galaxy Challenge}%
\item%
\href{https://www.kaggle.com/c/pf2012-diabetes}{Practice Fusion Diabetes Classification}%
\item%
\href{https://www.kaggle.com/c/bioresponse}{Predicting a Biological Response}%
\end{itemize}%
Many Kaggle competitions include computer vision datasets, such as:%
\index{computer vision}%
\index{dataset}%
\index{Kaggle}%
\par%
\begin{itemize}[noitemsep]%
\item%
\href{https://www.kaggle.com/c/diabetic-retinopathy-detection}{Diabetic Retinopathy Detection}%
\item%
\href{https://www.kaggle.com/c/dogs-vs-cats}{Cats vs Dogs}%
\item%
\href{https://www.kaggle.com/c/state-farm-distracted-driver-detection}{State Farm Distracted Driver Detection}%
\end{itemize}

\subsection{Module 8 Assignment}%
\label{subsec:Module8Assignment}%
You can find the first assignment here:%
\href{https://github.com/jeffheaton/t81_558_deep_learning/blob/master/assignments/assignment_yourname_class8.ipynb}{ assignment 8}%
\par

\section{Part 8.2: Building Ensembles with Scikit{-}Learn and Keras}%
\label{sec:Part8.2BuildingEnsembleswithScikit{-}LearnandKeras}%
\subsection{Evaluating Feature Importance}%
\label{subsec:EvaluatingFeatureImportance}%
Feature importance tells us how important each feature (from the feature/import vector) is to predicting a neural network or another model. There are many different ways to evaluate the feature importance of neural networks. The following paper presents an excellent (and readable) overview of the various means of assessing the significance of neural network inputs/features.%
\index{feature}%
\index{input}%
\index{model}%
\index{neural network}%
\index{predict}%
\index{vector}%
\par%
\begin{itemize}[noitemsep]%
\item%
An accurate comparison of methods for quantifying variable importance in artificial neural networks using simulated data%
\index{neural network}%
\cite{olden2004accurate}%
.%
\textit{ Ecological Modelling}%
, 178(3), 389{-}397.%
\end{itemize}%
In summary, the following methods are available to neural networks:%
\index{neural network}%
\par%
\begin{itemize}[noitemsep]%
\item%
Connection Weights Algorithm%
\index{algorithm}%
\index{connection}%
\item%
Partial Derivatives%
\index{derivative}%
\index{partial derivative}%
\item%
Input Perturbation%
\index{input}%
\index{perturb}%
\item%
Sensitivity Analysis%
\index{sensitivity}%
\item%
Forward Stepwise Addition%
\item%
Improved Stepwise Selection 1%
\item%
Backward Stepwise Elimination%
\item%
Improved Stepwise Selection%
\end{itemize}%
For this chapter, we will use the input Perturbation feature ranking algorithm. This algorithm will work with any regression or classification network. In the next section, I provide an implementation of the input perturbation algorithm for scikit{-}learn. This code implements a function below that will work with any scikit{-}learn model.%
\index{algorithm}%
\index{classification}%
\index{feature}%
\index{input}%
\index{model}%
\index{perturb}%
\index{regression}%
\par%
\href{https://en.wikipedia.org/wiki/Leo_Breiman}{Leo Breiman }%
provided this algorithm in his seminal paper on random forests.%
\index{algorithm}%
\index{random}%
\href{https://www.stat.berkeley.edu/~breiman/randomforest2001.pdf}{ {[}Citebreiman2001random:{]} }%
Although he presented this algorithm in conjunction with random forests, it is model{-}independent and appropriate for any supervised learning model.  This algorithm, known as the input perturbation algorithm, works by evaluating a trained model's accuracy with each input individually shuffled from a data set. Shuffling an input causes it to become useless{-}{-}{-}effectively removing it from the model. More important inputs will produce a less accurate score when they are removed by shuffling them. This process makes sense because important features will contribute to the model's accuracy. I first presented the TensorFlow implementation of this algorithm in the following paper.%
\index{algorithm}%
\index{feature}%
\index{input}%
\index{learning}%
\index{model}%
\index{perturb}%
\index{random}%
\index{ROC}%
\index{ROC}%
\index{TensorFlow}%
\par%
\begin{itemize}[noitemsep]%
\item%
Early stabilizing feature importance for TensorFlow deep neural networks%
\index{feature}%
\index{neural network}%
\index{TensorFlow}%
\cite{heaton2017early}%
\end{itemize}%
This algorithm will use log loss to evaluate a classification problem and RMSE for regression.%
\index{algorithm}%
\index{classification}%
\index{MSE}%
\index{regression}%
\index{RMSE}%
\index{RMSE}%
\par%
\begin{tcolorbox}[size=title,title=Code,breakable]%
\begin{lstlisting}[language=Python, upquote=true]
from sklearn import metrics
import scipy as sp
import numpy as np
import math
from sklearn import metrics

def perturbation_rank(model, x, y, names, regression):
    errors = []

    for i in range(x.shape[1]):
        hold = np.array(x[:, i])
        np.random.shuffle(x[:, i])
        
        if regression:
            pred = model.predict(x)
            error = metrics.mean_squared_error(y, pred)
        else:
            pred = model.predict(x)
            error = metrics.log_loss(y, pred)
            
        errors.append(error)
        x[:, i] = hold
        
    max_error = np.max(errors)
    importance = [e/max_error for e in errors]

    data = {'name':names,'error':errors,'importance':importance}
    result = pd.DataFrame(data, columns = ['name','error','importance'])
    result.sort_values(by=['importance'], ascending=[0], inplace=True)
    result.reset_index(inplace=True, drop=True)
    return result\end{lstlisting}
\end{tcolorbox}

\subsection{Classification and Input Perturbation Ranking}%
\label{subsec:ClassificationandInputPerturbationRanking}%
We now look at the code to perform perturbation ranking for a classification neural network.  The implementation technique is slightly different for classification vs. regression, so I must provide two different implementations.  The primary difference between classification and regression is how we evaluate the accuracy of the neural network in each of these two network types.  We will use the Root Mean Square (RMSE) error calculation, whereas we will use log loss for classification.%
\index{classification}%
\index{error}%
\index{MSE}%
\index{neural network}%
\index{perturb}%
\index{regression}%
\index{RMSE}%
\index{RMSE}%
\par%
The code presented below creates a classification neural network that will predict the classic iris dataset.%
\index{classification}%
\index{dataset}%
\index{iris}%
\index{neural network}%
\index{predict}%
\par%
\begin{tcolorbox}[size=title,title=Code,breakable]%
\begin{lstlisting}[language=Python, upquote=true]
import pandas as pd
import io
import requests
import numpy as np
from sklearn import metrics
from tensorflow.keras.models import Sequential
from tensorflow.keras.layers import Dense, Activation
from tensorflow.keras.callbacks import EarlyStopping
from sklearn.model_selection import train_test_split

df = pd.read_csv(
    "https://data.heatonresearch.com/data/t81-558/iris.csv", 
    na_values=['NA', '?'])

# Convert to numpy - Classification
x = df[['sepal_l', 'sepal_w', 'petal_l', 'petal_w']].values
dummies = pd.get_dummies(df['species']) # Classification
species = dummies.columns
y = dummies.values

# Split into train/test
x_train, x_test, y_train, y_test = train_test_split(    
    x, y, test_size=0.25, random_state=42)

# Build neural network
model = Sequential()
model.add(Dense(50, input_dim=x.shape[1], activation='relu')) # Hidden 1
model.add(Dense(25, activation='relu')) # Hidden 2
model.add(Dense(y.shape[1],activation='softmax')) # Output
model.compile(loss='categorical_crossentropy', optimizer='adam')
model.fit(x_train,y_train,verbose=2,epochs=100)\end{lstlisting}
\end{tcolorbox}%
Next, we evaluate the accuracy of the trained model.  Here we see that the neural network performs great, with an accuracy of 1.0.  We might fear overfitting with such high accuracy for a more complex dataset.  However, for this example, we are more interested in determining the importance of each column.%
\index{dataset}%
\index{model}%
\index{neural network}%
\index{overfitting}%
\par%
\begin{tcolorbox}[size=title,title=Code,breakable]%
\begin{lstlisting}[language=Python, upquote=true]
from sklearn.metrics import accuracy_score

pred = model.predict(x_test)
predict_classes = np.argmax(pred,axis=1)
expected_classes = np.argmax(y_test,axis=1)
correct = accuracy_score(expected_classes,predict_classes)
print(f"Accuracy: {correct}")\end{lstlisting}
\tcbsubtitle[before skip=\baselineskip]{Output}%
\begin{lstlisting}[upquote=true]
Accuracy: 1.0
\end{lstlisting}
\end{tcolorbox}%
We are now ready to call the input perturbation algorithm.  First, we extract the column names and remove the target column.  The target column is not important, as it is the objective, not one of the inputs.  In supervised learning, the target is of the utmost importance.%
\index{algorithm}%
\index{input}%
\index{learning}%
\index{perturb}%
\par%
We can see the importance displayed in the following table.  The most important column is always 1.0, and lessor columns will continue in a downward trend.  The least important column will have the lowest rank.%
\par%
\begin{tcolorbox}[size=title,title=Code,breakable]%
\begin{lstlisting}[language=Python, upquote=true]
# Rank the features
from IPython.display import display, HTML

names = list(df.columns) # x+y column names
names.remove("species") # remove the target(y)
rank = perturbation_rank(model, x_test, y_test, names, False)
display(rank)\end{lstlisting}
\tcbsubtitle[before skip=\baselineskip]{Output}%
\begin{tabular}[hbt!]{l|l|l|l}%
\hline%
&name&error&importance\\%
\hline%
0&petal\_l&2.609378&1.000000\\%
1&petal\_w&0.480387&0.184100\\%
2&sepal\_l&0.223239&0.085553\\%
3&sepal\_w&0.128518&0.049252\\%
\hline%
\end{tabular}%
\vspace{2mm}%
\end{tcolorbox}

\subsection{Regression and Input Perturbation Ranking}%
\label{subsec:RegressionandInputPerturbationRanking}%
We now see how to use input perturbation ranking for a regression neural network.  We will use the MPG dataset as a demonstration.  The code below loads the MPG dataset and creates a regression neural network for this dataset.  The code trains the neural network and calculates an RMSE evaluation.%
\index{dataset}%
\index{input}%
\index{MSE}%
\index{neural network}%
\index{perturb}%
\index{regression}%
\index{RMSE}%
\index{RMSE}%
\par%
\begin{tcolorbox}[size=title,title=Code,breakable]%
\begin{lstlisting}[language=Python, upquote=true]
from tensorflow.keras.models import Sequential
from tensorflow.keras.layers import Dense, Activation
from sklearn.model_selection import train_test_split
import pandas as pd
import io
import os
import requests
import numpy as np
from sklearn import metrics

save_path = "."

df = pd.read_csv(
    "https://data.heatonresearch.com/data/t81-558/auto-mpg.csv", 
    na_values=['NA', '?'])

cars = df['name']

# Handle missing value
df['horsepower'] = df['horsepower'].fillna(df['horsepower'].median())

# Pandas to Numpy
x = df[['cylinders', 'displacement', 'horsepower', 'weight',
       'acceleration', 'year', 'origin']].values
y = df['mpg'].values # regression

# Split into train/test
x_train, x_test, y_train, y_test = train_test_split(    
    x, y, test_size=0.25, random_state=42)

# Build the neural network
model = Sequential()
model.add(Dense(25, input_dim=x.shape[1], activation='relu')) # Hidden 1
model.add(Dense(10, activation='relu')) # Hidden 2
model.add(Dense(1)) # Output
model.compile(loss='mean_squared_error', optimizer='adam')
model.fit(x_train,y_train,verbose=2,epochs=100)

# Predict
pred = model.predict(x)\end{lstlisting}
\end{tcolorbox}%
Just as before, we extract the column names and discard the target.  We can now create a ranking of the importance of each of the input features.  The feature with a ranking of 1.0 is the most important.%
\index{feature}%
\index{input}%
\par%
\begin{tcolorbox}[size=title,title=Code,breakable]%
\begin{lstlisting}[language=Python, upquote=true]
# Rank the features
from IPython.display import display, HTML

names = list(df.columns) # x+y column names
names.remove("name")
names.remove("mpg") # remove the target(y)
rank = perturbation_rank(model, x_test, y_test, names, True)
display(rank)\end{lstlisting}
\tcbsubtitle[before skip=\baselineskip]{Output}%
\begin{tabular}[hbt!]{l|l|l|l}%
\hline%
&name&error&importance\\%
\hline%
0&displacement&139.657598&1.000000\\%
1&acceleration&139.261508&0.997164\\%
2&origin&134.637690&0.964056\\%
3&year&134.177126&0.960758\\%
4&cylinders&132.747246&0.950519\\%
5&horsepower&121.501102&0.869993\\%
6&weight&75.244610&0.538779\\%
\hline%
\end{tabular}%
\vspace{2mm}%
\end{tcolorbox}

\subsection{Biological Response with Neural Network}%
\label{subsec:BiologicalResponsewithNeuralNetwork}%
The following sections will demonstrate how to use feature importance ranking and ensembling with a more complex dataset. Ensembling is the process where you combine multiple models for greater accuracy. Kaggle competition winners frequently make use of ensembling for high{-}ranking solutions.%
\index{dataset}%
\index{feature}%
\index{Kaggle}%
\index{model}%
\index{ROC}%
\index{ROC}%
\par%
We will use the biological response dataset, a Kaggle dataset, where there is an unusually high number of columns. Because of the large number of columns, it is essential to use feature ranking to determine the importance of these columns. We begin by loading the dataset and preprocessing. This Kaggle dataset is a binary classification problem. You must predict if certain conditions will cause a biological response.%
\index{classification}%
\index{dataset}%
\index{feature}%
\index{Kaggle}%
\index{predict}%
\index{ROC}%
\index{ROC}%
\par%
\begin{itemize}[noitemsep]%
\item%
\href{https://www.kaggle.com/c/bioresponse}{Predicting a Biological Response}%
\end{itemize}%
\begin{tcolorbox}[size=title,title=Code,breakable]%
\begin{lstlisting}[language=Python, upquote=true]
import pandas as pd
import os
import numpy as np
from sklearn import metrics
from scipy.stats import zscore
from sklearn.model_selection import KFold
from IPython.display import HTML, display

URL = "https://data.heatonresearch.com/data/t81-558/kaggle/"

df_train = pd.read_csv(
    URL+"bio_train.csv", 
    na_values=['NA', '?'])

df_test = pd.read_csv(
    URL+"bio_test.csv", 
    na_values=['NA', '?'])

activity_classes = df_train['Activity']\end{lstlisting}
\end{tcolorbox}%
A large number of columns is evident when we display the shape of the dataset.%
\index{dataset}%
\par%
\begin{tcolorbox}[size=title,title=Code,breakable]%
\begin{lstlisting}[language=Python, upquote=true]
print(df_train.shape)\end{lstlisting}
\tcbsubtitle[before skip=\baselineskip]{Output}%
\begin{lstlisting}[upquote=true]
(3751, 1777)
\end{lstlisting}
\end{tcolorbox}%
The following code constructs a classification neural network and trains it for the biological response dataset.  Once trained, the accuracy is measured.%
\index{classification}%
\index{dataset}%
\index{neural network}%
\par%
\begin{tcolorbox}[size=title,title=Code,breakable]%
\begin{lstlisting}[language=Python, upquote=true]
import os
import pandas as pd
import tensorflow as tf
from tensorflow.keras.models import Sequential
from tensorflow.keras.layers import Dense, Activation
from sklearn.model_selection import train_test_split
from tensorflow.keras.callbacks import EarlyStopping
import numpy as np
import sklearn

# Encode feature vector
# Convert to numpy - Classification
x_columns = df_train.columns.drop('Activity')
x = df_train[x_columns].values
y = df_train['Activity'].values # Classification
x_submit = df_test[x_columns].values.astype(np.float32)


# Split into train/test
x_train, x_test, y_train, y_test = train_test_split(    
    x, y, test_size=0.25, random_state=42) 

print("Fitting/Training...")
model = Sequential()
model.add(Dense(25, input_dim=x.shape[1], activation='relu'))
model.add(Dense(10))
model.add(Dense(1,activation='sigmoid'))
model.compile(loss='binary_crossentropy', optimizer='adam')
monitor = EarlyStopping(monitor='val_loss', min_delta=1e-3, 
                        patience=5, verbose=1, mode='auto')
model.fit(x_train,y_train,validation_data=(x_test,y_test),
          callbacks=[monitor],verbose=0,epochs=1000)
print("Fitting done...")

# Predict
pred = model.predict(x_test).flatten()


# Clip so that min is never exactly 0, max never 1
pred = np.clip(pred,a_min=1e-6,a_max=(1-1e-6)) 
print("Validation logloss: {}".format(
    sklearn.metrics.log_loss(y_test,pred)))

# Evaluate success using accuracy
pred = pred>0.5 # If greater than 0.5 probability, then true
score = metrics.accuracy_score(y_test, pred)
print("Validation accuracy score: {}".format(score))

# Build real submit file
pred_submit = model.predict(x_submit)

# Clip so that min is never exactly 0, max never 1 (would be a NaN score)
pred = np.clip(pred,a_min=1e-6,a_max=(1-1e-6)) 
submit_df = pd.DataFrame({'MoleculeId':[x+1 for x \
        in range(len(pred_submit))],'PredictedProbability':\
                          pred_submit.flatten()})
submit_df.to_csv("submit.csv", index=False)\end{lstlisting}
\tcbsubtitle[before skip=\baselineskip]{Output}%
\begin{lstlisting}[upquote=true]
Fitting/Training...
Epoch 7: early stopping
Fitting done...
Validation logloss: 0.5564708781752792
Validation accuracy score: 0.7515991471215352
\end{lstlisting}
\end{tcolorbox}

\subsection{What Features/Columns are Important}%
\label{subsec:WhatFeatures/ColumnsareImportant}%
The following uses perturbation ranking to evaluate the neural network.%
\index{neural network}%
\index{perturb}%
\par%
\begin{tcolorbox}[size=title,title=Code,breakable]%
\begin{lstlisting}[language=Python, upquote=true]
# Rank the features
from IPython.display import display, HTML

names = list(df_train.columns) # x+y column names
names.remove("Activity") # remove the target(y)
rank = perturbation_rank(model, x_test, y_test, names, False)
display(rank[0:10])\end{lstlisting}
\tcbsubtitle[before skip=\baselineskip]{Output}%
\begin{tabular}[hbt!]{l|l|l|l}%
\hline%
&name&error&importance\\%
\hline%
0&D27&0.603974&1.000000\\%
1&D1049&0.565997&0.937122\\%
2&D51&0.565883&0.936934\\%
3&D998&0.563872&0.933604\\%
4&D1059&0.563745&0.933394\\%
5&D961&0.563723&0.933357\\%
6&D1407&0.563532&0.933041\\%
7&D1309&0.562244&0.930908\\%
8&D1100&0.561902&0.930341\\%
9&D1275&0.561659&0.929940\\%
\hline%
\end{tabular}%
\vspace{2mm}%
\end{tcolorbox}

\subsection{Neural Network Ensemble}%
\label{subsec:NeuralNetworkEnsemble}%
A neural network ensemble combines neural network predictions with other models. The program determines the exact blend of these models by logistic regression. The following code performs this blend for a classification.  If you present the final predictions from the ensemble to Kaggle, you will see that the result is very accurate.%
\index{classification}%
\index{ensemble}%
\index{Kaggle}%
\index{model}%
\index{neural network}%
\index{predict}%
\index{regression}%
\par%
\begin{tcolorbox}[size=title,title=Code,breakable]%
\begin{lstlisting}[language=Python, upquote=true]
import numpy as np
import os
import pandas as pd
import math
from tensorflow.keras.wrappers.scikit_learn import KerasClassifier
from sklearn.neighbors import KNeighborsClassifier
from sklearn.model_selection import StratifiedKFold
from sklearn.ensemble import RandomForestClassifier 
from sklearn.ensemble import ExtraTreesClassifier
from sklearn.ensemble import GradientBoostingClassifier
from sklearn.linear_model import LogisticRegression

SHUFFLE = False
FOLDS = 10

def build_ann(input_size,classes,neurons):
    model = Sequential()
    model.add(Dense(neurons, input_dim=input_size, activation='relu'))
    model.add(Dense(1))
    model.add(Dense(classes,activation='softmax'))
    model.compile(loss='categorical_crossentropy', optimizer='adam')
    return model

def mlogloss(y_test, preds):
    epsilon = 1e-15
    sum = 0
    for row in zip(preds,y_test):
        x = row[0][row[1]]
        x = max(epsilon,x)
        x = min(1-epsilon,x)
        sum+=math.log(x)
    return( (-1/len(preds))*sum)

def stretch(y):
    return (y - y.min()) / (y.max() - y.min())


def blend_ensemble(x, y, x_submit):
    kf = StratifiedKFold(FOLDS)
    folds = list(kf.split(x,y))

    models = [
        KerasClassifier(build_fn=build_ann,neurons=20,
                    input_size=x.shape[1],classes=2),
        KNeighborsClassifier(n_neighbors=3),
        RandomForestClassifier(n_estimators=100, n_jobs=-1, 
                               criterion='gini'),
        RandomForestClassifier(n_estimators=100, n_jobs=-1, 
                               criterion='entropy'),
        ExtraTreesClassifier(n_estimators=100, n_jobs=-1, 
                             criterion='gini'),
        ExtraTreesClassifier(n_estimators=100, n_jobs=-1, 
                             criterion='entropy'),
        GradientBoostingClassifier(learning_rate=0.05, 
                subsample=0.5, max_depth=6, n_estimators=50)]

    dataset_blend_train = np.zeros((x.shape[0], len(models)))
    dataset_blend_test = np.zeros((x_submit.shape[0], len(models)))

    for j, model in enumerate(models):
        print("Model: {} : {}".format(j, model) )
        fold_sums = np.zeros((x_submit.shape[0], len(folds)))
        total_loss = 0
        for i, (train, test) in enumerate(folds):
            x_train = x[train]
            y_train = y[train]
            x_test = x[test]
            y_test = y[test]
            model.fit(x_train, y_train)
            pred = np.array(model.predict_proba(x_test))
            dataset_blend_train[test, j] = pred[:, 1]
            pred2 = np.array(model.predict_proba(x_submit))
            fold_sums[:, i] = pred2[:, 1]
            loss = mlogloss(y_test, pred)
            total_loss+=loss
            print("Fold #{}: loss={}".format(i,loss))
        print("{}: Mean loss={}".format(model.__class__.__name__,
                                        total_loss/len(folds)))
        dataset_blend_test[:, j] = fold_sums.mean(1)

    print()
    print("Blending models.")
    blend = LogisticRegression(solver='lbfgs')
    blend.fit(dataset_blend_train, y)
    return blend.predict_proba(dataset_blend_test)

if __name__ == '__main__':

    np.random.seed(42)  # seed to shuffle the train set

    print("Loading data...")
    URL = "https://data.heatonresearch.com/data/t81-558/kaggle/"

    df_train = pd.read_csv(
        URL+"bio_train.csv", 
        na_values=['NA', '?'])

    df_submit = pd.read_csv(
        URL+"bio_test.csv", 
        na_values=['NA', '?'])

    predictors = list(df_train.columns.values)
    predictors.remove('Activity')
    x = df_train[predictors].values
    y = df_train['Activity']
    x_submit = df_submit.values

    if SHUFFLE:
        idx = np.random.permutation(y.size)
        x = x[idx]
        y = y[idx]

    submit_data = blend_ensemble(x, y, x_submit)
    submit_data = stretch(submit_data)

    ####################
    # Build submit file
    ####################
    ids = [id+1 for id in range(submit_data.shape[0])]
    submit_df = pd.DataFrame({'MoleculeId': ids, 
                              'PredictedProbability': 
                              submit_data[:, 1]},
                             columns=['MoleculeId',
                            'PredictedProbability'])
    submit_df.to_csv("submit.csv", index=False)\end{lstlisting}
\end{tcolorbox}

\section{Part 8.3: Architecting Network: Hyperparameters}%
\label{sec:Part8.3ArchitectingNetworkHyperparameters}%
You have probably noticed several hyperparameters introduced previously in this course that you need to choose for your neural network. The number of layers, neuron counts per layer, layer types, and activation functions are all choices you must make to optimize your neural network. Some of the categories of hyperparameters for you to choose from coming from the following list:%
\index{activation function}%
\index{hyperparameter}%
\index{layer}%
\index{neural network}%
\index{neuron}%
\index{parameter}%
\index{SOM}%
\par%
\begin{itemize}[noitemsep]%
\item%
Number of Hidden Layers and Neuron Counts%
\index{hidden layer}%
\index{layer}%
\index{neuron}%
\item%
Activation Functions%
\index{activation function}%
\item%
Advanced Activation Functions%
\index{activation function}%
\item%
Regularization: L1, L2, Dropout%
\index{dropout}%
\index{L1}%
\index{L2}%
\index{regularization}%
\item%
Batch Normalization%
\item%
Training Parameters%
\index{parameter}%
\index{training}%
\end{itemize}%
The following sections will introduce each of these categories for Keras. While I will provide some general guidelines for hyperparameter selection, no two tasks are the same. You will benefit from experimentation with these values to determine what works best for your neural network. In the next part, we will see how machine learning can select some of these values independently.%
\index{hyperparameter}%
\index{Keras}%
\index{learning}%
\index{neural network}%
\index{parameter}%
\index{SOM}%
\par%
\subsection{Number of Hidden Layers and Neuron Counts}%
\label{subsec:NumberofHiddenLayersandNeuronCounts}%
The structure of Keras layers is perhaps the hyperparameters that most become aware of first. How many layers should you have? How many neurons are on each layer? What activation function and layer type should you use? These are all questions that come up when designing a neural network. There are many different%
\index{activation function}%
\index{hyperparameter}%
\index{Keras}%
\index{layer}%
\index{neural network}%
\index{neuron}%
\index{parameter}%
\href{https://keras.io/layers/core/}{ types of layer }%
in Keras, listed here:%
\index{Keras}%
\par%
\begin{itemize}[noitemsep]%
\item%
\textbf{Activation }%
{-} You can also add activation functions as layers.  Using the activation layer is the same as specifying the activation function as part of a Dense (or other) layer type.%
\index{activation function}%
\index{layer}%
\item%
\textbf{ActivityRegularization }%
Used to add L1/L2 regularization outside of a layer. You can specify L1 and L2 as part of a Dense (or other) layer type.%
\index{L1}%
\index{L2}%
\index{layer}%
\index{regularization}%
\item%
\textbf{Dense }%
{-} The original neural network layer type. In this layer type, every neuron connects to the next layer. The input vector is one{-}dimensional, and placing specific inputs next does not affect each other.%
\index{input}%
\index{input vector}%
\index{layer}%
\index{neural network}%
\index{neuron}%
\index{vector}%
\item%
\textbf{Dropout }%
{-} Dropout consists of randomly setting a fraction rate of input units to 0 at each update during training time, which helps prevent overfitting. Dropout only occurs during training.%
\index{dropout}%
\index{input}%
\index{overfitting}%
\index{random}%
\index{training}%
\item%
\textbf{Flatten }%
{-} Flattens the input to 1D and does not affect the batch size.%
\index{input}%
\item%
\textbf{Input }%
{-} A Keras tensor is a tensor object from the underlying back end (Theano, TensorFlow, or CNTK), which we augment with specific attributes to build a Keras by knowing the inputs and outputs of the model.%
\index{input}%
\index{Keras}%
\index{model}%
\index{output}%
\index{TensorFlow}%
\index{Theano}%
\item%
\textbf{Lambda }%
{-} Wraps arbitrary expression as a Layer object.%
\index{layer}%
\item%
\textbf{Masking }%
{-} Masks a sequence using a mask value to skip timesteps.%
\item%
\textbf{Permute }%
{-} Permutes the input dimensions according to a given pattern. Useful for tasks such as connecting RNNs and convolutional networks.%
\index{convolution}%
\index{convolutional}%
\index{input}%
\item%
\textbf{RepeatVector }%
{-} Repeats the input n times.%
\index{input}%
\item%
\textbf{Reshape }%
{-} Similar to Numpy reshapes.%
\index{NumPy}%
\item%
\textbf{SpatialDropout1D }%
{-} This version performs the same function as Dropout; however, it drops entire 1D feature maps instead of individual elements.%
\index{dropout}%
\index{feature}%
\item%
\textbf{SpatialDropout2D }%
{-} This version performs the same function as Dropout; however, it drops entire 2D feature maps instead of individual elements%
\index{dropout}%
\index{feature}%
\item%
\textbf{SpatialDropout3D }%
{-} This version performs the same function as Dropout; however, it drops entire 3D feature maps instead of individual elements.%
\index{dropout}%
\index{feature}%
\end{itemize}%
There is always trial and error for choosing a good number of neurons and hidden layers. Generally, the number of neurons on each layer will be larger closer to the hidden layer and smaller towards the output layer. This configuration gives the neural network a somewhat triangular or trapezoid appearance.%
\index{error}%
\index{hidden layer}%
\index{layer}%
\index{neural network}%
\index{neuron}%
\index{output}%
\index{output layer}%
\index{SOM}%
\par

\subsection{Activation Functions}%
\label{subsec:ActivationFunctions}%
Activation functions are a choice that you must make for each layer. Generally, you can follow this guideline:%
\index{activation function}%
\index{layer}%
\par%
\begin{itemize}[noitemsep]%
\item%
Hidden Layers {-} RELU%
\index{hidden layer}%
\index{layer}%
\index{ReLU}%
\item%
Output Layer {-} Softmax for classification, linear for regression.%
\index{classification}%
\index{layer}%
\index{linear}%
\index{output}%
\index{output layer}%
\index{regression}%
\index{softmax}%
\end{itemize}%
Some of the common activation functions in Keras are listed here:%
\index{activation function}%
\index{Keras}%
\index{SOM}%
\par%
\begin{itemize}[noitemsep]%
\item%
\textbf{softmax }%
{-} Used for multi{-}class classification.  Ensures all output neurons behave as probabilities and sum to 1.0.%
\index{classification}%
\index{neuron}%
\index{output}%
\index{output neuron}%
\item%
\textbf{elu }%
{-} Exponential linear unit.  Exponential Linear Unit or its widely known name ELU is a function that tends to converge cost to zero faster and produce more accurate results. Can produce negative outputs.%
\index{linear}%
\index{output}%
\item%
\textbf{selu }%
{-} Scaled Exponential Linear Unit (SELU), essentially%
\index{linear}%
\textbf{ elu }%
multiplied by a scaling constant.%
\item%
\textbf{softplus }%
{-} Softplus activation function. $log(exp(x) + 1)$%
\index{activation function}%
\href{https://papers.nips.cc/paper/1920-incorporating-second-order-functional-knowledge-for-better-option-pricing.pdf}{ Introduced }%
in 2001.%
\item%
\textbf{softsign }%
Softsign activation function. $x / (abs(x) + 1)$ Similar to tanh, but not widely used.%
\index{activation function}%
\item%
\textbf{relu }%
{-} Very popular neural network activation function.  Used for hidden layers, cannot output negative values. No trainable parameters.%
\index{activation function}%
\index{hidden layer}%
\index{layer}%
\index{neural network}%
\index{output}%
\index{parameter}%
\item%
\textbf{tanh }%
Classic neural network activation function, though often replaced by relu family on modern networks.%
\index{activation function}%
\index{neural network}%
\index{ReLU}%
\item%
\textbf{sigmoid }%
{-} Classic neural network activation.  Often used on output layer of a binary classifier.%
\index{layer}%
\index{neural network}%
\index{output}%
\index{output layer}%
\item%
\textbf{hard\_sigmoid }%
{-} Less computationally expensive variant of sigmoid.%
\index{sigmoid}%
\item%
\textbf{exponential }%
{-} Exponential (base e) activation function.%
\index{activation function}%
\item%
\textbf{linear }%
{-} Pass{-}through activation function. Usually used on the output layer of a regression neural network.%
\index{activation function}%
\index{layer}%
\index{neural network}%
\index{output}%
\index{output layer}%
\index{regression}%
\end{itemize}%
For more information about Keras activation functions refer to the following:%
\index{activation function}%
\index{Keras}%
\par%
\begin{itemize}[noitemsep]%
\item%
\href{https://keras.io/activations/}{Keras Activation Functions}%
\item%
\href{https://ml-cheatsheet.readthedocs.io/en/latest/activation_functions.html}{Activation Function Cheat Sheets}%
\end{itemize}

\subsection{Advanced Activation Functions}%
\label{subsec:AdvancedActivationFunctions}%
Hyperparameters are not changed when the neural network trains. You, the network designer, must define the hyperparameters. The neural network learns regular parameters during neural network training. Neural network weights are the most common type of regular parameter. The "%
\index{hyperparameter}%
\index{neural network}%
\index{parameter}%
\index{training}%
\href{https://keras.io/layers/advanced-activations/}{advanced activation functions}%
," as Keras call them, also contain parameters that the network will learn during training. These activation functions may give you better performance than RELU.%
\index{activation function}%
\index{Keras}%
\index{parameter}%
\index{ReLU}%
\index{training}%
\par%
\begin{itemize}[noitemsep]%
\item%
\textbf{LeakyReLU }%
{-} Leaky version of a Rectified Linear Unit. It allows a small gradient when the unit is not active, controlled by alpha hyperparameter.%
\index{gradient}%
\index{hyperparameter}%
\index{linear}%
\index{parameter}%
\item%
\textbf{PReLU }%
{-} Parametric Rectified Linear Unit, learns the alpha hyperparameter.%
\index{hyperparameter}%
\index{linear}%
\index{parameter}%
\end{itemize}

\subsection{Regularization: L1, L2, Dropout}%
\label{subsec:RegularizationL1,L2,Dropout}%
\begin{itemize}[noitemsep]%
\item%
\href{https://keras.io/regularizers/}{Keras Regularization}%
\item%
\href{https://keras.io/layers/core/}{Keras Dropout}%
\end{itemize}

\subsection{Batch Normalization}%
\label{subsec:BatchNormalization}%
\begin{itemize}[noitemsep]%
\item%
\href{https://keras.io/layers/normalization/}{Keras Batch Normalization}%
\par%
\item%
Ioffe, S.,  Szegedy, C. (2015).%
\href{https://arxiv.org/abs/1502.03167}{ Batch normalization: Accelerating deep network training by reducing internal covariate shift}%
.%
\textit{ arXiv preprint arXiv:1502.03167}%
.%
\par%
\end{itemize}%
Normalize the activations of the previous layer at each batch, i.e. applies a transformation that maintains the mean activation close to 0 and the activation standard deviation close to 1. Can allow learning rate to be larger.%
\index{layer}%
\index{learning}%
\index{learning rate}%
\index{standard deviation}%
\par

\subsection{Training Parameters}%
\label{subsec:TrainingParameters}%
\begin{itemize}[noitemsep]%
\item%
\href{https://keras.io/optimizers/}{Keras Optimizers}%
\par%
\item%
\textbf{Batch Size }%
{-} Usually small, such as 32 or so.%
\par%
\item%
\textbf{Learning Rate }%
{-} Usually small, 1e{-}3 or so.%
\par%
\end{itemize}

\section{Part 8.4: Bayesian Hyperparameter Optimization for Keras}%
\label{sec:Part8.4BayesianHyperparameterOptimizationforKeras}%
Bayesian Hyperparameter Optimization is a method of finding hyperparameters more efficiently than a grid search. Because each candidate set of hyperparameters requires a retraining of the neural network, it is best to keep the number of candidate sets to a minimum. Bayesian Hyperparameter Optimization achieves this by training a model to predict good candidate sets of hyperparameters.%
\index{Bayesian Hyperparameter Optimization}%
\index{hyperparameter}%
\index{model}%
\index{neural network}%
\index{optimization}%
\index{parameter}%
\index{predict}%
\index{training}%
\cite{snoek2012practical}%
\par%
\begin{itemize}[noitemsep]%
\item%
\href{https://github.com/fmfn/BayesianOptimization}{bayesian{-}optimization}%
\item%
\href{https://github.com/hyperopt/hyperopt}{hyperopt}%
\item%
\href{https://github.com/JasperSnoek/spearmint}{spearmint}%
\end{itemize}%
\begin{tcolorbox}[size=title,title=Code,breakable]%
\begin{lstlisting}[language=Python, upquote=true]
# Ignore useless W0819 warnings generated by TensorFlow 2.0.  
# Hopefully can remove this ignore in the future.
# See https://github.com/tensorflow/tensorflow/issues/31308
import logging, os
logging.disable(logging.WARNING)
os.environ["TF_CPP_MIN_LOG_LEVEL"] = "3"

import pandas as pd
from scipy.stats import zscore

# Read the data set
df = pd.read_csv(
    "https://data.heatonresearch.com/data/t81-558/jh-simple-dataset.csv",
    na_values=['NA','?'])

# Generate dummies for job
df = pd.concat([df,pd.get_dummies(df['job'],prefix="job")],axis=1)
df.drop('job', axis=1, inplace=True)

# Generate dummies for area
df = pd.concat([df,pd.get_dummies(df['area'],prefix="area")],axis=1)
df.drop('area', axis=1, inplace=True)

# Missing values for income
med = df['income'].median()
df['income'] = df['income'].fillna(med)

# Standardize ranges
df['income'] = zscore(df['income'])
df['aspect'] = zscore(df['aspect'])
df['save_rate'] = zscore(df['save_rate'])
df['age'] = zscore(df['age'])
df['subscriptions'] = zscore(df['subscriptions'])

# Convert to numpy - Classification
x_columns = df.columns.drop('product').drop('id')
x = df[x_columns].values
dummies = pd.get_dummies(df['product']) # Classification
products = dummies.columns
y = dummies.values\end{lstlisting}
\end{tcolorbox}%
Now that we've preprocessed the data, we can begin the hyperparameter optimization.  We start by creating a function that generates the model based on just three parameters.  Bayesian optimization works on a vector of numbers, not on a problematic notion like how many layers and neurons are on each layer.  To represent this complex neuron structure as a vector, we use several numbers to describe this structure.%
\index{hyperparameter}%
\index{layer}%
\index{model}%
\index{neuron}%
\index{optimization}%
\index{parameter}%
\index{ROC}%
\index{ROC}%
\index{vector}%
\par%
\begin{itemize}[noitemsep]%
\item%
\textbf{dropout }%
{-} The dropout percent for each layer.%
\index{dropout}%
\index{layer}%
\item%
\textbf{neuronPct }%
{-} What percent of our fixed 5,000 maximum number of neurons do we wish to use?  This parameter specifies the total count of neurons in the entire network.%
\index{neuron}%
\index{parameter}%
\item%
\textbf{neuronShrink }%
{-} Neural networks usually start with more neurons on the first hidden layer and then decrease this count for additional layers.  This percent specifies how much to shrink subsequent layers based on the previous layer.  We stop adding more layers once we run out of neurons (the count specified by neuronPct).%
\index{hidden layer}%
\index{layer}%
\index{neural network}%
\index{neuron}%
\end{itemize}%
These three numbers define the structure of the neural network.  The commends in the below code show exactly how the program constructs the network.%
\index{neural network}%
\par%
\begin{tcolorbox}[size=title,title=Code,breakable]%
\begin{lstlisting}[language=Python, upquote=true]
import pandas as pd
import os
import numpy as np
import time
import tensorflow.keras.initializers
import statistics
import tensorflow.keras
from sklearn import metrics
from sklearn.model_selection import StratifiedKFold
from tensorflow.keras.models import Sequential
from tensorflow.keras.layers import Dense, Activation, Dropout, InputLayer
from tensorflow.keras import regularizers
from tensorflow.keras.callbacks import EarlyStopping
from sklearn.model_selection import StratifiedShuffleSplit
from sklearn.model_selection import ShuffleSplit
from tensorflow.keras.layers import LeakyReLU,PReLU
from tensorflow.keras.optimizers import Adam

def generate_model(dropout, neuronPct, neuronShrink):
    # We start with some percent of 5000 starting neurons on 
    # the first hidden layer.
    neuronCount = int(neuronPct * 5000)
    
    # Construct neural network
    model = Sequential()

    # So long as there would have been at least 25 neurons and 
    # fewer than 10
    # layers, create a new layer.
    layer = 0
    while neuronCount>25 and layer<10:
        # The first (0th) layer needs an input input_dim(neuronCount)
        if layer==0:
            model.add(Dense(neuronCount, 
                input_dim=x.shape[1], 
                activation=PReLU()))
        else:
            model.add(Dense(neuronCount, activation=PReLU())) 
        layer += 1

        # Add dropout after each hidden layer
        model.add(Dropout(dropout))

        # Shrink neuron count for each layer
        neuronCount = neuronCount * neuronShrink

    model.add(Dense(y.shape[1],activation='softmax')) # Output
    return model\end{lstlisting}
\end{tcolorbox}%
We can test this code to see how it creates a neural network based on three such parameters.%
\index{neural network}%
\index{parameter}%
\par%
\begin{tcolorbox}[size=title,title=Code,breakable]%
\begin{lstlisting}[language=Python, upquote=true]
# Generate a model and see what the resulting structure looks like.
model = generate_model(dropout=0.2, neuronPct=0.1, neuronShrink=0.25)
model.summary()\end{lstlisting}
\tcbsubtitle[before skip=\baselineskip]{Output}%
\begin{lstlisting}[upquote=true]
Model: "sequential"
_________________________________________________________________
 Layer (type)                Output Shape              Param #
=================================================================
 dense (Dense)               (None, 500)               24500
 dropout (Dropout)           (None, 500)               0
 dense_1 (Dense)             (None, 125)               62750
 dropout_1 (Dropout)         (None, 125)               0
 dense_2 (Dense)             (None, 31)                3937
 dropout_2 (Dropout)         (None, 31)                0
 dense_3 (Dense)             (None, 7)                 224
=================================================================
Total params: 91,411
Trainable params: 91,411
Non-trainable params: 0
_________________________________________________________________
\end{lstlisting}
\end{tcolorbox}%
We will now create a function to evaluate the neural network using three such parameters.  We use bootstrapping because one training run might have "bad luck" with the assigned random weights.  We use this function to train and then evaluate the neural network.%
\index{bootstrapping}%
\index{neural network}%
\index{parameter}%
\index{random}%
\index{training}%
\par%
\begin{tcolorbox}[size=title,title=Code,breakable]%
\begin{lstlisting}[language=Python, upquote=true]
SPLITS = 2
EPOCHS = 500
PATIENCE = 10

def evaluate_network(dropout,learning_rate,neuronPct,neuronShrink):
    # Bootstrap

    # for Classification
    boot = StratifiedShuffleSplit(n_splits=SPLITS, test_size=0.1)
    # for Regression
    # boot = ShuffleSplit(n_splits=SPLITS, test_size=0.1)

    # Track progress
    mean_benchmark = []
    epochs_needed = []
    num = 0
    
    # Loop through samples
    for train, test in boot.split(x,df['product']):
        start_time = time.time()
        num+=1

        # Split train and test
        x_train = x[train]
        y_train = y[train]
        x_test = x[test]
        y_test = y[test]

        model = generate_model(dropout, neuronPct, neuronShrink)
        model.compile(loss='categorical_crossentropy', 
                      optimizer=Adam(learning_rate=learning_rate))
        monitor = EarlyStopping(monitor='val_loss', min_delta=1e-3, 
        patience=PATIENCE, verbose=0, mode='auto', 
                                restore_best_weights=True)

        # Train on the bootstrap sample
        model.fit(x_train,y_train,validation_data=(x_test,y_test),
                  callbacks=[monitor],verbose=0,epochs=EPOCHS)
        epochs = monitor.stopped_epoch
        epochs_needed.append(epochs)

        # Predict on the out of boot (validation)
        pred = model.predict(x_test)

        # Measure this bootstrap's log loss
        y_compare = np.argmax(y_test,axis=1) # For log loss calculation
        score = metrics.log_loss(y_compare, pred)
        mean_benchmark.append(score)
        m1 = statistics.mean(mean_benchmark)
        m2 = statistics.mean(epochs_needed)
        mdev = statistics.pstdev(mean_benchmark)

        # Record this iteration
        time_took = time.time() - start_time
        
    tensorflow.keras.backend.clear_session()
    return (-m1)\end{lstlisting}
\end{tcolorbox}%
You can try any combination of our three hyperparameters, plus the learning rate, to see how effective these four numbers are.  Of course, our goal is not to manually choose different combinations of these four hyperparameters; we seek to automate.%
\index{hyperparameter}%
\index{learning}%
\index{learning rate}%
\index{parameter}%
\par%
\begin{tcolorbox}[size=title,title=Code,breakable]%
\begin{lstlisting}[language=Python, upquote=true]
print(evaluate_network(
    dropout=0.2,
    learning_rate=1e-3,
    neuronPct=0.2,
    neuronShrink=0.2))\end{lstlisting}
\tcbsubtitle[before skip=\baselineskip]{Output}%
\begin{lstlisting}[upquote=true]
-0.6668764846259546
\end{lstlisting}
\end{tcolorbox}%
First, we must install the Bayesian optimization package if we are in Colab.%
\index{optimization}%
\par%
\begin{tcolorbox}[size=title,title=Code,breakable]%
\begin{lstlisting}[language=Python, upquote=true]
!pip install bayesian-optimization\end{lstlisting}
\end{tcolorbox}%
We will now automate this process. We define the bounds for each of these four hyperparameters and begin the Bayesian optimization. Once the program finishes, the best combination of hyperparameters found is displayed. The%
\index{hyperparameter}%
\index{optimization}%
\index{parameter}%
\index{ROC}%
\index{ROC}%
\textbf{ optimize }%
function accepts two parameters that will significantly impact how long the process takes to complete:%
\index{parameter}%
\index{ROC}%
\index{ROC}%
\par%
\begin{itemize}[noitemsep]%
\item%
\textbf{n\_iter }%
{-} How many steps of Bayesian optimization that you want to perform. The more steps, the more likely you will find a reasonable maximum.%
\index{optimization}%
\item%
\textbf{init\_points}%
: How many steps of random exploration that you want to perform. Random exploration can help by diversifying the exploration space.%
\index{random}%
\end{itemize}%
\begin{tcolorbox}[size=title,title=Code,breakable]%
\begin{lstlisting}[language=Python, upquote=true]
from bayes_opt import BayesianOptimization
import time

# Supress NaN warnings
import warnings
warnings.filterwarnings("ignore",category =RuntimeWarning)

# Bounded region of parameter space
pbounds = {'dropout': (0.0, 0.499),
           'learning_rate': (0.0, 0.1),
           'neuronPct': (0.01, 1),
           'neuronShrink': (0.01, 1)
          }

optimizer = BayesianOptimization(
    f=evaluate_network,
    pbounds=pbounds,
    verbose=2,  # verbose = 1 prints only when a maximum 
    # is observed, verbose = 0 is silent
    random_state=1,
)

start_time = time.time()
optimizer.maximize(init_points=10, n_iter=20,)
time_took = time.time() - start_time

print(f"Total runtime: {hms_string(time_took)}")
print(optimizer.max)\end{lstlisting}
\tcbsubtitle[before skip=\baselineskip]{Output}%
\begin{lstlisting}[upquote=true]
|   iter    |  target   |  dropout  | learni... | neuronPct |
neuron... |
----------------------------------------------------------------------
---
|  1        | -0.8092   |  0.2081   |  0.07203  |  0.01011  |  0.3093
|
|  2        | -0.7167   |  0.07323  |  0.009234 |  0.1944   |  0.3521
|
|  3        | -17.87    |  0.198    |  0.05388  |  0.425    |  0.6884
|
|  4        | -0.8022   |  0.102    |  0.08781  |  0.03711  |  0.6738
|
|  5        | -0.9209   |  0.2082   |  0.05587  |  0.149    |  0.2061
|
|  6        | -17.96    |  0.3996   |  0.09683  |  0.3203   |  0.6954

...

Total runtime: 1:36:11.56
{'target': -0.6955536706512794, 'params': {'dropout':
0.2504561773412203, 'learning_rate': 0.0076232346709142924,
'neuronPct': 0.012648791521811826, 'neuronShrink':
0.5229748831552032}}
\end{lstlisting}
\end{tcolorbox}%
As you can see, the algorithm performed 30 total iterations. This total iteration count includes ten random and 20 optimization iterations.%
\index{algorithm}%
\index{iteration}%
\index{optimization}%
\index{random}%
\par

\section{Part 8.5: Current Semester's Kaggle}%
\label{sec:Part8.5CurrentSemestersKaggle}%
Kaggke competition site for current semester:%
\par%
\begin{itemize}[noitemsep]%
\item%
Fall 2022 coming soon.%
\end{itemize}%
Previous Kaggle competition sites for this class (NOT this semester's assignment, feel free to use code):%
\index{Kaggle}%
\par%
\begin{itemize}[noitemsep]%
\item%
\href{https://www.kaggle.com/c/tsp-cv}{Spring 2022 Kaggle Assignment}%
\item%
\href{https://www.kaggle.com/c/applications-of-deep-learning-wustlfall-2021}{Fall 2021 Kaggle Assignment}%
\item%
\href{https://www.kaggle.com/c/applications-of-deep-learning-wustl-spring-2021b}{Spring 2021 Kaggle Assignment}%
\item%
\href{https://www.kaggle.com/c/applications-of-deep-learning-wustl-fall-2020}{Fall 2020 Kaggle Assignment}%
\item%
\href{https://www.kaggle.com/c/applications-of-deep-learningwustl-spring-2020}{Spring 2020 Kaggle Assignment}%
\item%
\href{https://kaggle.com/c/applications-of-deep-learningwustl-fall-2019}{Fall 2019 Kaggle Assignment}%
\item%
\href{https://www.kaggle.com/c/applications-of-deep-learningwustl-spring-2019}{Spring 2019 Kaggle Assignment}%
\item%
\href{https://www.kaggle.com/c/wustl-t81-558-washu-deep-learning-fall-2018}{Fall 2018 Kaggle Assignment}%
\item%
\href{https://www.kaggle.com/c/wustl-t81-558-washu-deep-learning-spring-2018}{Spring 2018 Kaggle Assignment}%
\item%
\href{https://www.kaggle.com/c/wustl-t81-558-washu-deep-learning-fall-2017}{Fall 2017 Kaggle Assignment}%
\item%
\href{https://inclass.kaggle.com/c/applications-of-deep-learning-wustl-spring-2017}{Spring 2017 Kaggle Assignment}%
\item%
\href{https://inclass.kaggle.com/c/wustl-t81-558-washu-deep-learning-fall-2016}{Fall 2016 Kaggle Assignment}%
\end{itemize}%
\subsection{Iris as a Kaggle Competition}%
\label{subsec:IrisasaKaggleCompetition}%
If I used the Iris data as a Kaggle, I would give you the following three files:%
\index{iris}%
\index{Kaggle}%
\par%
\begin{itemize}[noitemsep]%
\item%
\href{https://data.heatonresearch.com/data/t81-558/datasets/kaggle_iris_test.csv}{kaggle\_iris\_test.csv }%
{-} The data that Kaggle will evaluate you on. It contains only input; you must provide answers.  (contains x)%
\index{input}%
\index{Kaggle}%
\item%
\href{https://data.heatonresearch.com/data/t81-558/datasets/kaggle_iris_train.csv}{kaggle\_iris\_train.csv }%
{-} The data that you will use to train. (contains x and y)%
\item%
\href{https://data.heatonresearch.com/data/t81-558/datasets/kaggle_iris_sample.csv}{kaggle\_iris\_sample.csv }%
{-} A sample submission for Kaggle. (contains x and y)%
\index{Kaggle}%
\end{itemize}%
Important features of the Kaggle iris files (that differ from how we've previously seen files):%
\index{feature}%
\index{iris}%
\index{Kaggle}%
\par%
\begin{itemize}[noitemsep]%
\item%
The iris species is already index encoded.%
\index{iris}%
\index{species}%
\item%
Your training data is in a separate file.%
\index{training}%
\item%
You will load the test data to generate a submission file.%
\end{itemize}%
The following program generates a submission file for "Iris Kaggle". You can use it as a starting point for assignment 3.%
\index{iris}%
\index{Kaggle}%
\par%
\begin{tcolorbox}[size=title,title=Code,breakable]%
\begin{lstlisting}[language=Python, upquote=true]
import os
import pandas as pd
from sklearn.model_selection import train_test_split
import tensorflow as tf
import numpy as np
from tensorflow.keras.models import Sequential
from tensorflow.keras.layers import Dense, Activation
from tensorflow.keras.callbacks import EarlyStopping

df_train = pd.read_csv(
    "https://data.heatonresearch.com/data/t81-558/datasets/"+\
    "kaggle_iris_train.csv", na_values=['NA','?'])

# Encode feature vector
df_train.drop('id', axis=1, inplace=True)

num_classes = len(df_train.groupby('species').species.nunique())

print("Number of classes: {}".format(num_classes))

# Convert to numpy - Classification
x = df_train[['sepal_l', 'sepal_w', 'petal_l', 'petal_w']].values
dummies = pd.get_dummies(df_train['species']) # Classification
species = dummies.columns
y = dummies.values
    
# Split into train/test
x_train, x_test, y_train, y_test = train_test_split(    
    x, y, test_size=0.25, random_state=45)

# Train, with early stopping
model = Sequential()
model.add(Dense(50, input_dim=x.shape[1], activation='relu'))
model.add(Dense(25))
model.add(Dense(y.shape[1],activation='softmax'))
model.compile(loss='categorical_crossentropy', optimizer='adam')
monitor = EarlyStopping(monitor='val_loss', min_delta=1e-3, 
                        patience=5, verbose=1, mode='auto',
                       restore_best_weights=True)

model.fit(x_train,y_train,validation_data=(x_test,y_test),
          callbacks=[monitor],verbose=0,epochs=1000)\end{lstlisting}
\tcbsubtitle[before skip=\baselineskip]{Output}%
\begin{lstlisting}[upquote=true]
Number of classes: 3
Restoring model weights from the end of the best epoch: 103.
Epoch 108: early stopping
\end{lstlisting}
\end{tcolorbox}%
Now that we've trained the neural network, we can check its log loss.%
\index{neural network}%
\par%
\begin{tcolorbox}[size=title,title=Code,breakable]%
\begin{lstlisting}[language=Python, upquote=true]
from sklearn import metrics

# Calculate multi log loss error
pred = model.predict(x_test)
score = metrics.log_loss(y_test, pred)
print("Log loss score: {}".format(score))\end{lstlisting}
\tcbsubtitle[before skip=\baselineskip]{Output}%
\begin{lstlisting}[upquote=true]
Log loss score: 0.10988010508939623
\end{lstlisting}
\end{tcolorbox}%
Now we are ready to generate the Kaggle submission file.  We will use the iris test data that does not contain a $y$ target value.  It is our job to predict this value and submit it to Kaggle.%
\index{iris}%
\index{Kaggle}%
\index{predict}%
\par%
\begin{tcolorbox}[size=title,title=Code,breakable]%
\begin{lstlisting}[language=Python, upquote=true]
# Generate Kaggle submit file

# Encode feature vector
df_test = pd.read_csv(
    "https://data.heatonresearch.com/data/t81-558/datasets/"+\
    "kaggle_iris_test.csv", na_values=['NA','?'])

# Convert to numpy - Classification
ids = df_test['id']
df_test.drop('id', axis=1, inplace=True)
x = df_test[['sepal_l', 'sepal_w', 'petal_l', 'petal_w']].values
y = dummies.values

# Generate predictions
pred = model.predict(x)
#pred

# Create submission data set

df_submit = pd.DataFrame(pred)
df_submit.insert(0,'id',ids)
df_submit.columns = ['id','species-0','species-1','species-2']

# Write submit file locally
df_submit.to_csv("iris_submit.csv", index=False) 

print(df_submit[:5])\end{lstlisting}
\tcbsubtitle[before skip=\baselineskip]{Output}%
\begin{lstlisting}[upquote=true]
id  species-0  species-1  species-2
0  100   0.022300   0.777859   0.199841
1  101   0.001309   0.273849   0.724842
2  102   0.001153   0.319349   0.679498
3  103   0.958006   0.041989   0.000005
4  104   0.976932   0.023066   0.000002
\end{lstlisting}
\end{tcolorbox}

\subsection{MPG as a Kaggle Competition (Regression)}%
\label{subsec:MPGasaKaggleCompetition(Regression)}%
If the Auto MPG data were used as a Kaggle, you would be given the following three files:%
\index{Kaggle}%
\par%
\begin{itemize}[noitemsep]%
\item%
\href{https://data.heatonresearch.com/data/t81-558/datasets/kaggle_auto_test.csv}{kaggle\_mpg\_test.csv }%
{-} The data that Kaggle will evaluate you on.  Contains only input, you must provide answers.  (contains x)%
\index{input}%
\index{Kaggle}%
\item%
\href{https://data.heatonresearch.com/data/t81-558/datasets/kaggle_auto_test.csv}{kaggle\_mpg\_train.csv }%
{-} The data that you will use to train. (contains x and y)%
\item%
\href{https://data.heatonresearch.com/data/t81-558/datasets/kaggle_auto_sample.csv}{kaggle\_mpg\_sample.csv }%
{-} A sample submission for Kaggle. (contains x and y)%
\index{Kaggle}%
\end{itemize}%
Important features of the Kaggle iris files (that differ from how we've previously seen files):%
\index{feature}%
\index{iris}%
\index{Kaggle}%
\par%
The following program generates a submission file for "MPG Kaggle".%
\index{Kaggle}%
\par%
\begin{tcolorbox}[size=title,title=Code,breakable]%
\begin{lstlisting}[language=Python, upquote=true]
from tensorflow.keras.models import Sequential
from tensorflow.keras.layers import Dense, Activation
from sklearn.model_selection import train_test_split
from tensorflow.keras.callbacks import EarlyStopping
import pandas as pd
import io
import os
import requests
import numpy as np
from sklearn import metrics

save_path = "."

df = pd.read_csv(
    "https://data.heatonresearch.com/data/t81-558/datasets/"+\
    "kaggle_auto_train.csv", 
    na_values=['NA', '?'])

cars = df['name']

# Handle missing value
df['horsepower'] = df['horsepower'].fillna(df['horsepower'].median())

# Pandas to Numpy
x = df[['cylinders', 'displacement', 'horsepower', 'weight',
       'acceleration', 'year', 'origin']].values
y = df['mpg'].values # regression

# Split into train/test
x_train, x_test, y_train, y_test = train_test_split(    
    x, y, test_size=0.25, random_state=42)

# Build the neural network
model = Sequential()
model.add(Dense(25, input_dim=x.shape[1], activation='relu')) # Hidden 1
model.add(Dense(10, activation='relu')) # Hidden 2
model.add(Dense(1)) # Output
model.compile(loss='mean_squared_error', optimizer='adam')
monitor = EarlyStopping(monitor='val_loss', min_delta=1e-3, patience=5, 
                        verbose=1, mode='auto', restore_best_weights=True)
model.fit(x_train,y_train,validation_data=(x_test,y_test),
          verbose=2,callbacks=[monitor],epochs=1000)

# Predict
pred = model.predict(x_test)\end{lstlisting}
\end{tcolorbox}%
Now that we've trained the neural network, we can check its RMSE error.%
\index{error}%
\index{MSE}%
\index{neural network}%
\index{RMSE}%
\index{RMSE}%
\par%
\begin{tcolorbox}[size=title,title=Code,breakable]%
\begin{lstlisting}[language=Python, upquote=true]
import numpy as np

# Measure RMSE error.  RMSE is common for regression.
score = np.sqrt(metrics.mean_squared_error(pred,y_test))
print("Final score (RMSE): {}".format(score))\end{lstlisting}
\tcbsubtitle[before skip=\baselineskip]{Output}%
\begin{lstlisting}[upquote=true]
Final score (RMSE): 6.023776405947501
\end{lstlisting}
\end{tcolorbox}%
Now we are ready to generate the Kaggle submission file.  We will use the MPG test data that does not contain a $y$ target value.  It is our job to predict this value and submit it to Kaggle.%
\index{Kaggle}%
\index{predict}%
\par%
\begin{tcolorbox}[size=title,title=Code,breakable]%
\begin{lstlisting}[language=Python, upquote=true]
import pandas as pd

# Generate Kaggle submit file

# Encode feature vector
df_test = pd.read_csv(
    "https://data.heatonresearch.com/data/t81-558/datasets/"+\
    "kaggle_auto_test.csv", na_values=['NA','?'])

# Convert to numpy - regression
ids = df_test['id']
df_test.drop('id', axis=1, inplace=True)

# Handle missing value
df_test['horsepower'] = df_test['horsepower'].\
    fillna(df['horsepower'].median())

x = df_test[['cylinders', 'displacement', 'horsepower', 'weight',
       'acceleration', 'year', 'origin']].values

# Generate predictions
pred = model.predict(x)
#pred

# Create submission data set

df_submit = pd.DataFrame(pred)
df_submit.insert(0,'id',ids)
df_submit.columns = ['id','mpg']

# Write submit file locally
df_submit.to_csv("auto_submit.csv", index=False) 

print(df_submit[:5])\end{lstlisting}
\tcbsubtitle[before skip=\baselineskip]{Output}%
\begin{lstlisting}[upquote=true]
id        mpg
0  350  27.158819
1  351  24.450621
2  352  24.913355
3  353  26.994867
4  354  26.669268
\end{lstlisting}
\end{tcolorbox}

\chapter{Transfer Learning}%
\label{chap:TransferLearning}%
\section{Part 9.1: Introduction to Keras Transfer Learning}%
\label{sec:Part9.1IntroductiontoKerasTransferLearning}%
Human beings learn new skills throughout their entire lives. However, this learning is rarely from scratch. No matter what task a human learns, they are most likely drawing on experiences to learn this new skill early in life. In this way, humans learn much differently than most deep learning projects.%
\index{learning}%
\par%
A human being learns to tell the difference between a cat and a dog at some point. To teach a neural network, you would obtain many cat pictures and dog pictures. The neural network would iterate over all of these pictures and train on the differences. The human child that learned to distinguish between the two animals would probably need to see a few examples when parents told them the name of each type of animal. The human child would use previous knowledge of looking at different living and non{-}living objects to help make this classification. The child would already know the physical appearance of sub{-}objects, such as fur, eyes, ears, noses, tails, and teeth.%
\index{classification}%
\index{neural network}%
\index{SOM}%
\par%
Transfer learning attempts to teach a neural network by similar means. Rather than training your neural network from scratch, you begin training with a preloaded set of weights. Usually, you will remove the topmost layers of the pretrained neural network and retrain it with new uppermost layers. The layers from the previous neural network will be locked so that training does not change these weights. Only the newly added layers will be trained.%
\index{layer}%
\index{learning}%
\index{neural network}%
\index{training}%
\index{transfer learning}%
\par%
It can take much computing power to train a neural network for a large image dataset. Google, Facebook, Microsoft, and other tech companies have utilized GPU arrays for training high{-}quality neural networks for various applications. Transferring these weights into your neural network can save considerable effort and compute time. It is unlikely that a pretrained model will exactly fit the application that you seek to implement. Finding the closest pretrained model and using transfer learning is essential for a deep learning engineer.%
\index{dataset}%
\index{GPU}%
\index{GPU}%
\index{learning}%
\index{model}%
\index{neural network}%
\index{training}%
\index{transfer learning}%
\par%
\subsection{Transfer Learning Example}%
\label{subsec:TransferLearningExample}%
Let's look at a simple example of using transfer learning to build upon an imagenet neural network. We will begin by training a neural network for Fisher's Iris Dataset. This network takes four measurements and classifies each observation into three iris species. However, what if later we received a data set that included the four measurements, plus a cost as the target? This dataset does not contain the species; as a result, it uses the same four inputs as the base model we just trained.%
\index{dataset}%
\index{input}%
\index{iris}%
\index{learning}%
\index{model}%
\index{neural network}%
\index{species}%
\index{training}%
\index{transfer learning}%
\par%
We can take our previously trained iris network and transfer the weights to a new neural network that will learn to predict the cost through transfer learning. Also of note, the original neural network was a classification network, yet we now use it to build a regression neural network. Such a transformation is common for transfer learning. As a reference point, I randomly created this iris cost dataset.%
\index{classification}%
\index{dataset}%
\index{iris}%
\index{learning}%
\index{neural network}%
\index{predict}%
\index{random}%
\index{regression}%
\index{transfer learning}%
\par%
The first step is to train our neural network for the regular Iris Dataset. The code presented here is the same as we saw in Module 3.%
\index{dataset}%
\index{iris}%
\index{neural network}%
\par%
\begin{tcolorbox}[size=title,title=Code,breakable]%
\begin{lstlisting}[language=Python, upquote=true]
import pandas as pd
import io
import requests
import numpy as np
from sklearn import metrics
from tensorflow.keras.models import Sequential
from tensorflow.keras.layers import Dense, Activation
from tensorflow.keras.callbacks import EarlyStopping

df = pd.read_csv(
    "https://data.heatonresearch.com/data/t81-558/iris.csv", 
    na_values=['NA', '?'])

# Convert to numpy - Classification
x = df[['sepal_l', 'sepal_w', 'petal_l', 'petal_w']].values
dummies = pd.get_dummies(df['species']) # Classification
species = dummies.columns
y = dummies.values


# Build neural network
model = Sequential()
model.add(Dense(50, input_dim=x.shape[1], activation='relu')) # Hidden 1
model.add(Dense(25, activation='relu')) # Hidden 2
model.add(Dense(y.shape[1],activation='softmax')) # Output

model.compile(loss='categorical_crossentropy', optimizer='adam')
model.fit(x,y,verbose=2,epochs=100)\end{lstlisting}
\tcbsubtitle[before skip=\baselineskip]{Output}%
\begin{lstlisting}[upquote=true]
...
5/5 - 0s - loss: 0.0868 - 15ms/epoch - 3ms/step
Epoch 100/100
5/5 - 0s - loss: 0.0892 - 8ms/epoch - 2ms/step
\end{lstlisting}
\end{tcolorbox}%
To keep this example simple, we are not setting aside a validation set.  The goal of this example is to show how to create a multi{-}layer neural network, where we transfer the weights to another network.  We begin by evaluating the accuracy of the network on the training set.%
\index{layer}%
\index{neural network}%
\index{training}%
\index{validation}%
\par%
\begin{tcolorbox}[size=title,title=Code,breakable]%
\begin{lstlisting}[language=Python, upquote=true]
from sklearn.metrics import accuracy_score
pred = model.predict(x)
predict_classes = np.argmax(pred,axis=1)
expected_classes = np.argmax(y,axis=1)
correct = accuracy_score(expected_classes,predict_classes)
print(f"Training Accuracy: {correct}")\end{lstlisting}
\tcbsubtitle[before skip=\baselineskip]{Output}%
\begin{lstlisting}[upquote=true]
Training Accuracy: 0.9866666666666667
\end{lstlisting}
\end{tcolorbox}%
Viewing the model summary is as expected; we can see the three layers previously defined.%
\index{layer}%
\index{model}%
\par%
\begin{tcolorbox}[size=title,title=Code,breakable]%
\begin{lstlisting}[language=Python, upquote=true]
model.summary()\end{lstlisting}
\tcbsubtitle[before skip=\baselineskip]{Output}%
\begin{lstlisting}[upquote=true]
Model: "sequential"
_________________________________________________________________
 Layer (type)                Output Shape              Param #
=================================================================
 dense (Dense)               (None, 50)                250
 dense_1 (Dense)             (None, 25)                1275
 dense_2 (Dense)             (None, 3)                 78
=================================================================
Total params: 1,603
Trainable params: 1,603
Non-trainable params: 0
_________________________________________________________________
\end{lstlisting}
\end{tcolorbox}

\subsection{Create a New Iris Network}%
\label{subsec:CreateaNewIrisNetwork}%
Now that we've trained a neural network on the iris dataset, we can transfer the knowledge of this neural network to other neural networks. It is possible to create a new neural network from some or all of the layers of this neural network. We will create a new neural network that is essentially a clone of the first neural network to demonstrate the technique. We now transfer all of the layers from the original neural network into the new one.%
\index{dataset}%
\index{iris}%
\index{layer}%
\index{neural network}%
\index{SOM}%
\par%
\begin{tcolorbox}[size=title,title=Code,breakable]%
\begin{lstlisting}[language=Python, upquote=true]
model2 = Sequential()
for layer in model.layers:
    model2.add(layer)
model2.summary()\end{lstlisting}
\tcbsubtitle[before skip=\baselineskip]{Output}%
\begin{lstlisting}[upquote=true]
Model: "sequential_1"
_________________________________________________________________
 Layer (type)                Output Shape              Param #
=================================================================
 dense (Dense)               (None, 50)                250
 dense_1 (Dense)             (None, 25)                1275
 dense_2 (Dense)             (None, 3)                 78
=================================================================
Total params: 1,603
Trainable params: 1,603
Non-trainable params: 0
_________________________________________________________________
\end{lstlisting}
\end{tcolorbox}%
As a sanity check, we would like to calculate the accuracy of the newly created model.  The in{-}sample accuracy should be the same as the previous model that the new model transferred.%
\index{model}%
\par%
\begin{tcolorbox}[size=title,title=Code,breakable]%
\begin{lstlisting}[language=Python, upquote=true]
from sklearn.metrics import accuracy_score
pred = model2.predict(x)
predict_classes = np.argmax(pred,axis=1)
expected_classes = np.argmax(y,axis=1)
correct = accuracy_score(expected_classes,predict_classes)
print(f"Training Accuracy: {correct}")\end{lstlisting}
\tcbsubtitle[before skip=\baselineskip]{Output}%
\begin{lstlisting}[upquote=true]
Training Accuracy: 0.9866666666666667
\end{lstlisting}
\end{tcolorbox}%
The in{-}sample accuracy of the newly created neural network is the same as the first neural network. We've successfully transferred all of the layers from the original neural network.%
\index{layer}%
\index{neural network}%
\par

\subsection{Transfering to a Regression Network}%
\label{subsec:TransferingtoaRegressionNetwork}%
The Iris Cost Dataset has measurements for samples of these flowers that conform to the predictors contained in the original iris dataset: sepal width, sepal length, petal width, and petal length. We present the cost dataset here.%
\index{dataset}%
\index{iris}%
\index{predict}%
\par%
\begin{tcolorbox}[size=title,title=Code,breakable]%
\begin{lstlisting}[language=Python, upquote=true]
df_cost = pd.read_csv(
    "https://data.heatonresearch.com/data/t81-558/iris_cost.csv", 
    na_values=['NA', '?'])

df_cost\end{lstlisting}
\tcbsubtitle[before skip=\baselineskip]{Output}%
\begin{tabular}[hbt!]{l|l|l|l|l|l}%
\hline%
&sepal\_l&sepal\_w&petal\_l&petal\_w&cost\\%
\hline%
0&7.8&3.0&6.2&2.0&10.740\\%
1&5.0&2.2&1.7&1.5&2.710\\%
2&6.9&2.6&3.7&1.4&4.624\\%
3&5.9&2.2&3.7&2.4&6.558\\%
4&5.1&3.9&6.8&0.7&7.395\\%
...&...&...&...&...&...\\%
245&4.7&2.1&4.0&2.3&5.721\\%
246&7.2&3.0&4.3&1.1&5.266\\%
247&6.6&3.4&4.6&1.4&5.776\\%
248&5.7&3.7&3.1&0.4&2.233\\%
249&7.6&4.0&5.1&1.4&7.508\\%
\hline%
\end{tabular}%
\vspace{2mm}%
\end{tcolorbox}%
For transfer learning to be effective, the input for the newly trained neural network most closely conforms to the first neural network we transfer.%
\index{input}%
\index{learning}%
\index{neural network}%
\index{transfer learning}%
\par%
We will strip away the last output layer that contains the softmax activation function that performs this final classification. We will create a new output layer that will output the cost prediction. We will only train the weights in this new layer. We will mark the first two layers as non{-}trainable. The hope is that the first few layers have learned to abstract the raw input data in a way that is also helpful to the new neural network.\newline%
This process is accomplished by looping over the first few layers and copying them to the new neural network. We output a summary of the new neural network to verify that Keras stripped the previous output layer.%
\index{activation function}%
\index{classification}%
\index{input}%
\index{Keras}%
\index{layer}%
\index{neural network}%
\index{output}%
\index{output layer}%
\index{predict}%
\index{ROC}%
\index{ROC}%
\index{softmax}%
\par%
\begin{tcolorbox}[size=title,title=Code,breakable]%
\begin{lstlisting}[language=Python, upquote=true]
model3 = Sequential()
for i in range(2):
    layer = model.layers[i]
    layer.trainable = False
    model3.add(layer)
model3.summary()\end{lstlisting}
\tcbsubtitle[before skip=\baselineskip]{Output}%
\begin{lstlisting}[upquote=true]
Model: "sequential_2"
_________________________________________________________________
 Layer (type)                Output Shape              Param #
=================================================================
 dense (Dense)               (None, 50)                250
 dense_1 (Dense)             (None, 25)                1275
=================================================================
Total params: 1,525
Trainable params: 0
Non-trainable params: 1,525
_________________________________________________________________
\end{lstlisting}
\end{tcolorbox}%
We add a final regression output layer to complete the new neural network.%
\index{layer}%
\index{neural network}%
\index{output}%
\index{output layer}%
\index{regression}%
\par%
\begin{tcolorbox}[size=title,title=Code,breakable]%
\begin{lstlisting}[language=Python, upquote=true]
model3.add(Dense(1)) # Output

model3.compile(loss='mean_squared_error', optimizer='adam')
model3.summary()\end{lstlisting}
\tcbsubtitle[before skip=\baselineskip]{Output}%
\begin{lstlisting}[upquote=true]
Model: "sequential_2"
_________________________________________________________________
 Layer (type)                Output Shape              Param #
=================================================================
 dense (Dense)               (None, 50)                250
 dense_1 (Dense)             (None, 25)                1275
 dense_3 (Dense)             (None, 1)                 26
=================================================================
Total params: 1,551
Trainable params: 26
Non-trainable params: 1,525
_________________________________________________________________
\end{lstlisting}
\end{tcolorbox}%
Now we train just the output layer to predict the cost. The cost in the made{-}up dataset is dependent on the species, so the previous learning should be helpful.%
\index{dataset}%
\index{layer}%
\index{learning}%
\index{output}%
\index{output layer}%
\index{predict}%
\index{species}%
\par%
\begin{tcolorbox}[size=title,title=Code,breakable]%
\begin{lstlisting}[language=Python, upquote=true]
# Convert to numpy - Classification
x = df_cost[['sepal_l', 'sepal_w', 'petal_l', 'petal_w']].values
y = df_cost.cost.values

# Train the last layer of the network
model3.fit(x,y,verbose=2,epochs=100)\end{lstlisting}
\tcbsubtitle[before skip=\baselineskip]{Output}%
\begin{lstlisting}[upquote=true]
...
8/8 - 0s - loss: 1.8851 - 17ms/epoch - 2ms/step
Epoch 100/100
8/8 - 0s - loss: 1.8838 - 9ms/epoch - 1ms/step
\end{lstlisting}
\end{tcolorbox}%
We can evaluate the in{-}sample RMSE for the new model containing transferred layers from the previous model.%
\index{layer}%
\index{model}%
\index{MSE}%
\index{RMSE}%
\index{RMSE}%
\par%
\begin{tcolorbox}[size=title,title=Code,breakable]%
\begin{lstlisting}[language=Python, upquote=true]
from sklearn.metrics import accuracy_score
pred = model3.predict(x)
score = np.sqrt(metrics.mean_squared_error(pred,y))
print(f"Final score (RMSE): {score}")\end{lstlisting}
\tcbsubtitle[before skip=\baselineskip]{Output}%
\begin{lstlisting}[upquote=true]
Final score (RMSE): 1.3716589625823072
\end{lstlisting}
\end{tcolorbox}

\subsection{Module 9 Assignment}%
\label{subsec:Module9Assignment}%
You can find the first assignment here:%
\href{https://github.com/jeffheaton/t81_558_deep_learning/blob/master/assignments/assignment_yourname_class9.ipynb}{ assignment 9}%
\par

\section{Part 9.2: Keras Transfer Learning for Computer Vision}%
\label{sec:Part9.2KerasTransferLearningforComputerVision}%
We will take a look at several popular pretrained neural networks for Keras. The following two sites, among others, can be great starting points to find pretrained models for use in your projects:%
\index{Keras}%
\index{model}%
\index{neural network}%
\par%
\begin{itemize}[noitemsep]%
\item%
\href{https://modelzoo.co/}{TensorFlow Model Zoo}%
\item%
\href{https://paperswithcode.com/}{Papers with Code}%
\end{itemize}%
Keras contains built{-}in support for several pretrained models. In the Keras documentation, you can find the%
\index{Keras}%
\index{model}%
\href{https://keras.io/applications/}{ complete list}%
.%
\par%
\subsection{Transfering Computer Vision}%
\label{subsec:TransferingComputerVision}%
There are many pretrained models for computer vision. This section will show you how to obtain a pretrained model for computer vision and train just the output layer. Additionally, once we train the output layer, we will fine{-}tune the entire network by training all weights using by applying a low learning rate.%
\index{computer vision}%
\index{layer}%
\index{learning}%
\index{learning rate}%
\index{model}%
\index{output}%
\index{output layer}%
\index{training}%
\par

\subsection{The Kaggle Cats vs. Dogs Dataset}%
\label{subsec:TheKaggleCatsvs.DogsDataset}%
We will train a neural network to recognize cats and dogs for this example. The {[}cats and dogs dataset{]} comes from a classic Kaggle competition. We can achieve a very high score on this data set through modern training techniques and ensemble learning. I based this module on a tutorial provided by%
\index{dataset}%
\index{ensemble}%
\index{ensemble learning}%
\index{Kaggle}%
\index{learning}%
\index{neural network}%
\index{training}%
\href{https://keras.io/guides/transfer_learning/}{ Francois Chollet}%
, one of the creators of Keras. I made some changes to his example to fit with this course.%
\index{Keras}%
\index{SOM}%
\par%
We begin by downloading this dataset from Keras. We do not need the entire dataset to achieve high accuracy. Using a portion also speeds up training. We will use 40\% of the original training data (25,000 images) for training and 10\% for validation.%
\index{dataset}%
\index{Keras}%
\index{training}%
\index{validation}%
\par%
The dogs and cats dataset is relatively large and will not easily fit into a less than 12GB system, such as Colab. Because of this memory size, you must take additional steps to handle the data. Rather than loading the dataset as a Numpy array, as done previously in this book, we will load it as a prefetched dataset so that only the portions of the dataset currently needed are in RAM. If you wish to load the dataset, in its entirety as a Numpy array, add the batch\_size={-}1 option to the load command below.%
\index{dataset}%
\index{NumPy}%
\par%
\begin{tcolorbox}[size=title,title=Code,breakable]%
\begin{lstlisting}[language=Python, upquote=true]
import tensorflow_datasets as tfds
import tensorflow as tf

tfds.disable_progress_bar()

train_ds, validation_ds = tfds.load(
    "cats_vs_dogs",
    split=["train[:40%]", "train[40%:50%]"],
    as_supervised=True,  # Include labels
)

num_train = tf.data.experimental.cardinality(train_ds)
num_test = tf.data.experimental.cardinality(validation_ds)

print(f"Number of training samples: {num_train}")
print(f"Number of validation samples: {num_test}")\end{lstlisting}
\tcbsubtitle[before skip=\baselineskip]{Output}%
\begin{lstlisting}[upquote=true]
Number of training samples: 9305
Number of validation samples: 2326
\end{lstlisting}
\end{tcolorbox}

\subsection{Looking at the Data and Augmentations}%
\label{subsec:LookingattheDataandAugmentations}%
We begin by displaying several of the images from this dataset. The labels are above each image. As can be seen from the images below, 1 indicates a dog, and 0 indicates a cat.%
\index{dataset}%
\par%
\begin{tcolorbox}[size=title,title=Code,breakable]%
\begin{lstlisting}[language=Python, upquote=true]
import matplotlib.pyplot as plt

plt.figure(figsize=(10, 10))
for i, (image, label) in enumerate(train_ds.take(9)):
    ax = plt.subplot(3, 3, i + 1)
    plt.imshow(image)
    plt.title(int(label))
    plt.axis("off")\end{lstlisting}
\tcbsubtitle[before skip=\baselineskip]{Output}%
\includegraphics[width=4in]{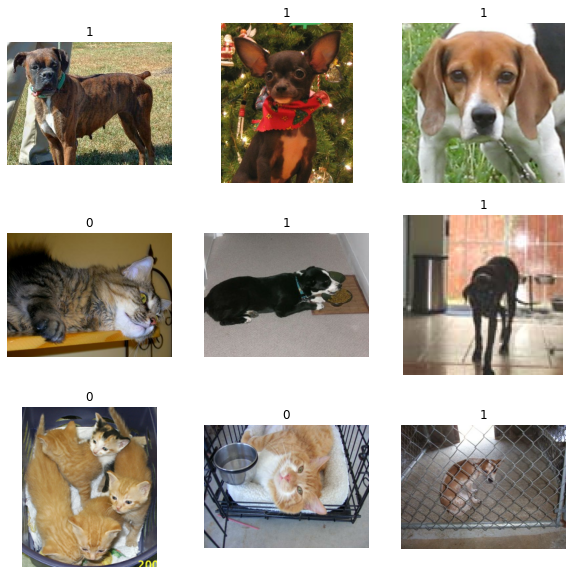}%
\end{tcolorbox}%
Upon examining the above images, another problem becomes evident. The images are of various sizes. We will standardize all images to 190x190 with the following code.%
\par%
\begin{tcolorbox}[size=title,title=Code,breakable]%
\begin{lstlisting}[language=Python, upquote=true]
size = (150, 150)

train_ds = train_ds.map(lambda x, y: (tf.image.resize(x, size), y))
validation_ds = validation_ds.map(lambda x, y: \
                                  (tf.image.resize(x, size), y))\end{lstlisting}
\end{tcolorbox}%
We will batch the data and use caching and prefetching to optimize loading speed.%
\par%
\begin{tcolorbox}[size=title,title=Code,breakable]%
\begin{lstlisting}[language=Python, upquote=true]
batch_size = 32

train_ds = train_ds.cache().batch(batch_size).prefetch(buffer_size=10)
validation_ds = validation_ds.cache() \
    .batch(batch_size).prefetch(buffer_size=10)\end{lstlisting}
\end{tcolorbox}%
Augmentation is a powerful computer vision technique that increases the amount of training data available to your model by altering the images in the training data. To use augmentation, we will allow horizontal flips of the images. A horizontal flip makes much more sense for cats and dogs in the real world than a vertical flip. How often do you see upside{-}down dogs or cats? We also include a limited degree of rotation.%
\index{computer vision}%
\index{model}%
\index{training}%
\par%
\begin{tcolorbox}[size=title,title=Code,breakable]%
\begin{lstlisting}[language=Python, upquote=true]
from tensorflow import keras
from tensorflow.keras import layers

data_augmentation = keras.Sequential(
    [layers.RandomFlip("horizontal"), layers.RandomRotation(0.1),]
)\end{lstlisting}
\end{tcolorbox}%
The following code allows us to visualize the augmentation.%
\par%
\begin{tcolorbox}[size=title,title=Code,breakable]%
\begin{lstlisting}[language=Python, upquote=true]
import numpy as np

for images, labels in train_ds.take(1):
    plt.figure(figsize=(10, 10))
    first_image = images[0]
    for i in range(9):
        ax = plt.subplot(3, 3, i + 1)
        augmented_image = data_augmentation(
            tf.expand_dims(first_image, 0), training=True
        )
        plt.imshow(augmented_image[0].numpy().astype("int32"))
        plt.title(int(labels[0]))
        plt.axis("off")\end{lstlisting}
\tcbsubtitle[before skip=\baselineskip]{Output}%
\includegraphics[width=4in]{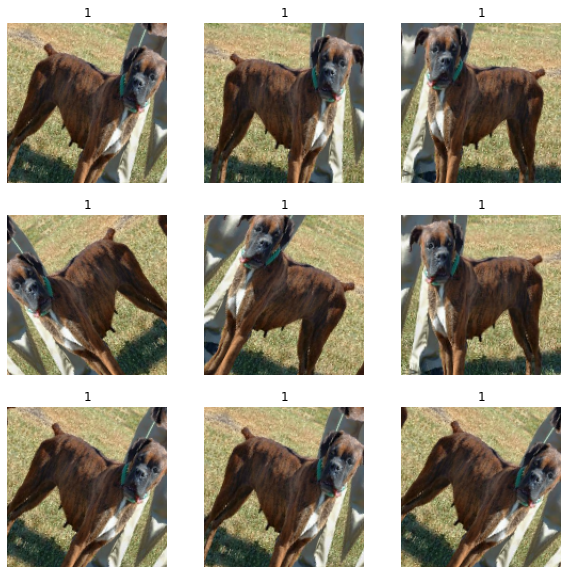}%
\end{tcolorbox}

\subsection{Create a Network and Transfer Weights}%
\label{subsec:CreateaNetworkandTransferWeights}%
We are now ready to create our new neural network with transferred weights. We will transfer the weights from an Xception neural network that contains weights trained for imagenet. We load the existing Xception neural network with%
\index{neural network}%
\textbf{ keras.applications}%
. There is quite a bit going on with the loading of the%
\textbf{ base\_model}%
, so we will examine this call piece by piece.%
\par%
The base Xception neural network accepts an image of 299x299. However, we would like to use 150x150. It turns out that it is relatively easy to overcome this difference. Convolutional neural networks move a kernel across an image tensor as they scan. Keras defines the number of weights by the size of the layer's kernel, not the image that the kernel scans. As a result, we can discard the old input layer and recreate an input layer consistent with our desired image size. We specify%
\index{convolution}%
\index{convolutional}%
\index{Convolutional Neural Networks}%
\index{input}%
\index{input layer}%
\index{Keras}%
\index{layer}%
\index{neural network}%
\textbf{ include\_top }%
as false and specify our input shape.%
\index{input}%
\par%
We freeze the base model so that the model will not update existing weights as training occurs. We create the new input layer that consists of 150x150 by 3 RGB color components. These RGB components are integer numbers between 0 and 255. Neural networks deal better with floating{-}point numbers when you distribute them around zero. To accomplish this neural network advantage, we normalize each RGB component to between {-}1 and 1.%
\index{input}%
\index{input layer}%
\index{layer}%
\index{model}%
\index{neural network}%
\index{training}%
\par%
The batch normalization layers do require special consideration. We need to keep these layers in inference mode when we unfreeze the base model for fine{-}tuning. To do this, we make sure that the base model is running in inference mode here.%
\index{layer}%
\index{model}%
\par%
\begin{tcolorbox}[size=title,title=Code,breakable]%
\begin{lstlisting}[language=Python, upquote=true]
base_model = keras.applications.Xception(
    weights="imagenet",  # Load weights pre-trained on ImageNet.
    input_shape=(150, 150, 3),
    include_top=False,
)  # Do not include the ImageNet classifier at the top.

# Freeze the base_model
base_model.trainable = False

# Create new model on top
inputs = keras.Input(shape=(150, 150, 3))
x = data_augmentation(inputs)  # Apply random data augmentation

# Pre-trained Xception weights requires that input be scaled
# from (0, 255) to a range of (-1., +1.), the rescaling layer
# outputs: `(inputs * scale) + offset`
scale_layer = keras.layers.Rescaling(scale=1 / 127.5, offset=-1)
x = scale_layer(x)

# The base model contains batchnorm layers. 
# We want to keep them in inference mode
# when we unfreeze the base model for fine-tuning, 
# so we make sure that the
# base_model is running in inference mode here.
x = base_model(x, training=False)
x = keras.layers.GlobalAveragePooling2D()(x)
x = keras.layers.Dropout(0.2)(x)  # Regularize with dropout
outputs = keras.layers.Dense(1)(x)
model = keras.Model(inputs, outputs)

model.summary()\end{lstlisting}
\tcbsubtitle[before skip=\baselineskip]{Output}%
\begin{lstlisting}[upquote=true]
Downloading data from https://storage.googleapis.com/tensorflow/keras-
applications/xception/xception_weights_tf_dim_ordering_tf_kernels_noto
p.h5
83689472/83683744 [==============================] - 1s 0us/step
83697664/83683744 [==============================] - 1s 0us/step
Model: "model"
_________________________________________________________________
 Layer (type)                Output Shape              Param #
=================================================================
 input_2 (InputLayer)        [(None, 150, 150, 3)]     0
 sequential (Sequential)     (None, 150, 150, 3)       0
 rescaling (Rescaling)       (None, 150, 150, 3)       0
 xception (Functional)       (None, 5, 5, 2048)        20861480
 global_average_pooling2d (G  (None, 2048)             0
 lobalAveragePooling2D)

...

=================================================================
Total params: 20,863,529
Trainable params: 2,049
Non-trainable params: 20,861,480
_________________________________________________________________
\end{lstlisting}
\end{tcolorbox}%
Next, we compile and fit the model. The fitting will use the Adam optimizer; because we are performing binary classification, we use the binary cross{-}entropy loss function, as we have done before.%
\index{ADAM}%
\index{classification}%
\index{model}%
\par%
\begin{tcolorbox}[size=title,title=Code,breakable]%
\begin{lstlisting}[language=Python, upquote=true]
model.compile(
    optimizer=keras.optimizers.Adam(),
    loss=keras.losses.BinaryCrossentropy(from_logits=True), 
    metrics=[keras.metrics.BinaryAccuracy()],
)

epochs = 20
model.fit(train_ds, epochs=epochs, validation_data=validation_ds)\end{lstlisting}
\tcbsubtitle[before skip=\baselineskip]{Output}%
\begin{lstlisting}[upquote=true]
...
291/291 [==============================] - 11s 37ms/step - loss:
0.0907 - binary_accuracy: 0.9627 - val_loss: 0.0718 -
val_binary_accuracy: 0.9729
Epoch 20/20
291/291 [==============================] - 11s 37ms/step - loss:
0.0899 - binary_accuracy: 0.9652 - val_loss: 0.0694 -
val_binary_accuracy: 0.9746
\end{lstlisting}
\end{tcolorbox}%
The training above shows that the validation accuracy reaches the mid 90\% range. This accuracy is good; however, we can do better.%
\index{training}%
\index{validation}%
\par

\subsection{Fine{-}Tune the Model}%
\label{subsec:Fine{-}TunetheModel}%
Finally, we will fine{-}tune the model. First, we set all weights to trainable and then train the neural network with a low learning rate (1e{-}5). This fine{-}tuning results in an accuracy in the upper 90\% range. The fine{-}tuning allows all weights in the neural network to adjust slightly to optimize for the dogs/cats data.%
\index{learning}%
\index{learning rate}%
\index{model}%
\index{neural network}%
\par%
\begin{tcolorbox}[size=title,title=Code,breakable]%
\begin{lstlisting}[language=Python, upquote=true]
# Unfreeze the base_model. Note that it keeps running in inference mode
# since we passed `training=False` when calling it. This means that
# the batchnorm layers will not update their batch statistics.
# This prevents the batchnorm layers from undoing all the training
# we've done so far.
base_model.trainable = True
model.summary()

model.compile(
    optimizer=keras.optimizers.Adam(1e-5),  # Low learning rate
    loss=keras.losses.BinaryCrossentropy(from_logits=True),
    metrics=[keras.metrics.BinaryAccuracy()],
)

epochs = 10
model.fit(train_ds, epochs=epochs, validation_data=validation_ds)\end{lstlisting}
\tcbsubtitle[before skip=\baselineskip]{Output}%
\begin{lstlisting}[upquote=true]
Model: "model"
_________________________________________________________________
 Layer (type)                Output Shape              Param #
=================================================================
 input_2 (InputLayer)        [(None, 150, 150, 3)]     0
 sequential (Sequential)     (None, 150, 150, 3)       0
 rescaling (Rescaling)       (None, 150, 150, 3)       0
 xception (Functional)       (None, 5, 5, 2048)        20861480
 global_average_pooling2d (G  (None, 2048)             0
 lobalAveragePooling2D)
 dropout (Dropout)           (None, 2048)              0
 dense (Dense)               (None, 1)                 2049
=================================================================
Total params: 20,863,529
Trainable params: 20,809,001

...

val_binary_accuracy: 0.9837
Epoch 10/10
291/291 [==============================] - 41s 140ms/step - loss:
0.0162 - binary_accuracy: 0.9944 - val_loss: 0.0548 -
val_binary_accuracy: 0.9819
\end{lstlisting}
\end{tcolorbox}

\section{Part 9.3: Transfer Learning for NLP with Keras}%
\label{sec:Part9.3TransferLearningforNLPwithKeras}%
You will commonly use transfer learning with Natural Language Processing (NLP). Word embeddings are a common means of transfer learning in NLP where network layers map words to vectors. Third parties trained neural networks on a large corpus of text to learn these embeddings. We will use these vectors as the input to the neural network rather than the actual characters of words.%
\index{input}%
\index{layer}%
\index{learning}%
\index{neural network}%
\index{ROC}%
\index{ROC}%
\index{transfer learning}%
\index{vector}%
\par%
This course has an entire module covering NLP; however, we use word embeddings to perform sentiment analysis in this module. We will specifically attempt to classify if a text sample is speaking in a positive or negative tone.%
\index{sentiment analysis}%
\par%
The following three sources were helpful for the creation of this section.%
\par%
\begin{itemize}[noitemsep]%
\item%
Universal sentence encoder%
\cite{cer2018universal}%
. arXiv preprint arXiv:1803.11175)%
\item%
Deep Transfer Learning for Natural Language Processing: Text Classification with Universal Embeddings%
\index{classification}%
\index{learning}%
\index{ROC}%
\index{ROC}%
\index{transfer learning}%
\cite{howard2018universal}%
\item%
\href{http://hunterheidenreich.com/blog/google-universal-sentence-encoder-in-keras/}{Keras Tutorial: How to Use Google's Universal Sentence Encoder for Spam Classification}%
\end{itemize}%
These examples use TensorFlow Hub, which allows pretrained models to be loaded into TensorFlow easily. To install TensorHub use the following commands.%
\index{model}%
\index{TensorFlow}%
\par%
\begin{tcolorbox}[size=title,title=Code,breakable]%
\begin{lstlisting}[language=Python, upquote=true]
!pip install tensorflow_hub\end{lstlisting}
\end{tcolorbox}%
It is also necessary to install TensorFlow Datasets, which you can install with the following command.%
\index{dataset}%
\index{TensorFlow}%
\par%
\begin{tcolorbox}[size=title,title=Code,breakable]%
\begin{lstlisting}[language=Python, upquote=true]
!pip install tensorflow_datasets\end{lstlisting}
\end{tcolorbox}%
Movie reviews are a good source of training data for sentiment analysis. These reviews are textual, and users give them a star rating which indicates if the viewer had a positive or negative experience with the movie. Load the Internet Movie DataBase (IMDB) reviews data set. This example is based on a TensorFlow example that you can%
\index{sentiment analysis}%
\index{TensorFlow}%
\index{training}%
\href{https://colab.research.google.com/github/tensorflow/hub/blob/master/examples/colab/tf2_text_classification.ipynb#scrollTo=2ew7HTbPpCJH}{ find here}%
.%
\par%
\begin{tcolorbox}[size=title,title=Code,breakable]%
\begin{lstlisting}[language=Python, upquote=true]
import tensorflow as tf
import tensorflow_hub as hub
import tensorflow_datasets as tfds

train_data, test_data = tfds.load(name="imdb_reviews", 
                                  split=["train", "test"], 
                                  batch_size=-1, as_supervised=True)

train_examples, train_labels = tfds.as_numpy(train_data)
test_examples, test_labels = tfds.as_numpy(test_data)

# /Users/jheaton/tensorflow_datasets/imdb_reviews/plain_text/0.1.0\end{lstlisting}
\end{tcolorbox}%
Load a pretrained embedding model called%
\index{model}%
\href{https://tfhub.dev/google/tf2-preview/gnews-swivel-20dim/1}{ gnews{-}swivel{-}20dim}%
.  Google trained this network on GNEWS data and can convert raw text into vectors.%
\index{vector}%
\par%
\begin{tcolorbox}[size=title,title=Code,breakable]%
\begin{lstlisting}[language=Python, upquote=true]
model = "https://tfhub.dev/google/tf2-preview/gnews-swivel-20dim/1"
hub_layer = hub.KerasLayer(model, output_shape=[20], input_shape=[], 
                           dtype=tf.string, trainable=True)\end{lstlisting}
\end{tcolorbox}%
The following code displays three movie reviews.  This display allows you to see the actual data.%
\par%
\begin{tcolorbox}[size=title,title=Code,breakable]%
\begin{lstlisting}[language=Python, upquote=true]
train_examples[:3]\end{lstlisting}
\tcbsubtitle[before skip=\baselineskip]{Output}%
\begin{lstlisting}[upquote=true]
array([b"This was an absolutely terrible movie. Don't be lured in by
Christopher Walken or Michael Ironside. Both are great actors, but
this must simply be their worst role in history. Even their great
acting could not redeem this movie's ridiculous storyline. This movie
is an early nineties US propaganda piece. The most pathetic scenes
were those when the Columbian rebels were making their cases for
revolutions. Maria Conchita Alonso appeared phony, and her pseudo-love
affair with Walken was nothing but a pathetic emotional plug in a
movie that was devoid of any real meaning. I am disappointed that
there are movies like this, ruining actor's like Christopher Walken's
good name. I could barely sit through it.",
       b'I have been known to fall asleep during films, but this is
usually due to a combination of things including, really tired, being
warm and comfortable on the sette and having just eaten a lot. However
on this occasion I fell asleep because the film was rubbish. The plot

...

rush. Mr. Mann and company appear to have mistaken Dawson City for
Deadwood, the Canadian North for the American Wild West.<br /><br
/>Canadian viewers be prepared for a Reefer Madness type of enjoyable
howl with this ludicrous plot, or, to shake your head in disgust.'],
      dtype=object)
\end{lstlisting}
\end{tcolorbox}%
The embedding layer can convert each to 20{-}number vectors, which the neural network receives as input in place of the actual words.%
\index{input}%
\index{layer}%
\index{neural network}%
\index{vector}%
\par%
\begin{tcolorbox}[size=title,title=Code,breakable]%
\begin{lstlisting}[language=Python, upquote=true]
hub_layer(train_examples[:3])\end{lstlisting}
\tcbsubtitle[before skip=\baselineskip]{Output}%
\begin{lstlisting}[upquote=true]
<tf.Tensor: shape=(3, 20), dtype=float32, numpy=
array([[ 1.7657859 , -3.882232  ,  3.913424  , -1.5557289 , -3.3362343
,
        -1.7357956 , -1.9954445 ,  1.298955  ,  5.081597  , -1.1041285
,
        -2.0503852 , -0.7267516 , -0.6567596 ,  0.24436145, -3.7208388
,
         2.0954835 ,  2.2969332 , -2.0689783 , -2.9489715 , -1.1315986
],
       [ 1.8804485 , -2.5852385 ,  3.4066994 ,  1.0982676 , -4.056685
,
        -4.891284  , -2.7855542 ,  1.3874227 ,  3.8476458 , -0.9256539
,
        -1.896706  ,  1.2113281 ,  0.11474716,  0.76209456, -4.8791065
,

...

        -2.2268343 ,  0.07446616, -1.4075902 , -0.706454  , -1.907037
,
         1.4419788 ,  1.9551864 , -0.42660046, -2.8022065 ,
0.43727067]],
      dtype=float32)>
\end{lstlisting}
\end{tcolorbox}%
We add additional layers to classify the movie reviews as either positive or negative.%
\index{layer}%
\par%
\begin{tcolorbox}[size=title,title=Code,breakable]%
\begin{lstlisting}[language=Python, upquote=true]
model = tf.keras.Sequential()
model.add(hub_layer)
model.add(tf.keras.layers.Dense(16, activation='relu'))
model.add(tf.keras.layers.Dense(1, activation='sigmoid'))

model.summary()\end{lstlisting}
\tcbsubtitle[before skip=\baselineskip]{Output}%
\begin{lstlisting}[upquote=true]
Model: "sequential"
_________________________________________________________________
 Layer (type)                Output Shape              Param #
=================================================================
 keras_layer (KerasLayer)    (None, 20)                400020
 dense (Dense)               (None, 16)                336
 dense_1 (Dense)             (None, 1)                 17
=================================================================
Total params: 400,373
Trainable params: 400,373
Non-trainable params: 0
_________________________________________________________________
\end{lstlisting}
\end{tcolorbox}%
We are now ready to compile the neural network. For this application, we use the adam training method for binary classification. We also save the initial random weights for later to start over easily.%
\index{ADAM}%
\index{classification}%
\index{neural network}%
\index{random}%
\index{training}%
\par%
\begin{tcolorbox}[size=title,title=Code,breakable]%
\begin{lstlisting}[language=Python, upquote=true]
model.compile(optimizer='adam',
              loss='binary_crossentropy',
              metrics=['accuracy'])
init_weights = model.get_weights()\end{lstlisting}
\end{tcolorbox}%
Before fitting, we split the training data into the train and validation sets.%
\index{training}%
\index{validation}%
\par%
\begin{tcolorbox}[size=title,title=Code,breakable]%
\begin{lstlisting}[language=Python, upquote=true]
x_val = train_examples[:10000]
partial_x_train = train_examples[10000:]

y_val = train_labels[:10000]
partial_y_train = train_labels[10000:]\end{lstlisting}
\end{tcolorbox}%
We can now fit the neural network. This fitting will run for 40 epochs and allow us to evaluate the effectiveness of the neural network, as measured by the training set.%
\index{neural network}%
\index{training}%
\par%
\begin{tcolorbox}[size=title,title=Code,breakable]%
\begin{lstlisting}[language=Python, upquote=true]
history = model.fit(partial_x_train,
                    partial_y_train,
                    epochs=40,
                    batch_size=512,
                    validation_data=(x_val, y_val),
                    verbose=1)\end{lstlisting}
\tcbsubtitle[before skip=\baselineskip]{Output}%
\begin{lstlisting}[upquote=true]
...
30/30 [==============================] - 1s 37ms/step - loss: 0.0711 -
accuracy: 0.9820 - val_loss: 0.3562 - val_accuracy: 0.8738
Epoch 40/40
30/30 [==============================] - 1s 37ms/step - loss: 0.0661 -
accuracy: 0.9847 - val_loss: 0.3626 - val_accuracy: 0.8728
\end{lstlisting}
\end{tcolorbox}%
\subsection{Benefits of Early Stopping}%
\label{subsec:BenefitsofEarlyStopping}%
While we used a validation set, we fit the neural network without early stopping. This dataset is complex enough to allow us to see the benefit of early stopping. We will examine how accuracy and loss progressed for training and validation sets. Loss measures the degree to which the neural network was confident in incorrect answers. Accuracy is the percentage of correct classifications, regardless of the neural network's confidence.%
\index{classification}%
\index{dataset}%
\index{early stopping}%
\index{neural network}%
\index{training}%
\index{validation}%
\par%
We begin by looking at the loss as we fit the neural network.%
\index{neural network}%
\par%
\begin{tcolorbox}[size=title,title=Code,breakable]%
\begin{lstlisting}[language=Python, upquote=true]
%matplotlib inline
import matplotlib.pyplot as plt

history_dict = history.history
acc = history_dict['accuracy']
val_acc = history_dict['val_accuracy']
loss = history_dict['loss']
val_loss = history_dict['val_loss']

epochs = range(1, len(acc) + 1)

plt.plot(epochs, loss, 'bo', label='Training loss')
plt.plot(epochs, val_loss, 'b', label='Validation loss')
plt.title('Training and validation loss')
plt.xlabel('Epochs')
plt.ylabel('Loss')
plt.legend()

plt.show()\end{lstlisting}
\tcbsubtitle[before skip=\baselineskip]{Output}%
\includegraphics[width=3in]{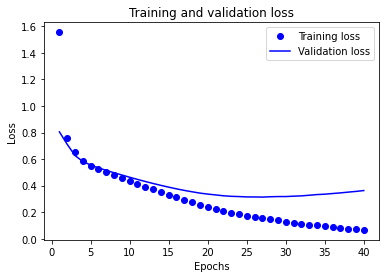}%
\end{tcolorbox}%
We can see that training and validation loss are similar early in the fitting. However, as fitting continues and overfitting sets in, training and validation loss diverge from each other. Training loss continues to fall consistently. However, once overfitting happens, the validation loss no longer falls and eventually begins to increase a bit. Early stopping, which we saw earlier in this course, can prevent some overfitting.%
\index{early stopping}%
\index{overfitting}%
\index{SOM}%
\index{training}%
\index{validation}%
\par%
\begin{tcolorbox}[size=title,title=Code,breakable]%
\begin{lstlisting}[language=Python, upquote=true]
plt.clf()   # clear figure

plt.plot(epochs, acc, 'bo', label='Training acc')
plt.plot(epochs, val_acc, 'b', label='Validation acc')
plt.title('Training and validation accuracy')
plt.xlabel('Epochs')
plt.ylabel('Accuracy')
plt.legend()

plt.show()\end{lstlisting}
\tcbsubtitle[before skip=\baselineskip]{Output}%
\includegraphics[width=3in]{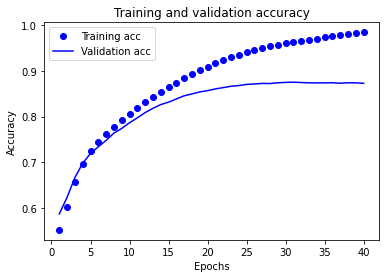}%
\end{tcolorbox}%
The accuracy graph tells a similar story. Now let's repeat the fitting with early stopping. We begin by creating an early stopping monitor and restoring the network's weights to random. Once this is complete, we can fit the neural network with the early stopping monitor enabled.%
\index{early stopping}%
\index{neural network}%
\index{random}%
\par%
\begin{tcolorbox}[size=title,title=Code,breakable]%
\begin{lstlisting}[language=Python, upquote=true]
from tensorflow.keras.callbacks import EarlyStopping

monitor = EarlyStopping(monitor='val_loss', min_delta=1e-3, 
        patience=5, verbose=1, mode='auto',
        restore_best_weights=True)

model.set_weights(init_weights)

history = model.fit(partial_x_train,
                    partial_y_train,
                    epochs=40,
                    batch_size=512,
                    callbacks=[monitor],
                    validation_data=(x_val, y_val),
                    verbose=1)\end{lstlisting}
\tcbsubtitle[before skip=\baselineskip]{Output}%
\begin{lstlisting}[upquote=true]
...
30/30 [==============================] - 1s 39ms/step - loss: 0.1475 -
accuracy: 0.9508 - val_loss: 0.3220 - val_accuracy: 0.8700
Epoch 34/40
29/30 [============================>.] - ETA: 0s - loss: 0.1419 -
accuracy: 0.9528Restoring model weights from the end of the best
epoch: 29.
30/30 [==============================] - 1s 38ms/step - loss: 0.1414 -
accuracy: 0.9531 - val_loss: 0.3231 - val_accuracy: 0.8704
Epoch 00034: early stopping
\end{lstlisting}
\end{tcolorbox}%
The training history chart is now shorter because we stopped earlier.%
\index{training}%
\par%
\begin{tcolorbox}[size=title,title=Code,breakable]%
\begin{lstlisting}[language=Python, upquote=true]
history_dict = history.history
acc = history_dict['accuracy']
val_acc = history_dict['val_accuracy']
loss = history_dict['loss']
val_loss = history_dict['val_loss']

epochs = range(1, len(acc) + 1)

plt.plot(epochs, loss, 'bo', label='Training loss')
plt.plot(epochs, val_loss, 'b', label='Validation loss')
plt.title('Training and validation loss')
plt.xlabel('Epochs')
plt.ylabel('Loss')
plt.legend()

plt.show()\end{lstlisting}
\tcbsubtitle[before skip=\baselineskip]{Output}%
\includegraphics[width=3in]{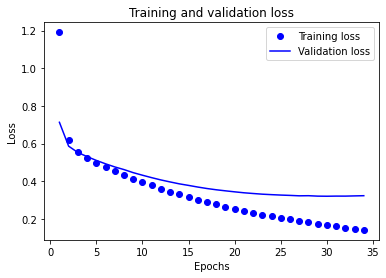}%
\end{tcolorbox}%
Finally, we evaluate the accuracy for the best neural network before early stopping occured.%
\index{early stopping}%
\index{neural network}%
\par%
\begin{tcolorbox}[size=title,title=Code,breakable]%
\begin{lstlisting}[language=Python, upquote=true]
from sklearn.metrics import accuracy_score
import numpy as np

pred = model.predict(x_val)
# Use 0.5 as the threshold
predict_classes = pred.flatten()>0.5

correct = accuracy_score(y_val,predict_classes)
print(f"Accuracy: {correct}")\end{lstlisting}
\tcbsubtitle[before skip=\baselineskip]{Output}%
\begin{lstlisting}[upquote=true]
Accuracy: 0.8685
\end{lstlisting}
\end{tcolorbox}

\section{Part 9.4: Transfer Learning for Facial Points and GANs}%
\label{sec:Part9.4TransferLearningforFacialPointsandGANs}%
I designed this notebook to work with Google Colab. You can run it locally; however, you might need to adjust some of the installation scripts contained in this notebook.%
\index{SOM}%
\par%
This part will see how we can use a 3rd party neural network to detect facial features, particularly the location of an individual's eyes. By locating eyes, we can crop portraits consistently. Previously, we saw that GANs could convert a random vector into a realistic{-}looking portrait. We can also perform the reverse and convert an actual photograph into a numeric vector. If we convert two images into these vectors, we can produce a video that transforms between the two images.%
\index{feature}%
\index{GAN}%
\index{neural network}%
\index{random}%
\index{vector}%
\index{video}%
\par%
NVIDIA trained StyleGAN on portraits consistently cropped with the eyes always in the same location. To successfully convert an image to a vector, we must crop the image similarly to how NVIDIA used cropping.%
\index{GAN}%
\index{NVIDIA}%
\index{StyleGAN}%
\index{vector}%
\par%
The code presented here allows you to choose a starting and ending image and use StyleGAN2 to produce a "morph" video between the two pictures. The preprocessing code will lock in on the exact positioning of each image, so your crop does not have to be perfect. The main point of your crop is for you to remove anything else that might be confused for a face. If multiple faces are detected, you will receive an error.%
\index{error}%
\index{GAN}%
\index{ROC}%
\index{ROC}%
\index{StyleGAN}%
\index{video}%
\par%
Also, make sure you have selected a GPU Runtime from CoLab. Choose "Runtime," then "Change Runtime Type," and choose GPU for "Hardware Accelerator."%
\index{GPU}%
\index{GPU}%
\par%
These settings allow you to change the high{-}level configuration. The number of steps determines how long your resulting video is. The video plays at 30 frames a second, so 150 is 5 seconds. You can also specify freeze steps to leave the video unchanged at the beginning and end. You will not likely need to change the network.%
\index{video}%
\par%
\begin{tcolorbox}[size=title,title=Code,breakable]%
\begin{lstlisting}[language=Python, upquote=true]
NETWORK = "https://nvlabs-fi-cdn.nvidia.com/"\
  "stylegan2-ada-pytorch/pretrained/ffhq.pkl"
STEPS = 150
FPS = 30
FREEZE_STEPS = 30\end{lstlisting}
\end{tcolorbox}%
\subsection{Upload Starting and Ending Images}%
\label{subsec:UploadStartingandEndingImages}%
We will begin by uploading a starting and ending image. The Colab service uploads these images. If you are running this code outside of Colab, these images are likely somewhere on your computer, and you provide the path to these files using the%
\index{SOM}%
\textbf{ SOURCE }%
and%
\textbf{ TARGET }%
variables.%
\par%
Choose your starting image.%
\par%
\begin{tcolorbox}[size=title,title=Code,breakable]%
\begin{lstlisting}[language=Python, upquote=true]
import os
from google.colab import files

uploaded = files.upload()

if len(uploaded) != 1:
  print("Upload exactly 1 file for source.")
else:
  for k, v in uploaded.items():
    _, ext = os.path.splitext(k)
    os.remove(k)
    SOURCE_NAME = f"source{ext}"
    open(SOURCE_NAME, 'wb').write(v)\end{lstlisting}
\end{tcolorbox}%
Also, choose your ending image.%
\par%
\begin{tcolorbox}[size=title,title=Code,breakable]%
\begin{lstlisting}[language=Python, upquote=true]
uploaded = files.upload()

if len(uploaded) != 1:
  print("Upload exactly 1 file for target.")
else:
  for k, v in uploaded.items():
    _, ext = os.path.splitext(k)
    os.remove(k)
    TARGET_NAME = f"target{ext}"
    open(TARGET_NAME, 'wb').write(v)\end{lstlisting}
\end{tcolorbox}

\subsection{Install Software}%
\label{subsec:InstallSoftware}%
Some software must be installed into Colab, for this notebook to work. We are specifically using these technologies:%
\index{SOM}%
\par%
\begin{itemize}[noitemsep]%
\item%
\href{https://arxiv.org/abs/2006.06676}{Training Generative Adversarial Networks with Limited Data}%
\end{itemize}%
Tero Karras, Miika Aittala, Janne Hellsten, Samuli Laine, Jaakko Lehtinen, Timo Aila%
\par%
\begin{itemize}[noitemsep]%
\item%
\href{https://www.cv-foundation.org/openaccess/content_cvpr_2014/papers/Kazemi_One_Millisecond_Face_2014_CVPR_paper.pdf}{One{-}millisecond face alignment with an ensemble of regression trees }%
Vahid Kazemi, Josephine Sullivan%
\end{itemize}%
\begin{tcolorbox}[size=title,title=Code,breakable]%
\begin{lstlisting}[language=Python, upquote=true]
!wget http://dlib.net/files/shape_predictor_5_face_landmarks.dat.bz2
!bzip2 -d shape_predictor_5_face_landmarks.dat.bz2\end{lstlisting}
\end{tcolorbox}%
\begin{tcolorbox}[size=title,title=Code,breakable]%
\begin{lstlisting}[language=Python, upquote=true]
import sys
!git clone https://github.com/NVlabs/stylegan2-ada-pytorch.git
!pip install ninja
sys.path.insert(0, "/content/stylegan2-ada-pytorch")\end{lstlisting}
\end{tcolorbox}

\subsection{Detecting Facial Features}%
\label{subsec:DetectingFacialFeatures}%
First, I will demonstrate how to detect the facial features we will use for consistent cropping and centering of the images. To accomplish this, we will use the%
\index{feature}%
\href{http://dlib.net/}{ dlib }%
package, a neural network library that gives us access to several pretrained models. The%
\index{model}%
\index{neural network}%
\href{https://github.com/davisking/dlib-models}{ DLIB Face Recognition ResNET Model V1 }%
is the model we will use; This is a 5{-}point landmarking model which identifies the corners of the eyes and bottom of the nose. The creators of this network trained it on the dlib 5{-}point face landmark dataset, which consists of 7198 faces.%
\index{dataset}%
\index{model}%
\par%
We begin by initializing dlib and loading the facial features neural network.%
\index{feature}%
\index{neural network}%
\par%
\begin{tcolorbox}[size=title,title=Code,breakable]%
\begin{lstlisting}[language=Python, upquote=true]
import cv2
import numpy as np
from PIL import Image
import dlib
from matplotlib import pyplot as plt

detector = dlib.get_frontal_face_detector()
predictor = dlib.shape_predictor('shape_predictor_5_face_landmarks.dat')\end{lstlisting}
\end{tcolorbox}%
Let's start by looking at the facial features of the source image. The following code detects the five facial features and displays their coordinates.%
\index{feature}%
\par%
\begin{tcolorbox}[size=title,title=Code,breakable]%
\begin{lstlisting}[language=Python, upquote=true]
img = cv2.imread(SOURCE_NAME)
if img is None:
    raise ValueError("Source image not found")

gray = cv2.cvtColor(img, cv2.COLOR_BGR2GRAY)
rects = detector(gray, 0)

if len(rects) == 0:
  raise ValueError("No faces detected")
elif len(rects) > 1:
  raise ValueError("Multiple faces detected")

shape = predictor(gray, rects[0])

w = img.shape[0]//50

for i in range(0, 5):
  pt1 = (shape.part(i).x, shape.part(i).y)
  pt2 = (shape.part(i).x+w, shape.part(i).y+w)
  cv2.rectangle(img,pt1,pt2,(0,255,255),4)
  print(pt1,pt2)\end{lstlisting}
\tcbsubtitle[before skip=\baselineskip]{Output}%
\begin{lstlisting}[upquote=true]
(1098, 546) (1128, 576)
(994, 554) (1024, 584)
(731, 556) (761, 586)
(833, 556) (863, 586)
(925, 729) (955, 759)
\end{lstlisting}
\end{tcolorbox}%
We can easily plot these features onto the source image. You can see the corners of the eyes and the base of the nose.%
\index{feature}%
\par%
\begin{tcolorbox}[size=title,title=Code,breakable]%
\begin{lstlisting}[language=Python, upquote=true]
img = cv2.cvtColor(img, cv2.COLOR_BGR2RGB)
plt.imshow(img)
plt.title('source')
plt.show()\end{lstlisting}
\tcbsubtitle[before skip=\baselineskip]{Output}%
\includegraphics[width=3in]{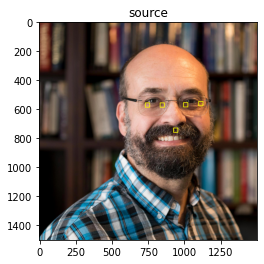}%
\end{tcolorbox}

\subsection{Preprocess Images for Best StyleGAN Results}%
\label{subsec:PreprocessImagesforBestStyleGANResults}%
Using dlib, we will center and crop the source and target image, using the eye positions as reference. I created two functions to accomplish this task. The first calls dlib and find the locations of the person's eyes. The second uses the eye locations to center the image around the eyes. We do not exactly center; we are offsetting slightly to center, similar to the original StyleGAN training set. I determined this offset by detecting the eyes of a generated StyleGAN face. The distance between the eyes gives us a means of telling how big the face is, which we use to scale the images consistently.%
\index{GAN}%
\index{StyleGAN}%
\index{training}%
\par%
\begin{tcolorbox}[size=title,title=Code,breakable]%
\begin{lstlisting}[language=Python, upquote=true]
def find_eyes(img):
  gray = cv2.cvtColor(img, cv2.COLOR_BGR2GRAY)
  rects = detector(gray, 0)
  
  if len(rects) == 0:
    raise ValueError("No faces detected")
  elif len(rects) > 1:
    raise ValueError("Multiple faces detected")

  shape = predictor(gray, rects[0])
  features = []

  for i in range(0, 5):
    features.append((i, (shape.part(i).x, shape.part(i).y)))

  return (int(features[3][1][0] + features[2][1][0]) // 2, \
    int(features[3][1][1] + features[2][1][1]) // 2), \
    (int(features[1][1][0] + features[0][1][0]) // 2, \
    int(features[1][1][1] + features[0][1][1]) // 2)

def crop_stylegan(img):
  left_eye, right_eye = find_eyes(img)
  # Calculate the size of the face
  d = abs(right_eye[0] - left_eye[0])
  z = 255/d
  # Consider the aspect ratio
  ar = img.shape[0]/img.shape[1]
  w = img.shape[1] * z
  img2 = cv2.resize(img, (int(w), int(w*ar)))
  bordersize = 1024
  img3 = cv2.copyMakeBorder(
      img2,
      top=bordersize,
      bottom=bordersize,
      left=bordersize,
      right=bordersize,
      borderType=cv2.BORDER_REPLICATE)

  left_eye2, right_eye2 = find_eyes(img3)

  # Adjust to the offset used by StyleGAN2
  crop1 = left_eye2[0] - 385 
  crop0 = left_eye2[1] - 490
  return img3[crop0:crop0+1024,crop1:crop1+1024]\end{lstlisting}
\end{tcolorbox}%
The following code will preprocess and crop your images.  If you receive an error indicating multiple faces were found, try to crop your image better or obscure the background.  If the program does not see a face, then attempt to obtain a clearer and more high{-}resolution image.%
\index{error}%
\index{ROC}%
\index{ROC}%
\par%
\begin{tcolorbox}[size=title,title=Code,breakable]%
\begin{lstlisting}[language=Python, upquote=true]
image_source = cv2.imread(SOURCE_NAME)
if image_source is None:
    raise ValueError("Source image not found")

image_target = cv2.imread(TARGET_NAME)
if image_target is None:
    raise ValueError("Source image not found")

cropped_source = crop_stylegan(image_source)
cropped_target = crop_stylegan(image_target)

img = cv2.cvtColor(cropped_source, cv2.COLOR_BGR2RGB)
plt.imshow(img)
plt.title('source')
plt.show()

img = cv2.cvtColor(cropped_target, cv2.COLOR_BGR2RGB)
plt.imshow(img)
plt.title('target')
plt.show()

cv2.imwrite("cropped_source.png", cropped_source)
cv2.imwrite("cropped_target.png", cropped_target)

#print(find_eyes(cropped_source))
#print(find_eyes(cropped_target))\end{lstlisting}
\tcbsubtitle[before skip=\baselineskip]{Output}%
\includegraphics[width=3in]{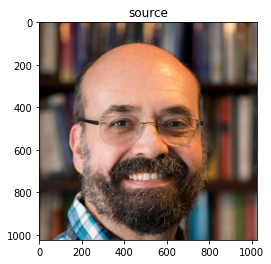}%
\includegraphics[width=3in]{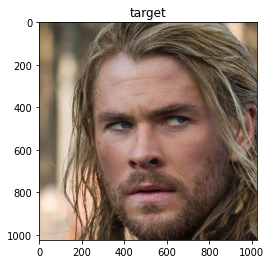}%
\begin{lstlisting}[upquote=true]
True
\end{lstlisting}
\end{tcolorbox}%
The two images are now 1024x1024 and cropped similarly to the ffhq dataset that NVIDIA used to train StyleGAN.%
\index{dataset}%
\index{GAN}%
\index{NVIDIA}%
\index{StyleGAN}%
\par

\subsection{Convert Source to a GAN}%
\label{subsec:ConvertSourcetoaGAN}%
We will use StyleGAN2, rather than the latest StyleGAN3, because StyleGAN2 contains a projector.py utility that converts images to latent vectors. StyleGAN3 does not have as good support for this%
\index{GAN}%
\index{latent vector}%
\index{StyleGAN}%
\index{vector}%
\href{https://github.com/NVlabs/stylegan3/issues/54}{ projection}%
. First, we convert the source to a GAN latent vector. This process will take several minutes.%
\index{GAN}%
\index{latent vector}%
\index{ROC}%
\index{ROC}%
\index{vector}%
\par%
\begin{tcolorbox}[size=title,title=Code,breakable]%
\begin{lstlisting}[language=Python, upquote=true]
cmd = f"python /content/stylegan2-ada-pytorch/projector.py "\
  f"--save-video 0 --num-steps 1000 --outdir=out_source "\
  f"--target=cropped_source.png --network={NETWORK}"
!{cmd}\end{lstlisting}
\end{tcolorbox}

\subsection{Convert Target to a GAN}%
\label{subsec:ConvertTargettoaGAN}%
Next, we convert the target to a GAN latent vector.  This process will also take several minutes.%
\index{GAN}%
\index{latent vector}%
\index{ROC}%
\index{ROC}%
\index{vector}%
\par%
\begin{tcolorbox}[size=title,title=Code,breakable]%
\begin{lstlisting}[language=Python, upquote=true]
cmd = f"python /content/stylegan2-ada-pytorch/projector.py "\
  f"--save-video 0 --num-steps 1000 --outdir=out_target "\
  f"--target=cropped_target.png --network={NETWORK}"
!{cmd}\end{lstlisting}
\end{tcolorbox}%
With the conversion complete, lets have a look at the two GANs.%
\index{GAN}%
\par%
\begin{tcolorbox}[size=title,title=Code,breakable]%
\begin{lstlisting}[language=Python, upquote=true]
img_gan_source = cv2.imread('/content/out_source/proj.png')
img = cv2.cvtColor(img_gan_source, cv2.COLOR_BGR2RGB)
plt.imshow(img)
plt.title('source-gan')
plt.show()\end{lstlisting}
\tcbsubtitle[before skip=\baselineskip]{Output}%
\includegraphics[width=3in]{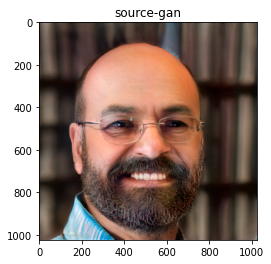}%
\end{tcolorbox}%
\begin{tcolorbox}[size=title,title=Code,breakable]%
\begin{lstlisting}[language=Python, upquote=true]
img_gan_target = cv2.imread('/content/out_target/proj.png')
img = cv2.cvtColor(img_gan_target, cv2.COLOR_BGR2RGB)
plt.imshow(img)
plt.title('target-gan')
plt.show()\end{lstlisting}
\tcbsubtitle[before skip=\baselineskip]{Output}%
\includegraphics[width=3in]{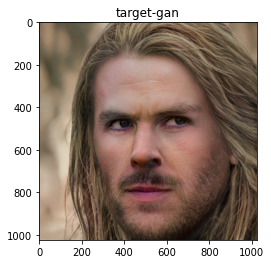}%
\end{tcolorbox}%
As you can see, the two GAN{-}generated images look similar to their real{-}world counterparts. However, they are by no means exact replicas.%
\index{GAN}%
\par

\subsection{Build the Video}%
\label{subsec:BuildtheVideo}%
The following code builds a transition video between the two latent vectors previously obtained.%
\index{latent vector}%
\index{vector}%
\index{video}%
\par%
\begin{tcolorbox}[size=title,title=Code,breakable]%
\begin{lstlisting}[language=Python, upquote=true]
import torch
import dnnlib
import legacy
import PIL.Image
import numpy as np
import imageio
from tqdm.notebook import tqdm

lvec1 = np.load('/content/out_source/projected_w.npz')['w']
lvec2 = np.load('/content/out_target/projected_w.npz')['w']

network_pkl = "https://nvlabs-fi-cdn.nvidia.com/stylegan2"\
  "-ada-pytorch/pretrained/ffhq.pkl"
device = torch.device('cuda')
with dnnlib.util.open_url(network_pkl) as fp:
    G = legacy.load_network_pkl(fp)['G_ema']\
      .requires_grad_(False).to(device) 

diff = lvec2 - lvec1
step = diff / STEPS
current = lvec1.copy()
target_uint8 = np.array([1024,1024,3], dtype=np.uint8)

video = imageio.get_writer('/content/movie.mp4', mode='I', fps=FPS, 
                           codec='libx264', bitrate='16M')

for j in tqdm(range(STEPS)):
  z = torch.from_numpy(current).to(device)
  synth_image = G.synthesis(z, noise_mode='const')
  synth_image = (synth_image + 1) * (255/2)
  synth_image = synth_image.permute(0, 2, 3, 1).clamp(0, 255)\
    .to(torch.uint8)[0].cpu().numpy()

  repeat = FREEZE_STEPS if j==0 or j==(STEPS-1) else 1
   
  for i in range(repeat):
    video.append_data(synth_image)
  current = current + step


video.close()\end{lstlisting}
\end{tcolorbox}

\subsection{Download your Video}%
\label{subsec:DownloadyourVideo}%
If you made it through all of these steps, you are now ready to download your video.%
\index{video}%
\par%
\begin{tcolorbox}[size=title,title=Code,breakable]%
\begin{lstlisting}[language=Python, upquote=true]
from google.colab import files
files.download("movie.mp4")\end{lstlisting}
\end{tcolorbox}

\section{Part 9.5: Transfer Learning for Keras Style Transfer}%
\label{sec:Part9.5TransferLearningforKerasStyleTransfer}%
In this part, we will implement style transfer. This technique takes two images as input and produces a third. The first image is the base image that we wish to transform. The second image represents the style we want to apply to the source image. Finally, the algorithm renders a third image that emulates the style characterized by the style image. This technique is called style transfer.%
\index{algorithm}%
\index{input}%
\cite{gatys2016image}%
\par%

\begin{figure}[h]%
\centering%
\includegraphics[width=4in]{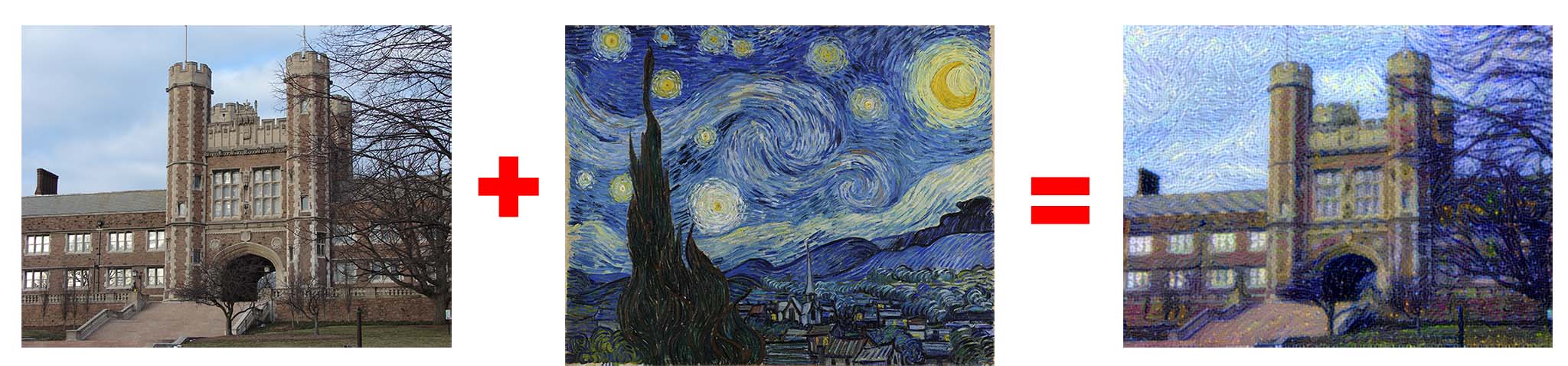}%
\caption{Style Transfer}%
\label{9.STYLE_TRANS}%
\end{figure}

\par%
I based the code presented in this part on a style transfer example in the Keras documentation created by%
\index{Keras}%
\href{https://keras.io/examples/generative/neural_style_transfer/}{ François Chollet}%
.%
\par%
We begin by uploading two images to Colab. If running this code locally, point these two filenames at the local copies of the images you wish to use.%
\par%
\begin{itemize}[noitemsep]%
\item%
\textbf{base\_image\_path }%
{-} The image to apply the style to.%
\item%
\textbf{style\_reference\_image\_path }%
{-} The image whose style we wish to copy.%
\end{itemize}%
First, we upload the base image.%
\par%
\begin{tcolorbox}[size=title,title=Code,breakable]%
\begin{lstlisting}[language=Python, upquote=true]
import os
from google.colab import files

uploaded = files.upload()

if len(uploaded) != 1:
  print("Upload exactly 1 file for source.")
else:
  for k, v in uploaded.items():
    _, ext = os.path.splitext(k)
    os.remove(k)
    base_image_path = f"source{ext}"
    open(base_image_path, 'wb').write(v)\end{lstlisting}
\end{tcolorbox}%
We also, upload the style image.%
\par%
\begin{tcolorbox}[size=title,title=Code,breakable]%
\begin{lstlisting}[language=Python, upquote=true]
uploaded = files.upload()

if len(uploaded) != 1:
  print("Upload exactly 1 file for target.")
else:
  for k, v in uploaded.items():
    _, ext = os.path.splitext(k)
    os.remove(k)
    style_reference_image_path = f"style{ext}"
    open(style_reference_image_path, 'wb').write(v)\end{lstlisting}
\end{tcolorbox}%
The loss function balances three different goals defined by the following three weights. Changing these weights allows you to fine{-}tune the image generation.%
\par%
\begin{itemize}[noitemsep]%
\item%
\textbf{total\_variation\_weight }%
{-} How much emphasis to place on the visual coherence of nearby pixels.%
\item%
\textbf{style\_weight }%
{-} How much emphasis to place on emulating the style of the reference image.%
\item%
\textbf{content\_weight }%
{-} How much emphasis to place on remaining close in appearance to the base image.%
\end{itemize}%
\begin{tcolorbox}[size=title,title=Code,breakable]%
\begin{lstlisting}[language=Python, upquote=true]
import numpy as np
import tensorflow as tf
from tensorflow import keras
from tensorflow.keras.applications import vgg19

result_prefix = "generated"

# Weights of the different loss components
total_variation_weight = 1e-6
style_weight = 1e-6
content_weight = 2.5e-8

# Dimensions of the generated picture.
width, height = keras.preprocessing.image.load_img(base_image_path).size
img_nrows = 400
img_ncols = int(width * img_nrows / height)\end{lstlisting}
\end{tcolorbox}%
We now display the two images we will use, first the base image followed by the style image.%
\par%
\begin{tcolorbox}[size=title,title=Code,breakable]%
\begin{lstlisting}[language=Python, upquote=true]
from IPython.display import Image, display

print("Source Image")
display(Image(base_image_path))\end{lstlisting}
\tcbsubtitle[before skip=\baselineskip]{Output}%
\includegraphics[width=4in]{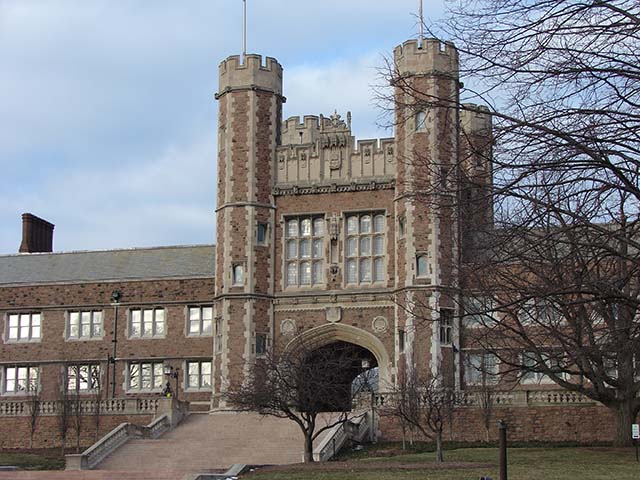}%
\begin{lstlisting}[upquote=true]
Source Image
\end{lstlisting}
\end{tcolorbox}%
\begin{tcolorbox}[size=title,title=Code,breakable]%
\begin{lstlisting}[language=Python, upquote=true]
print("Style Image")
display(Image(style_reference_image_path))\end{lstlisting}
\tcbsubtitle[before skip=\baselineskip]{Output}%
\includegraphics[width=4in]{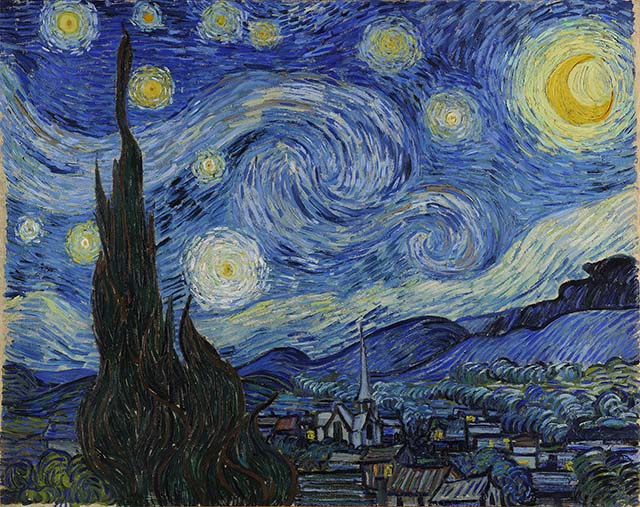}%
\begin{lstlisting}[upquote=true]
Style Image
\end{lstlisting}
\end{tcolorbox}%
\subsection{Image Preprocessing and Postprocessing}%
\label{subsec:ImagePreprocessingandPostprocessing}%
The preprocess\_image function begins by loading the image using Keras. We scale the image to the size specified by img\_nrows and img\_ncols. The img\_to\_array  converts the image to a Numpy array, to which we add dimension to account for batching. The dimensions expected by VGG are colors depth, height, width, and batch. Finally, we convert the Numpy array to a tensor.%
\index{Keras}%
\index{NumPy}%
\index{ROC}%
\index{ROC}%
\index{VGG}%
\par%
The deprocess\_image performs the reverse, transforming the output of the style transfer process back to a regular image. First, we reshape the image to remove the batch dimension. Next, The outputs are moved back into the 0{-}255 range by adding the mean value of the RGB colors. We must also convert the BGR (blue, green, red) colorspace of VGG to the more standard RGB encoding.%
\index{output}%
\index{ROC}%
\index{ROC}%
\index{VGG}%
\par%
\begin{tcolorbox}[size=title,title=Code,breakable]%
\begin{lstlisting}[language=Python, upquote=true]
def preprocess_image(image_path):
    # Util function to open, resize and format 
    # pictures into appropriate tensors
    img = keras.preprocessing.image.load_img(
        image_path, target_size=(img_nrows, img_ncols)
    )
    img = keras.preprocessing.image.img_to_array(img)
    img = np.expand_dims(img, axis=0)
    img = vgg19.preprocess_input(img)
    return tf.convert_to_tensor(img)


def deprocess_image(x):
    # Util function to convert a tensor into a valid image
    x = x.reshape((img_nrows, img_ncols, 3))
    # Remove zero-center by mean pixel
    x[:, :, 0] += 103.939
    x[:, :, 1] += 116.779
    x[:, :, 2] += 123.68
    # 'BGR'->'RGB'
    x = x[:, :, ::-1]
    x = np.clip(x, 0, 255).astype("uint8")
    return x\end{lstlisting}
\end{tcolorbox}

\subsection{Calculating the Style, Content, and Variation Loss}%
\begin{figure}[h]%
\centering%
\includegraphics[width=4in]{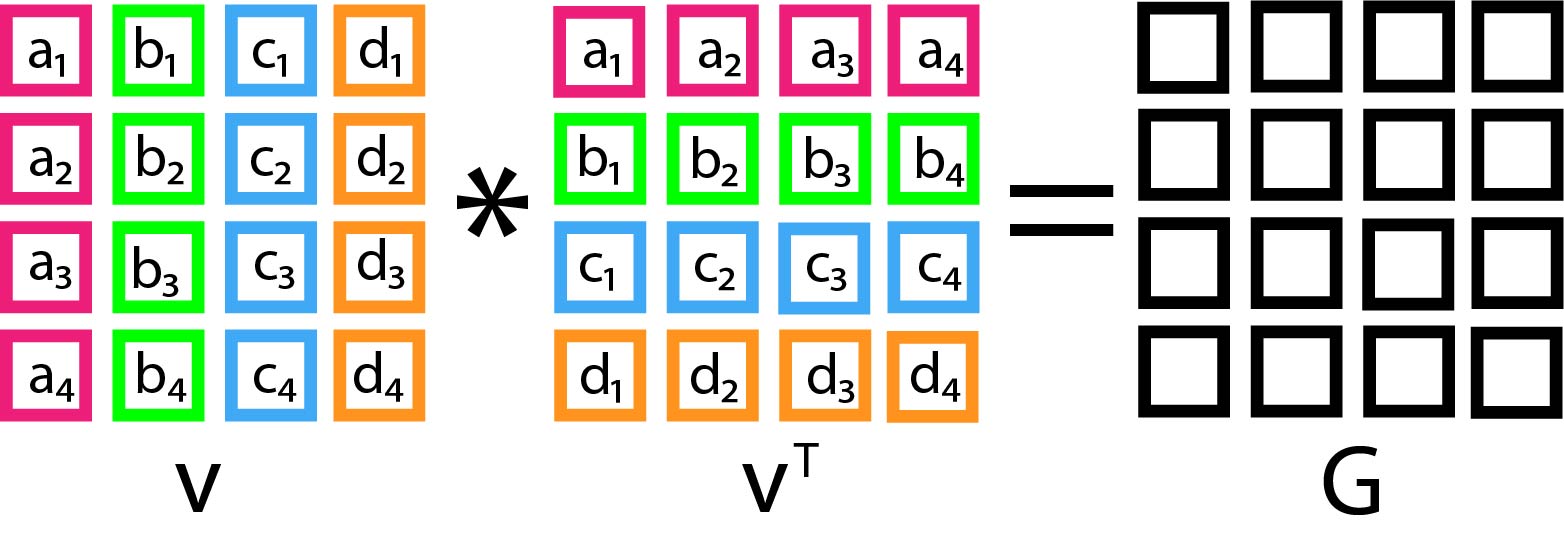}%
\caption{The Gram Matrix}%
\label{9.GRAM}%
\end{figure}

\label{subsec:CalculatingtheStyle,Content,andVariationLoss}%
Before we see how to calculate the 3{-}part loss function, I must introduce the Gram matrix's mathematical concept. Figure \ref{9.GRAM} demonstrates this concept.%
\index{matrix}%
\par%

\par%
We calculate the Gram matrix by multiplying a matrix by its transpose. To calculate two parts of the loss function, we will take the Gram matrix of the outputs from several convolution layers in the VGG network. To determine both style, and similarity to the original image, we will compare the convolution layer output of VGG rather than directly comparing the image pixels. In the third part of the loss function, we will directly compare pixels near each other.%
\index{convolution}%
\index{layer}%
\index{matrix}%
\index{output}%
\index{VGG}%
\par%
Because we are taking convolution output from several different levels of the VGG network, the Gram matrix provides a means of combining these layers. The Gram matrix of the VGG convolution layers represents the style of the image. We will calculate this style for the original image, the style{-}reference image, and the final output image as the algorithm generates it.%
\index{algorithm}%
\index{convolution}%
\index{layer}%
\index{matrix}%
\index{output}%
\index{VGG}%
\par%
\begin{tcolorbox}[size=title,title=Code,breakable]%
\begin{lstlisting}[language=Python, upquote=true]
# The gram matrix of an image tensor (feature-wise outer product)
def gram_matrix(x):
    x = tf.transpose(x, (2, 0, 1))
    features = tf.reshape(x, (tf.shape(x)[0], -1))
    gram = tf.matmul(features, tf.transpose(features))
    return gram


# The "style loss" is designed to maintain
# the style of the reference image in the generated image.
# It is based on the gram matrices (which capture style) of
# feature maps from the style reference image
# and from the generated image
def style_loss(style, combination):
    S = gram_matrix(style)
    C = gram_matrix(combination)
    channels = 3
    size = img_nrows * img_ncols
    return tf.reduce_sum(tf.square(S - C)) /\
      (4.0 * (channels ** 2) * (size ** 2))


# An auxiliary loss function
# designed to maintain the "content" of the
# base image in the generated image
def content_loss(base, combination):
    return tf.reduce_sum(tf.square(combination - base))


# The 3rd loss function, total variation loss,
# designed to keep the generated image locally coherent
def total_variation_loss(x):
    a = tf.square(
        x[:, : img_nrows - 1, : img_ncols - 1, :] \
          - x[:, 1:, : img_ncols - 1, :]
    )
    b = tf.square(
        x[:, : img_nrows - 1, : img_ncols - 1, :] \
          - x[:, : img_nrows - 1, 1:, :]
    )
    return tf.reduce_sum(tf.pow(a + b, 1.25))\end{lstlisting}
\end{tcolorbox}%
The style\_loss function compares how closely the current generated image (combination) matches the style of the reference style image. The Gram matrixes of the style and current generated image are subtracted and normalized to calculate this difference in style. Precisely, it consists in a sum of L2 distances between the Gram matrices of the representations of the base image and the style reference image, extracted from different layers of VGG. The general idea is to capture color/texture information at different spatial scales (fairly large scales, as defined by the depth of the layer considered).%
\index{L2}%
\index{layer}%
\index{matrix}%
\index{VGG}%
\par%
The content\_loss function compares how closely the current generated image matches the original image. You must subtract Gram matrixes of the original and generated images to calculate this difference. Here we calculate the L2 distance between the base image's VGG features and the generated image's features, keeping the generated image close enough to the original one.%
\index{feature}%
\index{L2}%
\index{matrix}%
\index{VGG}%
\par%
Finally, the total\_variation\_loss function imposes local spatial continuity between the pixels of the generated image, giving it visual coherence.%
\par

\subsection{The VGG Neural Network}%
\label{subsec:TheVGGNeuralNetwork}%
VGG19 is a convolutional neural network model proposed by K. Simonyan and A. Zisserman.%
\index{convolution}%
\index{convolutional}%
\index{model}%
\index{neural network}%
\index{VGG}%
\cite{simonyan2014very}%
The model achieves 92.7\% top{-}5 test accuracy in ImageNet, a dataset of over 14 million images belonging to 1000 classes. We will transfer the VGG16 weights into our style transfer model. Keras provides functions to load the VGG neural network.%
\index{dataset}%
\index{Keras}%
\index{model}%
\index{neural network}%
\index{VGG}%
\par%
\begin{tcolorbox}[size=title,title=Code,breakable]%
\begin{lstlisting}[language=Python, upquote=true]
# Build a VGG19 model loaded with pre-trained ImageNet weights
model = vgg19.VGG19(weights="imagenet", include_top=False)

# Get the symbolic outputs of each "key" layer (we gave them unique names).
outputs_dict = dict([(layer.name, layer.output) for layer in model.layers])

# Set up a model that returns the activation values for every layer in
# VGG19 (as a dict).
feature_extractor = keras.Model(inputs=model.inputs, outputs=outputs_dict)\end{lstlisting}
\end{tcolorbox}%
We can now generate the complete loss function. The following images are input to the compute\_loss function:%
\index{input}%
\par%
\begin{itemize}[noitemsep]%
\item%
\textbf{combination\_image }%
{-} The current iteration of the generated image.%
\index{iteration}%
\item%
\textbf{base\_image }%
{-} The starting image.%
\item%
\textbf{style\_reference\_image }%
{-} The image that holds the style to reproduce.%
\end{itemize}%
The layers specified by style\_layer\_names indicate which layers should be extracted as features from VGG for each of the three images.%
\index{feature}%
\index{layer}%
\index{VGG}%
\par%
\begin{tcolorbox}[size=title,title=Code,breakable]%
\begin{lstlisting}[language=Python, upquote=true]
# List of layers to use for the style loss.
style_layer_names = [
    "block1_conv1",
    "block2_conv1",
    "block3_conv1",
    "block4_conv1",
    "block5_conv1",
]
# The layer to use for the content loss.
content_layer_name = "block5_conv2"


def compute_loss(combination_image, base_image, style_reference_image):
    input_tensor = tf.concat(
        [base_image, style_reference_image, combination_image], axis=0
    )
    features = feature_extractor(input_tensor)

    # Initialize the loss
    loss = tf.zeros(shape=())

    # Add content loss
    layer_features = features[content_layer_name]
    base_image_features = layer_features[0, :, :, :]
    combination_features = layer_features[2, :, :, :]
    loss = loss + content_weight * content_loss(
        base_image_features, combination_features
    )
    # Add style loss
    for layer_name in style_layer_names:
        layer_features = features[layer_name]
        style_reference_features = layer_features[1, :, :, :]
        combination_features = layer_features[2, :, :, :]
        sl = style_loss(style_reference_features, combination_features)
        loss += (style_weight / len(style_layer_names)) * sl

    # Add total variation loss
    loss += total_variation_weight * \
      total_variation_loss(combination_image)
    return loss\end{lstlisting}
\end{tcolorbox}

\subsection{Generating the Style Transferred Image}%
\label{subsec:GeneratingtheStyleTransferredImage}%
The compute\_loss\_and\_grads function calls the loss function and computes the gradients. The parameters of this model are the actual RGB values of the current iteration of the generated images. These parameters start with the base image, and the algorithm optimizes them to the final rendered image. We are not training a model to perform the transformation; we are training/modifying the image to minimize the loss functions. We utilize gradient tape to allow Keras to modify the image in the same way the neural network training modifies weights.%
\index{algorithm}%
\index{gradient}%
\index{iteration}%
\index{Keras}%
\index{model}%
\index{neural network}%
\index{parameter}%
\index{training}%
\par%
\begin{tcolorbox}[size=title,title=Code,breakable]%
\begin{lstlisting}[language=Python, upquote=true]
@tf.function
def compute_loss_and_grads(combination_image, \
                  base_image, style_reference_image):
    with tf.GradientTape() as tape:
        loss = compute_loss(combination_image, \
                base_image, style_reference_image)
    grads = tape.gradient(loss, combination_image)
    return loss, grads\end{lstlisting}
\end{tcolorbox}%
We can now optimize the image according to the loss function.%
\par%
\begin{tcolorbox}[size=title,title=Code,breakable]%
\begin{lstlisting}[language=Python, upquote=true]
optimizer = keras.optimizers.SGD(
    keras.optimizers.schedules.ExponentialDecay(
        initial_learning_rate=100.0, decay_steps=100, decay_rate=0.96
    )
)

base_image = preprocess_image(base_image_path)
style_reference_image = preprocess_image(style_reference_image_path)
combination_image = tf.Variable(preprocess_image(base_image_path))

iterations = 4000
for i in range(1, iterations + 1):
    loss, grads = compute_loss_and_grads(
        combination_image, base_image, style_reference_image
    )
    optimizer.apply_gradients([(grads, combination_image)])
    if i % 100 == 0:
        print("Iteration %d: loss=%.2f" % (i, loss))
        img = deprocess_image(combination_image.numpy())
        fname = result_prefix + "_at_iteration_%d.png" % i
        keras.preprocessing.image.save_img(fname, img)\end{lstlisting}
\tcbsubtitle[before skip=\baselineskip]{Output}%
\begin{lstlisting}[upquote=true]
Iteration 100: loss=4890.20
Iteration 200: loss=3527.19
Iteration 300: loss=3022.59
Iteration 400: loss=2751.59
Iteration 500: loss=2578.63
Iteration 600: loss=2457.19
Iteration 700: loss=2366.39
Iteration 800: loss=2295.66
Iteration 900: loss=2238.67
Iteration 1000: loss=2191.59
Iteration 1100: loss=2151.88
Iteration 1200: loss=2117.95
Iteration 1300: loss=2088.56
Iteration 1400: loss=2062.86
Iteration 1500: loss=2040.14

...

Iteration 3600: loss=1840.82
Iteration 3700: loss=1836.87
Iteration 3800: loss=1833.16
Iteration 3900: loss=1829.65
Iteration 4000: loss=1826.34
\end{lstlisting}
\end{tcolorbox}%
We can display the image.%
\par%
\begin{tcolorbox}[size=title,title=Code,breakable]%
\begin{lstlisting}[language=Python, upquote=true]
display(Image(result_prefix + "_at_iteration_4000.png"))\end{lstlisting}
\tcbsubtitle[before skip=\baselineskip]{Output}%
\includegraphics[width=4in]{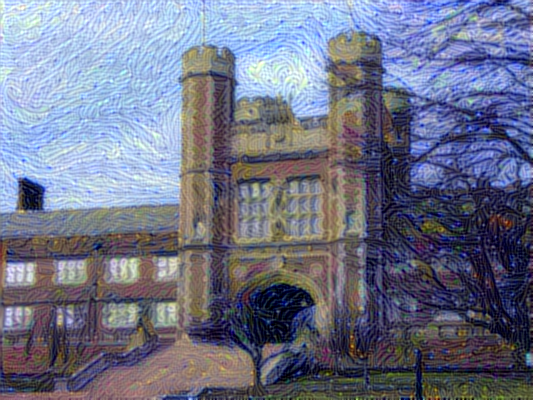}%
\end{tcolorbox}%
We can download this image.%
\par%
\begin{tcolorbox}[size=title,title=Code,breakable]%
\begin{lstlisting}[language=Python, upquote=true]
from google.colab import files
files.download(result_prefix + "_at_iteration_4000.png")\end{lstlisting}
\end{tcolorbox}

\chapter{Time Series in Keras}%
\label{chap:TimeSeriesinKeras}%
\section{Part 10.1: Time Series Data Encoding}%
\label{sec:Part10.1TimeSeriesDataEncoding}%
There are many different methods to encode data over time to a neural network. In this chapter, we will examine time series encoding and recurrent networks, two topics that are logical to put together because they are both methods for dealing with data that spans over time. Time series encoding deals with representing events that occur over time to a neural network. This encoding is necessary because a feedforward neural network will always produce the same output vector for a given input vector. Recurrent neural networks do not require encoding time series data because they can automatically handle data that occur over time.%
\index{feedforward}%
\index{input}%
\index{input vector}%
\index{neural network}%
\index{output}%
\index{recurrent}%
\index{recurrent network}%
\index{vector}%
\par%
The variation in temperature during the week is an example of time{-}series data. For instance, if we know that today's temperature is 25 degrees Fahrenheit and tomorrow's temperature is 27 degrees, the recurrent neural networks and time series encoding provide another option to predict the correct temperature for the week. Conversely, a traditional feedforward neural network will always respond with the same output for a given input. If we train a feedforward neural network to predict tomorrow's temperature, it should return a value of 27 for 25. It will always output 27 when given 25 might hinder its predictions. Surely the temperature of 27 will not always follow 25. It would be better for the neural network to consider the temperatures for days before the prediction. Perhaps the temperature over the last week might allow us to predict tomorrow's temperature. Therefore, recurrent neural networks and time series encoding represent two different approaches to representing data over time to a neural network.%
\index{feedforward}%
\index{input}%
\index{neural network}%
\index{output}%
\index{predict}%
\index{recurrent}%
\index{time{-}series}%
\par%
Previously we trained neural networks with input ($x$) and expected output ($y$). $X$ was a matrix, the rows were training examples, and the columns were values to be predicted. The $x$ value will now contain sequences of data. The definition of the $y$ value will stay the same.%
\index{input}%
\index{matrix}%
\index{neural network}%
\index{output}%
\index{predict}%
\index{training}%
\par%
Dimensions of the training set ($x$):%
\index{training}%
\par%
\begin{itemize}[noitemsep]%
\item%
Axis 1: Training set elements (sequences) (must be of the same size as $y$ size)%
\index{axis}%
\index{training}%
\item%
Axis 2: Members of sequence%
\index{axis}%
\item%
Axis 3: Features in data (like input neurons)%
\index{axis}%
\index{feature}%
\index{input}%
\index{input neuron}%
\index{neuron}%
\end{itemize}%
Previously, we might take as input a single stock price to predict if we should buy (1), sell ({-}1), or hold (0). The following code illustrates this encoding.%
\index{input}%
\index{predict}%
\par%
\begin{tcolorbox}[size=title,title=Code,breakable]%
\begin{lstlisting}[language=Python, upquote=true]
x = [
    [32],
    [41],
    [39],
    [20],
    [15]
]

y = [
    1,
    -1,
    0,
    -1,
    1
]

print(x)
print(y)\end{lstlisting}
\tcbsubtitle[before skip=\baselineskip]{Output}%
\begin{lstlisting}[upquote=true]
[[32], [41], [39], [20], [15]]
[1, -1, 0, -1, 1]
\end{lstlisting}
\end{tcolorbox}%
The following code builds a CSV file from scratch. To see it as a data frame, use the following:%
\index{CSV}%
\par%
\begin{tcolorbox}[size=title,title=Code,breakable]%
\begin{lstlisting}[language=Python, upquote=true]
from IPython.display import display, HTML
import pandas as pd
import numpy as np

x = np.array(x)
print(x[:,0])


df = pd.DataFrame({'x':x[:,0], 'y':y})
display(df)\end{lstlisting}
\tcbsubtitle[before skip=\baselineskip]{Output}%
\begin{tabular}[hbt!]{l|l|l}%
\hline%
&x&y\\%
\hline%
0&32&1\\%
1&41&{-}1\\%
2&39&0\\%
3&20&{-}1\\%
4&15&1\\%
\hline%
\end{tabular}%
\vspace{2mm}%
\begin{lstlisting}[upquote=true]
[32 41 39 20 15]
\end{lstlisting}
\end{tcolorbox}%
You might want to put volume in with the stock price.  The following code shows how to add a dimension to handle the volume.%
\par%
\begin{tcolorbox}[size=title,title=Code,breakable]%
\begin{lstlisting}[language=Python, upquote=true]
x = [
    [32,1383],
    [41,2928],
    [39,8823],
    [20,1252],
    [15,1532]
]

y = [
    1,
    -1,
    0,
    -1,
    1
]

print(x)
print(y)\end{lstlisting}
\tcbsubtitle[before skip=\baselineskip]{Output}%
\begin{lstlisting}[upquote=true]
[[32, 1383], [41, 2928], [39, 8823], [20, 1252], [15, 1532]]
[1, -1, 0, -1, 1]
\end{lstlisting}
\end{tcolorbox}%
Again, very similar to what we did before.  The following shows this as a data frame.%
\par%
\begin{tcolorbox}[size=title,title=Code,breakable]%
\begin{lstlisting}[language=Python, upquote=true]
from IPython.display import display, HTML
import pandas as pd
import numpy as np

x = np.array(x)
print(x[:,0])


df = pd.DataFrame({'price':x[:,0], 'volume':x[:,1], 'y':y})
display(df)\end{lstlisting}
\tcbsubtitle[before skip=\baselineskip]{Output}%
\begin{tabular}[hbt!]{l|l|l|l}%
\hline%
&price&volume&y\\%
\hline%
0&32&1383&1\\%
1&41&2928&{-}1\\%
2&39&8823&0\\%
3&20&1252&{-}1\\%
4&15&1532&1\\%
\hline%
\end{tabular}%
\vspace{2mm}%
\begin{lstlisting}[upquote=true]
[32 41 39 20 15]
\end{lstlisting}
\end{tcolorbox}%
Now we get to sequence format. We want to predict something over a sequence, so the data format needs to add a dimension. You must specify a maximum sequence length. The individual sequences can be of any size.%
\index{predict}%
\index{SOM}%
\par%
\begin{tcolorbox}[size=title,title=Code,breakable]%
\begin{lstlisting}[language=Python, upquote=true]
x = [
    [[32,1383],[41,2928],[39,8823],[20,1252],[15,1532]],
    [[35,8272],[32,1383],[41,2928],[39,8823],[20,1252]],
    [[37,2738],[35,8272],[32,1383],[41,2928],[39,8823]],
    [[34,2845],[37,2738],[35,8272],[32,1383],[41,2928]],
    [[32,2345],[34,2845],[37,2738],[35,8272],[32,1383]],
]

y = [
    1,
    -1,
    0,
    -1,
    1
]

print(x)
print(y)\end{lstlisting}
\tcbsubtitle[before skip=\baselineskip]{Output}%
\begin{lstlisting}[upquote=true]
[[[32, 1383], [41, 2928], [39, 8823], [20, 1252], [15, 1532]], [[35,
8272], [32, 1383], [41, 2928], [39, 8823], [20, 1252]], [[37, 2738],
[35, 8272], [32, 1383], [41, 2928], [39, 8823]], [[34, 2845], [37,
2738], [35, 8272], [32, 1383], [41, 2928]], [[32, 2345], [34, 2845],
[37, 2738], [35, 8272], [32, 1383]]]
[1, -1, 0, -1, 1]
\end{lstlisting}
\end{tcolorbox}%
Even if there is only one feature (price), you must use 3 dimensions.%
\index{feature}%
\par%
\begin{tcolorbox}[size=title,title=Code,breakable]%
\begin{lstlisting}[language=Python, upquote=true]
x = [
    [[32],[41],[39],[20],[15]],
    [[35],[32],[41],[39],[20]],
    [[37],[35],[32],[41],[39]],
    [[34],[37],[35],[32],[41]],
    [[32],[34],[37],[35],[32]],
]

y = [
    1,
    -1,
    0,
    -1,
    1
]

print(x)
print(y)\end{lstlisting}
\tcbsubtitle[before skip=\baselineskip]{Output}%
\begin{lstlisting}[upquote=true]
[[[32], [41], [39], [20], [15]], [[35], [32], [41], [39], [20]],
[[37], [35], [32], [41], [39]], [[34], [37], [35], [32], [41]], [[32],
[34], [37], [35], [32]]]
[1, -1, 0, -1, 1]
\end{lstlisting}
\end{tcolorbox}%
\subsection{Module 10 Assignment}%
\label{subsec:Module10Assignment}%
You can find the first assignment here:%
\href{https://github.com/jeffheaton/t81_558_deep_learning/blob/master/assignments/assignment_yourname_class10.ipynb}{ assignment 10}%
\par

\section{Part 10.2: Programming LSTM with Keras and TensorFlow}%
\label{sec:Part10.2ProgrammingLSTMwithKerasandTensorFlow}%
So far, the neural networks that we've examined have always had forward connections. Neural networks of this type always begin with an input layer connected to the first hidden layer. Each hidden layer always connects to the next hidden layer. The final hidden layer always connects to the output layer. This manner of connection is why these networks are called "feedforward."  Recurrent neural networks are not as rigid, as backward linkages are also allowed. A recurrent connection links a neuron in a layer to either a previous layer or the neuron itself. Most recurrent neural network architectures maintain the state in the recurrent connections. Feedforward neural networks don't keep any state.%
\index{architecture}%
\index{connection}%
\index{feedforward}%
\index{hidden layer}%
\index{input}%
\index{input layer}%
\index{layer}%
\index{link}%
\index{network architecture}%
\index{neural network}%
\index{neuron}%
\index{output}%
\index{output layer}%
\index{recurrent}%
\par%
\subsection{Understanding LSTM}%
\label{subsec:UnderstandingLSTM}%
Long Short Term Memory (LSTM) layers are a type of recurrent unit that you often use with deep neural networks.%
\index{layer}%
\index{long short term memory}%
\index{LSTM}%
\index{neural network}%
\index{recurrent}%
\cite{hochreiter1997long}%
For TensorFlow, you can think of LSTM as a layer type that you can combine with other layer types, such as dense. LSTM makes use of two transfer function types internally.%
\index{layer}%
\index{LSTM}%
\index{TensorFlow}%
\par%
The first type of transfer function is the sigmoid.  This transfer function type is used form gates inside of the unit.  The sigmoid transfer function is given by the following equation:%
\index{sigmoid}%
\par%
\vspace{2mm}%
\begin{equation*}
\mbox{S}(t) = \frac{1}{1 + e^{-t}}
\end{equation*}
\vspace{2mm}%
\par%
The second type of transfer function is the hyperbolic tangent (tanh) function, which allows you to scale the output of the LSTM. This functionality is similar to how we have used other transfer functions in this course.%
\index{hyperbolic tangent}%
\index{LSTM}%
\index{output}%
\par%
We provide the graphs for these functions here:%
\par%
\begin{tcolorbox}[size=title,title=Code,breakable]%
\begin{lstlisting}[language=Python, upquote=true]
%matplotlib inline

import matplotlib
import numpy as np
import matplotlib.pyplot as plt
import math

def sigmoid(x):
    a = []
    for item in x:
        a.append(1/(1+math.exp(-item)))
    return a

def f2(x):
    a = []
    for item in x:
        a.append(math.tanh(item))
    return a

x = np.arange(-10., 10., 0.2)
y1 = sigmoid(x)
y2 = f2(x)

print("Sigmoid")
plt.plot(x,y1)
plt.show()

print("Hyperbolic Tangent(tanh)")
plt.plot(x,y2)
plt.show()\end{lstlisting}
\tcbsubtitle[before skip=\baselineskip]{Output}%
\includegraphics[width=3in]{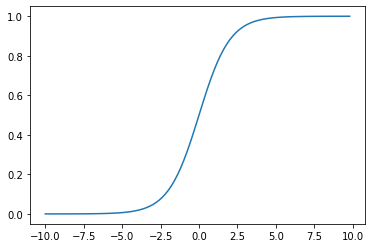}%
\begin{lstlisting}[upquote=true]
Sigmoid
\end{lstlisting}
\includegraphics[width=3in]{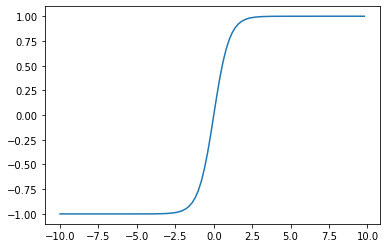}%
\begin{lstlisting}[upquote=true]
Hyperbolic Tangent(tanh)
\end{lstlisting}
\end{tcolorbox}%
Both of these two functions compress their output to a specific range.  For the sigmoid function, this range is 0 to 1.  For the hyperbolic tangent function, this range is {-}1 to 1.%
\index{hyperbolic tangent}%
\index{output}%
\index{sigmoid}%
\par
\begin{figure}[h]%
\centering%
\includegraphics[width=4in]{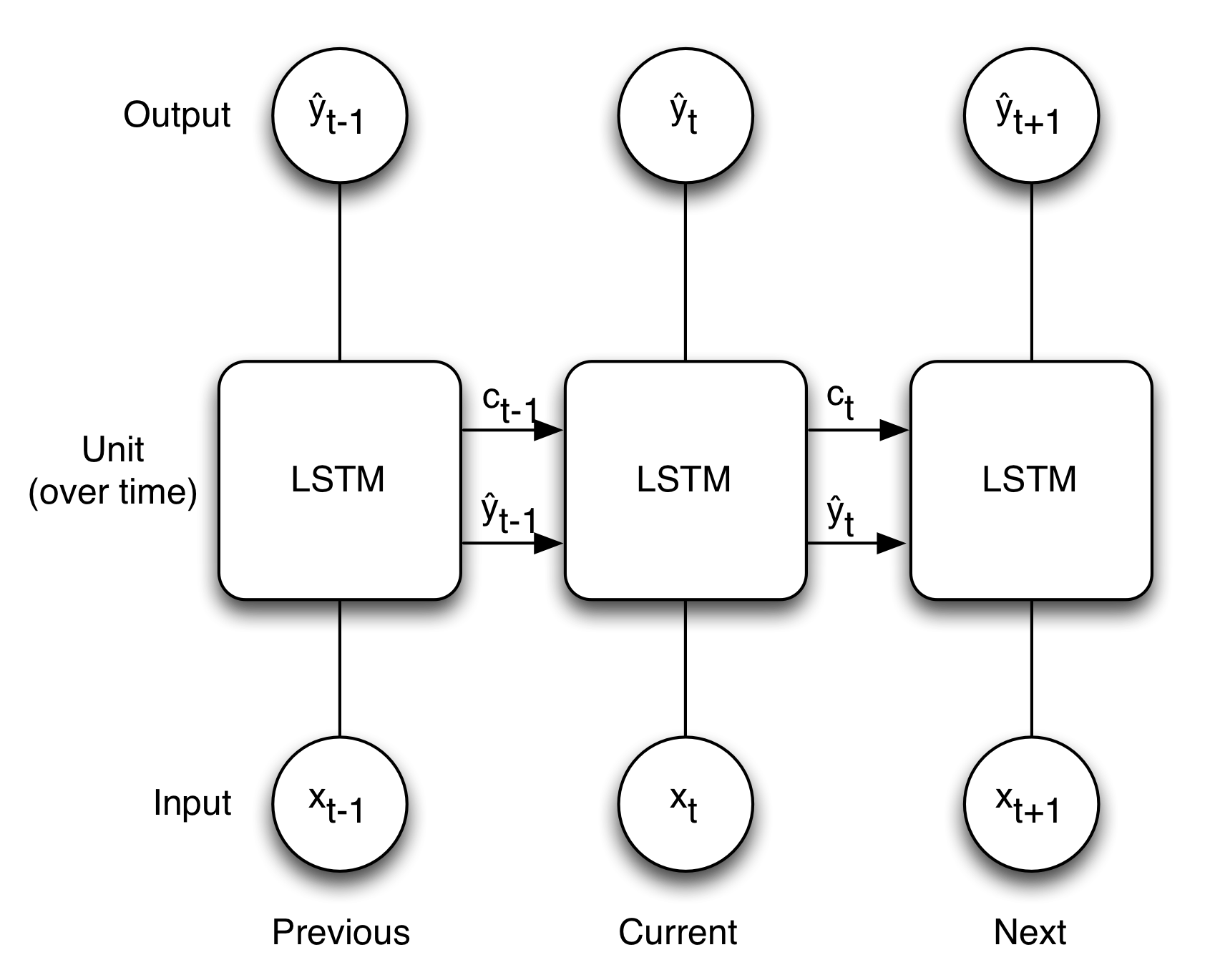}%
\caption{LSTM Layers}%
\label{10.LSTM}%
\end{figure}

%
LSTM maintains an internal state and produces an output.  The following diagram shows an LSTM unit over three timeslices: the current time slice (t), as well as the previous (t{-}1) and next (t+1) slice, as demonstrated by Figure \ref{10.LSTM}.%
\index{LSTM}%
\index{output}%
\par%

\par
\begin{figure}[h]%
\centering%
\includegraphics[width=4in]{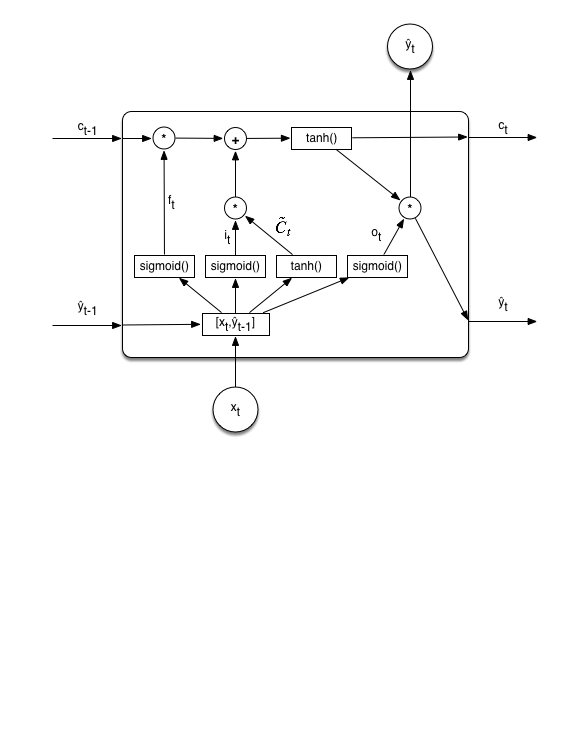}%
\caption{Inside a LSTM Layer}%
\label{10.ILSTM}%
\end{figure}

%
The values $\hat{y}$ are the output from the unit; the values ($x$) are the input to the unit, and the values $c$ are the context values.  The output and context values always feed their output to the next time slice.  The context values allow the network to maintain the state between calls.  Figure \ref{10.ILSTM} shows the internals of a LSTM layer.%
\index{context}%
\index{input}%
\index{layer}%
\index{LSTM}%
\index{output}%
\par%

\par%
A LSTM unit consists of three gates:%
\index{LSTM}%
\par%
\begin{itemize}[noitemsep]%
\item%
Forget Gate ($f_t$) {-} Controls if/when the context is forgotten. (MC)%
\index{context}%
\item%
Input Gate ($i_t$) {-} Controls if/when the context should remember a value. (M+/MS)%
\index{context}%
\index{input}%
\item%
Output Gate ($o_t$) {-} Controls if/when the remembered value is allowed to pass from the unit. (RM)%
\index{output}%
\end{itemize}

\subsection{Simple Keras LSTM Example}%
\label{subsec:SimpleKerasLSTMExample}%
The following code creates the LSTM network, an example of an RNN for classification.  The following code trains on a data set (x) with a max sequence size of 6 (columns) and six training elements (rows)%
\index{classification}%
\index{LSTM}%
\index{training}%
\par%
\begin{tcolorbox}[size=title,title=Code,breakable]%
\begin{lstlisting}[language=Python, upquote=true]
from tensorflow.keras.preprocessing import sequence
from tensorflow.keras.models import Sequential
from tensorflow.keras.layers import Dense, Embedding
from tensorflow.keras.layers import LSTM
import numpy as np

max_features = 4 # 0,1,2,3 (total of 4)
x = [
    [[0],[1],[1],[0],[0],[0]],
    [[0],[0],[0],[2],[2],[0]],
    [[0],[0],[0],[0],[3],[3]],
    [[0],[2],[2],[0],[0],[0]],
    [[0],[0],[3],[3],[0],[0]],
    [[0],[0],[0],[0],[1],[1]]
]
x = np.array(x,dtype=np.float32)
y = np.array([1,2,3,2,3,1],dtype=np.int32)

# Convert y2 to dummy variables
y2 = np.zeros((y.shape[0], max_features),dtype=np.float32)
y2[np.arange(y.shape[0]), y] = 1.0
print(y2)

print('Build model...')
model = Sequential()
model.add(LSTM(128, dropout=0.2, recurrent_dropout=0.2, \
               input_shape=(None, 1)))
model.add(Dense(4, activation='sigmoid'))

# try using different optimizers and different optimizer configs
model.compile(loss='binary_crossentropy',
              optimizer='adam',
              metrics=['accuracy'])

print('Train...')
model.fit(x,y2,epochs=200)
pred = model.predict(x)
predict_classes = np.argmax(pred,axis=1)
print("Predicted classes: {}",predict_classes)
print("Expected classes: {}",predict_classes)\end{lstlisting}
\tcbsubtitle[before skip=\baselineskip]{Output}%
\begin{lstlisting}[upquote=true]
[[0. 1. 0. 0.]
 [0. 0. 1. 0.]
 [0. 0. 0. 1.]
 [0. 0. 1. 0.]
 [0. 0. 0. 1.]
 [0. 1. 0. 0.]]
Build model...
Train...
...
1/1 [==============================] - 0s 66ms/step - loss: 0.2622 -
accuracy: 0.6667
Epoch 200/200
1/1 [==============================] - 0s 39ms/step - loss: 0.2329 -
accuracy: 0.6667
Predicted classes: {} [1 2 3 2 3 1]
Expected classes: {} [1 2 3 2 3 1]
\end{lstlisting}
\end{tcolorbox}%
We can now present a sequence directly to the model for classification.%
\index{classification}%
\index{model}%
\par%
\begin{tcolorbox}[size=title,title=Code,breakable]%
\begin{lstlisting}[language=Python, upquote=true]
def runit(model, inp):
    inp = np.array(inp,dtype=np.float32)
    pred = model.predict(inp)
    return np.argmax(pred[0])

print( runit( model, [[[0],[0],[0],[0],[0],[1]]] ))\end{lstlisting}
\tcbsubtitle[before skip=\baselineskip]{Output}%
\begin{lstlisting}[upquote=true]
1
\end{lstlisting}
\end{tcolorbox}

\subsection{Sun Spots Example}%
\label{subsec:SunSpotsExample}%
This section shows an example of RNN regression to predict sunspots.  You can find the data files needed for this example at the following location.%
\index{predict}%
\index{regression}%
\index{sunspots}%
\par%
\begin{itemize}[noitemsep]%
\item%
\href{http://www.sidc.be/silso/datafiles#total}{Sunspot Data Files}%
\item%
\href{http://www.sidc.be/silso/INFO/sndtotcsv.php}{Download Daily Sunspots }%
{-} 1/1/1818 to now.%
\end{itemize}%
The following code loads the sunspot file:%
\par%
\begin{tcolorbox}[size=title,title=Code,breakable]%
\begin{lstlisting}[language=Python, upquote=true]
import pandas as pd
import os
  
names = ['year', 'month', 'day', 'dec_year', 'sn_value' , 
         'sn_error', 'obs_num', 'unused1']
df = pd.read_csv(
    "https://data.heatonresearch.com/data/t81-558/SN_d_tot_V2.0.csv",
    sep=';',header=None,names=names,
    na_values=['-1'], index_col=False)

print("Starting file:")
print(df[0:10])

print("Ending file:")
print(df[-10:])\end{lstlisting}
\tcbsubtitle[before skip=\baselineskip]{Output}%
\begin{lstlisting}[upquote=true]
Starting file:
   year  month  day  dec_year  sn_value  sn_error  obs_num  unused1
0  1818      1    1  1818.001        -1       NaN        0        1
1  1818      1    2  1818.004        -1       NaN        0        1
2  1818      1    3  1818.007        -1       NaN        0        1
3  1818      1    4  1818.010        -1       NaN        0        1
4  1818      1    5  1818.012        -1       NaN        0        1
5  1818      1    6  1818.015        -1       NaN        0        1
6  1818      1    7  1818.018        -1       NaN        0        1
7  1818      1    8  1818.021        65      10.2        1        1
8  1818      1    9  1818.023        -1       NaN        0        1
9  1818      1   10  1818.026        -1       NaN        0        1
Ending file:
       year  month  day  dec_year  sn_value  sn_error  obs_num
unused1

...

0
72863  2017      6   29  2017.492        12       0.5       25
0
72864  2017      6   30  2017.495        11       0.5       30
0
\end{lstlisting}
\end{tcolorbox}%
As you can see, there is quite a bit of missing data near the end of the file.  We want to find the starting index where the missing data no longer occurs.  This technique is somewhat sloppy; it would be better to find a use for the data between missing values.  However, the point of this example is to show how to use LSTM with a somewhat simple time{-}series.%
\index{LSTM}%
\index{SOM}%
\index{time{-}series}%
\par%
\begin{tcolorbox}[size=title,title=Code,breakable]%
\begin{lstlisting}[language=Python, upquote=true]
start_id = max(df[df['obs_num'] == 0].index.tolist())+1  # Find the last zero and move one beyond
print(start_id)
df = df[start_id:] # Trim the rows that have missing observations\end{lstlisting}
\tcbsubtitle[before skip=\baselineskip]{Output}%
\begin{lstlisting}[upquote=true]
11314
\end{lstlisting}
\end{tcolorbox}%
\begin{tcolorbox}[size=title,title=Code,breakable]%
\begin{lstlisting}[language=Python, upquote=true]
df['sn_value'] = df['sn_value'].astype(float)
df_train = df[df['year']<2000]
df_test = df[df['year']>=2000]

spots_train = df_train['sn_value'].tolist()
spots_test = df_test['sn_value'].tolist()

print("Training set has {} observations.".format(len(spots_train)))
print("Test set has {} observations.".format(len(spots_test)))\end{lstlisting}
\tcbsubtitle[before skip=\baselineskip]{Output}%
\begin{lstlisting}[upquote=true]
Training set has 55160 observations.
Test set has 6391 observations.
\end{lstlisting}
\end{tcolorbox}%
To create an algorithm that will predict future values, we need to consider how to encode this data to be presented to the algorithm. The data must be submitted as sequences, using a sliding window algorithm to encode the data. We must define how large the window will be. Consider an n{-}sized window. Each sequence's $x$ values will be a $n$ data points sequence. The $y$'s will be the next value, after the sequence, that we are trying to predict. You can use the following function to take a series of values, such as sunspots, and generate sequences ($x$) and predicted values ($y$).%
\index{algorithm}%
\index{predict}%
\index{sunspots}%
\par%
\begin{tcolorbox}[size=title,title=Code,breakable]%
\begin{lstlisting}[language=Python, upquote=true]
import numpy as np

def to_sequences(seq_size, obs):
    x = []
    y = []

    for i in range(len(obs)-SEQUENCE_SIZE):
        #print(i)
        window = obs[i:(i+SEQUENCE_SIZE)]
        after_window = obs[i+SEQUENCE_SIZE]
        window = [[x] for x in window]
        #print("{} - {}".format(window,after_window))
        x.append(window)
        y.append(after_window)
        
    return np.array(x),np.array(y)
    
    
SEQUENCE_SIZE = 10
x_train,y_train = to_sequences(SEQUENCE_SIZE,spots_train)
x_test,y_test = to_sequences(SEQUENCE_SIZE,spots_test)

print("Shape of training set: {}".format(x_train.shape))
print("Shape of test set: {}".format(x_test.shape))\end{lstlisting}
\tcbsubtitle[before skip=\baselineskip]{Output}%
\begin{lstlisting}[upquote=true]
Shape of training set: (55150, 10, 1)
Shape of test set: (6381, 10, 1)
\end{lstlisting}
\end{tcolorbox}%
We can see the internal structure of the training data. The first dimension is the number of training elements, the second indicates a sequence size of 10, and finally, we have one data point per timeslice in the window.%
\index{training}%
\par%
\begin{tcolorbox}[size=title,title=Code,breakable]%
\begin{lstlisting}[language=Python, upquote=true]
x_train.shape\end{lstlisting}
\tcbsubtitle[before skip=\baselineskip]{Output}%
\begin{lstlisting}[upquote=true]
(55150, 10, 1)
\end{lstlisting}
\end{tcolorbox}%
We are now ready to build and train the model.%
\index{model}%
\par%
\begin{tcolorbox}[size=title,title=Code,breakable]%
\begin{lstlisting}[language=Python, upquote=true]
from tensorflow.keras.preprocessing import sequence
from tensorflow.keras.models import Sequential
from tensorflow.keras.layers import Dense, Embedding
from tensorflow.keras.layers import LSTM
from tensorflow.keras.datasets import imdb
from tensorflow.keras.callbacks import EarlyStopping
import numpy as np

print('Build model...')
model = Sequential()
model.add(LSTM(64, dropout=0.0, recurrent_dropout=0.0,\
                   input_shape=(None, 1)))
model.add(Dense(32))
model.add(Dense(1))
model.compile(loss='mean_squared_error', optimizer='adam')
monitor = EarlyStopping(monitor='val_loss', min_delta=1e-3, patience=5, 
                        verbose=1, mode='auto', restore_best_weights=True)
print('Train...')

model.fit(x_train,y_train,validation_data=(x_test,y_test),
          callbacks=[monitor],verbose=2,epochs=1000)\end{lstlisting}
\tcbsubtitle[before skip=\baselineskip]{Output}%
\begin{lstlisting}[upquote=true]
Build model...
Train...
...
1724/1724 - 10s - loss: 497.0393 - val_loss: 215.1721 - 10s/epoch -
6ms/step
Epoch 11/1000
Restoring model weights from the end of the best epoch: 6.
1724/1724 - 10s - loss: 495.1920 - val_loss: 220.1826 - 10s/epoch -
6ms/step
Epoch 11: early stopping
\end{lstlisting}
\end{tcolorbox}%
Finally, we evaluate the model with RMSE.%
\index{model}%
\index{MSE}%
\index{RMSE}%
\index{RMSE}%
\par%
\begin{tcolorbox}[size=title,title=Code,breakable]%
\begin{lstlisting}[language=Python, upquote=true]
from sklearn import metrics

pred = model.predict(x_test)
score = np.sqrt(metrics.mean_squared_error(pred,y_test))
print("Score (RMSE): {}".format(score))\end{lstlisting}
\end{tcolorbox}

\section{Part 10.3: Text Generation with LSTM}%
\label{sec:Part10.3TextGenerationwithLSTM}%
Recurrent neural networks are also known for their ability to generate text. As a result, the neural network's output can be free{-}form text. This section will demonstrate how to train an LSTM on a textual document, such as classic literature, and learn to output new text that appears to be of the same form as the training material. If you train your LSTM on%
\index{LSTM}%
\index{neural network}%
\index{output}%
\index{recurrent}%
\index{training}%
\href{https://en.wikipedia.org/wiki/William_Shakespeare}{ Shakespeare}%
, it will learn to crank out new prose similar to what Shakespeare had written.%
\par%
Don't get your hopes up. You will not teach your deep neural network to write the next%
\index{neural network}%
\href{https://en.wikipedia.org/wiki/Pulitzer_Prize_for_Fiction}{ Pulitzer Prize for Fiction}%
. The prose generated by your neural network will be nonsensical. However, the output text will usually be nearly grammatically correct and similar to the source training documents.%
\index{neural network}%
\index{output}%
\index{training}%
\par%
A neural network generating nonsensical text based on literature may not seem helpful. However, this technology gets so much interest because it forms the foundation for many more advanced technologies. The LSTM will typically learn human grammar from the source document opens a wide range of possibilities. You can use similar technology to complete sentences when entering text. The ability to output free{-}form text has become the foundation of many other technologies. In the next part, we will use this technique to create a neural network that can write captions for images to describe what is going on in the picture.%
\index{LSTM}%
\index{neural network}%
\index{output}%
\par%
\subsection{Additional Information}%
\label{subsec:AdditionalInformation}%
The following are some articles that I found helpful in putting this section together.%
\index{SOM}%
\par%
\begin{itemize}[noitemsep]%
\item%
\href{http://karpathy.github.io/2015/05/21/rnn-effectiveness/}{The Unreasonable Effectiveness of Recurrent Neural Networks}%
\item%
\href{https://keras.io/examples/lstm_text_generation/}{Keras LSTM Generation Example}%
\end{itemize}

\subsection{Character{-}Level Text Generation}%
\label{subsec:Character{-}LevelTextGeneration}%
There are several different approaches to teaching a neural network to output free{-}form text. The most basic question is if you wish the neural network to learn at the word or character level. Learning at the character level is the more interesting of the two. The LSTM is learning to construct its own words without even being shown what a word is. We will begin with character{-}level text generation. In the next module, we will see how we can use nearly the same technique to operate at the word level. We will implement word{-}level automatic captioning in the next module.%
\index{learning}%
\index{LSTM}%
\index{neural network}%
\index{output}%
\index{text generation}%
\par%
We import the needed Python packages and define the sequence length, named%
\index{Python}%
\textbf{ maxlen}%
. Time{-}series neural networks always accept their input as a fixed{-}length array. Because you might not use all of the sequence elements, filling extra pieces with zeros is common. You will divide the text into sequences of this length, and the neural network will train to predict what comes after this sequence.%
\index{input}%
\index{neural network}%
\index{predict}%
\index{time{-}series}%
\par%
\begin{tcolorbox}[size=title,title=Code,breakable]%
\begin{lstlisting}[language=Python, upquote=true]
from tensorflow.keras.callbacks import LambdaCallback
from tensorflow.keras.models import Sequential
from tensorflow.keras.layers import Dense
from tensorflow.keras.layers import LSTM
from tensorflow.keras.optimizers import RMSprop
from tensorflow.keras.utils import get_file
import numpy as np
import random
import sys
import io
import requests
import re\end{lstlisting}
\end{tcolorbox}%
We will train the neural network on the classic children's book%
\index{neural network}%
\href{https://en.wikipedia.org/wiki/Treasure_Island}{ Treasure Island}%
.  We begin by loading this text into a Python string and displaying the first 1,000 characters.%
\index{Python}%
\par%
\begin{tcolorbox}[size=title,title=Code,breakable]%
\begin{lstlisting}[language=Python, upquote=true]
r = requests.get("https://data.heatonresearch.com/data/t81-558/text/"\
                 "treasure_island.txt")
raw_text = r.text
print(raw_text[0:1000])\end{lstlisting}
\tcbsubtitle[before skip=\baselineskip]{Output}%
\begin{lstlisting}[upquote=true]
The Project Gutenberg EBook of Treasure Island, by Robert Louis
Stevenson
This eBook is for the use of anyone anywhere at no cost and with
almost no restrictions whatsoever.  You may copy it, give it away or
re-use it under the terms of the Project Gutenberg License included
with this eBook or online at www.gutenberg.net
Title: Treasure Island
Author: Robert Louis Stevenson
Illustrator: Milo Winter
Release Date: January 12, 2009 [EBook #27780]
Language: English
*** START OF THIS PROJECT GUTENBERG EBOOK TREASURE ISLAND ***
Produced by Juliet Sutherland, Stephen Blundell and the
Online Distributed Proofreading Team at http://www.pgdp.net
 THE ILLUSTRATED CHILDREN'S LIBRARY

...

            Milo Winter
           [Illustration]
           GRAMERCY BOOKS
              NEW YORK
 Foreword copyright  1986 by Random House V
\end{lstlisting}
\end{tcolorbox}%
We will extract all unique characters from the text and sort them.  This technique allows us to assign a unique ID to each character.  Because we sorted the characters, these IDs should remain the same.  The IDs will change if we add new characters to the original text.  We build two dictionaries.  The first%
\textbf{ char2idx }%
is used to convert a character into its ID.  The second%
\textbf{ idx2char }%
converts an ID back into its character.%
\par%
\begin{tcolorbox}[size=title,title=Code,breakable]%
\begin{lstlisting}[language=Python, upquote=true]
processed_text = raw_text.lower()
processed_text = re.sub(r'[^\x00-\x7f]',r'', processed_text) 

print('corpus length:', len(processed_text))

chars = sorted(list(set(processed_text)))
print('total chars:', len(chars))
char_indices = dict((c, i) for i, c in enumerate(chars))
indices_char = dict((i, c) for i, c in enumerate(chars))\end{lstlisting}
\tcbsubtitle[before skip=\baselineskip]{Output}%
\begin{lstlisting}[upquote=true]
corpus length: 397400
total chars: 60
\end{lstlisting}
\end{tcolorbox}%
We are now ready to build the actual sequences.  Like previous neural networks, there will be an $x$ and $y$.  However, for the LSTM, $x$ and $y$ will be sequences.  The $x$ input will specify the sequences where $y$ is the expected output.  The following code generates all possible sequences.%
\index{input}%
\index{LSTM}%
\index{neural network}%
\index{output}%
\par%
\begin{tcolorbox}[size=title,title=Code,breakable]%
\begin{lstlisting}[language=Python, upquote=true]
# cut the text in semi-redundant sequences of maxlen characters
maxlen = 40
step = 3
sentences = []
next_chars = []
for i in range(0, len(processed_text) - maxlen, step):
    sentences.append(processed_text[i: i + maxlen])
    next_chars.append(processed_text[i + maxlen])
print('nb sequences:', len(sentences))\end{lstlisting}
\tcbsubtitle[before skip=\baselineskip]{Output}%
\begin{lstlisting}[upquote=true]
nb sequences: 132454
\end{lstlisting}
\end{tcolorbox}%
\begin{tcolorbox}[size=title,title=Code,breakable]%
\begin{lstlisting}[language=Python, upquote=true]
sentences\end{lstlisting}
\tcbsubtitle[before skip=\baselineskip]{Output}%
\begin{lstlisting}[upquote=true]
['the project gutenberg ebook of treasure ',
 ' project gutenberg ebook of treasure isl',
 'oject gutenberg ebook of treasure island',
 'ct gutenberg ebook of treasure island, b',
 'gutenberg ebook of treasure island, by r',
 'enberg ebook of treasure island, by robe',
 'erg ebook of treasure island, by robert ',
 ' ebook of treasure island, by robert lou',
 'ook of treasure island, by robert louis ',
 ' of treasure island, by robert louis ste',
 ' treasure island, by robert louis steven',
 'easure island, by robert louis stevenson',
 'ure island, by robert louis stevenson\r\n\r',
 ' island, by robert louis stevenson\r\n\r\nth',
 'land, by robert louis stevenson\r\n\r\nthis ',

...

 'st of color plates_                     ',
 'of color plates_                        ',
 'color plates_                           ',
 'or plates_                              ',
 ...]
\end{lstlisting}
\end{tcolorbox}%
We can now convert the text into vectors.%
\index{vector}%
\par%
\begin{tcolorbox}[size=title,title=Code,breakable]%
\begin{lstlisting}[language=Python, upquote=true]
print('Vectorization...')
x = np.zeros((len(sentences), maxlen, len(chars)), dtype=np.bool)
y = np.zeros((len(sentences), len(chars)), dtype=np.bool)
for i, sentence in enumerate(sentences):
    for t, char in enumerate(sentence):
        x[i, t, char_indices[char]] = 1
    y[i, char_indices[next_chars[i]]] = 1\end{lstlisting}
\tcbsubtitle[before skip=\baselineskip]{Output}%
\begin{lstlisting}[upquote=true]
Vectorization...
\end{lstlisting}
\end{tcolorbox}%
Next, we create the neural network.  This neural network's primary feature is the LSTM layer, which allows the sequences to be processed.%
\index{feature}%
\index{layer}%
\index{LSTM}%
\index{neural network}%
\index{ROC}%
\index{ROC}%
\par%
\begin{tcolorbox}[size=title,title=Code,breakable]%
\begin{lstlisting}[language=Python, upquote=true]
# build the model: a single LSTM
print('Build model...')
model = Sequential()
model.add(LSTM(128, input_shape=(maxlen, len(chars))))
model.add(Dense(len(chars), activation='softmax'))

optimizer = RMSprop(lr=0.01)
model.compile(loss='categorical_crossentropy', optimizer=optimizer)\end{lstlisting}
\tcbsubtitle[before skip=\baselineskip]{Output}%
\begin{lstlisting}[upquote=true]
Build model...
\end{lstlisting}
\end{tcolorbox}%
\begin{tcolorbox}[size=title,title=Code,breakable]%
\begin{lstlisting}[language=Python, upquote=true]
model.summary()\end{lstlisting}
\tcbsubtitle[before skip=\baselineskip]{Output}%
\begin{lstlisting}[upquote=true]
Model: "sequential"
_________________________________________________________________
 Layer (type)                Output Shape              Param #
=================================================================
 lstm (LSTM)                 (None, 128)               96768
 dense (Dense)               (None, 60)                7740
=================================================================
Total params: 104,508
Trainable params: 104,508
Non-trainable params: 0
_________________________________________________________________
\end{lstlisting}
\end{tcolorbox}%
The LSTM will produce new text character by character.  We will need to sample the correct letter from the LSTM predictions each time.  The%
\index{LSTM}%
\index{predict}%
\textbf{ sample }%
function accepts the following two parameters:%
\index{parameter}%
\par%
\begin{itemize}[noitemsep]%
\item%
\textbf{preds }%
{-} The output neurons.%
\index{neuron}%
\index{output}%
\index{output neuron}%
\item%
\textbf{temperature }%
{-} 1.0 is the most conservative, 0.0 is the most confident (willing to make spelling and other errors).%
\index{error}%
\end{itemize}%
The sample function below essentially performs a softmax on the neural network predictions.  This process causes each output neuron to become a probability of its particular letter.%
\index{neural network}%
\index{neuron}%
\index{output}%
\index{output neuron}%
\index{predict}%
\index{probability}%
\index{ROC}%
\index{ROC}%
\index{softmax}%
\par%
\begin{tcolorbox}[size=title,title=Code,breakable]%
\begin{lstlisting}[language=Python, upquote=true]
def sample(preds, temperature=1.0):
    # helper function to sample an index from a probability array
    preds = np.asarray(preds).astype('float64')
    preds = np.log(preds) / temperature
    exp_preds = np.exp(preds)
    preds = exp_preds / np.sum(exp_preds)
    probas = np.random.multinomial(1, preds, 1)
    return np.argmax(probas)\end{lstlisting}
\end{tcolorbox}%
Keras calls the following function at the end of each training Epoch.  The code generates sample text generations that visually demonstrate the neural network better at text generation.  As the neural network trains, the generations should look more realistic.%
\index{Keras}%
\index{neural network}%
\index{text generation}%
\index{training}%
\par%
\begin{tcolorbox}[size=title,title=Code,breakable]%
\begin{lstlisting}[language=Python, upquote=true]
def on_epoch_end(epoch, _):
    # Function invoked at end of each epoch. Prints generated text.
    print("******************************************************")
    print('----- Generating text after Epoch: %d' % epoch)

    start_index = random.randint(0, len(processed_text) - maxlen - 1)
    for temperature in [0.2, 0.5, 1.0, 1.2]:
        print('----- temperature:', temperature)

        generated = ''
        sentence = processed_text[start_index: start_index + maxlen]
        generated += sentence
        print('----- Generating with seed: "' + sentence + '"')
        sys.stdout.write(generated)

        for i in range(400):
            x_pred = np.zeros((1, maxlen, len(chars)))
            for t, char in enumerate(sentence):
                x_pred[0, t, char_indices[char]] = 1.

            preds = model.predict(x_pred, verbose=0)[0]
            next_index = sample(preds, temperature)
            next_char = indices_char[next_index]

            generated += next_char
            sentence = sentence[1:] + next_char

            sys.stdout.write(next_char)
            sys.stdout.flush()
        print()\end{lstlisting}
\end{tcolorbox}%
We are now ready to train.  Depending on how fast your computer is, it can take up to an hour to train this network.  If you have a GPU available, please make sure to use it.%
\index{GPU}%
\index{GPU}%
\par%
\begin{tcolorbox}[size=title,title=Code,breakable]%
\begin{lstlisting}[language=Python, upquote=true]
# Ignore useless W0819 warnings generated by TensorFlow 2.0.  Hopefully can remove this ignore in the future.
# See https://github.com/tensorflow/tensorflow/issues/31308
import logging, os
logging.disable(logging.WARNING)
os.environ["TF_CPP_MIN_LOG_LEVEL"] = "3"

# Fit the model
print_callback = LambdaCallback(on_epoch_end=on_epoch_end)

model.fit(x, y,
          batch_size=128,
          epochs=60,
          callbacks=[print_callback])\end{lstlisting}
\tcbsubtitle[before skip=\baselineskip]{Output}%
\begin{lstlisting}[upquote=true]
...
1035/1035 [==============================] - 74s 71ms/step - loss:
1.1361
Epoch 60/60
1029/1035 [============================>.] - ETA: 0s - loss:
1.1339******************************************************
----- Generating text after Epoch: 59
----- temperature: 0.2
----- Generating with seed: "
tail of it on his unruly followers. th"
"it's a don't be men belief in my till be captain silver had been the
blows and the stockade, and a man of the boats was place and the
captain. he was a pairs and seemed to the barrows and the part, and i
saw he was a state before the pirates of his s
----- temperature: 0.5

...

----- Generating with seed: "
tail of it on his unruly followers. th"
levor, as i could now shee ehe me so come kint and so so
1035/1035 [==============================] - 74s 72ms/step - loss:
1.1339
\end{lstlisting}
\end{tcolorbox}

\section{Part 10.4: Introduction to Transformers}%
\label{sec:Part10.4IntroductiontoTransformers}%
Transformers are neural networks that provide state{-}of{-}the{-}art solutions for many of the problems previously assigned to recurrent neural networks.%
\index{neural network}%
\index{recurrent}%
\index{transformer}%
\cite{vaswani2017attention}%
Sequences can form both the input and the output of a neural network, examples of such configurations include::%
\index{input}%
\index{neural network}%
\index{output}%
\par%
\begin{itemize}[noitemsep]%
\item%
Vector to Sequence {-} Image captioning%
\index{vector}%
\item%
Sequence to Vector {-} Sentiment analysis%
\index{sentiment analysis}%
\index{vector}%
\item%
Sequence to Sequence {-} Language translation%
\index{language translation}%
\end{itemize}%
Sequence{-}to{-}sequence allows an input sequence to produce an output sequence based on an input sequence. Transformers focus primarily on this sequence{-}to{-}sequence configuration.%
\index{input}%
\index{output}%
\index{transformer}%
\par%
\subsection{High{-}Level Overview of Transformers}%
\label{subsec:High{-}LevelOverviewofTransformers}%
This course focuses primarily on the application of deep neural networks. The focus will be on presenting data to a transformer and a transformer's major components. As a result, we will not focus on implementing a transformer at the lowest level. The following section provides an overview of critical internal parts of a transformer, such as residual connections and attention. In the next chapter, we will use transformers from%
\index{connection}%
\index{neural network}%
\index{transformer}%
\href{https://huggingface.co/}{ Hugging Face }%
to perform natural language processing with transformers. If you are interested in implementing a transformer from scratch, Keras provides a comprehensive%
\index{Keras}%
\index{ROC}%
\index{ROC}%
\index{transformer}%
\href{https://www.tensorflow.org/text/tutorials/transformer}{ example}%
.%
\par
\begin{figure}[h]%
\centering%
\includegraphics[width=4in]{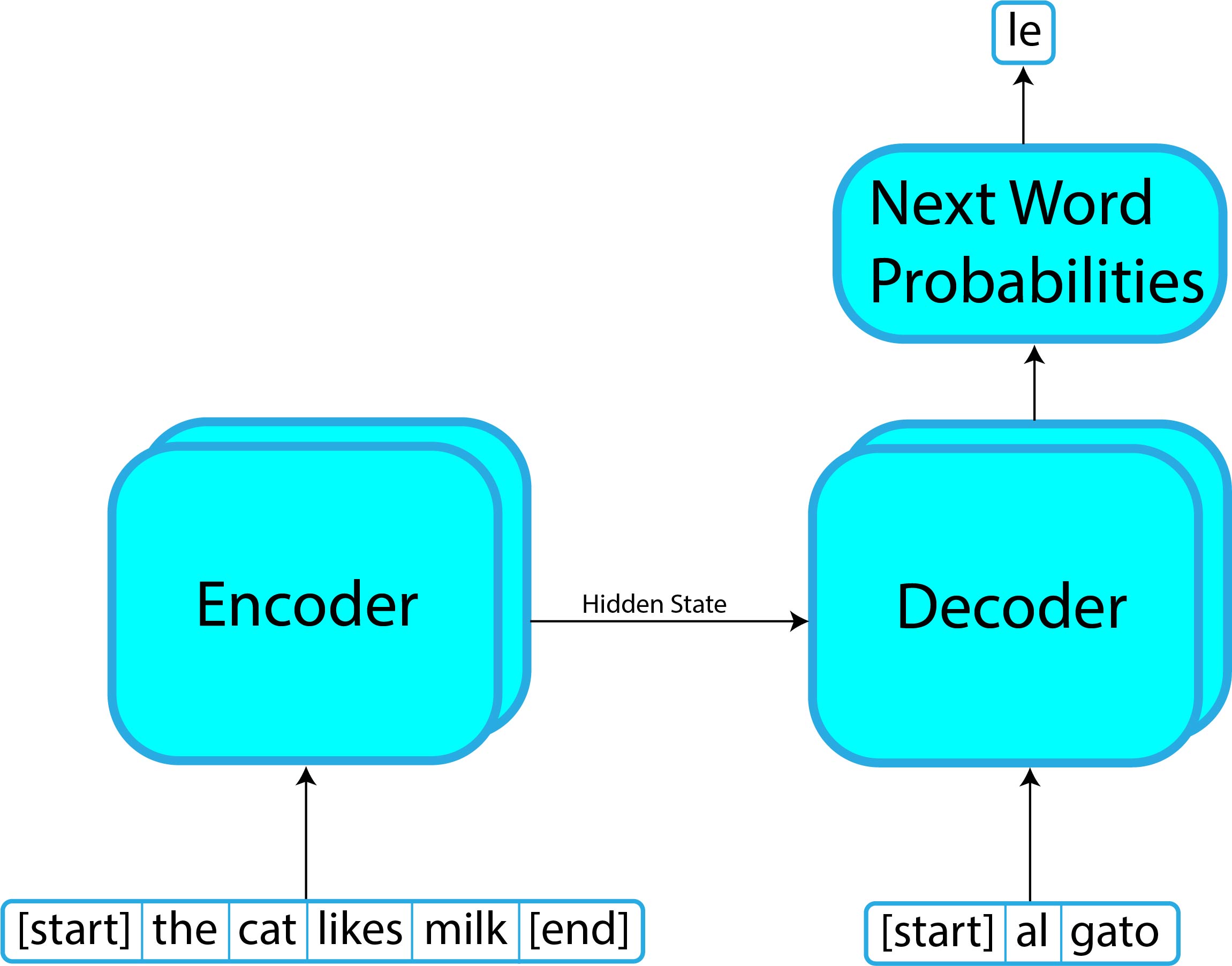}%
\caption{High Level View of a Translation Transformer}%
\label{10.TRANS-1}%
\end{figure}

%
Figure \ref{10.TRANS-1} presents a high{-}level view of a transformer for language translation.%
\index{language translation}%
\index{transformer}%
\par%

\par%
We use a transformer that translates between English and Spanish for this example. We present the English sentence "the cat likes milk" and receive a Spanish translation of "al gato le gusta la leche."%
\index{transformer}%
\par%
We begin by placing the English source sentence between the beginning and ending tokens. This input can be of any length, and we presented it to the neural network as a ragged Tensor. Because the Tensor is ragged, no padding is necessary. Such input is acceptable for the attention layer that will receive the source sentence. The encoder transforms this ragged input into a hidden state containing a series of key{-}value pairs representing the knowledge in the source sentence. The encoder understands to read English and convert to a hidden state. The decoder understands how to output Spanish from this hidden state.%
\index{input}%
\index{layer}%
\index{neural network}%
\index{output}%
\par%
We initially present the decoder with the hidden state and the starting token. The decoder will predict the probabilities of all words in its vocabulary. The word with the highest probability is the first word of the sentence.%
\index{predict}%
\index{probability}%
\par%
The highest probability word is attached concatenated to the translated sentence, initially containing only the beginning token. This process continues, growing the translated sentence in each iteration until the decoder predicts the ending token.%
\index{iteration}%
\index{predict}%
\index{probability}%
\index{ROC}%
\index{ROC}%
\par

\subsection{Transformer Hyperparameters}%
\label{subsec:TransformerHyperparameters}%
Before we describe how these layers fit together, we must consider the following transformer hyperparameters, along with default settings from the Keras transformer example:%
\index{hyperparameter}%
\index{Keras}%
\index{layer}%
\index{parameter}%
\index{transformer}%
\par%
\begin{itemize}[noitemsep]%
\item%
num\_layers = 4%
\index{layer}%
\item%
d\_model = 128%
\index{model}%
\item%
dff = 512%
\item%
num\_heads = 8%
\item%
dropout\_rate = 0.1%
\index{dropout}%
\end{itemize}%
Multiple encoder and decoder layers can be present. The%
\index{layer}%
\textbf{ num\_layers }%
hyperparameter specifies how many encoder and decoder layers there are. The expected tensor shape for the input to the encoder layer is the same as the output produced; as a result, you can easily stack these layers.%
\index{hyperparameter}%
\index{input}%
\index{layer}%
\index{output}%
\index{parameter}%
\par%
We will see embedding layers in the next chapter. However, you can think of an embedding layer as a dictionary for now. Each entry in the embedding corresponds to each word in a fixed{-}size vocabulary. Similar words should have similar vectors. The%
\index{layer}%
\index{vector}%
\textbf{ d\_model }%
hyperparameter specifies the size of the embedding vector. Though you will sometimes preload embeddings from a project such as%
\index{hyperparameter}%
\index{parameter}%
\index{SOM}%
\index{vector}%
\href{https://radimrehurek.com/gensim/models/word2vec.html}{ Word2vec }%
or%
\href{https://nlp.stanford.edu/projects/glove/}{ GloVe}%
, the optimizer can train these embeddings with the rest of the transformer. Training your embeddings allows the%
\index{training}%
\index{transformer}%
\textbf{ d\_model }%
hyperparameter to set to any desired value. If you transfer the embeddings, you must set the%
\index{hyperparameter}%
\index{parameter}%
\textbf{ d\_model }%
hyperparameter to the same value as the transferred embeddings.%
\index{hyperparameter}%
\index{parameter}%
\par%
The%
\textbf{ dff }%
hyperparameter specifies the size of the dense feedforward layers. The%
\index{feedforward}%
\index{hyperparameter}%
\index{layer}%
\index{parameter}%
\textbf{ num\_heads }%
hyperparameter sets the number of attention layers heads. Finally, the dropout\_rate specifies a dropout percentage to combat overfitting. We discussed dropout previously in this book.%
\index{dropout}%
\index{hyperparameter}%
\index{layer}%
\index{overfitting}%
\index{parameter}%
\par

\subsection{Inside a Transformer}%
\label{subsec:InsideaTransformer}%
In this section, we will examine the internals of a transformer so that you become familiar with essential concepts such as:%
\index{transformer}%
\par
\begin{figure}[h]%
\centering%
\includegraphics[width=4in]{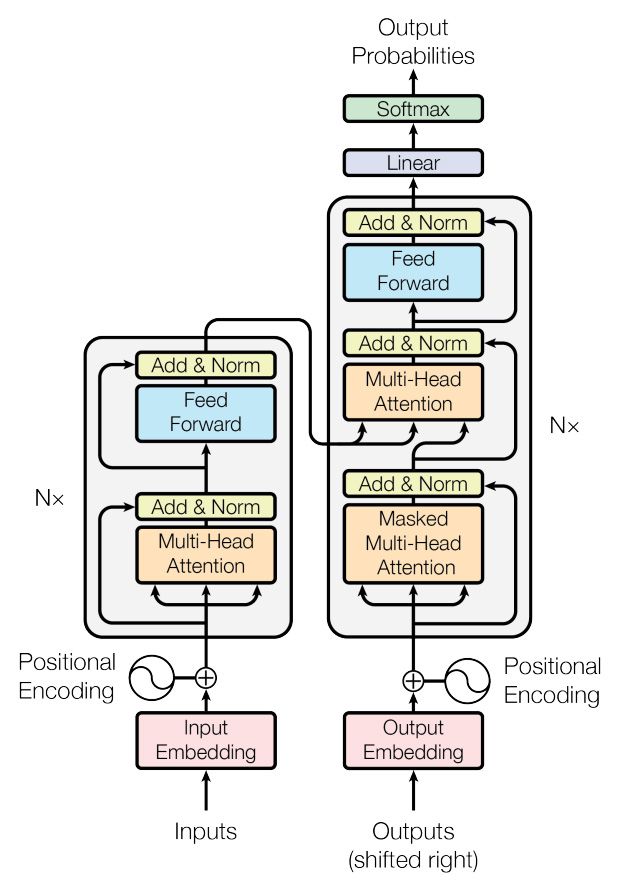}%
\caption{Architectural Diagram from the Paper}%
\label{10.TRANS-2}%
\end{figure}

%
\begin{itemize}[noitemsep]%
\item%
Embeddings%
\item%
Positional Encoding%
\item%
Attention and Self{-}Attention%
\item%
Residual Connection%
\index{connection}%
\end{itemize}%
You can see a lower{-}level diagram of a transformer in Figure \ref{10.TRANS-2}.%
\index{transformer}%
\par%

\par%
While the original transformer paper is titled "Attention is All you Need," attention isn't the only layer type you need. The transformer also contains dense layers. However, the title "Attention and Dense Layers are All You Need" isn't as catchy.%
\index{dense layer}%
\index{layer}%
\index{transformer}%
\par%
The transformer begins by tokenizing the input English sentence. Tokens may or may not be words. Generally, familiar parts of words are tokenized and become building blocks of longer words. This tokenization allows common suffixes and prefixes to be understood independently of their stem word. Each token becomes a numeric index that the transformer uses to look up the vector. There are several special tokens:%
\index{input}%
\index{tokenization}%
\index{transformer}%
\index{vector}%
\par%
\begin{itemize}[noitemsep]%
\item%
Index 0 = Pad%
\item%
Index 1 = Unknow%
\item%
Index 2 = Start token%
\item%
Index 3 = End token%
\end{itemize}%
The transformer uses index 0 when we must pad unused space at the end of a tensor. Index 1 is for unknown words. The starting and ending tokens are provided by indexes 2 and 3.%
\index{transformer}%
\par%
The token vectors are simply the inputs to the attention layers; there is no implied order or position. The transformer adds the slopes of a sine and cosine wave to the token vectors to encode position.%
\index{input}%
\index{layer}%
\index{slope}%
\index{transformer}%
\index{vector}%
\par%
Attention layers have three inputs: key (k), value(v), and query (q). This layer is self{-}attention if the query, key, and value are the same. The key and value pairs specify the information that the query operates upon. The attention layer learns what positions of data to focus upon.%
\index{input}%
\index{layer}%
\par%
The transformer presents the position encoded embedding vectors to the first self{-}attention segment in the encoder layer. The output from the attention is normalized and ultimately becomes the hidden state after all encoder layers are processed.%
\index{layer}%
\index{output}%
\index{ROC}%
\index{ROC}%
\index{transformer}%
\index{vector}%
\par%
The hidden state is only calculated once per query. Once the input Spanish sentence becomes a hidden state, this value is presented repeatedly to the decoder until the decoder forms the final Spanish sentence.%
\index{calculated}%
\index{input}%
\par%
This section presented a high{-}level introduction to transformers. In the next part, we will implement the encoder and apply it to time series. In the following chapter, we will use%
\index{transformer}%
\href{https://huggingface.co/}{ Hugging Face }%
transformers to perform natural language processing.%
\index{ROC}%
\index{ROC}%
\index{transformer}%
\par

\section{Part 10.5: Programming Transformers with Keras}%
\label{sec:Part10.5ProgrammingTransformerswithKeras}%
This section shows an example of a transformer encoder to predict sunspots.  You can find the data files needed for this example at the following location.%
\index{predict}%
\index{sunspots}%
\index{transformer}%
\par%
\begin{itemize}[noitemsep]%
\item%
\href{http://www.sidc.be/silso/datafiles#total}{Sunspot Data Files}%
\item%
\href{http://www.sidc.be/silso/INFO/sndtotcsv.php}{Download Daily Sunspots }%
{-} 1/1/1818 to now.%
\end{itemize}%
The following code loads the sunspot file:%
\par%
\begin{tcolorbox}[size=title,title=Code,breakable]%
\begin{lstlisting}[language=Python, upquote=true]
import pandas as pd
import os
  
names = ['year', 'month', 'day', 'dec_year', 'sn_value' , 
         'sn_error', 'obs_num', 'extra']
df = pd.read_csv(
    "https://data.heatonresearch.com/data/t81-558/SN_d_tot_V2.0.csv",
    sep=';',header=None,names=names,
    na_values=['-1'], index_col=False)

print("Starting file:")
print(df[0:10])

print("Ending file:")
print(df[-10:])\end{lstlisting}
\tcbsubtitle[before skip=\baselineskip]{Output}%
\begin{lstlisting}[upquote=true]
Starting file:
   year  month  day  dec_year  sn_value  sn_error  obs_num  extra
0  1818      1    1  1818.001        -1       NaN        0      1
1  1818      1    2  1818.004        -1       NaN        0      1
2  1818      1    3  1818.007        -1       NaN        0      1
3  1818      1    4  1818.010        -1       NaN        0      1
4  1818      1    5  1818.012        -1       NaN        0      1
5  1818      1    6  1818.015        -1       NaN        0      1
6  1818      1    7  1818.018        -1       NaN        0      1
7  1818      1    8  1818.021        65      10.2        1      1
8  1818      1    9  1818.023        -1       NaN        0      1
9  1818      1   10  1818.026        -1       NaN        0      1
Ending file:
       year  month  day  dec_year  sn_value  sn_error  obs_num  extra
72855  2017      6   21  2017.470        35       1.0       41      0

...

72860  2017      6   26  2017.484        21       1.1       25      0
72861  2017      6   27  2017.486        19       1.2       36      0
72862  2017      6   28  2017.489        17       1.1       22      0
72863  2017      6   29  2017.492        12       0.5       25      0
72864  2017      6   30  2017.495        11       0.5       30      0
\end{lstlisting}
\end{tcolorbox}%
As you can see, there is quite a bit of missing data near the end of the file.  We want to find the starting index where the missing data no longer occurs.  This technique is somewhat sloppy; it would be better to find a use for the data between missing values.  However, the point of this example is to show how to use a transformer encoder with a somewhat simple time series.%
\index{SOM}%
\index{transformer}%
\par%
\begin{tcolorbox}[size=title,title=Code,breakable]%
\begin{lstlisting}[language=Python, upquote=true]
# Find the last zero and move one beyond
start_id = max(df[df['obs_num'] == 0].index.tolist())+1  
print(start_id)
df = df[start_id:] # Trim the rows that have missing observations\end{lstlisting}
\tcbsubtitle[before skip=\baselineskip]{Output}%
\begin{lstlisting}[upquote=true]
11314
\end{lstlisting}
\end{tcolorbox}%
Divide into training and test/validation sets.%
\index{training}%
\index{validation}%
\par%
\begin{tcolorbox}[size=title,title=Code,breakable]%
\begin{lstlisting}[language=Python, upquote=true]
df['sn_value'] = df['sn_value'].astype(float)
df_train = df[df['year']<2000]
df_test = df[df['year']>=2000]

spots_train = df_train['sn_value'].tolist()
spots_test = df_test['sn_value'].tolist()

print("Training set has {} observations.".format(len(spots_train)))
print("Test set has {} observations.".format(len(spots_test)))\end{lstlisting}
\tcbsubtitle[before skip=\baselineskip]{Output}%
\begin{lstlisting}[upquote=true]
Training set has 55160 observations.
Test set has 6391 observations.
\end{lstlisting}
\end{tcolorbox}%
The%
\textbf{ to\_sequences }%
function takes linear time series data into an%
\index{linear}%
\textbf{ x }%
and%
\textbf{ y }%
where%
\textbf{ x }%
is all possible sequences of seq\_size. After each%
\textbf{ x }%
sequence, this function places the next value into the%
\textbf{ y }%
variable. These%
\textbf{ x }%
and%
\textbf{ y }%
data can train a time{-}series neural network.%
\index{neural network}%
\index{time{-}series}%
\par%
\begin{tcolorbox}[size=title,title=Code,breakable]%
\begin{lstlisting}[language=Python, upquote=true]
import numpy as np

def to_sequences(seq_size, obs):
    x = []
    y = []

    for i in range(len(obs)-SEQUENCE_SIZE):
        #print(i)
        window = obs[i:(i+SEQUENCE_SIZE)]
        after_window = obs[i+SEQUENCE_SIZE]
        window = [[x] for x in window]
        #print("{} - {}".format(window,after_window))
        x.append(window)
        y.append(after_window)
        
    return np.array(x),np.array(y)
    
    
SEQUENCE_SIZE = 10
x_train,y_train = to_sequences(SEQUENCE_SIZE,spots_train)
x_test,y_test = to_sequences(SEQUENCE_SIZE,spots_test)

print("Shape of training set: {}".format(x_train.shape))
print("Shape of test set: {}".format(x_test.shape))\end{lstlisting}
\tcbsubtitle[before skip=\baselineskip]{Output}%
\begin{lstlisting}[upquote=true]
Shape of training set: (55150, 10, 1)
Shape of test set: (6381, 10, 1)
\end{lstlisting}
\end{tcolorbox}%
We can view the results of the%
\textbf{ to\_sequences }%
encoding of the sunspot data.%
\par%
\begin{tcolorbox}[size=title,title=Code,breakable]%
\begin{lstlisting}[language=Python, upquote=true]
print(x_train.shape)\end{lstlisting}
\tcbsubtitle[before skip=\baselineskip]{Output}%
\begin{lstlisting}[upquote=true]
(55150, 10, 1)
\end{lstlisting}
\end{tcolorbox}%
Next, we create the transformer\_encoder; I obtained this function from a%
\index{transformer}%
\href{https://keras.io/examples/timeseries/timeseries_transformer_classification/}{ Keras example}%
. This layer includes residual connections, layer normalization, and dropout. This resulting layer can be stacked multiple times. We implement the projection layers with the Keras Conv1D.%
\index{connection}%
\index{dropout}%
\index{Keras}%
\index{layer}%
\index{stacked}%
\par%
\begin{tcolorbox}[size=title,title=Code,breakable]%
\begin{lstlisting}[language=Python, upquote=true]
from tensorflow import keras
from tensorflow.keras import layers

def transformer_encoder(inputs, head_size, num_heads, ff_dim, dropout=0):
    # Normalization and Attention
    x = layers.LayerNormalization(epsilon=1e-6)(inputs)
    x = layers.MultiHeadAttention(
        key_dim=head_size, num_heads=num_heads, dropout=dropout
    )(x, x)
    x = layers.Dropout(dropout)(x)
    res = x + inputs

    # Feed Forward Part
    x = layers.LayerNormalization(epsilon=1e-6)(res)
    x = layers.Conv1D(filters=ff_dim, kernel_size=1, activation="relu")(x)
    x = layers.Dropout(dropout)(x)
    x = layers.Conv1D(filters=inputs.shape[-1], kernel_size=1)(x)
    return x + res\end{lstlisting}
\end{tcolorbox}%
The following function is provided to build the model, including the attention layer.%
\index{layer}%
\index{model}%
\par%
\begin{tcolorbox}[size=title,title=Code,breakable]%
\begin{lstlisting}[language=Python, upquote=true]
def build_model(
    input_shape,
    head_size,
    num_heads,
    ff_dim,
    num_transformer_blocks,
    mlp_units,
    dropout=0,
    mlp_dropout=0,
):
    inputs = keras.Input(shape=input_shape)
    x = inputs
    for _ in range(num_transformer_blocks):
        x = transformer_encoder(x, head_size, num_heads, ff_dim, dropout)

    x = layers.GlobalAveragePooling1D(data_format="channels_first")(x)
    for dim in mlp_units:
        x = layers.Dense(dim, activation="relu")(x)
        x = layers.Dropout(mlp_dropout)(x)
    outputs = layers.Dense(1)(x)
    return keras.Model(inputs, outputs)\end{lstlisting}
\end{tcolorbox}%
We are now ready to build and train the model.%
\index{model}%
\par%
\begin{tcolorbox}[size=title,title=Code,breakable]%
\begin{lstlisting}[language=Python, upquote=true]
input_shape = x_train.shape[1:]

model = build_model(
    input_shape,
    head_size=256,
    num_heads=4,
    ff_dim=4,
    num_transformer_blocks=4,
    mlp_units=[128],
    mlp_dropout=0.4,
    dropout=0.25,
)

model.compile(
    loss="mean_squared_error",
    optimizer=keras.optimizers.Adam(learning_rate=1e-4)
)
#model.summary()

callbacks = [keras.callbacks.EarlyStopping(patience=10, \
    restore_best_weights=True)]

model.fit(
    x_train,
    y_train,
    validation_split=0.2,
    epochs=200,
    batch_size=64,
    callbacks=callbacks,
)

model.evaluate(x_test, y_test, verbose=1)\end{lstlisting}
\tcbsubtitle[before skip=\baselineskip]{Output}%
\begin{lstlisting}[upquote=true]
...
690/690 [==============================] - 11s 15ms/step - loss:
679.1320 - val_loss: 289.7046
Epoch 37/200
690/690 [==============================] - 11s 16ms/step - loss:
673.3400 - val_loss: 297.0687
200/200 [==============================] - 1s 5ms/step - loss:
214.5603
214.56031799316406
\end{lstlisting}
\end{tcolorbox}%
Finally, we evaluate the model with RMSE.%
\index{model}%
\index{MSE}%
\index{RMSE}%
\index{RMSE}%
\par%
\begin{tcolorbox}[size=title,title=Code,breakable]%
\begin{lstlisting}[language=Python, upquote=true]
from sklearn import metrics

pred = model.predict(x_test)
score = np.sqrt(metrics.mean_squared_error(pred,y_test))
print("Score (RMSE): {}".format(score))\end{lstlisting}
\tcbsubtitle[before skip=\baselineskip]{Output}%
\begin{lstlisting}[upquote=true]
Score (RMSE): 14.647875946283007
\end{lstlisting}
\end{tcolorbox}

\chapter{Natural Language Processing with Hugging Face}%
\label{chap:NaturalLanguageProcessingwithHuggingFace}%
\section{Part 11.1: Introduction to Hugging Face}%
\label{sec:Part11.1IntroductiontoHuggingFace}%
Transformers have become a mainstay of natural language processing. This module will examine the%
\index{ROC}%
\index{ROC}%
\index{transformer}%
\href{https://huggingface.co/}{ Hugging Face }%
Python library for natural language processing, bringing together pretrained transformers, data sets, tokenizers, and other elements. Through the Hugging Face API, you can quickly begin using sentiment analysis, entity recognition, language translation, summarization, and text generation.%
\index{hugging face}%
\index{language translation}%
\index{Python}%
\index{ROC}%
\index{ROC}%
\index{sentiment analysis}%
\index{summarization}%
\index{text generation}%
\index{transformer}%
\par%
Colab does not install Hugging face by default. Whether installing Hugging Face directly into a local computer or utilizing it through Colab, the following commands will install the library.%
\index{hugging face}%
\par%
\begin{tcolorbox}[size=title,title=Code,breakable]%
\begin{lstlisting}[language=Python, upquote=true]
!pip install transformers
!pip install transformers[sentencepiece]\end{lstlisting}
\end{tcolorbox}%
Now that we have Hugging Face installed, the following sections will demonstrate how to apply Hugging Face to a variety of everyday tasks. After this introduction, the remainder of this module will take a deeper look at several specific NLP tasks applied to Hugging Face.%
\index{hugging face}%
\par%
\subsection{Sentiment Analysis}%
\label{subsec:SentimentAnalysis}%
Sentiment analysis uses natural language processing, text analysis, computational linguistics, and biometrics to identify the tone of written text. Passages of written text can be into simple binary states of positive or negative tone. More advanced sentiment analysis might classify text into additional categories: sadness, joy, love, anger, fear, or surprise.%
\index{ROC}%
\index{ROC}%
\index{sentiment analysis}%
\par%
To demonstrate sentiment analysis, we begin by loading sample text, Shakespeare's%
\index{sentiment analysis}%
\href{https://en.wikipedia.org/wiki/Sonnet_18}{ 18th sonnet}%
, a famous poem.%
\par%
\begin{tcolorbox}[size=title,title=Code,breakable]%
\begin{lstlisting}[language=Python, upquote=true]
from urllib.request import urlopen

# Read sample text, a poem
URL = "https://data.heatonresearch.com/data/t81-558/"\
    "datasets/sonnet_18.txt"
f = urlopen(URL)
text = f.read().decode("utf-8")\end{lstlisting}
\end{tcolorbox}%
Usually, you have to preprocess text into embeddings or other vector forms before presentation to a neural network. Hugging Face provides a pipeline that simplifies this process greatly. The pipeline allows you to pass regular Python strings to the transformers and return standard Python values.%
\index{hugging face}%
\index{neural network}%
\index{Python}%
\index{ROC}%
\index{ROC}%
\index{transformer}%
\index{vector}%
\par%
We begin by loading a text{-}classification model. We do not specify the exact model type wanted, so Hugging Face automatically chooses a network from the Hugging Face hub named:%
\index{classification}%
\index{hugging face}%
\index{model}%
\par%
\begin{itemize}[noitemsep]%
\item%
distilbert{-}base{-}uncased{-}finetuned{-}sst{-}2{-}english%
\end{itemize}%
To specify the model to use, pass the model parameter, such as:%
\index{model}%
\index{parameter}%
\par%
\begin{tcolorbox}[size=title,breakable]%
\begin{lstlisting}[upquote=true]
pipe = pipeline(model="roberta-large-mnli")
\end{lstlisting}
\end{tcolorbox}%
The following code loads a model pipeline and a model for sentiment analysis.%
\index{model}%
\index{sentiment analysis}%
\par%
\begin{tcolorbox}[size=title,title=Code,breakable]%
\begin{lstlisting}[language=Python, upquote=true]
import pandas as pd
from transformers import pipeline

classifier = pipeline("text-classification")\end{lstlisting}
\end{tcolorbox}%
We can now display the sentiment analysis results with a Pandas dataframe.%
\index{sentiment analysis}%
\par%
\begin{tcolorbox}[size=title,title=Code,breakable]%
\begin{lstlisting}[language=Python, upquote=true]
outputs = classifier(text)
pd.DataFrame(outputs)\end{lstlisting}
\tcbsubtitle[before skip=\baselineskip]{Output}%
\begin{tabular}[hbt!]{l|l|l}%
\hline%
&label&score\\%
\hline%
0&POSITIVE&0.984666\\%
\hline%
\end{tabular}%
\vspace{2mm}%
\end{tcolorbox}%
As you can see, the poem was considered 0.98 positive.%
\par

\subsection{Entity Tagging}%
\label{subsec:EntityTagging}%
Entity tagging is the process that takes source text and finds parts of that text that represent entities, such as one of the following:%
\index{entity tagging}%
\index{ROC}%
\index{ROC}%
\par%
\begin{itemize}[noitemsep]%
\item%
Location (LOC)%
\item%
Organizations (ORG)%
\index{GAN}%
\item%
Person (PER)%
\item%
Miscellaneous (MISC)%
\end{itemize}%
The following code requests a "named entity recognizer" (ner) and processes the specified text.%
\index{ROC}%
\index{ROC}%
\par%
\begin{tcolorbox}[size=title,title=Code,breakable]%
\begin{lstlisting}[language=Python, upquote=true]
text2 = "Abraham Lincoln was a president who lived in the United States."

tagger = pipeline("ner", aggregation_strategy="simple")\end{lstlisting}
\end{tcolorbox}%
We similarly view the results as a Pandas data frame. As you can see, the person (PER) of Abraham Lincoln and location (LOC) of the United States is recognized.%
\par%
\begin{tcolorbox}[size=title,title=Code,breakable]%
\begin{lstlisting}[language=Python, upquote=true]
outputs = tagger(text2)
pd.DataFrame(outputs)\end{lstlisting}
\tcbsubtitle[before skip=\baselineskip]{Output}%
\begin{tabular}[hbt!]{l|l|l|l|l|l}%
\hline%
&entity\_group&score&word&start&end\\%
\hline%
0&PER&0.998893&Abraham Lincoln&0&15\\%
1&LOC&0.999651&United States&49&62\\%
\hline%
\end{tabular}%
\vspace{2mm}%
\end{tcolorbox}

\subsection{Question Answering}%
\label{subsec:QuestionAnswering}%
Another common task for NLP is question answering from a reference text. We load such a model with the following code.%
\index{model}%
\index{question answering}%
\par%
\begin{tcolorbox}[size=title,title=Code,breakable]%
\begin{lstlisting}[language=Python, upquote=true]
reader = pipeline("question-answering")
question = "What now shall fade?"\end{lstlisting}
\end{tcolorbox}%
For this example, we will pose the question "what shall fade" to Hugging Face for%
\index{hugging face}%
\href{https://en.wikipedia.org/wiki/Sonnet_18}{ Sonnet 18}%
. We see the correct answer of "eternal summer."%
\par%
\begin{tcolorbox}[size=title,title=Code,breakable]%
\begin{lstlisting}[language=Python, upquote=true]
outputs = reader(question=question, context=text)
pd.DataFrame([outputs])\end{lstlisting}
\tcbsubtitle[before skip=\baselineskip]{Output}%
\begin{tabular}[hbt!]{l|l|l|l|l}%
\hline%
&score&start&end&answer\\%
\hline%
0&0.471141&414&428&eternal summer\\%
\hline%
\end{tabular}%
\vspace{2mm}%
\end{tcolorbox}

\subsection{Language Translation}%
\label{subsec:LanguageTranslation}%
Language translation is yet another common task for NLP and Hugging Face.%
\index{hugging face}%
\index{language translation}%
\par%
\begin{tcolorbox}[size=title,title=Code,breakable]%
\begin{lstlisting}[language=Python, upquote=true]
translator = pipeline("translation_en_to_de",
                      model="Helsinki-NLP/opus-mt-en-de")\end{lstlisting}
\end{tcolorbox}%
The following code translates Sonnet 18 from English into German.%
\par%
\begin{tcolorbox}[size=title,title=Code,breakable]%
\begin{lstlisting}[language=Python, upquote=true]
outputs = translator(text, clean_up_tokenization_spaces=True,
                     min_length=100)
print(outputs[0]['translation_text'])\end{lstlisting}
\tcbsubtitle[before skip=\baselineskip]{Output}%
\begin{lstlisting}[upquote=true]
Sonnet 18 Originaltext William Shakespeare Soll ich dich mit einem
Sommertag vergleichen? Du bist schner und gemigter: Raue Winde
schtteln die lieblichen Knospen des Mai, Und der Sommervertrag hat zu
kurz ein Datum: Irgendwann zu hei das Auge des Himmels leuchtet, Und
oft ist sein Gold Teint dimm'd; Und jede faire von Fair irgendwann
sinkt, Durch Zufall oder die Natur wechselnden Kurs untrimm'd; Aber
dein ewiger Sommer wird nicht verblassen noch verlieren Besitz von dem
Schnen du schuld; noch wird der Tod prahlen du wandert in seinem
Schatten, Wenn in ewigen Linien zur Zeit wachsen: So lange die
Menschen atmen oder Augen sehen knnen, So lange lebt dies und dies
gibt dir Leben.
\end{lstlisting}
\end{tcolorbox}

\subsection{Summarization}%
\label{subsec:Summarization}%
Summarization is an NLP task that summarizes a more lengthy text into just a few sentences.%
\index{summarization}%
\par%
\begin{tcolorbox}[size=title,title=Code,breakable]%
\begin{lstlisting}[language=Python, upquote=true]
text2 = """
An apple is an edible fruit produced by an apple tree (Malus domestica). 
Apple trees are cultivated worldwide and are the most widely grown species 
in the genus Malus. The tree originated in Central Asia, where its wild 
ancestor, Malus sieversii, is still found today. Apples have been grown 
for thousands of years in Asia and Europe and were brought to North America 
by European colonists. Apples have religious and mythological significance 
in many cultures, including Norse, Greek, and European Christian tradition.
"""

summarizer = pipeline("summarization")\end{lstlisting}
\end{tcolorbox}%
The following code summarizes the Wikipedia entry for an "apple."%
\par%
\begin{tcolorbox}[size=title,title=Code,breakable]%
\begin{lstlisting}[language=Python, upquote=true]
outputs = summarizer(text2, max_length=45,
                     clean_up_tokenization_spaces=True)
print(outputs[0]['summary_text'])\end{lstlisting}
\tcbsubtitle[before skip=\baselineskip]{Output}%
\begin{lstlisting}[upquote=true]
An apple is an edible fruit produced by an apple tree (Malus
domestica) Apple trees are cultivated worldwide and are the most
widely grown species in the genus Malus. Apples have religious and
mythological
\end{lstlisting}
\end{tcolorbox}

\subsection{Text Generation}%
\label{subsec:TextGeneration}%
Finally, text generation allows us to take an input text and request the pretrained neural network to continue that text.%
\index{input}%
\index{neural network}%
\index{text generation}%
\par%
\begin{tcolorbox}[size=title,title=Code,breakable]%
\begin{lstlisting}[language=Python, upquote=true]
from urllib.request import urlopen

generator = pipeline("text-generation")\end{lstlisting}
\end{tcolorbox}%
Here an example is provided that generates additional text after Sonnet 18.%
\par%
\begin{tcolorbox}[size=title,title=Code,breakable]%
\begin{lstlisting}[language=Python, upquote=true]
outputs = generator(text, max_length=400)
print(outputs[0]['generated_text'])\end{lstlisting}
\tcbsubtitle[before skip=\baselineskip]{Output}%
\begin{lstlisting}[upquote=true]
Sonnet 18 original text
William Shakespeare
Shall I compare thee to a summer's day?
Thou art more lovely and more temperate:
Rough winds do shake the darling buds of May,
And summer's lease hath all too short a date:
Sometime too hot the eye of heaven shines,
And often is his gold complexion dimm'd;
And every fair from fair sometime declines,
By chance or nature's changing course untrimm'd;
But thy eternal summer shall not fade
Nor lose possession of that fair thou owest;
Nor shall Death brag thou wander'st in his shade,
When in eternal lines to time thou growest:
So long as men can breathe or eyes can see,

...

[Italian: The Tale of the
Cat].................................................................
 'Sir! sir la verde'~~~~~~~~~~~~~~~~~~~~~~~~~~~~~~~~~~~~~~~~~~~~~~~~~~
~~~~~~~~~~~~~~~~~~~~~~~~
[Irish: The Tale of
\end{lstlisting}
\end{tcolorbox}

\section{Part 11.2: Hugging Face Tokenizers}%
\label{sec:Part11.2HuggingFaceTokenizers}%
Tokenization is the task of chopping it up into pieces, called tokens, perhaps at the same time throwing away certain characters, such as punctuation. Consider how the program might break up the following sentences into words.%
\index{tokenization}%
\par%
\begin{itemize}[noitemsep]%
\item%
This is a test.%
\item%
Ok, but what about this?%
\item%
Is U.S.A. the same as USA.?%
\item%
What is the best data{-}set to use?%
\item%
I think I will do this{-}no wait; I will do that.%
\end{itemize}%
The hugging face includes tokenizers that can break these sentences into words and subwords. Because English, and some other languages, are made up of common word parts, we tokenize subwords. For example, a gerund word, such as "sleeping," will be tokenized into "sleep" and "\#\#ing".%
\index{hugging face}%
\index{SOM}%
\par%
We begin by installing Hugging Face if needed.%
\index{hugging face}%
\par%
\begin{tcolorbox}[size=title,title=Code,breakable]%
\begin{lstlisting}[language=Python, upquote=true]
!pip install transformers
!pip install transformers[sentencepiece]\end{lstlisting}
\end{tcolorbox}%
First, we create a Hugging Face tokenizer. There are several different tokenizers available from the Hugging Face hub. For this example, we will make use of the following tokenizer:%
\index{hugging face}%
\par%
\begin{itemize}[noitemsep]%
\item%
distilbert{-}base{-}uncased%
\end{itemize}%
This tokenizer is based on BERT and assumes case{-}insensitive English text.%
\par%
\begin{tcolorbox}[size=title,title=Code,breakable]%
\begin{lstlisting}[language=Python, upquote=true]
from transformers import AutoTokenizer
model = "distilbert-base-uncased"
tokenizer = AutoTokenizer.from_pretrained(model)\end{lstlisting}
\end{tcolorbox}%
We can now tokenize a sample sentence.%
\par%
\begin{tcolorbox}[size=title,title=Code,breakable]%
\begin{lstlisting}[language=Python, upquote=true]
encoded = tokenizer('Tokenizing text is easy.')
print(encoded)\end{lstlisting}
\tcbsubtitle[before skip=\baselineskip]{Output}%
\begin{lstlisting}[upquote=true]
{'input_ids': [101, 19204, 6026, 3793, 2003, 3733, 1012, 102],
'attention_mask': [1, 1, 1, 1, 1, 1, 1, 1]}
\end{lstlisting}
\end{tcolorbox}%
The result of this tokenization contains two elements:%
\index{tokenization}%
\par%
\begin{itemize}[noitemsep]%
\item%
input\_ids {-} The individual subword indexes, each index uniquely identifies a subword.%
\index{input}%
\item%
attention\_mask {-} Which values in%
\textit{ input\_ids }%
are meaningful and not  padding.%
\end{itemize}%
This sentence had no padding, so all elements have an attention mask of "1". Later, we will request the output to be of a fixed length, introducing padding, which always has an attention mask of "0". Though each tokenizer can be implemented differently, the attention mask of a tokenizer is generally either "0" or "1".%
\index{output}%
\par%
Due to subwords and special tokens, the number of tokens may not match the number of words in the source string. We can see the meanings of the individual tokens by converting these IDs back to strings.%
\par%
\begin{tcolorbox}[size=title,title=Code,breakable]%
\begin{lstlisting}[language=Python, upquote=true]
tokenizer.convert_ids_to_tokens(encoded.input_ids)\end{lstlisting}
\tcbsubtitle[before skip=\baselineskip]{Output}%
\begin{lstlisting}[upquote=true]
['[CLS]', 'token', '##izing', 'text', 'is', 'easy', '.', '[SEP]']
\end{lstlisting}
\end{tcolorbox}%
As you can see, there are two special tokens placed at the beginning and end of each sequence. We will soon see how we can include or exclude these special tokens. These special tokens can vary per tokenizer; however, {[}CLS{]} begins a sequence for this tokenizer, and {[}SEP{]} ends a sequence. You will also see that the gerund "tokening" is broken into "token" and "*ing".%
\par%
For this tokenizer, the special tokens occur between 100 and 103. Most Hugging Face tokenizers use this approximate range for special tokens. The value zero (0) typically represents padding. We can display all special tokens with this command.%
\index{hugging face}%
\par%
\begin{tcolorbox}[size=title,title=Code,breakable]%
\begin{lstlisting}[language=Python, upquote=true]
tokenizer.convert_ids_to_tokens([0, 100, 101, 102, 103])\end{lstlisting}
\tcbsubtitle[before skip=\baselineskip]{Output}%
\begin{lstlisting}[upquote=true]
['[PAD]', '[UNK]', '[CLS]', '[SEP]', '[MASK]']
\end{lstlisting}
\end{tcolorbox}%
This tokenizer supports these common tokens:%
\par%
\begin{itemize}[noitemsep]%
\item%
{[}CLS{]} {-} Sequence beginning.%
\item%
{[}SEP{]} {-} Sequence end.%
\item%
{[}PAD{]} {-} Padding.%
\item%
{[}UNK{]} {-} Unknown token.%
\item%
{[}MASK{]} {-} Mask out tokens for a neural network to predict. Not used in this book, see%
\index{neural network}%
\index{predict}%
\href{https://arxiv.org/abs/2109.01819}{ MLM paper}%
.%
\end{itemize}%
It is also possible to tokenize lists of sequences. We can pad and truncate sequences to achieve a standard length by tokenizing many sequences at once.%
\par%
\begin{tcolorbox}[size=title,title=Code,breakable]%
\begin{lstlisting}[language=Python, upquote=true]
text = [
    "This movie was great!",
    "I hated this move, waste of time!",
    "Epic?"
]

encoded = tokenizer(text, padding=True, add_special_tokens=True)

print("**Input IDs**")
for a in encoded.input_ids:
    print(a)

print("**Attention Mask**")
for a in encoded.attention_mask:
    print(a)\end{lstlisting}
\tcbsubtitle[before skip=\baselineskip]{Output}%
\begin{lstlisting}[upquote=true]
**Input IDs**
[101, 2023, 3185, 2001, 2307, 999, 102, 0, 0, 0, 0]
[101, 1045, 6283, 2023, 2693, 1010, 5949, 1997, 2051, 999, 102]
[101, 8680, 1029, 102, 0, 0, 0, 0, 0, 0, 0]
**Attention Mask**
[1, 1, 1, 1, 1, 1, 1, 0, 0, 0, 0]
[1, 1, 1, 1, 1, 1, 1, 1, 1, 1, 1]
[1, 1, 1, 1, 0, 0, 0, 0, 0, 0, 0]
\end{lstlisting}
\end{tcolorbox}%
Notice the%
\textbf{ input\_id}%
's for the three movie review text sequences. Each of these sequences begins with 101 and we pad with zeros. Just before the padding, each group of IDs ends with 102. The attention masks also have zeros for each of the padding entries.%
\par%
We used two parameters to the tokenizer to control the tokenization process. Some other useful%
\index{parameter}%
\index{ROC}%
\index{ROC}%
\index{SOM}%
\index{tokenization}%
\href{https://huggingface.co/docs/transformers/main_classes/tokenizer}{ parameters }%
include:%
\par%
\begin{itemize}[noitemsep]%
\item%
add\_special\_tokens (defaults to True) Whether or not to encode the sequences with the special tokens relative to their model.%
\index{model}%
\item%
padding (defaults to False) Activates and controls truncation.%
\item%
max\_length (optional) Controls the maximum length to use by one of the truncation/padding parameters.%
\index{parameter}%
\end{itemize}

\section{Part 11.3: Hugging Face Datasets}%
\label{sec:Part11.3HuggingFaceDatasets}%
The Hugging Face hub includes data sets useful for natural language processing (NLP). The Hugging Face library provides functions that allow you to navigate and obtain these data sets. When we access Hugging Face data sets, the data is in a format specific to Hugging Face. In this part, we will explore this format and see how to convert it to Pandas or TensorFlow data.%
\index{hugging face}%
\index{ROC}%
\index{ROC}%
\index{TensorFlow}%
\par%
We begin by installing Hugging Face if needed. It is also essential to install Hugging Face datasets.%
\index{dataset}%
\index{hugging face}%
\par%
\begin{tcolorbox}[size=title,title=Code,breakable]%
\begin{lstlisting}[language=Python, upquote=true]
!pip install transformers
!pip install transformers[sentencepiece]
!pip install datasets\end{lstlisting}
\end{tcolorbox}%
We begin by querying Hugging Face to obtain the total count and names of the data sets. This code obtains the total count and the names of the first five datasets.%
\index{dataset}%
\index{hugging face}%
\par%
\begin{tcolorbox}[size=title,title=Code,breakable]%
\begin{lstlisting}[language=Python, upquote=true]
from datasets import list_datasets

all_datasets = list_datasets()

print(f"Hugging Face hub currently contains {len(all_datasets)}")
print(f"datasets. The first 5 are:")
print("\n".join(all_datasets[:10]))\end{lstlisting}
\tcbsubtitle[before skip=\baselineskip]{Output}%
\begin{lstlisting}[upquote=true]
Hugging Face hub currently contains 3832
datasets. The first 5 are:
acronym_identification
ade_corpus_v2
adversarial_qa
aeslc
afrikaans_ner_corpus
ag_news
ai2_arc
air_dialogue
ajgt_twitter_ar
allegro_reviews
\end{lstlisting}
\end{tcolorbox}%
We begin by loading the emotion data set from the Hugging Face hub.%
\index{hugging face}%
\href{https://huggingface.co/datasets/emotion}{ Emotion }%
is a dataset of English Twitter messages with six basic emotions: anger, fear, joy, love, sadness, and surprise.%
\index{dataset}%
\cite{saravia2018carer}%
The following code loads the emotion data set from the Hugging Face hub.%
\index{hugging face}%
\par%
\begin{tcolorbox}[size=title,title=Code,breakable]%
\begin{lstlisting}[language=Python, upquote=true]
from datasets import load_dataset

emotions = load_dataset("emotion")\end{lstlisting}
\tcbsubtitle[before skip=\baselineskip]{Output}%
\begin{lstlisting}[upquote=true]
Downloading builder script:   0%|          | 0.00/1.66k [00:00<?,
?B/s]Downloading metadata:   0%|          | 0.00/1.61k [00:00<?,
?B/s]Downloading and preparing dataset emotion/default (download: 1.97
MiB, generated: 2.07 MiB, post-processed: Unknown size, total: 4.05
MiB) to /root/.cache/huggingface/datasets/emotion/default/0.0.0/348f63
ca8e27b3713b6c04d723efe6d824a56fb3d1449794716c0f0296072705...
Downloading data:   0%|          | 0.00/1.66M [00:00<?,
?B/s]Downloading data:   0%|          | 0.00/204k [00:00<?,
?B/s]Downloading data:   0%|          | 0.00/207k [00:00<?,
?B/s]Generating train split:   0%|          | 0/16000 [00:00<?, ?
examples/s]Generating validation split:   0%|          | 0/2000
[00:00<?, ? examples/s]Generating test split:   0%|          | 0/2000
[00:00<?, ? examples/s]Dataset emotion downloaded and prepared to /roo
t/.cache/huggingface/datasets/emotion/default/0.0.0/348f63ca8e27b3713b
6c04d723efe6d824a56fb3d1449794716c0f0296072705. Subsequent calls will
reuse this data.
  0%|          | 0/3 [00:00<?, ?it/s]
\end{lstlisting}
\end{tcolorbox}%
A quick scan of the downloaded data set reveals its structure. In this case, Hugging Face already separated the data into training, validation, and test data sets. The training set consists of 16,000 observations, while the test and validation sets contain 2,000 observations. The dataset is a Python dictionary that includes a Dataset object for each of these three divisions. The datasets only contain two columns, the text and the emotion label for each text sample.%
\index{dataset}%
\index{hugging face}%
\index{Python}%
\index{training}%
\index{validation}%
\par%
\begin{tcolorbox}[size=title,title=Code,breakable]%
\begin{lstlisting}[language=Python, upquote=true]
emotions\end{lstlisting}
\tcbsubtitle[before skip=\baselineskip]{Output}%
\begin{lstlisting}[upquote=true]
DatasetDict({
    train: Dataset({
        features: ['text', 'label'],
        num_rows: 16000
    })
    validation: Dataset({
        features: ['text', 'label'],
        num_rows: 2000
    })
    test: Dataset({
        features: ['text', 'label'],
        num_rows: 2000
    })
})
\end{lstlisting}
\end{tcolorbox}%
You can see a single observation from the training data set here. This observation includes both the text sample and the assigned emotion label. The label is a numeric index representing the assigned emotion.%
\index{training}%
\par%
\begin{tcolorbox}[size=title,title=Code,breakable]%
\begin{lstlisting}[language=Python, upquote=true]
emotions['train'][2]\end{lstlisting}
\tcbsubtitle[before skip=\baselineskip]{Output}%
\begin{lstlisting}[upquote=true]
{'label': 3, 'text': 'im grabbing a minute to post i feel greedy
wrong'}
\end{lstlisting}
\end{tcolorbox}%
We can display the labels in order of their index labels.%
\par%
\begin{tcolorbox}[size=title,title=Code,breakable]%
\begin{lstlisting}[language=Python, upquote=true]
emotions['train'].features\end{lstlisting}
\tcbsubtitle[before skip=\baselineskip]{Output}%
\begin{lstlisting}[upquote=true]
{'label': ClassLabel(num_classes=6, names=['sadness', 'joy', 'love',
'anger', 'fear', 'surprise'], id=None),
 'text': Value(dtype='string', id=None)}
\end{lstlisting}
\end{tcolorbox}%
Hugging face can provide these data sets in a variety of formats. The following code receives the emotion data set as a Pandas data frame.%
\index{hugging face}%
\par%
\begin{tcolorbox}[size=title,title=Code,breakable]%
\begin{lstlisting}[language=Python, upquote=true]
import pandas as pd

emotions.set_format(type='pandas')
df = emotions["train"][:]
df[:5]\end{lstlisting}
\tcbsubtitle[before skip=\baselineskip]{Output}%
\begin{tabular}[hbt!]{l|l|l}%
\hline%
&text&label\\%
\hline%
0&i didnt feel humiliated&0\\%
1&i can go from feeling so hopeless to so damned...&0\\%
2&im grabbing a minute to post i feel greedy wrong&3\\%
3&i am ever feeling nostalgic about the fireplac...&2\\%
4&i am feeling grouchy&3\\%
\hline%
\end{tabular}%
\vspace{2mm}%
\end{tcolorbox}%
We can use the Pandas "apply" function to add the textual label for each observation.%
\par%
\begin{tcolorbox}[size=title,title=Code,breakable]%
\begin{lstlisting}[language=Python, upquote=true]
def label_it(row):
  return emotions["train"].features["label"].int2str(row)


df['label_name'] = df["label"].apply(label_it)
df[:5]\end{lstlisting}
\tcbsubtitle[before skip=\baselineskip]{Output}%
\begin{tabular}[hbt!]{l|l|l|l}%
\hline%
&text&label&label\_name\\%
\hline%
0&i didnt feel humiliated&0&sadness\\%
1&i can go from feeling so hopeless to so damned...&0&sadness\\%
2&im grabbing a minute to post i feel greedy wrong&3&anger\\%
3&i am ever feeling nostalgic about the fireplac...&2&love\\%
4&i am feeling grouchy&3&anger\\%
\hline%
\end{tabular}%
\vspace{2mm}%
\end{tcolorbox}%
With the data in Pandas format and textually labeled, we can display a bar chart of the frequency of each of the emotions.%
\par%
\begin{tcolorbox}[size=title,title=Code,breakable]%
\begin{lstlisting}[language=Python, upquote=true]
import matplotlib.pyplot as plt

df["label_name"].value_counts(ascending=True).plot.barh()
plt.show()\end{lstlisting}
\tcbsubtitle[before skip=\baselineskip]{Output}%
\includegraphics[width=3in]{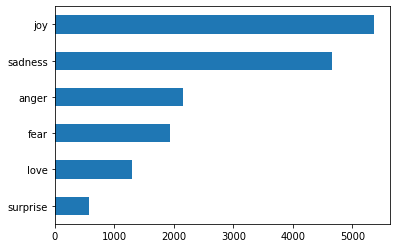}%
\end{tcolorbox}%
Finally, we utilize Hugging Face tokenizers and data sets together. The following code tokenizes the entire emotion data set. You can see below that the code has transformed the training set into subword tokens that are now ready to be used in conjunction with a transformer for either inference or training.%
\index{hugging face}%
\index{training}%
\index{transformer}%
\par%
\begin{tcolorbox}[size=title,title=Code,breakable]%
\begin{lstlisting}[language=Python, upquote=true]
from transformers import AutoTokenizer


def tokenize(rows):
    return tokenizer(rows['text'], padding=True, truncation=True)


model_ckpt = "distilbert-base-uncased"
tokenizer = AutoTokenizer.from_pretrained(model_ckpt)

emotions.set_format(type=None)

encoded = tokenize(emotions["train"][:2])

print("**Input IDs**")
for a in encoded.input_ids:
    print(a)\end{lstlisting}
\tcbsubtitle[before skip=\baselineskip]{Output}%
\begin{lstlisting}[upquote=true]
Downloading:   0%|          | 0.00/28.0 [00:00<?, ?B/s]Downloading:
0%|          | 0.00/483 [00:00<?, ?B/s]Downloading:   0%|          |
0.00/226k [00:00<?, ?B/s]Downloading:   0%|          | 0.00/455k
[00:00<?, ?B/s]**Input IDs**
[101, 1045, 2134, 2102, 2514, 26608, 102, 0, 0, 0, 0, 0, 0, 0, 0, 0,
0, 0, 0, 0, 0, 0, 0]
[101, 1045, 2064, 2175, 2013, 3110, 2061, 20625, 2000, 2061, 9636,
17772, 2074, 2013, 2108, 2105, 2619, 2040, 14977, 1998, 2003, 8300,
102]
\end{lstlisting}
\end{tcolorbox}

\section{Part 11.4: Training Hugging Face Models}%
\label{sec:Part11.4TrainingHuggingFaceModels}%
Up to this point, we've used data and models from the Hugging Face hub unmodified. In this section, we will transfer and train a Hugging Face model. We will use Hugging Face data sets, tokenizers, and pretrained models to achieve this training.%
\index{hugging face}%
\index{model}%
\index{training}%
\par%
We begin by installing Hugging Face if needed. It is also essential to install Hugging Face datasets.%
\index{dataset}%
\index{hugging face}%
\par%
\begin{tcolorbox}[size=title,title=Code,breakable]%
\begin{lstlisting}[language=Python, upquote=true]
!pip install transformers
!pip install transformers[sentencepiece]
!pip install datasets\end{lstlisting}
\end{tcolorbox}%
We begin by loading the emotion data set from the Hugging Face hub. Emotion is a dataset of English Twitter messages with six basic emotions: anger, fear, joy, love, sadness, and surprise. The following code loads the emotion data set from the Hugging Face hub.%
\index{dataset}%
\index{hugging face}%
\par%
\begin{tcolorbox}[size=title,title=Code,breakable]%
\begin{lstlisting}[language=Python, upquote=true]
from datasets import load_dataset

emotions = load_dataset("emotion")\end{lstlisting}
\end{tcolorbox}%
You can see a single observation from the training data set here. This observation includes both the text sample and the assigned emotion label. The label is a numeric index representing the assigned emotion.%
\index{training}%
\par%
\begin{tcolorbox}[size=title,title=Code,breakable]%
\begin{lstlisting}[language=Python, upquote=true]
emotions['train'][2]\end{lstlisting}
\tcbsubtitle[before skip=\baselineskip]{Output}%
\begin{lstlisting}[upquote=true]
{'label': 3, 'text': 'im grabbing a minute to post i feel greedy
wrong'}
\end{lstlisting}
\end{tcolorbox}%
We can display the labels in order of their index labels.%
\par%
\begin{tcolorbox}[size=title,title=Code,breakable]%
\begin{lstlisting}[language=Python, upquote=true]
emotions['train'].features\end{lstlisting}
\tcbsubtitle[before skip=\baselineskip]{Output}%
\begin{lstlisting}[upquote=true]
{'label': ClassLabel(num_classes=6, names=['sadness', 'joy', 'love',
'anger', 'fear', 'surprise'], id=None),
 'text': Value(dtype='string', id=None)}
\end{lstlisting}
\end{tcolorbox}%
Next, we utilize Hugging Face tokenizers and data sets together. The following code tokenizes the entire emotion data set. You can see below that the code has transformed the training set into subword tokens that are now ready to be used in conjunction with a transformer for either inference or training.%
\index{hugging face}%
\index{training}%
\index{transformer}%
\par%
\begin{tcolorbox}[size=title,title=Code,breakable]%
\begin{lstlisting}[language=Python, upquote=true]
from transformers import AutoTokenizer


def tokenize(rows):
    return tokenizer(rows['text'], padding="max_length", truncation=True)


model_ckpt = "distilbert-base-uncased"
tokenizer = AutoTokenizer.from_pretrained(model_ckpt)

emotions.set_format(type=None)

tokenized_datasets = emotions.map(tokenize, batched=True)\end{lstlisting}
\end{tcolorbox}%
We will utilize the Hugging Face%
\index{hugging face}%
\textbf{ DefaultDataCollator }%
to transform the emotion data set into TensorFlow type data that we can use to finetune a neural network.%
\index{neural network}%
\index{TensorFlow}%
\par%
\begin{tcolorbox}[size=title,title=Code,breakable]%
\begin{lstlisting}[language=Python, upquote=true]
from transformers import DefaultDataCollator

data_collator = DefaultDataCollator(return_tensors="tf")\end{lstlisting}
\end{tcolorbox}%
Now we generate a shuffled training and evaluation data set.%
\index{training}%
\par%
\begin{tcolorbox}[size=title,title=Code,breakable]%
\begin{lstlisting}[language=Python, upquote=true]
small_train_dataset = tokenized_datasets["train"].shuffle(seed=42)
small_eval_dataset = tokenized_datasets["test"].shuffle(seed=42)\end{lstlisting}
\end{tcolorbox}%
We can now generate the TensorFlow data sets. We specify which columns should map to the input features and labels. We do not need to shuffle because we previously shuffled the data.%
\index{feature}%
\index{input}%
\index{TensorFlow}%
\par%
\begin{tcolorbox}[size=title,title=Code,breakable]%
\begin{lstlisting}[language=Python, upquote=true]
tf_train_dataset = small_train_dataset.to_tf_dataset(
    columns=["attention_mask", "input_ids", "token_type_ids"],
    label_cols=["labels"],
    shuffle=True,
    collate_fn=data_collator,
    batch_size=8,
)

tf_validation_dataset = small_eval_dataset.to_tf_dataset(
    columns=["attention_mask", "input_ids", "token_type_ids"],
    label_cols=["labels"],
    shuffle=False,
    collate_fn=data_collator,
    batch_size=8,
)\end{lstlisting}
\end{tcolorbox}%
We will now load the distilbert model for classification. We will adjust the pretrained weights to predict the emotions of text lines.%
\index{classification}%
\index{model}%
\index{predict}%
\par%
\begin{tcolorbox}[size=title,title=Code,breakable]%
\begin{lstlisting}[language=Python, upquote=true]
import tensorflow as tf
from transformers import TFAutoModelForSequenceClassification

model = TFAutoModelForSequenceClassification.from_pretrained(\
    "distilbert-base-uncased", num_labels=6)\end{lstlisting}
\end{tcolorbox}%
We now train the neural network. Because the network is already pretrained, we use a small learning rate.%
\index{learning}%
\index{learning rate}%
\index{neural network}%
\par%
\begin{tcolorbox}[size=title,title=Code,breakable]%
\begin{lstlisting}[language=Python, upquote=true]
model.compile(
    optimizer=tf.keras.optimizers.Adam(learning_rate=5e-5),
    loss=tf.keras.losses.SparseCategoricalCrossentropy(from_logits=True),
    metrics=tf.metrics.SparseCategoricalAccuracy(),
)

model.fit(tf_train_dataset, validation_data=tf_validation_dataset,
          epochs=5)\end{lstlisting}
\tcbsubtitle[before skip=\baselineskip]{Output}%
\begin{lstlisting}[upquote=true]
...
2000/2000 [==============================] - 346s 173ms/step - loss:
0.1092 - sparse_categorical_accuracy: 0.9486 - val_loss: 0.1654 -
val_sparse_categorical_accuracy: 0.9295
Epoch 5/5
2000/2000 [==============================] - 347s 173ms/step - loss:
0.0960 - sparse_categorical_accuracy: 0.9585 - val_loss: 0.1830 -
val_sparse_categorical_accuracy: 0.9220
\end{lstlisting}
\end{tcolorbox}

\section{Part 11.5: What are Embedding Layers in Keras}%
\label{sec:Part11.5WhatareEmbeddingLayersinKeras}%
\href{https://keras.io/layers/embeddings/}{Embedding Layers }%
are a handy feature of Keras that allows the program to automatically insert additional information into the data flow of your neural network. In the previous section, you saw that Word2Vec could expand words to a 300 dimension vector. An embedding layer would automatically allow you to insert these 300{-}dimension vectors in the place of word indexes.%
\index{feature}%
\index{Keras}%
\index{layer}%
\index{neural network}%
\index{vector}%
\par%
Programmers often use embedding layers with Natural Language Processing (NLP); however, you can use these layers when you wish to insert a lengthier vector in an index value place. In some ways, you can think of an embedding layer as dimension expansion. However, the hope is that these additional dimensions provide more information to the model and provide a better score.%
\index{layer}%
\index{model}%
\index{ROC}%
\index{ROC}%
\index{SOM}%
\index{vector}%
\par%
\subsection{Simple Embedding Layer Example}%
\label{subsec:SimpleEmbeddingLayerExample}%
\begin{itemize}[noitemsep]%
\item%
\textbf{input\_dim }%
= How large is the vocabulary?  How many categories are you encoding? This parameter is the number of items in your "lookup table."%
\index{parameter}%
\item%
\textbf{output\_dim }%
= How many numbers in the vector you wish to return.%
\index{vector}%
\item%
\textbf{input\_length }%
= How many items are in the input feature vector that you need to transform?%
\index{feature}%
\index{input}%
\index{vector}%
\end{itemize}%
Now we create a neural network with a vocabulary size of 10, which will reduce those values between 0{-}9 to 4 number vectors. This neural network does nothing more than passing the embedding on to the output. But it does let us see what the embedding is doing. Each feature vector coming in will have two such features.%
\index{feature}%
\index{neural network}%
\index{output}%
\index{vector}%
\par%
\begin{tcolorbox}[size=title,title=Code,breakable]%
\begin{lstlisting}[language=Python, upquote=true]
from tensorflow.keras.models import Sequential
from tensorflow.keras.layers import Embedding
import numpy as np

model = Sequential()
embedding_layer = Embedding(input_dim=10, output_dim=4, input_length=2)
model.add(embedding_layer)
model.compile('adam', 'mse')\end{lstlisting}
\end{tcolorbox}%
Let's take a look at the structure of this neural network to see what is happening inside it.%
\index{neural network}%
\par%
\begin{tcolorbox}[size=title,title=Code,breakable]%
\begin{lstlisting}[language=Python, upquote=true]
model.summary()\end{lstlisting}
\tcbsubtitle[before skip=\baselineskip]{Output}%
\begin{lstlisting}[upquote=true]
Model: "sequential"
_________________________________________________________________
 Layer (type)                Output Shape              Param #
=================================================================
 embedding (Embedding)       (None, 2, 4)              40
=================================================================
Total params: 40
Trainable params: 40
Non-trainable params: 0
_________________________________________________________________
\end{lstlisting}
\end{tcolorbox}%
For this neural network, which is just an embedding layer, the input is a vector of size 2. These two inputs are integer numbers from 0 to 9 (corresponding to the requested input\_dim quantity of 10 values). Looking at the summary above, we see that the embedding layer has 40 parameters. This value comes from the embedded lookup table that contains four amounts (output\_dim) for each of the 10 (input\_dim) possible integer values for the two inputs. The output is 2 (input\_length) length 4 (output\_dim) vectors, resulting in a total output size of 8, which corresponds to the Output Shape given in the summary above.%
\index{input}%
\index{layer}%
\index{neural network}%
\index{output}%
\index{parameter}%
\index{vector}%
\par%
Now, let us query the neural network with two rows. The input is two integer values, as was specified when we created the neural network.%
\index{input}%
\index{neural network}%
\par%
\begin{tcolorbox}[size=title,title=Code,breakable]%
\begin{lstlisting}[language=Python, upquote=true]
input_data = np.array([
    [1, 2]
])

pred = model.predict(input_data)

print(input_data.shape)
print(pred)\end{lstlisting}
\tcbsubtitle[before skip=\baselineskip]{Output}%
\begin{lstlisting}[upquote=true]
(1, 2)
[[[-0.04494917  0.01937468 -0.00152863  0.04808659]
  [-0.04002655  0.03441895  0.04462588 -0.01472597]]]
\end{lstlisting}
\end{tcolorbox}%
Here we see two length{-}4 vectors that Keras looked up for each input integer. Recall that Python arrays are zero{-}based. Keras replaced the value of 1 with the second row of the 10 x 4 lookup matrix. Similarly, Keras returned the value of 2 by the third row of the lookup matrix. The following code displays the lookup matrix in its entirety. The embedding layer performs no mathematical operations other than inserting the correct row from the lookup table.%
\index{input}%
\index{Keras}%
\index{layer}%
\index{matrix}%
\index{Python}%
\index{vector}%
\par%
\begin{tcolorbox}[size=title,title=Code,breakable]%
\begin{lstlisting}[language=Python, upquote=true]
embedding_layer.get_weights()\end{lstlisting}
\tcbsubtitle[before skip=\baselineskip]{Output}%
\begin{lstlisting}[upquote=true]
[array([[-0.03164196,  0.02898774, -0.0273805 ,  0.01066511],
        [-0.04494917,  0.01937468, -0.00152863,  0.04808659],
        [-0.04002655,  0.03441895,  0.04462588, -0.01472597],
        [ 0.02480464, -0.02585896,  0.0099823 ,  0.02589831],
        [-0.02502655,  0.02517617, -0.03199299,  0.00127842],
        [-0.00205797,  0.02709344, -0.04335414, -0.01793201],
        [ 0.03926537,  0.0293855 ,  0.0445295 , -0.02160555],
        [-0.0075082 , -0.03241253,  0.04906586, -0.02384975],
        [ 0.00264529, -0.01921672, -0.0031809 ,  0.00151991],
        [-0.02407705, -0.04659952, -0.02667597, -0.04108504]],
       dtype=float32)]
\end{lstlisting}
\end{tcolorbox}%
The values above are random parameters that Keras generated as starting points.  Generally, we will transfer an embedding or train these random values into something useful.  The following section demonstrates how to embed a hand{-}coded embedding.%
\index{Keras}%
\index{parameter}%
\index{random}%
\index{SOM}%
\par

\subsection{Transferring An Embedding}%
\label{subsec:TransferringAnEmbedding}%
Now, we see how to hard{-}code an embedding lookup that performs a simple one{-}hot encoding.  One{-}hot encoding would transform the input integer values of 0, 1, and 2 to the vectors $[1,0,0]$, $[0,1,0]$, and $[0,0,1]$ respectively. The following code replaced the random lookup values in the embedding layer with this one{-}hot coding{-}inspired lookup table.%
\index{input}%
\index{layer}%
\index{random}%
\index{vector}%
\par%
\begin{tcolorbox}[size=title,title=Code,breakable]%
\begin{lstlisting}[language=Python, upquote=true]
from tensorflow.keras.models import Sequential
from tensorflow.keras.layers import Embedding
import numpy as np

embedding_lookup = np.array([
    [1, 0, 0],
    [0, 1, 0],
    [0, 0, 1]
])

model = Sequential()
embedding_layer = Embedding(input_dim=3, output_dim=3, input_length=2)
model.add(embedding_layer)
model.compile('adam', 'mse')

embedding_layer.set_weights([embedding_lookup])\end{lstlisting}
\end{tcolorbox}%
We have the following parameters for the Embedding layer:%
\index{layer}%
\index{parameter}%
\par%
\begin{itemize}[noitemsep]%
\item%
input\_dim=3 {-} There are three different integer categorical values allowed.%
\index{categorical}%
\index{input}%
\item%
output\_dim=3 {-} Three columns represent a categorical value with three possible values per one{-}hot encoding.%
\index{categorical}%
\index{output}%
\item%
input\_length=2 {-} The input vector has two of these categorical values.%
\index{categorical}%
\index{input}%
\index{input vector}%
\index{vector}%
\end{itemize}%
We query the neural network with two categorical values to see the lookup performed.%
\index{categorical}%
\index{neural network}%
\par%
\begin{tcolorbox}[size=title,title=Code,breakable]%
\begin{lstlisting}[language=Python, upquote=true]
input_data = np.array([
    [0, 1]
])

pred = model.predict(input_data)

print(input_data.shape)
print(pred)\end{lstlisting}
\tcbsubtitle[before skip=\baselineskip]{Output}%
\begin{lstlisting}[upquote=true]
(1, 2)
[[[1. 0. 0.]
  [0. 1. 0.]]]
\end{lstlisting}
\end{tcolorbox}%
The given output shows that we provided the program with two rows from the one{-}hot encoding table. This encoding is a correct one{-}hot encoding for the values 0 and 1, where there are up to 3 unique values possible.%
\index{output}%
\par%
The following section demonstrates how to train this embedding lookup table.%
\par

\subsection{Training an Embedding}%
\label{subsec:TraininganEmbedding}%
First, we make use of the following imports.%
\par%
\begin{tcolorbox}[size=title,title=Code,breakable]%
\begin{lstlisting}[language=Python, upquote=true]
from numpy import array
from tensorflow.keras.preprocessing.text import one_hot
from tensorflow.keras.preprocessing.sequence import pad_sequences
from tensorflow.keras.models import Sequential
from tensorflow.keras.layers import Flatten, Embedding, Dense\end{lstlisting}
\end{tcolorbox}%
We create a neural network that classifies restaurant reviews according to positive or negative.  This neural network can accept strings as input, such as given here.  This code also includes positive or negative labels for each review.%
\index{input}%
\index{neural network}%
\par%
\begin{tcolorbox}[size=title,title=Code,breakable]%
\begin{lstlisting}[language=Python, upquote=true]
# Define 10 resturant reviews.
reviews = [
    'Never coming back!',
    'Horrible service',
    'Rude waitress',
    'Cold food.',
    'Horrible food!',
    'Awesome',
    'Awesome service!',
    'Rocks!',
    'poor work',
    'Couldn\'t have done better']

# Define labels (1=negative, 0=positive)
labels = array([1, 1, 1, 1, 1, 0, 0, 0, 0, 0])\end{lstlisting}
\end{tcolorbox}%
Notice that the second to the last label is incorrect.  Errors such as this are not too out of the ordinary, as most training data could have some noise.%
\index{error}%
\index{SOM}%
\index{training}%
\par%
We define a vocabulary size of 50 words.  Though we do not have 50 words, it is okay to use a value larger than needed.  If there are more than 50 words, the least frequently used words in the training set are automatically dropped by the embedding layer during training.  For input, we one{-}hot encode the strings.  We use the TensorFlow one{-}hot encoding method here rather than Scikit{-}Learn. Scikit{-}learn would expand these strings to the 0's and 1's as we would typically see for dummy variables.  TensorFlow translates all words to index values and replaces each word with that index.%
\index{input}%
\index{layer}%
\index{TensorFlow}%
\index{training}%
\par%
\begin{tcolorbox}[size=title,title=Code,breakable]%
\begin{lstlisting}[language=Python, upquote=true]
VOCAB_SIZE = 50
encoded_reviews = [one_hot(d, VOCAB_SIZE) for d in reviews]
print(f"Encoded reviews: {encoded_reviews}")\end{lstlisting}
\tcbsubtitle[before skip=\baselineskip]{Output}%
\begin{lstlisting}[upquote=true]
Encoded reviews: [[40, 43, 7], [27, 31], [49, 46], [2, 28], [27, 28],
[20], [20, 31], [39], [18, 39], [11, 3, 18, 11]]
\end{lstlisting}
\end{tcolorbox}%
The program one{-}hot encodes these reviews to word indexes; however, their lengths are different.  We pad these reviews to 4 words and truncate any words beyond the fourth word.%
\par%
\begin{tcolorbox}[size=title,title=Code,breakable]%
\begin{lstlisting}[language=Python, upquote=true]
MAX_LENGTH = 4

padded_reviews = pad_sequences(encoded_reviews, maxlen=MAX_LENGTH,
                               padding='post')
print(padded_reviews)\end{lstlisting}
\tcbsubtitle[before skip=\baselineskip]{Output}%
\begin{lstlisting}[upquote=true]
[[40 43  7  0]
 [27 31  0  0]
 [49 46  0  0]
 [ 2 28  0  0]
 [27 28  0  0]
 [20  0  0  0]
 [20 31  0  0]
 [39  0  0  0]
 [18 39  0  0]
 [11  3 18 11]]
\end{lstlisting}
\end{tcolorbox}%
As specified by the%
\textbf{ padding=post }%
setting, each review is padded by appending zeros at the end, as specified by the%
\textbf{ padding=post }%
setting.%
\par%
Next, we create a neural network to learn to classify these reviews.%
\index{neural network}%
\par%
\begin{tcolorbox}[size=title,title=Code,breakable]%
\begin{lstlisting}[language=Python, upquote=true]
model = Sequential()
embedding_layer = Embedding(VOCAB_SIZE, 8, input_length=MAX_LENGTH)
model.add(embedding_layer)
model.add(Flatten())
model.add(Dense(1, activation='sigmoid'))
model.compile(optimizer='adam', loss='binary_crossentropy',
              metrics=['acc'])

print(model.summary())\end{lstlisting}
\tcbsubtitle[before skip=\baselineskip]{Output}%
\begin{lstlisting}[upquote=true]
Model: "sequential_2"
_________________________________________________________________
 Layer (type)                Output Shape              Param #
=================================================================
 embedding_2 (Embedding)     (None, 4, 8)              400
 flatten (Flatten)           (None, 32)                0
 dense (Dense)               (None, 1)                 33
=================================================================
Total params: 433
Trainable params: 433
Non-trainable params: 0
_________________________________________________________________
None
\end{lstlisting}
\end{tcolorbox}%
This network accepts four integer inputs that specify the indexes of a padded movie review. The first embedding layer converts these four indexes into four length vectors 8. These vectors come from the lookup table that contains 50 (VOCAB\_SIZE) rows of vectors of length 8. This encoding is evident by the 400 (8 times 50) parameters in the embedding layer. The output size from the embedding layer is 32 (4 words expressed as 8{-}number embedded vectors). A single output neuron is connected to the embedding layer by 33 weights (32 from the embedding layer and a single bias neuron). Because this is a single{-}class classification network, we use the sigmoid activation function and binary\_crossentropy.%
\index{activation function}%
\index{bias}%
\index{bias neuron}%
\index{classification}%
\index{input}%
\index{layer}%
\index{neuron}%
\index{output}%
\index{output neuron}%
\index{parameter}%
\index{sigmoid}%
\index{vector}%
\par%
The program now trains the neural network. The embedding lookup and dense 33 weights are updated to produce a better score.%
\index{neural network}%
\par%
\begin{tcolorbox}[size=title,title=Code,breakable]%
\begin{lstlisting}[language=Python, upquote=true]
# fit the model
model.fit(padded_reviews, labels, epochs=100, verbose=0)\end{lstlisting}
\tcbsubtitle[before skip=\baselineskip]{Output}%
\begin{lstlisting}[upquote=true]

\end{lstlisting}
\end{tcolorbox}%
We can see the learned embeddings.  Think of each word's vector as a location in the 8 dimension space where words associated with positive reviews are close to other words.  Similarly, training places negative reviews close to each other.  In addition to the training setting these embeddings, the 33 weights between the embedding layer and output neuron similarly learn to transform these embeddings into an actual prediction.  You can see these embeddings here.%
\index{layer}%
\index{neuron}%
\index{output}%
\index{output neuron}%
\index{predict}%
\index{training}%
\index{vector}%
\par%
\begin{tcolorbox}[size=title,title=Code,breakable]%
\begin{lstlisting}[language=Python, upquote=true]
print(embedding_layer.get_weights()[0].shape)
print(embedding_layer.get_weights())\end{lstlisting}
\tcbsubtitle[before skip=\baselineskip]{Output}%
\begin{lstlisting}[upquote=true]
(50, 8)
[array([[-0.11389559, -0.04778124,  0.10034387,  0.12887037,
0.05670259,
        -0.09982903, -0.15423775, -0.06774805],
       [-0.04839246,  0.00527745,  0.0084306 , -0.03498586,  0.010772
,
         0.04015711,  0.03564452, -0.00849336],
       [-0.11003157, -0.05829103,  0.12370535, -0.07124459, -0.0667479
,
        -0.14339209, -0.13791779, -0.13947721],
       [-0.15395765, -0.08560142, -0.15915371, -0.0882007 ,
0.15756004,
        -0.10337664, -0.12412377, -0.10282961],
       [ 0.04919637, -0.00870635, -0.02393281,  0.04445953,  0.0124351
,

...

0.04153964,
        -0.04445877, -0.00612149, -0.03430663],
       [-0.08493928, -0.10910758,  0.0605178 , -0.10072854,
-0.11677803,
        -0.05648913, -0.13342443, -0.08516318]], dtype=float32)]
\end{lstlisting}
\end{tcolorbox}%
We can now evaluate this neural network's accuracy, including the embeddings and the learned dense layer.%
\index{dense layer}%
\index{layer}%
\index{neural network}%
\par%
\begin{tcolorbox}[size=title,title=Code,breakable]%
\begin{lstlisting}[language=Python, upquote=true]
loss, accuracy = model.evaluate(padded_reviews, labels, verbose=0)
print(f'Accuracy: {accuracy}')\end{lstlisting}
\tcbsubtitle[before skip=\baselineskip]{Output}%
\begin{lstlisting}[upquote=true]
Accuracy: 1.0
\end{lstlisting}
\end{tcolorbox}%
The accuracy is a perfect 1.0, indicating there is likely overfitting. It would be good to use early stopping to not overfit for a more complex data set.%
\index{early stopping}%
\index{overfitting}%
\par%
\begin{tcolorbox}[size=title,title=Code,breakable]%
\begin{lstlisting}[language=Python, upquote=true]
print(f'Log-loss: {loss}')\end{lstlisting}
\tcbsubtitle[before skip=\baselineskip]{Output}%
\begin{lstlisting}[upquote=true]
Log-loss: 0.48446863889694214
\end{lstlisting}
\end{tcolorbox}%
However, the loss is not perfect. Even though the predicted probabilities indicated a correct prediction in every case, the program did not achieve absolute confidence in each correct answer. The lack of confidence was likely due to the small amount of noise (previously discussed) in the data set. Some words that appeared in both positive and negative reviews contributed to this lack of absolute certainty.%
\index{predict}%
\index{SOM}%
\par

\chapter{Reinforcement Learning}%
\label{chap:ReinforcementLearning}%
\section{Part 12.1: Introduction to the OpenAI Gym}%
\label{sec:Part12.1IntroductiontotheOpenAIGym}%
\href{https://gym.openai.com/}{OpenAI Gym }%
aims to provide an easy{-}to{-}setup general{-}intelligence benchmark with various environments. The goal is to standardize how environments are defined in AI research publications to make published research more easily reproducible. The project claims to provide the user with a simple interface. As of June 2017, developers can only use Gym with Python.%
\index{gym}%
\index{Python}%
\par%
OpenAI gym is pip{-}installed onto your local machine. There are a few significant limitations to be aware of:%
\index{gym}%
\index{OpenAI}%
\par%
\begin{itemize}[noitemsep]%
\item%
OpenAI Gym Atari only%
\index{gym}%
\index{OpenAI}%
\textbf{ directly }%
supports Linux and Macintosh%
\item%
OpenAI Gym Atari can be used with Windows; however, it requires a particular%
\index{gym}%
\index{OpenAI}%
\href{https://towardsdatascience.com/how-to-install-openai-gym-in-a-windows-environment-338969e24d30}{ installation procedure}%
\item%
OpenAI Gym can not directly render animated games in Google CoLab.%
\index{gym}%
\index{OpenAI}%
\end{itemize}%
Because OpenAI Gym requires a graphics display, an embedded video is the only way to display Gym in Google CoLab. The presentation of OpenAI Gym game animations in Google CoLab is discussed later in this module.%
\index{gym}%
\index{OpenAI}%
\index{video}%
\par%
\subsection{OpenAI Gym Leaderboard}%
\label{subsec:OpenAIGymLeaderboard}%
The OpenAI Gym does have a leaderboard, similar to Kaggle; however, the OpenAI Gym's leaderboard is much more informal compared to Kaggle. The user's local machine performs all scoring. As a result, the OpenAI gym's leaderboard is strictly an "honor system."  The leaderboard is maintained in the following GitHub repository:%
\index{GitHub}%
\index{gym}%
\index{Kaggle}%
\index{OpenAI}%
\par%
\begin{itemize}[noitemsep]%
\item%
\href{https://github.com/openai/gym/wiki/Leaderboard}{OpenAI Gym Leaderboard}%
\end{itemize}%
You must provide a write{-}up with sufficient instructions to reproduce your result if you submit a score. A video of your results is suggested but not required.%
\index{video}%
\par

\subsection{Looking at Gym Environments}%
\label{subsec:LookingatGymEnvironments}%
The centerpiece of Gym is the environment, which defines the "game" in which your reinforcement algorithm will compete. An environment does not need to be a game; however, it describes the following game{-}like features:%
\index{algorithm}%
\index{feature}%
\index{gym}%
\par%
\begin{itemize}[noitemsep]%
\item%
\textbf{action space}%
: What actions can we take on the environment at each step/episode to alter the environment.%
\item%
\textbf{observation space}%
: What is the current state of the portion of the environment that we can observe. Usually, we can see the entire environment.%
\end{itemize}%
Before we begin to look at Gym, it is essential to understand some of the terminology used by this library.%
\index{gym}%
\index{SOM}%
\par%
\begin{itemize}[noitemsep]%
\item%
\textbf{Agent }%
{-} The machine learning program or model that controls the actions.%
\index{learning}%
\index{model}%
\end{itemize}%
Step {-} One round of issuing actions that affect the observation space.%
\par%
\begin{itemize}[noitemsep]%
\item%
\textbf{Episode }%
{-} A collection of steps that terminates when the agent fails to meet the environment's objective or the episode reaches the maximum number of allowed steps.%
\item%
\textbf{Render }%
{-} Gym can render one frame for display after each episode.%
\index{gym}%
\item%
\textbf{Reward }%
{-} A positive reinforcement that can occur at the end of each episode, after the agent acts.%
\item%
\textbf{Non{-}deterministic }%
{-} For some environments, randomness is a factor in deciding what effects actions have on reward and changes to the observation space.%
\index{random}%
\index{SOM}%
\end{itemize}%
It is important to note that many gym environments specify that they are not non{-}deterministic even though they use random numbers to process actions. Based on the gym GitHub issue tracker, a non{-}deterministic property means a deterministic environment behaves randomly. Even when you give the environment a consistent seed value, this behavior is confirmed. The program can use the seed method of an environment to seed the random number generator for the environment.%
\index{GitHub}%
\index{gym}%
\index{random}%
\index{ROC}%
\index{ROC}%
\par%
The Gym library allows us to query some of these attributes from environments. I created the following function to query gym environments.%
\index{gym}%
\index{SOM}%
\par%
\begin{tcolorbox}[size=title,title=Code,breakable]%
\begin{lstlisting}[language=Python, upquote=true]
import gym


def query_environment(name):
    env = gym.make(name)
    spec = gym.spec(name)
    print(f"Action Space: {env.action_space}")
    print(f"Observation Space: {env.observation_space}")
    print(f"Max Episode Steps: {spec.max_episode_steps}")
    print(f"Nondeterministic: {spec.nondeterministic}")
    print(f"Reward Range: {env.reward_range}")
    print(f"Reward Threshold: {spec.reward_threshold}")\end{lstlisting}
\end{tcolorbox}%
We will look at the%
\textbf{ MountainCar{-}v0 }%
environment, which challenges an underpowered car to escape the valley between two mountains.  The following code describes the Mountian Car environment.%
\par%
\begin{tcolorbox}[size=title,title=Code,breakable]%
\begin{lstlisting}[language=Python, upquote=true]
query_environment("MountainCar-v0")\end{lstlisting}
\tcbsubtitle[before skip=\baselineskip]{Output}%
\begin{lstlisting}[upquote=true]
Action Space: Discrete(3)
Observation Space: Box(-1.2000000476837158, 0.6000000238418579, (2,),
float32)
Max Episode Steps: 200
Nondeterministic: False
Reward Range: (-inf, inf)
Reward Threshold: -110.0
\end{lstlisting}
\end{tcolorbox}%
This environment allows three distinct actions: accelerate forward, decelerate, or backward. The observation space contains two continuous (floating point) values, as evident by the box object. The observation space is simply the position and velocity of the car. The car has 200 steps to escape for each episode. You would have to look at the code, but the mountain car receives no incremental reward. The only reward for the vehicle occurs when it escapes the valley.%
\index{continuous}%
\par%
\begin{tcolorbox}[size=title,title=Code,breakable]%
\begin{lstlisting}[language=Python, upquote=true]
query_environment("CartPole-v1")\end{lstlisting}
\tcbsubtitle[before skip=\baselineskip]{Output}%
\begin{lstlisting}[upquote=true]
Action Space: Discrete(2)
Observation Space: Box(-3.4028234663852886e+38,
3.4028234663852886e+38, (4,), float32)
Max Episode Steps: 500
Nondeterministic: False
Reward Range: (-inf, inf)
Reward Threshold: 475.0
\end{lstlisting}
\end{tcolorbox}%
The%
\textbf{ CartPole{-}v1 }%
environment challenges the agent to balance a pole while the agent. The environment has an observation space of 4 continuous numbers:%
\index{continuous}%
\par%
\begin{itemize}[noitemsep]%
\item%
Cart Position%
\item%
Cart Velocity%
\item%
Pole Angle%
\item%
Pole Velocity At Tip%
\end{itemize}%
To achieve this goal, the agent can take the following actions:%
\par%
\begin{itemize}[noitemsep]%
\item%
Push cart to the left%
\item%
Push cart to the right%
\end{itemize}%
There is also a continuous variant of the mountain car. This version does not simply have the motor on or off. The action space is a single floating{-}point number for the continuous cart that specifies how much forward or backward force the cart currently utilizes.%
\index{continuous}%
\par%
\begin{tcolorbox}[size=title,title=Code,breakable]%
\begin{lstlisting}[language=Python, upquote=true]
query_environment("MountainCarContinuous-v0")\end{lstlisting}
\tcbsubtitle[before skip=\baselineskip]{Output}%
\begin{lstlisting}[upquote=true]
Action Space: Box(-1.0, 1.0, (1,), float32)
Observation Space: Box(-1.2000000476837158, 0.6000000238418579, (2,),
float32)
Max Episode Steps: 999
Nondeterministic: False
Reward Range: (-inf, inf)
Reward Threshold: 90.0
\end{lstlisting}
\end{tcolorbox}%
Note: If you see a warning above, you can safely ignore it; it is a relatively minor bug in OpenAI Gym.%
\index{gym}%
\index{OpenAI}%
\par%
Atari games, like breakout, can use an observation space that is either equal to the size of the Atari screen (210x160) or even use the RAM of the Atari (128 bytes) to determine the state of the game.  Yes, that's bytes, not kilobytes!%
\par%
\begin{tcolorbox}[size=title,title=Code,breakable]%
\begin{lstlisting}[language=Python, upquote=true]
!wget http://www.atarimania.com/roms/Roms.rar 
!unrar x -o+ /content/Roms.rar >/dev/nul
!python -m atari_py.import_roms /content/ROMS >/dev/nul\end{lstlisting}
\end{tcolorbox}%
\begin{tcolorbox}[size=title,title=Code,breakable]%
\begin{lstlisting}[language=Python, upquote=true]
query_environment("Breakout-v0")\end{lstlisting}
\tcbsubtitle[before skip=\baselineskip]{Output}%
\begin{lstlisting}[upquote=true]
Action Space: Discrete(4)
Observation Space: Box(0, 255, (210, 160, 3), uint8)
Max Episode Steps: 10000
Nondeterministic: False
Reward Range: (-inf, inf)
Reward Threshold: None
\end{lstlisting}
\end{tcolorbox}%
\begin{tcolorbox}[size=title,title=Code,breakable]%
\begin{lstlisting}[language=Python, upquote=true]
query_environment("Breakout-ram-v0")\end{lstlisting}
\tcbsubtitle[before skip=\baselineskip]{Output}%
\begin{lstlisting}[upquote=true]
Action Space: Discrete(4)
Observation Space: Box(0, 255, (128,), uint8)
Max Episode Steps: 10000
Nondeterministic: False
Reward Range: (-inf, inf)
Reward Threshold: None
\end{lstlisting}
\end{tcolorbox}

\subsection{Render OpenAI Gym Environments from CoLab}%
\label{subsec:RenderOpenAIGymEnvironmentsfromCoLab}%
It is possible to visualize the game your agent is playing, even on CoLab. This section provides information on generating a video in CoLab that shows you an episode of the game your agent is playing. I based this video process on suggestions found%
\index{ROC}%
\index{ROC}%
\index{video}%
\href{https://colab.research.google.com/drive/1flu31ulJlgiRL1dnN2ir8wGh9p7Zij2t}{ here}%
.%
\par%
Begin by installing%
\textbf{ pyvirtualdisplay }%
and%
\textbf{ python{-}opengl}%
.%
\par%
\begin{tcolorbox}[size=title,title=Code,breakable]%
\begin{lstlisting}[language=Python, upquote=true]
!pip install gym pyvirtualdisplay > /dev/null 2>&1
!apt-get install -y xvfb python-opengl ffmpeg > /dev/null 2>&1\end{lstlisting}
\end{tcolorbox}%
Next, we install the needed requirements to display an Atari game.%
\par%
\begin{tcolorbox}[size=title,title=Code,breakable]%
\begin{lstlisting}[language=Python, upquote=true]
!apt-get update > /dev/null 2>&1
!apt-get install cmake > /dev/null 2>&1
!pip install --upgrade setuptools 2>&1
!pip install ez_setup > /dev/null 2>&1
!pip install gym[atari] > /dev/null 2>&1\end{lstlisting}
\end{tcolorbox}%
Next, we define the functions used to show the video by adding it to the CoLab notebook.%
\index{video}%
\par%
\begin{tcolorbox}[size=title,title=Code,breakable]%
\begin{lstlisting}[language=Python, upquote=true]
import gym
from gym.wrappers import Monitor
import glob
import io
import base64
from IPython.display import HTML
from pyvirtualdisplay import Display
from IPython import display as ipythondisplay

display = Display(visible=0, size=(1400, 900))
display.start()

"""
Utility functions to enable video recording of gym environment 
and displaying it.
To enable video, just do "env = wrap_env(env)""
"""


def show_video():
    mp4list = glob.glob('video/*.mp4')
    if len(mp4list) > 0:
        mp4 = mp4list[0]
        video = io.open(mp4, 'r+b').read()
        encoded = base64.b64encode(video)
        ipythondisplay.display(HTML(data='''<video alt="test" autoplay 
                loop controls style="height: 400px;">
                <source src="data:video/mp4;base64,{0}" type="video/mp4" />
             </video>'''.format(encoded.decode('ascii'))))
    else:
        print("Could not find video")


def wrap_env(env):
    env = Monitor(env, './video', force=True)
    return env\end{lstlisting}
\end{tcolorbox}%
Now we are ready to play the game.  We use a simple random agent.%
\index{random}%
\par%
\begin{tcolorbox}[size=title,title=Code,breakable]%
\begin{lstlisting}[language=Python, upquote=true]
#env = wrap_env(gym.make("MountainCar-v0"))
env = wrap_env(gym.make("Atlantis-v0"))

observation = env.reset()

while True:

    env.render()

    # your agent goes here
    action = env.action_space.sample()

    observation, reward, done, info = env.step(action)

    if done:
        break

env.close()
show_video()\end{lstlisting}
\end{tcolorbox}

\section{Part 12.2: Introduction to Q{-}Learning}%
\label{sec:Part12.2IntroductiontoQ{-}Learning}%
Q{-}Learning is a foundational technology upon which deep reinforcement learning is based. Before we explore deep reinforcement learning, it is essential to understand Q{-}Learning. Several components make up any Q{-}Learning system.%
\index{learning}%
\index{reinforcement learning}%
\par%
\begin{itemize}[noitemsep]%
\item%
\textbf{Agent }%
{-} The agent is an entity that exists in an environment that takes actions to affect the state of the environment, to receive rewards.%
\item%
\textbf{Environment }%
{-} The environment is the universe that the agent exists in. The environment is always in a specific state that is changed by the agent's actions.%
\item%
\textbf{Actions }%
{-} Steps that the agent can perform to alter the environment%
\item%
\textbf{Step }%
{-} A step occurs when the agent performs an action and potentially changes the environment state.%
\item%
\textbf{Episode }%
{-} A chain of steps that ultimately culminates in the environment entering a terminal state.%
\item%
\textbf{Epoch }%
{-} A training iteration of the agent that contains some number of episodes.%
\index{iteration}%
\index{SOM}%
\index{training}%
\item%
\textbf{Terminal State }%
{-}  A state in which further actions do not make sense. A terminal state occurs when the agent has one, lost, or the environment exceeds the maximum number of steps in many environments.%
\end{itemize}%
Q{-}Learning works by building a table that suggests an action for every possible state. This approach runs into several problems. First, the environment is usually composed of several continuous numbers, resulting in an infinite number of states. Q{-}Learning handles continuous states by binning these numeric values into ranges.%
\index{continuous}%
\index{learning}%
\par%
Out of the box, Q{-}Learning does not deal with continuous inputs, such as a car's accelerator that can range from released to fully engaged. Additionally, Q{-}Learning primarily deals with discrete actions, such as pressing a joystick up or down. Researchers have developed clever tricks to allow Q{-}Learning to accommodate continuous actions.%
\index{continuous}%
\index{input}%
\index{learning}%
\par%
Deep neural networks can help solve the problems of continuous environments and action spaces. In the next section, we will learn more about deep reinforcement learning. For now, we will apply regular Q{-}Learning to the Mountain Car problem from OpenAI Gym.%
\index{continuous}%
\index{gym}%
\index{learning}%
\index{neural network}%
\index{OpenAI}%
\index{reinforcement learning}%
\par%
\subsection{Introducing the Mountain Car}%
\label{subsec:IntroducingtheMountainCar}%
This section will demonstrate how Q{-}Learning can create a solution to the mountain car gym environment. The Mountain car is an environment where a car must climb a mountain. Because gravity is stronger than the car's engine, it cannot merely accelerate up the steep slope even with full throttle. The vehicle is situated in a valley and must learn to utilize potential energy by driving up the opposite hill before the car can make it to the goal at the top of the rightmost hill.%
\index{gym}%
\index{learning}%
\index{slope}%
\par%
First, it might be helpful to visualize the mountain car environment. The following code shows this environment. This code makes use of TF{-}Agents to perform this render. Usually, we use TF{-}Agents for the type of deep reinforcement learning that we will see in the next module. However, TF{-}Agents is just used to render the mountain care environment for now.%
\index{learning}%
\index{reinforcement learning}%
\par%
\begin{tcolorbox}[size=title,title=Code,breakable]%
\begin{lstlisting}[language=Python, upquote=true]
import tf_agents
from tf_agents.environments import suite_gym
import PIL.Image
import pyvirtualdisplay

display = pyvirtualdisplay.Display(visible=0, size=(1400, 900)).start()

env_name = 'MountainCar-v0'
env = suite_gym.load(env_name)
env.reset()
PIL.Image.fromarray(env.render())\end{lstlisting}
\tcbsubtitle[before skip=\baselineskip]{Output}%
\includegraphics[width=4in]{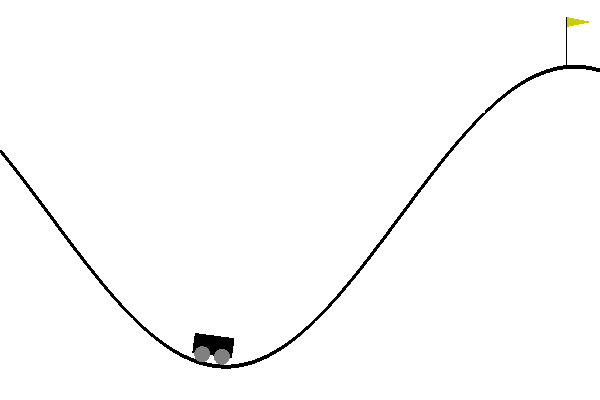}%
\end{tcolorbox}%
The mountain car environment provides the following discrete actions:%
\par%
\begin{itemize}[noitemsep]%
\item%
0 {-} Apply left force%
\item%
1 {-} Apply no force%
\item%
2 {-} Apply right force%
\end{itemize}%
The mountain car environment is made up of the following continuous values:%
\index{continuous}%
\par%
\begin{itemize}[noitemsep]%
\item%
state{[}0{]} {-} Position%
\item%
state{[}1{]} {-} Velocity%
\end{itemize}%
The cart is not strong enough. It will need to use potential energy from the mountain behind it. The following code shows an agent that applies full throttle to climb the hill.%
\par%
\begin{tcolorbox}[size=title,title=Code,breakable]%
\begin{lstlisting}[language=Python, upquote=true]
import gym
from gym.wrappers import Monitor
import glob
import io
import base64
from IPython.display import HTML
from pyvirtualdisplay import Display
from IPython import display as ipythondisplay

display = Display(visible=0, size=(1400, 900))
display.start()


def show_video():
    mp4list = glob.glob('video/*.mp4')
    if len(mp4list) > 0:
        mp4 = mp4list[0]
        video = io.open(mp4, 'r+b').read()
        encoded = base64.b64encode(video)
        ipythondisplay.display(HTML(data='''<video alt="test" autoplay 
                loop controls style="height: 400px;">
                <source src="data:video/mp4;base64,{0}" 
                type="video/mp4" />
             </video>'''.format(encoded.decode('ascii'))))
    else:
        print("Could not find video")


def wrap_env(env):
    env = Monitor(env, './video', force=True)
    return env\end{lstlisting}
\end{tcolorbox}%
We are now ready to train the agent.%
\par%
\begin{tcolorbox}[size=title,title=Code,breakable]%
\begin{lstlisting}[language=Python, upquote=true]
import gym

if COLAB:
    env = wrap_env(gym.make("MountainCar-v0"))
else:
    env = gym.make("MountainCar-v0")

env.reset()
done = False

i = 0
while not done:
    i += 1
    state, reward, done, _ = env.step(2)
    env.render()
    print(f"Step {i}: State={state}, Reward={reward}")

env.close()\end{lstlisting}
\tcbsubtitle[before skip=\baselineskip]{Output}%
\begin{lstlisting}[upquote=true]
Step 1: State=[-0.50905189  0.00089766], Reward=-1.0
Step 2: State=[-0.50726329  0.00178859], Reward=-1.0
Step 3: State=[-0.50459717  0.00266613], Reward=-1.0
Step 4: State=[-0.50107348  0.00352369], Reward=-1.0
Step 5: State=[-0.4967186   0.00435488], Reward=-1.0
Step 6: State=[-0.4915651  0.0051535], Reward=-1.0
Step 7: State=[-0.48565149  0.00591361], Reward=-1.0
Step 8: State=[-0.47902187  0.00662962], Reward=-1.0
Step 9: State=[-0.47172557  0.00729629], Reward=-1.0
Step 10: State=[-0.46381676  0.00790881], Reward=-1.0
Step 11: State=[-0.45535392  0.00846285], Reward=-1.0
Step 12: State=[-0.44639934  0.00895458], Reward=-1.0
Step 13: State=[-0.4370186   0.00938074], Reward=-1.0
Step 14: State=[-0.42727993  0.00973867], Reward=-1.0
Step 15: State=[-0.41725364  0.01002629], Reward=-1.0

...

Step 196: State=[-0.26463414 -0.00336818], Reward=-1.0
Step 197: State=[-0.26875498 -0.00412085], Reward=-1.0
Step 198: State=[-0.27360632 -0.00485134], Reward=-1.0
Step 199: State=[-0.27916172 -0.0055554 ], Reward=-1.0
Step 200: State=[-0.28539045 -0.00622873], Reward=-1.0
\end{lstlisting}
\end{tcolorbox}%
It helps to visualize the car. The following code shows a video of the car when run from a notebook.%
\index{video}%
\par%
\begin{tcolorbox}[size=title,title=Code,breakable]%
\begin{lstlisting}[language=Python, upquote=true]
show_video()\end{lstlisting}
\end{tcolorbox}

\subsection{Programmed Car}%
\label{subsec:ProgrammedCar}%
Now we will look at a car that I hand{-}programmed. This car is straightforward; however, it solves the problem. The programmed car always applies force in one direction or another. It does not break. Whatever direction the vehicle is currently rolling, the agent uses power in that direction. Therefore, the car begins to climb a hill, is overpowered, and turns backward. However, once it starts to roll backward, force is immediately applied in this new direction.%
\par%
The following code implements this preprogrammed car.%
\par%
\begin{tcolorbox}[size=title,title=Code,breakable]%
\begin{lstlisting}[language=Python, upquote=true]
import gym

if COLAB:
    env = wrap_env(gym.make("MountainCar-v0"))
else:
    env = gym.make("MountainCar-v0")

state = env.reset()
done = False

i = 0
while not done:
    i += 1

    if state[1] > 0:
        action = 2
    else:
        action = 0

    state, reward, done, _ = env.step(action)
    env.render()
    print(f"Step {i}: State={state}, Reward={reward}")

env.close()\end{lstlisting}
\tcbsubtitle[before skip=\baselineskip]{Output}%
\begin{lstlisting}[upquote=true]
Step 1: State=[-5.84581471e-01 -5.49227966e-04], Reward=-1.0
Step 2: State=[-0.58567588 -0.0010944 ], Reward=-1.0
Step 3: State=[-0.58730739 -0.00163151], Reward=-1.0
Step 4: State=[-0.58946399 -0.0021566 ], Reward=-1.0
Step 5: State=[-0.59212981 -0.00266582], Reward=-1.0
Step 6: State=[-0.59528526 -0.00315545], Reward=-1.0
Step 7: State=[-0.5989072  -0.00362194], Reward=-1.0
Step 8: State=[-0.60296912 -0.00406192], Reward=-1.0
Step 9: State=[-0.60744137 -0.00447225], Reward=-1.0
Step 10: State=[-0.61229141 -0.00485004], Reward=-1.0
Step 11: State=[-0.61748407 -0.00519267], Reward=-1.0
Step 12: State=[-0.62298187 -0.0054978 ], Reward=-1.0
Step 13: State=[-0.62874529 -0.00576342], Reward=-1.0
Step 14: State=[-0.63473313 -0.00598783], Reward=-1.0
Step 15: State=[-0.64090281 -0.00616968], Reward=-1.0

...

Step 149: State=[0.30975487 0.04947665], Reward=-1.0
Step 150: State=[0.35873547 0.0489806 ], Reward=-1.0
Step 151: State=[0.40752939 0.04879392], Reward=-1.0
Step 152: State=[0.45647027 0.04894088], Reward=-1.0
Step 153: State=[0.50591109 0.04944082], Reward=-1.0
\end{lstlisting}
\end{tcolorbox}%
We now visualize the preprogrammed car solving the problem.%
\par%
\begin{tcolorbox}[size=title,title=Code,breakable]%
\begin{lstlisting}[language=Python, upquote=true]
show_video()\end{lstlisting}
\end{tcolorbox}

\subsection{Reinforcement Learning}%
\label{subsec:ReinforcementLearning}%
Q{-}Learning is a system of rewards that the algorithm gives an agent for successfully moving the environment into a state considered successful. These rewards are the Q{-}values from which this algorithm takes its name. The final output from the Q{-}Learning algorithm is a table of Q{-}values that indicate the reward value of every action that the agent can take, given every possible environment state. The agent must bin continuous state values into a fixed finite number of columns.%
\index{algorithm}%
\index{continuous}%
\index{learning}%
\index{output}%
\par
\begin{figure}[h]%
\centering%
\includegraphics[width=4in]{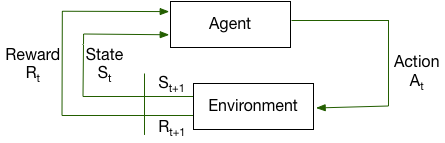}%
\caption{Reinforcement/Q Learning}%
\label{12.REINF}%
\end{figure}

%
Learning occurs when the algorithm runs the agent and environment through episodes and updates the Q{-}values based on the rewards received from actions taken; Figure \ref{12.REINF} provides a high{-}level overview of this reinforcement or Q{-}Learning loop.%
\index{algorithm}%
\index{learning}%
\par%

\par%
The Q{-}values can dictate action by selecting the action column with the highest Q{-}value for the current environment state. The choice between choosing a random action and a Q{-}value{-}driven action is governed by the epsilon ($\epsilon$) parameter, the probability of random action.%
\index{parameter}%
\index{probability}%
\index{random}%
\par%
Each time through the training loop, the training algorithm updates the Q{-}values according to the following equation.%
\index{algorithm}%
\index{training}%
\index{training algorithm}%
\par%
$Q^{new}(s_{t},a_{t}) \leftarrow \underbrace{Q(s_{t},a_{t})}_{\text{old value}} + \underbrace{\alpha}_{\text{learning rate}} \cdot  \overbrace{\bigg( \underbrace{\underbrace{r_{t}}_{\text{reward}} + \underbrace{\gamma}_{\text{discount factor}} \cdot \underbrace{\max_{a}Q(s_{t+1}, a)}_{\text{estimate of optimal future value}}}_{\text{new value (temporal difference target)}} - \underbrace{Q(s_{t},a_{t})}_{\text{old value}} \bigg) }^{\text{temporal difference}}$\newline%
\newline%
There are several parameters in this equation:\newline%
* alpha ($\alpha$) {-} The learning rate, how much should the current step cause the Q{-}values to be updated.\newline%
* lambda ($\lambda$) {-} The discount factor is the percentage of future reward that the algorithm should consider in this update.\newline%
\newline%
This equation modifies several values:\newline%
\newline%
* $Q(s_t,a_t)$ {-} The Q{-}table.  For each combination of states, what reward would the agent likely receive for performing each action?\newline%
* $s_t$ {-} The current state.\newline%
* $r_t$ {-} The last reward received.\newline%
* $a_t$ {-} The action that the agent will perform.\newline%
\newline%
The equation works by calculating a delta (temporal difference) that the equation should apply to the old state. This learning rate ($\alpha$) scales this delta. A learning rate of 1.0 would fully implement the temporal difference in the Q{-}values each iteration and would likely be very chaotic.\newline%
\newline%
There are two parts to the temporal difference: the new and old values. The new value is subtracted from the old value to provide a delta; the full amount we would change the Q{-}value by if the learning rate did not scale this value. The new value is a summation of the reward received from the last action and the maximum Q{-}values from the resulting state when the client takes this action. Adding the maximum of action Q{-}values for the new state is essential because it estimates the optimal future values from proceeding with this action. \newline%
\newline%
\#\# Q{-}Learning Car\newline%
\newline%
We will now use Q{-}Learning to produce a car that learns to drive itself. Look out, Tesla! We begin by defining two essential functions.%
\index{algorithm}%
\index{delta}%
\index{iteration}%
\index{learning}%
\index{learning rate}%
\index{parameter}%
\index{Q{-}Table}%
\index{ROC}%
\index{ROC}%
\index{temporal}%
\begin{tcolorbox}[size=title,title=Code,breakable]%
\begin{lstlisting}[language=Python, upquote=true]
import gym
import numpy as np

# This function converts the floating point state values into
# discrete values. This is often called binning.  We divide
# the range that the state values might occupy and assign
# each region to a bucket.
def calc_discrete_state(state):
    discrete_state = (state - env.observation_space.low)/buckets
    return tuple(discrete_state.astype(int))

# Run one game.  The q_table to use is provided.  We also
# provide a flag to indicate if the game should be
# rendered/animated.  Finally, we also provide
# a flag to indicate if the q_table should be updated.
def run_game(q_table, render, should_update):
    done = False
    discrete_state = calc_discrete_state(env.reset())
    success = False

    while not done:
        # Exploit or explore
        if np.random.random() > epsilon:
            # Exploit - use q-table to take current best action
            # (and probably refine)
            action = np.argmax(q_table[discrete_state])
        else:
            # Explore - t
            action = np.random.randint(0, env.action_space.n)

        # Run simulation step
        new_state, reward, done, _ = env.step(action)

        # Convert continuous state to discrete
        new_state_disc = calc_discrete_state(new_state)

        # Have we reached the goal position (have we won?)?
        if new_state[0] >= env.unwrapped.goal_position:
            success = True

        # Update q-table
        if should_update:
            max_future_q = np.max(q_table[new_state_disc])
            current_q = q_table[discrete_state + (action,)]
            new_q = (1 - LEARNING_RATE) * current_q + LEARNING_RATE * \
                (reward + DISCOUNT * max_future_q)
            q_table[discrete_state + (action,)] = new_q

        discrete_state = new_state_disc

        if render:
            env.render()

    return success\end{lstlisting}
\end{tcolorbox}%
Several hyperparameters are very important for Q{-}Learning. These parameters will likely need adjustment as you apply Q{-}Learning to other problems. Because of this, it is crucial to understand the role of each parameter.%
\index{hyperparameter}%
\index{learning}%
\index{parameter}%
\par%
\begin{itemize}[noitemsep]%
\item%
\textbf{LEARNING\_RATE }%
The rate at which previous Q{-}values are updated based on new episodes run during training.%
\index{training}%
\item%
\textbf{DISCOUNT }%
The amount of significance to give estimates of future rewards when added to the reward for the current action taken. A value of 0.95 would indicate a discount of 5\% on the future reward estimates.%
\item%
\textbf{EPISODES }%
The number of episodes to train over. Increase this for more complex problems; however, training time also increases.%
\index{training}%
\item%
\textbf{SHOW\_EVERY }%
How many episodes to allow to elapse before showing an update.%
\item%
\textbf{DISCRETE\_GRID\_SIZE }%
How many buckets to use when converting each continuous state variable. For example, {[}10, 10{]} indicates that the algorithm should use ten buckets for the first and second state variables.%
\index{algorithm}%
\index{continuous}%
\item%
\textbf{START\_EPSILON\_DECAYING }%
Epsilon is the probability that the agent will select a random action over what the Q{-}Table suggests. This value determines the starting probability of randomness.%
\index{probability}%
\index{Q{-}Table}%
\index{random}%
\item%
\textbf{END\_EPSILON\_DECAYING }%
How many episodes should elapse before epsilon goes to zero and no random actions are permitted. For example, EPISODES//10  means only the first 1/10th of the episodes might have random actions.%
\index{random}%
\end{itemize}%
\begin{tcolorbox}[size=title,title=Code,breakable]%
\begin{lstlisting}[language=Python, upquote=true]
LEARNING_RATE = 0.1
DISCOUNT = 0.95
EPISODES = 50000
SHOW_EVERY = 1000

DISCRETE_GRID_SIZE = [10, 10]
START_EPSILON_DECAYING = 0.5
END_EPSILON_DECAYING = EPISODES//10\end{lstlisting}
\end{tcolorbox}%
We can now make the environment.  If we are running in Google COLAB, we wrap the environment to be displayed inside the web browser.  Next, create the discrete buckets for state and build Q{-}table.%
\index{Q{-}Table}%
\par%
\begin{tcolorbox}[size=title,title=Code,breakable]%
\begin{lstlisting}[language=Python, upquote=true]
if COLAB:
    env = wrap_env(gym.make("MountainCar-v0"))
else:
    env = gym.make("MountainCar-v0")

epsilon = 1
epsilon_change = epsilon/(END_EPSILON_DECAYING - START_EPSILON_DECAYING)
buckets = (env.observation_space.high - env.observation_space.low) \
    / DISCRETE_GRID_SIZE
q_table = np.random.uniform(low=-3, high=0, size=(DISCRETE_GRID_SIZE
                                                  + [env.action_space.n]))
success = False\end{lstlisting}
\end{tcolorbox}%
We can now make the environment.  If we are running in Google COLAB, we wrap the environment to be displayed inside the web browser.  Next, create the discrete buckets for state and build Q{-}table.%
\index{Q{-}Table}%
\par%
\begin{tcolorbox}[size=title,title=Code,breakable]%
\begin{lstlisting}[language=Python, upquote=true]
episode = 0
success_count = 0

# Loop through the required number of episodes
while episode < EPISODES:
    episode += 1
    done = False

    # Run the game.  If we are local, display render animation
    # at SHOW_EVERY intervals.
    if episode % SHOW_EVERY == 0:
        print(f"Current episode: {episode}, success: {success_count}" +
              f" {(float(success_count)/SHOW_EVERY)}")
        success = run_game(q_table, True, False)
        success_count = 0
    else:
        success = run_game(q_table, False, True)

    # Count successes
    if success:
        success_count += 1

    # Move epsilon towards its ending value, if it still needs to move
    if END_EPSILON_DECAYING >= episode >= START_EPSILON_DECAYING:
        epsilon = max(0, epsilon - epsilon_change)

print(success)\end{lstlisting}
\tcbsubtitle[before skip=\baselineskip]{Output}%
\begin{lstlisting}[upquote=true]
Current episode: 1000, success: 0 0.0
Current episode: 2000, success: 0 0.0
Current episode: 3000, success: 0 0.0
Current episode: 4000, success: 31 0.031
Current episode: 5000, success: 321 0.321
Current episode: 6000, success: 602 0.602
Current episode: 7000, success: 620 0.62
Current episode: 8000, success: 821 0.821
Current episode: 9000, success: 707 0.707
Current episode: 10000, success: 714 0.714
Current episode: 11000, success: 574 0.574
Current episode: 12000, success: 443 0.443
Current episode: 13000, success: 480 0.48
Current episode: 14000, success: 458 0.458
Current episode: 15000, success: 327 0.327

...

Current episode: 47000, success: 1000 1.0
Current episode: 48000, success: 1000 1.0
Current episode: 49000, success: 1000 1.0
Current episode: 50000, success: 1000 1.0
True
\end{lstlisting}
\end{tcolorbox}%
As you can see, the number of successful episodes generally increases as training progresses. It is not advisable to stop the first time we observe 100\% success over 1,000 episodes. There is a randomness to most games, so it is not likely that an agent would retain its 100\% success rate with a new run. It might be safe to stop training once you observe that the agent has gotten 100\% for several update intervals.%
\index{random}%
\index{training}%
\par

\subsection{Running and Observing the Agent}%
\label{subsec:RunningandObservingtheAgent}%
Now that the algorithm has trained the agent, we can observe the agent in action. You can use the following code to see the agent in action.%
\index{algorithm}%
\par%
\begin{tcolorbox}[size=title,title=Code,breakable]%
\begin{lstlisting}[language=Python, upquote=true]

run_game(q_table, True, False)
show_video()\end{lstlisting}
\end{tcolorbox}

\subsection{Inspecting the Q{-}Table}%
\label{subsec:InspectingtheQ{-}Table}%
We can also display the Q{-}table. The following code shows the agent's action for each environment state. As the weights of a neural network, this table is not straightforward to interpret. Some patterns do emerge in that direction, as seen by calculating the means of rows and columns. The actions seem consistent at both velocity and position's upper and lower halves.%
\index{neural network}%
\index{Q{-}Table}%
\index{SOM}%
\par%
\begin{tcolorbox}[size=title,title=Code,breakable]%
\begin{lstlisting}[language=Python, upquote=true]
import pandas as pd

df = pd.DataFrame(q_table.argmax(axis=2))

df.columns = [f'v-{x}' for x in range(DISCRETE_GRID_SIZE[0])]
df.index = [f'p-{x}' for x in range(DISCRETE_GRID_SIZE[1])]
df\end{lstlisting}
\tcbsubtitle[before skip=\baselineskip]{Output}%
\begin{tabular}[hbt!]{l|l|l|l|l|l|l|l|l|l|l}%
\hline%
&v{-}0&v{-}1&v{-}2&v{-}3&v{-}4&v{-}5&v{-}6&v{-}7&v{-}8&v{-}9\\%
\hline%
p{-}0&2&2&2&2&2&2&2&0&2&0\\%
p{-}1&0&1&0&1&2&2&2&2&2&1\\%
p{-}2&1&0&0&2&2&2&2&1&1&0\\%
p{-}3&2&0&0&0&2&2&2&1&2&2\\%
p{-}4&2&0&0&0&0&2&0&2&2&2\\%
p{-}5&1&1&2&1&1&0&1&1&2&2\\%
p{-}6&2&2&0&0&0&0&2&2&2&2\\%
p{-}7&0&2&1&0&0&1&2&2&2&2\\%
p{-}8&2&0&1&2&0&0&2&2&1&2\\%
p{-}9&2&2&2&1&1&0&2&2&2&1\\%
\hline%
\end{tabular}%
\vspace{2mm}%
\end{tcolorbox}%
\begin{tcolorbox}[size=title,title=Code,breakable]%
\begin{lstlisting}[language=Python, upquote=true]
df.mean(axis=0)\end{lstlisting}
\tcbsubtitle[before skip=\baselineskip]{Output}%
\begin{lstlisting}[upquote=true]
v-0    1.4
v-1    1.0
v-2    0.8
v-3    0.9
v-4    1.0
v-5    1.1
v-6    1.7
v-7    1.5
v-8    1.8
v-9    1.4
dtype: float64
\end{lstlisting}
\end{tcolorbox}%
\begin{tcolorbox}[size=title,title=Code,breakable]%
\begin{lstlisting}[language=Python, upquote=true]
df.mean(axis=1)\end{lstlisting}
\tcbsubtitle[before skip=\baselineskip]{Output}%
\begin{lstlisting}[upquote=true]
p-0    1.6
p-1    1.3
p-2    1.1
p-3    1.3
p-4    1.0
p-5    1.2
p-6    1.2
p-7    1.2
p-8    1.2
p-9    1.5
dtype: float64
\end{lstlisting}
\end{tcolorbox}

\section{Part 12.3: Keras Q{-}Learning in the OpenAI Gym}%
\label{sec:Part12.3KerasQ{-}LearningintheOpenAIGym}%
As we covered in the previous part, Q{-}Learning is a robust machine learning algorithm. Unfortunately, Q{-}Learning requires that the Q{-}table contain an entry for every possible state that the environment can take. Traditional Q{-}learning might be a good learning algorithm if the environment only includes a handful of discrete state elements. However, the Q{-}table can become prohibitively large if the state space is large.%
\index{algorithm}%
\index{learning}%
\index{Q{-}Table}%
\par
\begin{figure}[h]%
\centering%
\includegraphics[width=4in]{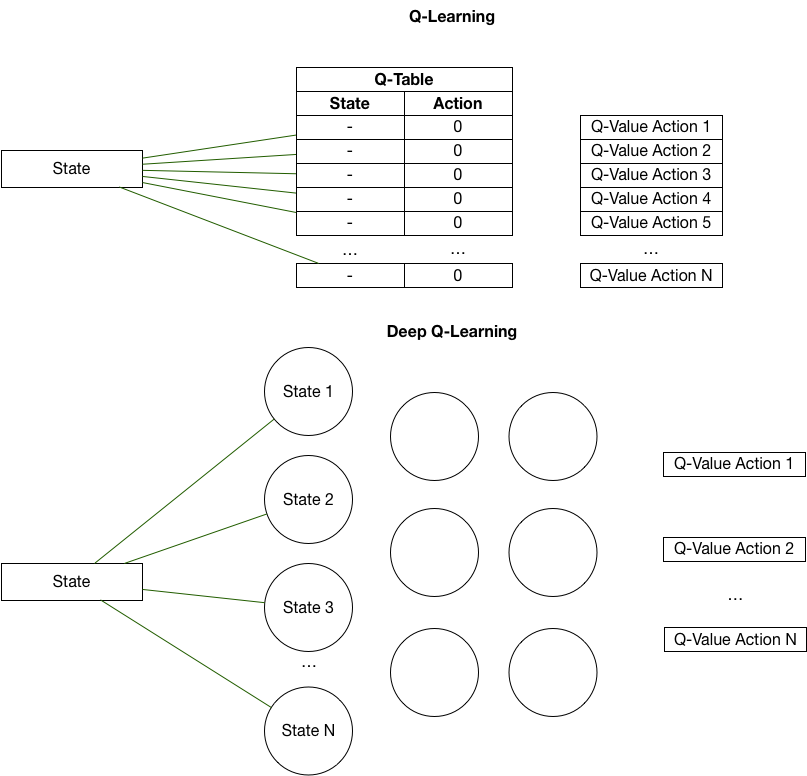}%
\caption{Deep Q{-}Learning (DQL)}%
\label{12.DQL}%
\end{figure}

%
Creating policies for large state spaces is a task that Deep Q{-}Learning Networks (DQN) can usually handle. Neural networks can generalize these states and learn commonalities. Unlike a table, a neural network does not require the program to represent every combination of state and action. A DQN maps the state to its input neurons and the action Q{-}values to the output neurons. The DQN effectively becomes a function that accepts the state and suggests action by returning the expected reward for each possible action. Figure \ref{12.DQL} demonstrates the DQN structure and mapping between state and action.%
\index{input}%
\index{input neuron}%
\index{learning}%
\index{neural network}%
\index{neuron}%
\index{output}%
\index{output neuron}%
\par%

\par%
As this diagram illustrates, the environment state contains several elements. For the basic DQN, the state can be a mix of continuous and categorical/discrete values. For the DQN, the discrete state elements the program typically encoded as dummy variables. The actions should be discrete when your program implements a DQN. Other algorithms support continuous outputs, which we will discuss later in this chapter.%
\index{algorithm}%
\index{categorical}%
\index{continuous}%
\index{output}%
\par%
This chapter will use%
\href{https://www.tensorflow.org/agents}{ TF{-}Agents }%
to implement a DQN to solve the cart{-}pole environment. TF{-}Agents makes designing, implementing, and testing new RL algorithms easier by providing well{-}tested modular components that can be modified and extended. It enables fast code iteration with functional test integration and benchmarking.%
\index{algorithm}%
\index{iteration}%
\par%
\subsection{DQN and the Cart{-}Pole Problem}%
\label{subsec:DQNandtheCart{-}PoleProblem}%
Barto (1983) first described the cart{-}pole problem.%
\cite{barto1983neuronlike}%
A cart is connected to a rigid hinged pole. The cart is free to move only in the vertical plane of the cart/track. The agent can apply an impulsive "left" or "right" force F of a fixed magnitude to the cart at discrete time intervals. The cart{-}pole environment simulates the physics behind keeping the pole reasonably upright position on the cart. The environment has four state variables:%
\par%
\begin{itemize}[noitemsep]%
\item%
$x$ The position of the cart on the track.%
\item%
$\theta$ The angle of the pole with the vertical%
\item%
$\dot{x}$ The cart velocity.%
\item%
$\dot{\theta}$ The rate of change of the angle.%
\end{itemize}%
The action space consists of discrete actions:%
\par%
\begin{itemize}[noitemsep]%
\item%
Apply force left%
\item%
Apply force right%
\end{itemize}%
To apply DQN to this problem, you need to create the following components for TF{-}Agents.%
\par%
\begin{itemize}[noitemsep]%
\item%
Environment%
\item%
Agent%
\item%
Policies%
\item%
Metrics and Evaluation%
\item%
Replay Buffer%
\item%
Data Collection%
\item%
Training%
\index{training}%
\end{itemize}%
These components are standard in most DQN implementations. Later, we will apply these same components to an Atari game, and after that, a problem with our design. This example is based on the%
\href{https://github.com/tensorflow/agents/blob/master/docs/tutorials/1_dqn_tutorial.ipynb}{ cart{-}pole tutorial }%
provided for TF{-}Agents.%
\par%
First, we must install TF{-}Agents.%
\par%
\begin{tcolorbox}[size=title,title=Code,breakable]%
\begin{lstlisting}[language=Python, upquote=true]
if COLAB:
  !sudo apt-get install -y xvfb ffmpeg x11-utils
  !pip install -q 'gym==0.10.11'
  !pip install -q 'imageio==2.4.0'
  !pip install -q PILLOW
  !pip install -q 'pyglet==1.3.2'
  !pip install -q pyvirtualdisplay
  !pip install -q tf-agents
  !pip install -q pygame\end{lstlisting}
\end{tcolorbox}%
We begin by importing needed Python libraries.%
\index{Python}%
\par%
\begin{tcolorbox}[size=title,title=Code,breakable]%
\begin{lstlisting}[language=Python, upquote=true]
import base64
import imageio
import IPython
import matplotlib
import matplotlib.pyplot as plt
import numpy as np
import PIL.Image
import pyvirtualdisplay

import tensorflow as tf

from tf_agents.agents.dqn import dqn_agent
from tf_agents.drivers import dynamic_step_driver
from tf_agents.environments import suite_gym
from tf_agents.environments import tf_py_environment
from tf_agents.eval import metric_utils
from tf_agents.metrics import tf_metrics
from tf_agents.networks import q_network
from tf_agents.policies import random_tf_policy
from tf_agents.replay_buffers import tf_uniform_replay_buffer
from tf_agents.trajectories import trajectory
from tf_agents.utils import common\end{lstlisting}
\end{tcolorbox}%
To allow this example to run in a notebook, we use a virtual display that will output an embedded video. If running this code outside a notebook, you could omit the virtual display and animate it directly to a window.%
\index{output}%
\index{video}%
\par%
\begin{tcolorbox}[size=title,title=Code,breakable]%
\begin{lstlisting}[language=Python, upquote=true]
# Set up a virtual display for rendering OpenAI gym environments.
display = pyvirtualdisplay.Display(visible=0, size=(1400, 900)).start()\end{lstlisting}
\end{tcolorbox}

\subsection{Hyperparameters}%
\label{subsec:Hyperparameters}%
We must define Several hyperparameters for the algorithm to train the agent.  The TF{-}Agent example provided reasonably well{-}tuned hyperparameters for cart{-}pole.  Later we will adapt these to an Atari game.%
\index{algorithm}%
\index{hyperparameter}%
\index{parameter}%
\par%
\begin{tcolorbox}[size=title,title=Code,breakable]%
\begin{lstlisting}[language=Python, upquote=true]
# How long should training run?
num_iterations = 20000
# How many initial random steps, before training start, to
# collect initial data.
initial_collect_steps = 1000
# How many steps should we run each iteration to collect
# data from.
collect_steps_per_iteration = 1
# How much data should we store for training examples.
replay_buffer_max_length = 100000

batch_size = 64
learning_rate = 1e-3
# How often should the program provide an update.
log_interval = 200

# How many episodes should the program use for each evaluation.
num_eval_episodes = 10
# How often should an evaluation occur.
eval_interval = 1000\end{lstlisting}
\end{tcolorbox}

\subsection{Environment}%
\label{subsec:Environment}%
TF{-}Agents use OpenAI gym environments to represent the task or problem to be solved. Standard environments can be created in TF{-}Agents using%
\index{gym}%
\index{OpenAI}%
\textbf{ tf\_agents.environments }%
suites. TF{-}Agents has suites for loading environments from sources such as the OpenAI Gym, Atari, and DM Control. We begin by loading the CartPole environment from the OpenAI Gym suite.%
\index{gym}%
\index{OpenAI}%
\par%
\begin{tcolorbox}[size=title,title=Code,breakable]%
\begin{lstlisting}[language=Python, upquote=true]
env_name = 'CartPole-v0'
env = suite_gym.load(env_name)\end{lstlisting}
\end{tcolorbox}%
We will quickly render this environment to see the visual representation.%
\par%
\begin{tcolorbox}[size=title,title=Code,breakable]%
\begin{lstlisting}[language=Python, upquote=true]
env.reset()
PIL.Image.fromarray(env.render())\end{lstlisting}
\tcbsubtitle[before skip=\baselineskip]{Output}%
\includegraphics[width=4in]{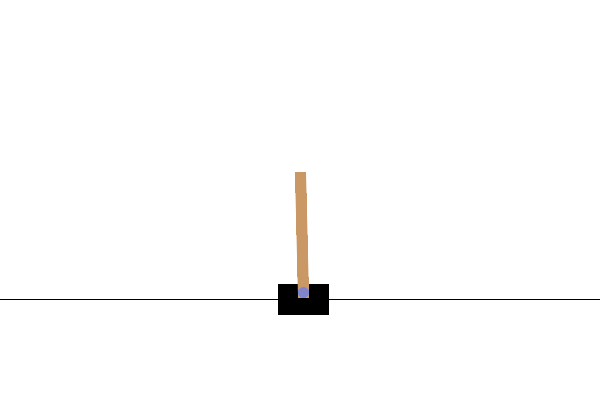}%
\end{tcolorbox}%
The%
\textbf{\texttt{ environment.step }}%
method takes an%
\textbf{\texttt{ action }}%
in the environment and returns a%
\textbf{\texttt{ TimeStep }}%
tuple containing the following observation of the environment and the reward for the action.%
\par%
The%
\textbf{\texttt{ time\_step\_spec() }}%
method returns the specification for the%
\textbf{\texttt{ TimeStep }}%
tuple. Its%
\textbf{\texttt{ observation }}%
attribute shows the shape of observations, the data types, and the ranges of allowed values. The%
\textbf{\texttt{ reward }}%
attribute shows the same details for the reward.%
\par%
\begin{tcolorbox}[size=title,title=Code,breakable]%
\begin{lstlisting}[language=Python, upquote=true]
print('Observation Spec:')
print(env.time_step_spec().observation)\end{lstlisting}
\tcbsubtitle[before skip=\baselineskip]{Output}%
\begin{lstlisting}[upquote=true]
Observation Spec:
BoundedArraySpec(shape=(4,), dtype=dtype('float32'),
name='observation', minimum=[-4.8000002e+00 -3.4028235e+38
-4.1887903e-01 -3.4028235e+38], maximum=[4.8000002e+00 3.4028235e+38
4.1887903e-01 3.4028235e+38])
\end{lstlisting}
\end{tcolorbox}%
\begin{tcolorbox}[size=title,title=Code,breakable]%
\begin{lstlisting}[language=Python, upquote=true]
print('Reward Spec:')
print(env.time_step_spec().reward)\end{lstlisting}
\tcbsubtitle[before skip=\baselineskip]{Output}%
\begin{lstlisting}[upquote=true]
Reward Spec:
ArraySpec(shape=(), dtype=dtype('float32'), name='reward')
\end{lstlisting}
\end{tcolorbox}%
The%
\textbf{\texttt{ action\_spec() }}%
method returns the shape, data types, and allowed values of valid actions.%
\par%
\begin{tcolorbox}[size=title,title=Code,breakable]%
\begin{lstlisting}[language=Python, upquote=true]
print('Action Spec:')
print(env.action_spec())\end{lstlisting}
\tcbsubtitle[before skip=\baselineskip]{Output}%
\begin{lstlisting}[upquote=true]
Action Spec:
BoundedArraySpec(shape=(), dtype=dtype('int64'), name='action',
minimum=0, maximum=1)
\end{lstlisting}
\end{tcolorbox}%
In the Cartpole environment:%
\par%
\begin{itemize}[noitemsep]%
\item%
\textbf{\texttt{observation }}%
is an array of 4 floats:%
\begin{itemize}[noitemsep]%
\item%
the position and velocity of the cart%
\item%
the angular position and velocity of the pole%
\item%
\item%
\end{itemize}%
\textbf{\texttt{reward }}%
is a scalar float value%
\textbf{\texttt{action }}%
is a scalar integer with only two possible values:%
\begin{itemize}[noitemsep]%
\item%
\textbf{\texttt{0 }}%
{-}{-}{-} "move left"%
\item%
\textbf{\texttt{1 }}%
{-}{-}{-} "move right"%
\end{itemize}%
\end{itemize}%
\begin{tcolorbox}[size=title,title=Code,breakable]%
\begin{lstlisting}[language=Python, upquote=true]
time_step = env.reset()
print('Time step:')
print(time_step)

action = np.array(1, dtype=np.int32)

next_time_step = env.step(action)
print('Next time step:')
print(next_time_step)\end{lstlisting}
\tcbsubtitle[before skip=\baselineskip]{Output}%
\begin{lstlisting}[upquote=true]
Time step:
TimeStep(
{'discount': array(1., dtype=float32),
 'observation': array([-0.03279859,  0.03562892, -0.04014493,
-0.04911802], dtype=float32),
 'reward': array(0., dtype=float32),
 'step_type': array(0, dtype=int32)})
Next time step:
TimeStep(
{'discount': array(1., dtype=float32),
 'observation': array([-0.03208601,  0.23130283, -0.04112729,
-0.35419184], dtype=float32),
 'reward': array(1., dtype=float32),
 'step_type': array(1, dtype=int32)})
\end{lstlisting}
\end{tcolorbox}%
Usually, the program instantiates two environments: one for training and one for evaluation.%
\index{training}%
\par%
\begin{tcolorbox}[size=title,title=Code,breakable]%
\begin{lstlisting}[language=Python, upquote=true]
train_py_env = suite_gym.load(env_name)
eval_py_env = suite_gym.load(env_name)\end{lstlisting}
\end{tcolorbox}%
The Cartpole environment, like most environments, is written in pure Python and is converted to TF{-}Agents and TensorFlow using the%
\index{Python}%
\index{TensorFlow}%
\textbf{ TFPyEnvironment }%
wrapper. The original environment's API uses Numpy arrays. The%
\index{NumPy}%
\textbf{ TFPyEnvironment }%
turns these to%
\textbf{ Tensors }%
to make them compatible with Tensorflow agents and policies.%
\index{TensorFlow}%
\par%
\begin{tcolorbox}[size=title,title=Code,breakable]%
\begin{lstlisting}[language=Python, upquote=true]
train_env = tf_py_environment.TFPyEnvironment(train_py_env)
eval_env = tf_py_environment.TFPyEnvironment(eval_py_env)\end{lstlisting}
\end{tcolorbox}

\subsection{Agent}%
\label{subsec:Agent}%
An Agent represents the algorithm used to solve an RL problem. TF{-}Agents provides standard implementations of a variety of Agents:%
\index{algorithm}%
\par%
\begin{itemize}[noitemsep]%
\item%
\href{https://storage.googleapis.com/deepmind-media/dqn/DQNNaturePaper.pdf}{DQN }%
(used in this example)%
\item%
\href{http://www-anw.cs.umass.edu/~barto/courses/cs687/williams92simple.pdf}{REINFORCE}%
\item%
\href{https://arxiv.org/pdf/1509.02971.pdf}{DDPG}%
\item%
\href{https://arxiv.org/pdf/1802.09477.pdf}{TD3}%
\item%
\href{https://arxiv.org/abs/1707.06347}{PPO}%
\item%
\href{https://arxiv.org/abs/1801.01290}{SAC}%
.%
\end{itemize}%
You can only use the DQN agent in environments with a discrete action space. The DQN uses a QNetwork, a neural network model that learns to predict Q{-}Values (expected returns) for all actions given a state from the environment.%
\index{model}%
\index{neural network}%
\index{predict}%
\par%
The following code uses%
\textbf{ tf\_agents.networks.q\_network }%
to create a QNetwork, passing in the%
\textbf{ observation\_spec}%
,%
\textbf{ action\_spec}%
, and a tuple describing the number and size of the model's hidden layers.%
\index{hidden layer}%
\index{layer}%
\index{model}%
\par%
\begin{tcolorbox}[size=title,title=Code,breakable]%
\begin{lstlisting}[language=Python, upquote=true]
fc_layer_params = (100,)

q_net = q_network.QNetwork(
    train_env.observation_spec(),
    train_env.action_spec(),
    fc_layer_params=fc_layer_params)\end{lstlisting}
\end{tcolorbox}%
Now we use%
\textbf{ tf\_agents.agents.dqn.dqn\_agent }%
to instantiate a%
\textbf{ DqnAgent}%
. In addition to the%
\textbf{ time\_step\_spec}%
,%
\textbf{ action\_spec }%
and the QNetwork, the agent constructor also requires an optimizer (in this case,%
\textbf{ AdamOptimizer}%
), a loss function, and an integer step counter.%
\par%
\begin{tcolorbox}[size=title,title=Code,breakable]%
\begin{lstlisting}[language=Python, upquote=true]
optimizer = tf.compat.v1.train.AdamOptimizer(learning_rate=learning_rate)

train_step_counter = tf.Variable(0)

agent = dqn_agent.DqnAgent(
    train_env.time_step_spec(),
    train_env.action_spec(),
    q_network=q_net,
    optimizer=optimizer,
    td_errors_loss_fn=common.element_wise_squared_loss,
    train_step_counter=train_step_counter)

agent.initialize()\end{lstlisting}
\end{tcolorbox}

\subsection{Policies}%
\label{subsec:Policies}%
A policy defines the way an agent acts in an environment. Typically, reinforcement learning aims to train the underlying model until the policy produces the desired outcome.%
\index{learning}%
\index{model}%
\index{reinforcement learning}%
\par%
In this example:%
\par%
\begin{itemize}[noitemsep]%
\item%
The desired outcome is keeping the pole balanced upright over the cart.%
\item%
The policy returns an action (left or right) for each%
\textbf{\texttt{ time\_step }}%
observation.%
\end{itemize}%
Agents contain two policies:%
\par%
\begin{itemize}[noitemsep]%
\item%
\textbf{agent.policy }%
{-}  The algorithm uses this main policy for evaluation and deployment.%
\index{algorithm}%
\item%
\textbf{agent.collect\_policy }%
{-} The algorithm this secondary policy for data collection.%
\index{algorithm}%
\end{itemize}%
\begin{tcolorbox}[size=title,title=Code,breakable]%
\begin{lstlisting}[language=Python, upquote=true]
eval_policy = agent.policy
collect_policy = agent.collect_policy\end{lstlisting}
\end{tcolorbox}%
You can create policies independently of agents. For example, use%
\textbf{ random\_tf\_policy }%
to create a policy that will randomly select an action for each%
\index{random}%
\textbf{ time\_step}%
. We will use this random policy to create initial collection data to begin training.%
\index{random}%
\index{training}%
\par%
\begin{tcolorbox}[size=title,title=Code,breakable]%
\begin{lstlisting}[language=Python, upquote=true]
random_policy = random_tf_policy.RandomTFPolicy(train_env.time_step_spec(),
                                                train_env.action_spec())\end{lstlisting}
\end{tcolorbox}%
To get an action from a policy, call the%
\textbf{ policy.action }%
method. The%
\textbf{ time\_step }%
contains the observation from the environment. This method returns a%
\textbf{ PolicyStep}%
, which is a named tuple with three components:%
\par%
\begin{itemize}[noitemsep]%
\item%
\textbf{action }%
{-} The action to be taken (in this case, 0 or 1).%
\item%
\textbf{state }%
{-} Used for stateful (that is, RNN{-}based) policies.%
\item%
\textbf{info }%
{-} Auxiliary data, such as log probabilities of actions.%
\end{itemize}%
Next, we create an environment and set up the random policy.%
\index{random}%
\par%
\begin{tcolorbox}[size=title,title=Code,breakable]%
\begin{lstlisting}[language=Python, upquote=true]
example_environment = tf_py_environment.TFPyEnvironment(
    suite_gym.load('CartPole-v0'))
time_step = example_environment.reset()
random_policy.action(time_step)\end{lstlisting}
\tcbsubtitle[before skip=\baselineskip]{Output}%
\begin{lstlisting}[upquote=true]
PolicyStep(action=<tf.Tensor: shape=(1,), dtype=int64,
numpy=array([0])>, state=(), info=())
\end{lstlisting}
\end{tcolorbox}

\subsection{Metrics and Evaluation}%
\label{subsec:MetricsandEvaluation}%
The most common metric used to evaluate a policy is the average return. The return is the sum of rewards obtained while running a policy in an environment for an episode. Several episodes are run, creating an average return. The following function computes the average return, given the policy, environment, and number of episodes. We will use this same evaluation for Atari.%
\par%
\begin{tcolorbox}[size=title,title=Code,breakable]%
\begin{lstlisting}[language=Python, upquote=true]
def compute_avg_return(environment, policy, num_episodes=10):

    total_return = 0.0
    for _ in range(num_episodes):

        time_step = environment.reset()
        episode_return = 0.0

        while not time_step.is_last():
            action_step = policy.action(time_step)
            time_step = environment.step(action_step.action)
            episode_return += time_step.reward
        total_return += episode_return

    avg_return = total_return / num_episodes
    return avg_return.numpy()[0]


# See also the metrics module for standard implementations
# of different metrics.
# https://github.com/tensorflow/agents/tree/master/tf_agents/metrics\end{lstlisting}
\end{tcolorbox}%
Running this computation on the%
\textbf{\texttt{ random\_policy }}%
shows a baseline performance in the environment.%
\par%
\begin{tcolorbox}[size=title,title=Code,breakable]%
\begin{lstlisting}[language=Python, upquote=true]
compute_avg_return(eval_env, random_policy, num_eval_episodes)\end{lstlisting}
\tcbsubtitle[before skip=\baselineskip]{Output}%
\begin{lstlisting}[upquote=true]
15.2
\end{lstlisting}
\end{tcolorbox}

\subsection{Replay Buffer}%
\label{subsec:ReplayBuffer}%
The replay buffer keeps track of data collected from the environment. This tutorial uses%
\textbf{ TFUniformReplayBuffer}%
. The constructor requires the specs for the data it will be collecting. This value is available from the agent using the%
\textbf{ collect\_data\_spec }%
method. The batch size and maximum buffer length are also required.%
\par%
\begin{tcolorbox}[size=title,title=Code,breakable]%
\begin{lstlisting}[language=Python, upquote=true]
replay_buffer = tf_uniform_replay_buffer.TFUniformReplayBuffer(
    data_spec=agent.collect_data_spec,
    batch_size=train_env.batch_size,
    max_length=replay_buffer_max_length)\end{lstlisting}
\end{tcolorbox}%
For most agents,%
\textbf{ collect\_data\_spec }%
is a named tuple called%
\textbf{ Trajectory}%
, containing the specs for observations, actions, rewards, and other items.%
\par%
\begin{tcolorbox}[size=title,title=Code,breakable]%
\begin{lstlisting}[language=Python, upquote=true]
agent.collect_data_spec\end{lstlisting}
\tcbsubtitle[before skip=\baselineskip]{Output}%
\begin{lstlisting}[upquote=true]
Trajectory(
{'action': BoundedTensorSpec(shape=(), dtype=tf.int64, name='action',
minimum=array(0), maximum=array(1)),
 'discount': BoundedTensorSpec(shape=(), dtype=tf.float32,
name='discount', minimum=array(0., dtype=float32), maximum=array(1.,
dtype=float32)),
 'next_step_type': TensorSpec(shape=(), dtype=tf.int32,
name='step_type'),
 'observation': BoundedTensorSpec(shape=(4,), dtype=tf.float32,
name='observation', minimum=array([-4.8000002e+00, -3.4028235e+38,
-4.1887903e-01, -3.4028235e+38],
      dtype=float32), maximum=array([4.8000002e+00, 3.4028235e+38,
4.1887903e-01, 3.4028235e+38],
      dtype=float32)),
 'policy_info': (),
 'reward': TensorSpec(shape=(), dtype=tf.float32, name='reward'),
 'step_type': TensorSpec(shape=(), dtype=tf.int32, name='step_type')})
\end{lstlisting}
\end{tcolorbox}

\subsection{Data Collection}%
\label{subsec:DataCollection}%
Now execute the random policy in the environment for a few steps, recording the data in the replay buffer.%
\index{random}%
\par%
\begin{tcolorbox}[size=title,title=Code,breakable]%
\begin{lstlisting}[language=Python, upquote=true]
def collect_step(environment, policy, buffer):
    time_step = environment.current_time_step()
    action_step = policy.action(time_step)
    next_time_step = environment.step(action_step.action)
    traj = trajectory.from_transition(time_step, action_step, \
                                      next_time_step)

    # Add trajectory to the replay buffer
    buffer.add_batch(traj)


def collect_data(env, policy, buffer, steps):
    for _ in range(steps):
        collect_step(env, policy, buffer)


collect_data(train_env, random_policy, replay_buffer, steps=100)

# This loop is so common in RL, that we provide standard implementations.
# For more details see the drivers module.
# https://www.tensorflow.org/agents/api_docs/python/tf_agents/drivers\end{lstlisting}
\end{tcolorbox}%
The replay buffer is now a collection of Trajectories. The agent needs access to the replay buffer. TF{-}Agents provides this access by creating an iterable%
\textbf{ tf.data.Dataset }%
pipeline, which will feed data to the agent.%
\par%
Each row of the replay buffer only stores a single observation step. But since the DQN Agent needs both the current and following observation to compute the loss, the dataset pipeline will sample two adjacent rows for each item in the batch (%
\index{dataset}%
\textbf{num\_steps=2}%
).%
\par%
The program also optimizes this dataset by running parallel calls and prefetching data.%
\index{dataset}%
\par%
\begin{tcolorbox}[size=title,title=Code,breakable]%
\begin{lstlisting}[language=Python, upquote=true]
# Dataset generates trajectories with shape [Bx2x...]
dataset = replay_buffer.as_dataset(
    num_parallel_calls=3, 
    sample_batch_size=batch_size, 
    num_steps=2).prefetch(3)


dataset\end{lstlisting}
\tcbsubtitle[before skip=\baselineskip]{Output}%
\begin{lstlisting}[upquote=true]
WARNING:tensorflow:From /usr/local/lib/python3.7/dist-
packages/tensorflow/python/autograph/impl/api.py:377:
ReplayBuffer.get_next (from tf_agents.replay_buffers.replay_buffer) is
deprecated and will be removed in a future version.
Instructions for updating:
Use `as_dataset(..., single_deterministic_pass=False) instead.
<PrefetchDataset element_spec=(Trajectory(
{'action': TensorSpec(shape=(64, 2), dtype=tf.int64, name=None),
 'discount': TensorSpec(shape=(64, 2), dtype=tf.float32, name=None),
 'next_step_type': TensorSpec(shape=(64, 2), dtype=tf.int32,
name=None),
 'observation': TensorSpec(shape=(64, 2, 4), dtype=tf.float32,
name=None),
 'policy_info': (),
 'reward': TensorSpec(shape=(64, 2), dtype=tf.float32, name=None),
 'step_type': TensorSpec(shape=(64, 2), dtype=tf.int32, name=None)}),
BufferInfo(ids=TensorSpec(shape=(64, 2), dtype=tf.int64, name=None),
probabilities=TensorSpec(shape=(64,), dtype=tf.float32, name=None)))>
\end{lstlisting}
\end{tcolorbox}%
\begin{tcolorbox}[size=title,title=Code,breakable]%
\begin{lstlisting}[language=Python, upquote=true]
iterator = iter(dataset)

print(iterator)\end{lstlisting}
\tcbsubtitle[before skip=\baselineskip]{Output}%
\begin{lstlisting}[upquote=true]
<tensorflow.python.data.ops.iterator_ops.OwnedIterator object at
0x7f05c0006c10>
\end{lstlisting}
\end{tcolorbox}

\subsection{Training the agent}%
\label{subsec:Trainingtheagent}%
Two things must happen during the training loop:%
\index{training}%
\par%
\begin{itemize}[noitemsep]%
\item%
Collect data from the environment%
\item%
Use that data to train the agent's neural network(s)%
\index{neural network}%
\end{itemize}%
This example also periodically evaluates the policy and prints the current score.%
\par%
The following will take \textasciitilde{}5 minutes to run.%
\par%
\begin{tcolorbox}[size=title,title=Code,breakable]%
\begin{lstlisting}[language=Python, upquote=true]
# (Optional) Optimize by wrapping some of the code in a graph
# using TF function.
agent.train = common.function(agent.train)

# Reset the train step
agent.train_step_counter.assign(0)

# Evaluate the agent's policy once before training.
avg_return = compute_avg_return(eval_env, agent.policy,
                                num_eval_episodes)
returns = [avg_return]

for _ in range(num_iterations):

    # Collect a few steps using collect_policy and 
    # save to the replay buffer.
    for _ in range(collect_steps_per_iteration):
        collect_step(train_env, agent.collect_policy, replay_buffer)

    # Sample a batch of data from the buffer and update 
    # the agent's network.
    experience, unused_info = next(iterator)
    train_loss = agent.train(experience).loss

    step = agent.train_step_counter.numpy()

    if step % log_interval == 0:
        print('step = {0}: loss = {1}'.format(step, train_loss))

    if step % eval_interval == 0:
        avg_return = compute_avg_return(eval_env, agent.policy,
                                        num_eval_episodes)
        print('step = {0}: Average Return = {1}'.format(step, avg_return))
        returns.append(avg_return)\end{lstlisting}
\tcbsubtitle[before skip=\baselineskip]{Output}%
\begin{lstlisting}[upquote=true]
WARNING:tensorflow:From /usr/local/lib/python3.7/dist-
packages/tensorflow/python/util/dispatch.py:1082: calling foldr_v2
(from tensorflow.python.ops.functional_ops) with back_prop=False is
deprecated and will be removed in a future version.
Instructions for updating:
back_prop=False is deprecated. Consider using tf.stop_gradient
instead.
Instead of:
results = tf.foldr(fn, elems, back_prop=False)
Use:
results = tf.nest.map_structure(tf.stop_gradient, tf.foldr(fn, elems))
step = 200: loss = 23.158374786376953
step = 400: loss = 7.158817768096924
step = 600: loss = 30.97699737548828
step = 800: loss = 9.831337928771973

...

step = 19400: loss = 16.59900665283203
step = 19600: loss = 16.253849029541016
step = 19800: loss = 124.63180541992188
step = 20000: loss = 22.45917320251465
step = 20000: Average Return = 198.3000030517578
\end{lstlisting}
\end{tcolorbox}

\subsection{Visualization and Plots}%
\label{subsec:VisualizationandPlots}%
Use%
\textbf{ matplotlib.pyplot }%
to chart how the policy improved during training.%
\index{training}%
\par%
One iteration of%
\index{iteration}%
\textbf{ Cartpole{-}v0 }%
consists of 200 time steps. The environment rewards%
\textbf{\texttt{ +1 }}%
for each step the pole stays up, so the maximum return for one episode is 200. The charts show the return increasing towards that maximum each time the algorithm evaluates it during training. (It may be a little unstable and not increase each time monotonically.)%
\index{algorithm}%
\index{training}%
\par%
\begin{tcolorbox}[size=title,title=Code,breakable]%
\begin{lstlisting}[language=Python, upquote=true]
iterations = range(0, num_iterations + 1, eval_interval)
plt.plot(iterations, returns)
plt.ylabel('Average Return')
plt.xlabel('Iterations')
plt.ylim(top=250)\end{lstlisting}
\tcbsubtitle[before skip=\baselineskip]{Output}%
\includegraphics[width=3in]{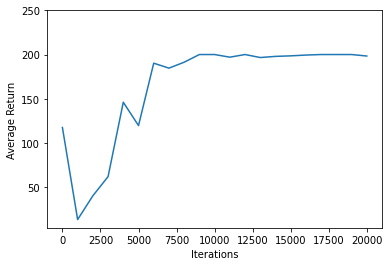}%
\begin{lstlisting}[upquote=true]
(3.859999799728394, 250.0)
\end{lstlisting}
\end{tcolorbox}

\subsection{Videos}%
\label{subsec:Videos}%
The charts are nice. But more exciting is seeing an agent performing a task in an environment.%
\par%
First, create a function to embed videos in the notebook.%
\index{video}%
\par%
\begin{tcolorbox}[size=title,title=Code,breakable]%
\begin{lstlisting}[language=Python, upquote=true]
def embed_mp4(filename):
    """Embeds an mp4 file in the notebook."""
    video = open(filename, 'rb').read()
    b64 = base64.b64encode(video)
    tag = '''
  <video width="640" height="480" controls>
    <source src="data:video/mp4;base64,{0}" type="video/mp4">
  Your browser does not support the video tag.
  </video>'''.format(b64.decode())

    return IPython.display.HTML(tag)\end{lstlisting}
\end{tcolorbox}%
Now iterate through a few episodes of the Cartpole game with the agent. The underlying Python environment (the one "inside" the TensorFlow environment wrapper) provides a%
\index{Python}%
\index{TensorFlow}%
\textbf{\texttt{ render() }}%
method, which outputs an image of the environment state. We can collect these frames into a video.%
\index{output}%
\index{video}%
\par%
\begin{tcolorbox}[size=title,title=Code,breakable]%
\begin{lstlisting}[language=Python, upquote=true]
def create_policy_eval_video(policy, filename, num_episodes=5, fps=30):
    filename = filename + ".mp4"
    with imageio.get_writer(filename, fps=fps) as video:
        for _ in range(num_episodes):
            time_step = eval_env.reset()
            video.append_data(eval_py_env.render())
            while not time_step.is_last():
                action_step = policy.action(time_step)
                time_step = eval_env.step(action_step.action)
                video.append_data(eval_py_env.render())
    return embed_mp4(filename)


create_policy_eval_video(agent.policy, "trained-agent")\end{lstlisting}
\end{tcolorbox}%
For fun, compare the trained agent (above) to an agent moving randomly. (It does not do as well.)%
\index{random}%
\par%
\begin{tcolorbox}[size=title,title=Code,breakable]%
\begin{lstlisting}[language=Python, upquote=true]
create_policy_eval_video(random_policy, "random-agent")\end{lstlisting}
\end{tcolorbox}

\section{Part 12.4: Atari Games with Keras Neural Networks}%
\label{sec:Part12.4AtariGameswithKerasNeuralNetworks}%
The Atari 2600 is a home video game console from Atari, Inc., Released on September 11, 1977. Most credit the Atari with popularizing microprocessor{-}based hardware and games stored on ROM cartridges instead of dedicated hardware with games built into the unit. Atari bundled their console with two joystick controllers, a conjoined pair of paddle controllers, and a game cartridge: initially%
\index{ROC}%
\index{ROC}%
\index{video}%
\href{https://en.wikipedia.org/wiki/Combat_(Atari_2600)}{ Combat}%
, and later%
\href{https://en.wikipedia.org/wiki/Pac-Man_(Atari_2600)}{ Pac{-}Man}%
.%
\par%
Atari emulators are popular and allow gamers to play many old Atari video games on modern computers. These emulators are even available as JavaScript.%
\index{Java}%
\index{JavaScript}%
\index{video}%
\par
\begin{figure}[h]%
\centering%
\includegraphics[width=4in]{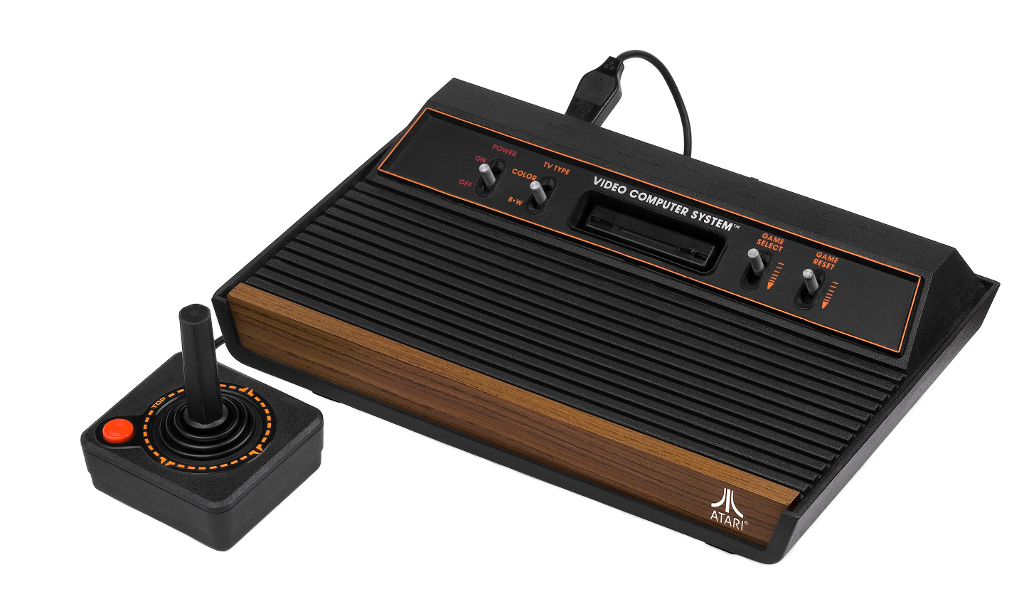}%
\caption{The Atari 2600}%
\label{12.ATARI}%
\end{figure}

%
\begin{itemize}[noitemsep]%
\item%
\href{http://www.virtualatari.org/listP.html}{Virtual Atari}%
\end{itemize}%
Atari games have become popular benchmarks for AI systems, particularly reinforcement learning. OpenAI Gym internally uses the%
\index{gym}%
\index{learning}%
\index{OpenAI}%
\index{reinforcement learning}%
\href{https://stella-emu.github.io/}{ Stella Atari Emulator}%
. You can see the Atari 2600 in Figure \ref{12.ATARI}.%
\par%

\par%
\subsection{Actual Atari 2600 Specs}%
\label{subsec:ActualAtari2600Specs}%
\begin{itemize}[noitemsep]%
\item%
CPU: 1.19 MHz MOS Technology 6507%
\item%
Audio + Video processor: Television Interface Adapter (TIA)%
\index{ROC}%
\index{ROC}%
\index{video}%
\item%
Playfield resolution: 40 x 192 pixels (NTSC). It uses a 20{-}pixel register that is mirrored or copied, left side to right side, to achieve the width of 40 pixels.%
\item%
Player sprites: 8 x 192 pixels (NTSC). Player, ball, and missile sprites use pixels 1/4 the width of playfield pixels (unless stretched).%
\index{layer}%
\item%
Ball and missile sprites: 1 x 192 pixels (NTSC).%
\item%
Maximum resolution: 160 x 192 pixels (NTSC). Max resolution is achievable only with programming tricks that combine sprite pixels with playfield pixels.%
\item%
128 colors (NTSC). 128 possible on screen. Max of 4 per line: background, playfield, player0 sprite, and player1 sprite. Palette switching between lines is common. Palette switching mid{-}line is possible but not common due to resource limitations.%
\index{layer}%
\item%
2 channels of 1{-}bit monaural sound with 4{-}bit volume control.%
\end{itemize}

\subsection{OpenAI Lab Atari Pong}%
\label{subsec:OpenAILabAtariPong}%
You can use OpenAI Gym with Windows; however, it requires a special%
\index{gym}%
\index{OpenAI}%
\href{https://towardsdatascience.com/how-to-install-openai-gym-in-a-windows-environment-338969e24d30}{ installation procedure}%
.%
\par%
This chapter demonstrates playing%
\href{https://github.com/wau/keras-rl2/blob/master/examples/dqn_atari.py}{ Atari Pong}%
. Pong is a two{-}dimensional sports game that simulates table tennis. The player controls an in{-}game paddle by moving it vertically across the left or right side of the screen. They can compete against another player controlling a second paddle on the opposing side. Players use the paddles to hit a ball back and forth. The goal is for each player to reach eleven points before the opponent; you earn points when one fails to return it to the other. For the Atari 2600 version of Pong, a computer player (controlled by the Atari 2600) is the opposing player.%
\index{layer}%
\index{two{-}dimensional}%
\par%
This section shows how to adapt TF{-}Agents to an Atari game. You can quickly adapt this example to any Atari game by simply changing the environment name. However, I tuned the code presented here for Pong, and it may not perform as well for other games. Some tuning will likely be necessary to produce a good agent for other games. Compared to the pole{-}cart game presented earlier in this chapter, some changes are required.%
\index{SOM}%
\par%
We begin by importing the needed Python packages.%
\index{Python}%
\par%
\begin{tcolorbox}[size=title,title=Code,breakable]%
\begin{lstlisting}[language=Python, upquote=true]
import base64
import imageio
import IPython
import matplotlib
import matplotlib.pyplot as plt
import numpy as np
import PIL.Image
import pyvirtualdisplay

import tensorflow as tf

from tf_agents.agents.dqn import dqn_agent
from tf_agents.drivers import dynamic_step_driver
from tf_agents.environments import suite_gym, suite_atari
from tf_agents.environments import tf_py_environment
from tf_agents.environments import batched_py_environment
from tf_agents.eval import metric_utils
from tf_agents.metrics import tf_metrics
from tf_agents.networks import q_network, network
from tf_agents.policies import random_tf_policy
from tf_agents.replay_buffers import tf_uniform_replay_buffer
from tf_agents.trajectories import trajectory
from tf_agents.utils import common
from tf_agents.agents.categorical_dqn import categorical_dqn_agent
from tf_agents.networks import categorical_q_network

from tf_agents.specs import tensor_spec
from tf_agents.trajectories import time_step as ts

# Set up a virtual display for rendering OpenAI gym environments.
display = pyvirtualdisplay.Display(visible=0, size=(1400, 900)).start()\end{lstlisting}
\end{tcolorbox}

\subsection{Hyperparameters}%
\label{subsec:Hyperparameters}%
The hyperparameter names are the same as the previous DQN example; however, I tuned the numeric values for the more complex Atari game.%
\index{hyperparameter}%
\index{parameter}%
\par%
\begin{tcolorbox}[size=title,title=Code,breakable]%
\begin{lstlisting}[language=Python, upquote=true]
# 10K already takes awhile to complete, with minimal results.
# To get an effective agent requires much more.
num_iterations = 10000

initial_collect_steps = 200
collect_steps_per_iteration = 10
replay_buffer_max_length = 100000

batch_size = 32
learning_rate = 2.5e-3
log_interval = 1000

num_eval_episodes = 5
eval_interval = 25000\end{lstlisting}
\end{tcolorbox}%
The algorithm needs more iterations for an Atari game. I also found that increasing the number of collection steps helped the algorithm train.%
\index{algorithm}%
\index{iteration}%
\par

\subsection{Atari Environment}%
\label{subsec:AtariEnvironment}%
You must handle Atari environments differently than games like cart{-}poll. Atari games typically use their 2D displays as the environment state. AI Gym represents Atari games as either a 3D (height by width by color) state spaced based on their screens or a vector representing the game's computer RAM state. To preprocess Atari games for greater computational efficiency, we skip several frames, decrease the resolution, and discard color information. The following code shows how we can set up an Atari environment.%
\index{gym}%
\index{ROC}%
\index{ROC}%
\index{vector}%
\par%
\begin{tcolorbox}[size=title,title=Code,breakable]%
\begin{lstlisting}[language=Python, upquote=true]
! wget http://www.atarimania.com/roms/Roms.rar
! mkdir /content/ROM/
! unrar e -o+ /content/Roms.rar /content/ROM/
! python -m atari_py.import_roms /content/ROM/\end{lstlisting}
\end{tcolorbox}%
\begin{tcolorbox}[size=title,title=Code,breakable]%
\begin{lstlisting}[language=Python, upquote=true]
#env_name = 'Breakout-v4'
env_name = 'Pong-v0'
#env_name = 'BreakoutDeterministic-v4'
#env = suite_gym.load(env_name)

# AtariPreprocessing runs 4 frames at a time, max-pooling over the last 2
# frames. We need to account for this when computing things like update
# intervals.
ATARI_FRAME_SKIP = 4

max_episode_frames=108000  # ALE frames

env = suite_atari.load(
    env_name,
    max_episode_steps=max_episode_frames / ATARI_FRAME_SKIP,
    gym_env_wrappers=suite_atari.DEFAULT_ATARI_GYM_WRAPPERS_WITH_STACKING)
#env = batched_py_environment.BatchedPyEnvironment([env])\end{lstlisting}
\end{tcolorbox}%
We can now reset the environment and display one step.  The following image shows how the Pong game environment appears to a user.%
\par%
\begin{tcolorbox}[size=title,title=Code,breakable]%
\begin{lstlisting}[language=Python, upquote=true]
env.reset()
PIL.Image.fromarray(env.render())\end{lstlisting}
\tcbsubtitle[before skip=\baselineskip]{Output}%
\includegraphics[width=1in]{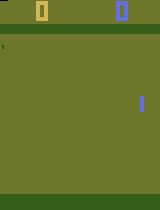}%
\end{tcolorbox}%
We are now ready to load and wrap the two environments for TF{-}Agents. The algorithm uses the first environment for evaluation and the second to train.%
\index{algorithm}%
\par%
\begin{tcolorbox}[size=title,title=Code,breakable]%
\begin{lstlisting}[language=Python, upquote=true]
train_py_env = suite_atari.load(
    env_name,
    max_episode_steps=max_episode_frames / ATARI_FRAME_SKIP,
    gym_env_wrappers=suite_atari.DEFAULT_ATARI_GYM_WRAPPERS_WITH_STACKING)

eval_py_env = suite_atari.load(
    env_name,
    max_episode_steps=max_episode_frames / ATARI_FRAME_SKIP,
    gym_env_wrappers=suite_atari.DEFAULT_ATARI_GYM_WRAPPERS_WITH_STACKING)

train_env = tf_py_environment.TFPyEnvironment(train_py_env)
eval_env = tf_py_environment.TFPyEnvironment(eval_py_env)\end{lstlisting}
\end{tcolorbox}

\subsection{Agent}%
\label{subsec:Agent}%
I used the following code from the TF{-}Agents examples to wrap up the regular Q{-}network class.  The AtariQNetwork class ensures that the pixel values from the Atari screen are divided by 255.  This division assists the neural network by normalizing the pixel values between 0 and 1.%
\index{neural network}%
\par%
\begin{tcolorbox}[size=title,title=Code,breakable]%
\begin{lstlisting}[language=Python, upquote=true]
# AtariPreprocessing runs 4 frames at a time, max-pooling over the last 2
# frames. We need to account for this when computing things like update
# intervals.
ATARI_FRAME_SKIP = 4


class AtariCategoricalQNetwork(network.Network):
    """CategoricalQNetwork subclass that divides observations by 255."""

    def __init__(self, input_tensor_spec, action_spec, **kwargs):
        super(AtariCategoricalQNetwork, self).__init__(
            input_tensor_spec, state_spec=())
        input_tensor_spec = tf.TensorSpec(
            dtype=tf.float32, shape=input_tensor_spec.shape)
        self._categorical_q_network = \
            categorical_q_network.CategoricalQNetwork(
                input_tensor_spec, action_spec, **kwargs)

    @property
    def num_atoms(self):
        return self._categorical_q_network.num_atoms

    def call(self, observation, step_type=None, network_state=()):
        state = tf.cast(observation, tf.float32)
        # We divide the grayscale pixel values by 255 here rather than
        # storing normalized values beause uint8s are 4x cheaper to
        # store than float32s.
        # TODO(b/129805821): handle the division by 255 for
        # train_eval_atari.py in
        # a preprocessing layer instead.
        state = state / 255
        return self._categorical_q_network(
            state, step_type=step_type, network_state=network_state)\end{lstlisting}
\end{tcolorbox}%
Next, we introduce two hyperparameters specific to the neural network we are about to define.%
\index{hyperparameter}%
\index{neural network}%
\index{parameter}%
\par%
\begin{tcolorbox}[size=title,title=Code,breakable]%
\begin{lstlisting}[language=Python, upquote=true]
fc_layer_params = (512,)
conv_layer_params = ((32, (8, 8), 4), (64, (4, 4), 2), (64, (3, 3), 1))

q_net = AtariCategoricalQNetwork(
    train_env.observation_spec(),
    train_env.action_spec(),
    conv_layer_params=conv_layer_params,
    fc_layer_params=fc_layer_params)\end{lstlisting}
\end{tcolorbox}%
Convolutional neural networks usually comprise several alternating pairs of convolution and max{-}pooling layers, ultimately culminating in one or more dense layers. These layers are the same types as previously seen in this course. The%
\index{convolution}%
\index{convolutional}%
\index{Convolutional Neural Networks}%
\index{dense layer}%
\index{layer}%
\index{neural network}%
\textbf{ QNetwork }%
accepts two parameters that define the convolutional neural network structure.%
\index{convolution}%
\index{convolutional}%
\index{neural network}%
\index{parameter}%
\par%
The more simple of the two parameters is%
\index{parameter}%
\textbf{ fc\_layer\_params}%
. This parameter specifies the size of each of the dense layers. A tuple specifies the size of each of the layers in a list.%
\index{dense layer}%
\index{layer}%
\index{parameter}%
\par%
The second parameter, named%
\index{parameter}%
\textbf{ conv\_layer\_params}%
, is a list of convolution layers parameters, where each item is a length{-}three tuple indicating (filters, kernel\_size, stride). This implementation of QNetwork supports only convolution layers. If you desire a more complex convolutional neural network, you must define your variant of the%
\index{convolution}%
\index{convolutional}%
\index{layer}%
\index{neural network}%
\index{parameter}%
\textbf{ QNetwork}%
.%
\par%
The%
\textbf{ QNetwork }%
defined here is not the agent. Instead, the%
\textbf{ QNetwork }%
is used by the DQN agent to implement the actual neural network. This technique allows flexibility as you can set your class if needed.%
\index{neural network}%
\par%
Next, we define the optimizer. For this example, I used RMSPropOptimizer. However, AdamOptimizer is another popular choice. We also created the DQN agent and referenced the Q{-}network.%
\index{ADAM}%
\par%
\begin{tcolorbox}[size=title,title=Code,breakable]%
\begin{lstlisting}[language=Python, upquote=true]
optimizer = tf.compat.v1.train.RMSPropOptimizer(
    learning_rate=learning_rate,
    decay=0.95,
    momentum=0.0,
    epsilon=0.00001,
    centered=True)

train_step_counter = tf.Variable(0)

observation_spec = tensor_spec.from_spec(train_env.observation_spec())
time_step_spec = ts.time_step_spec(observation_spec)

action_spec = tensor_spec.from_spec(train_env.action_spec())
target_update_period = 32000  # ALE frames
update_period = 16  # ALE frames
_update_period = update_period / ATARI_FRAME_SKIP


agent = categorical_dqn_agent.CategoricalDqnAgent(
    time_step_spec,
    action_spec,
    categorical_q_network=q_net,
    optimizer=optimizer,
    # epsilon_greedy=epsilon,
    n_step_update=1.0,
    target_update_tau=1.0,
    target_update_period=(
        target_update_period / ATARI_FRAME_SKIP / _update_period),
    gamma=0.99,
    reward_scale_factor=1.0,
    gradient_clipping=None,
    debug_summaries=False,
    summarize_grads_and_vars=False)

agent.initialize()\end{lstlisting}
\end{tcolorbox}

\subsection{Metrics and Evaluation}%
\label{subsec:MetricsandEvaluation}%
There are many different ways to measure the effectiveness of a model trained with reinforcement learning. The loss function of the internal Q{-}network is not a good measure of the entire DQN algorithm's overall fitness. The network loss function measures how close the Q{-}network fits the collected data and does not indicate how effectively the DQN maximizes rewards. The method used for this example tracks the average reward received over several episodes.%
\index{algorithm}%
\index{learning}%
\index{model}%
\index{reinforcement learning}%
\par%
\begin{tcolorbox}[size=title,title=Code,breakable]%
\begin{lstlisting}[language=Python, upquote=true]
def compute_avg_return(environment, policy, num_episodes=10):

    total_return = 0.0
    for _ in range(num_episodes):

        time_step = environment.reset()
        episode_return = 0.0

        while not time_step.is_last():
            action_step = policy.action(time_step)
            time_step = environment.step(action_step.action)
            episode_return += time_step.reward
        total_return += episode_return

    avg_return = total_return / num_episodes
    return avg_return.numpy()[0]


# See also the metrics module for standard implementations of
# different metrics.
# https://github.com/tensorflow/agents/tree/master/tf_agents/metrics\end{lstlisting}
\end{tcolorbox}

\subsection{Replay Buffer}%
\label{subsec:ReplayBuffer}%
DQN works by training a neural network to predict the Q{-}values for every possible environment state. A neural network needs training data, so the algorithm accumulates this training data as it runs episodes. The replay buffer is where this data is stored. Only the most recent episodes are stored; older episode data rolls off the queue as the queue accumulates new data.%
\index{algorithm}%
\index{neural network}%
\index{predict}%
\index{training}%
\par%
\begin{tcolorbox}[size=title,title=Code,breakable]%
\begin{lstlisting}[language=Python, upquote=true]
replay_buffer = tf_uniform_replay_buffer.TFUniformReplayBuffer(
    data_spec=agent.collect_data_spec,
    batch_size=train_env.batch_size,
    max_length=replay_buffer_max_length)

# Dataset generates trajectories with shape [Bx2x...]
dataset = replay_buffer.as_dataset(
    num_parallel_calls=3,
    sample_batch_size=batch_size,
    num_steps=2).prefetch(3)\end{lstlisting}
\tcbsubtitle[before skip=\baselineskip]{Output}%
\begin{lstlisting}[upquote=true]
WARNING:tensorflow:From /usr/local/lib/python3.7/dist-
packages/tensorflow/python/autograph/impl/api.py:377:
ReplayBuffer.get_next (from tf_agents.replay_buffers.replay_buffer) is
deprecated and will be removed in a future version.
Instructions for updating:
Use `as_dataset(..., single_deterministic_pass=False) instead.
\end{lstlisting}
\end{tcolorbox}

\subsection{Random Collection}%
\label{subsec:RandomCollection}%
The algorithm must prime the pump.  Training cannot begin on an empty replay buffer.  The following code performs a predefined number of steps to generate initial training data.%
\index{algorithm}%
\index{training}%
\par%
\begin{tcolorbox}[size=title,title=Code,breakable]%
\begin{lstlisting}[language=Python, upquote=true]
random_policy = random_tf_policy.RandomTFPolicy(train_env.time_step_spec(),
                                                train_env.action_spec())


def collect_step(environment, policy, buffer):
    time_step = environment.current_time_step()
    action_step = policy.action(time_step)
    next_time_step = environment.step(action_step.action)
    traj = trajectory.from_transition(time_step, action_step,\
                                      next_time_step)

    # Add trajectory to the replay buffer
    buffer.add_batch(traj)


def collect_data(env, policy, buffer, steps):
    for _ in range(steps):
        collect_step(env, policy, buffer)


collect_data(train_env, random_policy, replay_buffer,
             steps=initial_collect_steps)\end{lstlisting}
\end{tcolorbox}

\subsection{Training the Agent}%
\label{subsec:TrainingtheAgent}%
We are now ready to train the DQN. Depending on how many episodes you wish to run through, this process can take many hours. This code will update both the loss and average return as training occurs. As training becomes more successful, the average return should increase. The losses reported reflecting the average loss for individual training batches.%
\index{ROC}%
\index{ROC}%
\index{training}%
\index{training batch}%
\par%
\begin{tcolorbox}[size=title,title=Code,breakable]%
\begin{lstlisting}[language=Python, upquote=true]
iterator = iter(dataset)

# (Optional) Optimize by wrapping some of the code in a graph
# using TF function.
agent.train = common.function(agent.train)

# Reset the train step
agent.train_step_counter.assign(0)

# Evaluate the agent's policy once before training.
avg_return = compute_avg_return(eval_env, agent.policy,
                                num_eval_episodes)
returns = [avg_return]

for _ in range(num_iterations):

    # Collect a few steps using collect_policy and 
    # save to the replay buffer.
    for _ in range(collect_steps_per_iteration):
        collect_step(train_env, agent.collect_policy, replay_buffer)

    # Sample a batch of data from the buffer and 
    # update the agent's network.
    experience, unused_info = next(iterator)
    train_loss = agent.train(experience).loss

    step = agent.train_step_counter.numpy()

    if step % log_interval == 0:
        print('step = {0}: loss = {1}'.format(step, train_loss))

    if step % eval_interval == 0:
        avg_return = compute_avg_return(eval_env, agent.policy,
                                        num_eval_episodes)
        print('step = {0}: Average Return = {1}'.format(step, avg_return))
        returns.append(avg_return)\end{lstlisting}
\tcbsubtitle[before skip=\baselineskip]{Output}%
\begin{lstlisting}[upquote=true]
step = 1000: loss = 3.9279017448425293
step = 2000: loss = 3.9280214309692383
step = 3000: loss = 3.924931526184082
step = 4000: loss = 3.9209065437316895
step = 5000: loss = 3.919551134109497
step = 6000: loss = 3.919588327407837
step = 7000: loss = 3.9074008464813232
step = 8000: loss = 3.8954014778137207
step = 9000: loss = 3.8865578174591064
step = 10000: loss = 3.895845890045166
\end{lstlisting}
\end{tcolorbox}

\subsection{Videos}%
\label{subsec:Videos}%
Perhaps the most compelling way to view an Atari game's results is a video that allows us to see the agent play the game. We now have a trained model and observed its training progress on a graph. The following functions are defined to watch the agent play the game in the notebook.%
\index{model}%
\index{training}%
\index{video}%
\par%
\begin{tcolorbox}[size=title,title=Code,breakable]%
\begin{lstlisting}[language=Python, upquote=true]
def embed_mp4(filename):
    """Embeds an mp4 file in the notebook."""
    video = open(filename, 'rb').read()
    b64 = base64.b64encode(video)
    tag = '''
  <video width="640" height="480" controls>
    <source src="data:video/mp4;base64,{0}" type="video/mp4">
  Your browser does not support the video tag.
  </video>'''.format(b64.decode())

    return IPython.display.HTML(tag)


def create_policy_eval_video(policy, filename, num_episodes=5, fps=30):
    filename = filename + ".mp4"
    with imageio.get_writer(filename, fps=fps) as video:
        for _ in range(num_episodes):
            time_step = eval_env.reset()
            video.append_data(eval_py_env.render())
            while not time_step.is_last():
                action_step = policy.action(time_step)
                time_step = eval_env.step(action_step.action)
                video.append_data(eval_py_env.render())
    return embed_mp4(filename)\end{lstlisting}
\end{tcolorbox}%
First, we will observe the trained agent play the game.%
\par%
\begin{tcolorbox}[size=title,title=Code,breakable]%
\begin{lstlisting}[language=Python, upquote=true]
create_policy_eval_video(agent.policy, "trained-agent")\end{lstlisting}
\end{tcolorbox}%
For comparison, we observe a random agent play.  While the trained agent is far from perfect, with enough training, it does outperform the random agent considerably.%
\index{random}%
\index{training}%
\par%
\begin{tcolorbox}[size=title,title=Code,breakable]%
\begin{lstlisting}[language=Python, upquote=true]
create_policy_eval_video(random_policy, "random-agent")\end{lstlisting}
\end{tcolorbox}

\section{Part 12.5: Application of Reinforcement Learning}%
\label{sec:Part12.5ApplicationofReinforcementLearning}%
Creating an environment is the first step to applying TF{-}Agent{-}based reinforcement learning to a problem with your design. This part will see how to create your environment and apply it to an agent that allows actions to be floating{-}point values rather than the discrete actions employed by the Deep Q{-}Networks (DQN) that we used earlier in this chapter. This new type of agent is called a Deep Deterministic Policy Gradients (DDPG) network. From an application standpoint, the primary difference between DDPG and DQN is that DQN only supports discrete actions, whereas DDPG supports continuous actions; however, there are other essential differences that we will cover later in this chapter.%
\index{continuous}%
\index{gradient}%
\index{learning}%
\index{reinforcement learning}%
\par%
The environment that I will demonstrate in this chapter simulates paying off a mortgage and saving for retirement. This simulation allows the agent to allocate their income between several types of accounts, buying luxury items, and paying off their mortgage. The goal is to maximize net worth. Because we wish to provide the agent with the ability to distribute their income among several accounts, we provide continuous (floating point) actions that determine this distribution of the agent's salary.%
\index{continuous}%
\par%
Similar to previous TF{-}Agent examples in this chapter, we begin by importing needed packages.%
\par%
\begin{tcolorbox}[size=title,title=Code,breakable]%
\begin{lstlisting}[language=Python, upquote=true]
import base64
import imageio
import IPython
import matplotlib
import matplotlib.pyplot as plt
import numpy as np
import PIL.Image
import pyvirtualdisplay
import math
import numpy as np

import tensorflow as tf

from tf_agents.agents.ddpg import actor_network
from tf_agents.agents.ddpg import critic_network
from tf_agents.agents.ddpg import ddpg_agent

from tf_agents.agents.dqn import dqn_agent
from tf_agents.drivers import dynamic_step_driver
from tf_agents.environments import suite_gym
from tf_agents.environments import tf_py_environment
from tf_agents.eval import metric_utils
from tf_agents.metrics import tf_metrics
from tf_agents.networks import q_network
from tf_agents.policies import random_tf_policy
from tf_agents.replay_buffers import tf_uniform_replay_buffer
from tf_agents.trajectories import trajectory
from tf_agents.trajectories import policy_step
from tf_agents.utils import common

import gym
from gym import spaces
from gym.utils import seeding
from gym.envs.registration import register
import PIL.ImageDraw
import PIL.Image
from PIL import ImageFont\end{lstlisting}
\end{tcolorbox}%
If you get the following error, restart and rerun the Google CoLab environment.  Sometimes a restart is needed after installing TF{-}Agents.%
\index{error}%
\index{SOM}%
\par%
\begin{tcolorbox}[size=title,breakable]%
\begin{lstlisting}[upquote=true]
AttributeError: module 'google.protobuf.descriptor' has no 
    attribute '_internal_create_key'
\end{lstlisting}
\end{tcolorbox}%
We create a virtual display to view the simulation in a Jupyter notebook.%
\par%
\begin{tcolorbox}[size=title,title=Code,breakable]%
\begin{lstlisting}[language=Python, upquote=true]
# Set up a virtual display for rendering OpenAI gym environments.
vdisplay = pyvirtualdisplay.Display(visible=0, size=(1400, 900)).start()\end{lstlisting}
\end{tcolorbox}%
\subsection{Create an Environment of your Own}%
\label{subsec:CreateanEnvironmentofyourOwn}%
An environment is a simulator that your agent runs in. An environment must have a current state. Some of this state is visible to the agent. However, the environment also hides some aspects of the state from the agent. Likewise, the agent takes actions that will affect the state of the environment. There may also be internal actions outside the agent's control. For example, in the finance simulator demonstrated in this section, the agent does not control the investment returns or rate of inflation. Instead, the agent must react to these external actions and state components.%
\index{SOM}%
\par%
The environment class that you create must contain these elements:%
\par%
\begin{itemize}[noitemsep]%
\item%
Be a child class of%
\textbf{ gym.Env}%
\item%
Implement a%
\textbf{ seed }%
function that sets a seed that governs the simulation's random aspects. For this environment, the seed oversees the random fluctuations in inflation and rates of return.%
\index{random}%
\item%
Implement a%
\textbf{ reset }%
function that resets the state for a new episode.%
\item%
Implement a%
\textbf{ render }%
function that renders one frame of the simulation. The rendering is only for display and does not affect reinforcement learning.%
\index{learning}%
\index{reinforcement learning}%
\item%
Implement a%
\textbf{ step }%
function that performs one step of your simulation.%
\end{itemize}%
The class presented below implements a financial planning simulation. The agent must save for retirement and should attempt to amass the greatest possible net worth. The simulation includes the following key elements:%
\par%
\begin{itemize}[noitemsep]%
\item%
Random starting salary between 40K (USD) and 60K (USD).%
\index{random}%
\item%
Home loan for a house with a random purchase price between 1.5 and 4 times the starting salary.%
\index{random}%
\item%
Home loan is a standard amortized 30{-}year loan with a fixed monthly payment.%
\item%
Paying higher than the home's monthly payment pays the loan down quicker. Paying below the monthly payment results in late fees and eventually foreclosure.%
\item%
Ability to allocate income between luxury purchases and home payments (above or below payment amount) and a taxable and tax{-}advantaged savings account.%
\end{itemize}%
The state is composed of the following floating{-}point values:%
\par%
\begin{itemize}[noitemsep]%
\item%
\textbf{age }%
{-} The agent's current age in months (steps)%
\item%
\textbf{salary }%
{-} The agent's starting salary, increases relative to inflation.%
\item%
\textbf{home\_value }%
{-} The value of the agent's home, increases relative to inflation.%
\item%
\textbf{home\_loan }%
{-} How much the agent still owes on their home.%
\item%
\textbf{req\_home\_pmt }%
{-} The minimum required home payment.%
\item%
\textbf{acct\_tax\_adv }%
{-} The balance of the tax advantaged retirement account.%
\item%
\textbf{acct\_tax }%
{-} The balance of the taxable retuirement account.%
\end{itemize}%
The action space is composed of the following floating{-}point values (between 0 and 1):%
\par%
\begin{itemize}[noitemsep]%
\item%
\textbf{home\_loan }%
{-} The amount to apply to a home loan.%
\item%
\textbf{savings\_tax\_adv }%
{-} The amount to deposit in a tax{-}advantaged savings account.%
\item%
\textbf{savings taxable }%
{-} The amount to deposit in a taxable savings account.%
\item%
\textbf{luxury }%
{-} The amount to spend on luxury items/services.%
\end{itemize}%
The actions are weights that the program converts to a percentage of the total. For example, the home loan percentage is the home loan action value divided by all actions (including a home loan). The following code implements the environment and provides implementation details in the comments.%
\par%
\begin{tcolorbox}[size=title,title=Code,breakable]%
\begin{lstlisting}[language=Python, upquote=true]
class SimpleGameOfLifeEnv(gym.Env):
    metadata = {
        'render.modes': ['human', 'rgb_array'],
        'video.frames_per_second': 1
    }

    STATE_ELEMENTS = 7
    STATES = ['age', 'salary', 'home_value', 'home_loan', 'req_home_pmt',
              'acct_tax_adv', 'acct_tax', "expenses", "actual_home_pmt",
              "tax_deposit",
              "tax_adv_deposit", "net_worth"]
    STATE_AGE = 0
    STATE_SALARY = 1
    STATE_HOME_VALUE = 2
    STATE_HOME_LOAN = 3
    STATE_HOME_REQ_PAYMENT = 4
    STATE_SAVE_TAX_ADV = 5
    STATE_SAVE_TAXABLE = 6

    MEG = 1.0e6

    ACTION_ELEMENTS = 4
    ACTION_HOME_LOAN = 0
    ACTION_SAVE_TAX_ADV = 1
    ACTION_SAVE_TAXABLE = 2
    ACTION_LUXURY = 3

    INFLATION = (0.015)/12.0
    INTEREST = (0.05)/12.0
    TAX_RATE = (.142)/12.0
    EXPENSES = 0.6
    INVEST_RETURN = 0.065/12.0
    SALARY_LOW = 40000.0
    SALARY_HIGH = 60000.0
    START_AGE = 18
    RETIRE_AGE = 80

    def __init__(self, goal_velocity=0):
        self.verbose = False
        self.viewer = None

        self.action_space = spaces.Box(
            low=0.0,
            high=1.0,
            shape=(SimpleGameOfLifeEnv.ACTION_ELEMENTS,),
            dtype=np.float32
        )
        self.observation_space = spaces.Box(
            low=0,
            high=2,
            shape=(SimpleGameOfLifeEnv.STATE_ELEMENTS,),
            dtype=np.float32
        )

        self.seed()
        self.reset()

        self.state_log = []

    def seed(self, seed=None):
        self.np_random, seed = seeding.np_random(seed)
        return [seed]

    def _calc_net_worth(self):
        home_value = self.state[
            SimpleGameOfLifeEnv.STATE_HOME_VALUE]
        principal = self.state[
            SimpleGameOfLifeEnv.STATE_HOME_LOAN]
        worth = home_value - principal
        worth += self.state[
            SimpleGameOfLifeEnv.STATE_SAVE_TAX_ADV]
        worth += self.state[
            SimpleGameOfLifeEnv.STATE_SAVE_TAXABLE]
        return worth

    def _eval_action(self, action, payment):
        # Calculate actions
        act_home_payment = action[
            SimpleGameOfLifeEnv.ACTION_HOME_LOAN]
        act_tax_adv_pay = action[
            SimpleGameOfLifeEnv.ACTION_SAVE_TAX_ADV]
        act_taxable = action[
            SimpleGameOfLifeEnv.ACTION_SAVE_TAXABLE]
        act_luxury = action[
            SimpleGameOfLifeEnv.ACTION_LUXURY]
        if payment <= 0:
            act_home_payment = 0
        total_act = act_home_payment + act_tax_adv_pay\
            + act_taxable + \
            act_luxury + self.expenses

        if total_act < 1e-2:
            pct_home_payment = 0
            pct_tax_adv_pay = 0
            pct_taxable = 0
            pct_luxury = 0
        else:
            pct_home_payment = act_home_payment / total_act
            pct_tax_adv_pay = act_tax_adv_pay / total_act
            pct_taxable = act_taxable / total_act
            pct_luxury = act_luxury / total_act

        return pct_home_payment, pct_tax_adv_pay, pct_taxable, pct_luxury

    def step(self, action):
        self.last_action = action
        age = self.state[SimpleGameOfLifeEnv.STATE_AGE]
        salary = self.state[SimpleGameOfLifeEnv.STATE_SALARY]
        home_value = self.state[SimpleGameOfLifeEnv.STATE_HOME_VALUE]
        principal = self.state[SimpleGameOfLifeEnv.STATE_HOME_LOAN]
        payment = self.state[SimpleGameOfLifeEnv.STATE_HOME_REQ_PAYMENT]
        net1 = self._calc_net_worth()
        remaining_salary = salary

        # Calculate actions
        pct_home_payment, pct_tax_adv_pay, pct_taxable, pct_luxury = \
            self._eval_action(action, payment)

        # Expenses
        current_expenses = salary * self.expenses
        remaining_salary -= current_expenses
        if self.verbose:
            print(f"Expenses: {current_expenses}")
            print(f"Remaining Salary: {remaining_salary}")

        # Tax advantaged deposit action
        my_tax_adv_deposit = min(salary * pct_tax_adv_pay,
                                 remaining_salary)
        # Govt CAP
        my_tax_adv_deposit = min(my_tax_adv_deposit,
                                 self.year_tax_adv_deposit_left)
        self.year_tax_adv_deposit_left -= my_tax_adv_deposit
        remaining_salary -= my_tax_adv_deposit
        # Company match
        tax_adv_deposit = my_tax_adv_deposit * 1.05
        self.state[SimpleGameOfLifeEnv.STATE_SAVE_TAX_ADV] += \
            int(tax_adv_deposit)

        if self.verbose:
            print(f"IRA Deposit: {tax_adv_deposit}")
            print(f"Remaining Salary: {remaining_salary}")

        # Tax
        remaining_salary -= remaining_salary * \
            SimpleGameOfLifeEnv.TAX_RATE
        if self.verbose:
            print(f"Tax Salary: {remaining_salary}")

        # Home payment
        actual_payment = min(salary * pct_home_payment,
                             remaining_salary)

        if principal > 0:
            ipart = principal * SimpleGameOfLifeEnv.INTEREST
            ppart = actual_payment - ipart
            principal = int(principal-ppart)
            if principal <= 0:
                principal = 0
                self.state[SimpleGameOfLifeEnv.STATE_HOME_REQ_PAYMENT] = 0
            elif actual_payment < payment:
                self.late_count += 1
                if self.late_count > 15:
                    sell = (home_value-principal)/2
                    sell -= 20000
                    sell = max(sell, 0)
                    self.state[SimpleGameOfLifeEnv.STATE_SAVE_TAXABLE] \
                        += sell
                    principal = 0
                    home_value = 0
                    self.expenses += .3
                    self.state[SimpleGameOfLifeEnv.STATE_HOME_REQ_PAYMENT] \
                        = 0
                    if self.verbose:
                        print(f"Foreclosure!!")
                else:
                    late_fee = payment * 0.1
                    principal += late_fee
                    if self.verbose:
                        print(f"Late Fee: {late_fee}")

            self.state[SimpleGameOfLifeEnv.STATE_HOME_LOAN] = principal
            remaining_salary -= actual_payment

        if self.verbose:
            print(f"Home Payment: {actual_payment}")
            print(f"Remaining Salary: {remaining_salary}")

        # Taxable savings
        actual_savings = remaining_salary * pct_taxable
        self.state[SimpleGameOfLifeEnv.STATE_SAVE_TAXABLE] \
            += actual_savings
        remaining_salary -= actual_savings

        if self.verbose:
            print(f"Tax Save: {actual_savings}")
            print(f"Remaining Salary (goes to Luxury): {remaining_salary}")

        # Investment income
        return_taxable = self.state[
            SimpleGameOfLifeEnv.STATE_SAVE_TAXABLE]\
            * self.invest_return
        return_tax_adv = self.state[
            SimpleGameOfLifeEnv.STATE_SAVE_TAX_ADV]\
            * self.invest_return

        return_taxable *= 1-SimpleGameOfLifeEnv.TAX_RATE
        self.state[SimpleGameOfLifeEnv.STATE_SAVE_TAXABLE] \
            += return_taxable
        self.state[SimpleGameOfLifeEnv.STATE_SAVE_TAX_ADV] \
            += return_tax_adv

        # Yearly events
        if age > 0 and age % 12 == 0:
            self.perform_yearly()

        # Monthly events
        self.state[SimpleGameOfLifeEnv.STATE_AGE] += 1

        # Time to retire (by age?)
        done = self.state[SimpleGameOfLifeEnv.STATE_AGE] > \
            (SimpleGameOfLifeEnv.RETIRE_AGE*12)

        # Calculate reward
        net2 = self._calc_net_worth()
        reward = net2 - net1

        # Track progress
        if self.verbose:
            print(f"Networth: {nw}")
            print(f"*** End Step {self.step_num}: State={self.state}, \
          Reward={reward}")
        self.state_log.append(self.state + [current_expenses,
                                            actual_payment,
                                            actual_savings,
                                            my_tax_adv_deposit,
                                            net2])
        self.step_num += 1

        # Normalize state and finish up
        norm_state = [x/SimpleGameOfLifeEnv.MEG for x in self.state]
        return norm_state, reward/SimpleGameOfLifeEnv.MEG, done, {}

    def perform_yearly(self):
        salary = self.state[SimpleGameOfLifeEnv.STATE_SALARY]
        home_value = self.state[SimpleGameOfLifeEnv.STATE_HOME_VALUE]

        self.inflation = SimpleGameOfLifeEnv.INTEREST + \
            self.np_random.normal(loc=0, scale=1e-2)
        self.invest_return = SimpleGameOfLifeEnv.INVEST_RETURN + \
            self.np_random.normal(loc=0, scale=1e-2)

        self.year_tax_adv_deposit_left = 19000
        self.state[SimpleGameOfLifeEnv.STATE_SALARY] = \
            int(salary * (1+self.inflation))

        self.state[SimpleGameOfLifeEnv.STATE_HOME_VALUE] \
            = int(home_value * (1+self.inflation))

    def reset(self):
        self.expenses = SimpleGameOfLifeEnv.EXPENSES
        self.late_count = 0
        self.step_num = 0
        self.last_action = [0] * SimpleGameOfLifeEnv.ACTION_ELEMENTS
        self.state = [0] * SimpleGameOfLifeEnv.STATE_ELEMENTS
        self.state_log = []
        salary = float(self.np_random.randint(
            low=SimpleGameOfLifeEnv.SALARY_LOW,
            high=SimpleGameOfLifeEnv.SALARY_HIGH))
        house_mult = self.np_random.uniform(low=1.5, high=4)
        value = round(salary*house_mult)
        p = (value*0.9)
        i = SimpleGameOfLifeEnv.INTEREST
        n = 30 * 12
        m = float(int(p * (i * (1 + i)**n) / ((1 + i)**n - 1)))
        self.state[SimpleGameOfLifeEnv.STATE_AGE] = \
            SimpleGameOfLifeEnv.START_AGE * 12
        self.state[SimpleGameOfLifeEnv.STATE_SALARY] = salary / 12.0
        self.state[SimpleGameOfLifeEnv.STATE_HOME_VALUE] = value
        self.state[SimpleGameOfLifeEnv.STATE_HOME_LOAN] = p
        self.state[SimpleGameOfLifeEnv.STATE_HOME_REQ_PAYMENT] = m
        self.year_tax_adv_deposit_left = 19000
        self.perform_yearly()
        return np.array(self.state)

    def render(self, mode='human'):
        screen_width = 600
        screen_height = 400

        img = PIL.Image.new('RGB', (600, 400))
        d = PIL.ImageDraw.Draw(img)
        font = ImageFont.load_default()
        y = 0
        _, height = d.textsize("W", font)

        age = self.state[SimpleGameOfLifeEnv.STATE_AGE]
        salary = self.state[SimpleGameOfLifeEnv.STATE_SALARY]*12
        home_value = self.state[
            SimpleGameOfLifeEnv.STATE_HOME_VALUE]
        home_loan = self.state[
            SimpleGameOfLifeEnv.STATE_HOME_LOAN]
        home_payment = self.state[
            SimpleGameOfLifeEnv.STATE_HOME_REQ_PAYMENT]
        balance_tax_adv = self.state[
            SimpleGameOfLifeEnv.STATE_SAVE_TAX_ADV]
        balance_taxable = self.state[
            SimpleGameOfLifeEnv.STATE_SAVE_TAXABLE]
        net_worth = self._calc_net_worth()

        d.text((0, y), f"Age: {age/12}", fill=(0, 255, 0))
        y += height
        d.text((0, y), f"Salary: {salary:,}", fill=(0, 255, 0))
        y += height
        d.text((0, y), f"Home Value: {home_value:,}",
               fill=(0, 255, 0))
        y += height
        d.text((0, y), f"Home Loan: {home_loan:,}",
               fill=(0, 255, 0))
        y += height
        d.text((0, y), f"Home Payment: {home_payment:,}",
               fill=(0, 255, 0))
        y += height
        d.text((0, y), f"Balance Tax Adv: {balance_tax_adv:,}",
               fill=(0, 255, 0))
        y += height
        d.text((0, y), f"Balance Taxable: {balance_taxable:,}",
               fill=(0, 255, 0))
        y += height
        d.text((0, y), f"Net Worth: {net_worth:,}", fill=(0, 255, 0))
        y += height*2

        payment = self.state[SimpleGameOfLifeEnv.STATE_HOME_REQ_PAYMENT]
        pct_home_payment, pct_tax_adv_pay, pct_taxable, pct_luxury = \
            self._eval_action(self.last_action, payment)
        d.text((0, y), f"Percent Home Payment: {pct_home_payment}",
               fill=(0, 255, 0))
        y += height
        d.text((0, y), f"Percent Tax Adv: {pct_tax_adv_pay}",
               fill=(0, 255, 0))
        y += height
        d.text((0, y), f"Percent Taxable: {pct_taxable}", fill=(0, 255, 0))
        y += height
        d.text((0, y), f"Percent Luxury: {pct_luxury}", fill=(0, 255, 0))

        return np.array(img)

    def close(self):
        pass\end{lstlisting}
\end{tcolorbox}%
You must register the environment class with TF{-}Agents before your program can use it.%
\par%
\begin{tcolorbox}[size=title,title=Code,breakable]%
\begin{lstlisting}[language=Python, upquote=true]
register(
    id='simple-game-of-life-v0',
    entry_point=f'{__name__}:SimpleGameOfLifeEnv',
)\end{lstlisting}
\end{tcolorbox}

\subsection{Testing the Environment}%
\label{subsec:TestingtheEnvironment}%
This financial planning environment is complex.  It took me some degree of testing to perfect it.  Even at the current state of this simulator, it is far from a complete financial simulator.  The primary objective of this simulator is to demonstrate creating your environment for a non{-}video game project.%
\index{SOM}%
\index{video}%
\par%
I used the following code to help test this simulator.  I ran the simulator with fixed actions and then loaded the state into a Pandas data frame for easy viewing.%
\par%
\begin{tcolorbox}[size=title,title=Code,breakable]%
\begin{lstlisting}[language=Python, upquote=true]
env_name = 'simple-game-of-life-v0'
env = gym.make(env_name)

env.reset()
done = False

i = 0
env.verbose = False
while not done:
    i += 1
    state, reward, done, _ = env.step([1, 1, 0, 0])
    env.render()

env.close()\end{lstlisting}
\end{tcolorbox}%
\begin{tcolorbox}[size=title,title=Code,breakable]%
\begin{lstlisting}[language=Python, upquote=true]
import pandas as pd

df = pd.DataFrame(env.state_log, columns=SimpleGameOfLifeEnv.STATES)
df = df.round(0)
df['age'] = df['age']/12
df['age'] = df['age'].round(2)
for col in df.columns:
    df[col] = df[col].apply(lambda x: "{:,}".format(x))

pd.set_option('display.max_columns', 7)
pd.set_option('display.max_rows', 12)
display(df)\end{lstlisting}
\tcbsubtitle[before skip=\baselineskip]{Output}%
\begin{tabular}[hbt!]{l|l|l|l|l|l|l|l}%
\hline%
&age&salary&home\_value&...&tax\_deposit&tax\_adv\_deposit&net\_worth\\%
\hline%
0&18.08&4,876&214,749&...&0.0&1,880.0&24,578.0\\%
1&18.17&4,876&214,749&...&0.0&1,875.0&25,791.0\\%
2&18.25&4,876&214,749&...&0.0&1,875.0&27,039.0\\%
3&18.33&4,876&214,749&...&0.0&1,875.0&28,321.0\\%
4&18.42&4,876&214,749&...&0.0&1,875.0&29,640.0\\%
...&...&...&...&...&...&...&...\\%
740&79.75&6,830&302,304&...&0.0&683.0&3,990,102.0\\%
741&79.83&6,830&302,304&...&0.0&683.0&3,989,629.0\\%
742&79.92&6,830&302,304&...&0.0&683.0&3,989,157.0\\%
743&80.0&6,830&302,304&...&0.0&683.0&3,988,684.0\\%
744&80.08&6,816&301,724&...&0.0&683.0&3,987,632.0\\%
\hline%
\end{tabular}%
\vspace{2mm}%
\end{tcolorbox}%
1810888.5833333335%
\par

\subsection{Hyperparameters}%
\label{subsec:Hyperparameters}%
I tuned the following hyperparameters to get a reasonable result from training the agent.  Further optimization would be beneficial.%
\index{hyperparameter}%
\index{optimization}%
\index{parameter}%
\index{training}%
\par%
\begin{tcolorbox}[size=title,title=Code,breakable]%
\begin{lstlisting}[language=Python, upquote=true]
# How long should training run?
num_iterations = 3000
# How often should the program provide an update.
log_interval = 500

# How many initial random steps, before training start, to
# collect initial data.
initial_collect_steps = 1000
# How many steps should we run each iteration to collect
# data from.
collect_steps_per_iteration = 50
# How much data should we store for training examples.
replay_buffer_max_length = 100000

batch_size = 64

# How many episodes should the program use for each evaluation.
num_eval_episodes = 100
# How often should an evaluation occur.
eval_interval = 5000\end{lstlisting}
\end{tcolorbox}

\subsection{Instantiate the Environment}%
\label{subsec:InstantiatetheEnvironment}%
We are now ready to make use of our environment.  Because we registered the environment with TF{-}Agents the program can load the environment by its name "simple{-}game{-}of{-}life{-}v".%
\par%
\begin{tcolorbox}[size=title,title=Code,breakable]%
\begin{lstlisting}[language=Python, upquote=true]
env_name = 'simple-game-of-life-v0'
#env_name = 'MountainCarContinuous-v0'
env = suite_gym.load(env_name)\end{lstlisting}
\end{tcolorbox}%
We can now have a quick look at the first state rendered. Here we can see the random salary and home values are chosen for an agent.  The learned policy must be able to consider different starting salaries and home values and find an appropriate strategy.%
\index{random}%
\par%
\begin{tcolorbox}[size=title,title=Code,breakable]%
\begin{lstlisting}[language=Python, upquote=true]
env.reset()
PIL.Image.fromarray(env.render())\end{lstlisting}
\tcbsubtitle[before skip=\baselineskip]{Output}%
\includegraphics[width=4in]{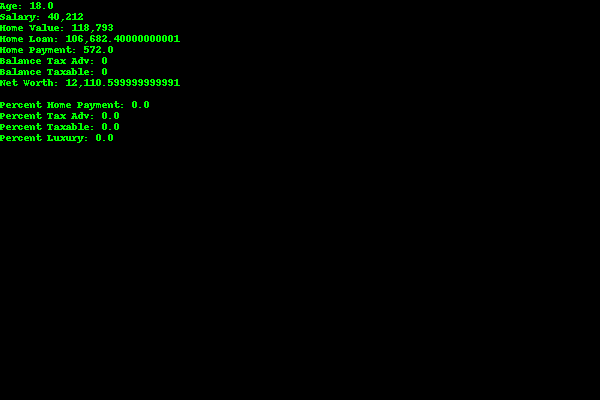}%
\end{tcolorbox}%
Just as before, the program instantiates two environments: one for training and one for evaluation.%
\index{training}%
\par%
\begin{tcolorbox}[size=title,title=Code,breakable]%
\begin{lstlisting}[language=Python, upquote=true]
train_py_env = suite_gym.load(env_name)
eval_py_env = suite_gym.load(env_name)

train_env = tf_py_environment.TFPyEnvironment(train_py_env)
eval_env = tf_py_environment.TFPyEnvironment(eval_py_env)\end{lstlisting}
\end{tcolorbox}%
You might be wondering why a DQN does not support continuous actions. This limitation is that the DQN algorithm maps each action as an output neuron. Each of these neurons predicts the likely future reward for taking each action. The algorithm knows the future rewards for each particular action. Generally, the DQN agent will perform the action that has the highest reward. However, because a continuous number represented in a computer has an effectively infinite number of possible values, it is not possible to calculate a future reward estimate for all of them.%
\index{algorithm}%
\index{continuous}%
\index{neuron}%
\index{output}%
\index{output neuron}%
\index{predict}%
\par
\begin{figure}[h]%
\centering%
\includegraphics[width=4in]{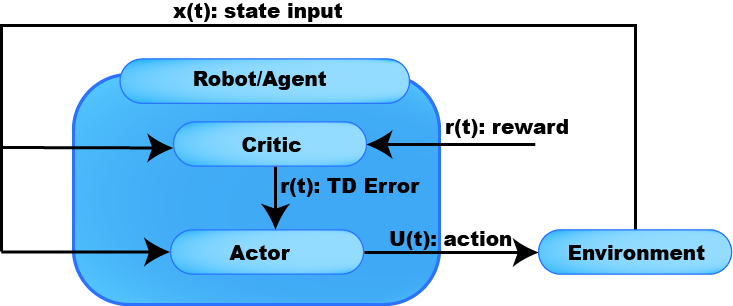}%
\caption{Actor Critic Model}%
\label{12.DDPG}%
\end{figure}

%
We will use the Deep Deterministic Policy Gradients (DDPG) algorithm to provide a continuous action space.%
\index{algorithm}%
\index{continuous}%
\index{gradient}%
\cite{lillicrap2015continuous}%
This technique uses two neural networks. The first neural network, called an actor, acts as the agent and predicts the expected reward for a given value of the action. The second neural network, called a critic, is trained to predict the accuracy of the actor{-}network. Training two neural networks in parallel that operate adversarially is a  popular technique. Earlier in this course, we saw that Generative Adversarial Networks (GAN) used a similar method. Figure \ref{12.DDPG} shows the structure of the DDPG network that we will use.%
\index{GAN}%
\index{neural network}%
\index{predict}%
\index{training}%
\par%

\par%
The environment provides the same input ($x(t)$) for each time step to both the actor and critic networks. The temporal difference error ($r(t)$) reports the difference between the estimated reward and the actual reward at any given state or time step.%
\index{error}%
\index{input}%
\index{temporal}%
\par%
The following code creates the actor and critic neural networks.%
\index{neural network}%
\par%
\begin{tcolorbox}[size=title,title=Code,breakable]%
\begin{lstlisting}[language=Python, upquote=true]
actor_fc_layers = (400, 300)
critic_obs_fc_layers = (400,)
critic_action_fc_layers = None
critic_joint_fc_layers = (300,)
ou_stddev = 0.2
ou_damping = 0.15
target_update_tau = 0.05
target_update_period = 5
dqda_clipping = None
td_errors_loss_fn = tf.compat.v1.losses.huber_loss
gamma = 0.995
reward_scale_factor = 1.0
gradient_clipping = None

actor_learning_rate = 1e-4
critic_learning_rate = 1e-3
debug_summaries = False
summarize_grads_and_vars = False

global_step = tf.compat.v1.train.get_or_create_global_step()

actor_net = actor_network.ActorNetwork(
    train_env.time_step_spec().observation,
    train_env.action_spec(),
    fc_layer_params=actor_fc_layers,
)

critic_net_input_specs = (train_env.time_step_spec().observation,
                          train_env.action_spec())

critic_net = critic_network.CriticNetwork(
    critic_net_input_specs,
    observation_fc_layer_params=critic_obs_fc_layers,
    action_fc_layer_params=critic_action_fc_layers,
    joint_fc_layer_params=critic_joint_fc_layers,
)

tf_agent = ddpg_agent.DdpgAgent(
    train_env.time_step_spec(),
    train_env.action_spec(),
    actor_network=actor_net,
    critic_network=critic_net,
    actor_optimizer=tf.compat.v1.train.AdamOptimizer(
        learning_rate=actor_learning_rate),
    critic_optimizer=tf.compat.v1.train.AdamOptimizer(
        learning_rate=critic_learning_rate),
    ou_stddev=ou_stddev,
    ou_damping=ou_damping,
    target_update_tau=target_update_tau,
    target_update_period=target_update_period,
    dqda_clipping=dqda_clipping,
    td_errors_loss_fn=td_errors_loss_fn,
    gamma=gamma,
    reward_scale_factor=reward_scale_factor,
    gradient_clipping=gradient_clipping,
    debug_summaries=debug_summaries,
    summarize_grads_and_vars=summarize_grads_and_vars,
    train_step_counter=global_step)
tf_agent.initialize()\end{lstlisting}
\end{tcolorbox}

\subsection{Metrics and Evaluation}%
\label{subsec:MetricsandEvaluation}%
Just as in previous examples, we will compute the average return over several episodes to evaluate performance.%
\par%
\begin{tcolorbox}[size=title,title=Code,breakable]%
\begin{lstlisting}[language=Python, upquote=true]
def compute_avg_return(environment, policy, num_episodes=10):

    total_return = 0.0
    for _ in range(num_episodes):

        time_step = environment.reset()
        episode_return = 0.0

        while not time_step.is_last():
            action_step = policy.action(time_step)
            time_step = environment.step(action_step.action)
            episode_return += time_step.reward
        total_return += episode_return

    avg_return = total_return / num_episodes
    return avg_return.numpy()[0]


# See also the metrics module for standard implementations of
# different metrics.
# https://github.com/tensorflow/agents/tree/master/tf_agents/metrics\end{lstlisting}
\end{tcolorbox}

\subsection{Data Collection}%
\label{subsec:DataCollection}%
Now execute the random policy in the environment for a few steps, recording the data in the replay buffer.%
\index{random}%
\par%
\begin{tcolorbox}[size=title,title=Code,breakable]%
\begin{lstlisting}[language=Python, upquote=true]
def collect_step(environment, policy, buffer):
    time_step = environment.current_time_step()
    action_step = policy.action(time_step)
    next_time_step = \
        environment.step(action_step.action)
    traj = trajectory.from_transition(\
        time_step, action_step,\
        next_time_step)

    # Add trajectory to the replay buffer
    buffer.add_batch(traj)


def collect_data(env, policy, buffer, steps):
    for _ in range(steps):
        collect_step(env, policy, buffer)


random_policy = random_tf_policy.RandomTFPolicy(\
    train_env.time_step_spec(),\
    train_env.action_spec())

replay_buffer = tf_uniform_replay_buffer.TFUniformReplayBuffer(
    data_spec=tf_agent.collect_data_spec,
    batch_size=train_env.batch_size,
    max_length=replay_buffer_max_length)

collect_data(train_env, random_policy, replay_buffer, steps=100)

# Dataset generates trajectories with shape [Bx2x...]
dataset = replay_buffer.as_dataset(
    num_parallel_calls=3,
    sample_batch_size=batch_size,
    num_steps=2).prefetch(3)\end{lstlisting}
\tcbsubtitle[before skip=\baselineskip]{Output}%
\begin{lstlisting}[upquote=true]
WARNING:tensorflow:From /usr/local/lib/python3.7/dist-
packages/tensorflow/python/autograph/impl/api.py:377:
ReplayBuffer.get_next (from tf_agents.replay_buffers.replay_buffer) is
deprecated and will be removed in a future version.
Instructions for updating:
Use `as_dataset(..., single_deterministic_pass=False) instead.
\end{lstlisting}
\end{tcolorbox}

\subsection{Training the agent}%
\label{subsec:Trainingtheagent}%
We are now ready to train the agent. Depending on how many episodes you wish to run through, this process can take many hours. This code will update on both the loss and average return as training occurs. As training becomes more successful, the average return should increase. The losses reported reflect the average loss for individual training batches.%
\index{ROC}%
\index{ROC}%
\index{training}%
\index{training batch}%
\par%
\begin{tcolorbox}[size=title,title=Code,breakable]%
\begin{lstlisting}[language=Python, upquote=true]
iterator = iter(dataset)

# (Optional) Optimize by wrapping some of the code in a graph using
# TF function.
tf_agent.train = common.function(tf_agent.train)

# Reset the train step
tf_agent.train_step_counter.assign(0)

# Evaluate the agent's policy once before training.
avg_return = compute_avg_return(eval_env, tf_agent.policy,
                                num_eval_episodes)
returns = [avg_return]

for _ in range(num_iterations):

    # Collect a few steps using collect_policy and 
    # save to the replay buffer.
    for _ in range(collect_steps_per_iteration):
        collect_step(train_env, tf_agent.collect_policy, replay_buffer)

    # Sample a batch of data from the buffer and update the
    # agent's network.
    experience, unused_info = next(iterator)
    train_loss = tf_agent.train(experience).loss

    step = tf_agent.train_step_counter.numpy()

    if step % log_interval == 0:
        print('step = {0}: loss = {1}'.format(step, train_loss))

    if step % eval_interval == 0:
        avg_return = compute_avg_return(eval_env, tf_agent.policy,
                                        num_eval_episodes)
        print('step = {0}: Average Return = {1}'.format(step, avg_return))
        returns.append(avg_return)\end{lstlisting}
\tcbsubtitle[before skip=\baselineskip]{Output}%
\begin{lstlisting}[upquote=true]
step = 500: loss = 0.00016351199883501977
step = 1000: loss = 6.34381067357026e-05
step = 1500: loss = 0.0012666243128478527
step = 2000: loss = 0.00041321030585095286
step = 2500: loss = 0.0006321941618807614
step = 3000: loss = 0.0006611005519516766
\end{lstlisting}
\end{tcolorbox}

\subsection{Visualization}%
\label{subsec:Visualization}%
The notebook can plot the average return over training iterations. The average return should increase as the program performs more training iterations.%
\index{iteration}%
\index{training}%
\par

\subsection{Videos}%
\label{subsec:Videos}%
We use the following functions to produce video in Jupyter notebook. As the person moves through their career, they focus on paying off the house and tax advantage investing.%
\index{video}%
\par%
\begin{tcolorbox}[size=title,title=Code,breakable]%
\begin{lstlisting}[language=Python, upquote=true]
def embed_mp4(filename):
    """Embeds an mp4 file in the notebook."""
    video = open(filename, 'rb').read()
    b64 = base64.b64encode(video)
    tag = '''
  <video width="640" height="480" controls>
    <source src="data:video/mp4;base64,{0}" type="video/mp4">
  Your browser does not support the video tag.
  </video>'''.format(b64.decode())

    return IPython.display.HTML(tag)


def create_policy_eval_video(policy, filename, num_episodes=5, fps=30):
    filename = filename + ".mp4"
    with imageio.get_writer(filename, fps=fps) as video:
        for _ in range(num_episodes):
            time_step = eval_env.reset()
            video.append_data(eval_py_env.render())
            while not time_step.is_last():
                action_step = policy.action(time_step)
                time_step = eval_env.step(action_step.action)
                video.append_data(eval_py_env.render())
    return embed_mp4(filename)


create_policy_eval_video(tf_agent.policy, "trained-agent")\end{lstlisting}
\end{tcolorbox}

\chapter{Advanced/Other Topics}%
\label{chap:Advanced/OtherTopics}%
\section{Part 13.1: Flask and Deep Learning Web Services}%
\label{sec:Part13.1FlaskandDeepLearningWebServices}%
Suppose you would like to create websites based on neural networks. In that case, we must expose the neural network so that Python and other programming languages can efficiently execute. The usual means for such integration is a web service. One of the most popular libraries for doing this in Python is%
\index{neural network}%
\index{Python}%
\href{https://palletsprojects.com/p/flask/}{ Flask}%
. This library allows you to quickly deploy your Python applications, including TensorFlow, as web services.%
\index{Python}%
\index{TensorFlow}%
\par%
Neural network deployment is a complex process, usually carried out by a company's%
\index{neural network}%
\index{ROC}%
\index{ROC}%
\href{https://en.wikipedia.org/wiki/Information_technology}{ Information Technology (IT) group}%
. When large numbers of clients must access your model, scalability becomes essential. The cloud usually handles this. The designers of Flask did not design for high{-}volume systems. When deployed to production, you will wrap models in%
\index{Flask}%
\index{model}%
\href{https://gunicorn.org/}{ Gunicorn }%
or TensorFlow Serving. We will discuss high{-}volume cloud deployment in the next section. Everything presented in this part ith Flask is directly compatible with the higher volume Gunicorn system. When early in the development process, it is common to use Flask directly.%
\index{Flask}%
\index{ROC}%
\index{ROC}%
\index{TensorFlow}%
\par%
\subsection{Flask Hello World}%
\label{subsec:FlaskHelloWorld}%
Flask is the server, and Jupyter usually fills the role of the client. It is uncommon to run Flask from a Jupyter notebook. However, we can run a simple web service from Jupyter. We will quickly move beyond this and deploy using a Python script (.py). Because we must use .py files, it won't be easy to use Google CoLab, as you will be running from the command line. For now, let's execute a Flask web container in Jupyter.%
\index{Flask}%
\index{Python}%
\par%
\begin{tcolorbox}[size=title,title=Code,breakable]%
\begin{lstlisting}[language=Python, upquote=true]
from werkzeug.wrappers import Request, Response
from flask import Flask

app = Flask(__name__)

@app.route("/")
def hello():
    return "Hello World!"

if __name__ == '__main__':
    from werkzeug.serving import run_simple
    run_simple('localhost', 9000, app)\end{lstlisting}
\end{tcolorbox}%
This program starts a web service on port 9000 of your computer.  This cell will remain running (appearing locked up).  However, it is merely waiting for browsers to connect.  If you point your browser at the following URL, you will interact with the Flask web service.%
\index{Flask}%
\par%
\begin{itemize}[noitemsep]%
\item%
http://localhost:9000/%
\end{itemize}%
You should see Hello World displayed.%
\par

\subsection{MPG Flask}%
\label{subsec:MPGFlask}%
Usually, you will interact with a web service through JSON.  A program will send a JSON message to your Flask application, and your Flask application will return a JSON.  Later, in module 13.3, we will see how to attach this web service to a web application that you can interact with through a browser.  We will create a Flask wrapper for a neural network that predicts the miles per gallon.  The sample JSON will look like this.%
\index{Flask}%
\index{neural network}%
\index{predict}%
\par%
\begin{tcolorbox}[size=title,breakable]%
\begin{lstlisting}[upquote=true]
{
  "cylinders": 8, 
  "displacement": 300,
  "horsepower": 78, 
  "weight": 3500,
  "acceleration": 20, 
  "year": 76,
  "origin": 1
}
\end{lstlisting}
\end{tcolorbox}%
We will see two different means of POSTing this JSON data to our web server.  First, we will use a utility called%
\href{https://www.getpostman.com/}{ POSTman}%
.  Secondly, we will use Python code to construct the JSON message and interact with Flask.%
\index{Flask}%
\index{Python}%
\par%
First, it is necessary to train a neural network with the MPG dataset.  This technique is very similar to what we've done many times before.  However, we will save the neural network so that we can load it later.  We do not want to have Flask train the neural network.  We wish to have the neural network already trained and deploy the already prepared .H5 file to save the neural network.  The following code trains an MPG neural network.%
\index{dataset}%
\index{Flask}%
\index{neural network}%
\par%
\begin{tcolorbox}[size=title,title=Code,breakable]%
\begin{lstlisting}[language=Python, upquote=true]
from tensorflow.keras.models import Sequential
from tensorflow.keras.layers import Dense, Activation
from sklearn.model_selection import train_test_split
from tensorflow.keras.callbacks import EarlyStopping
import pandas as pd
import io
import os
import requests
import numpy as np
from sklearn import metrics

df = pd.read_csv(
    "https://data.heatonresearch.com/data/t81-558/auto-mpg.csv", 
    na_values=['NA', '?'])

cars = df['name']

# Handle missing value
df['horsepower'] = df['horsepower'].fillna(df['horsepower'].median())

# Pandas to Numpy
x = df[['cylinders', 'displacement', 'horsepower', 'weight',
       'acceleration', 'year', 'origin']].values
y = df['mpg'].values # regression

# Split into validation and training sets
x_train, x_test, y_train, y_test = train_test_split(    
    x, y, test_size=0.25, random_state=42)

# Build the neural network
model = Sequential()
model.add(Dense(25, input_dim=x.shape[1], activation='relu')) # Hidden 1
model.add(Dense(10, activation='relu')) # Hidden 2
model.add(Dense(1)) # Output
model.compile(loss='mean_squared_error', optimizer='adam')

monitor = EarlyStopping(monitor='val_loss', min_delta=1e-3, patience=5, \
        verbose=1, mode='auto',\
        restore_best_weights=True)
model.fit(x_train,y_train,validation_data=(x_test,y_test),\
          callbacks=[monitor],verbose=2,epochs=1000)\end{lstlisting}
\tcbsubtitle[before skip=\baselineskip]{Output}%
\begin{lstlisting}[upquote=true]
Train on 298 samples, validate on 100 samples
...
298/298 - 0s - loss: 39.0555 - val_loss: 31.4981
Epoch 52/1000
Restoring model weights from the end of the best epoch.
298/298 - 0s - loss: 37.9472 - val_loss: 32.6139
Epoch 00052: early stopping
\end{lstlisting}
\end{tcolorbox}%
Next, we evaluate the score.  This evaluation is more of a sanity check to ensure the code above worked as expected.%
\par%
\begin{tcolorbox}[size=title,title=Code,breakable]%
\begin{lstlisting}[language=Python, upquote=true]
pred = model.predict(x_test)
# Measure RMSE error.  RMSE is common for regression.
score = np.sqrt(metrics.mean_squared_error(pred,y_test))
print(f"After load score (RMSE): {score}")\end{lstlisting}
\tcbsubtitle[before skip=\baselineskip]{Output}%
\begin{lstlisting}[upquote=true]
After load score (RMSE): 5.465193688130732
\end{lstlisting}
\end{tcolorbox}%
Next, we save the neural network to a .H5 file.%
\index{neural network}%
\par%
\begin{tcolorbox}[size=title,title=Code,breakable]%
\begin{lstlisting}[language=Python, upquote=true]
model.save(os.path.join("./dnn/","mpg_model.h5"))\end{lstlisting}
\end{tcolorbox}%
We want the Flask web service to check that the input JSON is valid.  To do this, we need to know what values we expect and their logical ranges.  The following code outputs the expected fields and their ranges, and packages all of this information into a JSON object that you should copy to the Flask web application.  This code allows us to validate the incoming JSON requests.%
\index{Flask}%
\index{input}%
\index{output}%
\par%
\begin{tcolorbox}[size=title,title=Code,breakable]%
\begin{lstlisting}[language=Python, upquote=true]
cols = [x for x in df.columns if x not in ('mpg','name')]

print("{")
for i,name in enumerate(cols):
    print(f'"{name}":{{"min":{df[name].min()},\
          "max":{df[name].max()}}}{"," if i<(len(cols)-1) else ""}')
print("}")\end{lstlisting}
\tcbsubtitle[before skip=\baselineskip]{Output}%
\begin{lstlisting}[upquote=true]
{
"cylinders":{"min":3,"max":8},
"displacement":{"min":68.0,"max":455.0},
"horsepower":{"min":46.0,"max":230.0},
"weight":{"min":1613,"max":5140},
"acceleration":{"min":8.0,"max":24.8},
"year":{"min":70,"max":82},
"origin":{"min":1,"max":3}
}
\end{lstlisting}
\end{tcolorbox}%
Finally, we set up the Python code to call the model for a single car and get a prediction.  You should also copy this code to the Flask web application.%
\index{Flask}%
\index{model}%
\index{predict}%
\index{Python}%
\par%
\begin{tcolorbox}[size=title,title=Code,breakable]%
\begin{lstlisting}[language=Python, upquote=true]
import os
from tensorflow.keras.models import load_model
import numpy as np

model = load_model(os.path.join("./dnn/","mpg_model.h5"))
x = np.zeros( (1,7) )

x[0,0] = 8 # 'cylinders', 
x[0,1] = 400 # 'displacement', 
x[0,2] = 80 # 'horsepower', 
x[0,3] = 2000 # 'weight',
x[0,4] = 19 # 'acceleration', 
x[0,5] = 72 # 'year', 
x[0,6] = 1 # 'origin'


pred = model.predict(x)
float(pred[0])\end{lstlisting}
\tcbsubtitle[before skip=\baselineskip]{Output}%
\begin{lstlisting}[upquote=true]
6.212100505828857
\end{lstlisting}
\end{tcolorbox}%
The completed web application can be found here:%
\par%
\begin{itemize}[noitemsep]%
\item%
\href{./py/mpg_server_1.py}{mpg\_server\_1.py}%
\end{itemize}%
You can run this server from the command line with the following command:%
\par%
\begin{tcolorbox}[size=title,breakable]%
\begin{lstlisting}[upquote=true]
python mpg_server_1.py
\end{lstlisting}
\end{tcolorbox}%
If you are using a virtual environment (described in Module 1.1), use the%
\textbf{\texttt{ activate tensorflow }}%
command for Windows or%
\textbf{\texttt{ source activate tensorflow }}%
for Mac before executing the above command.%
\par

\subsection{Flask MPG Client}%
\label{subsec:FlaskMPGClient}%
Now that we have a web service running, we would like to access it.  This server is a bit more complicated than the "Hello World" web server we first saw in this part.  The request to display was an HTTP GET.  We must now do an HTTP POST.  To accomplish access to a web service, you must use a client.  We will see how to use%
\href{https://www.getpostman.com/}{ PostMan }%
and directly through a Python program in Jupyter.%
\index{Python}%
\par%
We will begin with PostMan.  If you have not already done so, install PostMan.%
\par%
To successfully use PostMan to query your web service, you must enter the following settings:%
\par
\begin{figure}[h]%
\centering%
\includegraphics[width=4in]{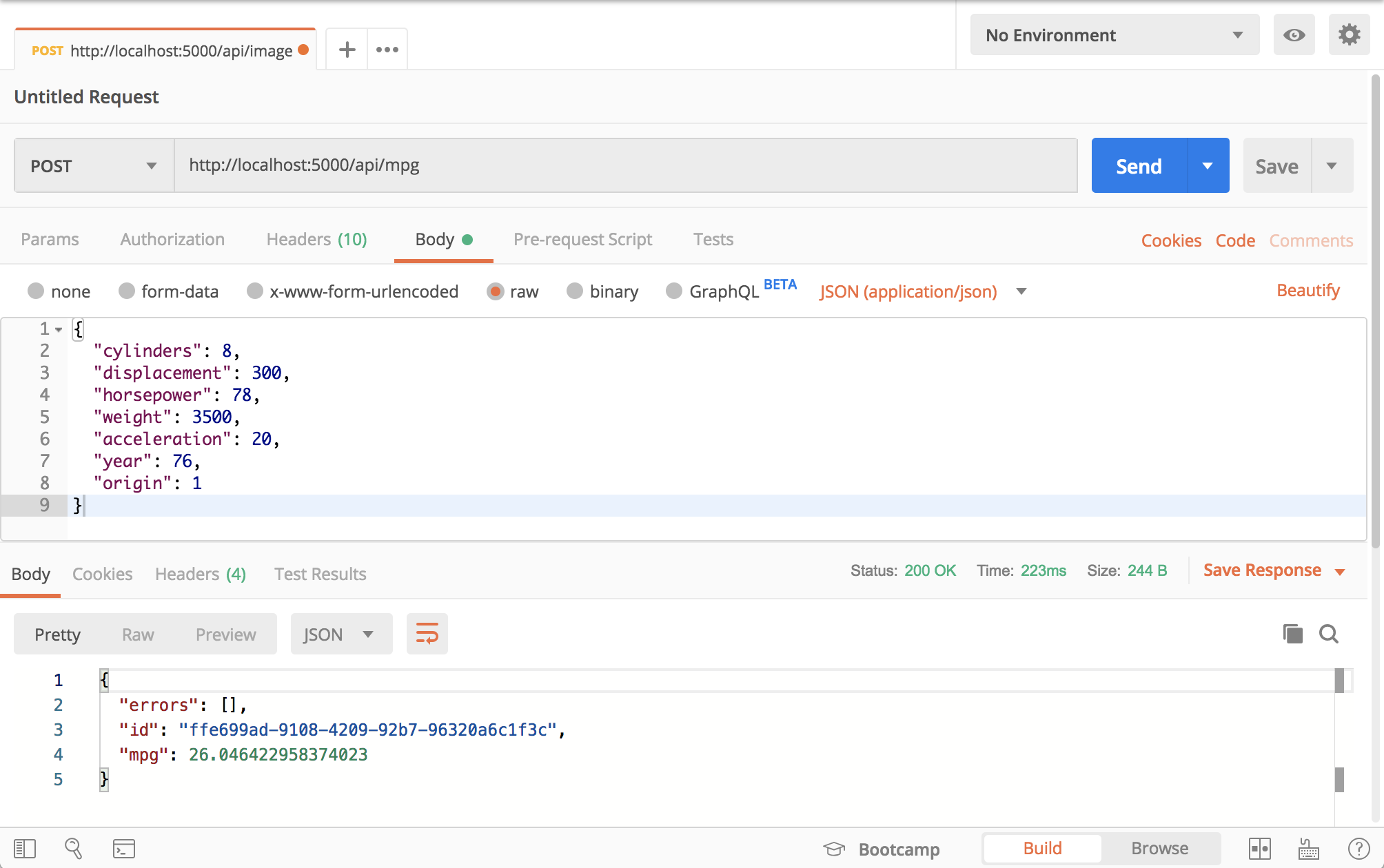}%
\caption{PostMan JSON}%
\label{13.PM}%
\end{figure}

%
\begin{itemize}[noitemsep]%
\item%
POST Request to http://localhost:5000/api/mpg%
\item%
RAW JSON and paste in JSON from above%
\item%
Click Send and you should get a correct result%
\end{itemize}%
Figure \ref{13.PM} shows a successful result.%
\par%

\par%
This same process can be done programmatically in Python.%
\index{Python}%
\index{ROC}%
\index{ROC}%
\par%
\begin{tcolorbox}[size=title,title=Code,breakable]%
\begin{lstlisting}[language=Python, upquote=true]
import requests

json = {
  "cylinders": 8, 
  "displacement": 300,
  "horsepower": 78, 
  "weight": 3500,
  "acceleration": 20, 
  "year": 76,
  "origin": 1
}

r = requests.post("http://localhost:5000/api/mpg",json=json)
if r.status_code == 200:
    print("Success: {}".format(r.text))
else: print("Failure: {}".format(r.text))\end{lstlisting}
\tcbsubtitle[before skip=\baselineskip]{Output}%
\begin{lstlisting}[upquote=true]
Success: {
  "errors": [],
  "id": "643d027e-554f-4401-ba5f-78592ae7e070",
  "mpg": 23.885438919067383
}
\end{lstlisting}
\end{tcolorbox}

\subsection{Images and Web Services}%
\label{subsec:ImagesandWebServices}%
We can also accept images from web services. We will create a web service that accepts images and classifies them using MobileNet. You will follow the same process; load your network as we did for the MPG example. You can find the completed web service can here:%
\index{ROC}%
\index{ROC}%
\par%
\href{./py/image_server_1.py}{image\_server\_1.py}%
\par%
You can run this server from the command line with:%
\par%
\begin{tcolorbox}[size=title,breakable]%
\begin{lstlisting}[upquote=true]
python mpg_server_1.py
\end{lstlisting}
\end{tcolorbox}%
If you are using a virtual environment (described in Module 1.1), use the%
\textbf{\texttt{ activate tensorflow }}%
command for Windows or%
\textbf{\texttt{ source activate tensorflow }}%
for Mac before executing the above command.%
\par%
To successfully use PostMan to query your web service, you must enter the following settings:%
\par
\begin{figure}[h]%
\centering%
\includegraphics[width=4in]{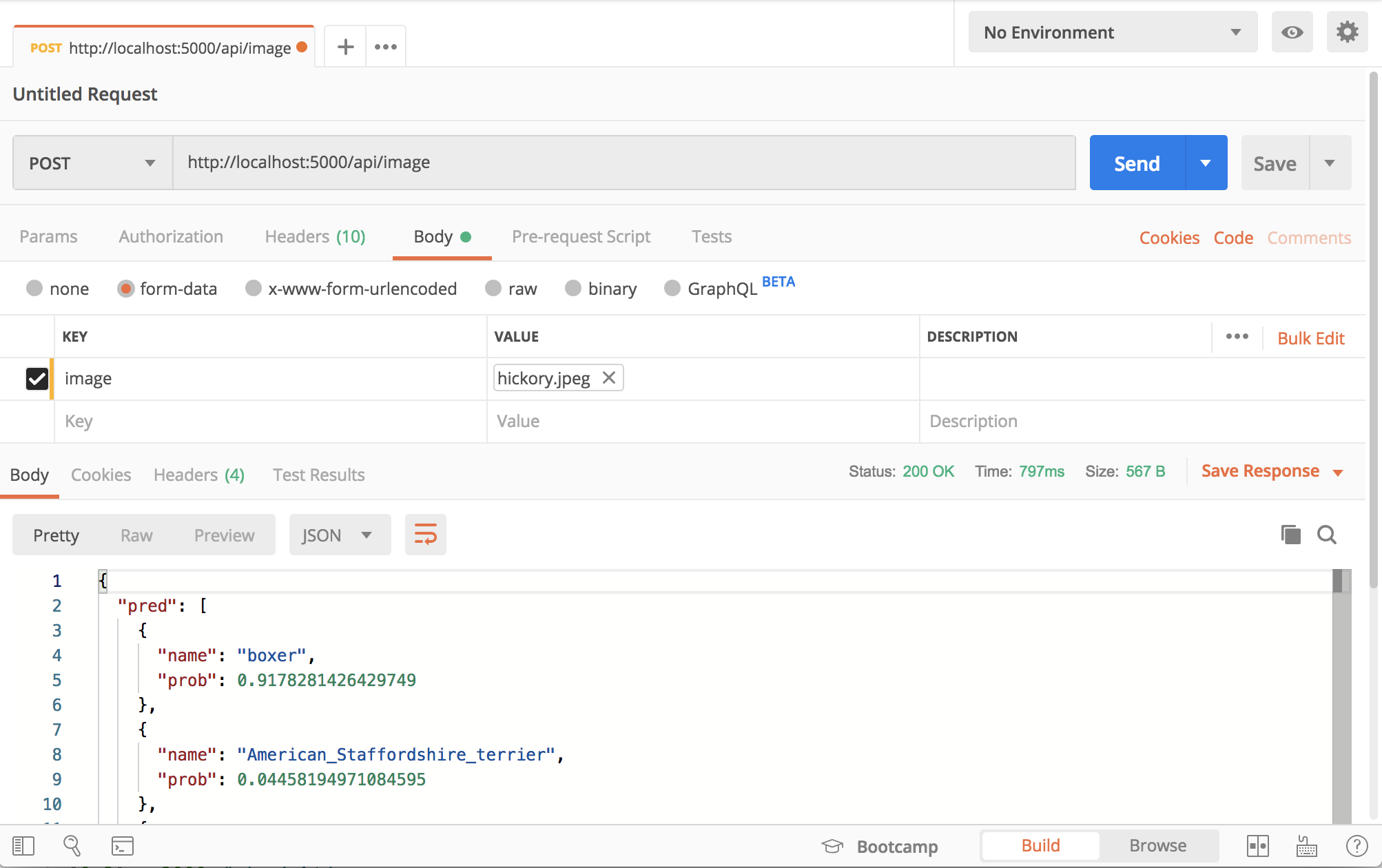}%
\caption{PostMan Images}%
\label{13.PMI}%
\end{figure}

%
\begin{itemize}[noitemsep]%
\item%
POST Request to http://localhost:5000/api/image%
\item%
Use "Form Data" and create one entry named "image" that is a file. Choose an image file to classify.%
\item%
Click Send, and you should get a correct result%
\end{itemize}%
Figure \ref{13.PMI} shows a successful result.%
\par%

\par%
This same process can be done programmatically in Python.%
\index{Python}%
\index{ROC}%
\index{ROC}%
\par%
\begin{tcolorbox}[size=title,title=Code,breakable]%
\begin{lstlisting}[language=Python, upquote=true]
import requests
response = requests.post('http://localhost:5000/api/image', files=\
        dict(image=('hickory.jpeg',open('./photos/hickory.jpeg','rb'))))
if response.status_code == 200:
    print("Success: {}".format(response.text))
else: print("Failure: {}".format(response.text))\end{lstlisting}
\tcbsubtitle[before skip=\baselineskip]{Output}%
\begin{lstlisting}[upquote=true]
Success: {
  "pred": [
    {
      "name": "boxer",
      "prob": 0.9178281426429749
    },
    {
      "name": "American_Staffordshire_terrier",
      "prob": 0.04458194971084595
    },
    {
      "name": "French_bulldog",
      "prob": 0.018736232072114944
    },
    {

...

      "name": "pug",
      "prob": 0.0009862519800662994
    }
  ]
}
\end{lstlisting}
\end{tcolorbox}

\section{Part 13.2: Interrupting and Continuing Training}%
\label{sec:Part13.2InterruptingandContinuingTraining}%
We would train our Keras models in one pass in an ideal world, utilizing as much GPU and CPU power as we need. The world in which we train our models is anything but ideal. In this part, we will see that we can stop and continue and even adjust training at later times. We accomplish this continuation with checkpoints. We begin by creating several utility functions. The first utility generates an output directory that has a unique name. This technique allows us to organize multiple runs of our experiment. We provide the Logger class to route output to a log file contained in the output directory.%
\index{GAN}%
\index{GPU}%
\index{GPU}%
\index{Keras}%
\index{model}%
\index{output}%
\index{training}%
\par%
\begin{tcolorbox}[size=title,title=Code,breakable]%
\begin{lstlisting}[language=Python, upquote=true]
import os
import re
import sys
import time
import numpy as np
from typing import Any, List, Tuple, Union
from tensorflow.keras.datasets import mnist
from tensorflow.keras import backend as K
import tensorflow as tf
import tensorflow.keras
import tensorflow as tf
from tensorflow.keras.callbacks import EarlyStopping, \
  LearningRateScheduler, ModelCheckpoint
from tensorflow.keras import regularizers
from tensorflow.keras.models import Sequential
from tensorflow.keras.layers import Dense, Dropout, Flatten
from tensorflow.keras.layers import Conv2D, MaxPooling2D
from tensorflow.keras.models import load_model
import pickle

def generate_output_dir(outdir, run_desc):
    prev_run_dirs = []
    if os.path.isdir(outdir):
        prev_run_dirs = [x for x in os.listdir(outdir) if os.path.isdir(\
            os.path.join(outdir, x))]
    prev_run_ids = [re.match(r'^\d+', x) for x in prev_run_dirs]
    prev_run_ids = [int(x.group()) for x in prev_run_ids if x is not None]
    cur_run_id = max(prev_run_ids, default=-1) + 1
    run_dir = os.path.join(outdir, f'{cur_run_id:05d}-{run_desc}')
    assert not os.path.exists(run_dir)
    os.makedirs(run_dir)
    return run_dir

# From StyleGAN2
class Logger(object):
    """Redirect stderr to stdout, optionally print stdout to a file, and 
    optionally force flushing on both stdout and the file."""

    def __init__(self, file_name: str = None, file_mode: str = "w", \
                 should_flush: bool = True):
        self.file = None

        if file_name is not None:
            self.file = open(file_name, file_mode)

        self.should_flush = should_flush
        self.stdout = sys.stdout
        self.stderr = sys.stderr

        sys.stdout = self
        sys.stderr = self

    def __enter__(self) -> "Logger":
        return self

    def __exit__(self, exc_type: Any, exc_value: Any, \
                 traceback: Any) -> None:
        self.close()

    def write(self, text: str) -> None:
        """Write text to stdout (and a file) and optionally flush."""
        if len(text) == 0: 
            return

        if self.file is not None:
            self.file.write(text)

        self.stdout.write(text)

        if self.should_flush:
            self.flush()

    def flush(self) -> None:
        """Flush written text to both stdout and a file, if open."""
        if self.file is not None:
            self.file.flush()

        self.stdout.flush()

    def close(self) -> None:
        """Flush, close possible files, and remove 
            stdout/stderr mirroring."""
        self.flush()

        # if using multiple loggers, prevent closing in wrong order
        if sys.stdout is self:
            sys.stdout = self.stdout
        if sys.stderr is self:
            sys.stderr = self.stderr

        if self.file is not None:
            self.file.close()

def obtain_data():
    (x_train, y_train), (x_test, y_test) = mnist.load_data()
    print("Shape of x_train: {}".format(x_train.shape))
    print("Shape of y_train: {}".format(y_train.shape))
    print()
    print("Shape of x_test: {}".format(x_test.shape))
    print("Shape of y_test: {}".format(y_test.shape))

    # input image dimensions
    img_rows, img_cols = 28, 28
    if K.image_data_format() == 'channels_first':
        x_train = x_train.reshape(x_train.shape[0], 1, img_rows, img_cols)
        x_test = x_test.reshape(x_test.shape[0], 1, img_rows, img_cols)
        input_shape = (1, img_rows, img_cols)
    else:
        x_train = x_train.reshape(x_train.shape[0], img_rows, img_cols, 1)
        x_test = x_test.reshape(x_test.shape[0], img_rows, img_cols, 1)
        input_shape = (img_rows, img_cols, 1)
    x_train = x_train.astype('float32')
    x_test = x_test.astype('float32')
    x_train /= 255
    x_test /= 255
    print('x_train shape:', x_train.shape)
    print("Training samples: {}".format(x_train.shape[0]))
    print("Test samples: {}".format(x_test.shape[0]))
    # convert class vectors to binary class matrices
    y_train = tf.keras.utils.to_categorical(y_train, num_classes)
    y_test = tf.keras.utils.to_categorical(y_test, num_classes)
    
    return input_shape, x_train, y_train, x_test, y_test\end{lstlisting}
\end{tcolorbox}%
We define the basic training parameters and where we wish to write the output.%
\index{output}%
\index{parameter}%
\index{training}%
\par%
\begin{tcolorbox}[size=title,title=Code,breakable]%
\begin{lstlisting}[language=Python, upquote=true]
outdir = "./data/"
run_desc = "test-train"
batch_size = 128
num_classes = 10

run_dir = generate_output_dir(outdir, run_desc)
print(f"Results saved to: {run_dir}")\end{lstlisting}
\tcbsubtitle[before skip=\baselineskip]{Output}%
\begin{lstlisting}[upquote=true]
Results saved to: ./data/00000-test-train
\end{lstlisting}
\end{tcolorbox}%
Keras provides a prebuilt checkpoint class named%
\index{Keras}%
\textbf{ ModelCheckpoint }%
that contains most of our desired functionality. This built{-}in class can save the model's state repeatedly as training progresses. Stopping neural network training is not always a controlled event. Sometimes this stoppage can be abrupt, such as a power failure or a network resource shutting down. If Microsoft Windows is your operating system of choice, your training can also be interrupted by a high{-}priority system update. Because of all of this uncertainty, it is best to save your model at regular intervals. This process is similar to saving a game at critical checkpoints, so you do not have to start over if something terrible happens to your avatar in the game.%
\index{model}%
\index{neural network}%
\index{ROC}%
\index{ROC}%
\index{SOM}%
\index{training}%
\par%
We will create our checkpoint class, named%
\textbf{ MyModelCheckpoint}%
. In addition to saving the model, we also save the state of the training infrastructure. Why save the training infrastructure, in addition to the weights? This technique eases the transition back into training for the neural network and will be more efficient than a cold start.%
\index{model}%
\index{neural network}%
\index{training}%
\par%
Consider if you interrupted your college studies after the first year. Sure, your brain (the neural network) will retain all the knowledge. But how much rework will you have to do? Your transcript at the university is like the training parameters. It ensures you do not have to start over when you come back.%
\index{neural network}%
\index{parameter}%
\index{training}%
\par%
\begin{tcolorbox}[size=title,title=Code,breakable]%
\begin{lstlisting}[language=Python, upquote=true]
class MyModelCheckpoint(ModelCheckpoint):
  def __init__(self, *args, **kwargs):
    super().__init__(*args, **kwargs)

  def on_epoch_end(self, epoch, logs=None):
    super().on_epoch_end(epoch,logs)\

    # Also save the optimizer state
    filepath = self._get_file_path(epoch=epoch, 
        logs=logs, batch=None)
    filepath = filepath.rsplit( ".", 1 )[ 0 ] 
    filepath += ".pkl"

    with open(filepath, 'wb') as fp:
      pickle.dump(
        {
          'opt': model.optimizer.get_config(),
          'epoch': epoch+1
         # Add additional keys if you need to store more values
        }, fp, protocol=pickle.HIGHEST_PROTOCOL)
    print('\nEpoch %05d: saving optimizaer to %s' % (epoch + 1, filepath))\end{lstlisting}
\end{tcolorbox}%
The optimizer applies a step decay schedule during training to decrease the learning rate as training progresses.  It is essential to preserve the current epoch that we are on to perform correctly after a training resume.%
\index{learning}%
\index{learning rate}%
\index{training}%
\par%
\begin{tcolorbox}[size=title,title=Code,breakable]%
\begin{lstlisting}[language=Python, upquote=true]
def step_decay_schedule(initial_lr=1e-3, decay_factor=0.75, step_size=10):
    def schedule(epoch):
        return initial_lr * (decay_factor ** np.floor(epoch/step_size))
    return LearningRateScheduler(schedule)\end{lstlisting}
\end{tcolorbox}%
We build the model just as we have in previous sessions.  However, the training function requires a few extra considerations.  We specify the maximum number of epochs; however, we also allow the user to select the starting epoch number for training continuation.%
\index{model}%
\index{training}%
\par%
\begin{tcolorbox}[size=title,title=Code,breakable]%
\begin{lstlisting}[language=Python, upquote=true]
def build_model(input_shape, num_classes):
    model = Sequential()
    model.add(Conv2D(32, kernel_size=(3, 3),
                     activation='relu',
                     input_shape=input_shape))
    model.add(Conv2D(64, (3, 3), activation='relu'))
    model.add(MaxPooling2D(pool_size=(2, 2)))
    model.add(Dropout(0.25))
    model.add(Flatten())
    model.add(Dense(128, activation='relu'))
    model.add(Dropout(0.5))
    model.add(Dense(num_classes, activation='softmax'))
    model.compile(
        loss='categorical_crossentropy', 
        optimizer=tf.keras.optimizers.Adam(),
        metrics=['accuracy'])
    return model

def train_model(model, initial_epoch=0, max_epochs=10):
    start_time = time.time()

    checkpoint_cb = MyModelCheckpoint(
        os.path.join(run_dir, 'model-{epoch:02d}-{val_loss:.2f}.hdf5'),
        monitor='val_loss',verbose=1)

    lr_sched_cb = step_decay_schedule(initial_lr=1e-4, decay_factor=0.75, \
                                      step_size=2)
    cb = [checkpoint_cb, lr_sched_cb]

    model.fit(x_train, y_train,
              batch_size=batch_size,
              epochs=max_epochs,
              initial_epoch = initial_epoch,
              verbose=2, callbacks=cb,
              validation_data=(x_test, y_test))
    score = model.evaluate(x_test, y_test, verbose=0, callbacks=cb)
    print('Test loss: {}'.format(score[0]))
    print('Test accuracy: {}'.format(score[1]))

    elapsed_time = time.time() - start_time
    print("Elapsed time: {}".format(hms_string(elapsed_time)))\end{lstlisting}
\end{tcolorbox}%
We now begin training, using the%
\index{training}%
\textbf{ Logger }%
class to write the output to a log file in the output directory.%
\index{output}%
\par%
\begin{tcolorbox}[size=title,title=Code,breakable]%
\begin{lstlisting}[language=Python, upquote=true]
with Logger(os.path.join(run_dir, 'log.txt')):
    input_shape, x_train, y_train, x_test, y_test = obtain_data()
    model = build_model(input_shape, num_classes)
    train_model(model, max_epochs=3)\end{lstlisting}
\tcbsubtitle[before skip=\baselineskip]{Output}%
\begin{lstlisting}[upquote=true]
Downloading data from https://storage.googleapis.com/tensorflow/tf-
keras-datasets/mnist.npz
11493376/11490434 [==============================] - 0s 0us/step
11501568/11490434 [==============================] - 0s 0us/step
Shape of x_train: (60000, 28, 28)
Shape of y_train: (60000,)
Shape of x_test: (10000, 28, 28)
Shape of y_test: (10000,)
x_train shape: (60000, 28, 28, 1)
Training samples: 60000
Test samples: 10000
...
469/469 - 2s - loss: 0.2284 - accuracy: 0.9332 - val_loss: 0.1087 -
val_accuracy: 0.9677 - lr: 1.0000e-04 - 2s/epoch - 5ms/step
Epoch 3/3

...

469/469 - 2s - loss: 0.1575 - accuracy: 0.9541 - val_loss: 0.0837 -
val_accuracy: 0.9746 - lr: 7.5000e-05 - 2s/epoch - 5ms/step
Test loss: 0.08365701138973236
Test accuracy: 0.9746000170707703
Elapsed time: 0:00:22.09
\end{lstlisting}
\end{tcolorbox}%
You should notice that the above output displays the name of the hdf5 and pickle (pkl) files produced at each checkpoint. These files serve the following functions:%
\index{output}%
\par%
\begin{itemize}[noitemsep]%
\item%
Pickle files contain the state of the optimizer.%
\item%
HDF5 files contain the saved model.%
\index{model}%
\end{itemize}%
For this training run, which went for 3 epochs, these two files were named:%
\index{training}%
\par%
\begin{itemize}[noitemsep]%
\item%
./data/00013{-}test{-}train/model{-}03{-}0.08.hdf5%
\index{model}%
\item%
./data/00013{-}test{-}train/model{-}03{-}0.08.pkl%
\index{model}%
\end{itemize}%
We can inspect the output from the training run. Notice we can see a folder named "00000{-}test{-}train". This new folder was the first training run. The program will call the next training run "00001{-}test{-}train", and so on. Inside this directory, you will find the pickle and hdf5 files for each checkpoint.%
\index{output}%
\index{training}%
\par%
\begin{tcolorbox}[size=title,title=Code,breakable]%
\begin{lstlisting}[language=Python, upquote=true]
!ls ./data\end{lstlisting}
\tcbsubtitle[before skip=\baselineskip]{Output}%
\begin{lstlisting}[upquote=true]
00000-test-train
\end{lstlisting}
\end{tcolorbox}%
\begin{tcolorbox}[size=title,title=Code,breakable]%
\begin{lstlisting}[language=Python, upquote=true]
!ls ./data/00000-test-train\end{lstlisting}
\tcbsubtitle[before skip=\baselineskip]{Output}%
\begin{lstlisting}[upquote=true]
log.txt             model-01-0.20.pkl   model-02-0.11.pkl
model-03-0.08.pkl
model-01-0.20.hdf5  model-02-0.11.hdf5  model-03-0.08.hdf5
\end{lstlisting}
\end{tcolorbox}%
Keras stores the model itself in an HDF5, which includes the optimizer. Because of this feature, it is not generally necessary to restore the internal state of the optimizer (such as ADAM). However, we include the code to do so. We can obtain the internal state of an optimizer by calling%
\index{ADAM}%
\index{feature}%
\index{Keras}%
\index{model}%
\textbf{ get\_config}%
, which will return a dictionary similar to the following:%
\par%
\begin{tcolorbox}[size=title,breakable]%
\begin{lstlisting}[upquote=true]
{'name': 'Adam', 'learning_rate': 7.5e-05, 'decay': 0.0, 
'beta_1': 0.9, 'beta_2': 0.999, 'epsilon': 1e-07, 'amsgrad': False}
\end{lstlisting}
\end{tcolorbox}%
In practice, I've found that different optimizers implement get\_config differently. This function will always return the training hyperparameters. However, it may not always capture the complete internal state of an optimizer beyond the hyperparameters. The exact implementation of get\_config can vary per optimizer implementation.%
\index{hyperparameter}%
\index{parameter}%
\index{training}%
\par%
\subsection{Continuing Training}%
\label{subsec:ContinuingTraining}%
We are now ready to continue training. You will need the paths to both your HDF5 and PKL files. You can find these paths in the output above. Your values may differ from mine, so perform a copy/paste.%
\index{output}%
\index{training}%
\par%
\begin{tcolorbox}[size=title,title=Code,breakable]%
\begin{lstlisting}[language=Python, upquote=true]
MODEL_PATH = './data/00000-test-train/model-03-0.08.hdf5'
OPT_PATH = './data/00000-test-train/model-03-0.08.pkl'\end{lstlisting}
\end{tcolorbox}%
The following code loads the HDF5 and PKL files and then recompiles the model based on the PKL file.  Depending on the optimizer in use, you might have to recompile the model.%
\index{model}%
\par%
\begin{tcolorbox}[size=title,title=Code,breakable]%
\begin{lstlisting}[language=Python, upquote=true]
import tensorflow as tf
from tensorflow.keras.models import load_model
import pickle

def load_model_data(model_path, opt_path):
    model = load_model(model_path)
    with open(opt_path, 'rb') as fp:
      d = pickle.load(fp)
      epoch = d['epoch']
      opt = d['opt']
      return epoch, model, opt

epoch, model, opt = load_model_data(MODEL_PATH, OPT_PATH)

# note: often it is not necessary to recompile the model
model.compile(
    loss='categorical_crossentropy', 
    optimizer=tf.keras.optimizers.Adam.from_config(opt),
    metrics=['accuracy'])\end{lstlisting}
\end{tcolorbox}%
Finally, we train the model for additional epochs.  You can see from the output that the new training starts at a higher accuracy than the first training run.  Further, the accuracy increases with additional training.  Also, you will notice that the epoch number begins at four and not one.%
\index{model}%
\index{output}%
\index{training}%
\par%
\begin{tcolorbox}[size=title,title=Code,breakable]%
\begin{lstlisting}[language=Python, upquote=true]
outdir = "./data/"
run_desc = "cont-train"
num_classes = 10

run_dir = generate_output_dir(outdir, run_desc)
print(f"Results saved to: {run_dir}")

with Logger(os.path.join(run_dir, 'log.txt')):
  input_shape, x_train, y_train, x_test, y_test = obtain_data()
  train_model(model, initial_epoch=epoch, max_epochs=6)\end{lstlisting}
\tcbsubtitle[before skip=\baselineskip]{Output}%
\begin{lstlisting}[upquote=true]
Results saved to: ./data/00001-cont-train
Shape of x_train: (60000, 28, 28)
Shape of y_train: (60000,)
Shape of x_test: (10000, 28, 28)
Shape of y_test: (10000,)
x_train shape: (60000, 28, 28, 1)
Training samples: 60000
Test samples: 10000
...
469/469 - 2s - loss: 0.1099 - accuracy: 0.9677 - val_loss: 0.0612 -
val_accuracy: 0.9818 - lr: 5.6250e-05 - 2s/epoch - 5ms/step
Epoch 6/6
Epoch 6: saving model to ./data/00001-cont-train/model-06-0.06.hdf5
Epoch 00006: saving optimizaer to ./data/00001-cont-
train/model-06-0.06.pkl
469/469 - 2s - loss: 0.0990 - accuracy: 0.9711 - val_loss: 0.0561 -
val_accuracy: 0.9827 - lr: 5.6250e-05 - 2s/epoch - 5ms/step
Test loss: 0.05610647052526474
Test accuracy: 0.982699990272522
Elapsed time: 0:00:11.72
\end{lstlisting}
\end{tcolorbox}

\section{Part 13.3: Using a Keras Deep Neural Network with a Web Application}%
\begin{figure}[h]%
\centering%
\includegraphics[width=4in]{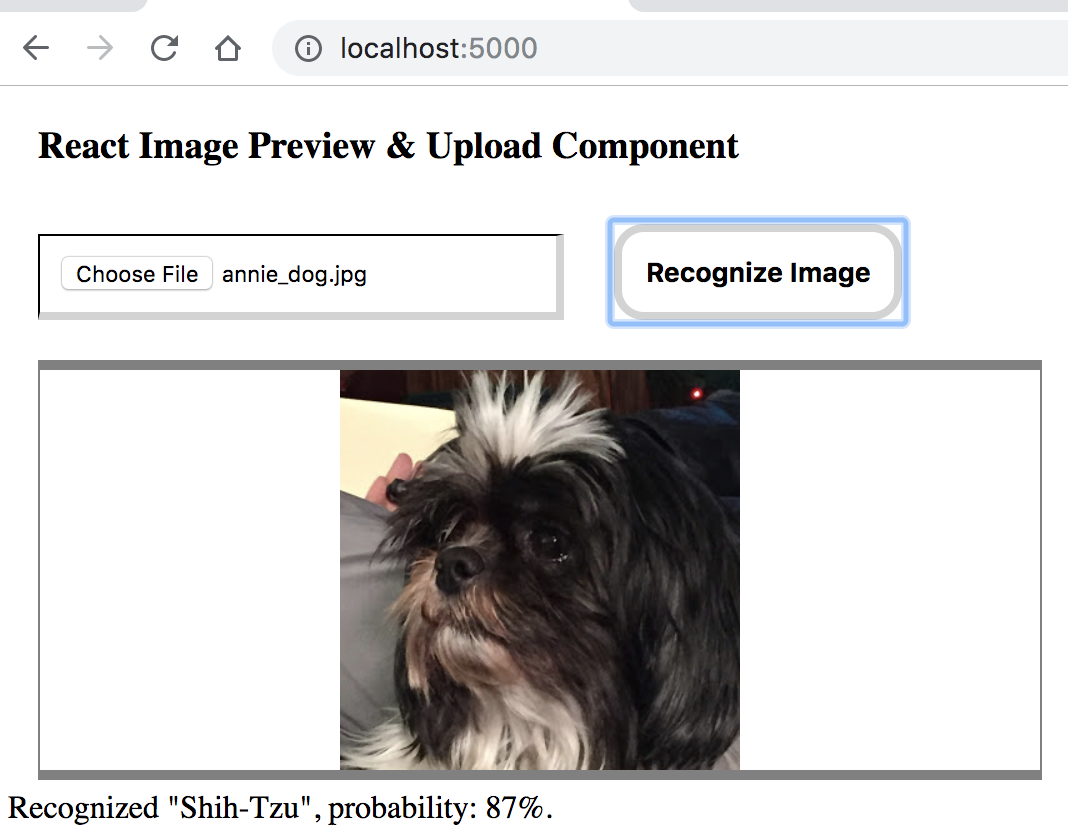}%
\caption{AI Web Application}%
\label{13.WEB}%
\end{figure}

\label{sec:Part13.3UsingaKerasDeepNeuralNetworkwithaWebApplication}%
In this part, we will extend the image API developed in Part 13.1 to work with a web application. This technique allows you to use a simple website to upload/predict images, such as in Figure \ref{13.WEB}.%
\index{predict}%
\par%

\par%
I added neural network functionality to a simple ReactJS image upload and preview example. To do this, we will use the same API developed in Module 13.1. However, we will now add a%
\index{neural network}%
\href{https://reactjs.org/}{ ReactJS }%
website around it. This single{-}page web application allows you to upload images for classification by the neural network. If you would like to read more about ReactJS and image uploading, you can refer to the%
\index{classification}%
\index{neural network}%
\href{http://www.hartzis.me/react-image-upload/}{ blog post }%
that provided some inspiration for this example.%
\index{SOM}%
\par%
I built this example from the following components:%
\par%
\begin{itemize}[noitemsep]%
\item%
\href{./py/}{GitHub Location for Web App}%
\item%
\href{./py/image_web_server_1.py}{image\_web\_server\_1.py }%
{-} The code both to start Flask and serve the HTML/JavaScript/CSS needed to provide the web interface.%
\index{Flask}%
\index{Java}%
\index{JavaScript}%
\item%
Directory WWW {-} Contains web assets.%
\begin{itemize}[noitemsep]%
\item%
\href{./py/www/index.html}{index.html }%
{-} The main page for the web application.%
\item%
\href{./py/www/style.css}{style.css }%
{-} The stylesheet for the web application.%
\item%
\href{./py/www/script.js}{script.js }%
{-} The JavaScript code for the web application.%
\index{Java}%
\index{JavaScript}%
\end{itemize}%
\end{itemize}

\section{Part 13.4: When to Retrain Your Neural Network}%
\label{sec:Part13.4WhentoRetrainYourNeuralNetwork}%
Dataset drift is a problem frequently seen in real{-}world applications of machine learning. Academic problems that courses typically present in school assignments usually do not experience this problem. For a class assignment, your instructor provides a single data set representing all of the data you will ever see for a task. In the real world, you obtain initial data to train your model; then, you will acquire new data over time that you use your model to predict.%
\index{dataset}%
\index{learning}%
\index{model}%
\index{predict}%
\par%
Consider this example. You create a startup company that develops a mobile application that helps people find jobs. To train your machine learning model, you collect attributes about people and their careers. Once you have your data, you can prepare your neural network to suggest the best jobs for individuals.%
\index{learning}%
\index{model}%
\index{neural network}%
\par%
Once your application is released, you will hopefully obtain new data. This data will come from job seekers using your app. These people are your customers. You have x values (their attributes), but you do not have y{-}values (their jobs). Your customers have come to you to find out what their be jobs will be. You will provide the customer's attributes to the neural network, and then it will predict their jobs. Usually, companies develop neural networks on initial data and then use the neural network to perform predictions on new data obtained over time from their customers.%
\index{neural network}%
\index{predict}%
\par%
Your job prediction model will become less relevant as the industry introduces new job types and the demographics of your customers change. However, companies must look if their model is still relevant as time passes. This change in your underlying data is called dataset drift. In this section, we will see ways that you can measure dataset drift.%
\index{dataset}%
\index{model}%
\index{predict}%
\par%
You can present your model with new data and see how its accuracy changes over time. However, to calculate efficiency, you must know the expected outputs from the model (y{-}values). You may not know the correct outcomes for new data that you are obtaining in real{-}time. Therefore, we will look at algorithms that examine the x{-}inputs and determine how much they have changed in distribution from the original x{-}inputs that we trained on. These changes are called dataset drift.%
\index{algorithm}%
\index{dataset}%
\index{input}%
\index{model}%
\index{output}%
\par%
Let's begin by creating generated data that illustrates drift. We present the following code to create a chart that shows such drift.%
\par%
\begin{tcolorbox}[size=title,title=Code,breakable]%
\begin{lstlisting}[language=Python, upquote=true]
import numpy as np

import matplotlib.pyplot as plot
from sklearn.linear_model import LinearRegression

def true_function(x):
    x2 = (x*8) - 1
    return ((np.sin(x2)/x2)*0.6)+0.3
    
#
x_train  = np.arange(0, 0.6, 0.01)
x_test  = np.arange(0.6, 1.1, 0.01)
x_true = np.concatenate( (x_train, x_test) )

#
y_true_train = true_function(x_train)
y_true_test = true_function(x_test)
y_true   = np.concatenate( (y_true_train, y_true_test) )

#
y_train = y_true_train + (np.random.rand(*x_train.shape)-0.5)*0.4
y_test  = y_true_test + (np.random.rand(*x_test.shape)-0.5)*0.4

#
lr_x_train = x_train.reshape((x_train.shape[0],1))
reg = LinearRegression().fit(lr_x_train, y_train)
reg_pred = reg.predict(lr_x_train)
print(reg.coef_[0])
print(reg.intercept_)

#
plot.xlim([0,1.5])
plot.ylim([0,1])
l1 = plot.scatter(x_train, y_train, c="g", label="Training Data")
l2 = plot.scatter(x_test, y_test, c="r", label="Testing Data")
l3, = plot.plot(lr_x_train, reg_pred, color='black', linewidth=3, 
                label="Trained Model")
l4, = plot.plot(x_true, y_true, label = "True Function")
plot.legend(handles=[l1, l2, l3, l4])

#
plot.title('Drift')
plot.xlabel('Time')
plot.ylabel('Sales')
plot.grid(True, which='both')
plot.show()\end{lstlisting}
\tcbsubtitle[before skip=\baselineskip]{Output}%
\includegraphics[width=3in]{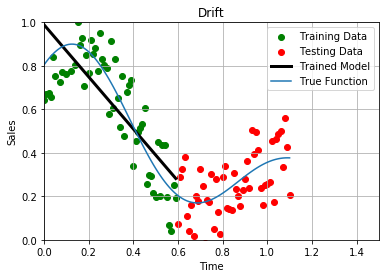}%
\begin{lstlisting}[upquote=true]
-1.1979470956001936
0.9888340153211445
\end{lstlisting}
\end{tcolorbox}%
The "True function" represents what the data does over time. Unfortunately, you only have the training portion of the data. Your model will do quite well on the data that you trained it with; however, it will be very inaccurate on the new test data presented. The prediction line for the model fits the training data well but does not fit the est data well.%
\index{model}%
\index{predict}%
\index{training}%
\par%
\subsection{Preprocessing the Sberbank Russian Housing Market Data}%
\label{subsec:PreprocessingtheSberbankRussianHousingMarketData}%
The examples provided in this section use a Kaggle dataset named The Sberbank Russian Housing Market, which you can access from the following link.%
\index{dataset}%
\index{Kaggle}%
\index{link}%
\par%
\begin{itemize}[noitemsep]%
\item%
\href{https://www.kaggle.com/c/sberbank-russian-housing-market/data}{Sberbank Russian Housing Market}%
\end{itemize}%
Because Kaggle provides datasets as training and test, we must load both of these files.%
\index{dataset}%
\index{Kaggle}%
\index{training}%
\par%
\begin{tcolorbox}[size=title,title=Code,breakable]%
\begin{lstlisting}[language=Python, upquote=true]
import os
import numpy as np
import pandas as pd
from sklearn.preprocessing import LabelEncoder

PATH = "/Users/jheaton/Downloads/sberbank-russian-housing-market"


train_df = pd.read_csv(os.path.join(PATH,"train.csv"))
test_df = pd.read_csv(os.path.join(PATH,"test.csv"))\end{lstlisting}
\end{tcolorbox}%
I provide a simple preprocess function that converts all numerics to z{-}scores and all categoricals to dummies.%
\index{categorical}%
\index{ROC}%
\index{ROC}%
\index{Z{-}Score}%
\par%
\begin{tcolorbox}[size=title,title=Code,breakable]%
\begin{lstlisting}[language=Python, upquote=true]
def preprocess(df):
    for i in df.columns:
        if df[i].dtype == 'object':
            df[i] = df[i].fillna(df[i].mode().iloc[0])
        elif (df[i].dtype == 'int' or df[i].dtype == 'float'):
            df[i] = df[i].fillna(np.nanmedian(df[i]))

    enc = LabelEncoder()
    for i in df.columns:
        if (df[i].dtype == 'object'):
            df[i] = enc.fit_transform(df[i].astype('str'))
            df[i] = df[i].astype('object')\end{lstlisting}
\end{tcolorbox}%
Next, we run the training and test datasets through the preprocessing function.%
\index{dataset}%
\index{ROC}%
\index{ROC}%
\index{training}%
\par%
\begin{tcolorbox}[size=title,title=Code,breakable]%
\begin{lstlisting}[language=Python, upquote=true]
preprocess(train_df)
preprocess(test_df)\end{lstlisting}
\end{tcolorbox}%
Finally, we remove thr target variable.  We are only looking for drift on the $x$ (input data).%
\index{input}%
\par%
\begin{tcolorbox}[size=title,title=Code,breakable]%
\begin{lstlisting}[language=Python, upquote=true]
train_df.drop('price_doc',axis=1,inplace=True)\end{lstlisting}
\end{tcolorbox}

\subsection{KS{-}Statistic}%
\label{subsec:KS{-}Statistic}%
We will use the KS{-}Statistic to determine the difference in distribution between columns in the training and test sets. As a baseline, consider if we compare the same field to itself. In this case, we are comparing the%
\index{training}%
\textbf{ kitch\_sq }%
in the training set. Because there is no difference in distribution between a field in itself, the p{-}value is 1.0, and the KS{-}Statistic statistic is 0. The P{-}Value is the probability of no difference between the two distributions. Typically some lower threshold is used for how low a P{-}Value is needed to reject the null hypothesis and assume there is a difference. The value of 0.05 is a standard threshold for p{-}values. Because the p{-}value is NOT below 0.05, we expect the two distributions to be the same. If the p{-}value were below the threshold, the%
\index{probability}%
\index{SOM}%
\index{training}%
\textbf{ statistic }%
value becomes interesting. This value tells you how different the two distributions are. A value of 0.0, in this case, means no differences.%
\par%
\begin{tcolorbox}[size=title,title=Code,breakable]%
\begin{lstlisting}[language=Python, upquote=true]
from scipy import stats

stats.ks_2samp(train_df['kitch_sq'], train_df['kitch_sq'])\end{lstlisting}
\tcbsubtitle[before skip=\baselineskip]{Output}%
\begin{lstlisting}[upquote=true]
Ks_2sampResult(statistic=-0.0, pvalue=1.0)
\end{lstlisting}
\end{tcolorbox}%
Now let's do something more interesting.  We will compare the same field%
\index{SOM}%
\textbf{ kitch\_sq }%
between the test and training sets.  In this case, the p{-}value is below 0.05, so the%
\index{training}%
\textbf{ statistic }%
value now contains the amount of difference detected.%
\par%
\begin{tcolorbox}[size=title,title=Code,breakable]%
\begin{lstlisting}[language=Python, upquote=true]
stats.ks_2samp(train_df['kitch_sq'], test_df['kitch_sq'])\end{lstlisting}
\tcbsubtitle[before skip=\baselineskip]{Output}%
\begin{lstlisting}[upquote=true]
Ks_2sampResult(statistic=0.25829078867676714, pvalue=0.0)
\end{lstlisting}
\end{tcolorbox}%
Next, we pull the KS{-}Stat for every field.  We also establish a boundary for the maximum p{-}value and how much of a difference is needed before we display the column.%
\par%
\begin{tcolorbox}[size=title,title=Code,breakable]%
\begin{lstlisting}[language=Python, upquote=true]
for col in train_df.columns:
    ks = stats.ks_2samp(train_df[col], test_df[col])
    if ks.pvalue < 0.05 and ks.statistic>0.1:
        print(f'{col}: {ks}')\end{lstlisting}
\tcbsubtitle[before skip=\baselineskip]{Output}%
\begin{lstlisting}[upquote=true]
id: Ks_2sampResult(statistic=1.0, pvalue=0.0)
timestamp: Ks_2sampResult(statistic=0.8982081426022823, pvalue=0.0)
life_sq: Ks_2sampResult(statistic=0.2255084471628891,
pvalue=7.29401465948424e-271)
max_floor: Ks_2sampResult(statistic=0.17313454154786817,
pvalue=7.82000315371674e-160)
build_year: Ks_2sampResult(statistic=0.3176883950430345, pvalue=0.0)
num_room: Ks_2sampResult(statistic=0.1226755470309048,
pvalue=1.8622542043144584e-80)
kitch_sq: Ks_2sampResult(statistic=0.25829078867676714, pvalue=0.0)
state: Ks_2sampResult(statistic=0.13641341252952505,
pvalue=2.1968159319271184e-99)
preschool_quota: Ks_2sampResult(statistic=0.2364160801236304,
pvalue=1.1710777340471466e-297)
school_quota: Ks_2sampResult(statistic=0.25657342859882415,

...

cafe_sum_2000_max_price_avg:
Ks_2sampResult(statistic=0.10732529051140638,
pvalue=1.1100804327460878e-61)
cafe_avg_price_2000: Ks_2sampResult(statistic=0.1081218037860151,
pvalue=1.3575759911857293e-62)
\end{lstlisting}
\end{tcolorbox}

\subsection{Detecting Drift between Training and Testing Datasets by Training}%
\label{subsec:DetectingDriftbetweenTrainingandTestingDatasetsbyTraining}%
Sample the training and test into smaller sets to train.  We want 10K elements from each; however, the test set only has 7,662, so we only sample that amount from each side.%
\index{training}%
\par%
\begin{tcolorbox}[size=title,title=Code,breakable]%
\begin{lstlisting}[language=Python, upquote=true]
SAMPLE_SIZE = min(len(train_df),len(test_df))
SAMPLE_SIZE = min(SAMPLE_SIZE,10000)
print(SAMPLE_SIZE)\end{lstlisting}
\tcbsubtitle[before skip=\baselineskip]{Output}%
\begin{lstlisting}[upquote=true]
7662
\end{lstlisting}
\end{tcolorbox}%
We take the random samples from the training and test sets and add a flag called%
\index{random}%
\index{training}%
\textbf{ source\_training }%
to tell the two apart.%
\par%
\begin{tcolorbox}[size=title,title=Code,breakable]%
\begin{lstlisting}[language=Python, upquote=true]
training_sample = train_df.sample(SAMPLE_SIZE, random_state=49)
testing_sample = test_df.sample(SAMPLE_SIZE, random_state=48)

# Is the data from the training set?
training_sample['source_training'] = 1
testing_sample['source_training'] = 0\end{lstlisting}
\end{tcolorbox}%
Next, we combine the data that we sampled from the training and test data sets and shuffle them.%
\index{sampled}%
\index{training}%
\par%
\begin{tcolorbox}[size=title,title=Code,breakable]%
\begin{lstlisting}[language=Python, upquote=true]
# Build combined training set
combined = testing_sample.append(training_sample)
combined.reset_index(inplace=True, drop=True)

# Now randomize
combined = combined.reindex(np.random.permutation(combined.index))
combined.reset_index(inplace=True, drop=True)\end{lstlisting}
\end{tcolorbox}%
We will now generate $x$ and $y$ to train.  We attempt to predict the%
\index{predict}%
\textbf{ source\_training }%
value as $y$, which indicates if the data came from the training or test set.  If the model successfully uses the data to predict if it came from training or testing, then there is likely drift.  Ideally, the train and test data should be indistinguishable.%
\index{model}%
\index{predict}%
\index{training}%
\par%
\begin{tcolorbox}[size=title,title=Code,breakable]%
\begin{lstlisting}[language=Python, upquote=true]
# Get ready to train
y = combined['source_training'].values
combined.drop('source_training',axis=1,inplace=True)
x = combined.values

y\end{lstlisting}
\tcbsubtitle[before skip=\baselineskip]{Output}%
\begin{lstlisting}[upquote=true]
array([1, 1, 1, ..., 1, 0, 0])
\end{lstlisting}
\end{tcolorbox}%
We will consider anything above a 0.75 AUC as having a good chance of drift.%
\par%
\begin{tcolorbox}[size=title,title=Code,breakable]%
\begin{lstlisting}[language=Python, upquote=true]
from sklearn.ensemble import RandomForestClassifier
from sklearn.model_selection import cross_val_score

model = RandomForestClassifier(n_estimators = 60, max_depth = 7,
    min_samples_leaf = 5)
lst = []

for i in combined.columns:
    score = cross_val_score(model,pd.DataFrame(combined[i]),y,cv=2,
                            scoring='roc_auc')
    if (np.mean(score) > 0.75):
        lst.append(i)
        print(i,np.mean(score))\end{lstlisting}
\tcbsubtitle[before skip=\baselineskip]{Output}%
\begin{lstlisting}[upquote=true]
id 1.0
timestamp 0.9601862111975688
full_sq 0.7966785611424911
life_sq 0.8724218330166038
build_year 0.8004825176688191
kitch_sq 0.9070093804672634
cafe_sum_500_min_price_avg 0.8435920036035689
cafe_avg_price_500 0.8453533835344671
\end{lstlisting}
\end{tcolorbox}

\section{Part 13.5: Tensor Processing Units (TPUs)}%
\label{sec:Part13.5TensorProcessingUnits(TPUs)}%
This book focuses primarily on NVIDIA Graphics Processing Units (GPUs) for deep learning acceleration. NVIDIA GPUs are not the only option for deep learning acceleration. TensorFlow continues to gain additional support for AMD and Intel GPUs. TPUs are also available from Google cloud platforms to accelerate deep learning. The focus of this book and course is on NVIDIA GPUs because of their wide availability on both local and cloud systems.%
\index{GPU}%
\index{GPU}%
\index{learning}%
\index{NVIDIA}%
\index{ROC}%
\index{ROC}%
\index{TensorFlow}%
\par%
Though this book focuses on NVIDIA GPUs, we will briefly examine Google Tensor Processing Units (TPUs). These devices are an AI accelerator Application{-}Specific Integrated Circuit (ASIC) developed by Google. They were designed specifically for neural network machine learning, mainly using Google's TensorFlow software. Google began using TPUs internally in 2015 and in 2018 made them available for third{-}party use, both as part of its cloud infrastructure and by offering a smaller version of the chip for sale.%
\index{GAN}%
\index{GPU}%
\index{GPU}%
\index{learning}%
\index{neural network}%
\index{NVIDIA}%
\index{ROC}%
\index{ROC}%
\index{TensorFlow}%
\par%
The full use of a TPU is a complex topic that I only introduced in this part. Supporting TPUs is slightly more complicated than GPUs because specialized coding is needed. Changes are rarely required to adapt CPU code to GPU for most relatively simple mainstream GPU tasks in TensorFlow. I will cover the mild code changes needed to utilize in this part.%
\index{GPU}%
\index{GPU}%
\index{TensorFlow}%
\par%
We will create a regression neural network to count paper clips in this part. I demonstrated this dataset and task several times previously in this book. This part focuses on the utilization of TPUs and not the creation of neural networks. I covered the design of computer vision previously in this book.%
\index{computer vision}%
\index{dataset}%
\index{neural network}%
\index{regression}%
\par%
\begin{tcolorbox}[size=title,title=Code,breakable]%
\begin{lstlisting}[language=Python, upquote=true]
import os
import pandas as pd

URL = "https://github.com/jeffheaton/data-mirror/"
DOWNLOAD_SOURCE = URL+"releases/download/v1/paperclips.zip"
DOWNLOAD_NAME = DOWNLOAD_SOURCE[DOWNLOAD_SOURCE.rfind('/')+1:]

if COLAB:
  PATH = "/content"
else:
  # I used this locally on my machine, you may need different
  PATH = "/Users/jeff/temp"

EXTRACT_TARGET = os.path.join(PATH,"clips")
SOURCE = os.path.join(EXTRACT_TARGET, "paperclips")

# Download paperclip data
!wget -O {os.path.join(PATH,DOWNLOAD_NAME)} {DOWNLOAD_SOURCE}
!mkdir -p {SOURCE}
!mkdir -p {TARGET}
!mkdir -p {EXTRACT_TARGET}
!unzip -o -j -d {SOURCE} {os.path.join(PATH, DOWNLOAD_NAME)} >/dev/null

# Add filenames
df_train = pd.read_csv(os.path.join(SOURCE, "train.csv"))
df_train['filename'] = "clips-" + df_train.id.astype(str) + ".jpg"\end{lstlisting}
\end{tcolorbox}%
\subsection{Preparing Data for TPUs}%
\label{subsec:PreparingDataforTPUs}%
To present the paperclips dataset to the TPU, we will convert the images to a Keras Dataset. Because we will load the entire dataset to RAM, we will only utilize the first 1,000 images. We previously loaded the labels from the%
\index{dataset}%
\index{Keras}%
\textbf{ train.csv }%
file. The following code loads these images and converts them to a Keras dataset.%
\index{dataset}%
\index{Keras}%
\par%
\begin{tcolorbox}[size=title,title=Code,breakable]%
\begin{lstlisting}[language=Python, upquote=true]
import tensorflow as tf
import keras_preprocessing
import glob, os
import tqdm
import numpy as np
from PIL import Image

IMG_SHAPE = (128,128)
BATCH_SIZE = 32

# Resize each image and convert the 0-255 ranged RGB values to 0-1 range.
def load_images(files, img_shape):
  cnt = len(files)
  x = np.zeros((cnt,)+img_shape+(3,),dtype=np.float32)
  i = 0
  for file in tqdm.tqdm(files):
    img = Image.open(file)
    img = img.resize(img_shape)
    img = np.array(img)
    img = img/255
    x[i,:,:,:] = img
    i+=1
  return x

# Process training data 
df_train = pd.read_csv(os.path.join(SOURCE, "train.csv"))
df_train['filename'] = "clips-" + df_train.id.astype(str) + ".jpg"

# Use only the first 1000 images
df_train = df_train[0:1000]

# Load images
images = [os.path.join(SOURCE,x) for x in df_train.filename]
x = load_images(images, IMG_SHAPE)
y = df_train.clip_count.values

# Convert to dataset
dataset = tf.data.Dataset.from_tensor_slices((x, y))
dataset = dataset.batch(BATCH_SIZE)\end{lstlisting}
\end{tcolorbox}%
TPUs are typically Cloud TPU workers, different from the local process running the user's Python program. Thus, it would be best to do some initialization work to connect to the remote cluster and initialize the TPUs. The TPU argument to%
\index{Python}%
\index{ROC}%
\index{ROC}%
\index{SOM}%
\textbf{ tf.distribute.cluster\_resolver}%
.%
\textbf{ TPUClusterResolver }%
is a unique address just for Colab. If you are running your code on Google Compute Engine (GCE), you should instead pass in the name of your Cloud TPU. The following code performs this initialization.%
\par%
\begin{tcolorbox}[size=title,title=Code,breakable]%
\begin{lstlisting}[language=Python, upquote=true]
try:
    tpu = tf.distribute.cluster_resolver.TPUClusterResolver.connect()
    print("Device:", tpu.master())
    strategy = tf.distribute.TPUStrategy(tpu)
except:
    strategy = tf.distribute.get_strategy()
print("Number of replicas:", strategy.num_replicas_in_sync)\end{lstlisting}
\end{tcolorbox}%
We will now use a ResNet neural network as a basis for our neural network. We begin by loading, from Keras, the ResNet50 network. We will redefine both the input shape and output of the ResNet model, so we will not transfer the weights. Since we redefine the input, the weights are of minimal value. We specify%
\index{input}%
\index{Keras}%
\index{model}%
\index{neural network}%
\index{output}%
\index{ResNet}%
\textbf{ include\_top }%
as False because we will change the input resolution. We also specify%
\index{input}%
\textbf{ weights }%
as false because we must retrain the network after changing the top input layers.%
\index{input}%
\index{input layer}%
\index{layer}%
\par%
\begin{tcolorbox}[size=title,title=Code,breakable]%
\begin{lstlisting}[language=Python, upquote=true]
from tensorflow.keras.applications.resnet50 import ResNet50
from tensorflow.keras.layers import Input
from tensorflow.keras.layers import Dense, GlobalAveragePooling2D
from tensorflow.keras.models import Model
from tensorflow.keras.callbacks import EarlyStopping
from tensorflow.keras.metrics import RootMeanSquaredError

def create_model():
  input_tensor = Input(shape=IMG_SHAPE+(3,))

  base_model = ResNet50(
      include_top=False, weights=None, input_tensor=input_tensor,
      input_shape=None)

  x=base_model.output
  x=GlobalAveragePooling2D()(x)
  x=Dense(1024,activation='relu')(x) 
  x=Dense(1024,activation='relu')(x) 
  model=Model(inputs=base_model.input,outputs=Dense(1)(x))
  return model

with strategy.scope():
  model = create_model()

  model.compile(loss = 'mean_squared_error', optimizer='adam', 
              metrics=[RootMeanSquaredError(name="rmse")])

  history = model.fit(dataset, epochs=100, steps_per_epoch=32, 
                    verbose = 1)\end{lstlisting}
\tcbsubtitle[before skip=\baselineskip]{Output}%
\begin{lstlisting}[upquote=true]
...
32/32 [==============================] - 1s 44ms/step - loss: 18.3960
- rmse: 4.2891
Epoch 100/100
32/32 [==============================] - 1s 44ms/step - loss: 10.4749
- rmse: 3.2365
\end{lstlisting}
\end{tcolorbox}%
You might receive the following error while fitting the neural network.%
\index{error}%
\index{neural network}%
\par%
\begin{tcolorbox}[size=title,breakable]%
\begin{lstlisting}[upquote=true]
InvalidArgumentError: Unable to parse tensor proto
\end{lstlisting}
\end{tcolorbox}%
If you do receive this error, it is likely because you are missing proper authentication to access Google Drive to store your datasets.%
\index{dataset}%
\index{error}%
\par

\chapter{Other Neural Network Techniques}%
\label{chap:OtherNeuralNetworkTechniques}%
\section{Part 14.1: What is AutoML}%
\label{sec:Part14.1WhatisAutoML}%
Automatic Machine Learning (AutoML) attempts to use machine learning to automate itself.  Data is passed to the AutoML application in raw form, and models are automatically generated.%
\index{AutoML}%
\index{learning}%
\index{model}%
\par%
\subsection{AutoML from your Local Computer}%
\label{subsec:AutoMLfromyourLocalComputer}%
The following AutoML applications are free:%
\index{AutoML}%
\par%
\begin{itemize}[noitemsep]%
\item%
\href{https://autokeras.com/}{AutoKeras}%
\item%
\href{https://automl.github.io/auto-sklearn/master/}{Auto{-}SKLearn}%
\item%
\href{https://github.com/automl/Auto-PyTorch}{Auto PyTorch}%
\item%
\href{http://epistasislab.github.io/tpot/}{TPOT}%
\end{itemize}%
The following AutoML applications are commercial:%
\index{AutoML}%
\par%
\begin{itemize}[noitemsep]%
\item%
\href{https://rapidminer.com/educational-program/}{Rapid Miner }%
{-} Free student version available.%
\item%
\href{https://www.dataiku.com/dss/editions/}{Dataiku }%
{-} Free community version available.%
\item%
\href{https://www.datarobot.com/}{DataRobot }%
{-} Commercial%
\item%
\href{https://www.h2o.ai/products/h2o-driverless-ai/}{H2O Driverless }%
{-} Commercial%
\end{itemize}

\subsection{AutoML from Google Cloud}%
\label{subsec:AutoMLfromGoogleCloud}%
There are also cloud{-}hosted AutoML platforms:%
\index{AutoML}%
\par%
\begin{itemize}[noitemsep]%
\item%
\href{https://cloud.google.com/vision/automl/docs/tutorial}{Google Cloud AutoML Tutorial}%
\item%
\href{https://docs.microsoft.com/en-us/azure/machine-learning/how-to-use-automated-ml-for-ml-models}{Azure AutoML}%
\end{itemize}%
This module will show how to use%
\href{https://autokeras.com/}{ AutoKeras}%
. First, we download the paperclips counting dataset that you saw previously in this book.%
\index{dataset}%
\par%
\begin{tcolorbox}[size=title,title=Code,breakable]%
\begin{lstlisting}[language=Python, upquote=true]
import os
import pandas as pd

URL = "https://github.com/jeffheaton/data-mirror/"
DOWNLOAD_SOURCE = URL+"releases/download/v1/paperclips.zip"
DOWNLOAD_NAME = DOWNLOAD_SOURCE[DOWNLOAD_SOURCE.rfind('/')+1:]

if COLAB:
  PATH = "/content"
else:
  # I used this locally on my machine, you may need different
  PATH = "/Users/jeff/temp"

EXTRACT_TARGET = os.path.join(PATH,"clips")
SOURCE = os.path.join(EXTRACT_TARGET, "paperclips")

# Download paperclip data
!wget -O {os.path.join(PATH,DOWNLOAD_NAME)} {DOWNLOAD_SOURCE}
!mkdir -p {SOURCE}
!mkdir -p {TARGET}
!mkdir -p {EXTRACT_TARGET}
!unzip -o -j -d {SOURCE} {os.path.join(PATH, DOWNLOAD_NAME)} >/dev/null

# Process training data 
df_train = pd.read_csv(os.path.join(SOURCE, "train.csv"))
df_train['filename'] = "clips-" + df_train.id.astype(str) + ".jpg"

# Use only the first 1000 images
df_train = df_train[0:1000]\end{lstlisting}
\end{tcolorbox}%
One limitation of AutoKeras is that it cannot directly utilize generators. Without resorting to complex techniques, all training data must reside in RAM. We will use the following code to load the image data to RAM.%
\index{Keras}%
\index{training}%
\par%
\begin{tcolorbox}[size=title,title=Code,breakable]%
\begin{lstlisting}[language=Python, upquote=true]
import tensorflow as tf
import keras_preprocessing
import glob, os
import tqdm
import numpy as np
from PIL import Image

IMG_SHAPE = (128,128)

def load_images(files, img_shape):
  cnt = len(files)
  x = np.zeros((cnt,)+img_shape+(3,))
  i = 0
  for file in tqdm.tqdm(files):
    img = Image.open(file)
    img = img.resize(img_shape)
    img = np.array(img)
    img = img/255
    x[i,:,:,:] = img
    i+=1
  return x

images = [os.path.join(SOURCE,x) for x in df_train.filename]
x = load_images(images, IMG_SHAPE)
y = df_train.clip_count.values\end{lstlisting}
\end{tcolorbox}

\subsection{Using AutoKeras}%
\label{subsec:UsingAutoKeras}%
\href{https://autokeras.com/}{AutoKeras }%
is an AutoML system based on Keras. The goal of AutoKeras is to make machine learning accessible to everyone.%
\index{AutoML}%
\index{Keras}%
\index{learning}%
\href{http://people.tamu.edu/~guangzhou92/Data_Lab/}{ DATA Lab }%
develops it at%
\href{https://www.tamu.edu/}{ Texas A\&M University}%
. We will see how to provide the paperclips dataset to AutoKeras and create an automatically tuned Keras deep learning model from this dataset. This automatic process frees you from choosing layer types and neuron counts.%
\index{dataset}%
\index{Keras}%
\index{layer}%
\index{learning}%
\index{model}%
\index{neuron}%
\index{ROC}%
\index{ROC}%
\par%
We begin by installing AutoKeras.%
\index{Keras}%
\par%
\begin{tcolorbox}[size=title,title=Code,breakable]%
\begin{lstlisting}[language=Python, upquote=true]
!pip install autokeras\end{lstlisting}
\end{tcolorbox}%
AutoKeras contains several%
\index{Keras}%
\href{https://autokeras.com/tutorial/overview/}{ examples }%
demonstrating image, tabular, and time{-}series data. We will make use of the%
\index{time{-}series}%
\textbf{ ImageRegressor}%
. Refer to the AutoKeras documentation for other classifiers and regressors to fit specific uses.%
\index{Keras}%
\par%
We define several variables to determine the AutoKeras operation:%
\index{Keras}%
\par%
\begin{itemize}[noitemsep]%
\item%
\textbf{MAX\_TRIALS }%
{-} Determines how many different models to see.%
\index{model}%
\item%
\textbf{SEED }%
{-} You can try different random seeds to obtain different results.%
\index{random}%
\item%
\textbf{VAL\_SPLIT }%
{-} What percent of the dataset should we use for validation.%
\index{dataset}%
\index{validation}%
\item%
\textbf{EPOCHS }%
{-} How many epochs to try each model for training.%
\index{model}%
\index{training}%
\item%
\textbf{BATCH\_SIZE }%
{-} Training batch size.%
\index{training}%
\index{training batch}%
\end{itemize}%
Setting MAX\_TRIALS and EPOCHS will have a great impact on your total runtime. You must balance how many models to try (MAX\_TRIALS) and how deeply to try to train each (EPOCHS). AutoKeras utilize early stopping, so setting EPOCHS too high will mean early stopping will prevent you from reaching the EPOCHS number of epochs.%
\index{early stopping}%
\index{Keras}%
\index{model}%
\par%
One strategy is to do a broad, shallow search. Set TRIALS high and EPOCHS low. The resulting model likely has the best hyperparameters. Finally, train this resulting model fully.%
\index{hyperparameter}%
\index{model}%
\index{parameter}%
\par%
\begin{tcolorbox}[size=title,title=Code,breakable]%
\begin{lstlisting}[language=Python, upquote=true]
import numpy as np
import autokeras as ak

MAX_TRIALS = 2
SEED = 42
VAL_SPLIT = 0.1
EPOCHS = 1000
BATCH_SIZE = 32

auto_reg = ak.ImageRegressor(overwrite=True, 
  max_trials=MAX_TRIALS,
  seed=42)
auto_reg.fit(x, y, validation_split=VAL_SPLIT, batch_size=BATCH_SIZE, 
        epochs=EPOCHS)
print(auto_reg.evaluate(x,y))\end{lstlisting}
\tcbsubtitle[before skip=\baselineskip]{Output}%
\begin{lstlisting}[upquote=true]
Trial 2 Complete [00h 04m 17s]
val_loss: 36.5126953125
Best val_loss So Far: 36.123992919921875
Total elapsed time: 01h 05m 46s
INFO:tensorflow:Oracle triggered exit
...
32/32 [==============================] - 3s 85ms/step - loss: 24.9218
- mean_squared_error: 24.9218
Epoch 1000/1000
32/32 [==============================] - 2s 78ms/step - loss: 24.9141
- mean_squared_error: 24.9141
INFO:tensorflow:Assets written to: ./image_regressor/best_model/assets
32/32 [==============================] - 2s 30ms/step - loss: 24.9077
- mean_squared_error: 24.9077
[24.90774917602539, 24.90774917602539]
\end{lstlisting}
\end{tcolorbox}%
We can now display the best model.%
\index{model}%
\par%
\begin{tcolorbox}[size=title,title=Code,breakable]%
\begin{lstlisting}[language=Python, upquote=true]
print(type(auto_reg))
model = auto_reg.export_model()
model.summary()\end{lstlisting}
\tcbsubtitle[before skip=\baselineskip]{Output}%
\begin{lstlisting}[upquote=true]
<class 'autokeras.tasks.image.ImageRegressor'>
Model: "model"
_________________________________________________________________
 Layer (type)                Output Shape              Param #
=================================================================
 input_1 (InputLayer)        [(None, 128, 128, 3)]     0
 cast_to_float32 (CastToFloa  (None, 128, 128, 3)      0
 t32)
 resnet50 (Functional)       (None, None, None, 2048)  23587712
 flatten (Flatten)           (None, 32768)             0
 regression_head_1 (Dense)   (None, 1)                 32769
=================================================================
Total params: 23,620,481
Trainable params: 32,769
Non-trainable params: 23,587,712
_________________________________________________________________
\end{lstlisting}
\end{tcolorbox}%
This top model can be saved and either utilized or trained further.%
\index{model}%
\par%
\begin{tcolorbox}[size=title,title=Code,breakable]%
\begin{lstlisting}[language=Python, upquote=true]
from keras.models import load_model

print(type(model))  

try:
    model.save("model_autokeras", save_format="tf")
except Exception:
    model.save("model_autokeras.h5")


loaded_model = load_model("model_autokeras",\
    custom_objects=ak.CUSTOM_OBJECTS)
print(loaded_model.evaluate(x,y))\end{lstlisting}
\tcbsubtitle[before skip=\baselineskip]{Output}%
\begin{lstlisting}[upquote=true]
<class 'keras.engine.functional.Functional'>
INFO:tensorflow:Assets written to: model_autokeras/assets
32/32 [==============================] - 2s 21ms/step - loss: 24.9077
- mean_squared_error: 24.9077
[24.90774917602539, 24.90774917602539]
\end{lstlisting}
\end{tcolorbox}

\section{Part 14.2: Using Denoising AutoEncoders in Keras}%
\label{sec:Part14.2UsingDenoisingAutoEncodersinKeras}%
Function approximation is perhaps the original task of machine learning. Long before computers and even the notion of machine learning, scientists came up with equations to fit their observations of nature. Scientists find equations to demonstrate correlations between observations. For example, various equations relate mass, acceleration, and force.%
\index{learning}%
\par%
Looking at complex data and deriving an equation does take some technical expertise. The goal of function approximation is to remove intuition from the process and instead depend on an algorithmic method to automatically generate an equation that describes data. A regression neural network performs this task.%
\index{algorithm}%
\index{neural network}%
\index{regression}%
\index{ROC}%
\index{ROC}%
\index{SOM}%
\par%
We begin by creating a function that we will use to chart a regression function.%
\index{regression}%
\par%
\begin{tcolorbox}[size=title,title=Code,breakable]%
\begin{lstlisting}[language=Python, upquote=true]
# Regression chart.
def chart_regression(pred, y, sort=True):
    t = pd.DataFrame({'pred': pred, 'y': y.flatten()})
    if sort:
        t.sort_values(by=['y'], inplace=True)
    plt.plot(t['y'].tolist(), label='expected')
    plt.plot(t['pred'].tolist(), label='prediction')
    plt.ylabel('output')
    plt.legend()
    plt.show()\end{lstlisting}
\end{tcolorbox}%
Next, we will attempt to approximate a slightly random variant of the trigonometric sine function.%
\index{random}%
\par%
\begin{tcolorbox}[size=title,title=Code,breakable]%
\begin{lstlisting}[language=Python, upquote=true]
import tensorflow as tf
import numpy as np
import pandas as pd
from tensorflow.keras.models import Sequential
from tensorflow.keras.layers import Dense, Activation
from tensorflow.keras.callbacks import EarlyStopping
import matplotlib.pyplot as plt

rng = np.random.RandomState(1)
x = np.sort((360 * rng.rand(100, 1)), axis=0)
y = np.array([np.sin(x*(np.pi/180.0)).ravel()]).T

model = Sequential()
model.add(Dense(100, input_dim=x.shape[1], activation='relu'))
model.add(Dense(50, activation='relu'))
model.add(Dense(25, activation='relu'))
model.add(Dense(1))
model.compile(loss='mean_squared_error', optimizer='adam')
model.fit(x,y,verbose=0,batch_size=len(x),epochs=25000)

pred = model.predict(x)

print("Actual")
print(y[0:5])

print("Pred")
print(pred[0:5])

chart_regression(pred.flatten(),y,sort=False)\end{lstlisting}
\tcbsubtitle[before skip=\baselineskip]{Output}%
\includegraphics[width=3in]{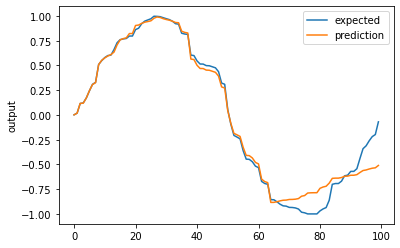}%
\begin{lstlisting}[upquote=true]
Actual
[[0.00071864]
 [0.01803382]
 [0.11465593]
 [0.1213861 ]
 [0.1712333 ]]
Pred
[[0.00078334]
 [0.0180243 ]
 [0.11705872]
 [0.11838552]
 [0.17200738]]
\end{lstlisting}
\end{tcolorbox}%
As you can see, the neural network creates a reasonably close approximation of the random sine function.%
\index{neural network}%
\index{random}%
\par%
\subsection{Multi{-}Output Regression}%
\label{subsec:Multi{-}OutputRegression}%
Unlike most models, neural networks can provide multiple regression outputs.  This feature allows a neural network to generate various outputs for the same input.  For example, you might train the MPG data set to predict MPG and horsepower.  One area in that multiple regression outputs can be helpful is autoencoders.  The following diagram shows a multi{-}regression neural network.  As you can see, there are multiple output neurons.  Usually, you will use multiple output neurons for classification.  Each output neuron will represent the probability of one of the classes.  However, in this case, it is a regression neural network.  Figure 13.MRG shows multi{-}output regression.%
\index{classification}%
\index{feature}%
\index{input}%
\index{model}%
\index{multiple output}%
\index{neural network}%
\index{neuron}%
\index{output}%
\index{output neuron}%
\index{predict}%
\index{probability}%
\index{regression}%
\par%

\begin{figure}[h]%
\centering%
\includegraphics[width=3in]{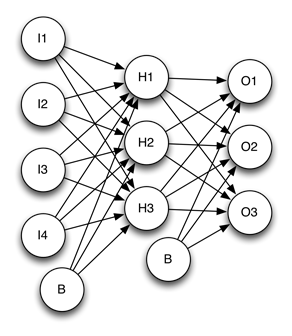}%
\caption{Multi{-}Output Regression}%
\label{14.MRG}%
\end{figure}

\par%
The following program uses a multi{-}output regression to predict both%
\index{output}%
\index{predict}%
\index{regression}%
\href{https://en.wikipedia.org/wiki/Trigonometric_functions#Sine.2C_cosine_and_tangent}{ sin }%
and%
\href{https://en.wikipedia.org/wiki/Trigonometric_functions#Sine.2C_cosine_and_tangent}{ cos }%
from the same input data.%
\index{input}%
\par%
\begin{tcolorbox}[size=title,title=Code,breakable]%
\begin{lstlisting}[language=Python, upquote=true]
from sklearn import metrics

rng = np.random.RandomState(1)
x = np.sort((360 * rng.rand(100, 1)), axis=0)
y = np.array([np.pi * np.sin(x*(np.pi/180.0)).ravel(), np.pi \
              * np.cos(x*(np.pi/180.0)).ravel()]).T

model = Sequential()
model.add(Dense(100, input_dim=x.shape[1], activation='relu'))
model.add(Dense(50, activation='relu'))
model.add(Dense(25, activation='relu'))
model.add(Dense(2)) # Two output neurons
model.compile(loss='mean_squared_error', optimizer='adam')
model.fit(x,y,verbose=0,batch_size=len(x),epochs=25000)


# Fit regression DNN model.
pred = model.predict(x)

score = np.sqrt(metrics.mean_squared_error(pred, y))
print("Score (RMSE): {}".format(score))

np.set_printoptions(suppress=True)

print("Predicted:")
print(np.array(pred[20:25]))

print("Expected:")
print(np.array(y[20:25]))\end{lstlisting}
\tcbsubtitle[before skip=\baselineskip]{Output}%
\begin{lstlisting}[upquote=true]
Score (RMSE): 0.06136952220466956
Predicted:
[[2.720404   1.590426  ]
 [2.7611256  1.5165515 ]
 [2.9106038  1.2454026 ]
 [3.005532   1.0359662 ]
 [3.0415256  0.90731066]]
Expected:
[[2.70765313 1.59317888]
 [2.75138445 1.51640628]
 [2.89299999 1.22480835]
 [2.97603942 1.00637655]
 [3.01381723 0.88685404]]
\end{lstlisting}
\end{tcolorbox}

\subsection{Simple Autoencoder}%
\begin{figure}[h]%
\centering%
\includegraphics[width=3in]{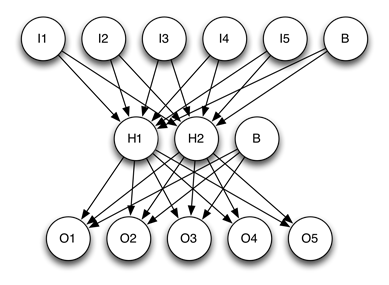}%
\caption{Simple Auto Encoder}%
\label{14.AUTO}%
\end{figure}

\label{subsec:SimpleAutoencoder}%
An autoencoder is a neural network with the same number of input neurons as it does outputs. The hidden layers of the neural network will have fewer neurons than the input/output neurons. Because there are fewer neurons, the auto{-}encoder must learn to encode the input to the fewer hidden neurons. The predictors (x) and output (y) are precisely the same in an autoencoder. Because of this, we consider autoencoders to be unsupervised. Figure \ref{14.AUTO} shows an autoencoder.%
\index{hidden layer}%
\index{hidden neuron}%
\index{input}%
\index{input neuron}%
\index{layer}%
\index{neural network}%
\index{neuron}%
\index{output}%
\index{output neuron}%
\index{predict}%
\par%

\par%
The following program demonstrates a very simple autoencoder that learns to encode a sequence of numbers. Fewer hidden neurons will make it more difficult for the autoencoder to understand.%
\index{hidden neuron}%
\index{neuron}%
\par%
\begin{tcolorbox}[size=title,title=Code,breakable]%
\begin{lstlisting}[language=Python, upquote=true]
from sklearn import metrics
import numpy as np
import pandas as pd
from IPython.display import display, HTML 
import tensorflow as tf

x = np.array([range(10)]).astype(np.float32)
print(x)

model = Sequential()
model.add(Dense(3, input_dim=x.shape[1], activation='relu'))
model.add(Dense(x.shape[1])) # Multiple output neurons
model.compile(loss='mean_squared_error', optimizer='adam')
model.fit(x,x,verbose=0,epochs=1000)

pred = model.predict(x)
score = np.sqrt(metrics.mean_squared_error(pred,x))
print("Score (RMSE): {}".format(score))
np.set_printoptions(suppress=True)
print(pred)\end{lstlisting}
\tcbsubtitle[before skip=\baselineskip]{Output}%
\begin{lstlisting}[upquote=true]
[[0. 1. 2. 3. 4. 5. 6. 7. 8. 9.]]
Score (RMSE): 0.024245187640190125
[[0.00000471 1.0009701  2.0032287  3.000911   4.0012217  5.0025473
  6.025212   6.9308095  8.014739   9.014762  ]]
\end{lstlisting}
\end{tcolorbox}

\subsection{Autoencode (single image)}%
\label{subsec:Autoencode(singleimage)}%
We are now ready to build a simple image autoencoder.  The program below learns a capable encoding for the image.  You can see the distortions that occur.%
\par%
\begin{tcolorbox}[size=title,title=Code,breakable]%
\begin{lstlisting}[language=Python, upquote=true]
%matplotlib inline
from PIL import Image, ImageFile
from matplotlib.pyplot import imshow
from tensorflow.keras.optimizers import SGD
import requests
from io import BytesIO

url = "https://data.heatonresearch.com/images/jupyter/brookings.jpeg"

response = requests.get(url,headers={'User-Agent': 'Mozilla/5.0'})
    
img = Image.open(BytesIO(response.content))
img.load()
img = img.resize((128,128), Image.ANTIALIAS)
img_array = np.asarray(img)
img_array = img_array.flatten()
img_array = np.array([ img_array ])
img_array = img_array.astype(np.float32)
print(img_array.shape[1])
print(img_array)

model = Sequential()
model.add(Dense(10, input_dim=img_array.shape[1], activation='relu'))
model.add(Dense(img_array.shape[1])) # Multiple output neurons
model.compile(loss='mean_squared_error', optimizer='adam')
model.fit(img_array,img_array,verbose=0,epochs=20)

print("Neural network output")
pred = model.predict(img_array)
print(pred)
print(img_array)
cols,rows = img.size
img_array2 = pred[0].reshape(rows,cols,3)
img_array2 = img_array2.astype(np.uint8)
img2 = Image.fromarray(img_array2, 'RGB')
img2\end{lstlisting}
\tcbsubtitle[before skip=\baselineskip]{Output}%
\includegraphics[width=1in]{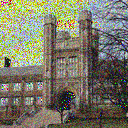}%
\begin{lstlisting}[upquote=true]
49152
[[203. 217. 240. ...  94.  92.  68.]]
Neural network output
[[238.31088 239.55913 194.47536 ...  67.12295  66.15083  74.94332]]
[[203. 217. 240. ...  94.  92.  68.]]
\end{lstlisting}
\end{tcolorbox}

\subsection{Standardize Images}%
\label{subsec:StandardizeImages}%
When processing several images together, it is sometimes essential to standardize them.  The following code reads a sequence of images and causes them to all be of the same size and perfectly square.  If the input images are not square, cropping will occur.%
\index{input}%
\index{ROC}%
\index{ROC}%
\index{SOM}%
\par%
\begin{tcolorbox}[size=title,title=Code,breakable]%
\begin{lstlisting}[language=Python, upquote=true]
%matplotlib inline
from PIL import Image, ImageFile
from matplotlib.pyplot import imshow
import requests
import numpy as np
from io import BytesIO
from IPython.display import display, HTML


images = [
    "https://data.heatonresearch.com/images/jupyter/Brown_Hall.jpeg",
    "https://data.heatonresearch.com/images/jupyter/brookings.jpeg",
    "https://data.heatonresearch.com/images/jupyter/WUSTLKnight.jpeg"
]


def make_square(img):
    cols,rows = img.size
    
    if rows>cols:
        pad = (rows-cols)/2
        img = img.crop((pad,0,cols,cols))
    else:
        pad = (cols-rows)/2
        img = img.crop((0,pad,rows,rows))
    
    return img
    
x = [] 
    
for url in images:
    ImageFile.LOAD_TRUNCATED_IMAGES = False
    response = requests.get(url,headers={'User-Agent': 'Mozilla/5.0'})
    img = Image.open(BytesIO(response.content))
    img.load()
    img = make_square(img)
    img = img.resize((128,128), Image.ANTIALIAS)
    print(url)
    display(img)
    img_array = np.asarray(img)
    img_array = img_array.flatten()
    img_array = img_array.astype(np.float32)
    img_array = (img_array-128)/128
    x.append(img_array)
    

x = np.array(x)

print(x.shape)\end{lstlisting}
\tcbsubtitle[before skip=\baselineskip]{Output}%
\includegraphics[width=1in]{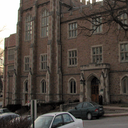}%
\begin{lstlisting}[upquote=true]
https://data.heatonresearch.com/images/jupyter/Brown_Hall.jpeg
\end{lstlisting}
\includegraphics[width=1in]{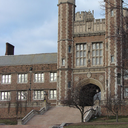}%
\begin{lstlisting}[upquote=true]
https://data.heatonresearch.com/images/jupyter/brookings.jpeg
\end{lstlisting}
\includegraphics[width=1in]{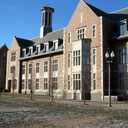}%
\begin{lstlisting}[upquote=true]
https://data.heatonresearch.com/images/jupyter/WUSTLKnight.jpeg
\end{lstlisting}
\begin{lstlisting}[upquote=true]
...
\end{lstlisting}
\end{tcolorbox}

\subsection{Image Autoencoder (multi{-}image)}%
\label{subsec:ImageAutoencoder(multi{-}image)}%
Autoencoders can learn the same encoding for multiple images.  The following code learns a single encoding for numerous images.%
\par%
\begin{tcolorbox}[size=title,title=Code,breakable]%
\begin{lstlisting}[language=Python, upquote=true]
%matplotlib inline
from PIL import Image, ImageFile
from matplotlib.pyplot import imshow
import requests
from io import BytesIO
from sklearn import metrics
import numpy as np
import pandas as pd
import tensorflow as tf
from IPython.display import display, HTML 

# Fit regression DNN model.
print("Creating/Training neural network")
model = Sequential()
model.add(Dense(50, input_dim=x.shape[1], activation='relu'))
model.add(Dense(x.shape[1])) # Multiple output neurons
model.compile(loss='mean_squared_error', optimizer='adam')
model.fit(x,x,verbose=0,epochs=1000)

print("Score neural network")
pred = model.predict(x)

cols,rows = img.size
for i in range(len(pred)):
    print(pred[i])
    img_array2 = pred[i].reshape(rows,cols,3)
    img_array2 = (img_array2*128)+128
    img_array2 = img_array2.astype(np.uint8)
    img2 = Image.fromarray(img_array2, 'RGB')
    display(img2)\end{lstlisting}
\tcbsubtitle[before skip=\baselineskip]{Output}%
\includegraphics[width=1in]{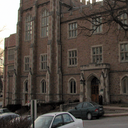}%
\begin{lstlisting}[upquote=true]
Creating/Training neural network
Score neural network
WARNING:tensorflow:5 out of the last 11 calls to <function
Model.make_predict_function.<locals>.predict_function at
0x7fe605654320> triggered tf.function retracing. Tracing is expensive
and the excessive number of tracings could be due to (1) creating
@tf.function repeatedly in a loop, (2) passing tensors with different
shapes, (3) passing Python objects instead of tensors. For (1), please
define your @tf.function outside of the loop. For (2), @tf.function
has experimental_relax_shapes=True option that relaxes argument shapes
that can avoid unnecessary retracing. For (3), please refer to
https://www.tensorflow.org/guide/function#controlling_retracing and
https://www.tensorflow.org/api_docs/python/tf/function for  more
details.
[ 0.98446846  0.9844943   0.98456836 ... -0.17971231 -0.20315537
 -0.20320868]
\end{lstlisting}
\includegraphics[width=1in]{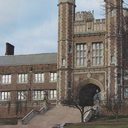}%
\begin{lstlisting}[upquote=true]
[ 0.5140943   0.59271055  0.6633089  ... -0.40498623 -0.40472946
 -0.54082954]
\end{lstlisting}
\includegraphics[width=1in]{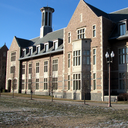}%
\begin{lstlisting}[upquote=true]
[-0.40605062  0.08633238  0.6571716  ... -0.12500083 -0.22656606
 -0.3437891 ]
\end{lstlisting}
\begin{lstlisting}[upquote=true]
...
\end{lstlisting}
\end{tcolorbox}

\subsection{Adding Noise to an Image}%
\label{subsec:AddingNoisetoanImage}%
Autoencoders can handle noise.  First, it is essential to see how to add noise to an image.  There are many ways to add such noise.  The following code adds random black squares to the image to produce noise.%
\index{random}%
\par%
\begin{tcolorbox}[size=title,title=Code,breakable]%
\begin{lstlisting}[language=Python, upquote=true]
from PIL import Image, ImageFile
from matplotlib.pyplot import imshow
import requests
from io import BytesIO

%matplotlib inline


def add_noise(a):
    a2 = a.copy()
    rows = a2.shape[0]
    cols = a2.shape[1]
    s = int(min(rows,cols)/20) # size of spot is 1/20 of smallest dimension
    
    for i in range(100):
        x = np.random.randint(cols-s)
        y = np.random.randint(rows-s)
        a2[y:(y+s),x:(x+s)] = 0
        
    return a2

url = "https://data.heatonresearch.com/images/jupyter/brookings.jpeg"
#url = "http://www.heatonresearch.com/images/about-jeff.jpg"

response = requests.get(url,headers={'User-Agent': 'Mozilla/5.0'})
img = Image.open(BytesIO(response.content))
img.load()

img_array = np.asarray(img)
rows = img_array.shape[0]
cols = img_array.shape[1]

print("Rows: {}, Cols: {}".format(rows,cols))

# Create new image
img2_array = img_array.astype(np.uint8)
print(img2_array.shape)
img2_array = add_noise(img2_array)
img2 = Image.fromarray(img2_array, 'RGB')
img2\end{lstlisting}
\tcbsubtitle[before skip=\baselineskip]{Output}%
\includegraphics[width=4in]{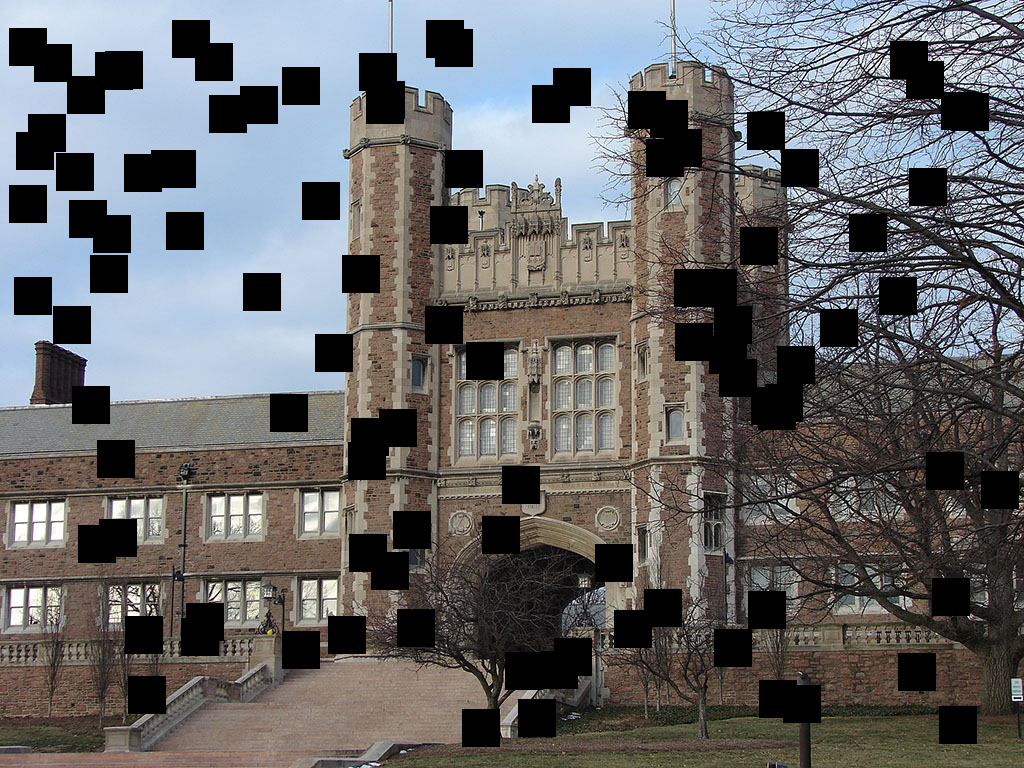}%
\begin{lstlisting}[upquote=true]
Rows: 768, Cols: 1024
(768, 1024, 3)
\end{lstlisting}
\end{tcolorbox}

\subsection{Denoising Autoencoder}%
\label{subsec:DenoisingAutoencoder}%
You design a denoising autoencoder to remove noise from input signals. You train the network to convert noisy data ($x$) to the original input ($y$). The $y$ becomes each image/signal (just like a normal autoencoder); however, the $x$ becomes a version of $y$ with noise added.  Noise is artificially added to the images to produce $x$.  The following code creates ten noisy versions of each of the images.%
\index{input}%
\par%
\begin{tcolorbox}[size=title,title=Code,breakable]%
\begin{lstlisting}[language=Python, upquote=true]
%matplotlib inline
from PIL import Image, ImageFile
from matplotlib.pyplot import imshow
import requests
import numpy as np
from io import BytesIO
from IPython.display import display, HTML

#url = "http://www.heatonresearch.com/images/about-jeff.jpg"

images = [
    "https://data.heatonresearch.com/images/jupyter/Brown_Hall.jpeg",
    "https://data.heatonresearch.com/images/jupyter/brookings.jpeg",
    "https://data.heatonresearch.com/images/jupyter/WUSTLKnight.jpeg"
]

def make_square(img):
    cols,rows = img.size
    
    if rows>cols:
        pad = (rows-cols)/2
        img = img.crop((pad,0,cols,cols))
    else:
        pad = (cols-rows)/2
        img = img.crop((0,pad,rows,rows))
    
    return img
    
x = []    
y = []
loaded_images = []
    
for url in images:
    ImageFile.LOAD_TRUNCATED_IMAGES = False
    response = requests.get(url,headers={'User-Agent': 'Mozilla/5.0'})
    img = Image.open(BytesIO(response.content))
    img.load()
    img = make_square(img)
    img = img.resize((128,128), Image.ANTIALIAS)
    
    loaded_images.append(img)
    print(url)
    display(img)
    for i in range(10):
        img_array = np.asarray(img)
        img_array_noise = add_noise(img_array)
        
        img_array = img_array.flatten()
        img_array = img_array.astype(np.float32)
        img_array = (img_array-128)/128
        
        img_array_noise = img_array_noise.flatten()
        img_array_noise = img_array_noise.astype(np.float32)
        img_array_noise = (img_array_noise-128)/128
        
        x.append(img_array_noise)
        y.append(img_array)
    
x = np.array(x)
y = np.array(y)

print(x.shape)
print(y.shape)\end{lstlisting}
\tcbsubtitle[before skip=\baselineskip]{Output}%
\includegraphics[width=1in]{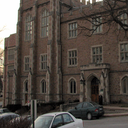}%
\begin{lstlisting}[upquote=true]
https://data.heatonresearch.com/images/jupyter/Brown_Hall.jpeg
\end{lstlisting}
\includegraphics[width=1in]{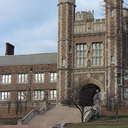}%
\begin{lstlisting}[upquote=true]
https://data.heatonresearch.com/images/jupyter/brookings.jpeg
\end{lstlisting}
\includegraphics[width=1in]{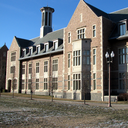}%
\begin{lstlisting}[upquote=true]
https://data.heatonresearch.com/images/jupyter/WUSTLKnight.jpeg
\end{lstlisting}
\begin{lstlisting}[upquote=true]
...
\end{lstlisting}
\end{tcolorbox}%
We now train the autoencoder neural network to transform the noisy images into clean images.%
\index{neural network}%
\par%
\begin{tcolorbox}[size=title,title=Code,breakable]%
\begin{lstlisting}[language=Python, upquote=true]
%matplotlib inline
from PIL import Image, ImageFile
from matplotlib.pyplot import imshow
import requests
from io import BytesIO
from sklearn import metrics
import numpy as np
import pandas as pd
import tensorflow as tf
from IPython.display import display, HTML 

# Fit regression DNN model.
print("Creating/Training neural network")
model = Sequential()
model.add(Dense(100, input_dim=x.shape[1], activation='relu'))
model.add(Dense(50, activation='relu'))
model.add(Dense(100, activation='relu'))
model.add(Dense(x.shape[1])) # Multiple output neurons
model.compile(loss='mean_squared_error', optimizer='adam')
model.fit(x,y,verbose=1,epochs=20)

print("Neural network trained")\end{lstlisting}
\tcbsubtitle[before skip=\baselineskip]{Output}%
\begin{lstlisting}[upquote=true]
Creating/Training neural network
...
1/1 [==============================] - 0s 105ms/step - loss: 0.0068
Epoch 20/20
1/1 [==============================] - 0s 110ms/step - loss: 0.0056
Neural network trained
\end{lstlisting}
\end{tcolorbox}%
We are now ready to evaluate the results.%
\par%
\begin{tcolorbox}[size=title,title=Code,breakable]%
\begin{lstlisting}[language=Python, upquote=true]
for z in range(3):
    print("*** Trial {}".format(z+1))
    
    # Choose random image
    i = np.random.randint(len(loaded_images))
    img = loaded_images[i]
    img_array = np.asarray(img)
    cols, rows = img.size

    # Add noise
    img_array_noise = add_noise(img_array)    

    #Display noisy image
    img2 = img_array_noise.astype(np.uint8)
    img2 = Image.fromarray(img2, 'RGB')
    print("With noise:")
    display(img2)

    # Present noisy image to auto encoder
    img_array_noise = img_array_noise.flatten()
    img_array_noise = img_array_noise.astype(np.float32)
    img_array_noise = (img_array_noise-128)/128
    img_array_noise = np.array([img_array_noise])
    pred = model.predict(img_array_noise)[0]

    # Display neural result
    img_array2 = pred.reshape(rows,cols,3)
    img_array2 = (img_array2*128)+128
    img_array2 = img_array2.astype(np.uint8)
    img2 = Image.fromarray(img_array2, 'RGB')
    print("After auto encode noise removal")
    display(img2)\end{lstlisting}
\tcbsubtitle[before skip=\baselineskip]{Output}%
\includegraphics[width=1in]{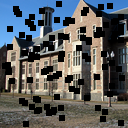}%
\begin{lstlisting}[upquote=true]
*** Trial 1
With noise:
\end{lstlisting}
\includegraphics[width=1in]{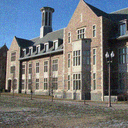}%
\begin{lstlisting}[upquote=true]
After auto encode noise removal
\end{lstlisting}
\includegraphics[width=1in]{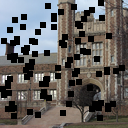}%
\begin{lstlisting}[upquote=true]
*** Trial 2
With noise:
\end{lstlisting}
\includegraphics[width=1in]{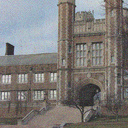}%
\begin{lstlisting}[upquote=true]
After auto encode noise removal
\end{lstlisting}
\includegraphics[width=1in]{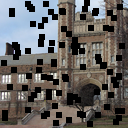}%
\begin{lstlisting}[upquote=true]
*** Trial 3
With noise:
\end{lstlisting}
\includegraphics[width=1in]{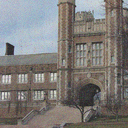}%
\begin{lstlisting}[upquote=true]
After auto encode noise removal
\end{lstlisting}
\end{tcolorbox}

\section{Part 14.3: Anomaly Detection in Keras}%
\label{sec:Part14.3AnomalyDetectioninKeras}%
Anomaly detection is an unsupervised training technique that analyzes the degree to which incoming data differs from the data you used to train the neural network. Traditionally, cybersecurity experts have used anomaly detection to ensure network security. However, you can use anomalies in data science to detect input for which you have not trained your neural network.%
\index{input}%
\index{neural network}%
\index{Supervised training}%
\index{training}%
\index{unsupervised training}%
\par%
There are several data sets that many commonly use to demonstrate anomaly detection. In this part, we will look at the KDD{-}99 dataset.%
\index{dataset}%
\par%
\begin{itemize}[noitemsep]%
\item%
\href{https://www.stratosphereips.org/category/dataset.html}{Stratosphere IPS Dataset}%
\item%
\href{https://www.unsw.adfa.edu.au/unsw-canberra-cyber/cybersecurity/ADFA-IDS-Datasets/}{The ADFA Intrusion Detection Datasets (2013) {-} for HIDS}%
\item%
\href{https://westpoint.edu/centers-and-research/cyber-research-center/data-sets}{ITOC CDX (2009)}%
\item%
\href{http://kdd.ics.uci.edu/databases/kddcup99/kddcup99.html}{KDD{-}99 Dataset}%
\end{itemize}%
\subsection{Read in KDD99 Data Set}%
\label{subsec:ReadinKDD99DataSet}%
Although the KDD99 dataset is over 20 years old, it is still widely used to demonstrate Intrusion Detection Systems (IDS) and Anomaly detection. KDD99 is the data set used for The Third International Knowledge Discovery and Data Mining Tools Competition, held in conjunction with KDD{-}99, The Fifth International Conference on Knowledge Discovery and Data Mining. The competition task was to build a network intrusion detector, a predictive model capable of distinguishing between "bad" connections, called intrusions or attacks, and "good" normal connections. This database contains a standard set of data to be audited, including various intrusions simulated in a military network environment.%
\index{connection}%
\index{dataset}%
\index{model}%
\index{predict}%
\par%
The following code reads the KDD99 CSV dataset into a Pandas data frame. The standard format of KDD99 does not include column names. Because of that, the program adds them.%
\index{CSV}%
\index{dataset}%
\par%
\begin{tcolorbox}[size=title,title=Code,breakable]%
\begin{lstlisting}[language=Python, upquote=true]
import pandas as pd
from tensorflow.keras.utils import get_file

pd.set_option('display.max_columns', 6)
pd.set_option('display.max_rows', 5)

try:
    path = get_file('kdd-with-columns.csv', origin=\
    'https://github.com/jeffheaton/jheaton-ds2/raw/main/'\
    'kdd-with-columns.csv',archive_format=None)
except:
    print('Error downloading')
    raise
    
print(path) 

# Origional file: http://kdd.ics.uci.edu/databases/kddcup99/kddcup99.html
df = pd.read_csv(path)

print("Read {} rows.".format(len(df)))
# df = df.sample(frac=0.1, replace=False) # Uncomment this line to 
# sample only 10% of the dataset
df.dropna(inplace=True,axis=1) 
# For now, just drop NA's (rows with missing values)


# display 5 rows
pd.set_option('display.max_columns', 5)
pd.set_option('display.max_rows', 5)
df\end{lstlisting}
\tcbsubtitle[before skip=\baselineskip]{Output}%
\begin{tabular}[hbt!]{l|l|l|l|l|l}%
\hline%
&duration&protocol\_type&...&dst\_host\_srv\_rerror\_rate&outcome\\%
\hline%
0&0&tcp&...&0.0&normal.\\%
1&0&tcp&...&0.0&normal.\\%
...&...&...&...&...&...\\%
494019&0&tcp&...&0.0&normal.\\%
494020&0&tcp&...&0.0&normal.\\%
\hline%
\end{tabular}%
\vspace{2mm}%
\begin{lstlisting}[upquote=true]
Downloading data from https://github.com/jeffheaton/jheaton-
ds2/raw/main/kdd-with-columns.csv
68132864/68132668 [==============================] - 1s 0us/step
68141056/68132668 [==============================] - 1s 0us/step
/root/.keras/datasets/kdd-with-columns.csv
Read 494021 rows.
\end{lstlisting}
\end{tcolorbox}%
The KDD99 dataset contains many columns that define the network state over time intervals during which a cyber attack might have taken place.  The " outcome " column specifies either "normal," indicating no attack, or the type of attack performed.  The following code displays the counts for each type of attack and "normal".%
\index{dataset}%
\par%
\begin{tcolorbox}[size=title,title=Code,breakable]%
\begin{lstlisting}[language=Python, upquote=true]
df.groupby('outcome')['outcome'].count()\end{lstlisting}
\tcbsubtitle[before skip=\baselineskip]{Output}%
\begin{lstlisting}[upquote=true]
outcome
back.               2203
buffer_overflow.      30
                    ...
warezclient.        1020
warezmaster.          20
Name: outcome, Length: 23, dtype: int64
\end{lstlisting}
\end{tcolorbox}

\subsection{Preprocessing}%
\label{subsec:Preprocessing}%
We must perform some preprocessing before we can feed the KDD99 data into the neural network. We provide the following two functions to assist with preprocessing. The first function converts numeric columns into Z{-}Scores. The second function replaces categorical values with dummy variables.%
\index{categorical}%
\index{neural network}%
\index{ROC}%
\index{ROC}%
\index{SOM}%
\index{Z{-}Score}%
\par%
\begin{tcolorbox}[size=title,title=Code,breakable]%
\begin{lstlisting}[language=Python, upquote=true]
# Encode a numeric column as zscores
def encode_numeric_zscore(df, name, mean=None, sd=None):
    if mean is None:
        mean = df[name].mean()

    if sd is None:
        sd = df[name].std()

    df[name] = (df[name] - mean) / sd
    
# Encode text values to dummy variables(i.e. [1,0,0],[0,1,0],[0,0,1] 
# for red,green,blue)
def encode_text_dummy(df, name):
    dummies = pd.get_dummies(df[name])
    for x in dummies.columns:
        dummy_name = f"{name}-{x}"
        df[dummy_name] = dummies[x]
    df.drop(name, axis=1, inplace=True)\end{lstlisting}
\end{tcolorbox}%
This code converts all numeric columns to Z{-}Scores and all textual columns to dummy variables. We now use these functions to preprocess each of the columns. Once the program preprocesses the data, we display the results.%
\index{ROC}%
\index{ROC}%
\index{Z{-}Score}%
\par%
\begin{tcolorbox}[size=title,title=Code,breakable]%
\begin{lstlisting}[language=Python, upquote=true]
# Now encode the feature vector

pd.set_option('display.max_columns', 6)
pd.set_option('display.max_rows', 5)

for name in df.columns:
  if name == 'outcome':
    pass
  elif name in ['protocol_type','service','flag','land','logged_in',
                'is_host_login','is_guest_login']:
    encode_text_dummy(df,name)
  else:
    encode_numeric_zscore(df,name)    

# display 5 rows

df.dropna(inplace=True,axis=1)
df[0:5]\end{lstlisting}
\tcbsubtitle[before skip=\baselineskip]{Output}%
\begin{tabular}[hbt!]{l|l|l|l|l|l|l|l}%
\hline%
&duration&src\_bytes&dst\_bytes&...&is\_host\_login{-}0&is\_guest\_login{-}0&is\_guest\_login{-}1\\%
\hline%
0&{-}0.067792&{-}0.002879&0.138664&...&1&1&0\\%
1&{-}0.067792&{-}0.002820&{-}0.011578&...&1&1&0\\%
2&{-}0.067792&{-}0.002824&0.014179&...&1&1&0\\%
3&{-}0.067792&{-}0.002840&0.014179&...&1&1&0\\%
4&{-}0.067792&{-}0.002842&0.035214&...&1&1&0\\%
\hline%
\end{tabular}%
\vspace{2mm}%
\end{tcolorbox}%
We divide the data into two groups, "normal" and the various attacks to perform anomaly detection. The following code divides the data into two data frames and displays each of these two groups' sizes.%
\par%
\begin{tcolorbox}[size=title,title=Code,breakable]%
\begin{lstlisting}[language=Python, upquote=true]
normal_mask = df['outcome']=='normal.'
attack_mask = df['outcome']!='normal.'

df.drop('outcome',axis=1,inplace=True)

df_normal = df[normal_mask]
df_attack = df[attack_mask]

print(f"Normal count: {len(df_normal)}")
print(f"Attack count: {len(df_attack)}")\end{lstlisting}
\tcbsubtitle[before skip=\baselineskip]{Output}%
\begin{lstlisting}[upquote=true]
Normal count: 97278
Attack count: 396743
\end{lstlisting}
\end{tcolorbox}%
Next, we convert these two data frames into Numpy arrays. Keras requires this format for data.%
\index{Keras}%
\index{NumPy}%
\par%
\begin{tcolorbox}[size=title,title=Code,breakable]%
\begin{lstlisting}[language=Python, upquote=true]
# This is the numeric feature vector, as it goes to the neural net
x_normal = df_normal.values
x_attack = df_attack.values\end{lstlisting}
\end{tcolorbox}

\subsection{Training the Autoencoder}%
\label{subsec:TrainingtheAutoencoder}%
It is important to note that we are not using the outcome column as a label to predict. We will train an autoencoder on the normal data and see how well it can detect that the data not flagged as "normal" represents an anomaly. This anomaly detection is unsupervised; there is no target (y) value to predict.%
\index{predict}%
\par%
Next, we split the normal data into a 25\% test set and a 75\% train set. The program will use the test data to facilitate early stopping.%
\index{early stopping}%
\par%
\begin{tcolorbox}[size=title,title=Code,breakable]%
\begin{lstlisting}[language=Python, upquote=true]
from sklearn.model_selection import train_test_split

x_normal_train, x_normal_test = train_test_split(
    x_normal, test_size=0.25, random_state=42)\end{lstlisting}
\end{tcolorbox}%
We display the size of the train and test sets.%
\par%
\begin{tcolorbox}[size=title,title=Code,breakable]%
\begin{lstlisting}[language=Python, upquote=true]
print(f"Normal train count: {len(x_normal_train)}")
print(f"Normal test count: {len(x_normal_test)}")\end{lstlisting}
\tcbsubtitle[before skip=\baselineskip]{Output}%
\begin{lstlisting}[upquote=true]
Normal train count: 72958
Normal test count: 24320
\end{lstlisting}
\end{tcolorbox}%
We are now ready to train the autoencoder on the normal data. The autoencoder will learn to compress the data to a vector of just three numbers. The autoencoder should be able to also decompress with reasonable accuracy. As is typical for autoencoders, we are merely training the neural network to produce the same output values as were fed to the input layer.%
\index{input}%
\index{input layer}%
\index{layer}%
\index{neural network}%
\index{output}%
\index{training}%
\index{vector}%
\par%
\begin{tcolorbox}[size=title,title=Code,breakable]%
\begin{lstlisting}[language=Python, upquote=true]
from sklearn import metrics
import numpy as np
import pandas as pd
from IPython.display import display, HTML 
import tensorflow as tf
from tensorflow.keras.models import Sequential
from tensorflow.keras.layers import Dense, Activation

model = Sequential()
model.add(Dense(25, input_dim=x_normal.shape[1], activation='relu'))
model.add(Dense(3, activation='relu')) # size to compress to
model.add(Dense(25, activation='relu'))
model.add(Dense(x_normal.shape[1])) # Multiple output neurons
model.compile(loss='mean_squared_error', optimizer='adam')
model.fit(x_normal_train,x_normal_train,verbose=1,epochs=100)\end{lstlisting}
\tcbsubtitle[before skip=\baselineskip]{Output}%
\begin{lstlisting}[upquote=true]
...
2280/2280 [==============================] - 6s 3ms/step - loss:
0.0512
Epoch 100/100
2280/2280 [==============================] - 5s 2ms/step - loss:
0.0562
\end{lstlisting}
\end{tcolorbox}

\subsection{Detecting an Anomaly}%
\label{subsec:DetectinganAnomaly}%
We are now ready to see if the abnormal data is an anomaly. The first two scores show the in{-}sample and out of sample RMSE errors. These two scores are relatively low at around 0.33 because they resulted from normal data. The much higher 0.76 error occurred from the abnormal data. The autoencoder is not as capable of encoding data that represents an attack. This higher error indicates an anomaly.%
\index{error}%
\index{MSE}%
\index{RMSE}%
\index{RMSE}%
\par%
\begin{tcolorbox}[size=title,title=Code,breakable]%
\begin{lstlisting}[language=Python, upquote=true]
pred = model.predict(x_normal_test)
score1 = np.sqrt(metrics.mean_squared_error(pred,x_normal_test))
pred = model.predict(x_normal)
score2 = np.sqrt(metrics.mean_squared_error(pred,x_normal))
pred = model.predict(x_attack)
score3 = np.sqrt(metrics.mean_squared_error(pred,x_attack))
print(f"Out of Sample Normal Score (RMSE): {score1}")
print(f"Insample Normal Score (RMSE): {score2}")
print(f"Attack Underway Score (RMSE): {score3}")\end{lstlisting}
\tcbsubtitle[before skip=\baselineskip]{Output}%
\begin{lstlisting}[upquote=true]
Out of Sample Normal Score (RMSE): 0.27485267641044275
Insample Normal Score (RMSE): 0.24613762509093587
Attack Underway Score (RMSE): 0.6398492471974858
\end{lstlisting}
\end{tcolorbox}

\section{Part 14.4: Training an Intrusion Detection System with KDD99}%
\label{sec:Part14.4TraininganIntrusionDetectionSystemwithKDD99}%
The%
\href{http://kdd.ics.uci.edu/databases/kddcup99/kddcup99.html}{ KDD{-}99 dataset }%
is very famous in the security field and almost a "hello world" of Intrusion Detection Systems (IDS) in machine learning. An intrusion detection system (IDS) is a program that monitors computers and network systems for malicious activity or policy violations. Any intrusion activity or violation is typically reported to an administrator or collected centrally. IDS types range in scope from single computers to large networks. Although the KDD99 dataset is over 20 years old, it is still widely used to demonstrate Intrusion Detection Systems (IDS). KDD99 is the data set used for The Third International Knowledge Discovery and Data Mining Tools Competition, which was held in conjunction with KDD{-}99, The Fifth International Conference on Knowledge Discovery and Data Mining. The competition task was to build a network intrusion detector, a predictive model capable of distinguishing between "bad" connections, called intrusions or attacks, and "good" normal connections. This database contains a standard set of data to be audited, including various intrusions simulated in a military network environment.%
\index{connection}%
\index{dataset}%
\index{learning}%
\index{model}%
\index{predict}%
\par%
\subsection{Read in Raw KDD{-}99 Dataset}%
\label{subsec:ReadinRawKDD{-}99Dataset}%
The following code reads the KDD99 CSV dataset into a Pandas data frame. The standard format of KDD99 does not include column names. Because of that, the program adds them.%
\index{CSV}%
\index{dataset}%
\par%
\begin{tcolorbox}[size=title,title=Code,breakable]%
\begin{lstlisting}[language=Python, upquote=true]
import pandas as pd
from tensorflow.keras.utils import get_file

pd.set_option('display.max_columns', 6)
pd.set_option('display.max_rows', 5)

try:
    path = get_file('kdd-with-columns.csv', origin=\
    'https://github.com/jeffheaton/jheaton-ds2/raw/main/'\
    'kdd-with-columns.csv',archive_format=None)
except:
    print('Error downloading')
    raise
    
print(path) 

# Origional file: http://kdd.ics.uci.edu/databases/kddcup99/kddcup99.html
df = pd.read_csv(path)

print("Read {} rows.".format(len(df)))
# df = df.sample(frac=0.1, replace=False) # Uncomment this line to 
# sample only 10% of the dataset
df.dropna(inplace=True,axis=1) 
# For now, just drop NA's (rows with missing values)


# display 5 rows
pd.set_option('display.max_columns', 5)
pd.set_option('display.max_rows', 5)
df\end{lstlisting}
\tcbsubtitle[before skip=\baselineskip]{Output}%
\begin{tabular}[hbt!]{l|l|l|l|l|l}%
\hline%
&duration&protocol\_type&...&dst\_host\_srv\_rerror\_rate&outcome\\%
\hline%
0&0&tcp&...&0.0&normal.\\%
1&0&tcp&...&0.0&normal.\\%
...&...&...&...&...&...\\%
494019&0&tcp&...&0.0&normal.\\%
494020&0&tcp&...&0.0&normal.\\%
\hline%
\end{tabular}%
\vspace{2mm}%
\begin{lstlisting}[upquote=true]
Downloading data from https://github.com/jeffheaton/jheaton-
ds2/raw/main/kdd-with-columns.csv
68132864/68132668 [==============================] - 1s 0us/step
68141056/68132668 [==============================] - 1s 0us/step
/root/.keras/datasets/kdd-with-columns.csv
Read 494021 rows.
\end{lstlisting}
\end{tcolorbox}

\subsection{Analyzing a Dataset}%
\label{subsec:AnalyzingaDataset}%
Before we preprocess the KDD99 dataset, let's look at the individual columns and distributions.  You can use the following script to give a high{-}level overview of how a dataset appears.%
\index{dataset}%
\index{ROC}%
\index{ROC}%
\par%
\begin{tcolorbox}[size=title,title=Code,breakable]%
\begin{lstlisting}[language=Python, upquote=true]
import pandas as pd
import os
import numpy as np
from sklearn import metrics
from scipy.stats import zscore

def expand_categories(values):
    result = []
    s = values.value_counts()
    t = float(len(values))
    for v in s.index:
        result.append("{}:{}%".format(v,round(100*(s[v]/t),2)))
    return "[{}]".format(",".join(result))
        
def analyze(df):
    print()
    cols = df.columns.values
    total = float(len(df))

    print("{} rows".format(int(total)))
    for col in cols:
        uniques = df[col].unique()
        unique_count = len(uniques)
        if unique_count>100:
            print("** {}:{} ({}%)".format(col,unique_count,\
                int(((unique_count)/total)*100)))
        else:
            print("** {}:{}".format(col,expand_categories(df[col])))
            expand_categories(df[col])\end{lstlisting}
\end{tcolorbox}%
The analysis looks at how many unique values are present.  For example, duration, a numeric value, has 2495 unique values, and there is a 0\% overlap.  A text/categorical value such as protocol\_type only has a few unique values, and the program shows the percentages of each.  Columns with many unique values do not have their item counts shown to save display space.%
\index{categorical}%
\par%
\begin{tcolorbox}[size=title,title=Code,breakable]%
\begin{lstlisting}[language=Python, upquote=true]
# Analyze KDD-99
analyze(df)\end{lstlisting}
\tcbsubtitle[before skip=\baselineskip]{Output}%
\begin{lstlisting}[upquote=true]
494021 rows
** duration:2495 (0%)
** protocol_type:[icmp:57.41%,tcp:38.47%,udp:4.12%]
** service:[ecr_i:56.96%,private:22.45%,http:13.01%,smtp:1.97%,other:1
.46%,domain_u:1.19%,ftp_data:0.96%,eco_i:0.33%,ftp:0.16%,finger:0.14%,
urp_i:0.11%,telnet:0.1%,ntp_u:0.08%,auth:0.07%,pop_3:0.04%,time:0.03%,
csnet_ns:0.03%,remote_job:0.02%,gopher:0.02%,imap4:0.02%,discard:0.02%
,domain:0.02%,iso_tsap:0.02%,systat:0.02%,shell:0.02%,echo:0.02%,rje:0
.02%,whois:0.02%,sql_net:0.02%,printer:0.02%,nntp:0.02%,courier:0.02%,
sunrpc:0.02%,netbios_ssn:0.02%,mtp:0.02%,vmnet:0.02%,uucp_path:0.02%,u
ucp:0.02%,klogin:0.02%,bgp:0.02%,ssh:0.02%,supdup:0.02%,nnsp:0.02%,log
in:0.02%,hostnames:0.02%,efs:0.02%,daytime:0.02%,link:0.02%,netbios_ns
:0.02%,pop_2:0.02%,ldap:0.02%,netbios_dgm:0.02%,exec:0.02%,http_443:0.
02%,kshell:0.02%,name:0.02%,ctf:0.02%,netstat:0.02%,Z39_50:0.02%,IRC:0
.01%,urh_i:0.0%,X11:0.0%,tim_i:0.0%,pm_dump:0.0%,tftp_u:0.0%,red_i:0.0

...

** outcome:[smurf.:56.84%,neptune.:21.7%,normal.:19.69%,back.:0.45%,sa
tan.:0.32%,ipsweep.:0.25%,portsweep.:0.21%,warezclient.:0.21%,teardrop
.:0.2%,pod.:0.05%,nmap.:0.05%,guess_passwd.:0.01%,buffer_overflow.:0.0
1%,land.:0.0%,warezmaster.:0.0%,imap.:0.0%,rootkit.:0.0%,loadmodule.:0
.0%,ftp_write.:0.0%,multihop.:0.0%,phf.:0.0%,perl.:0.0%,spy.:0.0%]
\end{lstlisting}
\end{tcolorbox}

\subsection{Encode the feature vector}%
\label{subsec:Encodethefeaturevector}%
We use the same two functions provided earlier to preprocess the data. The first encodes Z{-}Scores, and the second creates dummy variables from categorical columns.%
\index{categorical}%
\index{ROC}%
\index{ROC}%
\index{Z{-}Score}%
\par%
\begin{tcolorbox}[size=title,title=Code,breakable]%
\begin{lstlisting}[language=Python, upquote=true]
# Encode a numeric column as zscores
def encode_numeric_zscore(df, name, mean=None, sd=None):
    if mean is None:
        mean = df[name].mean()

    if sd is None:
        sd = df[name].std()

    df[name] = (df[name] - mean) / sd
    
# Encode text values to dummy variables(i.e. [1,0,0],
# [0,1,0],[0,0,1] for red,green,blue)
def encode_text_dummy(df, name):
    dummies = pd.get_dummies(df[name])
    for x in dummies.columns:
        dummy_name = f"{name}-{x}"
        df[dummy_name] = dummies[x]
    df.drop(name, axis=1, inplace=True)\end{lstlisting}
\end{tcolorbox}%
Again, just as we did for anomaly detection, we preprocess the data set.  We convert all numeric values to Z{-}Score and translate all categorical to dummy variables.%
\index{categorical}%
\index{ROC}%
\index{ROC}%
\index{Z{-}Score}%
\par%
\begin{tcolorbox}[size=title,title=Code,breakable]%
\begin{lstlisting}[language=Python, upquote=true]
# Now encode the feature vector

pd.set_option('display.max_columns', 6)
pd.set_option('display.max_rows', 5)

for name in df.columns:
  if name == 'outcome':
    pass
  elif name in ['protocol_type','service','flag','land','logged_in',
                'is_host_login','is_guest_login']:
    encode_text_dummy(df,name)
  else:
    encode_numeric_zscore(df,name)    

# display 5 rows

df.dropna(inplace=True,axis=1)
df[0:5]


# Convert to numpy - Classification
x_columns = df.columns.drop('outcome')
x = df[x_columns].values
dummies = pd.get_dummies(df['outcome']) # Classification
outcomes = dummies.columns
num_classes = len(outcomes)
y = dummies.values\end{lstlisting}
\end{tcolorbox}%
We will attempt to predict what type of attack is underway.  The outcome column specifies the attack type.  A value of normal indicates that there is no attack underway.  We display the outcomes; some attack types are much rarer than others.%
\index{predict}%
\index{SOM}%
\par%
\begin{tcolorbox}[size=title,title=Code,breakable]%
\begin{lstlisting}[language=Python, upquote=true]
df.groupby('outcome')['outcome'].count()\end{lstlisting}
\tcbsubtitle[before skip=\baselineskip]{Output}%
\begin{lstlisting}[upquote=true]
outcome
back.               2203
buffer_overflow.      30
                    ...
warezclient.        1020
warezmaster.          20
Name: outcome, Length: 23, dtype: int64
\end{lstlisting}
\end{tcolorbox}

\subsection{Train the Neural Network}%
\label{subsec:TraintheNeuralNetwork}%
We now train the neural network to classify the different KDD99 outcomes.  The code provided here implements a relatively simple neural with two hidden layers.  We train it with the provided KDD99 data.%
\index{hidden layer}%
\index{layer}%
\index{neural network}%
\par%
\begin{tcolorbox}[size=title,title=Code,breakable]%
\begin{lstlisting}[language=Python, upquote=true]
import pandas as pd
import io
import requests
import numpy as np
import os
from sklearn.model_selection import train_test_split
from sklearn import metrics
from tensorflow.keras.models import Sequential
from tensorflow.keras.layers import Dense, Activation
from tensorflow.keras.callbacks import EarlyStopping

# Create a test/train split.  25% test
# Split into train/test
x_train, x_test, y_train, y_test = train_test_split(
    x, y, test_size=0.25, random_state=42)

# Create neural net
model = Sequential()
model.add(Dense(10, input_dim=x.shape[1], activation='relu'))
model.add(Dense(50, input_dim=x.shape[1], activation='relu'))
model.add(Dense(10, input_dim=x.shape[1], activation='relu'))
model.add(Dense(1, kernel_initializer='normal'))
model.add(Dense(y.shape[1],activation='softmax'))
model.compile(loss='categorical_crossentropy', optimizer='adam')
monitor = EarlyStopping(monitor='val_loss', min_delta=1e-3, 
                        patience=5, verbose=1, mode='auto',
                           restore_best_weights=True)
model.fit(x_train,y_train,validation_data=(x_test,y_test),
          callbacks=[monitor],verbose=2,epochs=1000)\end{lstlisting}
\tcbsubtitle[before skip=\baselineskip]{Output}%
\begin{lstlisting}[upquote=true]
...
11579/11579 - 22s - loss: 0.0139 - val_loss: 0.0153 - 22s/epoch -
2ms/step
Epoch 19/1000
Restoring model weights from the end of the best epoch: 14.
11579/11579 - 23s - loss: 0.0141 - val_loss: 0.0152 - 23s/epoch -
2ms/step
Epoch 19: early stopping
\end{lstlisting}
\end{tcolorbox}%
We can now evaluate the neural network.  As you can see, the neural network achieves a 99\% accuracy rate.%
\index{neural network}%
\par%
\begin{tcolorbox}[size=title,title=Code,breakable]%
\begin{lstlisting}[language=Python, upquote=true]
# Measure accuracy
pred = model.predict(x_test)
pred = np.argmax(pred,axis=1)
y_eval = np.argmax(y_test,axis=1)
score = metrics.accuracy_score(y_eval, pred)
print("Validation score: {}".format(score))\end{lstlisting}
\tcbsubtitle[before skip=\baselineskip]{Output}%
\begin{lstlisting}[upquote=true]
Validation score: 0.9977005165740935
\end{lstlisting}
\end{tcolorbox}

\section{Part 14.5: New Technologies}%
\label{sec:Part14.5NewTechnologies}%
This course changes often to keep up with the rapidly evolving deep learning landscape. If you would like to continue to monitor this class, I suggest following me on the following:%
\index{learning}%
\par%
\begin{itemize}[noitemsep]%
\item%
\href{https://github.com/jeffheaton}{GitHub }%
{-} I post all changes to GitHub.%
\index{GitHub}%
\item%
\href{https://www.youtube.com/user/HeatonResearch}{Jeff Heaton's YouTube Channel }%
{-} I add new videos for this class on my channel.%
\index{video}%
\end{itemize}%
\subsection{New Technology Radar}%
\label{subsec:NewTechnologyRadar}%
Currently, these new technologies are on my radar for possible future inclusion in this course:%
\par%
\begin{itemize}[noitemsep]%
\item%
More advanced uses of transformers%
\index{transformer}%
\item%
More Advanced Transfer Learning%
\index{learning}%
\index{transfer learning}%
\item%
Augmentation%
\item%
Reinforcement Learning beyond TF{-}Agents%
\index{learning}%
\index{reinforcement learning}%
\end{itemize}%
This section seeks only to provide a high{-}level overview of these emerging technologies. I provide links to supplemental material and code in each subsection. I describe these technologies in the following sections.%
\index{link}%
\par%
Transformers are a relatively new technology that I will soon add to this course. They have resulted in many NLP applications. Projects such as the Bidirectional Encoder Representations from Transformers (BERT) and Generative Pre{-}trained Transformer (GPT{-}1,2,3) received much attention from practitioners. Transformers allow the sequence to sequence machine learning, allowing the model to utilize variable length, potentially textual, input. The output from the transformer is also a variable{-}length sequence. This feature enables the transformer to learn to perform such tasks as translation between human languages or even complicated NLP{-}based classification. Considerable compute power is needed to take advantage of transformers; thus, you should be taking advantage of transfer learning to train and fine{-}tune your transformers.%
\index{classification}%
\index{feature}%
\index{input}%
\index{learning}%
\index{model}%
\index{output}%
\index{transfer learning}%
\index{transformer}%
\par%
Complex models can require considerable training time. It is not unusual to see GPU clusters trained for days to achieve state{-}of{-}the{-}art results. This complexity requires a substantial monetary cost to train a state{-}of{-}the{-}art model. Because of this cost, you must consider transfer learning. Services, such as Hugging Face and NVIDIA GPU Cloud (NGC), contain many advanced pretrained neural networks for you to implement.%
\index{GPU}%
\index{GPU}%
\index{hugging face}%
\index{learning}%
\index{model}%
\index{neural network}%
\index{NVIDIA}%
\index{training}%
\index{transfer learning}%
\par%
Augmentation is a technique where algorithms generate additional training data augmenting the training data with new items that are modified versions of the original training data. This technique has seen many applications in computer vision. In this most basic example, the algorithm can flip images vertically and horizontally to quadruple the training set's size. Projects such as NVIDIA StyleGAN3 ADA have implemented augmentation to substantially decrease the amount of training data that the algorithm needs.%
\index{algorithm}%
\index{computer vision}%
\index{GAN}%
\index{NVIDIA}%
\index{StyleGAN}%
\index{training}%
\par%
Currently, this course makes use of TF{-}Agents to implement reinforcement learning. TF{-}Agents is convenient because it is based on TensorFlow. However, TF{-}Agents has been slow to update compared to other frameworks. Additionally, when TF{-}Agents is updated, internal errors are often introduced that can take months for the TF{-}Agents team to fix. When I compare simple "Hello World" type examples for Atari games on platforms like Stable Baselines to their TF{-}Agents equivalents, I am left wanting more from TF{-}Agents.%
\index{error}%
\index{learning}%
\index{reinforcement learning}%
\index{TensorFlow}%
\par

\subsection{Programming Language Radar}%
\label{subsec:ProgrammingLanguageRadar}%
Python has an absolute lock on the industry as a machine learning programming language. Python is not going anywhere any time soon. My main issue with Python is end{-}to{-}end deployment. Python will be your go{-}to language unless you are dealing with Jupyter notebooks or training/pipeline scripts. However, you will certainly need to utilize other languages to create edge applications, such as web pages and mobile apps. I do not suggest replacing Python with any of the following languages; however, these are some alternative languages and domains that you might choose to use them.%
\index{learning}%
\index{Python}%
\index{SOM}%
\index{training}%
\par%
\begin{itemize}[noitemsep]%
\item%
\textbf{IOS Application Development }%
{-} Swift%
\item%
\textbf{Android Development }%
{-} Kotlin and Java%
\index{Java}%
\item%
\textbf{Web Development }%
{-} NodeJS and JavaScript%
\index{Java}%
\index{JavaScript}%
\item%
\textbf{Mac Application Development }%
{-} Swift or JavaScript with Electron or React Native%
\index{Java}%
\index{JavaScript}%
\item%
\textbf{Windows Application Development }%
{-} C\# or JavaScript with Electron or React Native%
\index{Java}%
\index{JavaScript}%
\item%
\textbf{Linux Application Development }%
{-} C/C++ w with Tcl/Tk or JavaScript with Electron or React Native%
\index{Java}%
\index{JavaScript}%
\end{itemize}

\subsection{What About PyTorch?}%
\label{subsec:WhatAboutPyTorch?}%
Technical folks love debates that can reach levels of fervor generally reserved for religion or politics. Python and TensorFlow are approaching this level of spirited competition. There is no clear winner, at least at this point. Why did I base this class on Keras/TensorFlow, as opposed to PyTorch? There are two primary reasons. The first reason is a fact; the second is my opinion.%
\index{Keras}%
\index{Python}%
\index{PyTorch}%
\index{TensorFlow}%
\par%
PyTorch was not available in early 2016 when I introduced/developed this course.\newline%
PyTorch exposes lower{-}level details that would be distracting for applications of deep learning course.\newline%
I recommend being familiar with core deep learning techniques and being adaptable to switch between these two frameworks.%
\index{learning}%
\index{PyTorch}%
\par

\subsection{Where to From Here?}%
\label{subsec:WheretoFromHere?}%
So what's next? Here are some ideas.%
\index{SOM}%
\par%
\begin{itemize}[noitemsep]%
\item%
\href{https://colab.research.google.com/signup}{Google CoLab Pro }%
{-} If you need more GPU power; but are not yet ready to buy a GPU of your own.%
\index{GPU}%
\index{GPU}%
\item%
\href{https://www.tensorflow.org/certificate}{TensorFlow Certification}%
\item%
\href{https://www.coursera.org/}{Coursera}%
\end{itemize}%
I hope that you have enjoyed this course. If you have any suggestions for improvement or technology suggestions, please get in touch with me. This course is always evolving, and I invite you to subscribe to my%
\href{https://www.youtube.com/user/HeatonResearch}{ YouTube channel }%
for my latest updates. I also frequently post videos beyond the scope of this course, so the channel itself is a good next step. Thank you very much for your interest and focus on this course. Other social media links for me include:%
\index{link}%
\index{video}%
\par%
\begin{itemize}[noitemsep]%
\item%
\href{https://github.com/jeffheaton}{Jeff Heaton GitHub}%
\item%
\href{https://twitter.com/jeffheaton}{Jeff Heaton Twitter}%
\item%
\href{https://medium.com/@heatonresearch}{Jeff Heaton Medium}%
\end{itemize}

\backmatter%
\bibliographystyle{acm}%
\bibliography{citations}{}%
\printindex%
\end{document}